\documentclass[oneside]{book}

\usepackage{MyBookStyle}

\title{Trustworthy Machine Learning}

\author[1,*]{Bálint Mucsányi}
\author[1]{Michael Kirchhof}
\author[1,2]{Elisa Nguyen}
\author[1,2]{\\Alexander Rubinstein}
\author[1,2]{Seong Joon Oh}
\affil[1]{University of Tübingen}
\affil[2]{Tübingen AI Center}
\affil[*]{\texttt{balint.mucsanyi@student.uni-tuebingen.de}}

\date{}

\hypersetup{
    pdftitle={Trustworthy Machine Learning},
    pdfauthor={Bálint Mucsányi, Michael Kirchhof, Elisa Nguyen, Alexander Rubinstein, Seong Joon Oh},
    pdfkeywords={Machine Learning, Scalability, Trustworthiness, OOD Generalization, Explainability, Uncertainty, Evaluation},
}

\begin{document}
\pagestyle{empty}
\pagenumbering{alph}
\begin{titlepage}
\newgeometry{margin=60pt,top=45pt}
\begingroup
\singlespacing
\sffamily
\thispagecolor{TPBackColor}
\color{TPColor}
\begin{raggedright}
  \setmyfontsize{12}
    Bálint Mucsányi, Michael Kirchhof, Elisa Nguyen,\newline
    Alexander Rubinstein, Seong Joon Oh
\end{raggedright}

\vfill
\noindent{\color{TPColor!20!TPBackColor}\rule{\textwidth}{1pt}}
\vfill

\begin{center}
\singlespacing
\begin{minipage}[t]{0.7\linewidth}
  \setmyfontsize{43}\bfseries
  Trustworthy\newline 
Machine\newline
Learning
\end{minipage}\hfill%
\setbox\tempbox\hbox{\setmyfontsize{18}\bfseries Applications}%
\begin{minipage}[t]{\wd\tempbox}
  \setmyfontsize{18}\vskip-\dimexpr48pt-18pt\relax
Theory\newline
Applications\newline
Intuitions
\end{minipage}

\vfill

\includegraphics[width=\textwidth]{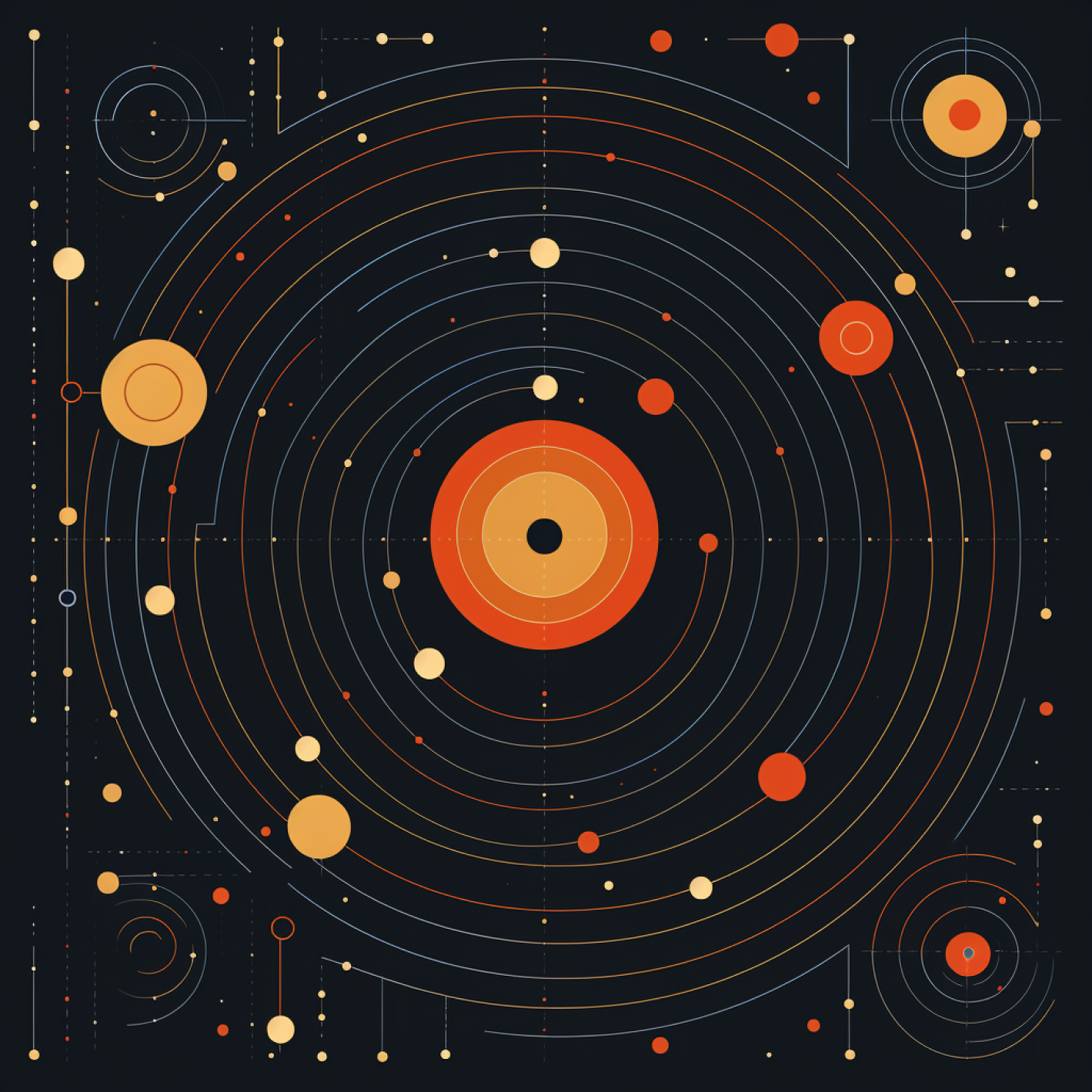}
\end{center}
\endgroup
\end{titlepage}\restoregeometry
\cleardoublepage%
\setcounter{page}{0}%
\pagenumbering{arabic}%
\pagestyle{MyBookPS}
\maketitle
\cleardoublepage


\tableofcontents

\newpage
\sloppy
\chapter*{Preface}
As machine learning technology gets applied to actual products and solutions, new challenges have emerged. Models unexpectedly fail to generalize to small changes in the distribution; some models are found to utilize sensitive features that could treat certain demographic user groups unfairly; models tend to be confident on novel data they have never seen, or models cannot communicate the rationale behind their decisions effectively with the end users like medical staff to maximize the human-machine synergies. Collectively, we face a trustworthiness issue with the current machine learning technology. A large fraction of machine learning research nowadays is dedicated to expanding the frontier of Trustworthy Machine Learning (TML). TML has been an explicit topic in the \href{https://icml.cc/Conferences/2023/CallForPapers}{call for papers of the ICML conference} since 2020, and other relatively young conferences dealing with TML topics emerged like \href{https://facctconference.org/}{FAccT}, or \href{https://www.aies-conference.com/2023/}{AIES}.

This textbook on TML is an end product of the homonymous course at the University of Tübingen, first offered in the Winter Semester of 2022/23. The book covers a theoretical and technical background for key topics in TML as well as underlying intuitions. We conduct a critical review of important classical and contemporary research papers on related topics. The book is meant to be a stand-alone product accompanied by code snippets and various pointers to further sources on topics of TML.

The goal of this book is to prepare readers to critically read, assess, and discuss research work in TML. Through the provided code snippets, readers will gain the technical background to implement basic TML techniques and, eventually,  conduct their own research in TML.

The book has the following prerequisites:
\begin{itemize}
    \item Familiarity with Python and PyTorch coding.
    \item Basic knowledge of machine learning concepts and deep learning.
    \item Basic maths: multivariate calculus, linear algebra, probability, statistics, and optimization.
\end{itemize}

Throughout the book, definitions will be provided in blue boxes in the following form:
\begin{definition}{Mitochondrion}
Mitochondria are the powerhouse of the cell.
\end{definition}
These will be displayed right before encountering the notion in the text.

Similarly, yellow boxes will contain additional information that is not crucial for understanding the main concepts introduced in the book. An example is provided below.
\begin{information}{Nests of Scarlet Tanagers}
Nests of scarlet tanagers are typically built on horizontal tree branches.~\cite{scarletwiki}
\end{information}

We hope that this book will pique readers' interest in TML and encourage them to contribute to this beautiful field of research.

\chapter{Introduction to Trustworthy Machine Learning}
\section{Scale is all we need?}

\begin{definition}{Generalization}
An ML model generalizes well if the rules found on the training set can be applied to new test situations we are interested in.
\end{definition}

\begin{figure}
    \centering
    \includegraphics[width=0.9\linewidth]{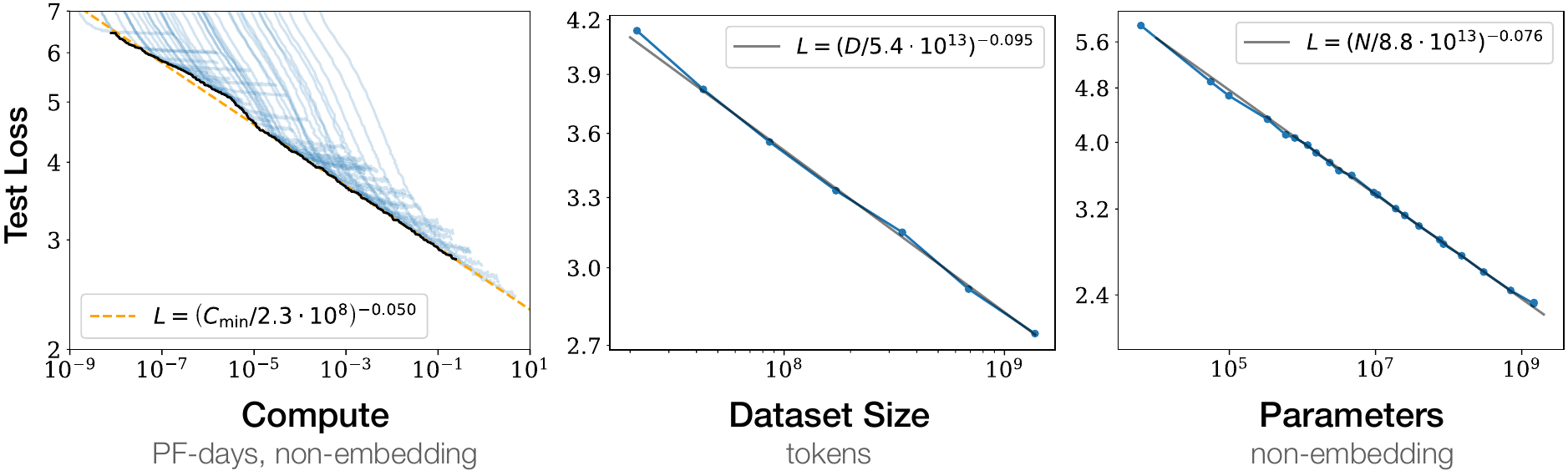}
    \caption{``Language modeling performance improves smoothly as we increase the model size, dataset [...] size, and amount of compute [with sufficiently small batch size] used for training. For optimal performance all three factors must be scaled up in tandem. Empirical performance has a power-law [i.e., \(y = a \cdot x^b\)] relationship with each individual factor when not bottlenecked by the two.''~\cite{DBLP:journals/corr/abs-2001-08361}. \(1 \text{ PF-day} = 10^{15} \cdot 24 \cdot 3600 \text{ floating point operations}\). Figure taken from~\cite{DBLP:journals/corr/abs-2001-08361}.}
    \label{fig:scaling}
\end{figure}

The story of Machine Learning (ML) seems to be that a bigger model with more data implies better test loss, as shown in Figure~\ref{fig:scaling}. Such models generalize well. Of course, more computing resources are needed, but more prominent tech companies possess them.

\begin{figure}
    \centering
    \includegraphics[width=\linewidth]{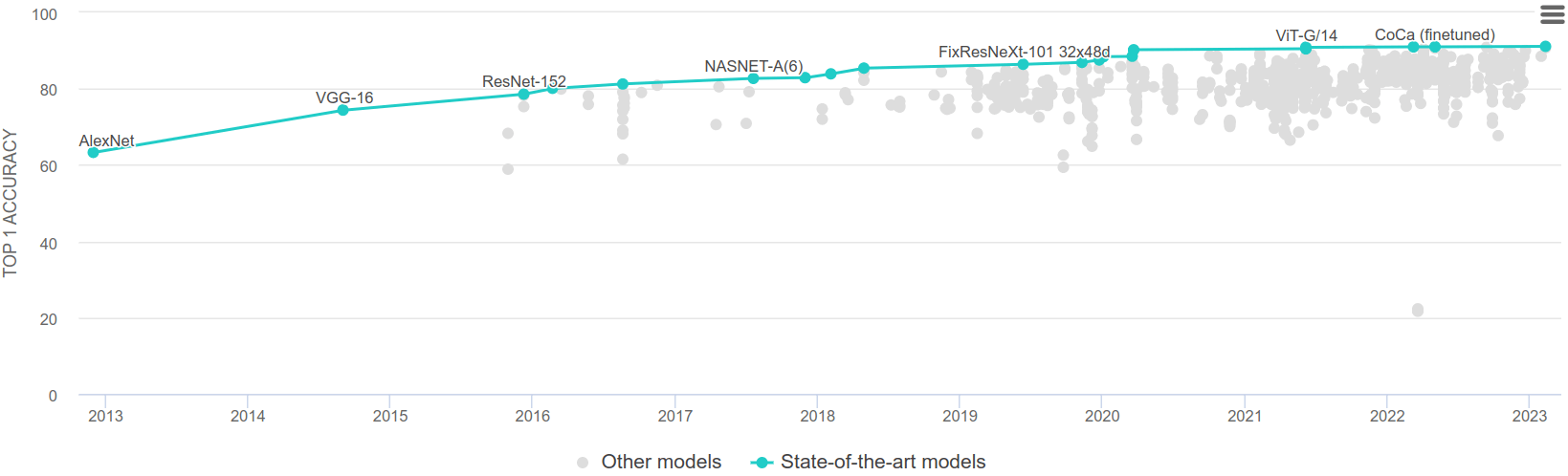}
    \caption{ImageNet top 1 accuracy leaderboard on 05.03.2023~\cite{imagenetleaderboard}. The performance of state-of-the-art methods plateaued over time.}
    \label{fig:leaderboard}
\end{figure}

Between 2013 and 2020, there was a steady increase in ImageNet~\cite{5206848} top-1 accuracy (Figure~\ref{fig:leaderboard}). This increase slowed over time, and between 2020 and 2023, we see a plateau in the top-1 accuracy -- seemingly, we ``solved ImageNet.''

\subsection{Are we done with ML?}

So, are we done with ML? If the reader's answer is `yes', then the following questions naturally follow:
\begin{itemize}
    \item Why do we not see ML used in every business?
    \item Why is ML not changing our lives yet?
    \item Why have we not gone through a quantum leap in productivity (results, profits, products) owing to ML?
\end{itemize}
If the reader's answer is `no', then we ask:
\begin{itemize}
    \item What are the remaining challenges in ML?
    \item How can we capture and measure those challenges?
\end{itemize}

This book aims to answer these questions while showcasing current state-of-the-art approaches in the field of TML.

\section{Key Limitations of ML}

Our answer is `no': Not all businesses use ML, and we have not yet gone through a quantum leap in productivity because of ML. Let us review the \emph{fundamental limitations of ML}.

\subsection{ML often does not work.}

ML models \emph{do} generalize, but not in the way one would expect. They tend to generalize well, given
\begin{enumerate}
    \item sufficient amount of data,
    \item appropriate inductive biases, and
    \item if we stay in the \emph{same distribution} as the training set (in-distribution (ID) generalization).
\end{enumerate}

Our models, however, need to cope with \emph{new situations} in practice. Whenever there are changes in the deployment conditions, our model will usually work \emph{much worse}.

\subsection{ML has high operating costs.}

So, we usually need to constantly adjust our model to the new settings. This requires
\begin{enumerate}
    \item an ML engineer (\(\cO\)(100k USD/year)),
    \item collecting fresh data (on dedicated pipelines) or buying specialized proprietary data, and
    \item computing resources or credits for an ML cloud to adjust the models on the new data.
\end{enumerate}

From a business perspective, these points boil down to a money issue. ML has high operating costs if our model constantly has to be adapted to new scenarios. If we had a model that generalized well, we would have less or even none of these costs.

\subsection{ML is currently not trustworthy.}

Even if we address the previous concerns, broad use of ML is not just a matter of whether our model works well or not -- \emph{it is difficult to trust ML models}. Extreme cases are when our \emph{life}, \emph{health}, or \emph{money} is at stake.

\textbf{Example 1}: Ten AI doctors say we have stomach cancer and recommend chemo- and radiotherapy. Could we trust this diagnosis and start these treatments? The majority of people would want to have the cancer pointed out in the MRT images. This is an example of \emph{explanability}.

\textbf{Example 2}: We are in a self-driving car driving through a curvy road along a cliff. Should we lift our hands off the wheel? Probably not. We likely \emph{could} not even do that because these cars would insist on human intervention (e.g., by giving warning signs). The automatic detection of an unusual environment is an example of \emph{uncertainty quantification}.

\textbf{Example 3}: It is also hard to trust images generated by \href{https://openai.com/product/dall-e-2}{DALL-E} to be sensible: We often see absurd artifacts in otherwise great ML-generated art. This is a problem of \emph{OOD generalization}, as our model only gives high-quality images for a restricted set of prompts.

\section{Topics of the Book}

The topics this book covers are as follows:
\begin{enumerate}
    \item \textbf{Out-Of-Distribution (OOD) Generalization.} Can we train a model to work well beyond the training distribution?
    \item \textbf{Explainability.} Can we make a model explain itself?
    \item \textbf{Uncertainty.} Can we make a model know when it does not know?
    \item \textbf{Evaluation.} How to quantify trustworthiness? How to measure progress?
\end{enumerate}

\medskip

The topics we do not cover but are also core parts of Trustworthy Machine Learning:
\begin{enumerate}
    \item \textbf{Fairness.} Demographic disparity is a core concern of fairness, which is the difference between the proportion of rejects and accepts for each population subgroup. The use of sensitive attributes (often implicitly) is also a significant problem regarding trustworthiness.
    \item \textbf{Privacy and Security.} Data are often proprietary and private. How to keep the data safe? Often we can reverse-engineer the original samples of the training set, e.g., in language models. This way, one can obtain sensitive, private information as well, e.g., medical records of patients.
    \item \textbf{Abuse of AI tools.} One can use ML to create deepfakes, e.g., to swap faces of people. Disseminating falsehood, e.g., via Large Language Models (LLMs), is also an alarming problem.
    \item \textbf{Environmental concern.} Accelerated computing consumes much energy.
    \item \textbf{Governance.} It is important to regulate the use of AI and formalize boundaries of AI usage.
\end{enumerate}

\section{Trustworthiness: Transition from ``What'' to ``How''}

To give an introduction to trustworthiness in ML, let us first define the ``What'' and ``How'' parts of an ML problem.

\begin{definition}{The ``What'' Part of a Problem}
The ``What'' part of a problem is learning the task we want to solve, i.e., the relationship between \(X\) and \(Y\). For example, the ``What'' part might be categorizing images into classes. The ``What'' point of view is that predicting \(Y\) given \(X\) is sufficient.
\end{definition}

\begin{definition}{The ``How'' Part of a Problem}
The ``How'' part of a problem specifies how a system comes to its prediction, what cues it is basing its decision-making on, and how it reasons about the prediction.

For robust AI systems, whether we solve a problem is not enough. How we solve it matters more.
\end{definition}

We currently have a ``What'' to ``How'' \emph{paradigm shift} in ML. Solving the ``What'' part is often \emph{not enough}, as detailed in the following section.

\subsection{Why Solving ``What'' is Not Enough}

A model can use multiple \emph{recognition cues} \(Z\) to make its prediction. These cues determine what the model bases its prediction on and what it exploits. There are \emph{two categories} of cues:
\begin{enumerate}
    \item \textbf{Causal, robust cue.} Such cues are robust to environmental changes, as the prediction is not based on that. Indeed, the label is \emph{caused} by this cue. We need to rely on causal, robust cues because otherwise, we will not generalize well to new domains. As an example, consider a car classification task. Then \(Z\) could be the car body region of the image, which is a robust cue.
    \item \textbf{Non-causal, spurious cue.} Such cues are hurtful for generalization. The label is not causally related to this cue, but they are \emph{highly correlated} in the dataset. In the car classification example, a highway in the background would be a spurious cue.
\end{enumerate}
When using vanilla training, nothing stops the model from using only non-causal, spurious cues, e.g., for recognition. The model can achieve high training accuracy (and even high in-distribution test accuracy) if the spurious cues are highly correlated with the label. Whenever the model faces an OOD dataset, however, it can perform arbitrarily poorly based on how predictive the learned bias cue is in the new setting.

\subsubsection{Shifted Focus in ML: The ``How'' parts of problems in Computer Vision}

In Computer Vision (CV), we might often be interested in whether an ML system is robust to perturbations. Examples include Gaussian noise, motion blur, zoom blur, brightness, and contrast changes. However, there are even more creative perturbations. For example, we might measure whether the ML system can still classify objects accurately in quite improbable positions.

Spurious cues that are highly correlated with the task cue but are otherwise semantically irrelevant can greatly harm a model's performance when not acted against. We often want to test whether our classifier exploits spurious cues. This can lead to it breaking down on OOD samples. For example, we can observe the behavior of the classifier in cases where the background is changed, the foreground object is deleted/changed, or the backgrounds and foregrounds are mixed across categories. If our model uses the image background as a spurious cue to make its predictions, it will showcase poor performance in these tests.

\subsubsection{Shifted Focus in ML: The ``How'' parts of problems in Natural Language Processing}

\begin{figure}
    \centering
    \includegraphics[width=\linewidth]{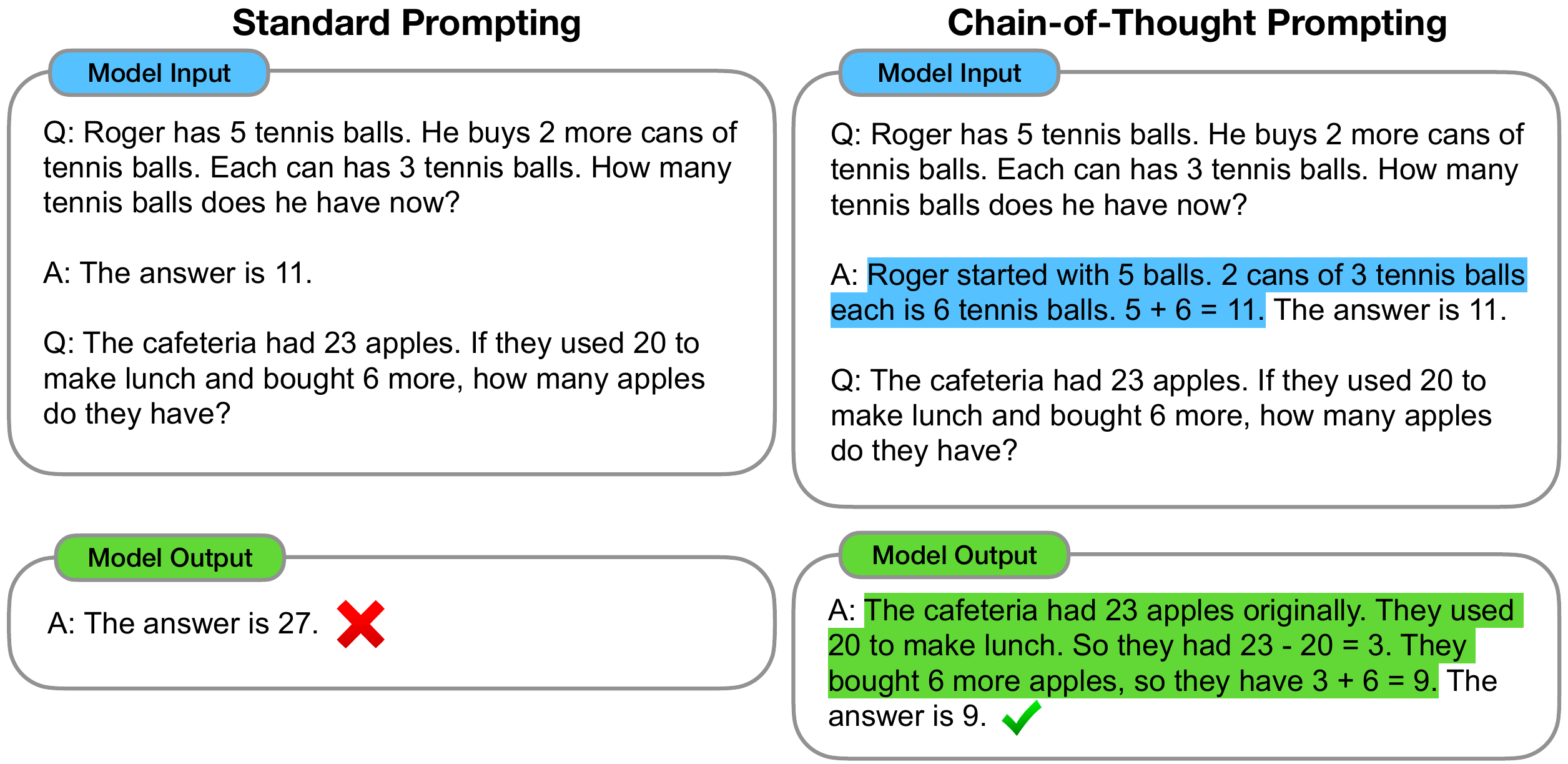}
    \caption{CoT Prompting can lead to better model answers. Figure taken from~\cite{wei2023chainofthought}.}
    \label{fig:cot}
\end{figure}

We would like to briefly mention Chain-of-Thought (CoT) Prompting. An example is given in Figure~\ref{fig:cot}. If we want to teach our Natural Language Processing (NLP) model a new task, we can provide it with some examples of the task and the correct answer and then ask a follow-up question. We supply no explanation of the answer in this case. What happens often is that the LLMs give incorrect answers to the next question. However, when prompting the model with exemplary detailed explanations of each correct answer, called CoT Prompting, the model also explains its prediction and even gets the answer right. It learns to rely on the right cues to provide the answer (and also provides an explanation).

\subsection{Machine Learning 2.0}

We distinguish two ML paradigms regarding what question they seek answers for: ML 1.0 and 2.0.

\begin{definition}{Machine Learning 1.0}
In ML 1.0, we learn the distribution \(P(X, Y)\) (or derivative distributions, such as \(P(Y \mid X)\)), either implicitly or explicitly, from \((X, Y)\) (``What'') data. ML 1.0 only considers the ``What'' task: It does not include the used cues, explanations, or reasoning, i.e., the ``How'' aspect \(Z\).
\end{definition}

\begin{definition}{Machine Learning 2.0}
In ML 2.0, we learn the distribution \(P(X, Z, Y)\) (or derivative distributions), either implicitly or explicitly, from \((X, Y)\) (``What'') data:

\[\text{Input } X \xrightarrow{\makebox[1.4cm]{}} \stackanchor{\text{Selection of cue, exact mechanism, reasoning.}}{\text{The ``How'' aspect \(Z\)}} \xrightarrow{\makebox[1.4cm]{}} \text{Output } Y\]
\end{definition}

The motivation of ML 2.0 is clear: we want to use the same kind of data to get more knowledge. However, the \(Z\)-problem is \emph{not guaranteed to be solvable} from \((X, Y)\) data. Learning \(P(X, Z, Y)\) contains all kinds of derivative tasks (a new set of tasks compared to what we had in ML 1.0): Now, we are trying to learn some distribution of \(X\), \(Z\), and \(Y\). For example, we may wish to be able to predict the Ground Truth (GT) \(Z\) from input \(X\) correctly (learn \(P(Z \mid X)\)), i.e., to make sure that given an input, the model is choosing the right cue for input \(X\).

In the following chapters, we aim to introduce the reader to various scalable trustworthy ML solutions with a focus on both theory and applications.

\clearpage

\chapter{OOD Generalization}
\section{Introduction to OOD Generalization}

OOD generalization stands as a pivotal challenge in modern ML research. It seeks to construct robust models that perform accurately even on data not represented in the training set. This branch of research not only elevates the trustworthiness and reliability of ML systems but also broadens their applicability in real-world scenarios.

Before we get our hands dirty, we have to discuss some terms that are often used in OOD generalization. Let us start with the most basic one: the \emph{task} we want to solve.

\begin{figure}
    \centering
    \includegraphics[width=0.8\linewidth]{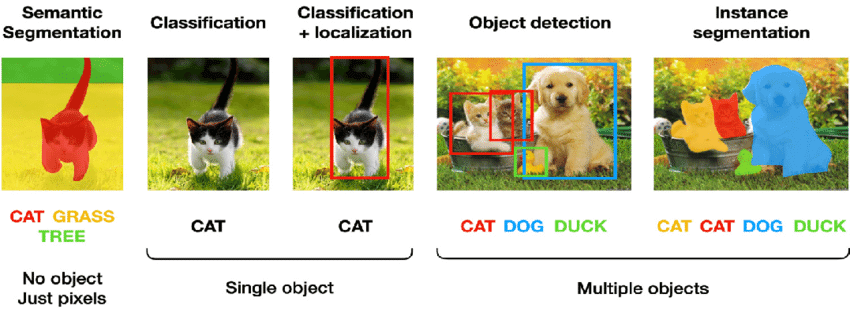}
    \caption{Illustrations of various computer vision tasks, taken from~\cite{article}. The field of computer vision is vast.}
    \label{fig:cvexamples}
\end{figure}

\begin{definition}{Task}
Task refers to the ground truth (GT), possibly non-deterministic (see aleatoric uncertainty in Section~\ref{ssec:types}) function that maps from the input space \(\cX\) to the output space \(\cY\) that a model is learning, or is a description thereof. Equivalently, the task is the GT distribution \(P(Y \mid X = x)\) we wish to model.

\medskip

\textbf{Alternative definition}: Task is the factor of variation (cue) that matters for us, i.e., the factor we want to recognize at deployment. Tasks are not inherent to the data; they are always defined by humans. This slightly differs from the previous definition, but both explain the same concept.
\end{definition}

\subsection{Examples of Tasks}

\begin{definition}{ImageNet}
ImageNet~\cite{deng2009imagenet} is a large-scale, diverse dataset initially created for object recognition research. Nowadays, it is popular to use ImageNet for classification, omitting the prediction of a bounding box. It contains millions of annotated images collected from the web and spans thousands of object categories that are organized according to the WordNet hierarchy for nouns. The dataset contains hundreds to thousands of samples per node in the hierarchy.

\medskip

\textbf{Ambiguity with ``the'' ImageNet Dataset}: The term ``ImageNet dataset'' has been used to refer to mainly two variants of the dataset which has caused a great deal of confusion:
\begin{itemize}
    \item \textbf{Full ImageNet Dataset/ImageNet-21K/ImageNet-22K}: The full ImageNet dataset contains 14,197,122 images associated with 21,841 WordNet categories~\cite{imagenetwiki}. However, not all of these images are used in typical computer vision benchmarks. ImageNet-21K is equivalent to ImageNet-22K, the difference is that some researchers round up the number of classes to 22,000 in the name.
    \item \textbf{ImageNet Large Scale Visual Recognition Challenge (ILSVRC) Dataset/ImageNet-1K/ILSVRC2017}: This is the most widely used subset of the ImageNet-21K dataset, involving 1,000 object categories. It contains 1,281,167 training data points, 50,000 validation samples, and 100,000 test images.~\cite{imagenetwiki}. However, the labels for the test set are not released. Therefore, one can only use the validation performance for evaluation when writing a paper, making the evaluation process less trustworthy. The annual ILSVRC competition, especially the 2012 challenge, which was won by the deep learning model AlexNet~\cite{10.5555/2999134.2999257}, played a pivotal role in the rise of deep learning.
\end{itemize}
\end{definition}

Even within ``classification'', there exist various tasks: different sets of classes correspond to different tasks.
\begin{itemize}
    \item The Pascal VOC datasets~\cite{Everingham10} consider 20 classes. These are datasets for object detection, instance segmentation, semantic segmentation, action classification, and image classification.
    \item The COCO datasets~\cite{cocodataset} contain 80 object categories and 91 stuff categories. Object categories strictly contain the Pascal VOC classes. These are datasets for object detection, instance segmentation, panoptic segmentation, semantic segmentation, and captioning. Crowd labels are added when there are too many (more than ten) instances of a class in an image. These aggregate multiple objects.\footnote{COCO is collected from Flickr. ImageNet is partly also from Flickr and other databases.}
\end{itemize}

\subsubsection{Examples of Tasks in Computer Vision (CV)}

An overview of CV tasks is given in Figure~\ref{fig:cvexamples}.
\begin{itemize}
    \item \textit{Semantic segmentation} aims to predict a semantic label for each pixel in an image.
    \item \textit{Classification} is the problem of categorizing a single object in the image.
    \item \textit{Classification + localization} aims classify \emph{and} localize a single object in the image.
    \item \textit{Object detection} classifies and localizes \emph{all} objects in the image. Now we have no restrictions on the number of objects the image might contain.
    \item \textit{Instance segmentation} assigns a semantic label and an instance label to every pixel in the image. The instance label differentiates between unique instances with the same semantic label.
\end{itemize}

\subsubsection{Examples of Tasks in NLP}

\begin{definition}{Semantic Analysis}
Semantic analysis in natural language processing (NLP) analyzes the conceptual meaning of morphemes, words, phrases, sentences, grammar, and vocabulary.
\end{definition}

\begin{definition}{Pragmatic Analysis}
Pragmatic analysis in NLP analyzes semantic meaning but also analyzes context. Instead of examining what an expression means, it studies what the speaker means in a specific context.
\end{definition}

\textit{Analysis tasks} aim to uncover syntactic, semantic, and pragmatic relationships between words/phrases/sentences in a document.
\begin{itemize}
    \item Tokenization is an essential syntactic analysis technique.
    \item The semantic analysis of a document might involve sentence classification (like sentiment analysis) or named-entity recognition.
    \item Word sense disambiguation is a particular example of pragmatic analysis. It aims to unfold which sense of a word is meant in a certain context.
    \item Part of speech tagging is can be both deemed a semantic and a pragmatic analysis technique. It marks up words in a document with the corresponding part of speech (e.g., noun or verb).
\end{itemize}

\textit{Generation tasks} involve generating text.
\begin{itemize}
    \item Machine translation is an example of conditional text generation where a translation in language \(B\) is generated given the original document in language \(A\).
    \item Question answering is also a conditional text generation problem where the model generates a coherent answer given a natural language question.
    \item Language modeling is the task of predicting the next word/character in a document or, equivalently, the task of assigning a probability to any text. Here, we condition on the partial sequence we have generated so far.
\end{itemize}

\subsection{Generalization Types}

Now, we are ready to consider a general overview of generalization types. First, let us introduce some terms that will play a crucial role in our discussion of OOD generalization.

\begin{definition}{Environment (Domain)}
The environment is the distribution from which our data are sampled.
\end{definition}

\begin{definition}{Cue (Feature, Attribute)}
Cues, features, and attributes all refer to the factors of variation in the data sample. Examples include color, shape, and size.

\medskip

\textbf{Note}: A cue is not necessarily a feature in a vector representation. Cues are also entirely independent of the model. They are characteristics of the dataset.
\end{definition}

\begin{definition}{In-Distribution (ID) and Out-of-Distribution (OOD) Samples}
In-distribution (ID) samples come from a test dataset which is used to gauge the model's performance on familiar data (in-distribution generalization). Out-of-distribution (OOD) samples, on the other hand, are drawn from a different test dataset to assess the model's performance on unfamiliar or unexpected data (out-of-distribution generalization).
\end{definition}

\begin{definition}{Generalization Types}

\medskip

\begin{center}
\label{tab:gentypes}
\begin{tabular}{l l m{0.32\linewidth} p{0.32\linewidth}} 
\toprule
\multicolumn{2}{c}{\textbf{Generalization type}} & \multicolumn{1}{c}{\textbf{How is training \(\boldsymbol{\approx}\) test?}} & \bfseries How is training \(\boldsymbol{\ne}\) test? \\
\midrule
\multicolumn{2}{l}{ID} & Training and test sets come from the same distribution. & We have different samples. \\ 
\midrule
\multirow{6}{*}{OOD} & Cross-Domain & \multirow{3}{\linewidth}{Training and test sets are for the same task.} & They are from different domains. \\
& Cross-Bias &  & They have different cue-correlations. \\
& Adversarial &  & Test samples are worst-case scenarios. \\
\bottomrule
\end{tabular}
\end{center}

\medskip

\textbf{Note}: This is not a comprehensive list of OOD generalization variants.
\end{definition}

Let us give examples for each scenario and consider some remarks.

\subsubsection{Example of ID Generalization}

We consider the task of recognizing a set of people from an office. They might be in different poses or situations, but always the same people, both in dev and deployment. The office theme will be common in the subsequent examples for different generalization types to highlight and emphasize the main differences between these.

\subsubsection{Example of Cross-Domain OOD Generalization}

Here, we might consider the task of recognizing person \(A\) from the office, but for the first time in a party costume during deployment. We have the same (or even different) people from dev in new, unseen clothes. One of the features is changing from training to test, meaning the training and test sets are from different domains. This generalization scenario mixes many factors; we will focus on cross-bias generalization more.

\subsubsection{Example of Cross-Bias OOD Generalization}

Persons \(A\) and \(B\) work in the office of the previous examples. We want to recognize person \(A\) for the first time in person \(B\)'s jacket. We have the same people but in exchanged clothes. The biased cue for person \(A\) has changed from their jacket to person \(B\)'s jacket. More formally, the cue that was highly correlated with person \(B\) in the training set now co-occurs with person \(A\) in the test dataset. The ML system will likely predict person \(B\) if we do not counteract the bias. This is because of the well-known shortcut bias of ML systems, which we will discuss later.

In practice, we are usually interested in changing the cue from training to test the model is likely to look at when making a prediction (because of shortcut bias), e.g., clothing. Such benchmarks test whether the model is focusing on a cue that is irrelevant to the task (e.g., a person's clothing is irrelevant to their identity).

\subsubsection{Example of Adversarial OOD Generalization}

Consider the problem of recognizing person \(A\) even when they hide their identity with a face mask (with someone else's face on it or using other tricks). Now person \(A\) is the adversary against our face recognition system, but this does not necessarily mean that person \(A\) has malicious intentions. Person \(A\) might wish to hide their true identity by making the model fail to recognize his face. There are also adversarial patterns to avoid facial recognition systems, e.g., to avoid surveillance. Adversarial generalization is a tough task, and it is even more challenging to obtain guarantees for this generalization type.

\section{Why do we even care about OOD generalization?}

In the YouTube video ``\href{https://www.youtube.com/watch?v=Zl9rM8D3k34&list=LL}{Self Driving Collision (Analysis)}''~\cite{collisionanalysis}, we see perfect weather and visibility, with low traffic. Nevertheless, as the Tesla turns onto the road, it does not detect a row of plastic bollards and hits them. This accident is not a one-off occurrence, as later in the video, it tries to hit other bollards too. Why does this happen? Because this is a new street arrangement that the model has not seen before, and it fails to generalize to this situation. To be sure that the model is robust in many situations, we need some kind of OOD generalization.

Many things constantly change in the world. New, unseen events happen all the time, like the Covid pandemic. If we trained a model before the pandemic to predict loungewear sales for a particular date,
it might have extrapolated well until national lockdowns were announced. These lockdowns caused a substantial domain shift, in which loungewear sales increased considerably. The model we trained before the lockdowns failed to reflect reality after this environmental change.

\begin{figure}
    \centering
    \includegraphics[width=0.8\linewidth]{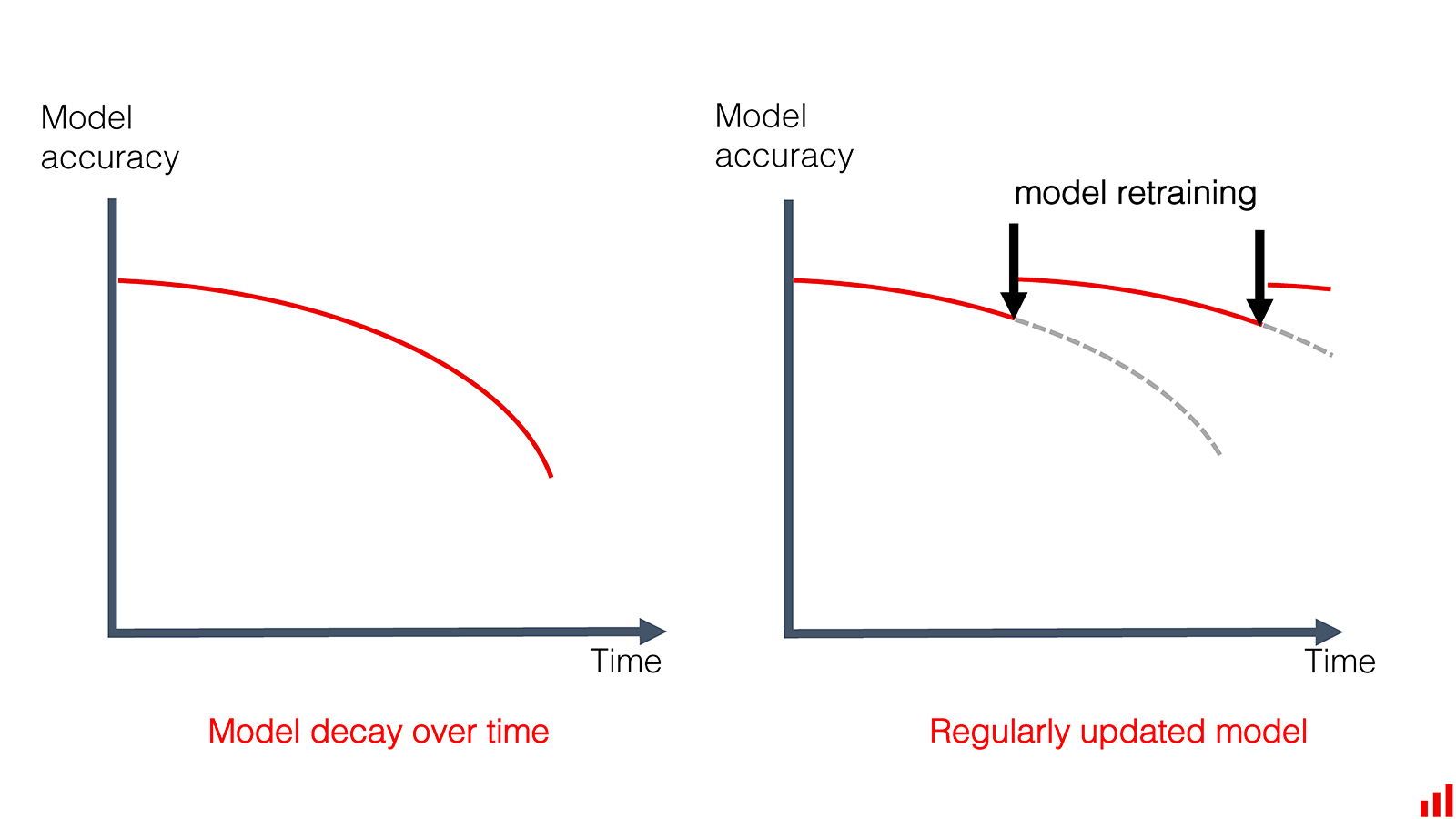}
    \caption{Illustration of the use of regular model updates to preserve deployment accuracy, taken from~\cite{modelupdate}. In many cases, model accuracy would plummet over time if we did not update it regularly.}
    \label{fig:decay}
\end{figure}

The typical solution to domain shifts is model retraining. Things inevitably change over time, and the model accuracy drop over time is unavoidable if the model is kept fixed. People thus often recollect data, annotate new samples, and retrain the model on new data. We can use this procedure to keep the model's accuracy above a certain threshold, illustrated in Figure~\ref{fig:decay}.

\begin{definition}{Model Selection}
Model selection is the process of selecting the best model after the individual models are evaluated based on the required criteria. One usually has a pool of models specialized for various domains. The expert chooses the best model for the current deployment scenario. For example, Amazon often performs model selection in its cloud services.
\end{definition}

\begin{figure}
    \centering
    \includegraphics[width=\linewidth]{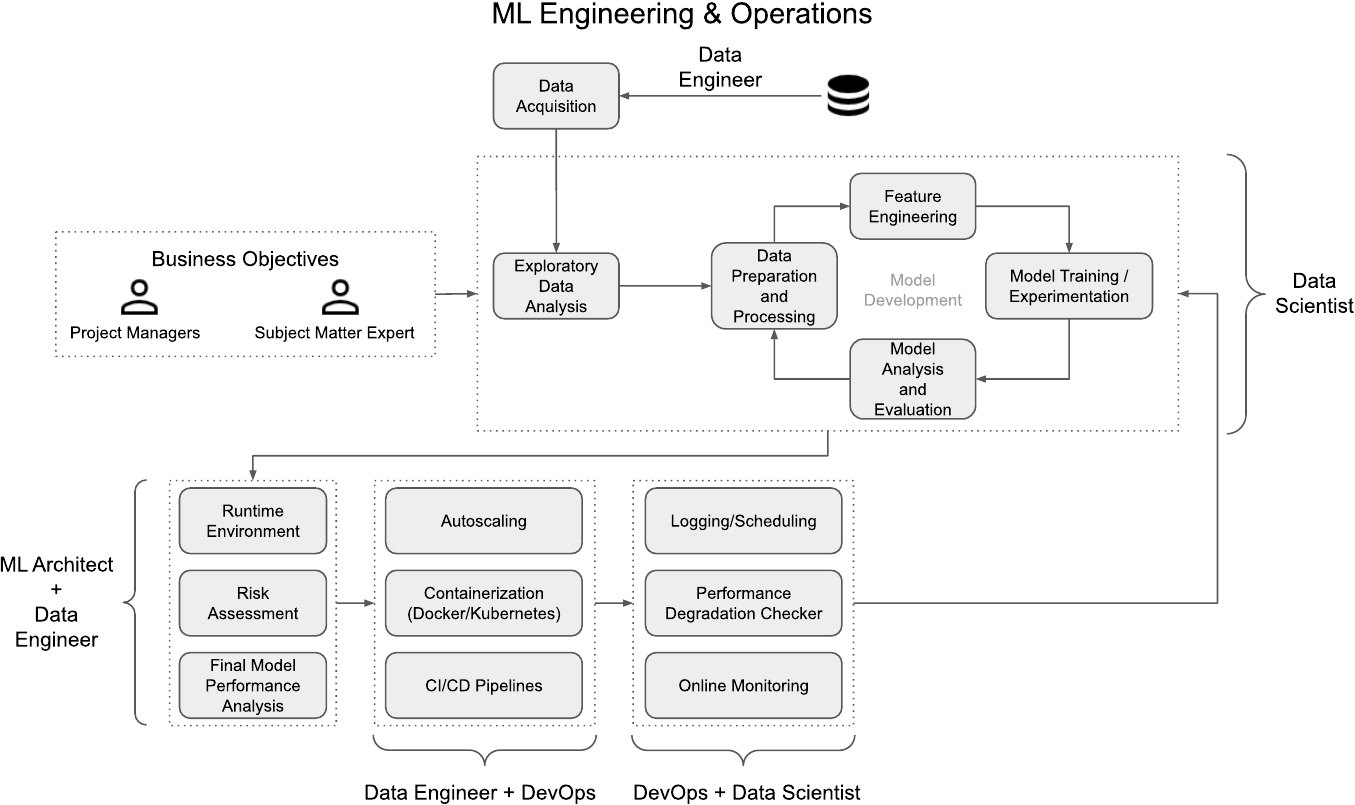}
    \caption{MLOps is a complex discipline with multiple participants. \textbf{Note}: Data Acquisition is not just a DB query. It also includes the collection of data. The data curation procedure can take a long time. One must keep track of shifting data (data versions), keep annotators in the loop, and update models accordingly. This procedure can be very costly. Figure adapted from~\cite{mlops}.}
    \label{fig:mlops}
\end{figure}

\begin{definition}{MLOps}
MLOps is an engineering discipline that aims to unify ML systems development and deployment to standardize and streamline the \emph{continuous delivery} of high-performing models in production~\cite{mlops}. An overview is given in Figure~\ref{fig:mlops}.
\end{definition}

However, the constant retraining of models and the model selection expertise (MLOps) is costly.
\begin{itemize}
    \item \textbf{Manpower}: 100k EUR/person/year at least.
    \item \textbf{GPUs, electricity}: 25k EUR/year + 8000 kg \(\text{CO}_2\)/year (\href{https://cloud.google.com/products/calculator#id=457292aa-54c3-471e-91bb-d418e7dd7032}{considering a single NVIDIA Tesla A100 unit and Google Cloud}).
    \item \textbf{Data management} (schema, maintenance) is also expensive.
\end{itemize}

\begin{information}{NVIDIA Tesla A100}
The NVIDIA Tesla A100 is a tensor core GPU often used for training ML models. It can be partitioned into 7 GPU instances so multiple networks can efficiently operate simultaneously (training or inference) on a single A100. In early 2023, it has one of the world's fastest memory bandwidths, with over \(2\) TB/s. Training BERT is possible \emph{under a minute} using a cluster of 2048 A100 GPUs~\cite{nvidiaa100}.
\end{information}

\begin{definition}{DevOps}
A set of practices intended to reduce the time between committing a change to a system and the change being placed into production while ensuring high quality.~\cite{devopsdef}
\end{definition}

ML problems arise from business goals. If there is no distribution shift and no need for model selection, there is no need for MLOps, and we only need DevOps. We need MLOps (continuous updates of models) because the data, user, and environment shift continuously. Ideally, we only have to perform continuous updates semi-automatically: We only need a few people to maintain the system. Eventually, however, we wish to get over MLOps as well. We need models that are very robust to domain shifts to achieve this.

\subsection{Greater Levels of Automation}

First, we define \textit{diagonal datasets} that will help us understand the levels of automation in ML and the ill-defined behavior in OOD generalization (Section~\ref{ssec:spurious}).

\begin{definition}{Diagonal Dataset}
A dataset in which all (or multiple in general) cues vary at the same time (i.e., they are perfectly correlated) that can be used to achieve 100\% accuracy. Thus, it is impossible to infer what the deployment task is from the label variation. A model using either of the perfectly correlated cues could achieve 100\% training accuracy.
\end{definition}

Next, we need to describe the Amazon Mechanical Turk service to reason about annotation costs and crowdsourcing.

\begin{definition}{Amazon Mechanical Turk}
The \href{https://www.mturk.com/}{Amazon Mechanical Turk} (AMT) is an online labor market for dataset annotation, where one can crowdsource their annotation task.
\end{definition}

We consider five levels of automation (1: lowest, 5: highest) in problem-solving.

\subsubsection{Level 1: No ML}

In this case, we have no ML model to use for our particular problem. The human effort is gigantic: A center with hundreds of personnel is constantly required (which was a common case 40-50 years ago). They take care of input streams on the fly, i.e., they are processing a continuous data stream with \emph{human intelligence}. This procedure is \emph{very costly} and \emph{inefficient}.

\subsubsection{Level 2: MLOps with Periodic Annotation}

In this setup, we have an ML model available to help with our problem. However, this model can only generalize to the same distribution based on the annotated samples. The human effort is reduced but still considerable: A group of people annotates thousands (possibly millions across projects) of samples every month, as the world is changing quickly. Options for annotations include in-house annotators, outsourcing to annotation companies, or crowdsourcing through AMT. Annotation costs 10-30 USD per hour per person on AMT. (Slightly above minimum wage for US workers.) Harder tasks, e.g., instance segmentation, cost more. For the browser-based annotation of 1 million images, we estimate up to 1 million USD for AMT crowdsourcing. An ML engineer's market price is 100-300k USD per year per person. These costs are prohibitively expensive for small businesses.

\subsubsection{Level 3: MLOps with Reduced Annotation}

Now, we have an ML model that is minimally resilient to distribution shifts. The human effort is reduced even more: Annotation is required only every year. This resilience reduces the cost of MLOps quite a bit.

\subsubsection{Level 4: MLOps with No Annotation}

In this hypothetical scenario, our ML model -- once trained -- is so robust against distribution shifts that it only requires minimal human engineering (e.g., hyperparameter adaptation and model selection). Regarding the human effort, annotation is not needed anymore. Only ML engineers are needed to select the right model suitable for the task at that particular time (based on the needs of business executives). They are also constantly looking for the best models.

\subsubsection{Level 5: ML without MLOps}
\label{sssec:level5}

Here, even the ML engineer functionality is (partly) automatized. The model can alter its hyperparameters to adapt to changing distributions. Adapting hyperparameters usually requires fine-tuning; however, the way we choose hyperparameters can be made very efficient, e.g., requiring only very few observations of training sessions and data samples. (In ML, we always need observations.) It can even be automatized with, e.g., Bayesian optimization. Importantly, this does not refer to a meta-model that can automatically choose between the set of candidate models. We cannot even be sure that is possible, as certain factors cannot be inferred from the data. As an example, let us consider a diagonal dataset in which the shape and color cues co-occur perfectly. At one point, users might want a shape-based classifier. Later they might change their mind and want a color-based classifier. This requirement is not reflected in the data stream for a diagonal dataset: it is part of the human specification. This is precisely why model selection always involves human feedback.

Why is an expert still needed for model selection? One might wonder why an expert decision-maker cannot be replaced in this very idealistic hypothetical scenario. This is because some metrics are often unreliable (that look good on paper, but the model that performs well on them might not be what we want), and there are requirements from a model that are often hard to quantify. An ML engineer might also be needed to keep the pool of models up to date, including the latest innovations in ML. There are also always new model architectures and general technologies that appear. This pool needs to be constantly curated and updated to the general needs of the users. These new models might also not be better than previous ones on \emph{all} criteria, just some of them (e.g., better accuracy at the cost of less interpretability).

There might also be many criteria to adhere to. For example, we might be interested in the performance, computational resources, fairness, calibration, or explainability. Accuracy is not the only criterion, and there is no \emph{single} criterion. The single \emph{best} model does not exist in general, no matter how robust our pool of models is; and even if our pool of models is robust, some models might perform (slightly) better in exact deployment scenarios on certain metrics -- we want to squeeze out performance. Model selection is not just an argmax. With multiple criteria, it is often too difficult to put some weights on these metrics and use thresholding. Automating model selection is, therefore, a challenging problem with fundamental limitations.

Finally, an expert is always needed to \emph{give the final word}. They must make an executive decision and choose the best model based on the business needs. When there are problems with a new model (e.g., fairness), a human must intervene and roll it back to a previous state. \textbf{Note}: The expert discussed here does not have to be an ML expert. The main decisions usually come from business executives.

\subsection{Once we ``solve'' OOD generalization...}

What happens if we ``solve'' OOD generalization (i.e., our models become robust to distribution shifts)?
\begin{itemize}
    \item Our model will work well even under new situations.
    \item MLOps will not be needed at the current scale. (However, model selection and ML expertise will probably be needed for a long time.)
    \item Small businesses will be able to adopt ML more easily.
    \item ML can be extended to more risky applications because we can be sure that it will work in novel situations, too.
    \item ML will drive the risky applications, e.g., the industry of healthcare, finance, or transportation. Robust models gain trust. However, we will see later that \emph{explainability} is just as important.
\end{itemize}

To summarize our introduction to OOD generalization and drive the key points across:
\begin{itemize}
    \item \red{ML is still costly because it requires periodic annotation and maintenance. There are huge human costs involved.}
    \item \green{When ML models generalize well to novel situations, costs will be reduced.}
\end{itemize}

\section{Formal Setup of OOD Generalization}
\label{ssec:formal}

\subsection{Stages of ML Systems}

To discuss a more formal setup of OOD generalization, let us first consider two stages of ML systems: \emph{development} and \emph{deployment}.

\begin{definition}{Development (dev)}
Development is the stage where we train our model and make design choices (for hyperparameters) within some resource constraints.
\end{definition}

\begin{definition}{Deployment}
Deployment is the stage where our final model is facing the real-world environment. This environment is called a \emph{deployment environment} and can change over time.
\end{definition}

\begin{definition}{Training}
Training is the particular action of fitting the model's parameters within the dev stage to the training set, with a fixed hyperparameter setting.

\medskip

We do not separate the training phase from the rest of the dev phase, but we \emph{do} separate dev from deployment.
\end{definition}

\begin{definition}{Testing}
Testing is a lab setup designed to mimic the deployment scenario closely -- scientists evaluate their final inventions on test benchmarks and report their results.

\medskip

\textbf{Practice point of view}:
\begin{itemize}
    \item This is different from deployment and still a part of development.
    \begin{itemize}
        \item If we want to be precise: As soon as we have labeled samples from deployment (and we make any design choices based on these or just test our model), we are using information from the deployment setup in dev. We cannot talk about true (domain or task) generalization anymore. The deployment scenario should stay fictitious and unobserved in such settings.
    \end{itemize}
\end{itemize}

\medskip

\textbf{Academia point of view}:
\begin{itemize}
    \item The test set (\ref{ssec:splits}) and the action of testing is treated as a part of the deployment.
\end{itemize}
\end{definition}

The specification of these stages can be bundled into one \emph{setting}.

\begin{definition}{Setting/Setup}
A setting specifies the available resources (during development) and an ML system's surrounding (deployment) environment.

\medskip

\textbf{Essential components of a setting}:
\begin{itemize}
    \item \textbf{Development resources}: What types of datasets, samples, labels, supervisions, guidance, explanation, tools, knowledge, or inductive bias are available?
    \begin{itemize}
        \item ML engineers are also resources. They have their own knowledge to optimize an ML model the right way. If we have better engineers with better intuition of what to do in a scenario, we can train the model quicker and better.
    \end{itemize}
    \item \textbf{Deployment environment}: What kind of distribution will our ML model be deployed on?
    \item \textbf{Time}: Resource availability changes over time. The deployment environment changes over time. We can only deploy after development, but sometimes we keep developing after deployment.
\end{itemize}
\end{definition}

\subsubsection{Example of a Setting}

Consider an ID supervised learning setup. This is an ideal scenario ML research has started its exploration in. Various strong results about consistency, convergence rates, and error bounds can be given in this setup~\cite{jiang2019non,NEURIPS2021_0e1ebad6} that break in OOD settings.

Our \textit{development resources} are labeled \((X, Y)\) samples from distribution \(P\). Our \textit{deployment environment} contains unlabeled samples \(X\) from distribution \(P\) presented one by one. \textbf{Note}: This is an incomplete description of the development resources and the deployment environment that aims to drive the main points across. In scientific papers, a much more thorough description is required.

We usually specify settings when we have an actual task we want to work on, i.e., we have a \emph{real-world scenario} at hand.

\begin{definition}{Real-World Scenario}
A real-world scenario is a projection of a setting onto a hypothetical or actual convincing real-world example. This is a particular situation that fits the setting.
\end{definition}

\subsubsection{Example of a Real-World Scenario}

Consider an ID supervised learning setup again (the simplest case). Our \textit{task} is to build a system for detecting defects (e.g., dents) in wafers (semiconductors, pieces of silicon used to create integrated circuits) through image analysis. Our \textit{development resources} contain a dataset of wafer images with corresponding labels -- defective or normal. In our \textit{deployment environment}, the images are of the same distribution, as the wafer products and camera sensors are identical between the dev dataset and the data stream from deployment.

\begin{information}{How to compare methods with different resources?}
We always want to compare methods fairly. If one method uses fewer resources in development, we cannot compare the two methods fairly.
\end{information}

\subsection{Dataset Splits in ML}
\label{ssec:splits}

Next, we discuss different general dataset splits used in ML.

\begin{definition}{Training Set}
The training set is a (usually large) collection of samples whose purpose is to train the model.

\medskip

\textbf{What is optimized?} Model parameters.

\medskip

\textbf{What is the objective?} The training loss, possibly with regularization.

\medskip

\textbf{What is the optimization algorithm?} A gradient descent variant using Tensor Processing Units (TPUs), or GPUs.

\medskip

\textbf{How frequent are updates of the model?} \(\cO\)(milliseconds-seconds).
\end{definition}

\begin{definition}{Validation/Dev Set}

The purpose of the validation set is to roughly simulate the deployment scenario by using samples the model has not seen yet and measure ID generalization.

\medskip

\textbf{What is optimized?} Model hyperparameters and design choices.

\medskip

\textbf{What is the objective?} Generalization metrics.
\begin{itemize}
    \item If we consider true OOD generalization without having access to the target domain (i.e., not domain adaptation (\ref{sssec:da}) or test-time training (\ref{sssec:ttt})), we cannot measure OOD generalizability on the validation set. Therefore, the validation set usually comes from the same domain(s) as the training set. Otherwise, we would already tune our hyperparameters on the domain we wish to generalize to; thus, whether we measure OOD generalization on the test set later is questionable. Such scenarios exhibit `leakage', which we will cover in Section~\ref{sssec:leakage}.
\end{itemize}
\textbf{What is the optimization algorithm?} For example, Bayesian optimization, ``Grad student descent'', random search.

\medskip

\textbf{How frequent are updates of the model?} \(\cO\)(minutes-days).  
\end{definition}

\begin{definition}{Test Set}
The test set is used to simulate the deployment scenario more accurately than during validation by using samples from the distribution we believe the model will face during deployment. The test set can, therefore, also measure OOD generalization.

\medskip

\textbf{What is optimized?} The methodology and overall approach through the shift of the field.
\begin{itemize}
    \item For example, the shift from CNNs towards ViTs.
    \item The line is unclear between the change of methodology and design choices; this is more like a spectrum.
\end{itemize}
\textbf{What is the objective?} Generalization metrics.
\begin{itemize}
    \item The test set can be any type of OOD dataset.
\end{itemize}
\textbf{What is the optimization algorithm?} Paradigm shifts, updating the evaluation or the evaluation standard.
\begin{itemize}
    \item As the field changes, the set test also changes. For example, for ImageNet, many test sets
    are available (e.g., for generalization to different OOD scenarios), and more have been added over time.
    \item We are setting a new goal for the field that many researchers will follow.
    \item Standard refers to the benchmark, metric, or protocol according to which we evaluate our models. (It has a close connection to the test sets we use.)
\end{itemize}
\textbf{How frequent are updates of the model?} \(\cO\)(months-years).
\begin{itemize}
    \item In the scale of months and years, methods are \emph{meant to be optimized} to the test set. The problems this optimization entails are crucial to understand and are discussed in detail in Section~\ref{ssec:idealism}.
    \item The test set must be updated to the user and societal needs over time. Naturally, the training set and validation set also change over time.
\end{itemize}
\end{definition}

\subsection{Why Idealists Cannot Evaluate on the Test Set}
\label{ssec:idealism}

We measure accuracy on the test set because we wish to \emph{compare} our method to previous methods. This is an implicit way of choosing a model over other methods, which is part of the methodology. Therefore, the test set is still a part of development in practice in the most precise sense.\footnote{For domain generalization (Section~\ref{ssec:domain}), we never get any annotations from deployment in reality. We consider the deployment scenario as a fictitious entity.}

Whenever we make any decisions based on test results (be it ours or others'), we cannot measure generalizability on the test set anymore. This is almost always violated in practice. However, there is no clear workaround, as benchmarks are essential to progress in any field of ML research. We can only ``spoil the test set less,'' but we can never \emph{not} spoil it if we want to advance the field.

\section{Common Settings for OOD Generalization}

There is no such thing as \emph{the} OOD generalization setting. There are many different scenarios for it. Let us first explain why differentiating between these learning settings is important.

\subsection{Why are the learning settings important?}

\begin{figure}
    \centering
    \includegraphics[width=0.6\linewidth]{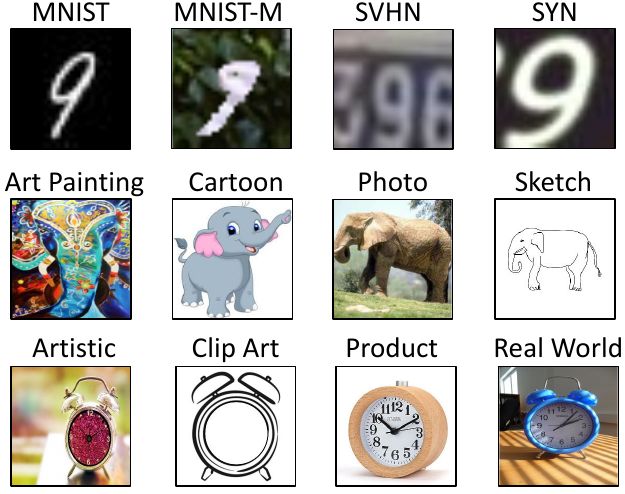}
    \caption{Collage of different domain labels and corresponding images, taken from~\cite{https://doi.org/10.48550/arxiv.2003.06054}. Images of the same kind of objects can be surprisingly different when considering different domains.}
    \label{fig:domainlabels}
\end{figure}

Let us first define the notion of \emph{domain labels}.

\begin{definition}{Domain Label}
The domain label is an indicator of the source distribution of each data point in the form of a categorical variable (e.g., dataset name).
\end{definition}

\textbf{Example}: ``MNIST'' can be a domain label for an image from the MNIST dataset~\cite{lecun2010mnist}. Samples in different datasets are (almost always) coming from different distributions. Other valid domain labels include ``Art Painting'', ``Cartoon'', ``Sketch'', or ``Real World'', as illustrated in Figure~\ref{fig:domainlabels}.

Distinguishing various learning settings is of crucial importance for the following reasons.
\begin{enumerate}
    \item To figure out which techniques can be used for the given learning scenario. We want to understand the given ingredients precisely and know the relevant search keywords for googling the papers.
    \item To compare against previous methods in the same learning setting. It is key to enumerate the exact (and sometimes hidden) ingredients used by a method and compare it only with methods that use the same ingredients. Some authors may give misleading information about the setting their method operates in. For example, if one claims to have not used domain labels but has used some equivalent form of them, we must be able to notice that and voice our concerns. Comparing methods based on their ingredients is much more justified than comparing based on the name of the settings the authors \emph{claim} to adhere to.
\end{enumerate}

\subsection{ID generalization}

\begin{figure}
    \centering
    \includegraphics[width=\linewidth]{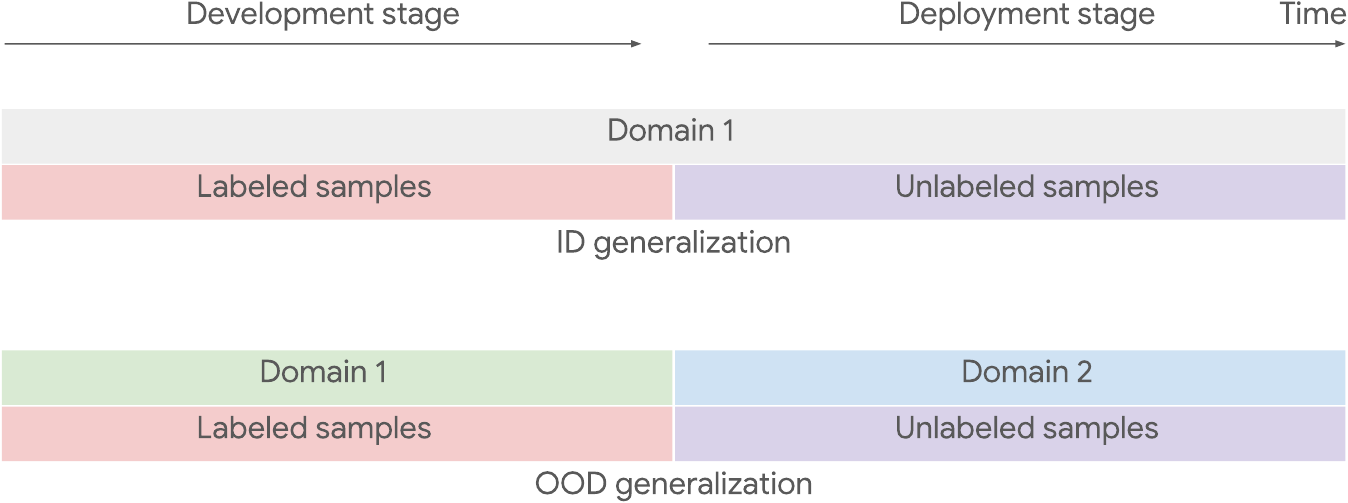}
    \caption{Illustration of the ID generalization setting (top) and the general OOD generalization setting (bottom). OOD generalization showcases a change of domain.}
    \label{fig:id_ood}
\end{figure}

For the sake of comparison, let us start with ID generalization. An illustration of this setting is depicted in Figure~\ref{fig:id_ood}. We have the same domain all the way through development and deployment. During deployment, we get unlabeled samples to which we wish our model to generalize.

\subsection{Domain-Dependent OOD Generalization}  

A general view of this setting is shown in Figure~\ref{fig:id_ood}. There are different domains for development and deployment. One needs to generalize to the deployment domain. This is the most general setting for OOD generalization.
There are many names of settings. Exact definitions of settings for OOD generalization fluctuate. Therefore, understanding the exact ingredients for each setting matters much more than ``which name to put''. For the purpose of the book, we will still go over the settings and try to put definite boundaries. We discuss different categorizations of OOD generalization settings below.

\subsubsection{Categorizing according to the nature of the difference between dev and deployment}
\begin{itemize}
    \item In \textit{cross-domain} generalization, the deployment environment contains completely unseen cues in dev.
    \item In \textit{cross-bias} generalization, deployment contains unseen compositions of seen cues in dev.
    \item \textit{Adversarial} generalization considers a (real/hypothetical) adversary in deployment who tries to choose the worst-case domain.
\end{itemize}

\subsubsection{Categorizing according to the extra information provided to address the ill-posedness}
\begin{itemize}
    \item In \textit{domain generalization}, domain labels are provided.
    \item In \textit{domain adaptation}, some (un-)labeled target domain samples are available in dev.
    \item \textit{Test-time training}'s dev continues even after deployment. We get access to deployment (target domain) samples. We may or may not label them.
    \item \textit{Domain-incremental continual learning} considers a single domain during dev. Domains are added over time during deployment.
\end{itemize}
These settings all come with different ingredients, and one should not compare methods across different settings. The two axes of variation above are independent.

\subsection{Domain Adaptation}
\label{sssec:da}

\begin{figure}
    \centering
    \includegraphics[width=\linewidth]{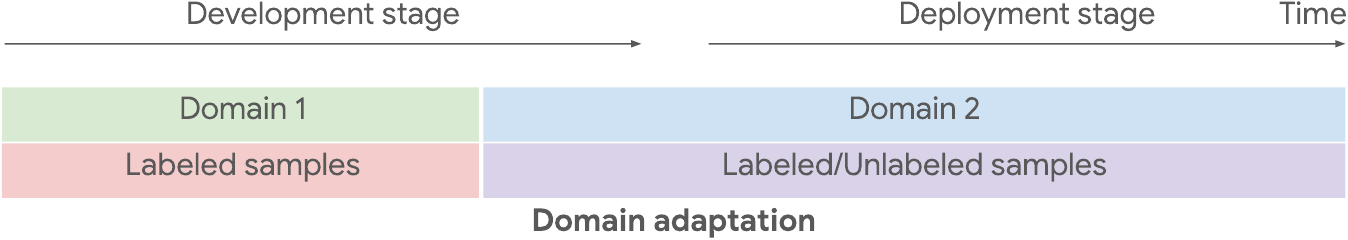}
    \caption{Domain adaptation setting. The development stage also comprises samples from domain 2.}
    \label{fig:da}
\end{figure}

Domain adaptation is illustrated in Figure~\ref{fig:da}. During the dev stage, we have access to some labeled or unlabeled samples (depending on the exact situation) from the deployment environment. We can, e.g., align our features with the target domain statistics using moment matching.

\subsubsection{Moment Matching in Domain Adaptation}
\begin{figure}
    \centering
    \includegraphics{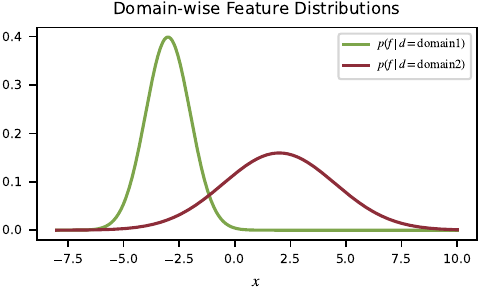}
    \caption{Feature embedding distributions can be notably different between the domains available in the development stage of domain adaptation. The figure shows Gaussians fit to domain-wise empirical feature distributions of samples from different domains for feature alignment in domain adaptation using moment matching: we aim to match these Gaussians among domains. \(f\): feature, \(d\): domain.}
    \label{fig:dist}
\end{figure}

An example of domain-wise feature distributions is illustrated in Figure~\ref{fig:dist}. These can, e.g., correspond to the penultimate layer's sample-wise activations in a ResNet-50. We represent the empirical distribution of the domain-wise feature values by their expectation and variance (approximated by the sample mean and variance). For domain 2, we have a few unlabeled samples that can be used for this computation. We place, e.g., an \(L_2\) penalty on the differences between domain 1 and 2 statistics (sample mean and variance in our example) of intermediate features, or we can also consider the Wasserstein distance between the Gaussians as a penalty. During training, for samples from domain 1, we compute the task loss (e.g., cross-entropy) and the penalty term. For samples from domain 2, we only compute the penalty, as there are no labels for these samples. We only backpropagate gradients of the penalty through the labeled domain 1 samples and use the unlabeled samples only to calculate the penalty. This way, we only directly train on domain 1 but adapt our model based on domain 2 samples to generalize to domain 2. This tends to give us a small amount of improvement in robustness.

\subsection{Domain Generalization}
\label{ssec:domain}

\begin{figure}
    \centering
    \includegraphics[width=\linewidth]{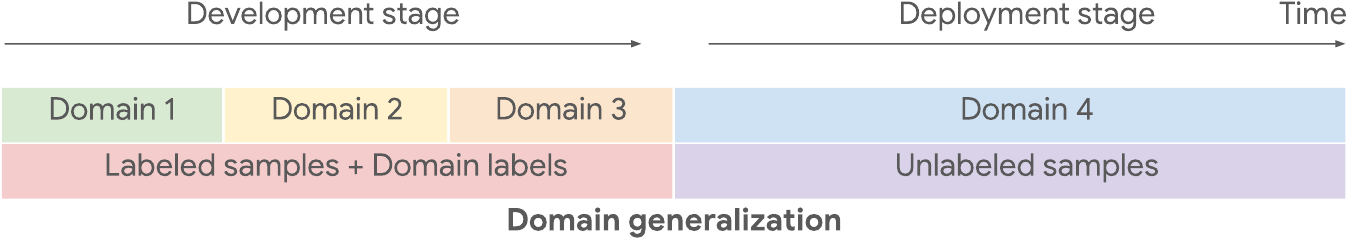}
    \caption{Domain generalization setting. We have access to multiple domains during the development stage, but we have to generalize to a novel, unseen domain in the deployment stage.}
    \label{fig:dg}
\end{figure}

An overview of domain generalization is given in Figure~\ref{fig:dg}. During the dev stage, we have access to labeled samples from multiple domains. We also know the domain label for every sample. Knowing domain labels is usually a hidden assumption; not many papers talk about this. If we do not know the domain labels, there are techniques for detecting the domains without them, but these are never perfect and come with additional assumptions.

\subsubsection{Moment Matching in Domain Generalization}

We can ``unlearn'' domain-related characteristics in our representation by performing \emph{moment matching} similarly to domain adaptation, but now between all domains available in the development stage. Similarly to the domain adaptation case, we compute the sample mean and variance separately for each domain as we have domain labels.\footnote{The task labels are not used for moment matching, only to compute the task loss.} We fit Gaussians to the features of samples from different domains. We align the Gaussians for different domains by, e.g., placing an \(L_2\) penalty on the pairwise differences between their corresponding means and covariance matrices. We backpropagate gradients through all samples, using both the task labels and domain labels. If we succeed, we ignore differences among domains in the training set based on moments. We hope that the model becomes independent of domain information (of any kind), so it will probably work well on the next (unknown) domain. 

\subsection{Test-Time Training}
\label{sssec:ttt}

\begin{figure}
    \centering
    \includegraphics[width=\linewidth]{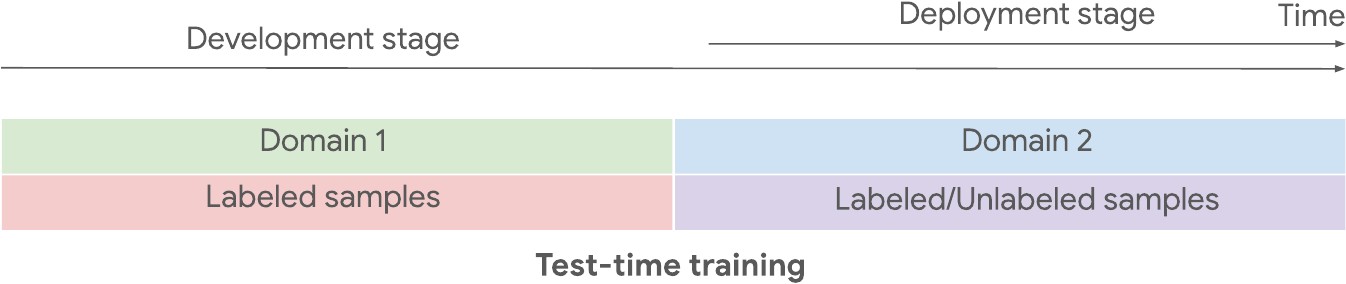}
    \caption{Test-time training setting. The development stage continues in the deployment stage.}
    \label{fig:ttt}
\end{figure}

Test-time training is shown in Figure~\ref{fig:ttt}. After training our model, we keep updating it according to the labeled or unlabeled samples (depending on the exact setting) from the deployment environment. Thus, dev continues even into the deployment because our model keeps being updated. We might not do labeling in domain 2, but it helps to have access to incoming domain 2 samples and correct the feature model on the fly (e.g., by performing moment matching).

This paradigm is becoming more popular these days. A key figure of the approach is Alexei Efros.

\subsection{Domain-Incremental Continual Learning}

\begin{figure}
    \centering
    \includegraphics[width=\linewidth]{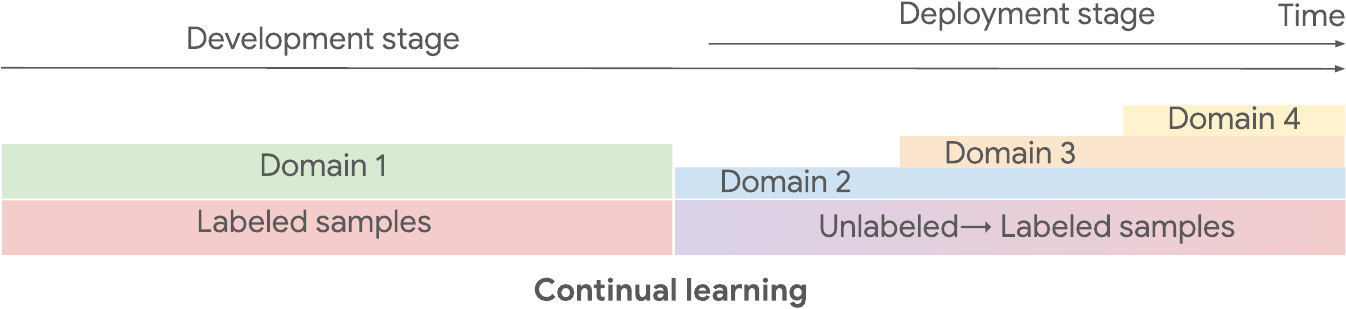}
    \caption{Domain-incremental continual learning. New domains are added over time in the deployment stage.}
    \label{fig:cldi}
\end{figure}

An overview of the domain-incremental continual learning setting is given in Figure~\ref{fig:cldi}. We train on a single domain before deployment. Domains are added over time during deployment. We label a few samples over time and update our model on the way. Only the labeled samples are used for improving our model. We hope that the model also does not forget previous domains. Performance should remain as high as possible for previous domains as well. Data keeps coming from all deployment domains, but we must adapt quickly to the new domain.

\subsection{Task-Dependent OOD Generalization}  

In general, the task being different is a lot harder than the domain being different. Usually, a different task also means a different domain.\footnote{However, we can also come up with counterexamples. When task 1 is to predict numbers 0-4 on MNIST and task 2 is to predict numbers 5-9, the domain stays the same, but the task changes.}

\begin{figure}
    \centering
    \includegraphics[width=\linewidth]{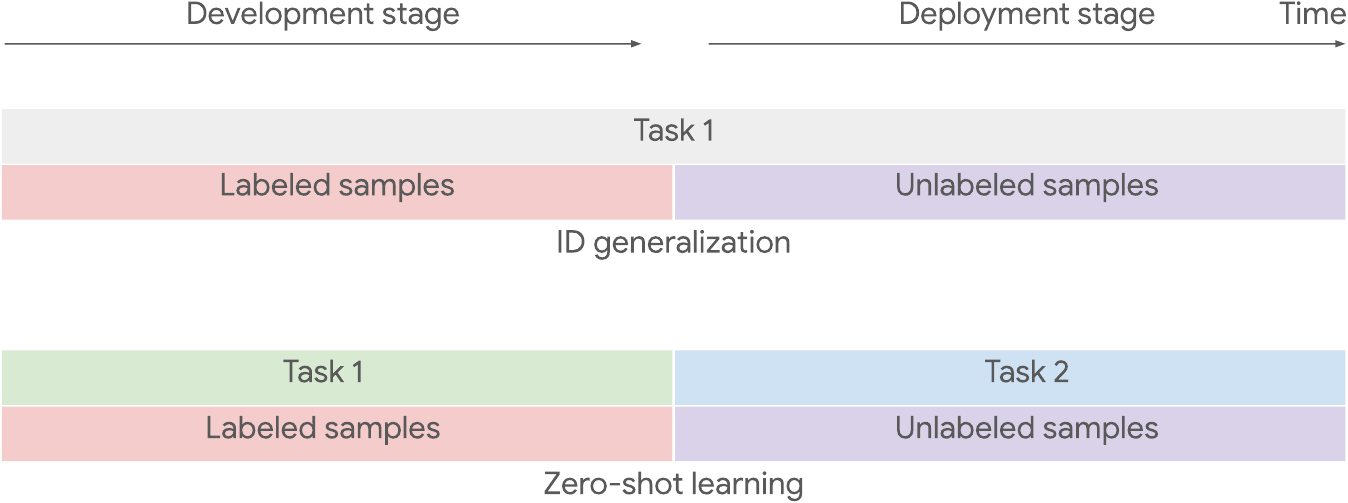}
    \caption{Comparison of ID generalization and zero-shot learning. Zero-shot learning aims to tackle a novel, unseen task in the deployment stage.}
    \label{fig:zeroshot}
\end{figure}

So far, the task stayed the same across development and deployment. However, the task can also change over time. The best-known scenario of this is \emph{zero-shot learning}, which is compared to ID generalization in Figure~\ref{fig:zeroshot}. In ID generalization, the task stays the same. In zero-shot learning, we have a different task for deployment about which we have no information in dev. 

\subsubsection{Large Language Models and Zero-Shot Learning}

Large Language Models (LLMs) are capable of performing zero-shot learning~\cite{https://doi.org/10.48550/arxiv.2205.11916} (called zero-shot-CoT prompting). They can encode the task description in natural language, so there is sufficient information for the model to solve the problem in principle. It is, however, still fascinating how LLMs can figure out how to solve new kinds of tasks not presented before that are an output of human creativity.

Nevertheless, we almost never have any guarantees about benchmarks truly being zero shot for LLMs -- their datasets are \emph{huge}, and we can never be certain that the model did not have the same task in its training dataset. \href{https://openai.com/research/clip}{CLIP} also has zero-shot learning capabilities.

\subsubsection{Categorizing according to which tasks are available at the development and deployment stages}
\begin{itemize}
    \item In \textit{ID generalization}, the task stays the same in dev and deployment.
    \item In \textit{zero-shot learning}, during deployment, we are faced with a new task not present in dev.
    \item \textit{\(K\)-shot learning} gives a ``softened'' setting where we have \(K\) labeled samples of the deployment task in dev.
    \item \textit{Meta-learning} has different tasks available during dev. This can also be combined with \(K\)-shot learning.
\end{itemize}

\subsection{\(K\)-Shot Learning}

\begin{figure}
    \centering
    \includegraphics[width=\linewidth]{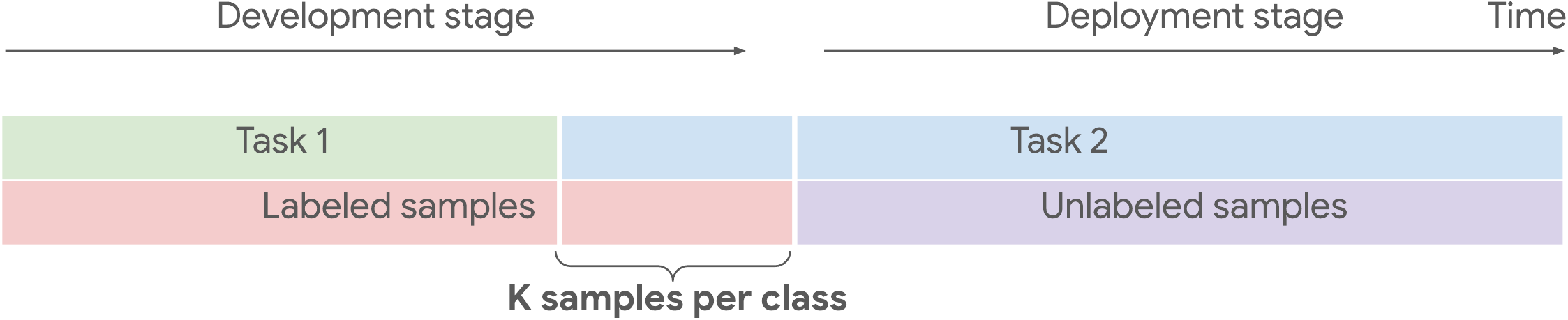}
    \caption{\(K\)-shot (few-shot) learning setting. \(K\) samples are available from task 2 during the development stage to aid the model towards robust generalization.}
    \label{fig:kshot}
\end{figure}

K-shot learning is illustrated in Figure~\ref{fig:kshot}. People try to make zero-shot learning easier by introducing some labeled samples for the target task during development. The \(K\) samples per class are for the target task. We learn to fit our model to the deployment task using a large number of task 1 samples and a few (\(K \times \#\text{class}\)) task 2 samples.

\textbf{Example 1}: ImageNet pretraining followed by fine-tuning on a downstream task.

\textbf{Example 2}: Linear probing in Self-Supervised Learning (SSL). Here, we do not even need labeled samples for domain 1. We train a strong feature representation in a self-supervised fashion, then we apply a linear classifier on the learned features and fine-tune the model with labeled task 2 data.

\subsection{Meta-Learning + \(K\)-Shot Learning}

\begin{figure}
    \centering
    \includegraphics[width=\linewidth]{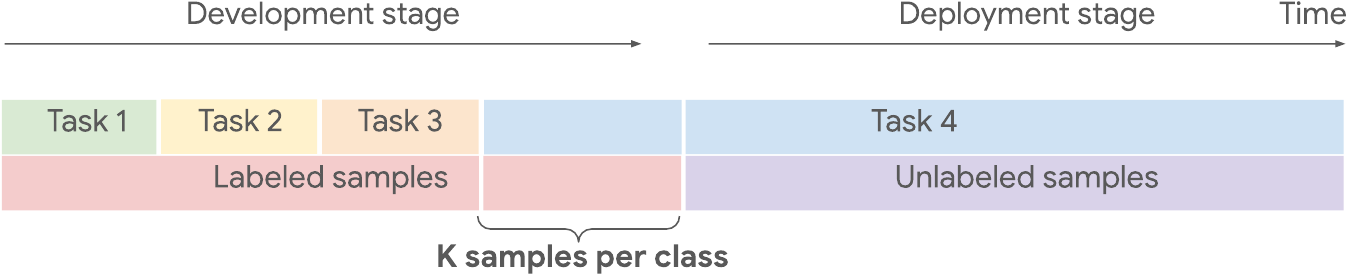}
    \caption{Meta-learning + \(K\)-shot learning setting. Multiple (proxy) tasks are available in the development stage. We further have access to \(K\) samples per class from the deployment stage task.}
    \label{fig:metakshot}
\end{figure}

Meta-learning can be combined with \(K\)-shot learning, as shown in Figure~\ref{fig:metakshot}. In this case, we have multiple tasks during development, and we wish to learn features that generalize across tasks, but we still need samples from the target task. 
In essence, we ``learn to learn a new task'' with tasks 1-3 (that give rise to a compound task). We then adapt our model to the deployment task 4, using the \(K\) samples per class for task 4.\footnote{For task changes, we also need to change the output head in the parametric case (e.g., linear probing). It is not needed for the non-parametric case (kNN) and CLIP~\cite{https://doi.org/10.48550/arxiv.2103.00020}. In CLIP, we need no information about the exact target task (zero-shot learning), but we need an LLM. The information comes from large-scale pretraining.}

\subsection{Task-Incremental Continual Learning}

\begin{figure}
    \centering
    \includegraphics[width=\linewidth]{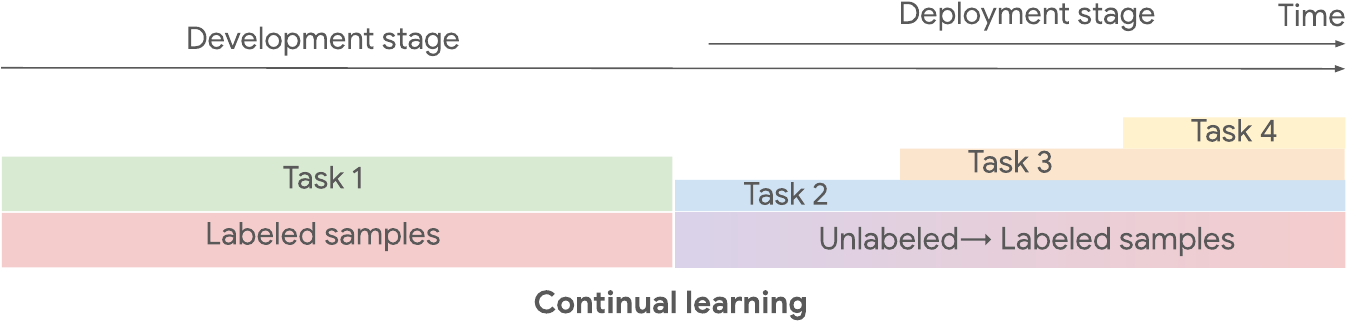}
    \caption{Task-incremental continual learning setting. The development stage continues in deployment, adding new tasks over time.}
    \label{fig:clti}
\end{figure}

Here, we consider a task-incremental version of continual learning, which is illustrated in Figure~\ref{fig:clti}. Tasks are added over time. We label only a few samples over time. (We can only utilize these.)\footnote{The output layer is always switched for the task accordingly. Sometimes very shallow output heads are enough (e.g., linear probing) if we have a strong backbone feature representation.} We update our model on the way. Ideally, the model should not forget the previous task.

\section{ML Dev as a Closed System of Information}

\begin{figure}
    \centering
    \includegraphics[width=\linewidth]{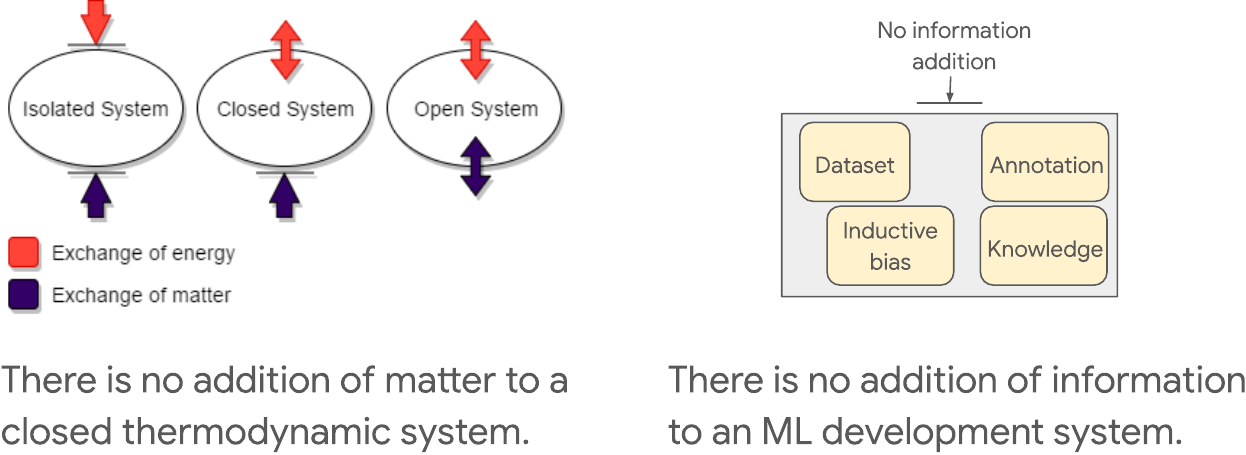}
    \caption{Comparison between a closed thermodynamic system~\cite{wikithermo} (left) and an ML development system (right).}
    \label{fig:thermo}
\end{figure}

To better illustrate the flow of information in the ML development stage, we draw a parallel between it and a closed thermodynamic system. This is illustrated in Figure~\ref{fig:thermo}.

Here, dev is represented as a closed system that consists of four main parts: \emph{dataset}, \emph{annotation}, \emph{inductive bias} and \emph{knowledge}. Inductive bias can appear in the form of the model architecture or the way we pre-process the data. A good example of knowledge is the expertise of people with much experience in training neural networks (NNs). 

In a closed lab environment where no dataset, annotation, or anything else is given to the system, there should be no additional information that suddenly appears. We should not expect new information to be born out of this system. Equivalently: There is no change in the maximal generalization performance we can get out of this system. There are lots of papers that \emph{violate} this principle (see \ref{sssec:leakage})~\cite{DBLP:journals/corr/abs-2007-01434,DBLP:journals/corr/abs-2007-02454}.

Note that it is possible to \emph{kill} information by, e.g., averaging things or replacing measurements with summary statistics.

\subsection{Information Leakage from Deployment}

\begin{definition}{Information leakage}
\emph{Information leakage} refers to the situation where information intended exclusively for the deployment stage becomes accessible during the development stage. It is an influx of information into a closed system.
\end{definition}

\begin{figure}
    \centering
    \includegraphics[width=\linewidth]{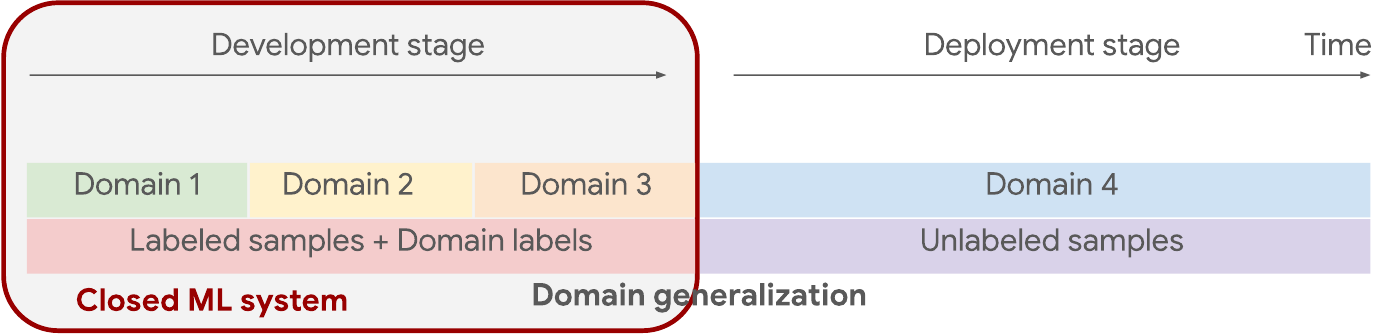}
    \caption{Closed ML system in the development stage for the domain generalization setting. Information about domain 4 must not be available in this closed system.}
    \label{fig:closedsystem}
\end{figure}

Let us consider information leakage in the domain generalization setting, illustrated in Figure~\ref{fig:closedsystem}. When one defines domain generalization as in Section~\ref{ssec:domain}, information about domain 4 must not be available during development. That is, we cannot inject new information into this closed ML system that comprises the resources \emph{at dev stage}. If we \emph{do} inject new information, we have to treat it as a new setting: When information about domain 4 is available, we cannot call it a domain generalization setup anymore. This also means we cannot compare against previous domain generalization methods. We need to set up a new setting, build a new benchmark, and compare against methods with the same setting.\footnote{Reducing the amount of information gained from evaluation helps in not spoiling the test set too much. For example, we might use a hidden server for benchmarking where only the ranking of submissions is shown but not the exact results.}

\subsubsection{Examples of Information Leakage in the Domain Generalization Setting}

Consider the domain generalization setting from~\ref{ssec:domain}. There are several ways how information leakage can surface and spoil our results.

\textbf{Scenario 1.} Some hyperparameters are chosen based on labeled samples from domain 4. In a sense, our dev set is partly taken from domain 4. We cannot talk about true generalization.

\textbf{Scenario 2.} Some hyperparameters are chosen by visually inspecting domain 4. This is still information leakage, just in a less automated way.

\textbf{Scenario 3.} The model is trained on labeled samples from domains 1-3 and unlabeled samples from domain 4. In the particular case of only two domains -- domain 1 in the development stage and domain 2 in the deployment stage -- and labeled or unlabeled samples being used from domain 2, we are performing domain adaptation, not domain generalization.

\textbf{Scenario 4.} Some hyperparameters are chosen to maximize publicly available scores after evaluation on some benchmarks with domain 4. Strictly speaking, these scores contain information about domain 4. One way to overcome this information leakage is to provide only a ranking of methods but not the scores.

\subsubsection{Information Leakage from Domain Generalization Evaluation}

Let us consider the particular problem of evaluation domain generalization methods in a bit more detail. It makes perfect sense to have a few labeled samples from deployment because we have to evaluate our system on a new domain anyway. Still, strictly speaking, as soon as we evaluate our model on the new domain, we \emph{use} our target domain (domain 4), so we cannot talk about generalization. We may need to shift the definition of domain generalization into something that allows some validation in the target domain (like domain adaptation). \textbf{Evaluating or benchmarking domain generalization is, therefore, contradictory.} Researchers still evaluate their methods on domain generalization benchmarks (observing the test set corresponding to the deployment domain multiple times through their lifetime), as we need to monitor progress somehow. 

\subsubsection{Information Leakage from Pretraining}
\label{sssec:test}

Another interesting question arises if we consider pretrained models. As soon as there is a pretrained system in dev resources, we introduce not just a single model but the entire pretraining dataset (which is gigantic in the case of large language models (LLMs) or CLIP~\cite{https://doi.org/10.48550/arxiv.2103.00020}). Much expertise is put into the dev scenario, which is also a dev resource. The consequence is that it is hard to say that anything is zero-shot learning in such settings. For example, if we give a new task to an LLM, we can never be sure that the model has never seen that task during training. The use of ImageNet-1K pretraining for zero-shot learning is also criticized: the 1k classes contain much information, and for certain classes we evaluate our model on, we do not have true zero-shot learning at all.

\subsubsection{When is information leakage not a problem?}

The other side of the argument about the severity of information leakage is that it might not always matter whether something is zero-shot learning. If an LLM contains all information about the world, then there is nothing new in the deployment stage, so we cannot have true zero-shot learning. Nevertheless, the model works very well, so we can make good use of it in many real-world scenarios. Take the example of face recognition. Suppose there is a person our system has never seen before. We want it to recognize (or verify) that this is the same person on subsequent days. This setup is zero-shot verification. However, as soon as our training set contains billions of identities, this does not matter anymore. In the most extreme case of having seen all people in the world, we do not have to generalize to unseen people. Nevertheless, we are still happy with the system if it works well for everyone. Zero-shot learning thus becomes less meaningful at a large scale.

\subsection{A Case Study on Information Leakage}
\label{sssec:leakage}

We consider an example of a paper that is leaking information from deployment, titled ``\href{https://arxiv.org/abs/2007.02454}{Self-Challenging Improves Cross-Domain Generalization}''~\cite{DBLP:journals/corr/abs-2007-02454}. It is straightforward to find such papers, even from highly regarded research groups.

\begin{definition}{Ablation Study}
We are changing one factor at a time in our method, as we want to see the contribution of each factor towards the final performance. Everything else is kept fixed. Then we can understand the effect of the factor better, and we can also optimize that factor (hyperparameter) separately.
\end{definition}

\begin{table}
\centering
\caption{Benchmark results of various feature drop strategies. Explanation of columns: e.g., for the art painting column, we train the model on \{cartoon, sketch, photo\} and test it on art painting. Table taken from~\cite{DBLP:journals/corr/abs-2007-02454}.}
\label{tab:reftab}
\begin{tabular}{c c c c c c c}
\toprule
\bfseries Feature Drop Strategies & \bfseries backbone & \bfseries artpaint & \bfseries cartoon & \bfseries sketch & \bfseries photo & \bfseries Avg $\boldsymbol{\uparrow}$ \\ [0.5ex]
\midrule
 Baseline & ResNet18  & 78.96 & 73.93  & 70.59 & \textbf{96.28} & 79.94 \\
 Random & ResNet18  & 79.32 & 75.27 & 74.06 & 95.54 & 81.05\\
 Top-Activation& ResNet18  & 80.31 & 76.05 & 76.13 & \textbf{95.72}&  82.03\\
 Top-Gradient& ResNet18  & \textbf{81.23} & \textbf{77.23} & \textbf{77.56} & 95.61& \textbf{82.91} \\
\bottomrule
\end{tabular}
\end{table}

In Tables 1-5 of~\cite{DBLP:journals/corr/abs-2007-02454}, an ablation study is conducted. We show Table 1 of the paper in Table~\ref{tab:reftab} for convenience. We see various hyperparameters chosen based on the performance on the domain they want to generalize to. (For example, the ``Feature Drop Strategy'' hyperparameter considers different ways to drop features to make the model better regularized.) They are looking at the generalization performance to each of the domains using leave-one-out domain generalization. They finally choose the hyperparameters based on the average accuracy on the left-out domains. If we also validate on the test set, we cannot talk about domain generalization anymore, as we have information leakage. (Even if we consider the academic point of view of the test set belonging to deployment.)

This hyperparameter configuration will be pretty good for the PACS dataset~\cite{DBLP:journals/corr/abs-1710-03077} (see below). However, this does not guarantee that this is the best ingredient for non-PACS cases. We might overfit to PACS severely by making such choices. (Of course, this overfitting can also happen even if we do not use the test set as a part of the validation set, but rather as a criterion for method selection across different papers. However, that is a much less severe case of overfitting. Here, the authors make use of the test set many times in a \emph{single} paper.)

\textbf{Takeaway}: Ablation studies are generally great for ID generalization tasks, but one should be very careful with ablation studies for OOD generalization.

\subsection{Solutions to Information Leakage}

There are many (partial) solutions to combat information leakage, discussed below.

\textbf{Select hyperparameters within dev resources.} Section 3 of ``\href{https://arxiv.org/abs/2007.01434}{In Search of Lost Domain Generalization}''~\cite{DBLP:journals/corr/abs-2007-01434} discusses information leakage and provides possible solutions for it. Selecting hyperparameters, design choices, checkpoints, and other parts of the system must be a part of the learning problem (i.e., part of the ML dev system). When we propose a new domain generalization algorithm, we must specify a method for selecting the hyperparameters rather than relying on an unclear methodology that invites potential information leakage.

\textbf{Use the test set once per project.} By using a specific test set multiple times, we can always overfit to it. If our goal is to go towards a distribution outside dev, then evaluating on the test set multiple times can be harmful. However, as discussed previously, we \emph{do} have to use it multiple times to compare methods and evaluate our approach. Solution: At least do not use the test set for hyperparameter tuning; tune them on the validation set. For example, if we want to generalize well to the art painting dataset of PACS, tune the hyperparameters on \{cartoon, sketch, photo\}. Then we measure performance on the art painting test set. We will do the same thing for a new, genuinely unknown domain in deployment: find the hyperparameters on the known domains. Thus, one should use the test set sparingly. A good rule of thumb might be to use it once per paper. This way, we are less likely to overfit to it. (The State-of-the-Art (SotA) architectures are also likely to overfit to standard benchmarks, e.g., to ImageNet-1K.~\cite{https://doi.org/10.48550/arxiv.1902.10811})

\textbf{Update benchmarks.} Even if the test set labels are unknown, we can overfit to the test set just based on the reported performance, e.g., on leaderboards. Thus, for many reasons, the test set has to be changed every once in a while. Another helpful idea is to use a non-fixed benchmark, where the data stream changes over time. For comparability, this is an issue: we have a continuously changing target over time. However, it is usually not problematic: In human studies, researchers have been dealing with a changing evaluation set. By using statistical tests, they could always argue about statistical significance. One particular example is the case of \emph{clinical trials}. It is physically impossible to test two related drugs on exactly the same set of people: The test set changes from experiment to experiment. However, statistical tests give a principled way to determine if the observed changes are significant or if they could have happened by chance. 

\textbf{Modify evaluation methods.} A different approach is to use a differential-privacy-based evaluation method. This method adds a Laplace noise to the accuracy before reporting it to the practitioner. This is better than overfitting to a single benchmark.

\textbf{Summary}: We are trying to address an impossible problem: to truly generalize to new domains, which requires them to be previously unseen. We can never keep them completely unseen, as then we cannot \emph{measure} generalizability. However, as soon as we measure generalizability, we cannot talk about generalization anymore. This is an unavoidable dilemma for many ML fields, even more so for Trustworthy Machine Learning (TML), as they deal with more challenging cases of generalization where evaluation is very tricky.

\section{Domain Generalization Benchmarks}

\begin{figure}
    \centering
    \includegraphics[width=0.5\linewidth]{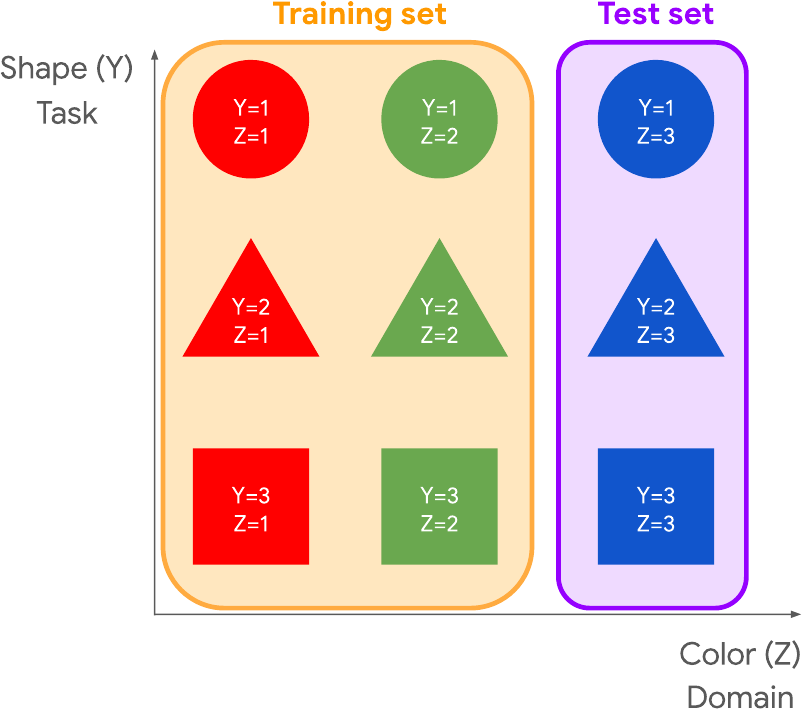}
    \caption{Training (yellow) and test (blue) datasets in the domain generalization setting. Shape is the task, color is the domain.}
    \label{fig:dg2}
\end{figure}

\begin{definition}{Subpopulation Shift Benchmarks}
In subpopulation shift benchmarks, we consider test distributions that are subpopulations of the training distribution and seek to perform well even on the \emph{worst-case} subpopulation.
\end{definition}

In the previous section, we highlighted the importance of paying special attention to benchmarks in the domain generalization setting. Now, let us take a closer look and discuss some prominent examples in more detail. Some of these benchmarks will be related to the problem of subpopulation shift which is partially connected to domain generalization.

\subsection{Examples of Domain Generalization Benchmarks}

A toy domain generalization problem is shown in Figure~\ref{fig:dg2}. Our \textit{goal} is to generalize well to blue images: This is a particular instance of cross-domain generalization. The \textit{inputs} are images with mono-colored objects of some shape. The \textit{labels} are $\{0, 1, 2\}$ -- we have a three-way classification problem. The set of classes is shared across the domains and is assigned according to the object's \textit{shape} (circle, triangle, or square). The model's \textit{task} is to predict the label of a given input. Here, we consider three domains: \red{red}, \green{green}, and \blue{blue} colored objects. This is not a real problem but we refer to it to illustrate the scheme shared among the subsequent benchmarks.

\begin{figure}
    \centering
    \includegraphics[width=0.6\linewidth]{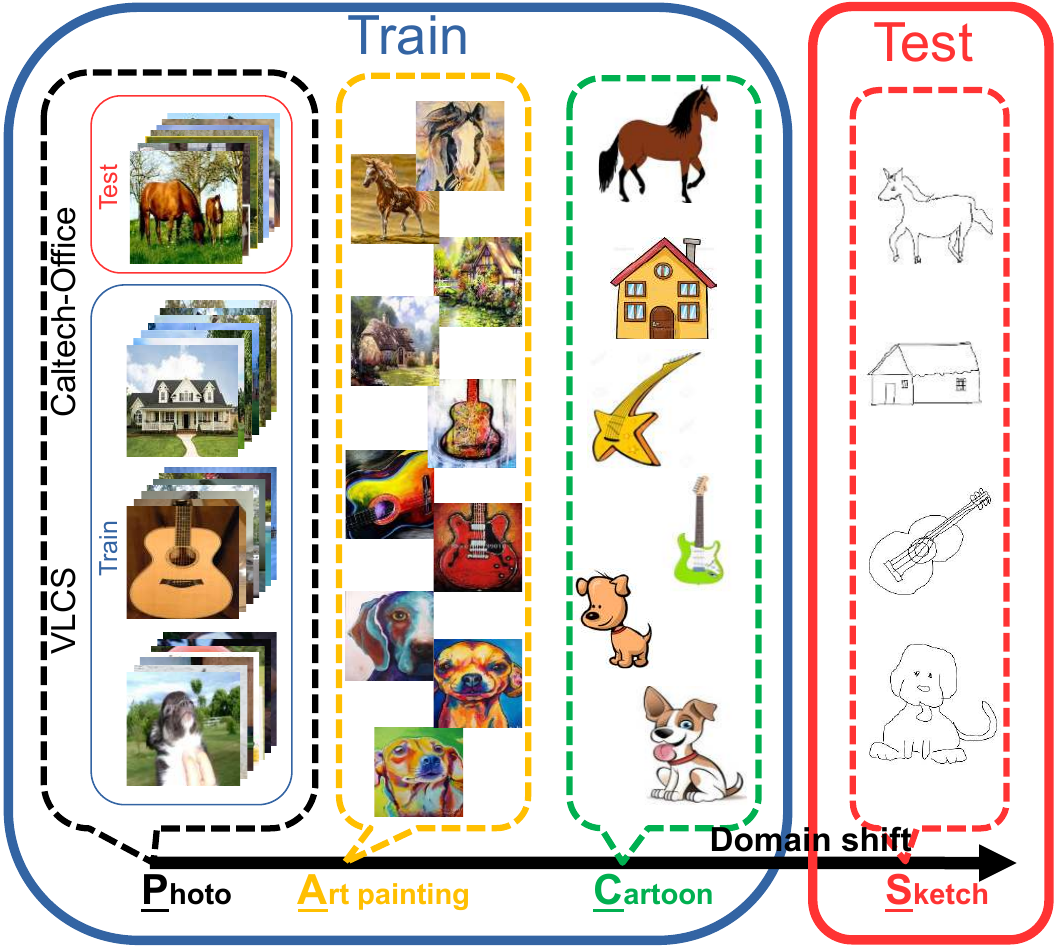}
    \caption{The PACS dataset can be used for domain generalization. Figure taken from~\cite{https://doi.org/10.48550/arxiv.1710.03077}.}
    \label{fig:pacs}
\end{figure}

\subsubsection{The PACS Dataset}

The PACS dataset considers four domains: \textbf{P}hotos, \textbf{A}rt Paintings, \textbf{C}artoons, and \textbf{S}ketches. Samples from each domain are shown in Figure~\ref{fig:pacs}. The set of classes is shared across the domains. To benchmark domain generalization, leave-one-out evaluation is used. For example, one might train on \textbf{PAC} and test on \textbf{S}.

\subsubsection{DomainBed}

\begin{figure}
    \centering
    \includegraphics[width=0.6\linewidth]{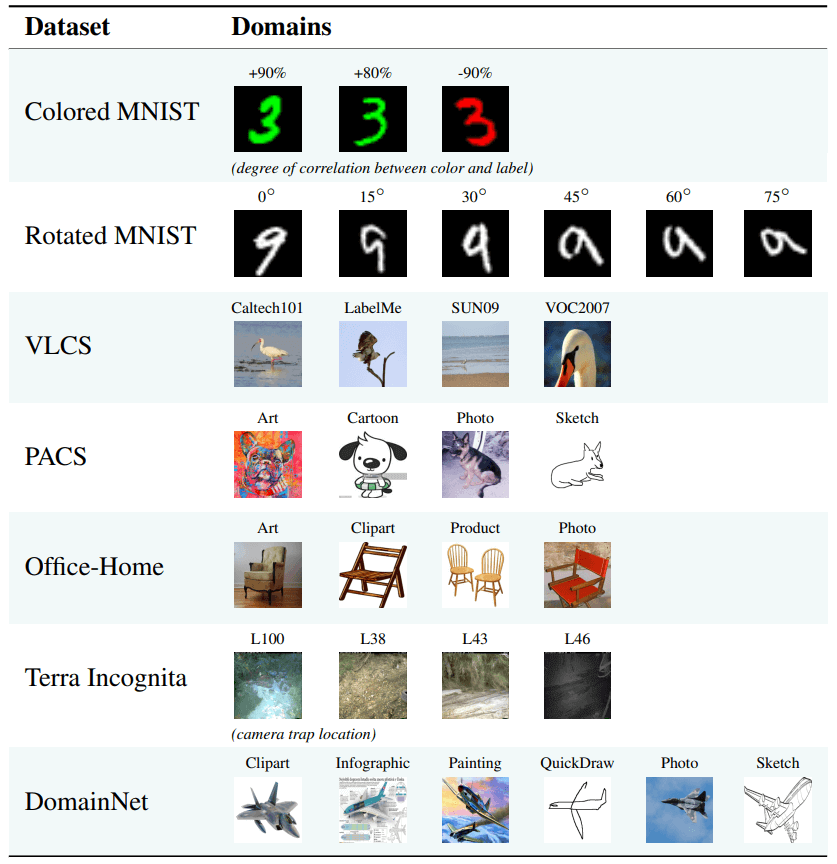}
    \caption{The DomainBed suite. Illustration taken from~\cite{https://doi.org/10.48550/arxiv.2007.01434}.}
    \label{fig:domainbed}
\end{figure}

DomainBed is a combination of some popular domain generalization benchmarks into a single suite. It subsumes, e.g., PACS, Colored MNIST, Rotated MNIST, and Office-Home. Office-Home contains \href{https://paperswithcode.com/dataset/office-home}{four domains}. These are (1) \textit{art} that contains artistic images in the form of sketches, paintings, ornamentation, and other styles; (2) \textit{clipart} that is a collection of clipart images; (3) \textit{product} that contains images of objects without a background; and (4) \textit{real-world} that collects images of objects captured with a regular camera. For each domain, the dataset contains images of 65 object categories found typically in Office and Home settings. Samples from each subsumed benchmark are shown in Figure~\ref{fig:domainbed}.

\subsubsection{The Wilds Benchmark}

\begin{figure}
    \centering
    \includegraphics[width=\linewidth]{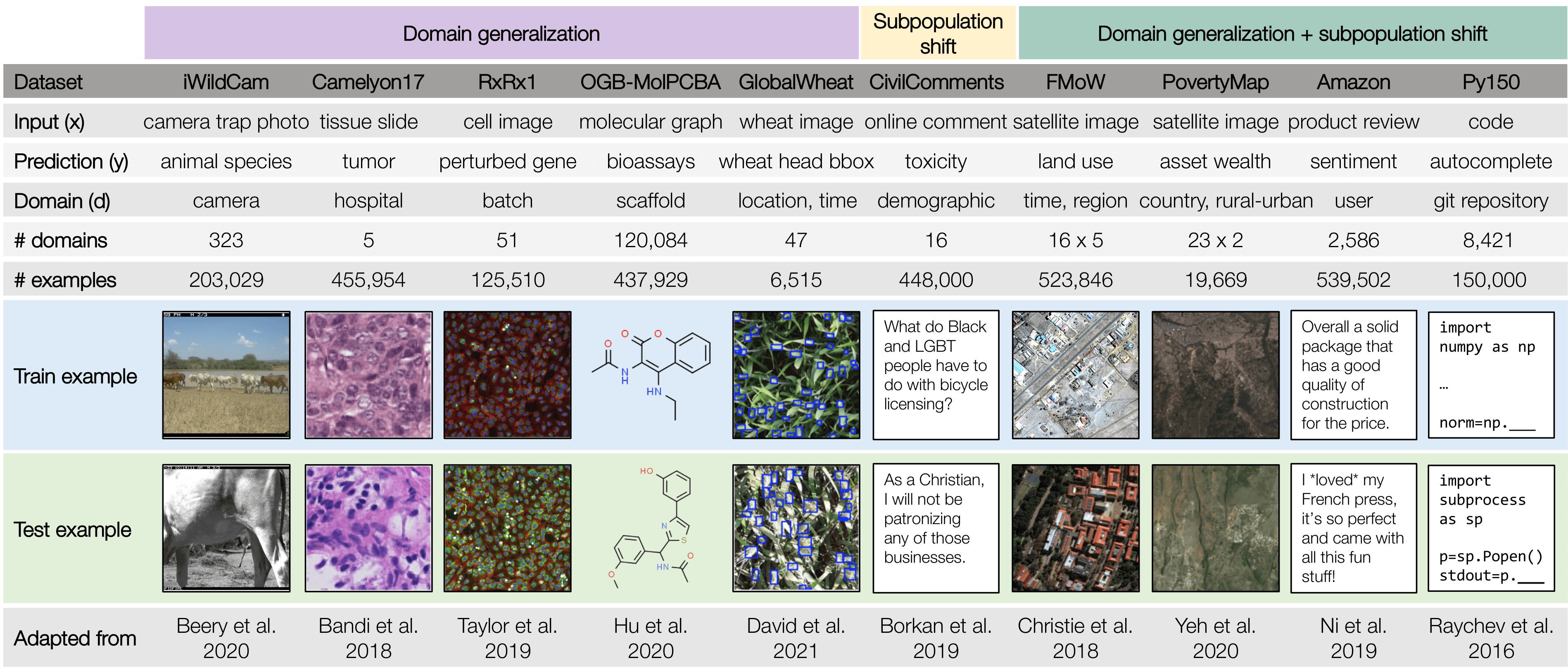}
    \caption{The Wilds suite. Figure taken from~\cite{stanfordai}.}
    \label{fig:wilds}
\end{figure}

The Wilds benchmark~\cite{pmlr-v139-koh21a} comprises several tasks and domains for each task. IT contains domain generalization benchmarks and also subpopulation shift benchmarks. A detailed illustration of the dataset is shown in Figure~\ref{fig:wilds}.

\subsubsection{ImageNet-C}

\begin{figure}
    \centering
    \includegraphics[width=0.6\linewidth]{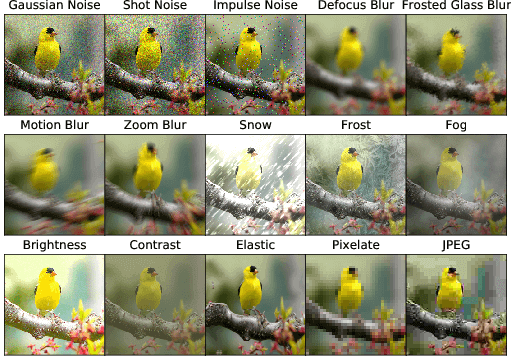}
    \caption{Illustration of the various corruptions ImageNet-C employs, taken from~\cite{hendrycksgithub}.}
    \label{fig:imgnetc}
\end{figure}

ImageNet-C~\cite{hendrycks2019robustness} is an extension to the ImageNet dataset~\cite{deng2009imagenet} with a focus on robustness. For the same image, the dataset contains various corruptions. Corruptions include Gaussian Noise, Defocus Blur, Frosted Glass Blur, Motion Blur, Zoom Blur, JPEG Encoding-Decoding, Brightness Change, and Contrast Change. Examples of these corruption types are shown in Figure~\ref{fig:imgnetc}. The ImageNet-C dataset consists of 75 corruptions, all applied to the ImageNet test set images. It simulates possible corruptions under the deployment scenario, thereby measuring the robustness of the model to the perturbation of the data generating process.

\subsubsection{ImageNet-A}

\begin{figure}
    \centering
    \includegraphics[width=0.4\linewidth]{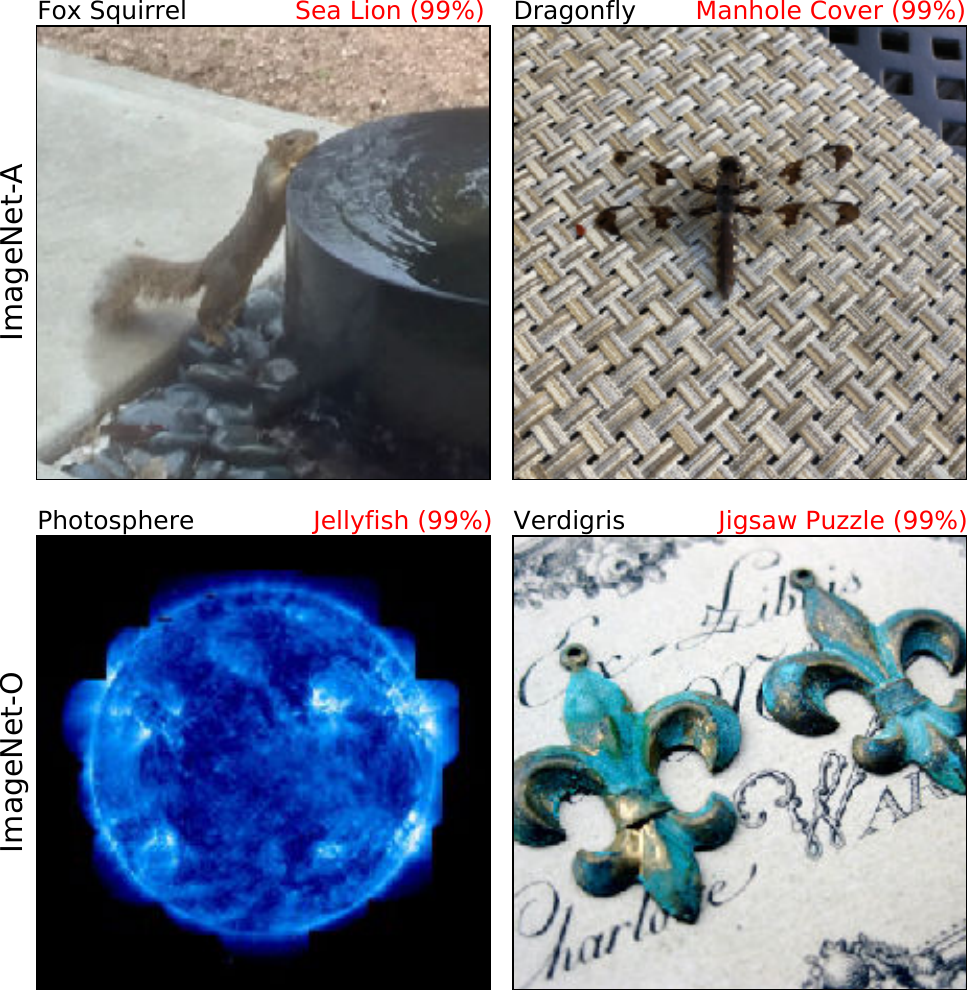}
    \caption{Sample from the ImageNet-A and ImageNet-O datasets, taken from~\cite{hendrycks2019nae}.}
    \label{fig:imgnetao}
\end{figure}

ImageNet-A~\cite{hendrycks2019nae} collects common failure cases of the PyTorch ResNet-50~\cite{https://doi.org/10.48550/arxiv.1512.03385} on ImageNet.\footnote{We explicitly mention the used implementation of ResNet-50 because there are \href{https://stackoverflow.com/questions/67365237/imagenet-pretrained-resnet50-backbones-are-different-between-pytorch-and-tensorf}{subtle differences} between versions.} It contains images that classifiers should be able to predict correctly but cannot. Examples from ImageNet-A are shown in Figure~\ref{fig:imgnetao}.

\subsubsection{ImageNet-O}

ImageNet-O is another extension to ImageNet that contains anomalies of unforeseen classes which should result in low-confidence predictions, as the true class labels are not ImageNet-1K labels. ImageNet-O examples are shown in Figure~\ref{fig:imgnetao}.

\section{Domain Generalization Difficulties}

We have discussed how easy it is to confuse a setting with domain generalization just by not being careful enough with how one uses information about the target distribution. For those who are ready to accept this difficulty, we would like to point out that there are even more complications with domain generalization. However, we hope that these difficulties will not be an obstacle but rather an invitation to challenge, which is why we gathered the most important ones in this section.

\subsection{Ill-Defined Behavior and Spurious Correlations}
\label{ssec:spurious}

\begin{figure}
    \centering
    \includegraphics[width=0.5\linewidth]{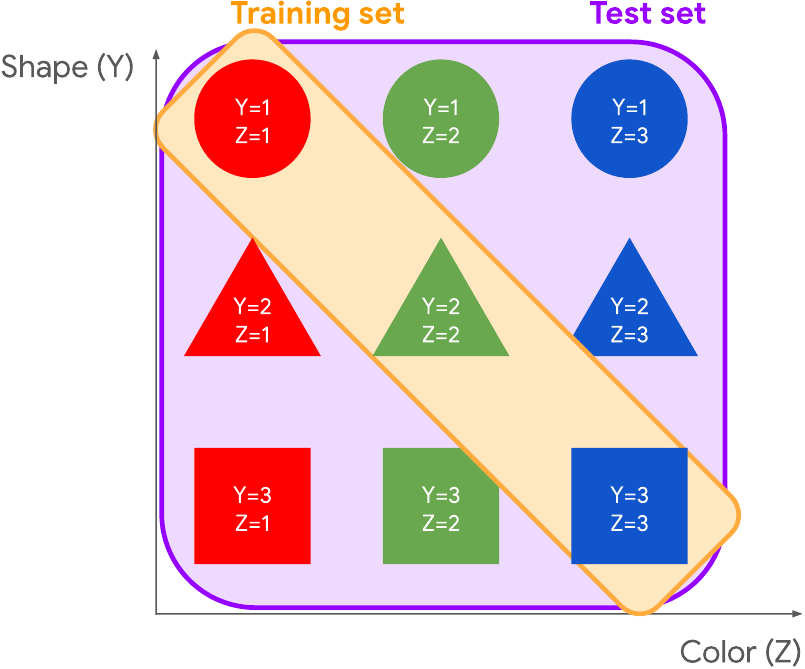}
    \caption{Diagonal training dataset and unbiased test set in cross-domain generalization. During training, the model is only exposed to samples where the shape and color labels coincide.}
    \label{fig:diagdomain}
\end{figure}

Consider Figure~\ref{fig:dg}. For this setting, we outline two main difficulties: the ill-defined behavior on novel domains and the spurious correlations between task labels and domain labels.

\subsubsection{Ill-Defined Behavior on Novel Domains}

The model does not know what to do in regions without any training data. One could ask how domain generalization is even possible. It works in practice, but there are no rigorous theories as to why. We take it at face value, without any guarantees of the model's behavior on novel domains. This problem can be addressed through calibrated epistemic uncertainty estimation (\ref{ssec:epi}) to make the model ``know when it does not know''.

\subsubsection{Spurious Correlations between Task Labels and Domain Labels}

\begin{definition}{Spurious correlation}
A spurious correlation is the co-occurrence of some cues, features, or labels, which happens in the development stage but not in the deployment stage.
\end{definition}

For example, our prediction of shape may depend a lot on the color. If we have a diagonal dataset, this can become a huge problem, as depicted in Figure~\ref{fig:diagdomain}. Here, we have a perfect correlation between the two cues in the training dataset. In other words, there are spurious correlations between \emph{task labels} and \emph{domain labels}. This results in an ill-defined behavior on novel domains. 

The problem of spurious correlations is also present in cross-bias generalization. We will consider this setting as it is easier than domain generalization, and ill-definedness is out of the picture.

\section{Cross-Bias Generalization}

We will now discuss cross-bias generalization from Table \ref{tab:gentypes} that has a particular focus on the problem of spurious correlations. As seen before, we can amplify the spurious correlation between domain (bias) and target label (task) for OOD generalization to arrive at a scenario like Figure~\ref{fig:diagdomain}. We also remove the issue with unseen attributes: a model is guaranteed to encounter each attribute (e.g., possible shapes, colors) at least once, but in a heavily correlated fashion.

\begin{figure}
    \centering
    \includegraphics[width=\linewidth]{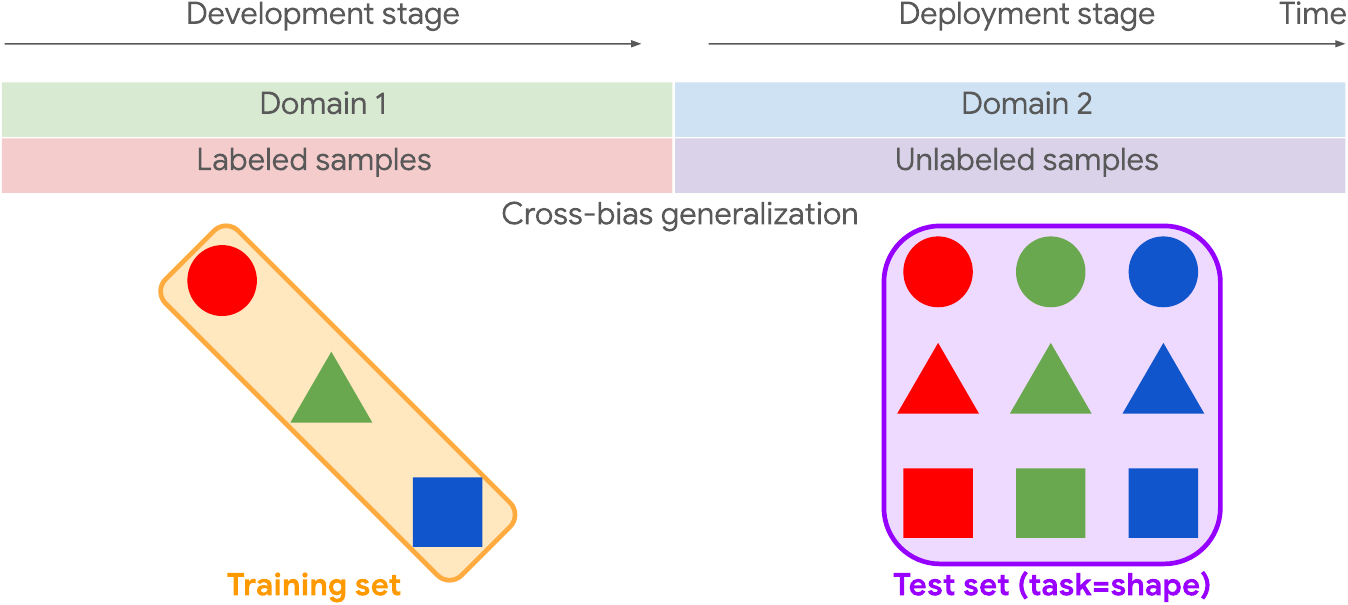}
    \caption{Cross-bias generalization setting with an unbiased deployment domain. In the deployment stage, the model has to do well on samples where the correlation between color and shape is broken.}
    \label{fig:cbg}
\end{figure}

This leads us to textbook cross-bias generalization, a cleaner setup for addressing the spurious correlations, for which an overview is given in Figure~\ref{fig:cbg}. In the test set, we have to recognize a diverse set of combinations of cues that we have not seen during training.

In general, the situation could be better described as ``We still have more dominance along the diagonal, but we have a requirement that every single subgroup (i.e., (color, shape) combination) has to have a similar level of performance.''. This formulation is roughly equivalent to having an equal number of samples in each grid cell. It is also possible that the deployment scenario is \emph{still biased}, just has a different bias. We impose no restrictions on the deployment distribution.

\begin{information}{Compositionality and Cross-Bias Generalization}
Cross-bias generalization has close ties to compositionality~\cite{andreas2019measuring,lake2014towards} that aims to disentangle semantically different parts of the input in the representation of neural networks. If a network leverages compositionality, i.e., treats semantically independent parts of the input independently when making a prediction, spurious correlations cannot arise by definition. This leads to robust cross-bias generalization. Of course, achieving this in practice is much more complicated.
\end{information}

\subsection{Why is cross-bias generalization still challenging?}

ID generalization is already an ill-posed problem. The No Free Lunch Theorem states that without extra inductive bias in the dev scenario, we cannot train a model that generalizes to the same distribution. We need inductive biases to find well-generalizing models ID. Without inductive biases, any model is equally likely to generalize well ID~\cite{wolpert1997no,mitchell1980need}.

OOD generalization (in particular, cross-bias generalization) poses another layer of difficulty: the \emph{ambiguity of cues}, discussed next. We need further information in the ML dev system to solve it.

\subsection{The Feature Selection Problem}

We mentioned that the ambiguity of cues brings an additional challenge to cross-bias generalization. We would like to formally define this ambiguity.

\begin{definition}[label=def:underspec]{Underspecification}
An ML setting is underspecified when multiple features (e.g., color, shape, scale) let us achieve 100\% accuracy on the training set. The training set does not specify what kind of cue the model should be looking at and how to generalize to new samples that do not have a perfect correlation. If the model chooses the incorrect cue, we say a \emph{misspecification} happens.

\medskip

\textbf{Note}: We assume a network with very high capacity that can get 100\% accuracy for every cue in the training set. For complex cues, the decision boundary tends to be wiggly, but under our assumption, even this decision boundary can be learned.
\end{definition}

Underspecification in the cross-bias generalization setting necessitates the selection of the suitable feature(s) for good generalization to the deployment scenario.

A model under the vanilla OOD (e.g., cross-bias) generalization setting with a diagonal dataset lacks the information to generalize to an arbitrary deployment task well. When predicting in the deployment scenario (considering an uncorrelated dataset), the model cannot simultaneously use all perfectly aligned cues on the training set, as they contradict each other. Any cue the model adopts from training could be correct; the answer depends on the deployment task (chosen by a human, e.g., they can choose the most challenging cue for the model), which is arbitrary out of the perfectly correlated cues.
\begin{itemize}
    \item If we have an adversarial deployment task selector, it can always fool the system into performing badly by choosing the most difficult cue for the model as the task.
\end{itemize}

\textbf{Without any knowledge about the deployment task, cross-bias generalization is not solvable with a diagonal training set.} Yet, it happens a lot that someone claims this in ML conference papers. They usually have a hidden ingredient that they implicitly assume. This is a prime example of \emph{information leakage}.

\begin{figure}
    \centering
     \includegraphics[width=0.8\linewidth]{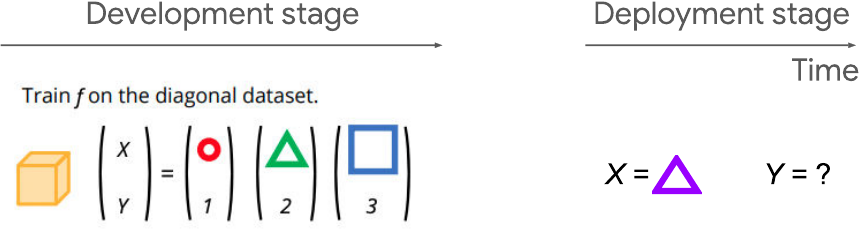}
    \caption{Underspecification in a more general toy OOD setting than cross-bias generalization. We are faced with the same problem: We now know that color is not the task, but shape and size can still be tasks. Figure inspired by~\cite{https://doi.org/10.48550/arxiv.2110.03095}.}
    \label{fig:underspec}
\end{figure}

To select the right feature for the task, more information is needed. This also holds for more general OOD settings: an example is shown in Figure~\ref{fig:underspec}.

\subsubsection{The Feature Selection Problem in Fairness}

The feature selection problem is also closely connected to the problem of fairness. What is fairness? From the viewpoint of the equality of opportunity as a notion of individual fairness: people who are similar \wrt a task should be \emph{treated} similarly. There can be attributes for individuals that are relevant to the task and attributes that are supposed to be irrelevant, e.g., demographic details, such as race or gender. We want the model to only look at relevant features (task cue), not sensitive/prohibited attributes (bias cue). This notion of fairness is comparative: We are determining whether there are differences in how similar people (according to the task cues) are treated.

Decision-makers should automatically avoid differential treatment according to people's race, gender, or other possibly discriminatory factors if we accept in advance that none of these characteristics can be relevant to the task at hand.

\subsection{Extra Information to Make Cross-Bias Generalization Possible}

As we have seen, without extra information, cross-bias generalization is not solvable. We suggest considering a simplified generalization setting where such information is available in the development stage.
This is much less exciting than true generalization, but we need this simplification to make the problem feasible. We will consider two ways to add extra information to the setting that makes the problem well-posed.

\begin{figure}
    \centering
    \includegraphics[width=\linewidth]{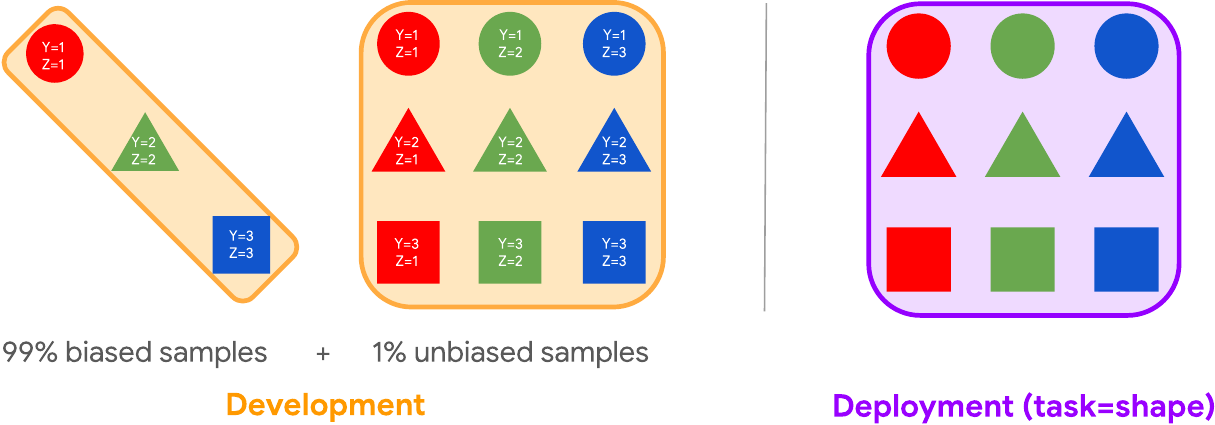}
    \caption{New setting that makes cross-bias generalization possible, referred to as the ``First way'' in the text. We have a few unbiased samples in the development resources and bias labels are also available.}
    \label{fig:first}
\end{figure}

\subsubsection{First way to make cross-bias generalization feasible: adding unbiased samples}

This approach is illustrated in Figure~\ref{fig:first}. A small number of non-correlated samples are added to the dev resources (these samples are not necessarily deployment samples). We have attribute (\(Z\)s -- here bias, but it could also be domain) labels for each sample that specify which bias category a sample corresponds to. For example, \(Z\)s can correspond to different jackets. It is useful to explicitly tell the model what \emph{not to} use as cues (see DANN in Section~\ref{sssec:dann}) in the form of bias labels.

People control the percentage of unbiased samples using \(\rho \in [0, 1]\) in papers. We have to know what \(\rho\) they are using; it is a part of the setting. The lower the percentage of unbiased samples, the harder the task becomes. The task can be made arbitrarily hard, up to the point that it is impossible again (\(\rho = 0\)). The test set is unbiased in this example. However, in the deployment domain, we might just as well have biased samples that are biased in a different way than the dev samples.

\subsubsection{A word about domain generalization}

As we discussed in \ref{ssec:domain}, \emph{domain generalization} is supplying additional information by providing domain labels. However, if we simply treat the bias labels (color) as our domain labels for domain generalization, we are sadly still not able to solve the problem: for such a diagonal dataset, the task labels are the same as the bias labels. Under this interpretation, domain generalization does not directly make the problem solvable. That is why we still need access to unbiased samples. In this case, we treat the domain labels as `unbiased' and `biased', and the problem is solvable again. (This is very similar to the first way, only the interpretation is different.)

\subsubsection{Second way to make cross-bias generalization feasible: converting the problem to domain adaptation/test-time training}

We can also consider \emph{domain adaptation}. Here, the source of extra information is access to the target distribution. This is different from before when we only had unbiased samples that did not necessarily come from the target domain. By performing domain adaptation, we make the target distribution more accessible to the dev stage. Here, we assume labeled samples from the target domain, and knowledge about the domain of each sample.

The same logic applies if we convert the problem to \emph{test-time training}. The only difference is that in test-time training, the target distribution changes continuously during deployment, therefore, we constantly adapt our model to new situations.

\subsection{How to determine what cue our model learns to recognize?}

To understand how well we solved the problem of cross-bias generalization or to gain insights into the model's inner workings, it is often helpful to understand which cue our model uses for predictions. However, answering this question is not straightforward in general. 

To diagnose our model, we require labels for different cues (e.g., labels \(Y\) and \(Z\) from Figure~\ref{fig:first}).
In that case, after training on our close-to-diagonal dataset, we label unbiased (off-diagonal) samples from a test set according to different cues and calculate the model's accuracy \wrt each labeling scheme on this unbiased test set. The model should achieve high accuracy for the cue it learned on the training dataset and perform close to random guessing for all other cues.

\section{Shortcut (Simplicity) Bias}
\label{ssec:simplicity}

We have seen that due to underspecification (Definition~\ref{def:underspec}), models can learn different equally plausible cues. But do models prioritize learning one cue over others? It turns out the answer is yes, simpler cues are learned first. This property is usually called \emph{shortcut/simplicity bias}, defined below.

\begin{definition}{Shortcut Bias/Simplicity Bias}
The shortcut bias is the ML models' inborn preference for ``simpler'' cues (features) over ``complex'' ones.

When there are multiple candidates of cues for the model to choose from for achieving 100\% accuracy (i.e., the setting is underspecified), the model chooses the \emph{easier} cue.
\end{definition}

\subsection{Examples of Shortcut Bias}

Let us first define the \emph{Kolmogorov complexity}, which is needed for the details of the first example.

\begin{information}{Kolmogorov Complexity}
The Kolmogorov complexity measures the complexity of strings (or objects in general) based on the minimal length among programs that generate that string.

\medskip

Kolmogorov Complexity of a cue \(p_{Y \mid X}\) (KCC)~\cite{https://doi.org/10.48550/arxiv.2110.03095}:
\[K(p_{Y \mid X}) = \min_{f:\cL(f; X, Y) < \delta} K(f) \qquad \delta > 0, f: X \rightarrow Y.\]
Intuitively, \(K(p_{Y \mid X})\) measures the \emph{minimal} complexity of the function \(f\) required to memorize the labeling \(p_{Y \mid X}\) on the training set (i.e., \(\cL < \delta\)).
\end{information}

As a toy example, according to the paper ``\href{https://arxiv.org/abs/2110.03095}{Which Shortcut Cues Will DNNs Choose? A Study from the Parameter-Space Perspective}''~\cite{https://doi.org/10.48550/arxiv.2110.03095}, Color \(>\) Scale \(>\) Shape \(>\) Orientation in the order of models' preference, regardless of the network architecture and the training algorithm. Why could this be? The reason, according to the authors, is that color is a simpler cue than the others, as measured by the Kolmogorov complexity of the cues. The authors approximate \(K(f)\) by the minimal number of parameters of model \(f\) to memorize the training set with labels \wrt the cue in question.

To better illustrate what simplicity bias is, we provide several examples below. An overview is shown in Table~\ref{tab:overview}, which is further detailed in the individual sections.

\begin{table}
\centering
\caption{Overview of bias types and corresponding cues.}
\label{tab:overview}
\begin{tabular}{p{3.3cm}p{3cm}p{3.3cm}p{3.5cm}}
\toprule
\textbf{Problem} & \textbf{Task} & \textbf{Bias Cue} & \textbf{Task Cue} \\
\midrule
Context bias & Classify object & Background context & Foreground object(s) \\
\midrule
Texture bias & Classify object & Texture of object & Shape of object \\
\midrule
Not understanding sentence structure & Natural language inference & Set of words in a sentence, lexical overlap cue, subsequence cue, constituent cue & The entire sentence \\
\midrule
Biased action recognition & Recognize action that human is performing & Scene, instrument, static frames & Human movement \\
\midrule
Using single modality for multi-modal tasks & Visual question answering & Question only & Question and image \\
\midrule
Use of sensitive attributes & Predict possibility of future defaults & Sensitive attributes (disability, gender, ethnicity, religion, etc.) & Size of the loans, history of repayment, income level, age, etc. \\
\bottomrule
\end{tabular}
\end{table}

\subsubsection{Context Bias}

\begin{figure}
    \centering
    \includegraphics[width=0.4\linewidth]{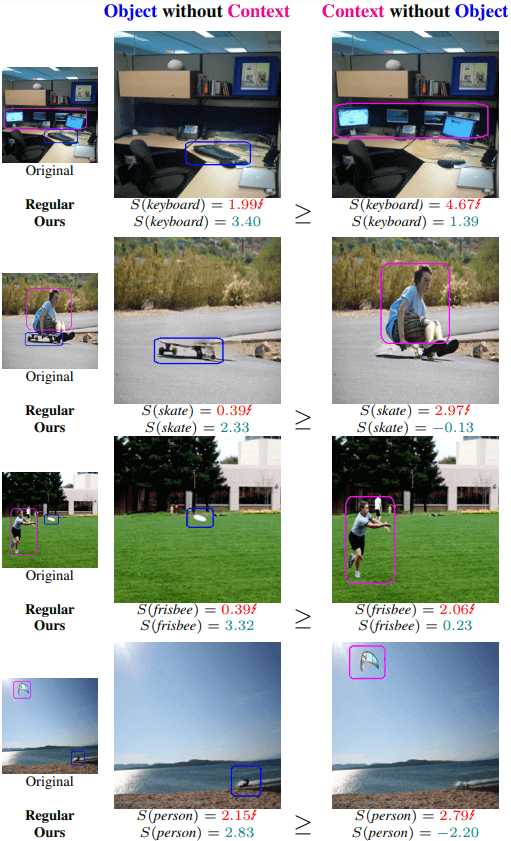}
    \caption{Context bias can arise in various settings. Figure taken from~\cite{https://doi.org/10.48550/arxiv.1812.06707}.}
    \label{fig:context}
\end{figure}

Consider the task of object classification. The task cues are the foreground objects, but a classifier focusing on the background context bias cues can achieve high accuracy when the background is highly correlated with the foreground. The examples, shown in Figure~\ref{fig:context}, are from~\cite{https://doi.org/10.48550/arxiv.1812.06707}.

\textbf{Example 1}: We have a classification problem where one of the classes is `keyboard'. On nearly all images, keyboards are accompanied by monitors. The model might learn a shortcut bias for detecting monitors (detecting these might be easier than detecting keyboards): Then, the context (monitor pixels) will influence the keyboard score (logit) more than the actual keyboard presence. This process will not generalize to novel scenes where keyboards and monitors appear separately. If we remove the monitors from the image, the score for `keyboard' will go down. If we remove the keyboard from the image, the score for `keyboard' will stay quite high because the monitors are still present. Generally, co-occurring cues/features (diagonal samples) often lead to spurious correlations.

\textbf{Example 2}: The task is `frisbee', and the bias is `person'. It is easier to detect people because they are usually larger in images. The same phenomenon can be observed here as in \textbf{Example 1}.

\textbf{Note}: We humans also often look at the context to predict what is present in an image (or scene).

\subsubsection{Texture Bias}

Consider the task of object classification again. In this case, the task cue is the shape of the object and the bias cue is the texture of the object.

\textbf{Example}: Training a cat/dog classifier on a diagonal dataset, where the texture and shape are highly correlated. At test time, we want to predict cats when changing their texture (e.g., to greyscale, silhouette, edges, or to a marginally different texture). The accuracy of humans stays consistently high because we like to look at global shapes. Popular CNN models break down in such scenarios. However, when only the true texture of the original object (cat) is presented, models stay perfectly accurate while humans make more mistakes (90\% accuracy). The example is inspired by~\cite{https://doi.org/10.48550/arxiv.1811.12231}.

\textbf{Note}: Networks are prone to be biased towards textures because it is much easier to learn. If the task is `shape', such networks will generalize poorly to no/different textures.

\subsubsection{NLP Models Not Understanding the Exact Structure of the Sentence}

Our task of interest is natural language inference: Given premise and hypothesis, determine whether (1) the premise implies the hypothesis, (2) they contradict each other, or (3) they are neutral. The task cue is the whole sentence pair. However, the model might only use the set of words in the sentences, the lexical overlap cue, the subsequence cue, or the constituent cue. These are explained in the examples below, taken from~\cite{mccoy-etal-2019-right}. We consider three bias cues and corresponding premise-implication pairs for each.

\textbf{Example 1}: Lexical overlap cue. Assumes that a premise entails all hypotheses constructed from words in the premise. 
\begin{center}
The doctor was paid by the actor. \(\implies\) The doctor paid the actor.
\end{center}

\textbf{Example 2}: Subsequence cue. Assumes that a premise entails all of its contiguous subsequences.
\begin{center}
The doctor near the actor danced. \(\implies\) The actor danced.
\end{center}

\textbf{Example 3}: Constituent cue. Assumes that a premise entails all complete subtrees in its parse tree.
\begin{center}
If the artist slept, the actor ran. \(\implies\) The artist slept.
\end{center}
These can all lead to wrong implications, as seen above.

\subsubsection{Biased Action Recognition}

The model's task is to recognize the action that a human is performing on a video. The task cue is the human movement, e.g., swinging, jumping, or sliding. The bias cues might be the scene, the instrument (on/with which the action is performed), or the static frames. The quiz below is taken from~\cite{https://doi.org/10.48550/arxiv.1912.05534}.

\begin{figure}
    \centering
    \includegraphics[width=\linewidth]{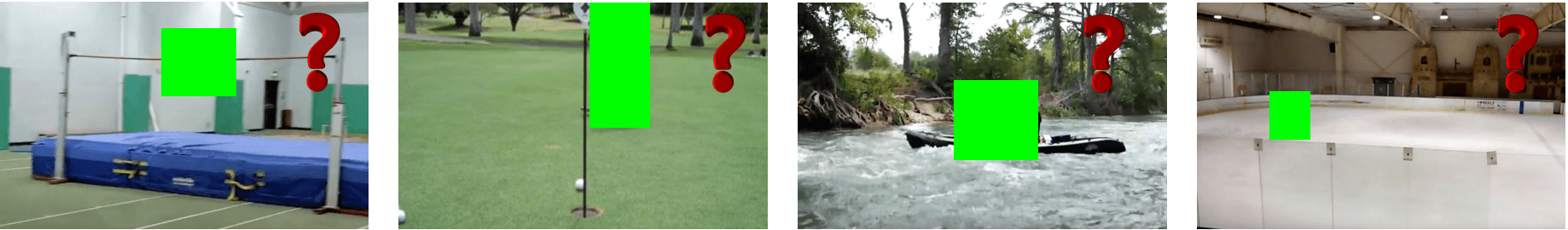}
    \caption{Example of four frames in videos where it is remarkably easy to predict a human's (very likely) action based on a single, static frame.}
    \label{fig:quiz}
\end{figure}

\textbf{Quiz}: Can the reader guess what action the blocked person is doing in the videos of Figure~\ref{fig:quiz}? Even from the scene alone, we as humans can have a good guess about what the person is likely doing. This tells us that humans also use many cues in the context to make predictions. However, we also know that there are many other possibilities; we are just giving the most likely prediction. When we observe the actual task cue, we can make predictions based on that. Machines fail miserably because they \emph{only rely on bias cues} from the dataset. We want ML models to be aware that they can be tricked in such cases; a notion of uncertainty and well-calibratedness is needed.

\subsubsection{Using a Single Domain for Multi-modal Tasks: Visual Question Answering}

The task is to answer a question in natural language using both the question and a visual aid (an image). The task cue is, therefore, both the question and the image. The bias cue is \emph{only} the question. When one of the modalities is already sufficient for making good predictions on the training set, the model can choose to only look at that cue because of the simplicity bias. This generalizes poorly to situations where both modalities are needed. The example below is inspired by~\cite{https://doi.org/10.48550/arxiv.1906.10169}.

\textbf{Example}: The question is ``What color are the bananas?''. In the image, we see a couple of green bananas. When the model only relies on the question, it will probably get this question wrong. (Correct answer: green, not yellow.)

\subsubsection{ML-based Credit Evaluation System using Sensitive Attributes}

The model is tasked to predict the possibility of future defaults for each individual. (Will the person go bankrupt, or will they be able to repay the loan?) The task cue is the size of the loans, history of repayment, income level, age, and similar factors. The bias cues are sensitive attributes that are not allowed to be used for the prediction, such as disability, gender, ethnicity, or religion. When an ML system learns to use bias cues to predict credit risks (that might not be explicit features in a vector representation), the model is not fair. The ML system requires further guidance to not use sensitive cues.

\subsection{Is the simplicity bias a bad thing?}

Whether shortcut bias is a good or a bad thing depends on the task.

\subsubsection{Simplicity Bias in ID Generalization}

Simplicity bias is actually \emph{praised} in ML in general, especially in ID generalization. There are reports saying
\begin{center}
\text{``Deep learning generalizes because the parameter-function map [\(\theta \mapsto f_\theta\)]}\\
\text{is biased towards simple functions.'' (\href{https://arxiv.org/abs/1805.08522}{Valle-Perez et al. 2019})}
\end{center}
The parameter space is enormous. If there is no inductive bias (from the training algorithm or the architecture), we can find whatever solution in the parameter space, many of which do not generalize well. But because of the simplicity bias, we will find some simpler rules that are very likely to generalize well \emph{to the same distribution}. (Here, we use the assumption that preference for simple cues usually leads to simple functions.)

\begin{figure}
    \centering
    \includegraphics[width=0.8\linewidth]{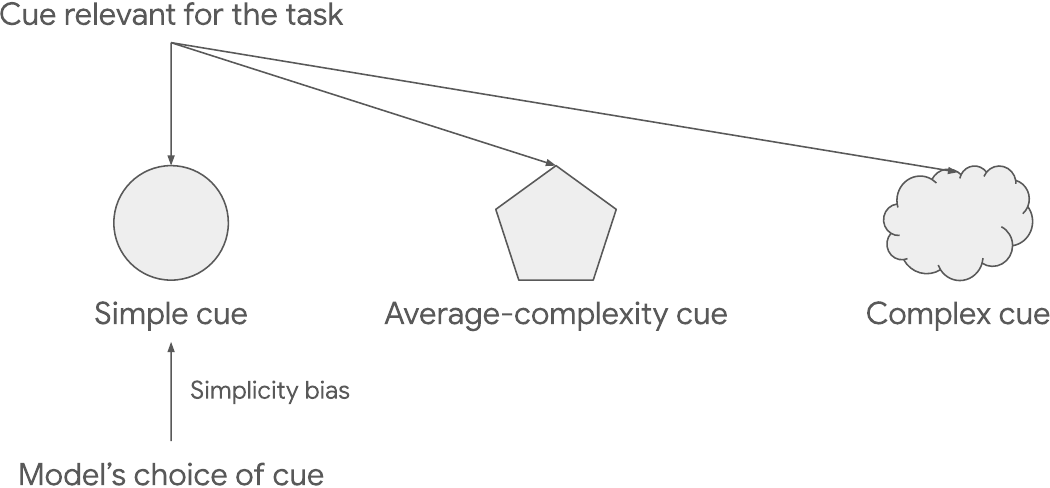}
    \caption{Example where the shortcut bias is favorable. For ID generalization tasks, simple cues are often sufficient for generalization.}
    \label{fig:favorable}
\end{figure}

The usefulness of the shortcut bias in ID generalization is illustrated in Figure~\ref{fig:favorable}. In the diagonal dataset case, any of the perfectly correlated cues are valid for performing well in deployment, considering ID generalization.

\subsubsection{Simplicity Bias in OOD Generalization}

\begin{figure}
    \centering
    \includegraphics[width=0.8\linewidth]{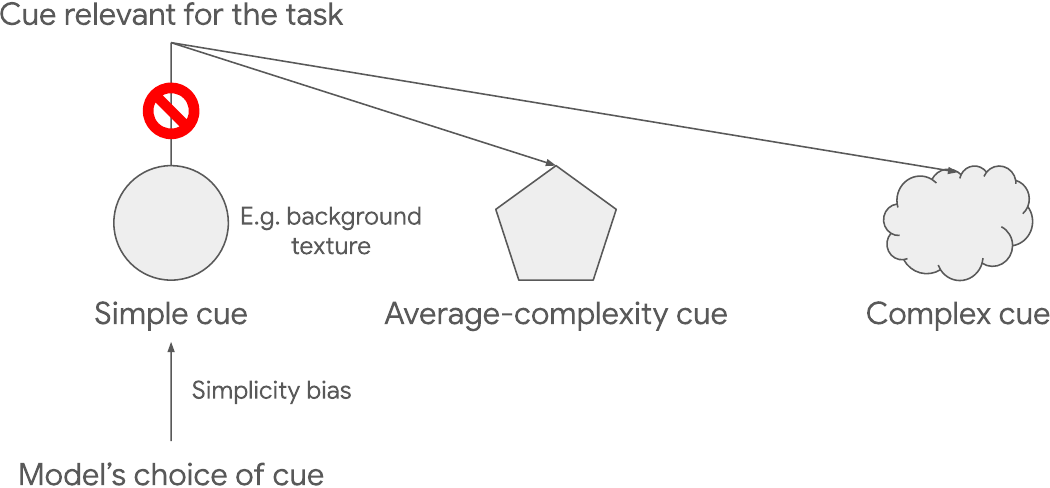}
    \caption{Example where the shortcut bias is unfavorable because of misspecification and does not lead to robust generalization. For OOD generalization tasks, simple cues may not work anymore.}
    \label{fig:unfavorable}
\end{figure}

For OOD generalization, the picture is a bit different. Simplicity bias is usually not welcome here because there are many OOD cases where the simplest cue is not good for generalization, as it is not relevant to the task. This causes problems during deployment, as the model's natural choice does not necessarily concur with the cue that would let the model generalize. For example, the background texture tends to be simple to recognize because we only have to look at very local parts of the image. The model might be able to use it to fit the training data well, but it will not usually generalize to different domains. For fairness, simple cues (e.g., parent's income) may also not be \emph{ethical} to use. We wish to prevent the model from using these cues. An example where the shortcut bias is unfavorable is given in Figure~\ref{fig:unfavorable}.

\section{Identifying and Evaluating Misspecification}
\label{sssec:identify}

As discussed in earlier sections, underspecification (as defined in Definition~\ref{def:underspec}) poses significant challenges to domain and cross-bias generalization. Therefore, it is crucial to diagnose whether our ML system suffers from misspecification.
There are two main strategies to evaluate misspecification~\cite{https://doi.org/10.48550/arxiv.2011.03395} (e.g., to determine whether the model uses too much context). Both are \emph{counterfactual} evaluation methods, i.e., they manipulate the input to determine what cue the model is looking at. (Counterfactual evaluation always seeks answers to questions of the form ``What would be the prediction if we changed \dots?''.) We either alter the task cue or the bias cue to observe the behavior of the model.

\textbf{Altering the task cue.} Here, the needed ingredients are the test set with task labels\footnote{The task label is needed to calculate the loss.} and cue disentanglement (the ability to change cues in the input independently). The evaluation method is as follows.
\begin{itemize}
    \item \textbf{Alter}: For every test sample, alter (or remove) the task-relevant cue. 
    \item \textbf{Decide}: If the model performance \textit{does not drop} significantly, our model is biased towards an irrelevant cue, meaning our system is misspecified.
\end{itemize}

\textbf{Altering the bias cue.} The needed ingredients are the same as when altering the task cue. The evaluation method is detailed below.
\begin{itemize}
    \item \textbf{Alter}: For every test sample, alter (or remove) the bias cue. 
    \item \textbf{Decide}: If the model performance \textit{drops} significantly, our model is biased towards the altered cue, which, again, means that our system is misspecified.
\end{itemize}
\textbf{Note}: This way, we also know \emph{what} our model is biased towards. With the previous method, we could only determine \emph{whether} our model is biased.

These desiderata can be formulated in terms of \emph{differences in accuracy/loss}. As long as there is a straightforward method that ranks biased and unbiased models correctly, it works well. Different papers do it differently.

\begin{figure}
    \centering
    \includegraphics[width=0.5\linewidth]{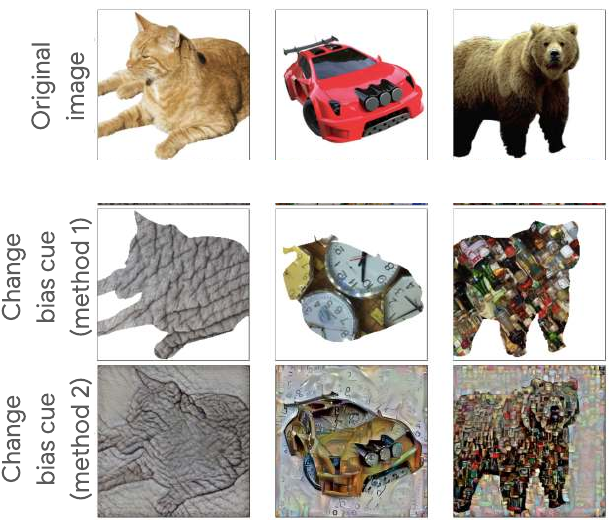}
    \caption{Example of two ways to change the (possible) bias cue of texture while preserving the task cue of shape.}
    \label{fig:twoways}
\end{figure}

\subsubsection{Examples of changing the bias cue}

\textbf{Example 1} (Figure~\ref{fig:context}): The task cue is `skateboard', and the bias cue is `person'. It is improbable to see a skateboard on the road without a person on it: the task cue is highly correlated with the bias cue. We remove the bias cue and see how the score for `skateboard' changes for a trained model. The needed ingredients are bounding box annotations/segmentation masks for objects and a good inpainting model. If the score for `skateboard' drops a lot, the model has been relying on the bias cue.

\textbf{Example 2} (Figure~\ref{fig:twoways}): The task cue is `shape', and the bias cue is `texture'. We consider two ways to change the bias cue: (1) Obtain a segmentation mask of the object and overlay a texture image of choice. (2) Style-transfer~\cite{https://doi.org/10.48550/arxiv.1508.06576} original image with a texture image of choice. The latter causes a less abrupt change: The image stays more reasonable. If the score of the true object drops significantly, the model has been relying a lot on the texture bias.

\subsubsection{Example of changing the task cue}

The following example is taken from~\cite{https://doi.org/10.48550/arxiv.1909.12434}. The task cue is the overall positivity/negativity (sentiment) of the review. The bias cue is ``anything but the task cue,'' e.g., the bag of words representation of a review. We let a human change the task cue (the sentiment analysis labels) by introducing minimal changes (a few words) in the sentences. If the score of the positive label does not change significantly after the update, the model does not rely on the overall meaning of the inputs.

To further illustrate the possible interventions the human annotator can make, we list some examples of changes made to the reviews:
\begin{itemize}
    \item Recasting fact as ``hoped for''.
    \item Suggesting sarcasm.
    \item Inserting modifiers.
    \item Replacing modifiers.
    \item Inserting negative phrases.
    \item Diminishing via qualifiers.
    \item Changing the perspective.
    \item Changing the rating and some words.
\end{itemize}
Some of these are indeed very subtle and a model that is biased to the bag of words that appear in the review cannot react to such changes.

\section{Overview of Scenarios for Selecting the Right Features}
\label{sssec:overview}

\begin{figure}
    \centering
    \includegraphics[width=\linewidth]{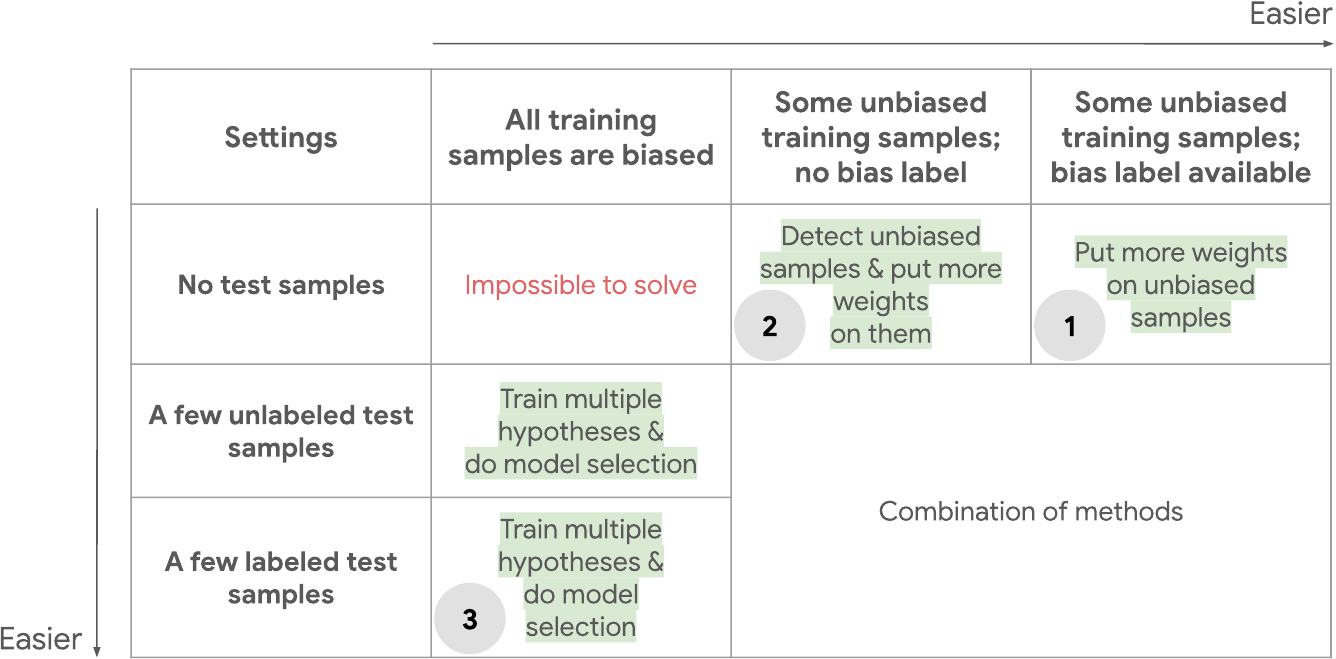}
    \caption{Overview of possible cross-bias generalization scenarios where the problem is made feasible by using different kinds of additional information. Scenario 2 can use prior knowledge about what the bias will be in the dataset. Under these assumptions, it can either detect unbiased samples and put more weight on them or make the final and intentionally biased models different (independent) in other ways. We will not discuss the upper version of scenario 3 (paper: ``\href{https://arxiv.org/abs/1909.13231}{Test-Time Training with Self-Supervision for Generalization under Distribution Shifts}''~\cite{sun2020testtime}), as we are not yet convinced that it is a possible case to solve in general deployment scenarios. We have no full trust yet.}
    \label{fig:scenarios}
\end{figure}

So far, we have seen that predictions of models are often based on \emph{bias} cues, while a key to generalization lies in their reliance on the \emph{task} cues. How could we ensure that our model uses the task cue for its predictions? We will see approaches to selecting the right features for many settings. Let us quickly review some possible scenarios with extra information in Figure~\ref{fig:scenarios}. The figure considers several settings that vary in their access to unbiased training samples or test samples as well as corresponding labels. It is important to understand that if we have no information apart from the diagonal dataset, the problem is conceptually unsolvable (top left cell). All the remaining cells describe different scenarios where generalization becomes possible again and we will discuss them in the next sections.

\section{Scenario 1 for Selecting the Right Features}

An example of this scenario is given in Figure~\ref{fig:first}. In this case, we have a small number of unbiased training samples (1\% or even less) with bias labels. This is the easiest setting, as we know which samples are unbiased: we simply compare \(Y\) with \(Z\). When they are equal, we have an on-diagonal sample. When they are unequal, the sample is off-diagonal (unbiased). We up-weight the off-diagonal samples and perform regular Empirical Risk minimization (ERM). This is the most naive approach, but it can perform well.

\begin{information}{How to find unbiased samples?}
What we discuss is, of course, a very simplistic setup. It is much more challenging to tell what samples are unbiased for the COCO dataset with 80 categories. However, if we know all target and bias labels, we can compute a matrix of co-occurrences between classes. We can then infer which images are more typical or atypical (e.g. a skateboard without a person is very unlikely), depending on the co-occurrence statistics of the labels. For very unlikely samples, we can, e.g., give a large weight during training. We generally weight samples more where the bias is either missing or different. However, there is an important caveat detailed in the example below.

\medskip

\textbf{Example}: We have a dataset with many images of cats, dogs, and humans appearing together. The task is to predict whether an image contains a cat. If we see a sample with both a cat and a dog present, can we call it an atypical (unbiased, off-diagonal) sample and give it a large weight? \emph{Only if the model is actually biased towards `human'.} If the model is biased towards `dog', this only aggravates the problem. Co-occurrence statistics are useful to give initial weights to samples but are usually only coarse proxies. Many biases are subtle and do not arise in an ``interpretable'' way. Determining weights post-hoc can directly act upon the problems of our model.

\medskip

We can only determine weights in such cases using the following routine:
\begin{enumerate}
    \item Train the network normally.
    \item Determine to which combination of cues (such as `dog' and `human' jointly, just `dog', or just `human') it is biased towards using the unbiased test set.
    \item Combat these biases by increasing the weights of samples that contain unlikely combinations of cues \wrt the present biases.
\end{enumerate}

\medskip

The computationally complex part here is annotation. Generally, it is a very strong assumption that we have labels for all possible cues! Once we have the task and bias labels, we create a counting matrix for co-occurrences which is easily computable on the CPU. In the COCO object detection dataset, there are many objects on a single image usually, so co-occurrences are easy to calculate. (Our assumption here is that labeling is complete.)
\end{information}

\begin{information}{Model becoming biased again}
What happens if we have biased fish images (i.e., fish are always in the hands of fishermen on the images) and we get unbiased images (e.g., fish in water), but the model learns shortcuts again (water background \(\implies\) fish)? There are two solutions in general.

\medskip

\textbf{Bottom-up, incremental approach.} We continuously search for the current model's biases by testing it for different sets of correlations (like testing our model's performance on fish images for a set of potential biases using unit tests). Such sets can be constructed by removing/replacing possible shortcuts (e.g., water background) in the original images. If we find that our model now uses some shortcuts, we incorporate new samples without the corresponding biases. We continue doing this until the possible ways to learn shortcuts are saturated (i.e., it becomes more complicated than the task itself), and we are happy with the model. This approach does not guarantee that the ultimate model is unbiased, and usually, it is quite complicated to extensively test our model for potential biases.

\medskip

\textbf{Top-down approach.} Let us assume that some explainability method provided us with a comprehensive and complete list of cues the model is actually looking at. In such a case, we first determine what cues are task cues and what are bias cues by human inspection. Then, we remove the bias cues from our model and include others they should be looking at more. \textbf{Disclaimer}: There is no technique for this in general, but it would be very nice to have one. This is very much the frontier of research in explainability.
\end{information}

\subsection{Group DRO}

Let us see how the availability of a small set of unbiased samples can be exploited in practice.
In this section, we will discuss a method introduced in the paper ``\href{https://arxiv.org/abs/1911.08731}{Distributionally Robust Neural Networks for Group Shifts: On the Importance of regularization for Worst-Case Generalization}''~\cite{https://doi.org/10.48550/arxiv.1911.08731}, called Group Distributionally Robust Optimization (Group DRO). The goal of this method is to have the same accuracy for different bias groups (elements of the bias-task matrix depicted in Figure~\ref{fig:diagdomain}). This goal is achieved by minimizing the maximum loss across the groups. In the following paragraphs, we will discuss how this minimization is performed.

\subsubsection{Optimization problem in Group DRO}

In vanilla Empirical Risk Minimization (ERM), we have the following optimization problem:
\[\argmin_{\theta \in \Theta} \nE_{(x, y) \sim \hat{P}}\left[\ell(\theta; (x, y))\right].\]
To achieve the goal of minimizing the maximum loss across the groups in DRO, the optimization problem is modified to the following one:
\[\argmin_{\theta \in \Theta} \left\{\mathcal{R}(\theta) := \sup_{Q \in \mathcal{Q}} \nE_{(x, y) \sim Q} \left[\ell(\theta; (x, y))\right]\right\}.\]
Here, \(\mathcal{Q}\) encodes the possible test distributions we want to do well on. It should be chosen such that we are robust to distribution shifts, but we also do not get overly pessimistic models that optimize for implausible worst-case distributions \(Q\).

Let us now choose \(\mathcal{Q} := \left\{\sum_{g = 1}^m q_gP_g : q \in \Delta_m\right\}\) where \(\Delta_m\) is the \((m - 1)\)-dimensional probability simplex and \(P_g\) are group distributions. These can correspond to arbitrary groups, but for our use case, the groups are based on spurious correlations. If we go back to our toy example of a (color, shape) dataset, then the individual groups can correspond to all possible (color, shape) combinations). Then
\[\mathcal{R}(\theta) = \max_{g \in \{1, \dotsc, m\}} \nE_{(x, y) \sim P_g}\left[\ell(\theta; (x, y))\right],\]
as the optimum of a linear program (the way we defined \(\mathcal{Q}\)) is always attained at a vertex (a particular \(P_g\)). Now, if we consider the empirical distributions \(\hat{P}_g\), we get \textbf{Group DRO}:
\[\argmin_{\theta \in \Theta}\left\{\hat{\mathcal{R}}(\theta) := \max_{g \in \{1, \dotsc, m\}} \nE_{(x, y) \sim \hat{P}_g}\left[\ell(\theta; (x, y))\right]\right\}.\]
The learner aims to make predictions for \emph{the worst-case group} better. Ideally, at the end of training, we have the same loss for each group (considering equal label noise across groups -- if one group has huge corresponding label noise, the learner either overfits to the noise severely, which is suboptimal, or we do not have the same loss for each group at the end of training).

\subsubsection{Examples for the groups in Group DRO}

\textbf{Toy example.} In our previous example (Figure~\ref{fig:first}), all possible combinations of shape and color can be treated as a group. This way, we take into account the underrepresented combinations appropriately. We can also treat the on-diag and off-diag samples as the two groups, which might be a more stable choice if there are very few off-diag samples.

\textbf{Faces.} Assume a dataset of celebrities where the task is to predict gender from the image. The hair color annotation is also available. Further, assume that we have access to many diagonal samples and a small amount of off-diagonal samples where
\begin{align*}
&P_1\colon \text{ blonde female} &\text{50\%}\\
&P_2\colon \text{ dark-haired male} &\text{40\%}\\
&P_3\colon \text{ blonde male} &\text{3\%}\\
&P_4\colon \text{ dark-haired female} &\text{7\%}.\hspace{-0.35em}
\end{align*}
If we just performed ERM/Regularized Risk Minimization (RRM), the model would usually predict based on a mixture of cues that would still favor the larger groups more and still be able to achieve high accuracy as we explicitly optimize on the average loss. For example, it could predict based on hair color: for dark-haired people, we could predict `male', and for blonde individuals, we could predict `female'. However, Group DRO helps us optimize on the worst-case combination, which can help prevent shortcuts.

\textbf{Humans and skateboards.} We consider one group comprising samples that contain a skateboard but not a human and another group comprising samples of skateboards with a human.

\subsubsection{The Group DRO algorithm}

Roughly speaking, Group DRO minimizes its optimization objective by performing the following steps:
\begin{enumerate}
    \item Calculate losses for all groups.
    \item Select the group with the maximal loss.
    \item Set the model's gradient active only on the training samples from the worst-performing group.
    \item Repeat.
\end{enumerate}

The actual algorithm (Algorithm~\ref{alg:group_dro}) is a bit more complicated. It considers an exponential moving average for the weights of different groups and performs gradient steps \wrt these weights. This modification allows the method to be trained with SGD. It also has nice convergence guarantees~\cite{https://doi.org/10.48550/arxiv.1911.08731}.

\begin{algorithm}
\caption{Online optimization algorithm for group DRO}
\label{alg:group_dro}
\DontPrintSemicolon
\KwData{Step sizes \(\eta_q, \eta_\theta; P_g\) for each \(g \in \{1, \dotsc, m\}\)}
Initialize \(\theta^{(0)}\) and \(q^{(0)}\)

\For{\(t = 1, \dots, T\)}{
  \(g \sim \operatorname{Uniform}(1, \dots, m)\) \Comment*[r]{Choose a group \(g\) at random}
  \(x, y \sim P_g\) \Comment*[r]{Sample \(x, y\) from group \(g\)}
  \(q' \gets q^{(t - 1)}; q'_g \gets q'_g \exp\left(\eta_q \ell\left(\theta^{(t - 1)}; (x, y)\right)\right)\) \Comment*[r]{Update weights for group \(g\)}
  \(q^{(t)} \gets q' / \sum_{g' \in \{1, \dotsc, m\}} q'_{g'}\) \Comment*[r]{Renormalize \(q\)}
  \(\theta^{(t)} \gets \theta^{(t - 1)} - \eta_\theta q^{(t)}_g \nabla \ell\left(\theta^{(t - 1)}; (x, y)\right)\) \Comment*[r]{Use \(q\) to update \(\theta\)}
}
\end{algorithm}

\begin{information}{Comments for the Group DRO algorithm}

\textbf{Smoothed group-wise updates.} In Algorithm~\ref{alg:group_dro}, \(q^{(t)}_g\) influences the step size for the sample (and the corresponding group in general). This formulation can be considered a smoothed version of the original one, as we do not select the worst-performing group but still base the update on the group-wise performances.

\medskip

\textbf{Looking at the worst-group metric.} In general, the method performs worse than ERM on the average accuracy metric, as ERM directly optimizes on that. However, Group DRO shines on the worst-group accuracy metric, which is directly optimized by the method. ERM usually breaks down completely on the worst-group accuracy metric when there are notable group imbalances in the dataset.

\end{information}

\subsubsection{Ingredients for Group DRO}

In Group DRO, we have
\[\text{samples for } (X, Y, \red{G}) = (\text{input}, \text{output}, \red{\text{group}})\]
where the groups come from, e.g., spurious correlations or demographic groups. As we have group labels in addition to the usual setup (the difference is highlighted in red), we expect better worst-case accuracy. By explicitly optimizing on the worst-case spurious correlation/group, our model might generalize better in deployment.

\begin{definition}[label=def:attrlab]{Attribute Label}
Attribute labels are indicators of all possible factors of variation in our data. Domain labels are a particular case of these.
\end{definition}

The group label can not only be a bias or domain label, but even a general attribute label (Definition~\ref{def:attrlab}). This additional cue makes cross-domain generalization less ill-posed.\footnote{This is a more general statement than only considering cross-bias generalization -- whenever we are presented with a (nearly) diagonal dataset, we need additional information, and this can happen in any cross-domain setting.}

\subsection{Domain-Adversarial Training of Neural Networks (DANN)}
\label{sssec:dann}

\begin{figure}
    \centering
    \includegraphics[width=\linewidth]{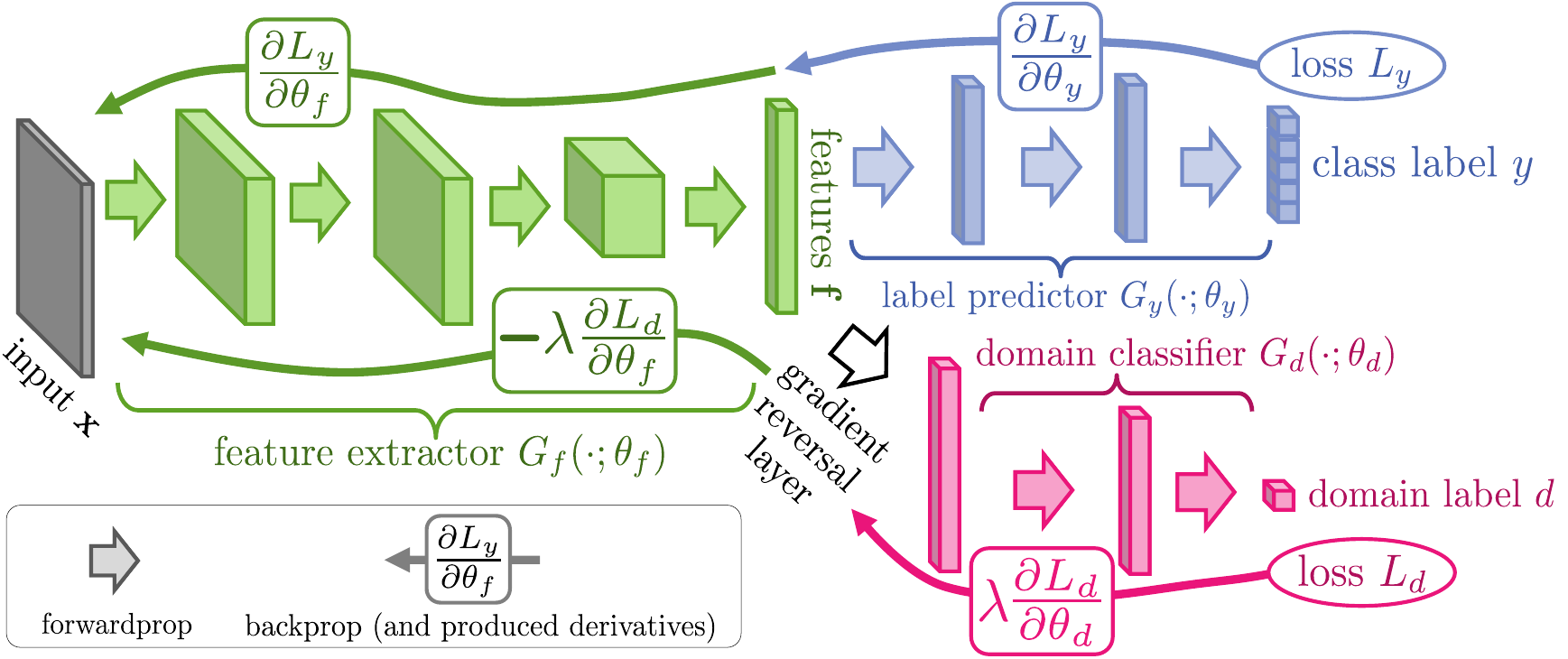}
    \caption{Overview of the DANN method. The feature extractor is encouraged to provide strong representations for predicting the class label and not contain any information about the domain label. Figure taken from the paper~\cite{https://doi.org/10.48550/arxiv.1505.07818}.}
    \label{fig:dann}
\end{figure}

Apart from Group DRO, we have one more algorithm for Scenario 1 to discuss, called ``Domain-Adversarial Training of Neural Networks'' (DANN).
The DANN method was introduced in the paper ``\href{https://arxiv.org/abs/1505.07818}{Domain-Adversarial Training of Neural Networks}''~\cite{https://doi.org/10.48550/arxiv.1505.07818} and is another method to select good cues given bias labels by removing domain information from the intermediate features. An overview of the method taken from the original paper is shown in Figure~\ref{fig:dann}.

The idea of the method is to add an additional head to the model ({\color{magenta} magenta} in the image) that would predict bias labels (named domain labels in the paper) and adversarially train a feature extractor (\green{green} in the image) such that the features it extracts are maximally non-informative for the additional head to predict bias labels but still informative for the original head (\blue{blue} in the image) to solve the main task.

This is achieved by splitting the training process into two parts. In the first part, the original head and feature extractor are jointly trained with gradient descent for the main task. In the second part, the bias-predicting head is trained with gradient descent for domain label prediction. The feature extractor parameters are adversarially trained with gradient \emph{ascent} to maximize the loss of the bias-predicting head. Intuitively, we optimize the bias-predicting head to ``squeeze out'' any domain information left in the extracted features.

DANN can be assigned to the group of methods that select task cues given bias labels by removing information about the bias from the intermediate features.

\subsubsection{DANN Optimization Objective}

We denote the prediction loss by
\[\cL^i_y(\theta_f, \theta_y) = \cL_y(G_y(G_f(x_i; \theta_f); \theta_y), y_i)\]
and the domain loss by
\[\cL^i_d(\theta_f, \theta_d) = \cL_d(G_d(G_f(x_i; \theta_f); \theta_d), d_i).\]
The training objective of DANN is
\[E(\theta_f, \theta_y, \theta_d) = \frac{1}{n}\sum_{i = 1}^n \cL^i_y(\theta_f, \theta_y) - \lambda \left(\frac{1}{n}\sum_{i = 1}^n \cL^i_d(\theta_f, \theta_d) + \frac{1}{n'}\sum_{i = n + 1}^N \cL^i_d(\theta_f, \theta_d)\right),\]
and the optimization problem is finding the saddle point \(\hat{\theta}_f, \hat{\theta}_y, \hat{\theta}_d\) such that
\begin{align*}
\left(\hat{\theta}_f, \hat{\theta}_y\right) &= \argmin_{\theta_f, \theta_y} E\left(\theta_f, \theta_y, \hat{\theta}_d\right),\\
\hat{\theta}_d &= \argmax_{\theta_d} E\left(\hat{\theta}_f, \hat{\theta}_y, \theta_d\right).
\end{align*}

\subsubsection{Breaking DANN apart}

First, we discuss the above formulation, which is for \emph{domain adaptation}. The DANN method was originally proposed for this task. The first term of the training objective is the loss term for correct task label prediction on domain 1. The second term is the loss term for correct domain label prediction. We have two sums in the second term for domain 1 and domain 2 samples, respectively. For domain 2, we only have \emph{unlabeled samples}, but we \emph{do} have domain labels. The set of domain labels we have is simply \{domain 1, domain 2\}. Obtaining \(\left(\hat{\theta}_f, \hat{\theta}_y\right)\) means minimizing the first term in \(\theta_f, \theta_y\) and maximizing the second term in \(\theta_f\). Similarly, we obtain \(\hat{\theta}_d\) by minimizing the second term in \(\theta_d\).

\subsubsection{Using DANN for cross-bias generalization}

We can easily adapt the DANN formulation to cross-bias generalization. In particular, we treat \(y\) as the task label (e.g., shape: \{circle, triangle, square\}) and \(d\) as the bias label (e.g., color: \{red, green, blue\}). Here, the first term enforces correct predictions on both the biased and unbiased samples and the second term is used to kill out information about the bias from the representation. On off-diagonal samples, the bias label is not the task label, thus, \(f\) will be optimized to ``forget'' the bias labels while predicting the task labels correctly. We do not need unbiased samples as long as we have access to labeled samples from the target domain. It could happen that, e.g., the target domain is also biased, just in a different way than the training set.

We could also treat the set of biased samples as domain 1, the set of unbiased samples as domain 2, and use the original formulation of DANN for cross-bias generalization. This approach also works with target domain samples instead of unbiased ones.

\subsubsection{Results of DANN for domain adaptation}

To obtain good results with DANN, it is crucial to choose the hyperparameters well. All hyperparameters are chosen fairly in the paper, and there is no information leakage (e.g., by using the test set for choosing hyperparameters). Most hyperparameters are chosen using cross-validation and grid search on a log scale. Some are kept fixed or chosen among a set of sensible values. This is a usual practice in machine learning research. For large-scale experiments, the authors give fixed formulas for the LR decay and the scheduler for the domain adaptation parameter \(\lambda\) for the feature extractor (from 0 to 1), and fixed values for the momentum and the domain adaptation parameter for the domain classifier (\(\lambda = 1\) to ensure that the domain classifier trains as fast as the label predictor).

The model is evaluated on generalizability between different \emph{Amazon review topics} on the sentiment analysis task. The results are shown in the top table of Table~\ref{tab:dannres}. There is no significant difference between how NNs, SVMs, and DANN generalize. DANN is very slightly better on most review topic combinations. DANN is also evaluated on generalizability between MNIST and MNIST-M, SVHN and MNIST, and other datasets for the same task. The results of these experiments are shown in Table~\ref{tab:dannres2}. On these benchmarks, DANN performed a lot better than NNs and SVMs.

\begin{table*}
\centering
\caption{Classification accuracy on the Amazon reviews dataset. DANN leads to very slight improvements on most (source, target) pairs. Table taken from~\cite{https://doi.org/10.48550/arxiv.1505.07818}.}
\label{tab:dannres}
\small {\small\sc\renewcommand{\arraystretch}{1.2}
\begin{tabular}{llcccccc}
\toprule
\multicolumn{2}{c}{  } & \multicolumn{3}{c}{\textbf{Original data}}  & \multicolumn{3}{c}{\textbf{mSDA representation}} \\
 \cmidrule(l r){3-5} \cmidrule(l r){6-8}
 Source  & Target  & DANN & NN  & SVM & DANN &  NN &  SVM  \\
\cmidrule(l r){1-2} \cmidrule(l r){3-5} \cmidrule(l r){6-8}
 
books & dvd &.784	 &.790	     &\textbf{.799} &.829	 &.824&\textbf{.830} 	 \\ 
books & electronics    &.733 &.747 &	\textbf{.748} 			 & \textbf{.804} 		 &		.770	 &		.766 \\ 
books & kitchen    & \textbf{.779} 		 &	.778		 &		.769		 & \textbf{.843} 		 &		.842	 &		.821 \\ 
dvd & books      & .723	 &	.720		 &	\textbf{.743} 			 &	.825	 &		.823	 &	\textbf{.826} 	 \\ 
dvd & electronics   & \textbf{.754} 		 &	.732		 &		.748		 & \textbf{.809} 		 &		.768	 &		.739 \\ 
dvd & kitchen    & \textbf{.783} 		 &	.778		 &		.746		 &	.849	 &	\textbf{.853} 		 &		.842 \\ 
electronics & books   &	\textbf{.713} 	 &	.709		 &	    .705		 & \textbf{.774} 		 &		.770	 &		.762 \\ 
electronics & dvd     &	\textbf{.738} 	 &	.733		 &		.726		 &	\textbf{.781} 	 &		.759	 &		.770 \\ 
electronics & kitchen    & \textbf{.854} 		 &\textbf{.854} 			 &		.847		 &	.881	 &	\textbf{.863} 		 &		.847 \\ 
kitchen & books  &	\textbf{.709} 	 &	.708		 &		.707		 &	.718	 &		.721	 &	\textbf{.769} 	\\ 
kitchen & dvd   & \textbf{.740} 		 &	.739		 &		.736		 & \textbf{.789} 		 &	\textbf{.789} 		 & .788 \\ 
kitchen & electronics   &\textbf{.843}	 &	.841		 &	.842	 &	.856	 &		.850	 &	\textbf{.861} 	 \\ 

\bottomrule
\end{tabular}}
\end{table*}

\begin{table*}
\centering
    \begin{small}
      \begin{sc}
        \renewcommand{\arraystretch}{1.3}
        \begin{tabular}{l r | c c c c}
          \toprule
          \multirow{2}{*}{Method} & {\scriptsize Source} & MNIST & Syn Numbers & SVHN & Syn Signs \\
          & {\scriptsize Target} & MNIST-M & SVHN & MNIST & GTSRB \\
          \midrule
          \multicolumn{2}{l |}{Source only} & 
          $ .5225 $                      & $ .8674 $                      & $ .5490 $                      & $ .7900 $                      \\
          \multicolumn{2}{l |}{SA (Fernando \etal, 2013)} & 
          $ .5690 \; (4.1\%) $           & $ .8644 \; (-5.5\%) $          & $ .5932 \; (9.9\%) $           & $ .8165 \; (12.7\%) $          \\
          \multicolumn{2}{l |}{DANN} & 
          $ \mathbf{.7666} \; (52.9\%) $ & $ \mathbf{.9109} \; (79.7\%) $ & $ \mathbf{.7385} \; (42.6\%) $ & $ \mathbf{.8865} \; (46.4\%) $ \\
          \multicolumn{2}{l |}{Train on target} & 
          $ .9596 $                      & $ .9220 $                      & $ .9942 $                      & $ .9980 $                      \\
          \bottomrule
        \end{tabular}
      \end{sc}
    \end{small}
    \caption{Classification accuracy of DANN compared to other methods on images of digits, assuming different source and target domains. ``MNIST-M corresponds to difference-blended digits over non-uniform background. The first row corresponds to the lower performance bound (i.e. if no adaptation is performed). The last row corresponds to training on the target domain data with known class labels (upper bound on the DA performance).''~\cite{https://doi.org/10.48550/arxiv.1505.07818} Table taken from~\cite{https://doi.org/10.48550/arxiv.1505.07818}.}
  \label{tab:dannres2}
\end{table*}

We should always take a look at how papers choose hyperparameters. For a more complicated model, like DANN, there are many hyperparameters to choose from. Depending on how smartly we choose them, we get dramatically different results. When comparing methods, we also need to make sure that we spend the same resources for tuning the hyperparameters of all methods. The DANN paper provides fair comparisons.

\subsubsection{Ingredients for DANN}

In DANN, like in Group DRO, we also have access to
\[\text{samples for } (X, Y, G) = (\text{input}, \text{output}, \text{group}).\]
The group label can again be a bias or domain label, but even a general attribute label. By using group supervision, we make cross-domain generalization less ill-posed.

\section{Scenario 2 for Selecting the Right Features}

Let us consider another cross-bias generalization setting from Figure~\ref{fig:scenarios}: Scenario 2. Here, we consider an abundance of biased samples, a few available unbiased training samples (\(<\) 1\%), and no bias labels. As we do not know which samples are biased (we only have task labels), we need additional assumptions/information on the bias to solve the problem.\footnote{Unless there are a lot of unbiased samples. Then we simply do ERM, and we basically have ID training.} The question becomes how to identify unbiased samples and how to amplify them.

Before answering this question, let us first think about what assumptions we can make about the bias. The usual assumption is that the bias cue is simple and the task cue (what we want to learn) is more complex. For example, when the task is `shape', and bias is `color', this assumption holds. When reversing the roles, the assumption is violated.

This assumption on simplicity leads us to the following possible additional assumptions:
\begin{enumerate}
    \item Bias is the first cue that a generic model learns.
    \item Bias is the cue that is learned by a model of a certain limited capacity (i.e., by a short-sighted, myopic model).
\end{enumerate}

\textbf{Note}: Sometimes, the assumption of the bias cue being a simpler cue than the task cue is violated. Practitioners have to understand the complexity of task cues and possible bias cues to successfully leverage methods with the above assumptions.

\begin{figure}
    \centering
    \includegraphics[width=\linewidth]{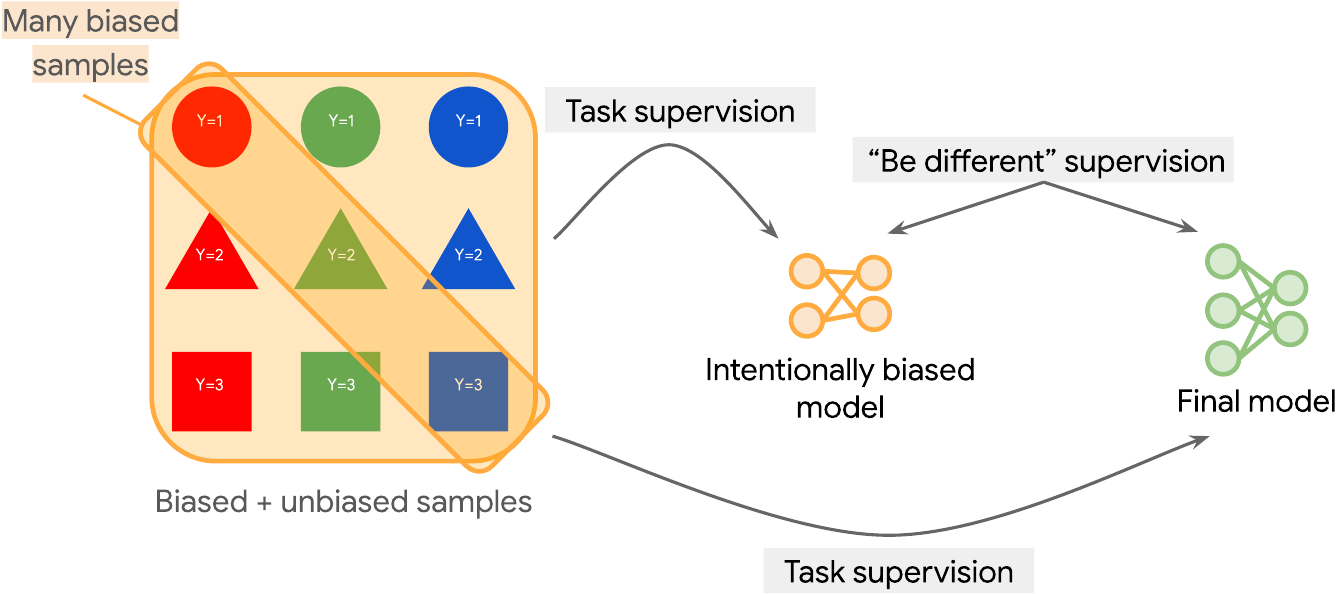}
    \caption{A general framework for selecting the right features, referred to as ``Scenario 2'' in the text. The \emph{intentionally based model} is trained on the entire training set using task supervision.}
    \label{fig:scenario2}
\end{figure}

In the next sections, we will describe a set of methods that identify unbiased samples based on these assumptions. The framework depicted in Figure~\ref{fig:scenario2} is a clear basis for our discussion. Before diving into it, we would like to explain two important modules from this framework: ``Intentionally biased model'' and ``Be different'' supervision.

\begin{definition}{Intentionally Biased Model}
An intentionally biased model is designed to learn bias cues quickly, based on the assumptions we made before. 

\medskip

We consider several examples of an intentionally biased model:
\begin{itemize}
    \item The model is trained for a small number of epochs. Whatever pattern that can already be learned in the first few epochs is considered bias.
    \item The model is not trained for a few epochs, but its initial correct predictions are amplified during training. This is conceptually very similar to the previous example but is perhaps more performant.
    \item The model has an architectural constraint: (1) CNN with a smaller receptive field. It can only extract very local information (e.g., texture patterns), not global shape. When the bias is `texture', this is the way to go. (2) Transformer with shallow depth. It can only learn very simplistic relationships. When our bias is simple, this can work. (3) Single-modality model. This is one way to go when the actual task requires looking at multiple modalities to solve the problem.
\end{itemize}
\end{definition}

\begin{definition}{``Be different'' Supervision}
``Be different'' supervision is a type of regularization that forces the final model to be different from the intentionally biased model. The final model is trained on the original task loss with regularization based on the biased model. The biased model might be trained \emph{before} the final model or \emph{in tandem} (Learning from Failure: Section~\ref{ssec:lff}, ReBias: Section~\ref{ssec:rebias}).

\medskip

Examples of the ``be different'' supervision:
\begin{itemize}
    \item Sample weighting based on biased model.
    \item Achieving representational independence.
\end{itemize}
\end{definition}

\subsection{Learning from Failure}
\label{ssec:lff}

\begin{figure}
    \centering
    \includegraphics[width=0.8\linewidth]{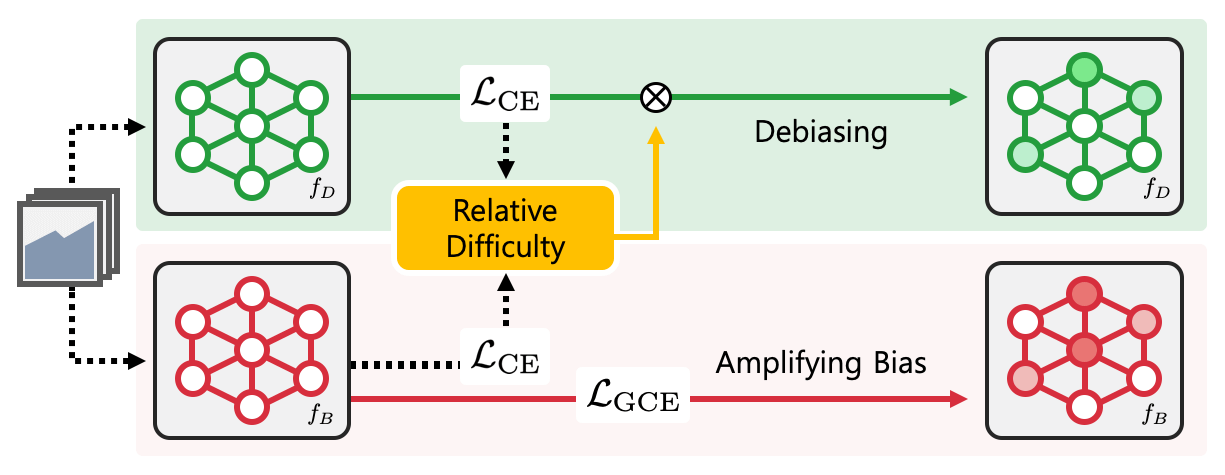}
    \caption{Overview of the Learning from Failure method. The intentionally biased model is used for determining the sample weights in the loss of the debiased model based on relative difficulty. Figure taken from the paper~\cite{DBLP:journals/corr/abs-2007-02561}.}
    \label{fig:lff}
\end{figure}

The method we consider now was introduced in the paper ``\href{https://arxiv.org/abs/2007.02561}{Learning from Failure: Training Debiased Classifier from Biased Classifier}''~\cite{DBLP:journals/corr/abs-2007-02561}. An overview, taken from the paper, is shown in Figure~\ref{fig:lff}.

Here, an \emph{intentionally biased} model is obtained by training with the following special loss that amplifies biases:
\[\cL_\mathrm{GCE}(p(x; \theta), y) = \frac{1 - p_y(x; \theta)^q}{q}\]
where \(y\) is the GT class and \(q > 0\). 

This loss forces the intentionally biased model to focus on samples for which the predicted ground truth probability is already high.
To understand why it happens, it can be shown that
\[\frac{\partial \cL_\mathrm{GCE}(p(x; \theta), y)}{\partial \theta} = p_y(x; \theta)^q \frac{\partial \cL_\mathrm{CE}(p(x; \theta), y)}{\partial \theta}\]
and as \(q \downarrow 0, \cL_\mathrm{GCE} \rightarrow \cL_\mathrm{CE}\).

The final model \(f_D\) is trained to be \emph{different} from the intentionally biased model by assigning the following sample weights:
\[\cW(x) = \frac{\cL_\mathrm{CE}(f_B(x), y)}{\cL_\mathrm{CE}(f_B(x), y) + \cL_\mathrm{CE}(f_D(x), y)}\]
where
\[\cL_\mathrm{CE}(p(x; \theta), y) = -\log p_y(x; \theta).\] 
Such weights force the final model to focus on the samples on which an intentionally biased model makes more mistakes.
The final training algorithm, as presented in the paper, is shown in Algorithm~\ref{alg:lff}.

\begin{algorithm}
\caption{Learning from Failure}
\label{alg:lff}
\KwData{\(\theta_B, \theta_D\), training set \(\cD\), learning rate \(\eta\), number of iterations \(T\)}
Initialize two networks \(f_B(x; \theta_B)\) and \(f_D(x; \theta_D)\)

\For{\(t = 1, \dots, T\)}{
    Draw a mini-batch \(\cB = \{(x^{(b)}, y^{(b)})\}_{b = 1}^B\) from \(\cD\)
    
    Update \(f_B(x; \theta_B)\) by \(\theta_B \gets \theta_B - \eta \nabla_{\theta_B} \sum_{(x, y) \in \cB} \cL_\mathrm{GCE}(f_B(x), y)\)
    
    Update \(f_D(x; \theta_D)\) by \(\theta_D \gets \theta_D - \eta \nabla_{\theta_D} \sum_{(x, y) \in \cB} \cW(x) \cdot \cL_\mathrm{CE}(f_D(x), y)\)
}
\end{algorithm}

\subsubsection{Breaking LfF Apart}

The intentionally biased model is trained with \(\cL_\mathrm{GCE}\). It amplifies whatever is predicted at the first iterations through the rest of the training. For example, if the model first learns `color', then the loss amplifies color-based predictions and enforces the same predictions throughout training.

The final model is then forced to think of different hypotheses than the first model. If the biased model correctly predicts a sample, it gets less weight in the loss for the final model.

With \(q > 0\), \(\cL_\mathrm{GCE}\) assigns more weight on confident samples, which results in larger gradient updates for these. The larger \(q\) is, the more the perfect predictions are weighted compared to imperfect ones. We train wrong predictions very slowly and initial predictions are strengthened over time.

\textbf{Assumption on bias}: Biases are the cues that are learned first. The method rewards easy samples to be learned quickly, and harder samples that were not predicted correctly to be given up by the intentionally biased model. Thus, this model is indeed biased towards easy cues.

\begin{figure}
    \centering
    \includegraphics{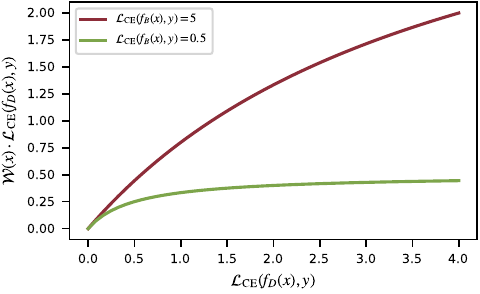}
    \caption{Illustration of \(\cW(x) \cdot \cL_\mathrm{CE}(f_D(x), y)\) as a function of \(\cL_\mathrm{CE}(f_D(x), y)\) for \(\cL_\mathrm{CE}(f_B(x), y) \in \{0.5, 5\}\). Samples with a higher loss for the biased model are more important for the unbiased model.}
    \label{fig:gce}
\end{figure}

For hard samples, \(\cL_\mathrm{CE}(f_B(x), y)\) is large throughout the training procedure. Both \(\cL_\mathrm{CE}(f_B(x), y)\) and \(\cL_\mathrm{CE}(f_D(x), y)\) are high for all \(x \in \cD\) in the beginning. The better \(f_D\) becomes on a sample, the more it is weighted (as \(\cL_\mathrm{CE}(f_D(x), y)\) decreases). However, the weight is multiplied by \(\cL_\mathrm{CE}(f_D(x), y)\), which balances this trend out. An illustration of \(\cW(x) \cdot \cL_\mathrm{CE}(f_D(x), y)\) is given in Figure~\ref{fig:gce}. Samples with high \(\cW(x)\) are ones that the biased model cannot handle well. Under our assumptions on the bias, samples with high \(\cW(x)\) are the unbiased ones. Thus, \(\cW(x)\) replaces the missing bias labels. Sample weights have a similar effect as the ``upweighting'' of the underrepresented group in Group DRO.

\textbf{Note}: In LfF, depending on the predictions of the first iteration, we choose the samples on which we wish and do not wish to train further. As a simpler baseline, we could also just train the intentionally biased model for 1-2 epochs but with the original cross-entropy loss. However, researchers usually prefer more `continuous' solutions rather than such thresholds and rules of thumb.

\subsubsection{Results of LfF}

\begin{figure}
    \centering
    \includegraphics[width=0.7\linewidth]{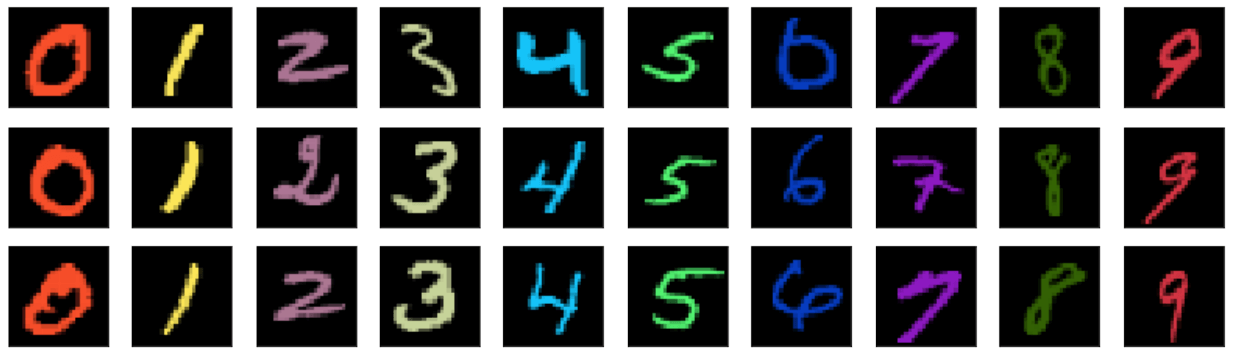}
    \caption{A representative sample from the Colored MNIST dataset~\cite{DBLP:journals/corr/abs-2007-02561}.}
    \label{fig:colormnist}
\end{figure}

\begin{figure}
    \centering
    \includegraphics[width=0.8\linewidth]{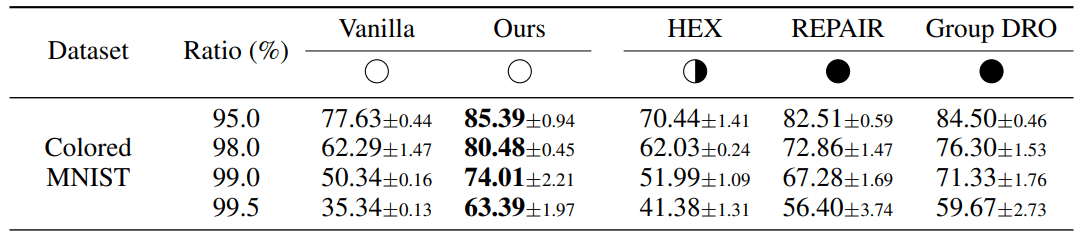}
    \caption{Results of LfF on Colored MNIST~\cite{https://doi.org/10.48550/arxiv.1907.02893}. LfF is significantly better than vanilla training but also shows improvements compared to other debiasing methods. There are [Ratio]\% biased samples and [1 - Ratio]\% unbiased samples. Table taken from the paper~\cite{DBLP:journals/corr/abs-2007-02561}.}
    \label{fig:lffcmnist}
\end{figure}

The paper showcases results on the Colored MNIST~\cite{https://doi.org/10.48550/arxiv.1907.02893} dataset where the task is the shape of the digit and the bias is the color of the digit. A sample from this dataset can be seen in Figure~\ref{fig:colormnist}, and the results are shown in Figure~\ref{fig:lffcmnist}. The results show that if we train a model for digit classification, it tends to pick up color much more quickly than the actual digit shape. The fact that we are improving performance by using LfF shows that
\begin{enumerate}
    \item color is indeed learned first; and
    \item color was indeed a bias that should be removed from consideration for digit recognition.
\end{enumerate}
The lower the percentage of unbiased samples we include, the larger the relative effect LfF has over vanilla ERM. As expected, if we change the bias cue to digit and the task cue to color, LfF fails.

\begin{information}{Changing the task on Colored MNIST}
If color were the task and we evaluated LfF on Colored MNIST, we would see a drop in accuracy, as color is learned first, not digit. Thus, compared to the vanilla baseline, the final model generalization performance can verify whether the biased model learned the bias cue and whether what was learned was indeed a bias cue.
\end{information}

\subsubsection{Ingredients for LfF}

In LfF, we use the usual ingredients for supervised learning (\(\text{samples for } (X, Y) = (\text{input}, \text{output})\)) plus an additional assumption:
\[\text{Biased samples are the ones that the intentionally biased model learns first.}\]
Simply put: the bias is the simplest cue out of the ones with high predictive performance on this biased dataset. This is sometimes true, sometimes not. However, whenever it \emph{is} true, we have a great solution for it. It can still happen, however, that the bias is not the easiest cue to learn. Then, the procedure misses the point.

\begin{information}{When is something a ``bias''?}
What is bias is defined by humans. It is not an algorithmic concept. Only when humans declare something as a bias does it become a bias. It depends on the task (i.e., the setting we wish to generalize to) that humans specify. Whatever is not the task is a potential bias. Once we have a fixed task, we identify biases by, e.g., performing counterfactual evaluation.
\end{information}

\begin{information}{Possible Extension of LfF}
In the first few epochs, we could already condition the intentionally biased model to look for parameter regions where there are a lot more correct solutions with a bit more complex cues. This is already achieved in a way for regular LfF: when a very simple cue results in very poor training performance, it will not be chosen, no matter how simple it is.
\end{information}

\subsection{ReBias: Representational regularization}
\label{ssec:rebias}

\begin{figure}
    \centering
    \includegraphics[width=0.8\linewidth]{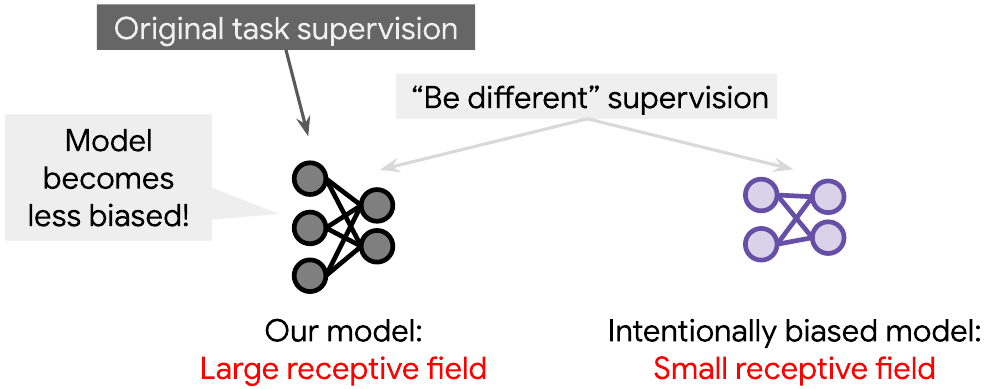}
    \caption{High-level and informal overview of the ReBias method. The intentionally biased model has a small receptive field to amplify texture bias. The debiased model is encouraged to be different from the intentionally biased one.}
    \label{fig:rebias}
\end{figure}

Another method which introduces a similar concept to LfF is``\href{https://arxiv.org/abs/1910.02806}{Learning De-biased Representations with Biased Representations}''~\cite{https://doi.org/10.48550/arxiv.1910.02806}. An intuitive overview is given in Figure~\ref{fig:rebias}. The paper considers texture bias as the key problem to solve. We build CNNs that are \emph{intentionally biased} towards texture by reducing their receptive fields. By constraining the intentionally biased model to this architecture, it is forced to capture local cues like texture.

The \emph{final model} has a large receptive field. It might be, e.g., a ResNet-50. The \emph{intentionally biased model} has a small receptive field, like the BagNet~\cite{https://doi.org/10.48550/arxiv.1904.00760} model. A large receptive field can capture both local and global cues. However, the model might not look at global cues if the dataset is structured so that the net can simply learn very local cues to perform well.

\begin{information}{Receptive Fields Beyond the Input Image}
We usually use padding to have the kernel centered at every pixel and influence the output dimensionality. If we use padding and regular (e.g., \(3 \times 3\)) convolutions, the receptive field of a deeper layer can be even beyond the image (but there, neurons only output zeros, constants, mirrors, or other redundant values). The field of view is huge in this case.
\end{information}

How can we perform ``be different'' supervision in this setup? The ReBias method leverages \emph{statistical independence} instead of giving specific weights to samples. We train a debiased representation by encouraging the final model's outputs to be statistically independent from the intentionally biased model's outputs. We measure this independence with the Hilbert-Schmidt Independence Criterion (HSIC) between two random variables \(U, V\):
\[\operatorname{HSIC}^{k, l}(U, V) = \Vert C_{UV}^{k, l} \Vert_{\mathrm{HS}}^2\]
where \(C\) is the cross-covariance operator in the Reproducing Kernel Hilbert Space (RKHS) corresponding to kernels \(k\) and \(l\), and \(\Vert \cdot \Vert_{\mathrm{HS}}\) is the Hilbert-Schmidt norm which is, intuitively, a ``non-linear version of the Frobenius norm of an infinite-dimensional covariance matrix.'' Kernels \(k\) and \(l\) correspond to random variables \(U\) and \(V\), respectively. Essentially, we embed \(U\) and \(V\) in the infinite-dimensional RKHS corresponding to the kernels \(k\) and \(l\), and compute their covariance there.

We use this criterion to make the invariances learned by these two models different. Our ``be different'' supervision is to minimize the HSIC between the two models.

\textbf{Important property}: It is well known~\cite{https://doi.org/10.48550/arxiv.1910.02806} that for two random variables \(U, V\) and RBF kernels \(k, l\),
\[\operatorname{HSIC}^{k, l}(U, V) = 0 \iff U \indep V.\]

\begin{information}{Why is HSIC needed?}
Why is making \(U\) and \(V\) uncorrelated not enough? If we have a covariance matrix and try to make it the identity matrix, we can enforce the correlation between the variables to be 0, but they will not necessarily be independent. There can be higher-order, non-linear dependencies. However, the HSIC lifts our random variables to an infinite-dimensional Hilbert space, and we consider the covariance ``matrix'' there. By doing so, we remove higher-order dependencies too at the same time, making the two variables truly independent.
\end{information}

If we just train a model \(f\) on some image classification dataset, it is very likely that the model finds a solution that is also representable by the small receptive field network \(g\), as the model can usually perform well by looking at very small patches for predictions and we have previously discussed the simplicity bias of DNNs. Therefore, for our final model \(f\) and the intentionally biased model \(g\), we want to enforce statistical independence \(f(X) \indep g(X)\) (that are random variables in \(\nR^C\)) to ensure that the model \(f\) we find is not equivalent to some other network \(g\) with a small receptive field. The paper uses a finite-sample unbiased estimator \(\operatorname{HSIC}^{k}_1(f(X), g(X))\) and the authors choose \(k\) and \(l\) to be both RBF kernels. Therefore, we consider the shorthand \(\operatorname{HSIC}_1(f, g)\).

We know that
\begin{align*}
    \operatorname{HSIC}(f(X), g(X)) = 0 &\iff \text{\(f(X)\) and \(g(X)\) are independent}\\
    &\iff \text{The models \(f, g\) have ``orthogonal invariances''.}
\end{align*}
Let us detail the last equality further. If \(g\) discriminates color (i.e., its decision boundary separates objects of different colors), then \(f\) should learn invariance for color (i.e., changing of object color does not influence the distance \wrt decision boundary of \(f\)), and vice versa: if \(g\) is treating two samples similarly, then \(f\) should consider these far away from each other in the feature representation.\footnote{This is somewhat like a metric learning objective for HSIC.} We train a de-biased representation by encouraging our model to be statistically independent of the intentionally biased representation.


\subsubsection{ReBias Optimization Problem}

The optimization problem in ReBias is
\[\argmin_{g \in G} \underbrace{\cL(g, x, y)}_{\text{Original task loss}} - \lambda_g \underbrace{\operatorname{HSIC}_1(f(x), g(x))}_{\text{Minimize independence}}\]
for the intentionally biased model and
\[\argmin_{f \in F} \underbrace{\cL(f, x, y)}_{\text{Original task loss}} + \lambda_f \underbrace{\operatorname{HSIC}_1(f(x), g(x))}_{\text{Maximize independence}}\]
for our model. The minimax game being solved is thus
\[\min_{f \in F}\max_{g \in G} \cL(f) - \cL(g) + \lambda \operatorname{HSIC}_1(f, g).\]
During training, we update \(f\) once, then update \(g\) for a fixed \(f\) \(n\) times (\(n = 1\) in the \href{https://github.com/clovaai/rebias/blob/master/trainer.py#L115}{official implementation}). There are many other options, e.g., training \(f\) and \(g\) together on the same loss value.

\subsubsection{Illustration of Training ReBias}

\begin{figure}
    \centering
    \includegraphics[width=\linewidth]{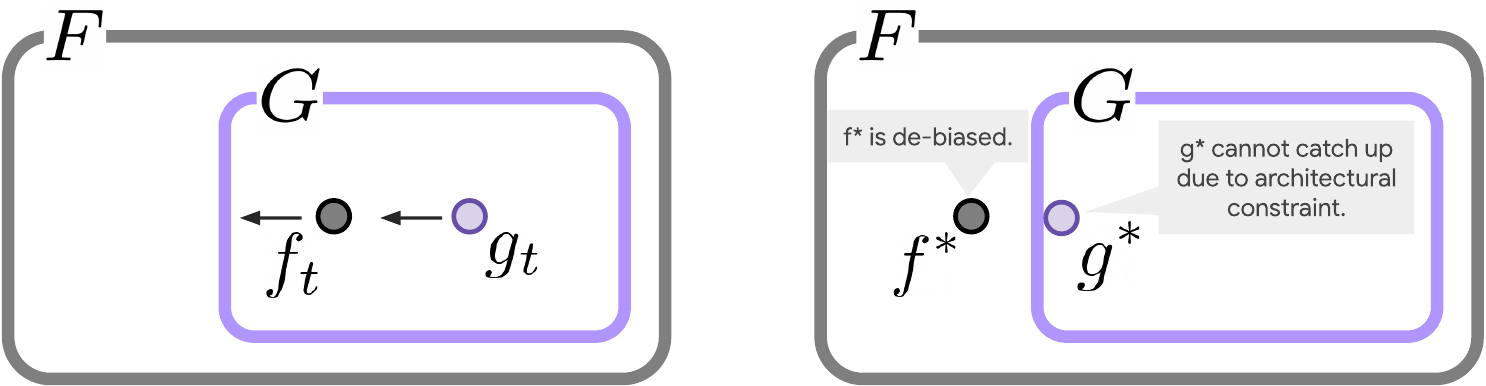}
    \caption{\emph{Left.} Illustration of the ReBias minimax optimization problem. The function \(f\) is optimized to be highly different from \(g\) while still solving the task. The function \(g\) is incentivized to stay as similar to \(f\) as possible. \emph{Right.} The optimal, de-biased function \(f^*\) leaves hypothesis space \(G\). Therefore, no function exists in \(G\) that can match \(f^*\).}
    \label{fig:rebias_t}
\end{figure}

The training procedure is illustrated in Figure~\ref{fig:rebias_t}. Functions \(f\) and \(g\) are elements of function spaces \(F\) and \(G\), respectively. The function \(g\) is architecturally constrained and we have \(G \subset F\). (We can pad kernels of \(g\) by zeros to get a valid model \(f \in F\) that simulates a model with a small receptive field.) During the optimization procedure, \(g\) tries to catch up to \(f\) (solve the task and maximize dependence). In turn, \(f\) tries to be different (run away) from \(g\) (solve the task and minimize dependence). Eventually, after doing this for a few iterations, \(f\) finally escapes the set of models \(G\). Thus, no function in \(G\) can represent \(f\) anymore (due to the architectural constraint), and \(f\) cannot leverage the simple cue that \(g\) uses. Now, e.g., \(f\) looks at global shapes instead of texture: \(f\) becomes debiased.

\begin{figure}
    \centering
    \includegraphics[width=0.6\linewidth]{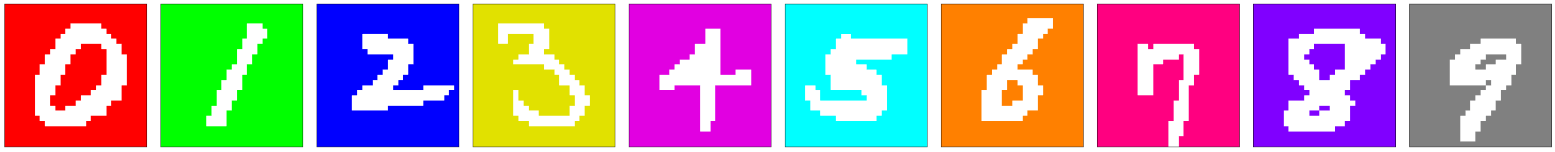}
    \caption{A versatile sample from the Colored MNIST dataset variant used in~\cite{https://doi.org/10.48550/arxiv.1910.02806}, taken from the paper.}
    \label{fig:colormnist2}
\end{figure}

\begin{table*}
\centering
\setlength\tabcolsep{0.4em}
\resizebox{\textwidth}{!}{
\begin{tabular}{ccccccccccccccc}
 & & \multicolumn{6}{c}{Biased} & & \multicolumn{6}{c}{Unbiased} \\
\cline{3-8} \cline{10-15}
& \vspace{-1em} \\
$\rho$ & & {\scriptsize Vanilla} & {\scriptsize Biased} & {\scriptsize\texttt{HEX}} & {\scriptsize\texttt{LearnedMixin}} & {\scriptsize\texttt{RUBi}} & {\scriptsize\texttt{ReBias}} & & {\scriptsize Vanilla} & {\scriptsize Biased} & {\scriptsize\texttt{HEX}} & {\scriptsize\texttt{LearnedMixin}} & {\scriptsize\texttt{RUBi}} & {\scriptsize\texttt{ReBias}} \\
\cline{1-1} \cline{3-8} \cline{10-15}
& \vspace{-1em} \\
\cline{1-1} \cline{3-8} \cline{10-15}
& \vspace{-1em} \\
.999             &  & \textbf{100.}           & \textbf{100.}        & 71.3 & 2.9        & 99.9  & \textbf{100.} &  & 10.4            & 10.  &      10.8       & 12.1       & 13.7 & \textbf{22.7} \\
.997             &  & \textbf{100.}           & \textbf{100.}        & 77.7  & 6.7        & 99.4  & \textbf{100.} &  & 33.4            & 10.     & 16.6        & 50.2       & 43.0 & \textbf{64.2} \\
.995             &  & \textbf{100.}           & \textbf{100.}         & 80.8 & 17.5       & 99.5  & \textbf{100.} &  & 72.1            & 10.     & 19.7         & 78.2       & \textbf{90.4} & 76.0 \\
.990          &  & \textbf{100.}           & \textbf{100.}      & 66.6   & 33.6       & \textbf{100.} & \textbf{100.} &  & 89.1         & 10.   & 24.7           & 88.3       & \textbf{93.6} & 88.1 \\
\cline{1-1} \cline{3-8} \cline{10-15}
& \vspace{-1em} \\
avg.           &  & \textbf{100.}           & \textbf{100.}     & 74.1     & 15.2       & 99.7  & \textbf{100.} &  & 51.2            & 10.    & 18.0          & 57.2       & 60.2 & \textbf{62.7} \\
\cline{1-1} \cline{3-8} \cline{10-15}
& \vspace{-1em} \\
\end{tabular}}
\caption{Results of ReBias on the Colored MNIST dataset, compared to various previous methods we do not cover in the book. ReBias shows notable improvements in some settings, resulting in the best average performance. \(\rho\) controls the fraction of unbiased samples in the training set. The vanilla and biased results show the performance of \(f \in F\) and \(g \in G\), respectively, trained using ERM. The results are taken from the paper~\cite{https://doi.org/10.48550/arxiv.1910.02806}.}
\label{tab:rebiasres1}
\end{table*}

\subsubsection{Results of ReBias}

Let us first consider the results of the method on the Colored MNIST dataset. In Colored MNIST, the color highly (or perfectly) correlates with the digit shape in the training set. Learning color is a shortcut to achieving high accuracy. Naively trained models will be biased towards color because of simplicity bias. The paper uses a variant of Colored MNIST in which all digits are white, but the background colors are perfectly correlated with the digits. A versatile sample from the dataset can be seen in Figure~\ref{fig:colormnist2}. The model we wish to debias is a LeNet architecture that can capture both color and shape. The intentionally biased model is a BagNet architecture that uses \(1 \times 1\) convolutions. This is very much liable to overfit to color. The evaluation is performed both on biased and unbiased test sets. When evaluating the trained model on a test set with bias identical to the training set, we measure ID generalization performance. When using a test set with unbiased samples (colors randomly assigned to samples), the model relying on the bias cue would perform poorly. The exact results are shown in Table~\ref{tab:rebiasres1}. ReBias improves unbiased accuracy while managing to retain biased accuracy. 

Let us now turn to the task of action recognition with a strong static bias. The authors use the Kinetics dataset~\cite{DBLP:journals/corr/abs-1907-06987} for training the model, which has a strong bias towards static cues. For evaluation, the Mimetics dataset~\cite{mimetics} is used that is ripped off the static cues and only contains the pure actions. The model to be debiased is a 3D-ResNet-18~\cite{DBLP:journals/corr/abs-1711-11248} that can capture both temporal and static cues. The intentionally biased model is a 2D-ResNet-18, which can only capture static cues (i.e., cues from individual frames). As the results in Table~\ref{tab:rebiasres2} show, ReBias improves unbiased accuracy while also managing to improve biased accuracy.

\begin{table}
\centering
\caption{Results of ReBias on the Kinetics (biased) and Mimetics (unbiased) datasets, compared to various previous methods we do not cover in the book. Notably, ReBias is the most performant approach on \emph{both} the biased and unbiased datasets. The vanilla and biased results show the performance of \(f \in F\) and \(g \in G\), respectively, trained using ERM. The results are taken from the paper~\cite{https://doi.org/10.48550/arxiv.1910.02806}.}
\label{tab:rebiasres2}
\setlength\tabcolsep{0.4em}
\small
\begin{tabular}{lcccc}
&& Biased     & Unbiased   \\
Model description && (Kinetics) & (Mimetics) \\
\cline{1-1} \cline{3-4}
\vspace{-1em} & \\ 
\cline{1-1} \cline{3-4}
\vspace{-1em} & \\
Vanilla {\scriptsize(\texttt{3D-ResNet18})} &&  54.5 &  18.9    \\  
Biased {\scriptsize(\texttt{2D-ResNet18})} &&  50.7 &  18.4    \\  
\cline{1-1} \cline{3-4}
& \vspace{-1em} \\
{\texttt{LearnedMixin}} {\scriptsize (Clark et al., 2019)} && 12.3 & 11.4 \\
{\texttt{RUBi}} {\scriptsize (Cadene et al., 2019)}  && 22.4  & 13.4 \\
\cline{1-1} \cline{3-4}
& \vspace{-1em} \\
\texttt{ReBias} && \textbf{55.8} & \textbf{22.4} \\ 
\cline{1-1} \cline{3-4}
\end{tabular}
\end{table}

\subsubsection{The Myopic Bias in Machine Learning}

Let us first provide a definition for a \emph{myopic model}.

\begin{definition}{Myopic Model}
A myopic (short-sighted) model in ML refers to a model that is limited in its scope or focus, and, therefore, may not be able to capture all of the relevant features or information needed for robust prediction and decision-making.

\medskip

For example, a myopic model that only looks at texture may not be able to capture other important visual cues such as shape, motion, or context, which can be critical for accurate image recognition or object detection. Similarly, a myopic model that only considers static frames in a video may miss important information conveyed by the temporal dynamics of the video, such as motion or changes over time, which can be critical for accurate action recognition or activity detection. A language model may also focus on word-level cues for the overall sentiment of the sentence (e.g., frequency of `not's).
\end{definition}

The intentionally biased models we are considering in ReBias are myopic. The myopic bias appears a lot in ML in general: A very large model that is capable of modeling all kinds of relationships in the data does not learn complex relationships if the data itself is too simple and very conducive to simple cues.

To avoid myopic models, we introduce a second network that is very myopic, and use ``be different'' supervision, just like in LfF or ReBias. Our model will then be able to leverage complex cues and relationships better.

\textbf{Example}: Considering a language model \(f \in F\) biased to word-level cues, we can ``subtract'' a simple Bag-of-Words (BoW) model (or a simple word embedding) \(g \in G\) from the language model by using ``be different'' supervision to obtain more global reasoning and a more robust model.

\subsubsection{Ingredients of ReBias}
In ReBias, we use the usual ingredients for supervised learning (\(\text{samples for } (X, Y) = (\text{input}, \text{output})\)), plus additional assumptions:
\begin{enumerate}
    \item The bias is ``myopic''.
    \item One can intentionally confine a family of functions to be myopic.
\end{enumerate}
Using ``be different'' supervision by enforcing statistical independence, we aim to obtain unbiased models that leverage robust cues.

\section{Scenario 3 for Selecting the Right Features}

\begin{figure}
    \centering
    \includegraphics[width=\linewidth]{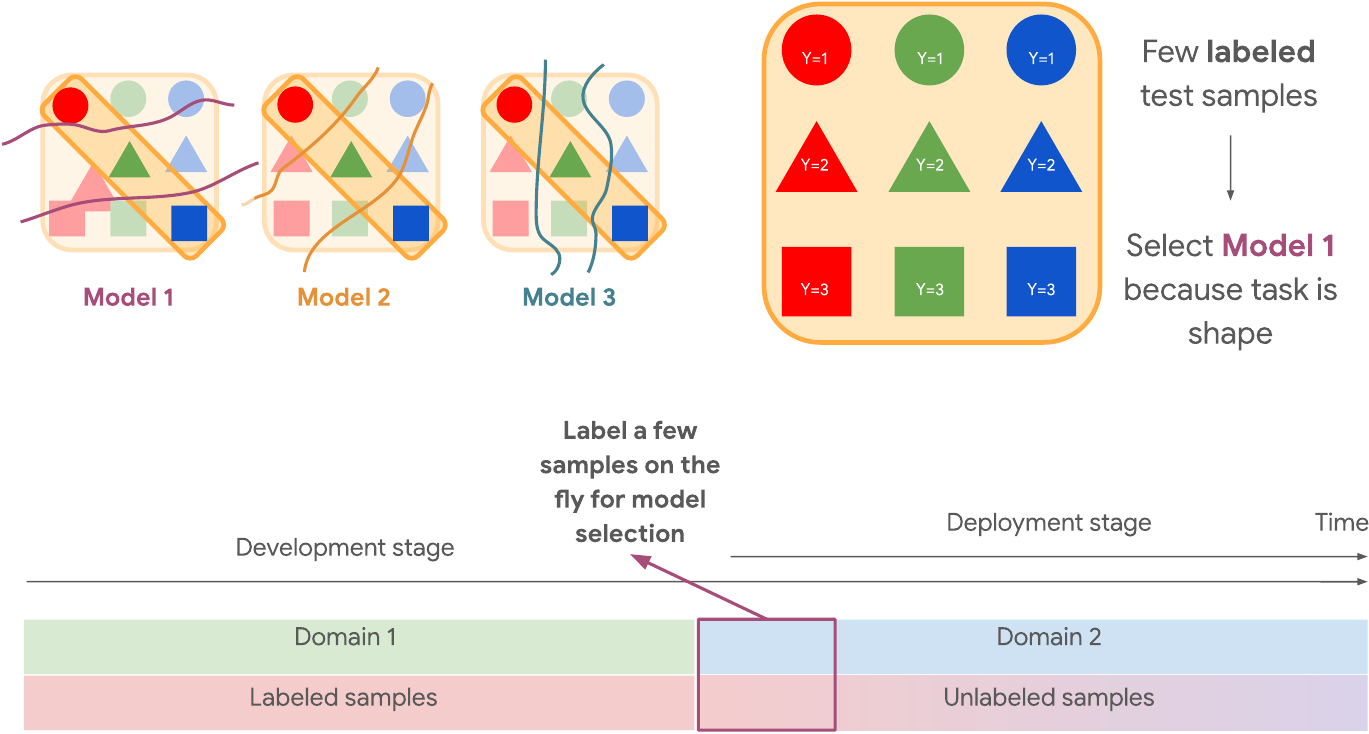}
    \caption{Overview of ``Scenario 3'' for selecting the right features.}
    \label{fig:scenario3}
\end{figure}

The last cross-bias generalization scenario from Figure~\ref{fig:scenarios} we would like to discuss is Scenario 3. A more detailed overview of this setting can be seen in Figure~\ref{fig:scenario3}. Here, we assume biased training samples (a labeled diagonal training set) without bias labels and a few labeled test samples. In such a case, we can train multiple models with diverse OOD behaviors, i.e., that have substantially different decision boundaries in the input space.\footnote{This is possible since the diagonal problem is highly ill-posed and the problem admits a versatile set of solutions.} Considering the shape-color dataset, the decision boundaries do not have to clearly cut any of the human-interpretable cues (predict only based on color vs. predict only based on shape). By having a diverse set of models, we can recognize samples according to many cues that \emph{might be} task cues in deployment. We hope that one of them encodes what we want in the deployment scenario. At deployment time, we choose the right model from this set based on \emph{a few labeled test samples}, then use it during deployment. This corresponds to \emph{domain adaptation} or \emph{test-time training} -- different OOD generalization types where we have access to labeled deployment (test) samples. In practice, this is usually done in the context of \emph{test-time training}, as the models are usually updated through the deployment procedure. By labeling samples on the fly (test-time training), one can perform model selection robustly.

\textbf{Note}: If we have deployment samples that are not unbiased and we also have bias labels, we can still use group DRO, sample weighting, and DANN.

In this scenario, the deployment domain is not necessarily unbiased. It can be equally biased, just in other ways. The labeled test samples decide the \emph{task}. We select the best-performing model on the test dataset (which is usually very small in size), e.g., based on accuracy. 

\begin{information}{Difference between having a few test samples and a few unbiased training samples}
In practice, we are unlikely to have unbiased samples at test time. When we do (e.g., as depicted in Figure~\ref{fig:scenario3}), these scenarios \emph{can} be the same, but there can also be other distributional shifts between train/test. The most likely case is that the deployment scenario contains many biased samples but with biases that differ from the training set biases. In this case, we aim to fine-tune/adapt our model to the specific bias at test time rather than aiming to do well on an unbiased set. Scenario 3 ensures that we can adapt to any shift at deployment (test) time, as we have direct access to deployment-time (test-time) data. This is a more straightforward setting, providing more information about the deployment scenario.
\end{information}

Here is one of the possible \emph{recipes} to deal with such a setting:
\begin{enumerate}
    \item Train an ensemble of models with some ``diversity'' regularization.
    \item At test time, use a few labeled samples or human inspection (if it costs less than annotation time or we have special selection criteria) to select the appropriate model that generalizes well.
\end{enumerate}

This recipe gives rise to two questions:
\begin{enumerate}
    \item How can we know that the samples we base our decision on are representative of the whole test domain as time progresses?
    \item How can we make sure that the set of models uses a diverse set of cues?
\end{enumerate}

For the first question, we have two possible answers:
\begin{itemize}
    \item Adapt the model very frequently (e.g., every batch of data we obtain).
    \item Trust that the deployment distribution is not going to change, e.g., for the next month, and update only every month.
\end{itemize}
By choosing either of the above, we also assume that these few labeled samples are enough to determine the most performant model in the deployment scenario.

For the second question, we cannot give a quick answer. If we naively train \(n\) models separately, all of them will likely focus on easy cues because of the simplicity bias of DNNs. That is why we need explicit regularization to enforce diversity. In the next section, we will focus on one of the methods that do exactly that.

\subsection{Predicting is not Understanding}

To look at one of the methods for diversifying models, let us discuss the paper ``\href{https://arxiv.org/abs/2207.02598}{Predicting is not Understanding: Recognizing and Addressing Underspecification in Machine Learning}''~\cite{https://doi.org/10.48550/arxiv.2207.02598}.
The intuition behind this method is that diverse ensemble training can be achieved by enforcing ``independence'' between models through the orthogonality of input gradients. 

One way to achieve this is to add an orthogonality constraint to the loss.\footnote{HSIC could also be used as an independence criterion.} Such a constraint can be represented as the squared cosine similarity of the input gradients for the same input:
\[\cL_\mathrm{indep}\left(\nabla_x f_{\theta_{m_1}}(x), \nabla_x f_{\theta_{m_2}}(x)\right) = \cos^2\left(\nabla_x f_{\theta_{m_1}}(x), \nabla_x f_{\theta_{m_2}}(x)\right).\]
Our goal is to have orthogonal input gradients. As this constraint is differentiable, we optimize it using Deep Learning (DL). 

\begin{information}{Shape of Gradients}
In the orthogonality constraint, the gradients are of the logits, not of the loss. This results in a 4D tensor for multi-class classification. We simply flatten this tensor and calculate the squared cosine similarity. We only have a 1D output for binary classification, so the gradients will have the same shape as the input image. The paper focuses on binary classification. The independence loss used by the paper requires \(\cO(M^2)\) network evaluations, where \(M\) is the number of models in our diverse set.
\end{information}

\subsubsection{Intuition for orthogonal input gradients}

Suppose that we have two models, \(m_1\) and \(m_2\), and two different regions of the image: background and foreground. If \(m_2\) is looking at the background, there is a significant focus on the background parts in the input gradient. We want the input gradient of \(m_1\) to be orthogonal to that of \(m_2\), as that will result in \(m_1\) focusing more on the foreground.

\subsubsection{Formal reasoning about independence}

We define ``independence'' as the statistical independence of model outputs for a local Gaussian perturbation around every \(x\) in the input space. We measure the change in output for model 1 and model 2 using this Gaussian perturbation. The perturbation is small enough to approximate a model via its linear tangent function (input gradient). For infinitesimally small perturbations (\(\sigma \downarrow 0\)), changes in logits between \(x\) and \(\tilde{x}\) can be approximated through linearization by the input gradients \(\nabla_x f\). In particular, for \(\sigma \downarrow 0\), the relative change in the logits from \(x\) to \(\tilde{x}\) is exactly given by the directional derivative \(\left\langle\nabla_x f(x), \frac{\tilde{x} - x}{\Vert \tilde{x} - x \Vert} \right\rangle\).

Why can we use the orthogonality of the input gradients for measuring statistical independence? It can be shown that the statistical independence of the model outputs is equivalent to the geometrical orthogonality of the input gradients when \(\sigma \downarrow 0\) for the local Gaussian perturbation. The local independence for a particular input \(x\) is defined as
\[f_{\theta_1}(\tilde{x}) \indep f_{\theta_2}(\tilde{x}), \tilde{x} \sim \cN(x, \sigma I) \in \nR^{d_{\mathrm{in}}},\]
and global independence for a particular input \(x\) means that in a set of predictors \(\{f_{\theta_1}, \dots, f_{\theta_M}\}\), all pairs are locally independent around \(x\).

We need one more ingredient to ensure that the models are diverse in meaningful ways. The set of orthogonal models increases exponentially with the input dimensionality. For images, we have overwhelmingly many orthogonal models -- the input space might be close to being 1M-dimensional. However, the relevant subset of images that make sense inside this space is quite low-dimensional. This low-dimensional subset is the \emph{data manifold}. We want to confine our exploration of decision boundaries to the manifold rather than the entire space. The reason is that diversification regularization without on-manifold constraints may result in models that are only diversified in the vast non-data-manifold dimensions, which means that they behave similarly on on-manifold samples.

\begin{figure}
    \centering
    \includegraphics[width=0.6\linewidth]{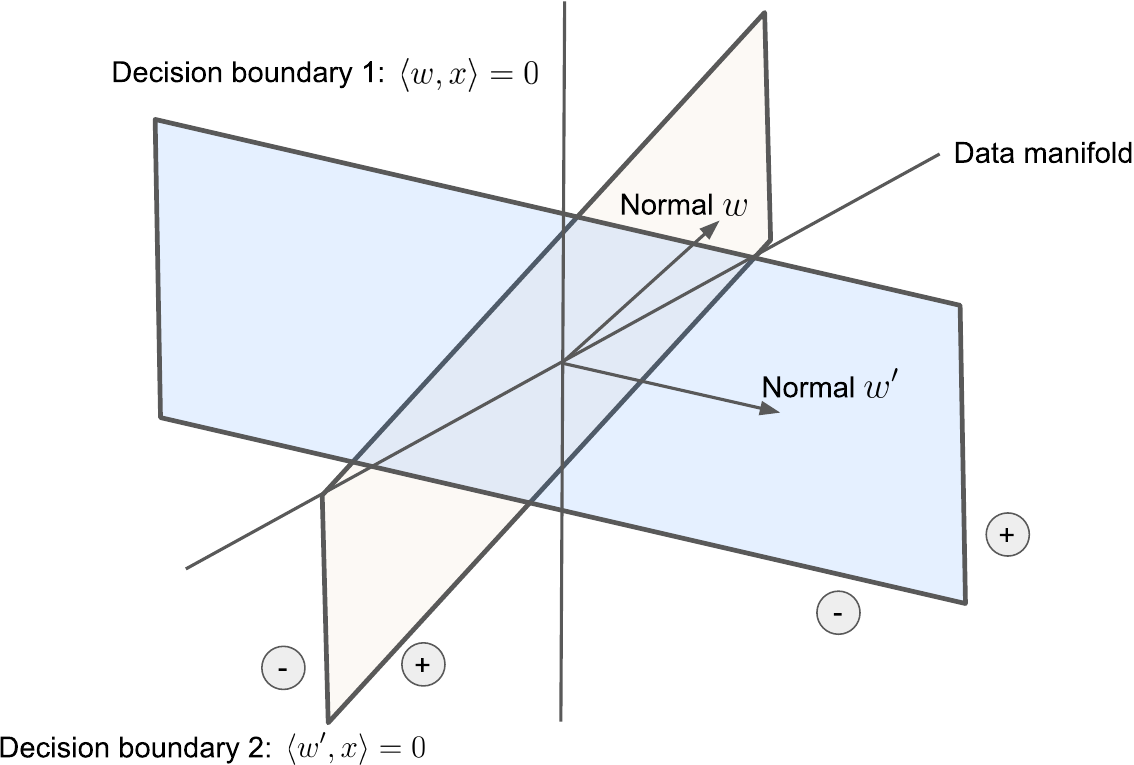}
    \caption{Example that highlights the importance of the on-manifold constraint in Predicting is not Understanding~\cite{https://doi.org/10.48550/arxiv.2207.02598}. We consider a 1D line as our data manifold and a binary classification problem. In the case of a linear classifier, the normal of the decision boundary is exactly the input gradient.  If we project the decision boundaries onto the data manifold, they become identical. This means that even though the weights of the two models are orthogonal, they make identical decisions on the data manifold.}
    \label{fig:onmanifold}
\end{figure}

We visualize an intuitive example of how models with orthogonal input gradients might still behave identically on the data manifold in Figure~\ref{fig:onmanifold}.

\subsubsection{On-Manifold Constraints}

In Predicting is not Understanding, the input gradient is regularized to be ``on'' the data manifold. We use a Variational Autoencoder (VAE)~\cite{https://doi.org/10.48550/arxiv.1312.6114} to learn an approximation of the data distribution from unlabeled samples, i.e., to learn the data manifold \(\cM\). One can then project any vector \(\in \nR^{d_{\mathrm{in}}}\) in the input space onto this data manifold by using the VAE \(\operatorname{proj}_\cM\colon \nR^{d_\mathrm{in}} \times \nR^{d_\mathrm{in}} \rightarrow \cM\). This VAE is trained to be capable of projecting a vector \(v\) (the gradient in the application) to the tangent plane of the manifold at point \(x\). For OOD samples, this means that we want
\[\operatorname{proj}_\cM(x, v) \approx v\quad \forall x \sim P_{\mathrm{OOD}}, x + v \sim P_\mathrm{OOD},\]
which is achieved by training the VAE to reconstruct the OOD images and applying a similar series of transformations to the vector \(v\) as well. Further details can be read in the paper.

The on-manifold constraint is
\[\cL_\mathrm{manifold}(\nabla f(x)) = \Vert \operatorname{proj}_{\cM}(x, \nabla_x f(x)) - \nabla_x f(x) \Vert_2^2,\]
where \(\operatorname{proj}_\cM\) is the projection of the gradient onto the tangent space of the manifold at point \(x\). This loss term forces the input gradient to be aligned with the data manifold.

Used together with the independence constraint, the model is constrained to have orthogonal gradients that are roughly inside the data manifold. Intuitively, when the independence constraint influences a model's gradients in dimensions oriented outwards from the manifold, it does not impact its predictions on natural data. Consequently, models that produce identical predictions on every natural input could satisfy the independence constraint because their decision boundaries are identical when projected onto the manifold. This drastically reduces the search space for new models and ensures that the next model in the ensemble will look at a \emph{meaningful} new cue.

\begin{information}{How to choose the dimensionality of the VAE latent space?}
We do not have to know the dimensionality of the manifold, as it is perfectly fine if we choose the number of models more than that. Adding new models can still lead to more diversity, but it will be impossible to enforce the perfect orthogonality of the input gradients. It is also not a problem if we have fewer dimensions than the actual number of dimensions of the manifold in the VAE latent space, as one can embed higher-dimensional factors of variation into lower dimensions.
\end{information}

\subsubsection{Putting it all together}
\begin{figure}
    \centering
    \includegraphics[width=0.6\linewidth]{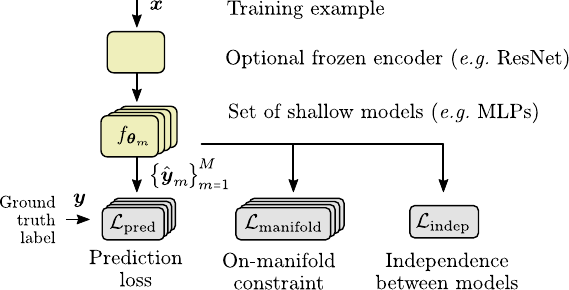}
    \caption{Overview of the Predicting is not Understanding method, taken from~\cite{https://doi.org/10.48550/arxiv.2207.02598}.}
    \label{fig:pinu}
\end{figure}
Our final loss function is
\begin{align*}
\cL(\cD_{\mathrm{tr}}, \theta_1, \dots, \theta_M) = \sum_{(x, y) \in \cD_{\mathrm{tr}}} &\Bigg[\frac{1}{M}\sum_{m = 1}^M \cL_\mathrm{pred}\left(y, \sigma(f_{\theta_m}(x))\right)\\
&+ \frac{1}{M^2} \sum_{m_1 = 1}^M \sum_{m_2 = 1}^M \lambda_{\mathrm{indep}} \cL_\mathrm{indep}\left(\nabla_x f_{\theta_{m_1}}(x), \nabla_x f_{\theta_{m_2}}(x)\right)\\
&+ \frac{1}{M} \sum_{m = 1}^M \lambda_\mathrm{manifold} \cL_{\mathrm{manifold}}\left(\nabla_x f_{\theta_m}(x)\right)\Bigg]
\end{align*}
that encapsulates the prediction losses, the independence losses, and the on-manifold losses for the \(M\) models.

\begin{figure}
    \centering
    \includegraphics[width=0.8\linewidth]{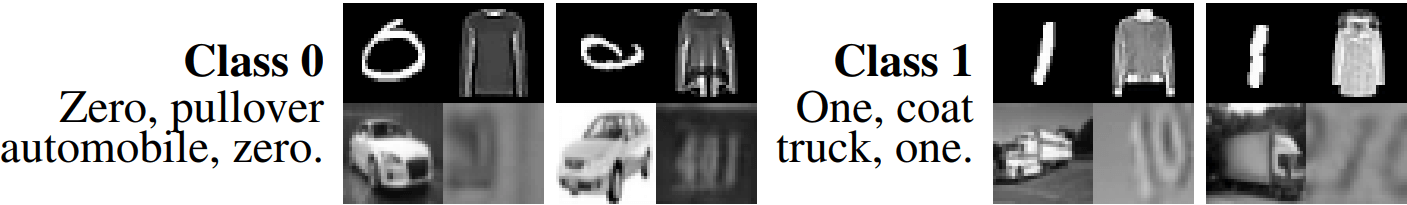}
    \caption{Examples of collages of four tiles in Predicting is not Understanding~\cite{https://doi.org/10.48550/arxiv.2207.02598}.}
    \label{fig:collage}
\end{figure}

\begin{table}
  \renewcommand{\tabcolsep}{0.50em}
  \renewcommand{\arraystretch}{1.0}
  \centering
  \caption{Results of Predicting is not Understanding~\cite{https://doi.org/10.48550/arxiv.2207.02598}. The ERM baseline only learns to look at the MNIST tile and performs random prediction for all other cues. The independence loss is also not enough by itself. \textbf{Fine-tuning}: After training a set of models, the authors remove the independence and on-manifold constraints and fine-tune the models by applying a binary mask on the pixels/channels such that each model is fine-tuned only on the parts of the image most relevant to themselves (as measured by the magnitude of the gradient \(\nabla_{\theta_m} f_{\theta_m}(x)\) among the models for each pixel/channel). \textbf{Pairwise combinations}: After training and fine-tuning a set of models, they combine the best of them (as given by our metric of choice on the OOD validation set) into a global one that uses all of the most relevant features. They train this global model from scratch, without regularizers, on masked data, using masks from the selected models combined with a logical OR. They repeat this pairwise combination as long as the accuracy of the global model increases. They always append the new models into the set of models. Independence + on-manifold constraints + VAE + FT + pairwise combinations (3x) performs best, achieving almost the upper bound (training on test-domain data). The upper bound accuracy can be achieved, e.g., if we have four models specializing perfectly in different quadrants. Based on these results, the models are indeed very diverse.}
  \label{tab:pinu_res}
  \begin{tabularx}{\linewidth}{Xccccc}
  \toprule
  \textbf{Collages} dataset (accuracy in \%) & \multicolumn{4}{c}{Best model on} \vspace{1.5pt}\\ \cline{2-5}
  ~ &  \rotatebox{90}{MNIST} & \rotatebox{90}{SVHN~~} & \rotatebox{90}{Fashion\phantom{e}} & \rotatebox{90}{CIFAR-10\phantom{e}} & \rotatebox{90}{Average}\\
  \midrule
  Upper bound (training on test-domain data) & 99.9 & 92.4 & 80.8 & 68.6 & 85.5 \\
  \midrule
  ERM Baseline & 99.8 & 50.0 & 50.0 & 50.0 & 62.5 \\
  Spectral decoupling~\cite{https://doi.org/10.48550/arxiv.2011.09468} & 99.9 & 49.8 & 50.6 & 49.9 & 62.5 \\ 
  With penalty on L1 norm of gradients & 98.5 & 49.6 & 50.5 & 50.0 & 62.1 \\ 
  With penalty on L2 norm of gradients~\cite{https://doi.org/10.48550/arxiv.1908.02729} & 96.6 & 52.1 & 52.3 & 54.3 & 63.8 \\ 
  Input dropout (best ratio: 0.9) & 97.4 & 50.7 & 56.1 & 52.1 & 64.1 \\ 
  Independence loss (cosine similarity) \cite{https://doi.org/10.48550/arxiv.1911.01291} & 99.7 & 50.4 & 51.5 & 50.2 & 63.0 \\ 
  Independence loss (dot product) \cite{teney2021evading} & 99.5 & 53.5 & 53.3 & 50.5 & 64.2 \\ 
  \midrule
  With {many more} models\\
  Independence loss (cosine similarity), \underline{1024} models & 99.5 & 58.1 & 66.8 & 63.0 & 71.9 \\ 
  Independence loss (dot product), \underline{128} models & 98.7 & 84.9 & 71.6 & 61.5 & 79.2 \\ 
  \midrule
  Proposed method (only 8 models)\\
  Independence + on-manifold constraints, PCA & 97.3 & 69.8 & 62.2 & 60.0 & 72.3 \\ 
  Independence + on-manifold constraints, VAE ($^\ast$) & 96.5 & 85.1 & 61.1 & 62.1 & 76.2 \\ 
($^\ast$) ~+~ FT ~(fine-tuning) & 99.7 & 90.9 & 81.4 & 67.4 & 84.8 \\ 
  ($^\ast$) ~+~ FT ~+~ pairwise combinations (1$\times$) & 99.9 & 92.2 & 79.3 & 66.3 & 84.4 \\ 
  ($^\ast$) ~+~ FT ~+~ pairwise combinations (2$\times$) & 99.9 & 92.5 & 80.2 & 67.5 & 85.0 \\ 
  \textbf{($^\ast$) ~+~ FT ~+~ pairwise combinations (3$\times$)} & \textbf{99.9} & \textbf{92.3} & \textbf{80.8} & \textbf{68.5} & \textbf{85.4} \\ 
  \bottomrule
  \end{tabularx}
\end{table}

\subsubsection{Results of Predicting is not Understanding}

The method is evaluated on a \emph{collage dataset} with controllable correlation among the four collage images. The four datasets used are MNIST, Fashion MNIST~\cite{https://doi.org/10.48550/arxiv.1708.07747}, CIFAR~\cite{krizhevsky2009learning}, and SVHN~\cite{37648}. We have ten classes for each dataset. We put one sample from each dataset in a window of four elements, as shown in Figure~\ref{fig:collage}. During training, a dataset with a perfect correlation between the four labels is used (e.g., the meta-class 0 is the quadruple (zero, pullover, automobile, zero)). This is a biased, diagonal dataset. For evaluation, we use a dataset with no correlation between the four labels. This is an unbiased dataset with off-diagonal samples as well. We label this \wrt, e.g., MNIST or CIFAR, and -- based on the results -- we find out what cue (which quadrant) each model learned.

This is \emph{cross-bias generalization}: At test time, we break the correlation among the four quadrants. Our expectation is that by training independent models, we should be able to get four different models that look at different quadrants. The results are shown in Table~\ref{tab:pinu_res}. To measure performance, the authors perform test-time oracle model selection: they are using test-time information by choosing the best model on each test-time dataset. A possible justification of test-time oracle model selection is that, in practice, we have a few labeled samples at test time to select the most performant model. These reported numbers show the upper bound on the performance in the scenario above because they are chosen based on the entire test set, not just a few samples.

\subsubsection{Ingredients for Predicting is not Understanding}

The Predicting is not Understanding method uses the usual ingredients for supervised learning (\(\text{samples for } (X, Y) = (\text{input}, \text{output})\)) to find a set of diverse hypotheses $m_1, \dots, m_N$. The model selection takes place at \emph{test time}. This work only shows the upper bound of attainable performance using the perfect test-time model selection.

\begin{definition}{Kullback-Leibler Divergence}
The Kullback-Leibler divergence from distribution \(Q\) to distribution \(P\) with densities \(q, p\) is given by
\[\operatorname{KL}\left(P \Vert Q\right) = \int_\cX p(x) \log \frac{p(x)}{q(x)}\ dx.\]
\end{definition}

\begin{information}{Independence and Input Gradients}
\begin{proposition}
A pair of predictors \(f_{\theta_1}, f_{\theta_2}\) are locally independent at \(x\) iff the mutual information \(\operatorname{MI}(f_{\theta_1}(\tilde{x}), f_{\theta_2}(\tilde{x})) = 0\) with \(\tilde{x} \sim \cN(x, \sigma^2I)\)~\cite{https://doi.org/10.48550/arxiv.2207.02598}.
\end{proposition}

\medskip

\begin{proof}
A pair of predictors \(f_{\theta_1}, f_{\theta_2}\) are defined to be locally independent at x iff their predictions are statistically independent for Gaussian perturbations around \(x\):
\[f_{\theta_1}(\tilde{x}) \indep f_{\theta_2}(\tilde{x})\]
with \(\tilde{x} \sim \cN(x, \sigma^2I)\).

The definition of mutual information is
\[\operatorname{MI}(f_{\theta_1}(\tilde{x}), f_{\theta_2}(\tilde{x})) = D_{\operatorname{KL}}(P_{f_{\theta_1}(\tilde{x}), f_{\theta_2}(\tilde{x})} \Vert P_{f_{\theta_1}(\tilde{x})} \otimes P_{f_{\theta_2}(\tilde{x})}).\]
It is a well-known fact that \(D_{\operatorname{KL}}(P \Vert Q) = 0 \iff P \equiv Q\).
From this, we immediately see that
\[\operatorname{MI}(f_{\theta_1}(\tilde{x}), f_{\theta_2}(\tilde{x})) = 0 \iff f_{\theta_1}(\tilde{x}) \indep f_{\theta_2}(\tilde{x}).\]
\end{proof}

For infinitesimally small perturbations (\(\sigma \downarrow 0\)), the variables \(f_{\theta_1}(\tilde{x}), f_{\theta_2}(\tilde{x})\) can be approximated through linearization wrt. the input gradients \(\nabla_x f\):
\[f(\tilde{x}) \approx f(x) + \nabla_x f(x)^\top (\tilde{x} - x) =: \hat{f}(\tilde{x}).\]

\begin{claim}
Following the above definition, \(\hat{f}_{\theta_1}(\tilde{x})\) and \(\hat{f}_{\theta_2}(\tilde{x})\) are 1D Gaussian random variables.
\end{claim}

\medskip

\begin{proof}
By definition, \(\hat{f}(\tilde{x}) = f(x) + \nabla_x f(x)^\top (\tilde{x} - x) = \underbrace{\nabla_x f(x)^\top}_{A :=}\tilde{x} + \underbrace{f(x) - \nabla_x f(x)^\top x}_{b :=}\).

As \(\tilde{x} \sim \cN(x, \sigma^2I)\), we know that
\begin{align*}
\hat{f}(\tilde{x}) = A\tilde{x} + b &\sim \cN(Ax + b, \sigma^2 AA^\top)\\
&\sim \cN(Ax + f(x) - Ax, \sigma^2 AA^\top)\\
&\sim \cN(f(x), \sigma^2 \nabla_x f(x)^\top \nabla_x f(x)).
\end{align*}
Substituting \(f_{\theta_1}\) or \(f_{\theta_2}\) into \(f\) directly gives the statement.
\end{proof}

\medskip

\begin{claim}
The correlation of \(\hat{f}_{\theta_1}(\tilde{x})\) and \(\hat{f}_{\theta_2}(\tilde{x})\) is given by \(\cos(\nabla_x f_{\theta_1}(x), \nabla_x f_{\theta_2}(x))\).
\end{claim}

\medskip

\begin{proof}
We know from before that
\begin{align*}
\hat{f}_{\theta_1}(\tilde{x}) &\sim \cN(f_{\theta_1}(x), \sigma^2 \nabla_x f_{\theta_1}(x)^\top \nabla_x f_{\theta_1}(x))\\
\hat{f}_{\theta_2}(\tilde{x}) &\sim \cN(f_{\theta_2}(x), \sigma^2 \nabla_x f_{\theta_2}(x)^\top \nabla_x f_{\theta_2}(x)).
\end{align*}
It follows that
\begin{align*}
\rho(\hat{f}_{\theta_1}(\tilde{x}), \hat{f}_{\theta_2}(\tilde{x})) &= \frac{\Cov(\hat{f}_{\theta_1}(\tilde{x}), \hat{f}_{\theta_2}(\tilde{x}))}{\sqrt{\sigma^4 \nabla_x f_{\theta_1}(x)^\top \nabla_x f_{\theta_1}(x)\nabla_x f_{\theta_2}(x)^\top \nabla_x f_{\theta_2}(x)}}.
\end{align*}
First, we calculate the covariance:
\begin{align*}
\Cov(\hat{f}_{\theta_1}(\tilde{x}), \hat{f}_{\theta_2}(\tilde{x})) &= \nE[\hat{f}_{\theta_1}(\tilde{x})\hat{f}_{\theta_2}(\tilde{x})] - \nE[\hat{f}_{\theta_1}(\tilde{x})]\nE[\hat{f}_{\theta_2}(\tilde{x})]\\
&= \nE[\hat{f}_{\theta_1}(\tilde{x})\hat{f}_{\theta_2}(\tilde{x})] - f_{\theta_1}(x)f_{\theta_2}(x)\\
&=\begin{multlined}[t] \nE[(f_{\theta_1}(x) + \nabla_x f_{\theta_1}(x)^\top (\tilde{x} - x))(f_{\theta_2}(x) + \nabla_x f_{\theta_2}(x)^\top (\tilde{x} - x))] \\- f_{\theta_1}(x)f_{\theta_2}(x)\end{multlined}\\
&= f_{\theta_1}(x)f_{\theta_2}(x) + f_{\theta_1}(x)\nE[\nabla_x f_{\theta_2}(x)^\top (\tilde{x} - x)] + f_{\theta_2}(x) \nE[\nabla_x f_{\theta_1}(x)^\top (\tilde{x} - x)]\\
&\quad+ \nE[\nabla_x f_{\theta_1}(x)^\top (\tilde{x} - x)\nabla_x f_{\theta_2}(x)^\top (\tilde{x} - x)] - f_{\theta_1}(x)f_{\theta_2}(x)\\
&= \nE[\nabla_x f_{\theta_1}(x)^\top (\tilde{x} - x)\nabla_x f_{\theta_2}(x)^\top (\tilde{x} - x)]\\
&= \nabla_x f_{\theta_1}(x)^\top \nE[(\tilde{x} - x)(\tilde{x} - x)^\top] \nabla_x f_{\theta_2}(x)\\
&= \nabla_x f_{\theta_1}(x)^\top \left(\Cov(\tilde{x} - x) + \nE[(\tilde{x} - x)]\nE[(\tilde{x} - x)]^\top\right) \nabla_x f_{\theta_2}(x)\\
&= \nabla_x f_{\theta_1}(x)^\top (\sigma^2I) \nabla_x f_{\theta_2}(x)\\
&= \sigma^2 \nabla_x f_{\theta_1}(x)^\top \nabla_x f_{\theta_2}(x).
\end{align*}
Plugging this back into the correlation formula, we obtain
\begin{align*}
\rho(\hat{f}_{\theta_1}(\tilde{x}), \hat{f}_{\theta_2}(\tilde{x})) &= \frac{\Cov(\hat{f}_{\theta_1}(\tilde{x}), \hat{f}_{\theta_2}(\tilde{x}))}{\sqrt{\sigma^4 \nabla_x f_{\theta_1}(x)^\top \nabla_x f_{\theta_1}(x)\nabla_x f_{\theta_2}(x)^\top \nabla_x f_{\theta_2}(x)}}\\
&= \frac{\sigma^2 \nabla_x f_{\theta_1}(x)^\top \nabla_x f_{\theta_2}(x)}{\sigma^2 \sqrt{\nabla_x f_{\theta_1}(x)^\top \nabla_x f_{\theta_1}(x)}\sqrt{\nabla_x f_{\theta_2}(x)^\top \nabla_x f_{\theta_2}(x)}}\\
&= \frac{\nabla_x f_{\theta_1}(x)^\top \nabla_x f_{\theta_2}(x)}{\sqrt{\nabla_x f_{\theta_1}(x)^\top \nabla_x f_{\theta_1}(x)}\sqrt{\nabla_x f_{\theta_2}(x)^\top \nabla_x f_{\theta_2}(x)}}\\
&= \cos(\nabla_x f_{\theta_1}(x), \nabla_x f_{\theta_2}(x)).
\end{align*}
\end{proof}

\begin{claim}
The mutual information of \(\hat{f}_{\theta_1}(x)\) and \(\hat{f}_{\theta_2}(x)\) is given by \(-\frac{1}{2} \log(1 -\cos^2(\nabla_x f_{\theta_1}(x), \nabla_x f_{\theta_2}(x)))\).
\end{claim}

\medskip

\begin{proof}
\begin{align*}
\operatorname{MI}(\hat{f}_{\theta_1}(\tilde{x}), \hat{f}_{\theta_2}(\tilde{x})) &= D_{\operatorname{KL}}(P_{\hat{f}_{\theta_1}(\tilde{x}), \hat{f}_{\theta_2}(\tilde{x})} \Vert P_{\hat{f}_{\theta_1}(\tilde{x})} \otimes P_{\hat{f}_{\theta_2}(\tilde{x})})\\
&= \int_{x_1 \in \cX} \int_{x_2 \in \cX} p(\hat{f}_{\theta_1}(x_1), \hat{f}_{\theta_2}(x_2)) \log \frac{p(\hat{f}_{\theta_1}(x_1), \hat{f}_{\theta_2}(x_2))}{p(\hat{f}_{\theta_1}(x_1))p(\hat{f}_{\theta_2}(x_2))}dx_2dx_1\\
&= \nH(\hat{f}_{\theta_1}(\tilde{x})) + \nH(\hat{f}_{\theta_2}(\tilde{x})) - \nH(\hat{f}_{\theta_1}(\tilde{x}), \hat{f}_{\theta_2}(\tilde{x})).
\end{align*}
Now we notice that \(\hat{f}_{\theta_1}(\tilde{x})\) and \(\hat{f}_{\theta_2}(\tilde{x})\) are also \emph{jointly} Gaussian:
\begin{align*}
\begin{pmatrix}\hat{f}_{\theta_1}(\tilde{x}) \\ \hat{f}_{\theta_2}(\tilde{x})\end{pmatrix} &\sim \cN\left(\begin{pmatrix} \mu_{\hat{f}_{\theta_1}(\tilde{x})} \\ \mu_{\hat{f}_{\theta_2}(\tilde{x})}\end{pmatrix}, \begin{bmatrix}\Var(\hat{f}_{\theta_1}(\tilde{x})) & \Cov(\hat{f}_{\theta_1}(\tilde{x}), \hat{f}_{\theta_2}(\tilde{x})) \\ \Cov(\hat{f}_{\theta_1}(\tilde{x}), \hat{f}_{\theta_2}(\tilde{x})) & \Var(\hat{f}_{\theta_2}(\tilde{x}))\end{bmatrix}\right)\\
&\sim \cN\left(\begin{pmatrix}f_{\theta_1}(x) \\ f_{\theta_2}(x)\end{pmatrix}, \begin{bmatrix}\sigma^2 \nabla_x f_{\theta_1}(x)^\top \nabla_x f_{\theta_1}(x) & \sigma^2 \nabla_x f_{\theta_1}(x)^\top \nabla_x f_{\theta_2}(x) \\ \sigma^2 \nabla_x f_{\theta_1}(x)^\top \nabla_x f_{\theta_2}(x) & \sigma^2 \nabla_x f_{\theta_2}(x)^\top \nabla_x f_{\theta_2}(x)\end{bmatrix}\right).
\end{align*}
Below, we derive the formula for the entropy of a multivariate Gaussian \(x \sim \cN(\mu, \Sigma) \in \nR^n\):
\begin{align*}
\nH(x) &= -\int p(x) \log p(x) dx\\
&= -\nE_x[\log \cN(x \mid \mu, \Sigma)]\\
&= -\nE_x\left[\log \left(\frac{1}{(2\pi)^{n/2}|\Sigma|^{1/2}} \exp\left(-\frac{1}{2}(x - \mu)^\top \Sigma^{-1}(x - \mu)\right)\right)\right]\\
&= \nE_x \left[\frac{n}{2} \log(2\pi) + \frac{1}{2}\log |\Sigma| + \frac{1}{2}(x - \mu)^\top \Sigma^{-1}(x - \mu)\right]\\
&= \frac{n}{2}\log(2\pi) + \frac{1}{2}\log |\Sigma| + \frac{1}{2}\nE_x[(x - \mu)^\top \Sigma^{-1} (x - \mu)]\\
&= \frac{n}{2}\log(2\pi) + \frac{1}{2}\log |\Sigma| + \frac{1}{2}\nE_x[\operatorname{tr}((x - \mu)^\top \Sigma^{-1} (x - \mu))]\\
&= \frac{n}{2}\log(2\pi) + \frac{1}{2}\log |\Sigma| + \frac{1}{2}\nE_x[\operatorname{tr}(\Sigma^{-1} (x - \mu)(x - \mu)^\top)]\\
&= \frac{n}{2}\log(2\pi) + \frac{1}{2}\log |\Sigma| + \frac{1}{2}\operatorname{tr}(\Sigma^{-1} \underbrace{\nE_x[(x - \mu)(x - \mu)^\top]}_{\Sigma})\\
&= \frac{n}{2}(1 + \log(2 \pi)) + \frac{1}{2} \log |\Sigma|.
\end{align*}
Finally, we plug this into our formula for the mutual information:
\begin{align*}
\operatorname{MI}(\hat{f}_{\theta_1}(\tilde{x})&, \hat{f}_{\theta_2}(\tilde{x})) = \nH(\hat{f}_{\theta_1}(\tilde{x})) + \nH(\hat{f}_{\theta_2}(\tilde{x})) - \nH(\hat{f}_{\theta_1}(\tilde{x}), \hat{f}_{\theta_2}(\tilde{x}))\\
&= \frac{1}{2}(1 + \log (2\pi)) + \frac{1}{2} \log \left(\sigma^2 \nabla_x f_{\theta_1}(x)^\top \nabla_x f_{\theta_1}(x)\right)\\
&\quad+ \frac{1}{2}(1 + \log (2\pi)) + \frac{1}{2} \log \left(\sigma^2 \nabla_x f_{\theta_2}(x)^\top \nabla_x f_{\theta_2}(x)\right)\\
&\quad- 1 - \log(2\pi) - \frac{1}{2} \log(\sigma^2 \nabla_x f_{\theta_1}(x)^\top \nabla_x f_{\theta_1}(x) \cdot \sigma^2 \nabla_x f_{\theta_2}(x)^\top \nabla_x f_{\theta_2}(x)\\
&\hspace{9.67em}- \sigma^2 \nabla_x f_{\theta_1}(x)^\top \nabla_x f_{\theta_2}(x) \cdot \sigma^2 \nabla_x f_{\theta_1}(x)^\top \nabla_x f_{\theta_2}(x))\\
&= \frac{1}{2}\log \frac{\sigma^4 \nabla_x f_{\theta_1}(x)^\top \nabla_x f_{\theta_1}(x) \cdot \nabla_x f_{\theta_2}(x)^\top \nabla_x f_{\theta_2}(x)}{\sigma^4 \left(\nabla_x f_{\theta_1}(x)^\top \nabla_x f_{\theta_1}(x) \cdot \nabla_x f_{\theta_2}(x)^\top \nabla_x f_{\theta_2}(x) - \left(\nabla_x f_{\theta_1}(x)^\top \nabla_x f_{\theta_2}(x)\right)^2\right)}\\
&= -\frac{1}{2}\log \left(1 - \frac{\left(\nabla_x f_{\theta_1}(x)^\top \nabla_x f_{\theta_2}(x)\right)^2}{\nabla_x f_{\theta_1}(x)^\top \nabla_x f_{\theta_1}(x) \cdot \nabla_x f_{\theta_2}(x)^\top \nabla_x f_{\theta_2}(x)}\right)\\
&= -\frac{1}{2}\log(1 - \cos^2(\nabla_x f_{\theta_1}(x), \nabla_x f_{\theta_2}(x))).
\end{align*}
\end{proof}

\textbf{Putting everything together}: For an infinitesimal perturbation (\(\sigma \downarrow 0\)), we know that \(\hat{f}_{\theta_1}(\tilde{x}) = f_{\theta_1}(\tilde{x})\) and \(\hat{f}_{\theta_2}(\tilde{x}) = f_{\theta_2}(\tilde{x})\), i.e., the linearization is exact. Of course, we have to re-linearize after every gradient step. By driving the mutual information to zero, we enforce statistical independence between \(f_{\theta_1}(\tilde{x})\) and \(f_{\theta_2}(\tilde{x})\). It is also easy to see that for \(\sigma \downarrow 0\),
\begin{align*}
\min_{\theta_1, \theta_2} \operatorname{MI}(f_{\theta_1}(\tilde{x}), f_{\theta_2}(\tilde{x})) &= \min_{\theta_1, \theta_2} -\frac{1}{2}\log(1 - \cos^2(\nabla_x f_{\theta_1}(x), \nabla_x f_{\theta_2}(x)))\\
&= \max_{\theta_1, \theta_2} \log(1 - \cos^2(\nabla_x f_{\theta_1}(x), \nabla_x f_{\theta_2}(x)))\\
&= \max_{\theta_1, \theta_2} 1 - \cos^2(\nabla_x f_{\theta_1}(x), \nabla_x f_{\theta_2}(x))\\
&= \min_{\theta_1, \theta_2} \cos^2(\nabla_x f_{\theta_1}(x), \nabla_x f_{\theta_2}(x)).
\end{align*}
Therefore, the local independence loss
\[\cL_{\mathrm{indep}}(\nabla_x f_{\theta_{m_1}}(x), \nabla_x f_{\theta_{m_2}}(x)) = \cos^2(\nabla_x f_{\theta_{m_1}}(x), \nabla_x f_{\theta_{m_2}}(x))\]
for a pair of models \((m_1, m_2)\) indeed encourages the statistical independence of the models' outputs, considering an infinitesimal Gaussian perturbation around the input \(x\). The obvious minimizer of the term is any constellation where the two input gradients are orthogonal.
 
\medskip
 
\textbf{Note}: It is also easy to see from the correlation and mutual information formulas that for Gaussian variables, zero correlation is equivalent to independence. This is, of course, not true in general.
\end{information}

\section{Adversarial OOD Generalization}

\begin{figure}
    \centering
    \includegraphics[width=0.8\linewidth]{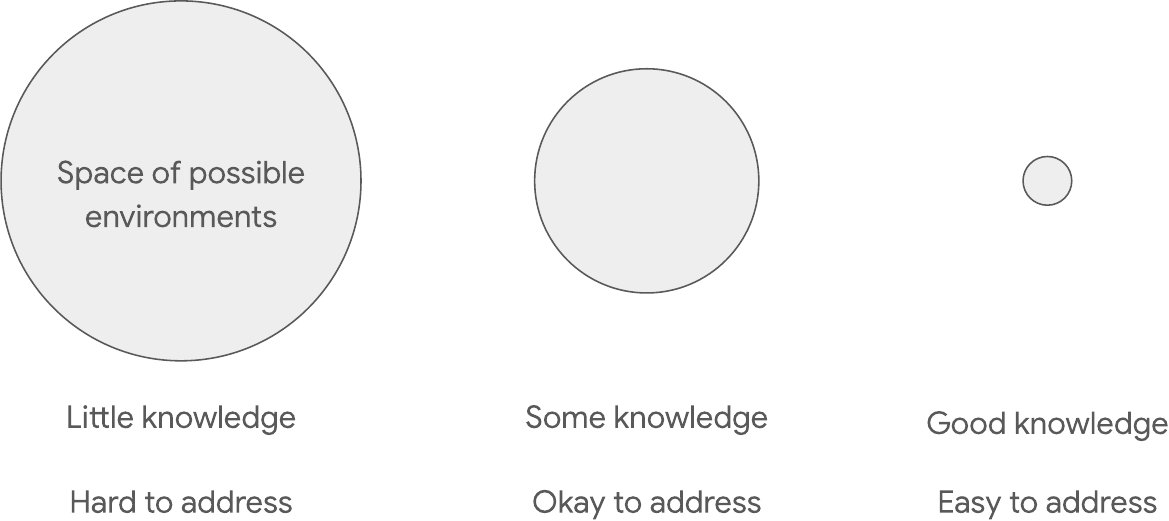}
    \caption{The size of the space of possible environments shrinks as we have more and more information about deployment.}
    \label{fig:knowledge}
\end{figure}

OOD generalization is about dealing with uncertainty. It is easy to make a model generalize well to a single possible environment. As we introduce more environments, this becomes harder and harder until we arrive at an infinite number of environments or ``any environment''. This tendency is illustrated in Figure~\ref{fig:knowledge}. As we have more and more knowledge about what will happen at deployment time, the space of possible environments shrinks, and thus we become more certain. A parallel can be drawn with the notion of entropy: If we already have much knowledge, additional information has a small entropy.

The question is: How can we take care of an infinite number of possible environments? There are two general methods for dealing with uncertainty when we do not know the deployment environment perfectly (or the \emph{enemy} who is trying to give us a hard environment):
\begin{enumerate}
    \item \textbf{Make an educated guess.} For a good guess, this is a nice, practical solution that is easy to carry out. However, we obtain no guarantees: We do not know if we made the right guess. In the worst case, we are not making any progress. Many methods seen so far fall into this category, e.g., ReBias, Predicting is not Understanding, and Learning from Failure (by guessing the bias).
    \item \textbf{Prepare for the worst.} Here, we have a so-called \emph{adversarial environment}. By following this principle, we can obtain theoretical lower bound guarantees: Our model's performance against the worst-case environment provides a lower bound on its performance against the space of possible environments. (We are safe for the worst-case scenario from a set of possible scenarios, so we are also safe for all of them.) An important caveat is that the guarantee is only within the pre-set space of possible environments (the strategy space). Outside of this, we have no guarantees. This approach can also lead to unrealistically pessimistic solutions.
\end{enumerate}
As they both have their pros and cons, there is no single right answer: it is a matter of choice and depends on our application.

So far, we have only considered OOD generalization methods for making an educated guess. Now, let us discuss \emph{adversarial generalization} that comprises methods that prepare for the worst.

\begin{definition}{Adversarial Generalization}
Adversarial generalization is an ML technique for ``preparing for the worst-case scenario'' when we do not know the target scenario/distribution in deployment.
\end{definition}

The following subsections will describe this type of OOD generalization in more detail.

\subsection{Formulation of a General Adversarial Environment}

Before discussing adversarial generalization, let us first introduce the notion of a \emph{devil}.

\begin{definition}{Devil}
The devil is a (known or unknown) adversary that actively tries to find the worst environment for us from the strategy space according to the adversarial goal and knowledge. The more knowledge it has, the worse environments it can specify for us.
\end{definition}

A general adversarial environment is specified by three parts: the \emph{adversarial goal}, the \emph{strategy space}, and the \emph{knowledge}.\footnote{If one of them is left unspecified, we are missing critical ingredients.} The exact definitions of these parts are given below.

\begin{definition}{Adversarial Goal}
The adversarial goal is a key component of an adversarial setting that specifies which environment is considered ``worst'' for our model.
\end{definition}

\begin{definition}{Strategy Space}
The strategy space in an adversarial setting defines the space of possible environments the devil can choose from.
\end{definition}

\begin{definition}{Knowledge}
The knowledge of the devil in an adversarial setting specifies the devil's ability to pick the worst environment for our model. In short, it defines what the devil knows about the model.
\end{definition}

In the next sections, we will discuss how exactly the devil can achieve their goals.

\subsection{Fast Gradient Sign Method (FGSM)}

First, we start with the definition of a \emph{white-box attack}, as the Fast Gradient Sign Method (FGSM) falls into this category.

\begin{definition}[label=def:whitebox]{White-Box Attack}
When the adversary knows the model architecture and the weights, we call the attack a white-box attack.
\end{definition}

\begin{figure}
    \centering
    \includegraphics[width=0.8\linewidth]{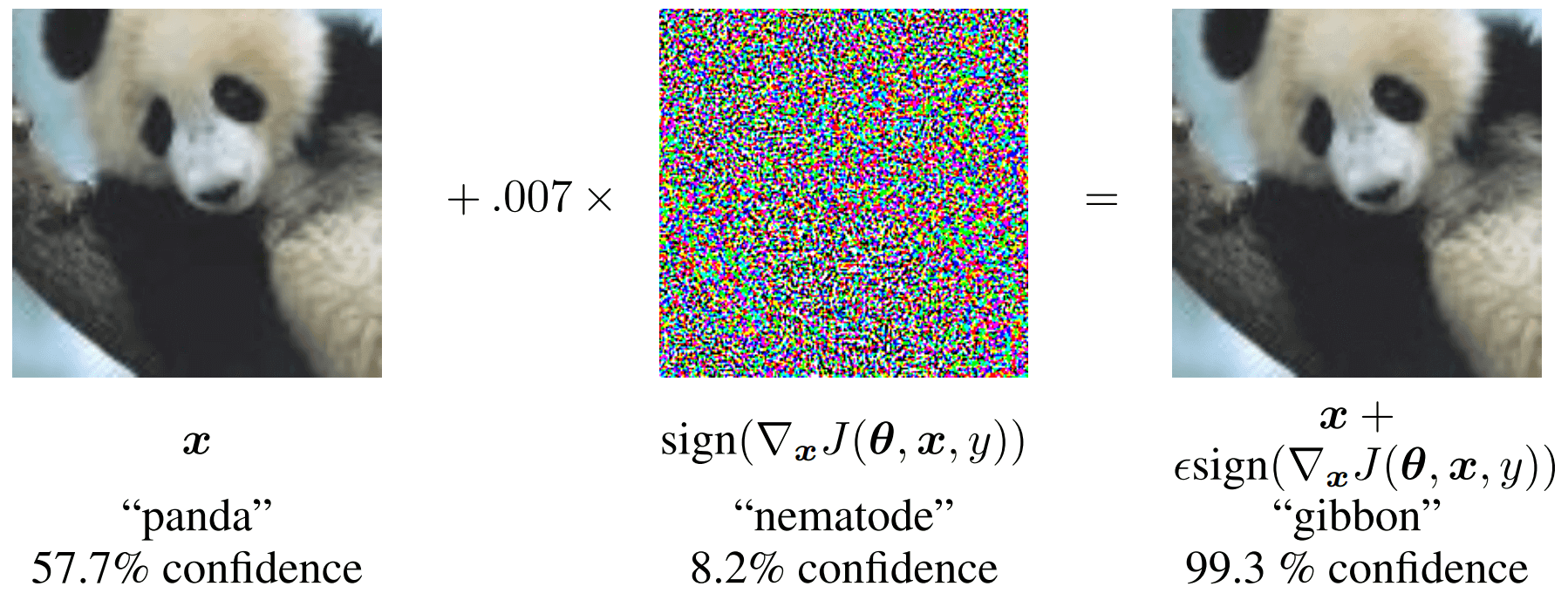}
    \caption{Demonstration of FGSM, taken from~\cite{https://doi.org/10.48550/arxiv.1412.6572}. By adding some noise of small magnitude, the network very confidently predicts an incorrect class, destroying the performance of the model. \(J\) is the cost function (loss) we wish to \emph{maximize}.}
    \label{fig:iconic}
\end{figure}

\begin{figure}
    \centering
    \includegraphics[width=0.35\linewidth]{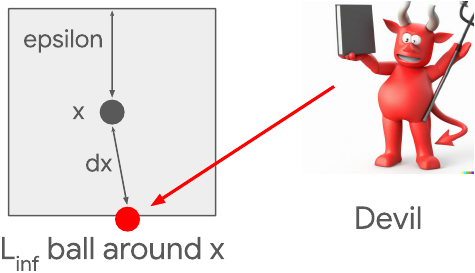}
    \caption{Informal illustration of the FGSM method's strategy space. The devil aims to find an adversarial sample in the \(L_\infty\) \(\epsilon\)-ball around the original input \(x\).}
    \label{fig:fgsm}
\end{figure}

If the devil does not want to think too much, then FGSM can be a popular first choice, as it is one of the simplest ways to achieve adversarial goals.
The FGSM attack, introduced in ``\href{https://arxiv.org/abs/1412.6572}{Explaining and Harnessing Adversarial Examples}''~\cite{https://doi.org/10.48550/arxiv.1412.6572} is a type of \(L_\infty\) adversarial attack. Its three ingredients are listed below.
\begin{itemize}
    \item \textbf{Adversarial Goal}: Reducing classification accuracy while being imperceptible to humans.
    \item \textbf{Strategy Space}: For every sample, the adversary may add a perturbation \(dx\) with norm \(\Vert dx \Vert_\infty \le \epsilon\) (to make sure it is imperceptible).
    \item \textbf{Knowledge}: Access to the model architecture, weights, and thus gradients. (White-box attack.)
\end{itemize}
The iconic image from~\cite{https://doi.org/10.48550/arxiv.1412.6572}, shown in Figure~\ref{fig:iconic}, depicts the attack, where a small perturbation applied completely destroys the model's prediction performance (``gibbon'' with 99.3\% confidence). A more general informal illustration of this scenario is given in Figure~\ref{fig:fgsm}.

\subsubsection{FGSM Method}



The FGSM attack perturbs the image \(x\) as
\[x + \epsilon\ \operatorname{sgn}\left(\nabla_x \cL(\theta, x, y)\right)\]
where \(\cL\) is the loss function used for model \(\theta\), \(y\) is the ground truth label, and the sign function is applied element-wise. One also has to take care about the image staying in the range \([0, 1]^{H \times W \times 3}\) by clipping or normalizing. \(\epsilon\) is the size of the perturbation, which is determined by the strategy space. It defines the maximal \(L_\infty\) norm of the perturbation.

Let us consider the pros and cons of FGSM below.
\begin{itemize}
    \item \textbf{Pros}: The method is very simple. We take a binary map of the gradient of the loss, i.e., the direction in which the loss increases the most around \(x\). The method is also cheap. It only requires one forward and backward pass per sample to create an adversarial perturbation which makes it swift to obtain.
    \item \textbf{Cons}: The method does not give an optimal result. The perturbed image does not necessarily correspond to the worst-case sample in the \(L_\infty\) ball (but it generally gives a good adversarial attack still for unprotected networks).
\end{itemize}

\subsection{Projected Gradient Descent (PGD)}

\begin{figure}
    \centering
    \includegraphics[width=0.35\linewidth]{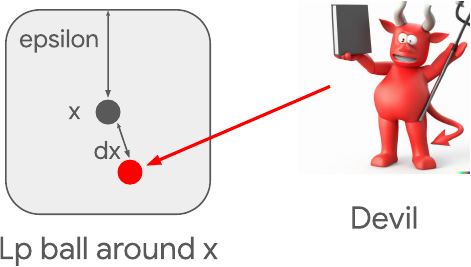}
    \caption{Informal illustration of the PGD method's strategy space. The devil aims to find a strong attack in the \(L_p\) ball around input \(x\).}
    \label{fig:pgd}
\end{figure}

If the devil wants to do something more sophisticated to succeed, the Projected Gradient Descent (PGD) might be a favorable choice for them.
The PGD attack, introduced in the paper ``\href{https://arxiv.org/abs/1706.06083}{Towards Deep Learning Models Resistant to Adversarial Attacks}''~\cite{https://doi.org/10.48550/arxiv.1706.06083}, is a type of \(L_p\) adversarial attack (\(1 \le p \le \infty\)), which is the strongest white-box attack to date (also because not many people are looking into strong attacks anymore). The three ingredients of it are detailed below.
\begin{itemize}
    \item \textbf{Adversarial Goal}: Same as for FGSM.
    \item \textbf{Strategy Space}: For every sample, the adversary may add a perturbation \(dx\) with norm \(\Vert dx \Vert_p \le \epsilon\).
    \item \textbf{Knowledge}: Same as for FGSM. The devil knows everything about the model, both structural details and the weights. It tries to generate a critical perturbation direction based on \(\theta\).
\end{itemize}
An illustration depicting this scenario is shown in Figure~\ref{fig:pgd}. The devil is \emph{trying to find} the worst-case sample for a fixed \(x\) in the \(L_p\) ball.

\subsubsection{PGD Method}

The PGD attack solves the optimization problem
\begin{align*}
&\max_{dx \in \nR^{H \times W \times 3}} \cL(f(x + dx), y; \theta)\\
&\text{s.t. } x + dx \in [0, 1]^{H \times W \times 3}\\
&\text{and } \Vert dx \Vert_p \le \epsilon.
\end{align*}
It perturbs the image \(x\) iteratively as
\[x^{t + 1} = \prod_{x + S} \left(x^t + \alpha \operatorname{sgn}\left(\nabla_x \cL(f(x^t), y; \theta)\right)\right)\]
where \(\cL\) is the loss function used for model \(\theta\), \(y\) is the ground truth label, \(t\) is the iteration index,
\(\alpha\) is the step size for each iteration, and \(\prod_{x + S}\) is the projection on the \(L_p\) \(\epsilon\)-sphere around \(x\).\footnote{\(S = \left\{y \in \nR^{H \times W \times 3} \middle| \Vert y \Vert_p \le \epsilon\right\}\).}
\(\epsilon\) is the size of the perturbation, which is determined by the strategy space. It defines the maximal \(L_p\) norm of the perturbation.

Being an iterative algorithm, PGD usually finds an even worse-case sample than FGSM (which only performs a single step). We iteratively follow the sign of the gradient with step size \(\alpha\) and project back onto the \(L_p\) \(\epsilon\)-ball around \(x\). According to the properties of the sign function, in each step, we go in an angle of \(\beta \in \{\pm 45^\circ, \pm 90^\circ, \pm 135^\circ, 0^\circ, 180^\circ\}\) from the previous \(x^t\) before projecting back onto the \(\epsilon\)-ball.\footnote{The projection \emph{can} change this angle.} (Usually, in visualization, this means traveling along the boundary of the \(L_p\) ball.) Now we can go out of the \(L_p\) ball of \(\epsilon\) even in a single step (especially around the `corners' of the \(L_p\) ball), depending on how we choose \(\alpha\). Like in FGSM, we also take care of the image staying in the range \([0, 1]^{H \times W \times 3}\) using clipping or normalization. Convergence happens when, e.g., \(\Vert x^{t + 1} - x^t \Vert_2 \le 1\mathrm{e}{-5}\) or some similar criterion is satisfied.

\begin{information}{Using the Gradient's Sign}
Why do we use the sign of the gradient in these methods and not the magnitude? Either case works. However, e.g., Adam~\cite{kingma2017adam} is also taking the sign of the gradient for updates (considering the formula without the exponential moving average) and is one of the SotA methods. In high dimensions, the choice of taking the sign does not matter much. This is usually a choice we make based on empirical observations.
\end{information}

\subsection{FGSM vs. PGD}

Let us briefly compare the two attacks we have seen so far, FGSM and PGD. In both cases, the optimization problem for the adversary is non-convex, as the loss surface is non-convex in \(x\). We also have no guarantee for the globally optimal solution, even within a small \(\epsilon\)-ball (which is very tiny in a high-dimensional space). The strength of the attack depends a lot on the optimization algorithm. We have many design choices, and not all Gradient Descent (GD) variants perform similarly. \textbf{PGD is generally much stronger than FGSM; it finds better local optima.} FGSM does not even find local optima in general, as it consists of just a single gradient step. PGD is generally a SotA white-box attack even as of 2023.

\begin{information}{Size of the \(\epsilon\)-ball in High-Dimensional Spaces and Distribution of Volume}
Why is the \(\epsilon\)-ball tiny in a high-dimensional space for a small value of \(\epsilon\)? The volume of a ball with radius \(r\) in \(\nR^d\) is
\[V_d = \frac{\pi^{d/2}}{\Gamma\left(1 + \frac{d}{2}\right)}r^d\]
and \(\Gamma(n) = (n - 1)!\) for a positive integer \(n\). Therefore, the denominator increases much faster than the numerator, driving the volume to 0 as \(d \rightarrow \infty\).

\medskip

The volume is thus concentrated near the surface in high-dimensional spaces: For a fixed dimension \(d\), the fraction of the volume of a smaller ball with radius \(r < 1\) inside a unit ball is \(r^d\) (as the scalar multiplier cancels). For \(r \approx 1\) but \(d\) very large, this is around 0.
\end{information}

\begin{information}{How to choose \(\epsilon\) in \(L_p\) attacks?}

The hyperparameter \(\epsilon\) is usually chosen to be very small. Even more importantly, one should fix it across studies, as we typically wish to compare against previous attacks/defenses. There are unified values in the community but the exact value does not matter that much, as below a certain threshold, the perturbations are (mostly) not visible to humans anyway.
\end{information}

\subsection{Different Strategy Spaces for Adversarial Attacks}

\begin{figure}
    \centering
    \includegraphics[width=0.7\linewidth]{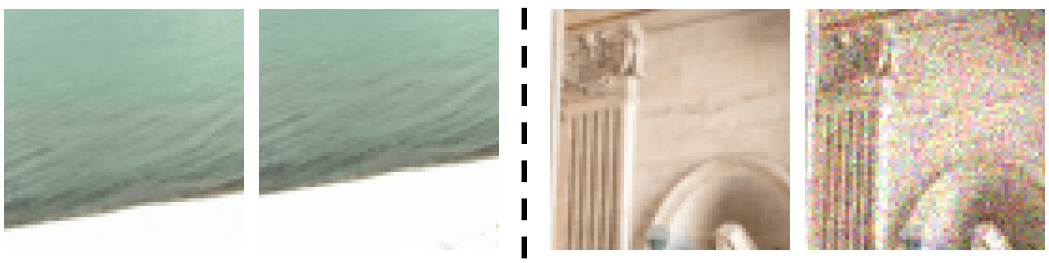}
    \caption{Example of two image pairs where humans would choose the \emph{left} pair as more similar, but regarding \(L_2\) distances, the \emph{right} pair is much closer. This is because of the translation in the first image pair. Figure-snippet taken from~\cite{DBLP:journals/corr/abs-1801-03924}.}
    \label{fig:unaligned}
\end{figure}

So far, we have considered perturbations inside an \(L_p\) ball determined by \(\epsilon\). The problem with this strategy space is that it is not aligned with human perception -- it is in the pixel space. It is missing some perturbations that are not visible to humans (such as shifting all pixels up by one), but it also captures some changes apparent to humans (such as additive noise at initially very clear and homogeneous surfaces in images). The \(L_p\) strategy space is thus not well-aligned with the adversarial goal. Sometimes it does not satisfy the goal (as the adversary's goal is to produce imperceptible perturbations), and sometimes it technically satisfies the goal but could do it even better (as the adversary's goal is usually also to decrease accuracy as much as possible). The misalignment of additive perturbations and the adversarial goal is further illustrated in Figure~\ref{fig:unaligned}. According to this observation, in the following subsections, we will consider strategy spaces that are different from the \(L_p\)-ball-based ones.

\subsubsection{Flow-Based Perturbations}

In general, images closer in perception space (ones that look more similar to humans) can have a larger \(L_p\) difference than obviously different image pairs. Suppose that we have a robust model against any \(L_p\) perturbations \wrt a ball parameterized by a small \(\epsilon\). In this case, the adversary could still be able to find a one-pixel shift of the image that destroys the model's predictions, even though this perturbation is imperceptible. This is a ``blind spot'' of an adversary that uses an \(L_p\)-ball-based strategy space.

\begin{definition}[label=def:totalvariation]{Total Variation}
The total variation of a vector field \(f\colon \nR^2 \rightarrow \nR^2\) is defined as
\[\Vert f \Vert_\mathrm{TV} = \int \Vert \nabla f_1(x) \Vert_2 + \Vert \nabla f_2(x) \Vert_2\ dx =: \int \Vert \nabla f(x) \Vert_2\ dx.\]

It is often considered a generalization of the \(L_2\) (or \(L_1\)) norm of the gradient to an entire vector space.
\end{definition}

Such small image translations (shifts) generally result in huge \(L_2\) distances (as images are not smooth, and along the object boundaries, we have a significant pixel distance), but correspond to perceptually minor differences. Luckily, we can define a metric that assigns small distances to small \emph{per-pixel} image translations. We consider \emph{optical flow} transformations, and we measure small changes by the \emph{total variation (TV)} norm, for which translations of an image have a ``size'' of zero. Such attacks are discussed in detail in Section~\ref{sssec:flow}.

\subsubsection{Physical Attacks}

\begin{figure}
    \centering
    \includegraphics[width=\linewidth]{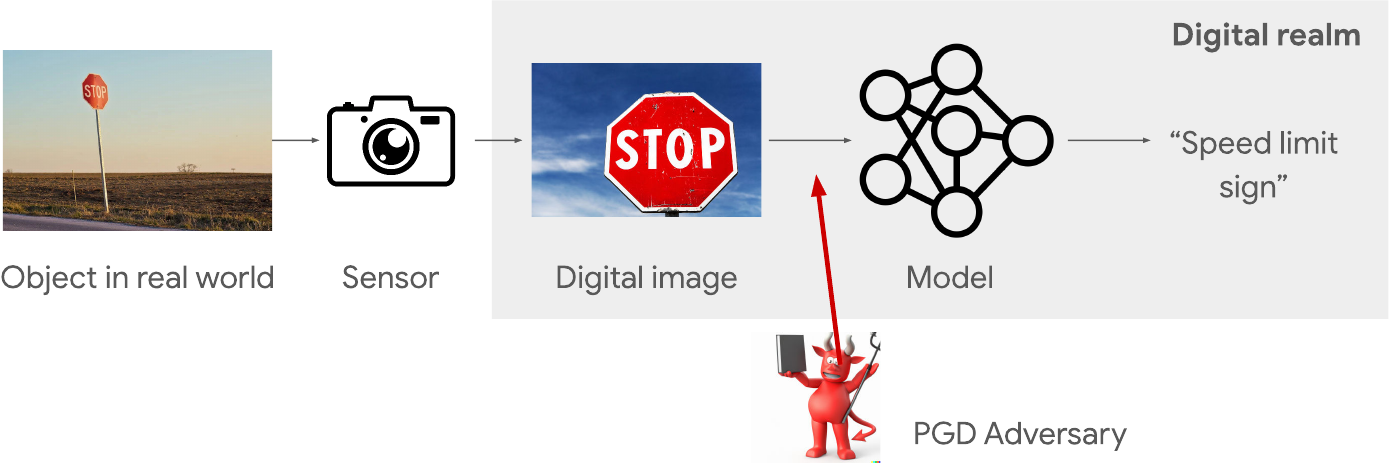}
    \caption{The PGD adversary, being a white-box attack, has access to the model in the digital realm.}
    \label{fig:plausibility}
\end{figure}

The plausibility of the previously discussed strategy spaces is questionable. \(L_p\) attacks and other attacks (e.g., flow-based ones) alter the \emph{digital image}. Do such adversaries exist in the real world? Are the previous attacks plausible at all? Basic security technology can already prevent such adversaries, with access as depicted in Figure~\ref{fig:plausibility}. Therefore, looking into other strategy spaces is well-motivated.

\begin{definition}[label=def:blackbox]{Black-Box Attack}
When the adversary only observes the inputs and outputs of a model and does not know the model architecture and the weights, we call the attack a black-box attack.
\end{definition}

The PGD adversary perturbs pixels of a digital image after it is captured in the real world. Does this scenario make sense? Should we even defend against such an adversary? We just have to ensure that no one gets to see our compiled code and that no one can change the data stream. This is basic information security. Furthermore, even if one gains access to the data stream, one also needs access to the exact model for white-box attacks. (Once the adversary is that deep in, they might as well just change the prediction directly\dots) For black-box attacks in the digital realm, this is not needed, but it is still a strange scenario where one has access to the data stream but not the model output. We even go one step further in the discussion about plausibility: When we use an API and have no access to internal data streams, we can indeed construct black-box attacks for the model (as we will see in Sections~\ref{sssec:sub} and \ref{sssec:zero}). However, this only ruins the accuracy for us, which seems to be a very poor adversarial goal. \textbf{By focusing on attacks in the digital realm, we are probably looking at a non-existent problem.}

\begin{figure}
    \centering
    \includegraphics[width=\linewidth]{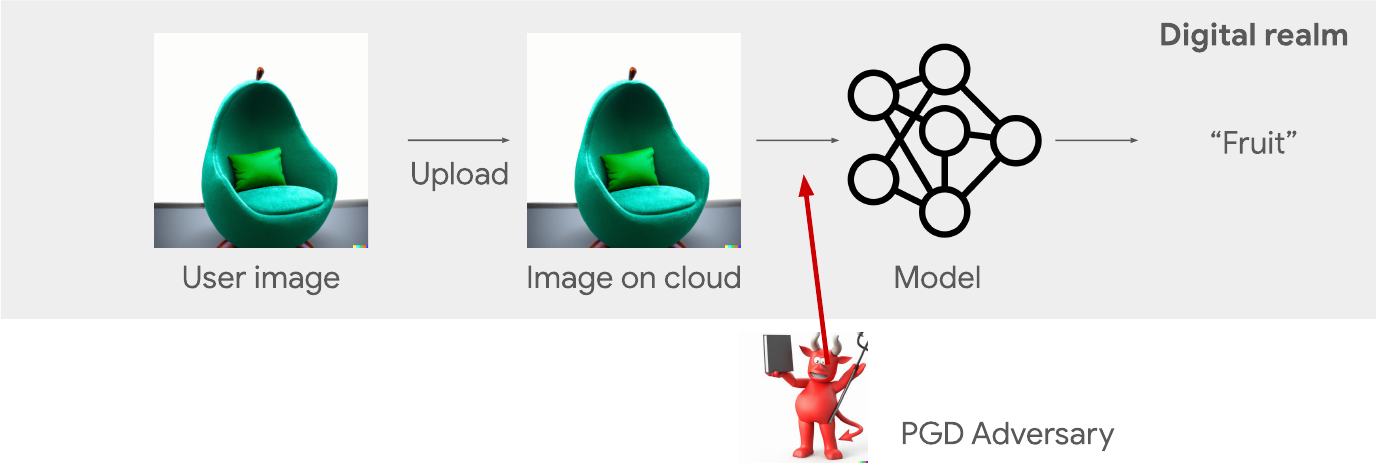}
    \caption{In a cloud setting, the PGD adversary still acts in the digital realm.}
    \label{fig:plausibility2}
\end{figure}

Another very similar scenario in the digital realm is when images are uploaded to the cloud, as shown in Figure~\ref{fig:plausibility2}. It is very unrealistic for an adversary to come into this pipeline and make changes.

\begin{figure}
    \centering
    \includegraphics[width=0.6\linewidth]{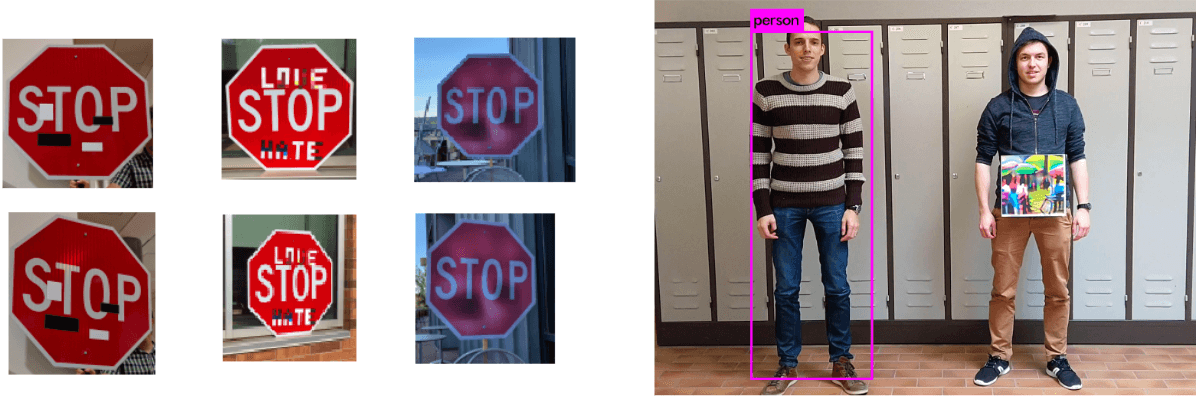}
    \caption{Physical attacks are more realistic than those in the digital realm~\cite{9025518, physicalreview}.}
    \label{fig:physical}
\end{figure}

\begin{figure}
    \centering
    \includegraphics[width=\linewidth]{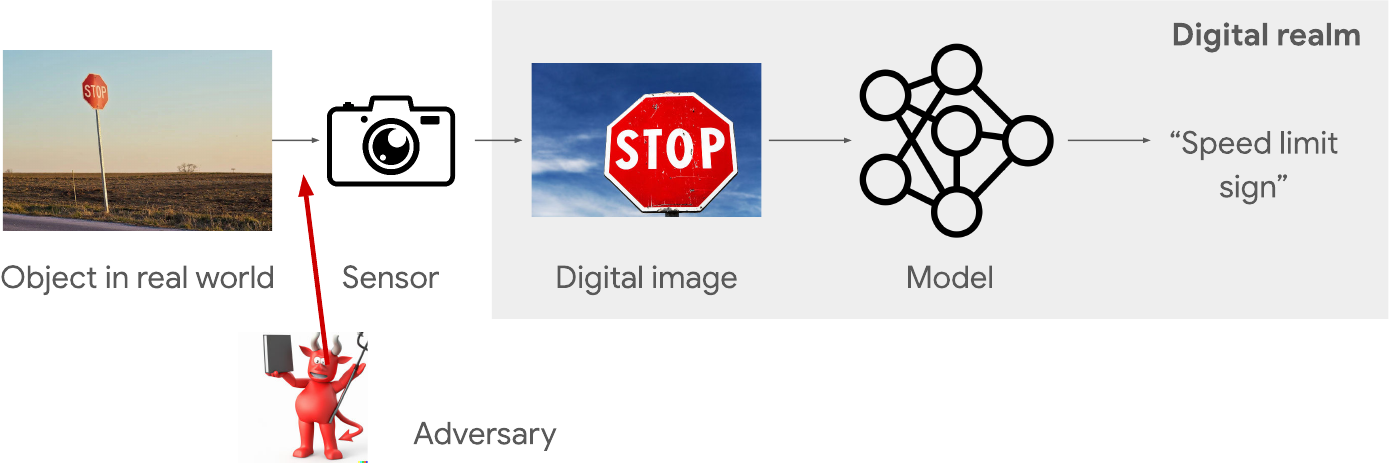}
    \caption{In a physical adversarial setting, the adversary has access to the object in the real world. The adversary might also know the internals of the model (considering a white-box setting), but still only intervene in the physical world.}
    \label{fig:plausibility3}
\end{figure}

This lack of realism in attacks in the digital realm inspires the search for a new strategy space in the real world: Let us discuss \emph{physical attacks}. In contrast to previously mentioned attacks, they induce physical changes in objects in the real world. These involve, e.g., putting a carefully constructed sticker (or graffiti) on stop signs to make sure that self-driving cars do not detect it or printing a pattern on cardboard (and e.g. wearing it around the neck) such that the person carrying the sign does not get detected. These options are illustrated in Figure~\ref{fig:physical}. This is much more realistic, as the adversary intervenes in the real world, not in a secure stage in a pipeline. The adversaries usually \emph{do} have the necessary access to real-world objects. We argue that we should instead be focusing on defending against such attacks, shown in Figure~\ref{fig:plausibility3}. \textbf{Note}: Such attacks can be both black-box and white-box attacks.

\subsubsection{Object Poses in the 3D World}

\begin{figure}
    \centering
    \includegraphics[width=0.6\linewidth]{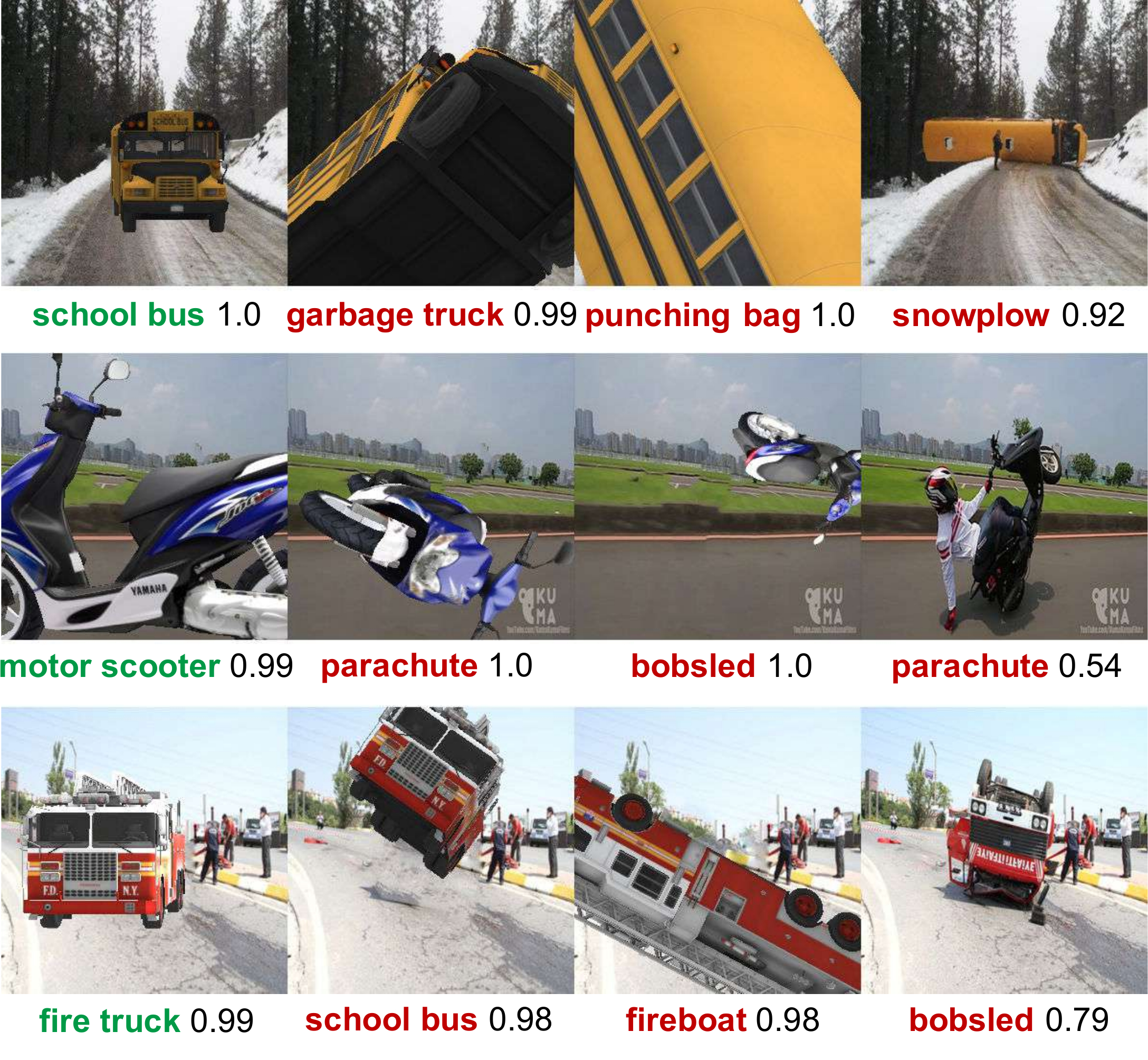}
    \caption{Collage of synthetic and real images with the model's corresponding max-probability predictions. According to the human eye, images in (row, column) positions (1, 4), (2, 2), (2, 4), (4, 1), (4, 4) are quite plausible.}
    \label{fig:poses}
\end{figure}

We briefly discuss an interesting boundary between adversarial robustness and OOD generalization that also introduces a new strategy space. This is the paper ``\href{https://arxiv.org/abs/1811.11553}{Strike (with) a Pose: Neural Networks Are Easily Fooled by Strange Poses of Familiar Objects}''~\cite{https://doi.org/10.48550/arxiv.1811.11553} which focuses on changing poses of objects in 3D space (which is similar to physical attacks but can also be done digitally given a sophisticated image synthesis tool). A collage of synthetic and real images the authors considered is shown in Figure~\ref{fig:poses}.

The three ingredients of this ``attack'' are as follows.
\begin{itemize}
    \item \textbf{Adversarial Goal}: Reducing classification accuracy by changing object poses.
    \item \textbf{Strategy Space}: For every sample, the adversary may arbitrarily change the object poses.
    \item \textbf{Knowledge}: Same as for FGSM. (White-box attack.)
\end{itemize}

Here, the adversary does not necessarily care about small changes in the object pose. Larger changes can still be plausible for the human eye. Once it becomes obvious to humans, they can, of course, intervene. The devil knows everything about the model, both structural details and the weights. It tries to generate a critical pose perturbation based on weights \(\theta\).

One may ask how this is a real threat at all. The threatening observation is that the model completely breaks down for the plausible examples, even though these could be observed in the real world. This work is on the boundary of adversarial robustness and OOD generalization to real-world domains.\footnote{This is a boundary between \emph{intentions}. The method can be used to construct adversarial examples that also correspond to plausible OOD domains we might wish to generalize to.} If the perturbation grows larger and we do not have a notion of a devil and worst-case samples anymore, we enter the realm of generalization to plausible real-world domains, across biases, or in other OOD generalization schemes.

\subsection{Optical Flow}
\label{sssec:flow}

Let us now discuss the optical flow approach in more detail. Optical flow is used a lot for visual tracking and videos. It provides the smallest warping of the underlying image mesh to transform image \(x_1\) into \(x_2\).\footnote{Warping refers to the pixel-wise displacements between the two image meshes.} It specifies the \emph{apparent} movement of pixels which is needed to transform image \(x_1\) into \(x_2\). We obtain it by performing (regularized) pixel matching between images/frames.

Optical flow is represented as a vector field over the 2D image plane. Each point of the 2D pixel plane corresponds to a 2D vector. Hence, the size of the warping may be readily computed via total variation (TV) (Definition~\ref{def:totalvariation}). This vector field is usually encoded by colors for visualization. The pixel intensity gives the 2D vector magnitude at the pixel, and the pixel color specifies the 2D vector direction at the pixel.

\textbf{Example}: Consider a ball flying across the sky. The ball pixels are translated across the frames by a tiny bit, but the \(L_2\) distance between the frames is large. Our task is to find \emph{pixel correspondences} between the two frames based on apparent motion. We set up a vector going from pixel \((i, j)\) in frame \(t\) to the corresponding pixel \((i', j')\) in frame \((t + 1)\). For example, if \((i, j) = (4, 5), (i', j') = (7, 2)\), then the forward flow is \((u, v) = (3, -3)\). We measure the distance between pixels by taking the \(L_2\) norm of this vector and taking the average of these distances for every pixel. This is precisely what we do when calculating TV. This gives an idea of how much warping has taken place between the two frames. A small flow, however, can also correspond to human-perceptible changes: a small ball flying fast between two frames on a huge, otherwise static image will have a low TV value, but humans are able to point out the differences quickly. Nevertheless, the perturbed images are still deemed plausible by human inspection.

\subsection{Adversarial Flow-Based Perturbation}

How can we use optical flow to find adversarial patterns? Instead of estimating the flow between 2 consecutive frames, we \emph{generate} a flow with a small total variation that fools our model, as done in the paper ``\href{https://openreview.net/forum?id=HyydRMZC-}{Spatially Transformed Adversarial Examples}''~\cite{xiao2018spatially}. The three ingredients of their method are:
\begin{itemize}
    \item \textbf{Adversarial Goal}: Reducing classification accuracy while being imperceptible to humans.
    \item \textbf{Strategy Space}: For every sample, the adversary may choose a flow \(f\) with \(\Vert f \Vert_\mathrm{TV} \le \epsilon\).
    \item \textbf{Knowledge}: Same as for FGSM. (White-box attack.)
\end{itemize}

This perturbation method is better aligned with human perception (i.e., it is a good proxy for it). It finds pixel-wise movement instead of additive perturbation. The adversary warps the underlying image mesh of image \(x\) according to \(f\) such that the classification result is wrong. If the vector field is aligned in the same direction (constant map), there is no total variation. On the contrary, if the vector field comprises vectors with large magnitudes that are closely spaced and point in different directions, it results in a large TV norm. These abrupt changes in nearby vectors correspond to steep gradients in the field. This is why penalizing total variation encourages images to be \emph{smoother}.

\begin{figure}
    \centering
    \includegraphics[width=0.9\linewidth]{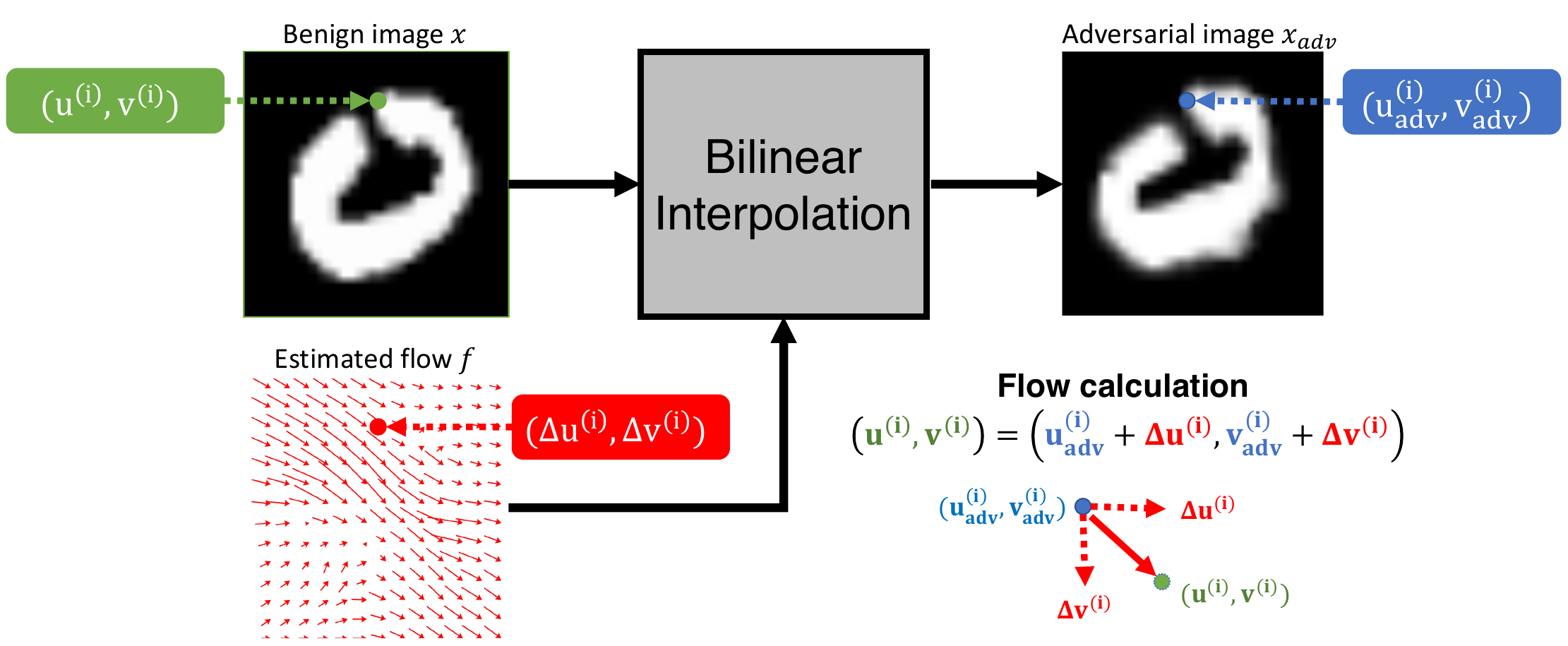}
    \caption{Overview of a flow-based adversarial attack using bilinear interpolation to obtain its final adversarial image from the \emph{backward} flow, taken from~\cite{xiao2018spatially}. See information~\ref{inf:flowadv} for details.}
    \label{fig:flowadv}
\end{figure}

How can we obtain the final adversarial image from the adversarial flow? Figure~\ref{fig:flowadv} shows a possible way using the \emph{backward flow} and \emph{bilinear interpolation}. 

\begin{information}{Interpolation between source and adversarial images in flow-based adversarial attack}
\label{inf:flowadv}
    During the adversarial attack in Figure~\ref{fig:flowadv}, the image \(x\) is fixed. The devil comes up with a \emph{backward} optical flow that takes the target pixels to the original pixels. The reason to predict backward flow instead of forward flow is easier bilinear interpolation. When the backward flow is available, each pixel of the adversarial image can be computed after querying some known pixel of the original image (source). On the contrary, using the forward flow to obtain the adversarial example would result in ``holes'' in the image. The actual \emph{magnitude} of the warps does not matter when calculating the TV norm, so translations of any kind are allowed. At borders, we might copy the pixels of the original image.
\end{information}

\subsection{White-Box vs. Black-Box Attacks}

So far, we have discussed white-box (Definition~\ref{def:whitebox}) and black-box (Definition~\ref{def:blackbox}) attacks. Let us discuss some pros and cons of these paradigms.

\textbf{White-box attacks are powerful}. The adversary can obtain the input gradients from the model. (Examples: FGSM, PGD, Flow-Based Perturbation.) White-box attacks are, however, not so realistic. For an ML model on the cloud/as an API, we are never allowed to look into the details of the model. It is intellectual property, and exposing it would make the model vulnerable to various attacks. The quick solution that most companies follow is to not open source their model. \emph{Black-box attacks are much weaker than white-box attacks but also much more realistic.}

Many real-world applications are based on API access. There are also further limitations to a realistic scenario:
\begin{itemize}
    \item The number of queries within a time window is limited (rate limit).
    \item Malicious query inputs are possibly blocked.
\end{itemize}
For example, consider a face model recognizing the user in a photo album: If we start sending strange patterns like random noise or non-face images, it can easily be detected, and we can be blocked from the service.

\textbf{Examples of black-box APIs.} \href{https://chat.openai.com/}{GPT-3.5/4}~\cite{https://doi.org/10.48550/arxiv.2005.14165,openai2023gpt4} produces text output given text input. It is an interesting objective to attack GPT-N based on only input/output observation pairs. One example is Jailbreak Prompts~\cite{shen2023do}: Here, the adversarial goal is making the model tell us information about immoral or illegal topics; the strategy space of the devil is giving any prompt to the model; and the knowledge of the adversary is the observed answers of the model. The attack is black-box by definition because we do not have access to the model's internal structure.
\href{https://labs.openai.com}{DALL-E}~\cite{https://doi.org/10.48550/arxiv.2102.12092} produces an image given a textual description.

In the following sections, we will discuss black-box attacks in more detail.

\subsection{Black-Box Attack via a Substitute Model}
\label{sssec:sub}

\begin{definition}{Substitute Model}
A substitute model is a network that is used to mimic a model we wish to attack. Prior knowledge about the attacked model is incorporated into the substitute model, such as the type of architecture, the size of the model, or the optimizer it was trained with.
\end{definition}

\begin{figure}
    \centering
    \includegraphics[width=0.8\linewidth]{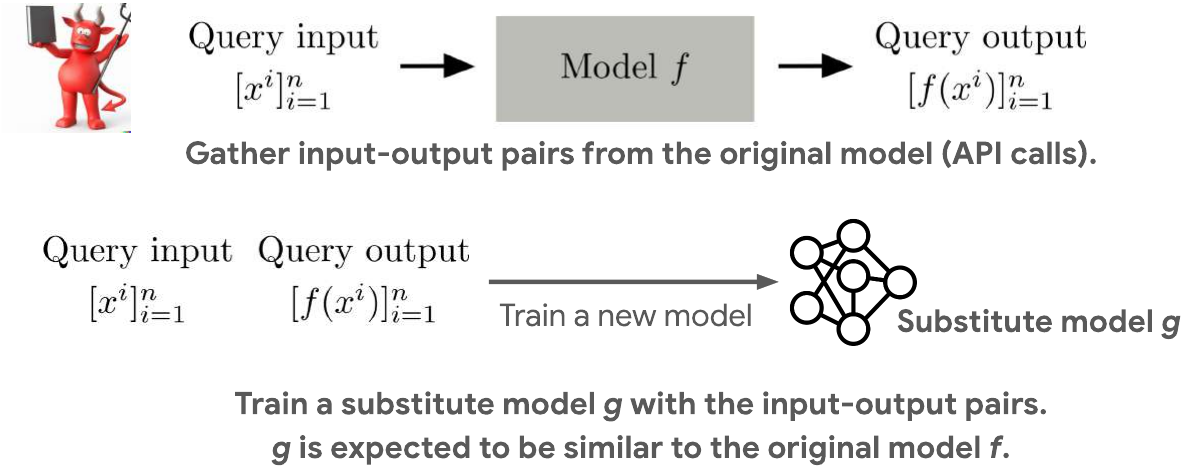}
    \caption{Illustration of a method for using substitute models to generate black-box adversarial attacks. We only need query inputs and outputs to train the substitute model.}
    \label{fig:substitute}
\end{figure}

We will start an overview of the black-box attacks with the seminal work ``\href{https://arxiv.org/abs/1602.02697}{Practical Black-Box Attacks against Machine Learning}''~\cite{https://doi.org/10.48550/arxiv.1602.02697}. It introduces the idea of using a substitute model to attack the original model. An overview of the method is given in Figure~\ref{fig:substitute}. Using this approach, we might need a lot of input-output pairs from the original model, depending on how complex the model is. Ideally, we want to follow the architecture of the target model. If we, e.g., know that the original model is a Transformer, we should also use one. We then attack the original model by creating adversarial inputs that attack the substitute model \(g\). By using \(g\), the adversary can generate \emph{white-box} attacks. The hope is that this attack also works for \(f\). Based on empirical observations, this can work quite well.

For smaller models, this method might be feasible to attack model \(f\) well. However, for larger models, we need tons of data and extreme computational effort to train the substitute model. In particular, only a handful of companies in the world could mimic GPT-3 with a substitute model. It would be easy to trace back who is responsible for the attacks. Usually, such large companies focus on problems other than these black-box attacks. They already have many other problems, e.g., private training data leakage by querying, bias issues, or explainability, that are way more realistic.

\subsection{Black-Box Attack via a Zeroth-Order Attack}
\label{sssec:zero}

Another type of black-box attack is based on the approximation of the model gradient with a lot of API calls. One way to do this is described in the work ``\href{https://arxiv.org/abs/1708.03999}{ZOO: Zeroth Order Optimization based Black-box Attacks to Deep Neural Networks without Training Substitute Models}''~\cite{Chen_2017}. The idea comes from the fact that one can approximate the gradient of the loss numerically using finite differences, so that for a small enough \(h \in \nR\):
\[\frac{\partial \cL(x)}{\partial x_i} \approx \frac{\cL(x + he_i) - \cL(x)}{h}\]
where \(e_i\) is the \(i\)th canonical basis vector.
One can also use a more stable symmetric version that gives better approximations in general (but requires more network evaluations):
\[\frac{\partial \cL(x)}{\partial x_i} \approx \frac{\cL(x + he_i) - \cL(x - he_i)}{2h},\]
where \(x \in \nR^d\) is a flattened image and \(\cL(x) \in \nR\) is the loss function of choice, based on our target class \(y\) that we want the model to classify \(x\) as.

\textbf{Example}: Consider a \(200 \times 200\) image. In this case, \(x \in \nR^{120,000}\) and \(\nabla \cL(x) \in \nR^{120,000}\). We need 120,001 API calls to approximate the gradient \emph{of a single image}, or 240,000 if we consider the symmetric approximation. No API will let us do this in a manageable time. For this to work, we also need access to the logits \(z(x)\) or the probabilities \(f(x)\) from the model to compute the objective \(\cL\), not just the predicted class label. For example, we might use the objective \(\cL(x) = \max\{\max_{i \ne y} z(x)_i - z(x)_y, -\kappa\}\), as given in~\cite{Chen_2017}.

Still, the worst case with such black-box attacks for the model owners is that the performance drops. This is not a realistic goal for an adversary, as it only happens to the attacker and only on the adversarial samples they create. If the attacker wants to decrease performance for others, too, they need access to the data stream.

\textbf{Note}: The paper was published in 2017. Back then, these attacks were focused on theoretical possibilities. Nowadays, the field is focusing more on realistic threats we have to tackle. The focus has shifted.

We can even be more imaginative and train a local model that predicts the pixel location most likely to generate the highest response by the attacked model. In this case, we need a dataset of (image, pixel) pairs where the pixel changes the prediction of many locally available models the most. Leveraging this dataset, we can get the model's pixel prediction and find the pixel's perturbation via API calls that result in the desired behavior (e.g., compute gradients for that pixel using finite differences and update that pixel of the image using the sign of the gradient).

We simplify the previous approach and pick random coordinates to perform stochastic coordinate descent, as shown in~\cite{Chen_2017}:
\begin{algorithm}
\caption{Stochastic Coordinate Descent}
\While{not converged}{
    Randomly pick a coordinate \(i \in \{1, \dots, d\}\)\\
    Compute an update \(dx_i^*\) by approximately minimizing
    \[\argmin_{dx_i} \cL(x + dx_i e_i)\]
    \(x_i \gets x_i + dx_i^*\)
}
\end{algorithm}

This is, of course, not very efficient. It is better to pick \(i\) smartly and perturb that pixel using a few API calls to determine a suitable perturbation.

\subsection{Defense against Attacks: Adversarial Training}

We have discussed many adversarial attacks. Is there any way to defend against them? To answer this, we will touch upon one instructive defense method that gave rise to the research direction of defense methods, called adversarial training. This method was introduced in the paper ``\href{https://arxiv.org/abs/1706.06083}{Towards Deep Learning Models Resistant to Adversarial Attacks}''~\cite{https://doi.org/10.48550/arxiv.1706.06083}. It is generally perceived as one of the best-working defenses against \(L_p\) attacks.

Adversarial training has a minimax formulation: Optimize \(\theta\) \wrt the worst-case perturbation of \(x\) as
\[\min_{\theta} \nE_{(x, y) \in \cD}\left[\max_{dx \in  S} \cL(x + dx, y; \theta)\right]\]
with, e.g., \(S = [-\epsilon, \epsilon]^N\) corresponding to an \(L_\infty\) attack.
In practice, we do a few PGD steps for each \(x\), generate an attack \(dx\), and use that for training \(\theta\).

\subsubsection{Results of Adversarial Training}

\begin{figure}
\begin{center}
{\setlength\tabcolsep{.06cm}
\begin{tabular}{c c c c}
 & \multicolumn{2}{c}{MNIST} &  \\
\begin{tikzpicture}[scale=0.4] 
    \begin{semilogxaxis}[
        xlabel=Capacity scale,
        ylabel=Accuracy,
        ylabel near ticks,
        xtick = data,
        log basis x = 2,
        log ticks with fixed point,
        grid = both,
        label style={font=\Huge},
        tick label style={font=\LARGE},
        every axis plot/.append style={ultra thick},
    ]
    \addplot plot coordinates {
        (1,  98.3)
        (2,  98.9)
        (4,  99.1)
        (8, 99.2)
        (16, 99.2)
    };
    \addplot plot coordinates {
        (1,  0.6)
        (2,  2.3)
        (4,  4.0)
        (8,  3.6)
        (16, 3.9)
    };
    \addplot plot coordinates {
        (1,  0.0)
        (2,  0.0)
        (4,  0.0)
        (8,  0.0)
        (16, 0.0)
    };
    \legend{}

    \end{semilogxaxis}
\end{tikzpicture} &
\begin{tikzpicture}[scale=0.4] 
    \begin{semilogxaxis}[
        xlabel=Capacity scale,
        ylabel near ticks,
        xtick = data,
        log basis x = 2,
        label style={font=\Huge},
        tick label style={font=\LARGE},
        log ticks with fixed point,
        grid = both,
        every axis plot/.append style={ultra thick},
    ]
    \addplot+[error bars/.cd,y dir=both,y explicit] plot coordinates {
        (1,  68.9) +- (0, 8.4)
        (2,  33.4) +- (0, 7.0)
        (4,  79.2) +- (0, 5.9)
        (8,  88.8) +- (0, 8.2)
        (16, 93.7) +- (0, 3.1)
    };
    \addplot+[error bars/.cd,y dir=both,y explicit] plot coordinates {
        (1,  94.8) +- (0, 3.4)
        (2,  99.5) +- (0, 0.3)
        (4,  99.8) +- (0, 0.1)
        (8,  99.3) +- (0, 0.5)
        (16, 99.6) +- (0, 0.3)
    };
    \addplot+[error bars/.cd,y dir=both,y explicit] plot coordinates {
        (1,  0.0)
        (2,  0.0)
        (4,  0.0) 
        (8,  0.0)
        (16, 0.0) 
    };
    \legend{}

    \end{semilogxaxis}
\end{tikzpicture} &
\begin{tikzpicture}[scale=0.4] 
    \begin{semilogxaxis}[
        xlabel=Capacity scale,
        ylabel near ticks,
        xtick = data,
        ymin = -10,
        label style={font=\Huge},
        tick label style={font=\LARGE},
        log ticks with fixed point,
        log basis x = 2,
        every axis plot/.append style={ultra thick},
        grid = both
    ]
    \addplot plot coordinates {
        (1,  11.4)
        (2,  11.4)
        (4,  97.6)
        (8,  98.0)
        (16, 98.9)
    };
    \addplot plot coordinates {
        (1,  11.4)
        (2,  11.4)
        (4,  89.6)
        (8, 93.9)
        (16, 95.5)
    };
    \addplot plot coordinates {
        (1,  11.4)
        (2,  11.4)
        (4,  87.7)
        (8, 91.5)
        (16, 93.3)
    };
    \legend{}

    \end{semilogxaxis}
\end{tikzpicture} &
\begin{tikzpicture}[scale=0.4] 
    \begin{loglogaxis}[
        xlabel=Capacity scale,
        ylabel=Average loss,
        ylabel near ticks,
        label style={font=\Huge},
        tick label style={font=\LARGE},
        log ticks with fixed point,
        xtick = data,
        log basis x = 2,
        grid = both,
        every axis plot/.append style={ultra thick},
        legend style = {at={(1.3,1.1)}, font=\Huge}
    ]
    \addplot plot coordinates {
        (1,  0.0537)
        (2,  0.0394)
        (4,  0.0371)
        (8,  0.0364)
        (16, 0.0351)
    };
    \addplot+[error bars/.cd,y dir=both,y explicit] plot coordinates {
        (1,  0.1647) +- (0, 0.1039)
        (2,  0.0175) +- (0, 0.0095)
        (4,  0.0067) +- (0, 0.0016)
        (8,  0.0206) +- (0, 0.0147)
        (16, 0.0110) +- (0, 0.0088)
    };
    \addplot plot coordinates {
        (1,  2.3011)
        (2,  2.3012)
        (4,  0.3650)
        (8,  0.2504)
        (16, 0.1770)
    };
    \legend{Natural\\ FGSM\\ PGD\\}

    \end{loglogaxis}
\end{tikzpicture} \\

 & \multicolumn{2}{c}{CIFAR10} &  \\
 
 \begin{tabular}{cc|c}
  & Simple & Wide \\ \hline
	Natural\quad & 92.7\% & 95.2\% \\
	FGSM\quad  & 27.5\% & 32.7\% \\
	PGD\quad  & 0.8\% & 3.5\% \\
\end{tabular}
 &
  \begin{tabular}{c|c}
  Simple & Wide \\ \hline
	 87.4\% & 90.3\% \\
	 90.9\% & 95.1\% \\
	 0.0\% & 0.0\% \\
\end{tabular}
 &
  \begin{tabular}{c|c}
   Simple & Wide \\ \hline
	 79.4\% & 87.3\% \\
	 51.7\% & 56.1\% \\
	 43.7\% & 45.8\% \\
\end{tabular}
 &
   \begin{tabular}{c|c}
   Simple & Wide \\ \hline
	 0.00357 & 0.00371 \\
	 0.0115 & 0.00557 \\
	 1.11 & 0.0218 \\
\end{tabular}
\\	
(a) Standard training & (b) FGSM training & (c) PGD training & (d) Training Loss

\end{tabular}}
\end{center}
\caption{Results of adversarial training considering the FGSM and PGD attacks when attacked during testing. Adversarial training with PGD is very effective, given enough capacity. `Natural' refers to test images that are not attacked. `Simple' and `wide' refer to the architecture used. A detailed analysis is given in the text. Figure adapted from~\cite{https://doi.org/10.48550/arxiv.1706.06083}.}
\label{fig:advres}
\end{figure}
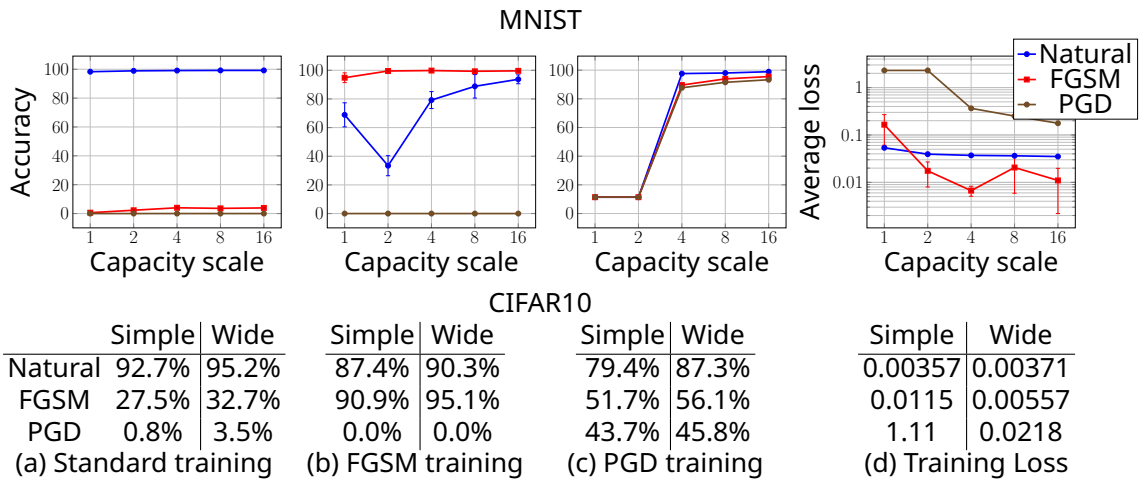

We would like to discuss Figure~\ref{fig:advres} from several aspects, as this extensive benchmark sheds light on many interesting tendencies in adversarial generalization.

\textbf{Impact of adversarial attacks.} The impact is very strong for standard training. Any kind of attack pushes accuracy to 0. FGSM is slightly weaker than PGD. PGD gives 0 accuracy for any capacity scale.

\textbf{Impact of adversarial training.} (b) If we train on FGSM samples, we become very robust to FGSM-attacked test inputs. The model accuracy for natural images decreases compared to standard training, especially for smaller capacities. There is always a trade-off, even if we mix attacked and natural images during training (which is not done here): adversarial training leads to a drop in natural accuracy, as we cannot be perfect at both. The method breaks down completely for PGD-attacked inputs that are very strong. (c) If we train on PGD samples, we get fairly nice adversarial accuracy. The method seems to work but only for high-capacity models. For low-capacity models, PGD training is useless. The higher the capacity, the more effective PGD training is. The general lesson of adversarial training: We really need a high-capacity model.

\textbf{Impact of model capacity.} For adversarial training, we need to be able to take care of much more complexity in the data distribution, which results in very complex decision boundaries. We must defend against attacks while learning the task itself, which is quite challenging.\footnote{In \emph{curriculum learning}, we are doing the exact opposite, which can help stabilize training.} The stronger the attack, the more apparent the need for a high-capacity model is.

\textbf{Transferability of attacks/defenses.} In general, to be able to defend against a particular attack, we need to train against it. Otherwise, the method often completely breaks down. PGD transfers very well to natural and FGSM samples, but only for large-capacity models. (a): most accurate on natural images. (b): most accurate on FGSM images. (c): most accurate on natural (!) images.

\subsubsection{Computational Complexity}
\label{ssec:complexity}

Adversarial training is perceived as a costly defense mechanism. Let us highlight where the main overhead is coming from and discuss complexities from various viewpoints.

\textbf{Per-batch complexity.} We define per-batch complexity as the number of forward and backward steps per batch. In vanilla training, we have one forward step and one backward step per batch. In adversarial training, we need \(T\) forward and backward steps for finding the adversarial PGD samples (batched), and one (forward, backward) pair for updating \(\theta\). Therefore, in total, we need \(T + 1\) forward and backward steps. (For FGSM, we naturally have \(T = 1\), but it does not work well for adversarial training.) \emph{For a single batch, adversarial training is \(T + 1\) times as expensive as vanilla training.}

\textbf{Training complexity.} Similarly to per-batch complexity, we define training complexity as the number of forward and backward steps \emph{per training}. As adversarial training is \(T + 1\) times as expensive as vanilla training for a single batch, the entire training is also \emph{at least} \(T + 1\) times more expensive. We might even need more epochs because we are solving a more complex problem. The apparent complexity is \(T + 1\) times as much as for vanilla training. The added complexity can arise from having to add more epochs.\footnote{If we also take into consideration that we might need a model of much higher capacity, the relative cost of adversarial training is even more extreme.} We generally control the complexity by varying the number of iterations \(T\) per attack.

\textbf{Inference complexity.} Inference complexity refers to the number of forward and backward steps per inference. We need one forward step and one backward step for both the vanilla-trained model and the adversarially trained model. After adversarial training, we use our model as usual. We can, e.g., open source it and let people use it without any overhead.

\begin{information}{Data Augmentation vs. Adversarial Training}
Is data augmentation considered adversarial training? Augmentation methods do not know anything about the model. The model can learn to be \emph{invariant} to flipping and other perturbations with data augmentation, but these are not considered adversarial attacks. Adversarial attacks are much stronger than simple data augmentation methods. Adversarial training also does not make the model invariant to many (possibly large-scale) perturbations, only to the adversarial attacks.
\end{information}

\subsection{Obfuscated Gradients: Breaking the defense Again!}

As we have seen in the previous sections, people came up with adversarial attacks and others made defenses against them. Of course, this was not a one-round process. In the field of adversarial robustness, there is a constant loop between attacks and defenses trying to improve upon each other. Let us consider attacks against types of previously popular defenses such as \emph{obfuscated gradients}.

\begin{definition}{Obfuscated Gradients}
Obfuscated gradient methods are a class of defense mechanisms designed so that the constructed defense necessarily causes gradient masking. We consider three types of obfuscated gradient methods~\cite{https://doi.org/10.48550/arxiv.1705.07263}:
\begin{itemize}
    \item \textbf{Methods shattering gradients}: Gradients are shattered when a defense makes the computation graph non-differentiable, introduces numeric instability, or otherwise causes a gradient to be non-existent or incorrect. 
    \item \textbf{Methods using stochastic gradients}: Stochastic gradients are caused by randomized defenses, where either the network itself is randomized, or the input is randomly transformed before being fed to the classifier, causing the gradients to become randomized.
    \item \textbf{Exploding and vanishing gradients}: These are often caused by defenses that consist of multiple iterations of neural network evaluation, feeding the output of one computation as the input of the next. This type of computation, when unrolled, can be viewed as an extremely deep neural network evaluation, which can cause vanishing/exploding gradients.
\end{itemize}
\end{definition}

2018 was a loud year for adversarial attacks and defenses. In particular, ICLR'18 was teeming with defense methods. The paper ``\href{https://arxiv.org/abs/1802.00420}{Obfuscated Gradients Give a False Sense of Security: Circumventing Defenses to Adversarial Examples}''~\cite{https://doi.org/10.48550/arxiv.1802.00420} from ICML'18 (which was three months later than ICLR'18) claimed:
\[\text{``7 of 9 ICLR'18 defenses do not work.''}\]
\newcolumntype{d}[1]{D{\%}{\%}{#1} }
\begin{table}
\small
\centering
\caption{
    Results of ICLR'18 defenses against the ICML'18 attack~\cite{https://doi.org/10.48550/arxiv.1802.00420} taken from the paper. ``Defenses denoted with * propose combining
adversarial training; we report here the defense alone [\ldots]. The fundamental principle behind the defense denoted with ** has 0\% accuracy; in practice, imperfections cause
the theoretically optimal attack to fail [\ldots].'' 
}
\label{tab:dontwork}
\begin{tabular}{p{9em}ccc}
  \toprule
    \textbf{Defense} & \textbf{Dataset} & \textbf{Distance} & \textbf{Accuracy} \\
  \midrule
    \citeauthor{buckman2018thermometer} & CIFAR & $0.031$ ($\ell_\infty$) & 0\%*\\
    \citeauthor{https://doi.org/10.48550/arxiv.1801.02613} & CIFAR & $0.031$ ($\ell_\infty$) & 5\%\\
    \citeauthor{DBLP:journals/corr/abs-1711-00117} & ImageNet & $0.005$ ($\ell_2$) & 0\%*\\
    \citeauthor{dhillon2018stochastic} & CIFAR & $0.031$ ($\ell_\infty$) & 0\%\\
    \citeauthor{xie2018mitigating} & ImageNet & $0.031$ ($\ell_\infty$) & 0\%*\\
    \citeauthor{https://doi.org/10.48550/arxiv.1710.10766} & CIFAR & $0.031$ ($\ell_\infty$) & 9\%*\\
    Samangouei \etal & MNIST & $0.005$ ($\ell_2$) & 55\%**\\
    \midrule
    \citeauthor{https://doi.org/10.48550/arxiv.1706.06083} & CIFAR & $0.031$ ($\ell_\infty$) & 47\%\\
    Na \etal & CIFAR & $0.015$ ($\ell_\infty$) & 15\%\\
 \bottomrule
\end{tabular}
\end{table}
As shown in Table~\ref{tab:dontwork}, after the authors' proposed attack, they mostly reached a very low (0 - 9\%) accuracy even when using the ICLR'18 defenses. Why were these obfuscated gradient methods so ineffective? These defenses are specifically targeted against gradient-based attacks. Obfuscated gradient methods break down the gradients (i.e., they make it malfunction). This misleads gradient-based attacks. The problem is that the model itself is \emph{still vulnerable} to most of these attacks using a simple trick. One can use slight modifications of gradient-based attacks to attack such defenses again, usually with great success.

\subsubsection{Breaking down gradients does not give us any guarantees.}

Even if the previously mentioned methods worked in making gradient-based adversarial attacks impossible, the model being safe is not equivalent to no gradient-based algorithm being able to find an attack. There can still be some adversarial image within the \(L_p\) ball (or neighborhood in general). When we break down the gradients, PGD cannot attack the image in the right way \emph{directly}. Using PGD naively results in a benign image, i.e., the network can still recognize it well.

\emph{The model is safe when there is absolutely no adversarial sample within the attack space.} If this is not guaranteed, there can still be some algorithm that can find the working attack.

\subsubsection{How can we make the gradients malfunction?}

One way to make the gradients malfunction is to transform the inputs before feeding them to the DNN. These are called input-transformation-based defenses. They apply image transformations (and possibly random combinations thereof) to the original input image. The idea is that if we just transform our image in different ways using a discrete set of transformations, that does not change the content of the image much, and if we have many variations of possible transformations (of which we select one at test time), that is supposed to be very effective against adversarial attacks. This is because adversarial attacks are minimal changes in the image, and if we are killing these small changes using transformations, the attack will probably not harm the model anymore. We want to remove adversarial effects from the input image before feeding the result to the DNN. As we will soon see, this intuitive reasoning is \emph{flawed} in most cases, as input-transformation-based defenses only work when considering chained random transformations with a combinatorial scaling of possibilities.

\subsubsection{Examples for Input Transformations}

\begin{figure}
    \centering
    \includegraphics[width=0.5\linewidth]{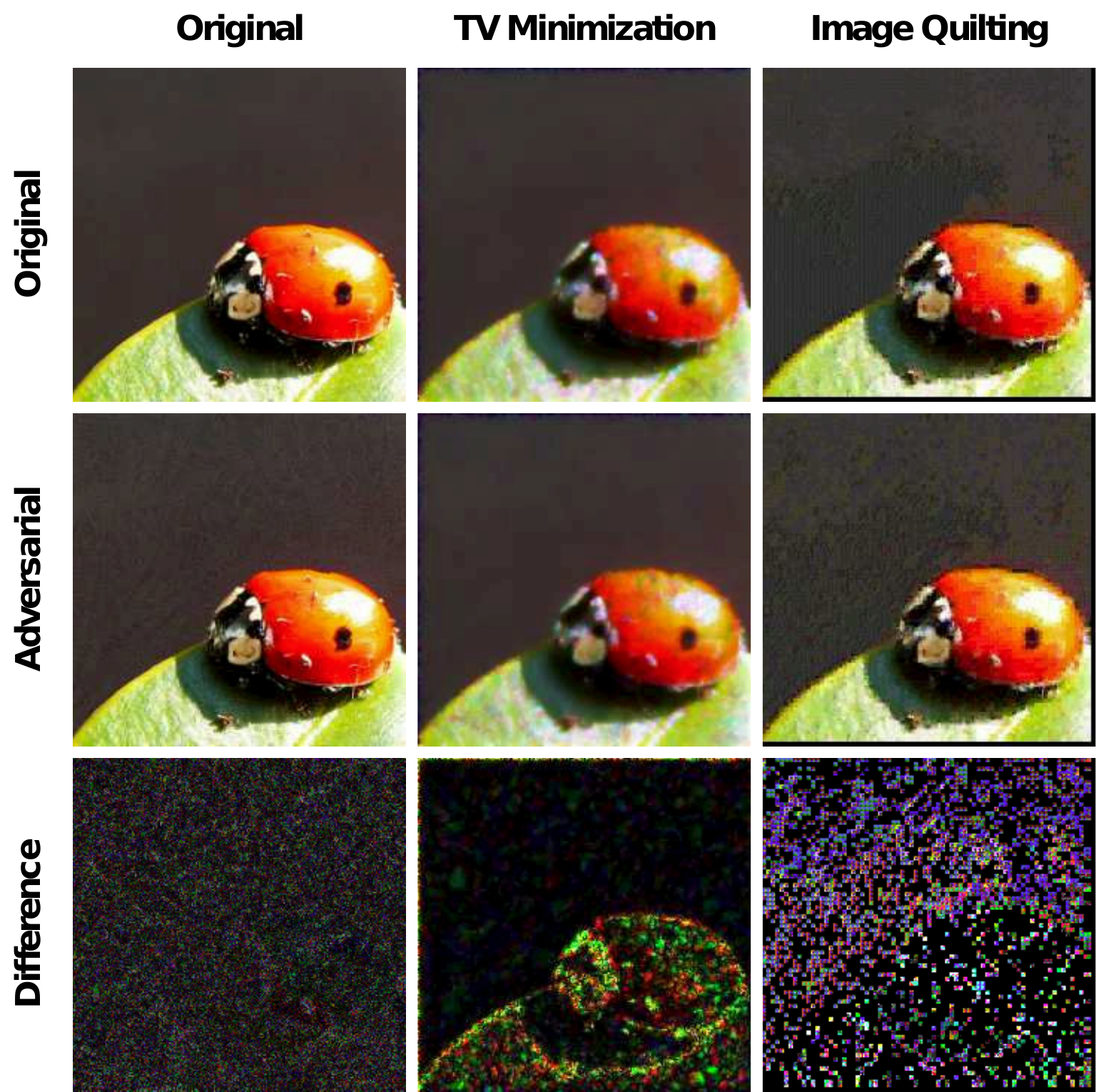}
    \caption{Example input transformations that can be used for gradient obfuscation. Figure taken from~\cite{DBLP:journals/corr/abs-1711-00117}.}
    \label{fig:input_transformations}
\end{figure}

Several examples of input transformations are shown in Figure~\ref{fig:input_transformations} that are detailed below.

\textbf{Cropping and rescaling of the original image.} We crop the part that contains the gist of what is going on in the image or rescale the image to the input size of the network.

\begin{definition}{Bit Depth}
The bit depth of an image refers to the number of colors a single pixel can represent. An 8-bit image can only contain \(2^8 = 256\) unique colors. A 24-bit image can contain \(2^{24} = 16,777,216\) unique colors.
\end{definition}

\textbf{Bit depth reduction.} Reducing the bit depth kills some information, but by doing this denoising (from the perturbation's viewpoint), we can also remove critical adversarial perturbations.

\textbf{JPEG encoding and decoding.} JPEG uses Discrete Cosine Transform (DCT). This is a typical transformation included in image viewers -- a natural way to defend against perturbations.

\textbf{Removing random pixels and inpainting them} The inpainting can be done, e.g., via TV (Definition~\ref{def:totalvariation}) minimization. When removing a boundary region, such inpainting will not result in a constant region (having the average value of the neighboring pixels) but rather a very smoothed version of the original image. (The boundary will be followed to some extent.)

\textbf{Image quilting.} This method reconstructs images using small patches from \emph{other images} in a database. The used patches are chosen to be similar to the original patches. These are also usually tiny. Before feeding it to the network, we replace the original image with the reconstruction.

\subsubsection{Results of naive FGSM, DeepFool~\cite{https://doi.org/10.48550/arxiv.1511.04599}, and Carlini-Wagner~\cite{https://doi.org/10.48550/arxiv.1608.04644} after input transformations}

\begin{figure}
    \centering
    \includegraphics[width=0.8\linewidth]{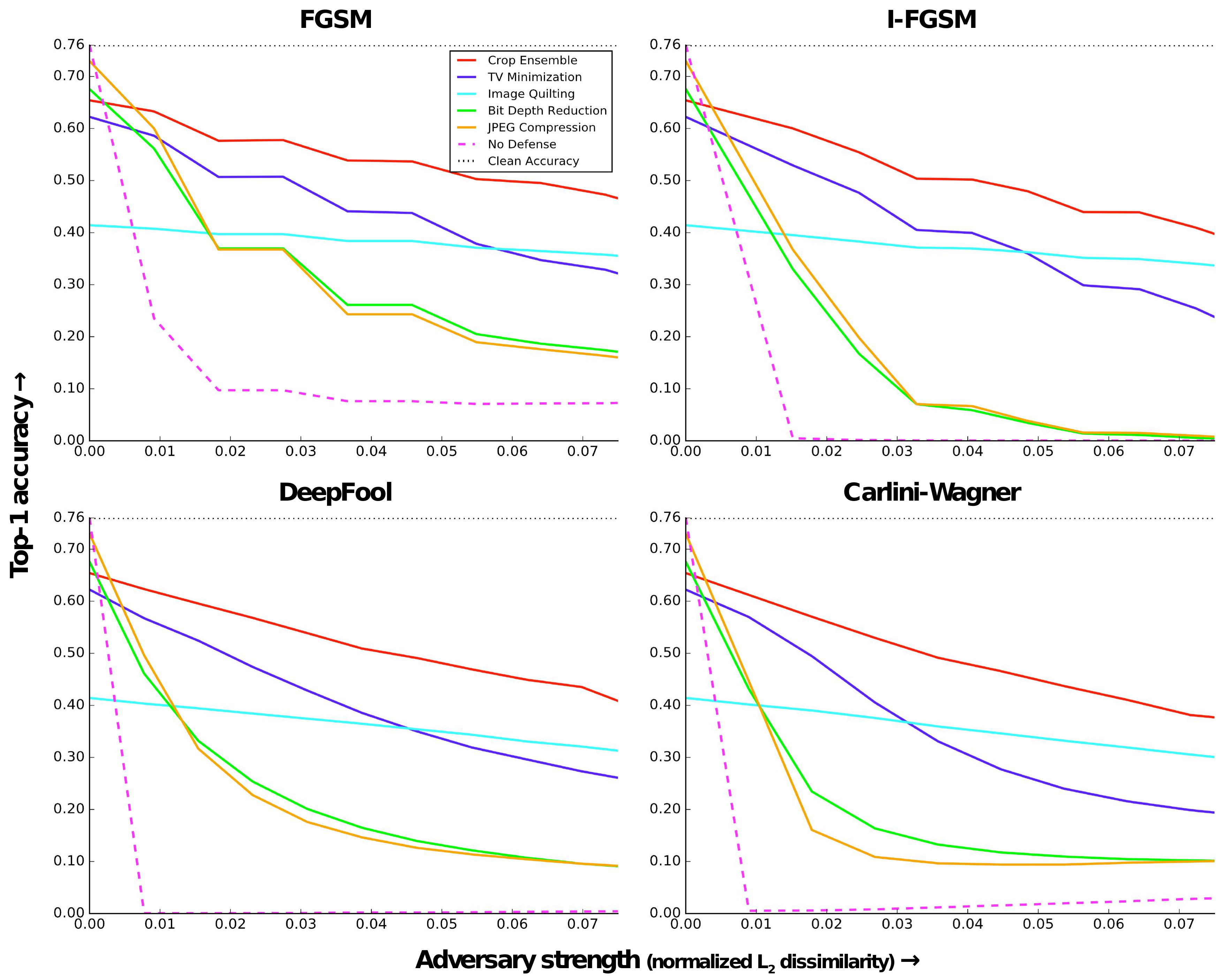}
    \caption{Top-1 classification accuracy of ResNet-50 on adversarial samples of various kinds. If we use no input transformations, the model's predictions break down completely. If we use the transformations listed in text \emph{individually}, the methods start failing. The stronger the adversary (i.e., the more \(L_2\) dissimilarity we allow), the better the attack methods do, but they still perform quite poorly. Figure taken from~\cite{DBLP:journals/corr/abs-1711-00117}.}
    \label{fig:carlini}
\end{figure}

To see whether the input transformation defense works, we take a look at the results of FGSM, DeepFool, and the Carlini-Wagner method in Figure~\ref{fig:carlini}. The general message of these results is that applying the previously listed input transformation to an image protects it against gradient-based adversarial attacks. We will see that this is an \emph{incorrect conclusion}.

\subsubsection{Straight-Through Gradient Estimator}

One of the reasons why previous methods using input transformations \emph{still fail to defend our networks} is the fact that we can still ``approximate'' the gradient of the defended model by using a straight-through gradient estimator.

\begin{definition}[label=def:stgrad]{Straight-Through Gradient Estimator}
The straight-through estimator generates gradients for a non-differentiable transformation as if the forward pass were the identity transformation; i.e., it lets the gradient flow through in the computational graph.
\end{definition}

A successful application of the straight-through estimator is attacking JPEG encoding/decoding defenses. The \emph{forward} pass is JPEG encoding and decoding, which is non-differentiable (because of quantization) but close to an identity mapping. In the \emph{backward} pass, we compute the gradient as if the forward were the identity mapping. The fact that this is a successful application of the estimator for an attack shows that this transformation only helped for gradient obfuscation because it made the computations non-differentiable. A Python example of a JPEG transformer is shown in Listing~\ref{lst:snippet}.

\begin{booklst}[Python example of a JPEG transformer.]{lst:snippet}
class JPEGTransformer(nn.Module):
    def forward(self, x):
        """JPEG encoding and decoding."""
        encoded_x = self.jpeg_encode(x)
        transformed_x = self.jpeg_decode(encoded_x)
        return transformed_x

    def backward(self, x, dy):
        """Straight-through estimator.
        Computes gradient as if self.forward = lambda x: x.
        """
        return dy
\end{booklst}

\subsubsection{The problem with naive gradient obfuscation methods}

When we attack models employing gradient obfuscation methods detailed above as a white box, we \emph{also have access to the transformations}.\footnote{Note, however, that BaRT~\cite{8954476} (\ref{sssec:bart}) works because it has such a large stochasticity internally.} First, assume that there is a single deterministic transformation. We do not have to know what this transformation precisely is; we just need access to it.

\begin{figure}
    \centering
    \includegraphics[width=0.8\linewidth]{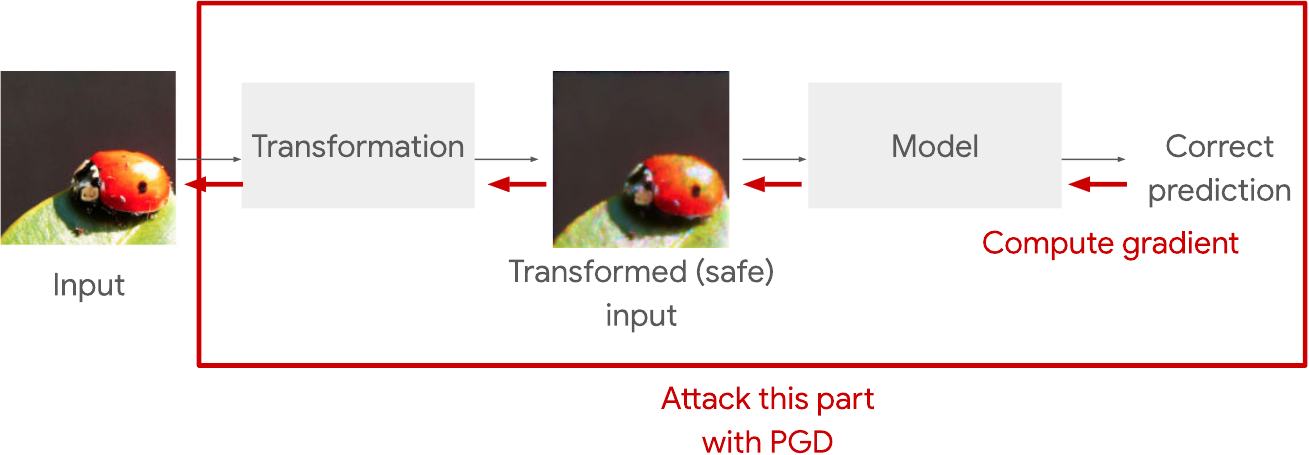}
    \caption{An easy way to circumvent obfuscated gradient defenses when the applied transformations are differentiable.}
    \label{fig:attackcrop}
\end{figure}

\textbf{Cropping and rescaling.} This is a differentiable transformation (cropping is just indexing, which is differentiable; rescaling is linear), therefore, we can attack the joint network, i.e., the entire pipeline. The defense does not work at all -- we can generate successful attacks again. This is depicted in Figure~\ref{fig:attackcrop}.

\begin{figure}
    \centering
    \includegraphics[width=0.8\linewidth]{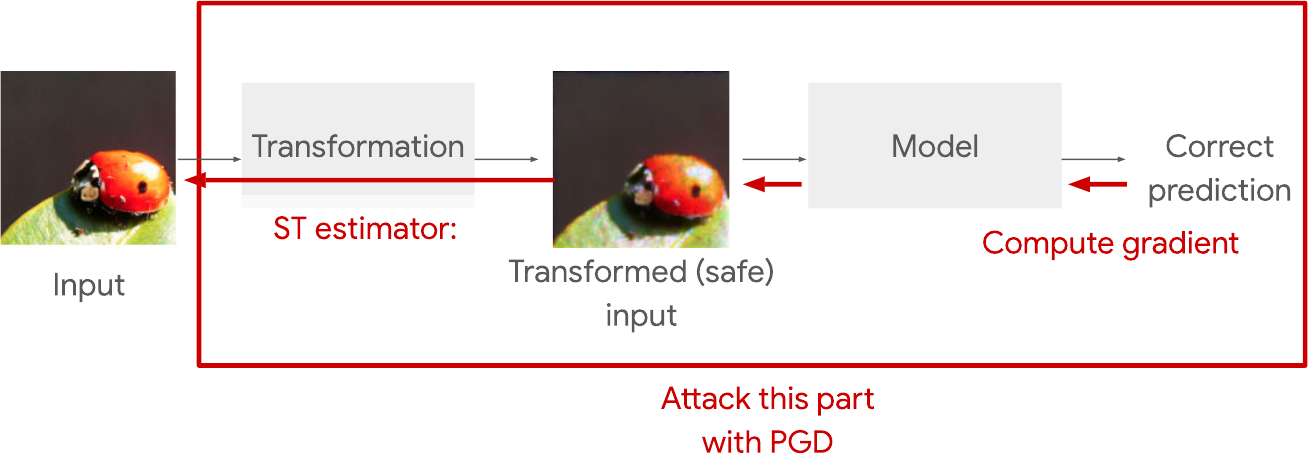}
    \caption{Circumventing obfuscated gradient defenses when the applied transformations are \emph{non-differentiable}, using the straight-through gradient estimator.}
    \label{fig:straightthrough}
\end{figure}

\textbf{Other discrete transformations.} For example, consider JPEG encoding and decoding. Such transformations are not differentiable. However, we can still ``differentiate through'' quantization layers, using the \emph{straight-through gradient estimator} (Definition~\ref{def:stgrad}). We can generate successful attacks again, as depicted in Figure~\ref{fig:straightthrough}.

\textbf{Mixture of random transformations.} Now, assume that there are multiple transformations, and one of them (or a mixture of them) is chosen randomly. When there is uncertainty in what transformation is used, the white box partially becomes a black box, as we do not know what is taking place in the random transformation. Still, for easier cases, the attacker can generate an attack that works for \emph{any} of the transformations (defenses) by performing \textit{Expectation over Transformations} (EoT).\footnote{Naive iterative gradient-based optimization would not work, as the gradients of the individual random transformations are simply too noisy.} 

\textbf{Expectation over Transformations (EoT).} One can observe that 
\[\nabla \nE_{t \sim T} f(t(x)) = \nE_{t \sim T} \nabla f(t(x)),\]
as the gradient and integral can be exchanged when a function is sufficiently smooth, which DNNs are. (For discrete transformations, we use the straight-through estimator anyway, which makes them also work.) The formula tells us that to attack the expected output of \(f\) \wrt \(t \sim T\), we take the gradient for each transformation and then take the expectation \wrt the transformations. This procedure can be trivially Monte Carlo estimated. We update the input \wrt the expected gradient's approximation iteratively. Python code for a simple EoT attack is given in Listing~\ref{lst:eot}.

With sufficient capacity for the attacker, the defense can become ineffective. This is pushing the limit of the capacity of the attacker. If the attacker has full capacity to address many possibilities for transformations at test time, we attack all of them simultaneously. The ICML'18 attack applies all the techniques we mentioned before. It destroys the defense that uses random transformations and makes the network have 0\% adversarial accuracy.

\begin{booklst}[Python pseudocode for generating an EoT attack. This is not PGD or FGSM, but they would work the same way with EoT.]{lst:eot}
def generate_eot_attack(x, model, transformation_list, num_samples):
    random_transformations = np.random.choice(transformation_list, num_samples)

    grad_eot = np.zeros_like(x)
    for transformation in random_transformations:
        y = model(transformation(x))
        grad_x = compute_input_gradient(y, x)

        # Approximate expectation by averaging.
        grad_eot += grad_x / num_samples

    return x + grad_eot
\end{booklst}

\subsection{Effectiveness of Adversarial Training}

Adversarial training (AT) does not introduce obfuscated gradients. It was hard for the ICML'18 method to attack adversarially trained models with greater attack success rates. AT is, therefore, an effective defense. Notably, the authors of~\cite{https://doi.org/10.48550/arxiv.1802.00420} use vanilla adversarial training without EoT. Performing EoT additionally would increase computation costs but would likely result in even stronger defenses.

\textbf{Note}: Even after adversarial training, there might still be some adversarial samples within the \(L_p\) ball -- we get no guarantees. However, adversarial training is still understood as a solid defense. The critical caveat of AT is that it is complicated to perform at scale. If we are dealing with an ImageNet scale, it is possible but also very impressive. The training time increases notably: adversarial training takes at least \(T + 1\) times as long as regular training (Subsection~\ref{ssec:complexity}), but here we also have to perform EoT, resulting in a triple \texttt{for} loop.

\subsection{Barrage of Random Transforms (BaRT)}
\label{sssec:bart}

As we have seen, we always have a loop of improvement in adversarial settings between attackers and defenders.
Once a defense with a mixture of random transformations is broken (e.g., EoT effectively beats a defender with a reasonable number of candidate transformations), the question naturally arises: What happens when the set of transformations is gigantic on the defense side? 

If the defender starts using random combinations of transformations, the number of possibilities grows exponentially as the number of individual transformations and the length of the transformation sequence grows.

The paper ``\href{https://openaccess.thecvf.com/content_CVPR_2019/papers/Raff_Barrage_of_Random_Transforms_for_Adversarially_Robust_Defense_CVPR_2019_paper.pdf}{Barrage of Random Transforms for Adversarially Robust Defense}''~\cite{8954476} was a ``reply'' to the EoT paper that introduced an enormous set of possible transformations.

\subsubsection{BaRT Method}

The method introduces ten groups of possible image transformations listed below.
\begin{itemize}
    \item Color Precision Reduction
    \item JPEG Noise
    \item Swirl
    \item Noise Injection
    \item FFT Perturbation
    \item Zoom
    \item Color Space
    \item Contrast
    \item Greyscale
    \item Denoising
\end{itemize}
Each group contains some number of transformations. In total, we have 25 transformations, each of which has parameters \(p\) that alter their behavior.

The choice of transformations is made as follows.
\begin{enumerate}
    \item Randomly select \(k\) out of \(n\) transforms where each transform by itself is randomized.
    \item Apply the selected transforms in a random sequence:
    \[f(x) = f(t_{\pi(1)}(t_{\pi(2)}(\dots(t_{\pi(k)}(A(x)))\dots))),\]
    where \(A\) is the adversary.
\end{enumerate}
Selecting the transformations randomly and applying them in a random sequence generates an exponential number of possibilities (\(n! / (n - k)!\)) that still do not change the semantic meaning of the image. Even after applying all transformations, the model can still recognize the objects pretty well. However, the sheer number of possibilities makes it very hard for the attacker to prepare against all kinds of defenses. It must have a large enough capacity and many samples are required to Monte Carlo sample the expectation. To establish resilience against such input transformations, they are applied both during training and inference. Therefore, this is \emph{not} a post-hoc algorithm.

The method has some overhead in the cost of training, but it boils down to selecting an input transformation sequence with can be done very efficiently on the CPU. The overhead is, therefore, similar to that of data augmentation. One can also influence this overhead by changing how often the transformations are resampled.

\subsubsection{BaRT Results}
\begin{figure}
    \centering
    \includegraphics[width=0.5\linewidth]{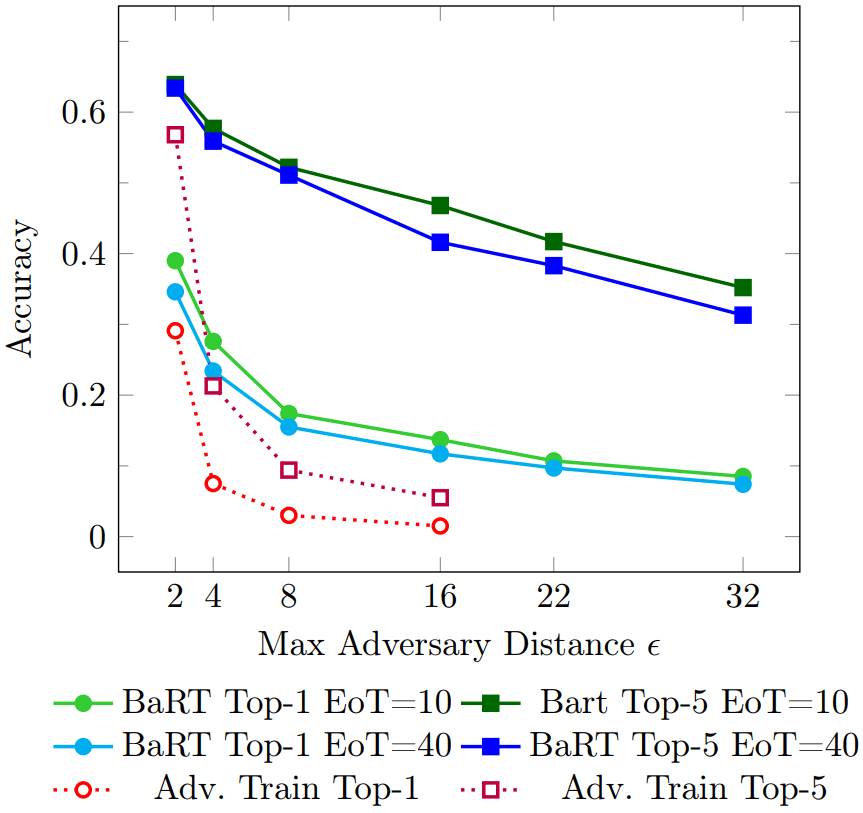}
    \caption{BaRT defends a model against PGD (which is not surprising). BaRT also defends a model against the ICML'18 methods with EoT (10 or 40 samples), designed to break gradient obfuscations. Using BaRT, performance does not drop too much by increasing the max adversary distance \(\epsilon\). It is even more effective than adversarial training -- the attacker cannot push the scores down to 0, not even for \(\Vert x - \hat{x} \Vert_\infty < 32\). (!) Top-k refers to top-k accuracy. Figure taken from~\cite{8954476}. }
    \label{fig:bartres}
\end{figure}

The results of BaRT are shown in Figure~\ref{fig:bartres}. The key message here is that \emph{BaRT is one of the SotA adversarial defense methods even in 2023.}

\subsection{Certified defenses}

Let us discuss \emph{certifications of robustness}. Certified defense methods make sure there is \emph{no successful attack} in the strategy space (e.g., the \(L_p\) ball) under some assumptions. The ``\href{https://arxiv.org/abs/1801.09344}{Certified Defenses against Adversarial Examples}''~\cite{https://doi.org/10.48550/arxiv.1801.09344} paper can give certifications of robustness by considering many simplifying assumptions for the network and the adversarial objective.

The typical chain of thought for certified defenses is to come up with a trainable objective, and then show that solving this trainable objective will ensure that there is no worse attack than a certain type.

The authors consider a binary classification setting and a two-layer neural network where the score is calculated as
\[f(x) = V\sigma(Wx).\]
Here, \(V \in \nR^{2 \times m}\), \(W \in \nR^{m \times d}\), and \(\sigma\) is an elementwise non-linearity with bounded gradients to \([0, 1]\), e.g., ReLU or sigmoid. Notably, the authors calculate the score of both positive and negative classes instead of considering a single score for the ease of formalism. A certificate of defense is given by bounding the margin of the incorrect class over the correct one for any adversarial perturbation inside the \(L_\infty\) 
\(\epsilon\)-ball centered at a particular input \(x\), denoted by \(B_\epsilon(x)\). Further details are discussed in Information~\ref{inf:cert_def}.

\begin{definition}[label=def:lineintegrals]{Fundamental Theorem of Line Integrals}
Consider a parametric curve \(r: [a, b] \rightarrow \nR^d\) and a differentiable function \(f: \nR^d \rightarrow \nR\). Then
\[\int_a^b \underbrace{\left\langle \nabla f(r(t)), r'(t) \right\rangle\ dt}_{\left\langle \nabla f(r(t)), dr \right\rangle,\ dr = r'(t)dt} = f(r(b)) - f(r(a)).\]
In words: The integral of directional derivatives along the curve \(r\) of the function \(f\) is equal to the difference of boundary values of \(f\). In short, the shape of the curve \(r\) does not matter.

\medskip

\textbf{Connection to single variable calculus}: The fundamental theorem of integrals states that for a differentiable \(f: \nR \rightarrow \nR\):
\[\int_a^b f'(x)\ dx = f(b) - f(a).\]
In this case, we have a single possible way from \(a\) to \(b\), which is generalized for line integrals.
\end{definition}

\begin{information}{Formulation of the Certified Defenses Method} \label{inf:cert_def}
The authors consider the following worst-case adversarial attack:
\(A_\mathrm{opt}(x) = \argmax_{\tilde{x} \in B_\epsilon(x)}\tilde{f}(\tilde{x}),\)
where
\[\tilde{f}(x) := \underbrace{f^1(x)}_{\text{score of incorrect label}} - \underbrace{f^2(x)}_{\text{score of correct label}}.\]
The attack is successful if
\(\tilde{f}(A_\mathrm{opt}(x)) > 0\) as the incorrect class is predicted.

\medskip

We derive the following upper bounds on the severity of any adversarial attack \(A(x)\):
\[\tilde{f}(A(x)) \overset{(i)}{\le} \tilde{f}(A_\mathrm{opt}(x)) \overset{(ii)}{\le} \tilde{f}(x) + \epsilon \max_{\tilde{x} \in B_\epsilon(x)}\Vert \nabla \tilde{f}(\tilde{x}) \Vert_1 \overset{(iii)}{\le} \tilde{f}_\mathrm{QP}(x) \overset{(iv)}{\le} \tilde{f}_\mathrm{SDP}(x).\]

(i) Arises from the optimality of \(A_\mathrm{opt}\). (ii) leverages the fundamental theorem of line intergrals (Definition~\ref{def:lineintegrals}):
\begin{align*}
\tilde{f}(\tilde{x}) &= \tilde{f}(x) + \int_0^1 \nabla \tilde{f}(t\tilde{x} + (1 - t)x)^\top(\tilde{x} - x)dt\\
&\le \tilde{f}(x) + \max_{\tilde{x}' \in B_\epsilon(x)} \epsilon \Vert \nabla \tilde{f}(\tilde{x}) \Vert_1
\end{align*}
where \(\tilde{x} \in B_\epsilon(x)\) and the inequality holds because the linear interpolation \(t\tilde{x} + (1 - t)x\) of two elements \(x\) and \(\tilde{x}\) of \(B_\epsilon(x)\) is also an element of \(B_\epsilon(x)\) for any \(t \in [0, 1]\). In (iii), \(\tilde{f}_\mathrm{QP}(x)\) denotes the optimal value of a (non-convex) quadratic program. This is a specific bound for two-layer networks where \(\tilde{f}(x) = f^1(x) - f^2(x) = v^\top \sigma(Wx)\) with \(v := V^1 - V^2\) being the difference of last-layer weights of the correct and incorrect class. In this specific case, we upper-bound \(\tilde{f}(x) + \epsilon \max_{\tilde{x} \in B_\epsilon(x)}\Vert \nabla \tilde{f}(\tilde{x}) \Vert_1\) by noting that for \(\tilde{x} \in B_\epsilon(x)\):
\[\Vert \nabla \tilde{f}(\tilde{x}) \Vert_1 = \Vert W^\top \operatorname{diag}(v)\sigma'(W\tilde{x})\Vert_1 \le \max_{s \in [0, 1]^m} \Vert W^\top \operatorname{diag}(v)s\Vert_1 = \max_{s \in [0, 1]^m, t \in [-1, 1]^d} t^\top W^\top \operatorname{diag}(v)s\]
where the last equality shows a different way to write the \(L_1\) norm. Therefore, \[\tilde{f}(x) + \epsilon \max_{\tilde{x} \in B_\epsilon(x)} \Vert \nabla \tilde{f}(\tilde{x}) \Vert_1 \le \tilde{f}(\tilde{x}) + \epsilon \max_{s \in [0, 1]^m, t \in [-1, 1]^d} t^\top W^\top \operatorname{diag}(v)s =: \tilde{f}_\mathrm{QP}(x).\] The reason why we do not stop here is that this quadratic program is still a non-convex optimization problem. This is why we turn to (iv), which gives a \emph{convex} semidefinite bound. First, the authors of~\cite{https://doi.org/10.48550/arxiv.1801.09344} reparameterize the optimization problem in \(s\) as
\[\tilde{f}_\mathrm{QP}(x) := \tilde{f}(x) + \epsilon \max_{s \in [-1, 1]^m, t \in [-1, 1]^d} \frac{1}{2} t^\top W^\top \operatorname{diag}(v)(\bone + s)\]
where \(\bone\) is a vector of ones. Then, one needs to define auxiliary vectors and matrices to obtain the form of a semidefinite program:
\begin{align*}
y &:= \begin{pmatrix}1 \\ t \\ s\end{pmatrix}\\
M(v, W) &:= \begin{bmatrix}0 & 0 & \bone^\top W^\top\operatorname{diag}(v)\\ 0 & 0 & W^\top\operatorname{diag}(v)\\ \operatorname{diag}(v)^\top W\bone & \operatorname{diag}(v)^\top W & 0\end{bmatrix}.
\end{align*}
Now, we rewrite \(\tilde{f}_\mathrm{QP}(x)\) as
\[\tilde{f}_\mathrm{QP}(x) = \tilde{f}(x) + \epsilon \max_{y \in [-1, 1]^{(m + d + 1)}} \frac{1}{4}y^\top M(v, W)y = \tilde{f}(x) + \frac{\epsilon}{4} \max_{y \in [-1, 1]^{(m + d + 1)}} \left\langle M(v, W), yy^\top \right\rangle.\]
Finally, we note that \(\forall y \in [-1, 1]^{(m+d+1)}\), \(yy^\top\) is a positive semidefinite matrix\footnote{This statement holds for arbitrary vectors \(y \in \nR^n\).} and the diagonal of \(yy^\top\) is a vector of ones. Defining \(P = yy^\top\), we obtain the convex semidefinite program
\[\max_{y \in [-1, 1]^{(m + d + 1)}} \frac{1}{4}\left\langle M(v, W), yy^\top \right\rangle \le \max_{P \succeq 0, \operatorname{diag}(P) \le 1} \frac{1}{4}\left\langle M(v, W), P\right\rangle\]
where the notation \(P \succeq 0\) refers to \(P\) being positive semidefinite,
which allows us to define \(\tilde{f}_\mathrm{SDP}(x)\) as
\[\tilde{f}_\mathrm{SDP}(x) := \tilde{f}(x) + \frac{\epsilon}{4} \max_{P \succeq 0, \operatorname{diag}(P) \le 1} \left\langle M(v, W), P\right\rangle.\]

\medskip

Notably, the optimization problem in \(f_\mathrm{SDP}(x)\) is fixed in the neural network weights \(v\) and \(W\) and does not depend on \(x\). Therefore, obtaining it is very much feasible, as we only need to calculate the input-agnostic upper bound once for each model.

\medskip

Sadly, our story does not end here. One may assume that the post-hoc application of the above upper bound is enough. While we can indeed calculate such a certificate post-hoc, it might be arbitrarily loose. Regular cross-entropy training encourages \(\tilde{f}(x)\) to be large in magnitude on training samples. However, the term \(\frac{\epsilon}{4} \max_{P \succeq 0, \operatorname{diag}(P) \le 1} \left\langle M(v, W), P\right\rangle\) is \emph{not encouraged to be small} to tighten the bound. One might naively consider the following, non-post-hoc objective instead to obtain tighter bounds:
\[W^*, V^* = \argmin_{W, V} \sum_n \cL(V, W; x_n, y_n) + \lambda \max_{P \succeq 0, \operatorname{diag}(P) \le 1} \left\langle M(v, W), P\right\rangle\]
where \(\lambda\) controls the regularization strength. Using this objective is clearly infeasible, however: For each gradient step, we need the solution to the inner semidefinite program. Without going too much into detail, one can obtain a \href{https://en.wikipedia.org/wiki/Duality_(optimization)}{\emph{dual formulation}} of the semidefinite program to eliminate the inner optimization problem. First, we state the dual formulation:
\[\max_{P \succeq 0, \operatorname{diag}(P) \le 1} \left\langle M(v, W), P\right\rangle = \min_{c \in \nR^{(d + m + 1)}} (d + m + 1) \cdot \lambda^+_\mathrm{max}(M(V, W) - \operatorname{diag}(c)) + \sum_i max(c_i, 0)\]
where \(\lambda_\mathrm{max}^+(\cdot)\) calculates the maximal eigenvalue of the input matrix or returns zero if all eigenvalues are negative. How can we use this to eliminate the inner optimization problem? As the inner problem becomes the unconstrained minimization of an objective in \(c \in \nR^{(d + m + 1)}\), we optimize \(c\) in the same optimization loop as parameters \(V\) and \(W\). Therefore, we only have an additional parameter we have to optimize over and we can still use gradient-based unconstrained optimization.

\medskip

This leads us to the final objective: We optimize
\begin{align*}
&(W^*, V^*, c^*)=\\
& \argmin_{W, V, c}\sum_n \cL(V, W; x_n, y_n) + \lambda \cdot \left[(d + m + 1) \cdot \lambda^+_\mathrm{max}\left(M(V, W) - \diag\left(c\right)\right) + \sum_i max(c_i, 0)\right]
\end{align*}
which can be done quite efficiently. This encourages the network to be robust while also allowing us to provide a certification of robustness.

\medskip

Given \(V[t], W[t]\) and \(c[t]\) values at iteration \(t\) solving the above optimization problem, one obtains the following guarantee for any attack \(A\):
\[\tilde{f}(A(x)) \le \tilde{f}(x) + \frac{\epsilon}{4} \left[D \cdot \lambda^+_\mathrm{max}\left(M(V[t], W[t]) - \diag\left(c[t]\right)\right) + + \sum_i max(c[t]_i, 0)\right].\]
We get a certificate of the defense: Whatever perturbation there is in the \(L_\infty\) \(\epsilon\)-ball, the loss is bounded from above. This is theoretically meaningful but not yet in practice: The study is confined to 2-layer networks.
\end{information}

\subsection{History and Possible Future of Adversarial Robustness in ML}

A Coarse Timeline of Adversarial Robustness is listed below.
\begin{itemize}
    \item \textbf{First attack}: L-BFGS attack (2014)~\cite{https://doi.org/10.48550/arxiv.1312.6199}. This is a complicated method that does not work too well.
    \item \textbf{First practical attack}: FGSM attack (2014). As discussed previously, this is a straightforward method that works reasonably well.
    \item \textbf{Stronger iterative attack}: DeepFool (2015)~\cite{https://doi.org/10.48550/arxiv.1511.04599}.
    \item \textbf{First defense}: Distillation (2015)~\cite{https://doi.org/10.48550/arxiv.1503.02531}. Training labels of the distilled network are the predictions of the initially trained network. Both networks are trained using temperature \(T\).
    \item \textbf{First black-box attack}: Substitute model (2016).
    \item \textbf{Strong attack}: PGD (2017).
    \item \textbf{Strong defense}: Adversarial Training (2017).
    \item \textbf{First detection mechanisms}: Adversarial input detection methods (2017). Instead of making the model stronger, we train a second model to detect adversarial patterns. However, the attackers can also generate patterns that avoid these detections (fool the detector in a white-box fashion).
    \item ``It is easy to bypass adversarial detection methods.'' (2017)~\cite{https://doi.org/10.48550/arxiv.1705.07263}.
    \item Defenses at ICLR'18 (2018): input perturbation, adversarial input detection, adversarial training, etc.
    \item ``Defenses at ICLR'18 are mostly ineffective.'': Obfuscated gradients (2018).
    \item Barrage of Random Transforms (2019).  One only needs to apply many transformations sequentially in a random fashion.
\end{itemize}

2020- : We should \emph{stop} the cat-and-mouse game between attacks and defenses. It is a dead end. We are spiraling around attack and defense. Diversifying and randomizing (e.g., BaRT) is a promising approach. However, the constant spiral of whether the attacker or defender has more capacity to generate attacks/defenses is not very interesting from an academic perspective.

There are two main alternatives one may choose to work on:
\begin{itemize}
    \item Certified defenses making sure there is \emph{no attack} in the \(L_p\) ball.
    \item Dealing with realistic threats rather than unrealistic worst-case threats.
\end{itemize}

\subsection{Towards Less Pessimistic defenses}

Usually, the considered attacks are way too strong. Instead, we should work more on (1) defenses against black-box attacks, which is an exciting subfield of adversarial attacks, or (2) defenses against non-adversarial, non-worst-case perturbations (OOD gen., domain gen., cross-bias gen.). These are what we have learned in the previous chapters. Many researchers who used to study adversarial perturbations are now working on general OOD generalization and naturally shifting distributions.

\clearpage

\chapter{Explainability}
\section{Introduction}

\begin{figure}
    \centering
    \includegraphics[width=0.6\linewidth]{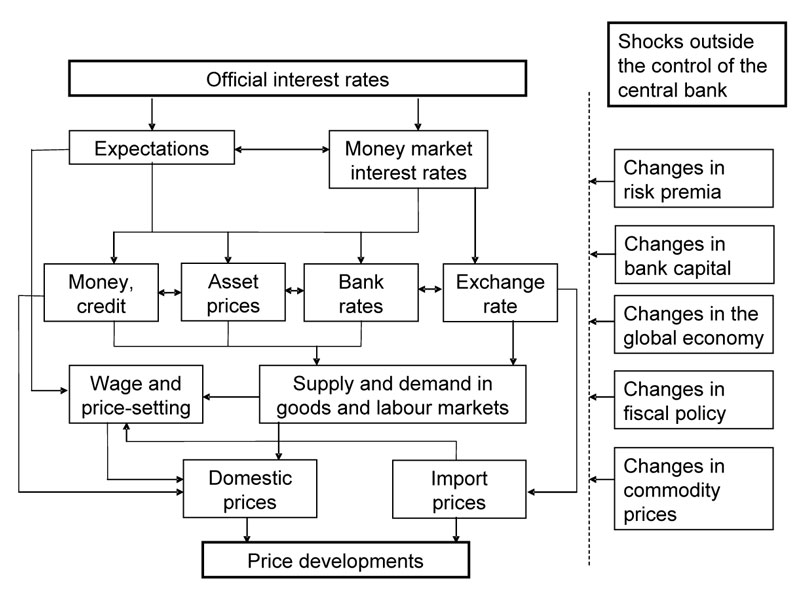}
    \caption{Schematic illustration of the main transmission channels of monetary policy decisions, taken from~\cite{asset}. Controlling price developments requires fine-grained control.}
    \label{fig:control}
\end{figure}

If we understand a system and its underlying mechanisms well, we can use the system to control something. An example is the economy: As we control official interest rates,\footnote{The central bank is directly in control of this through determining official interest rate policies. Similarly, other policy rates and asset purchases have a large effect on how prices develop.} we control the amount of money in the market. Official interest rates affect many components of the economy (e.g., bank rates, exchange rates, or asset prices) and finally also affect price developments (e.g., domestic prices and import prices), all through a highly complex procedure. This is illustrated in detail in Figure~\ref{fig:control}. There is always a new situation coming (shocks outside the central bank's control). One cannot solely rely on experience, as we do not have so much history of the market to base our decisions on previous experiences when we face new situations. It is essential to \emph{know what is happening in the system} to perform control.

When faced with a black box system, we do not understand its inner workings. For example, we do not exactly understand why a self-driving car is following the road in one case but not following it in the other. As we cannot control the system precisely, we cannot fix it when it is malfunctioning.

\subsection{Ways to Control Undefined Behavior}

\begin{definition}{Explanation}
An explanation is an answer to a \emph{why}-question.~\cite{DBLP:journals/corr/Miller17a}
\end{definition}

\begin{definition}{Interpretability}
Interpretability is the degree to which an observer can understand the cause of a decision.~\cite{biran2017explanation}
\end{definition}

\begin{definition}{Explainability}
Explainability is post-hoc interpretability.~\cite{lipton2018mythos} It is the degree to which an observer can understand the cause of a decision after receiving a particular explanation.
\end{definition}

\begin{definition}{Justification}
A justification explains why a decision/prediction is good but does not necessarily aim to explain the actual decision-making process.~\cite{biran2017explanation} It is also not necessarily sound.
\end{definition}

Two general ways exist to control undefined behavior in OOD (novel) situations: using unit tests and fixing only after understanding.

\textbf{An infinite list of unit tests and data augmentation.} We were looking into this in previous sections. In particular, in Section~\ref{sssec:identify}, we saw how we can identify spurious correlations in our model, and in Section~\ref{sssec:overview}, we saw how we can \emph{incorporate} samples from different domains (e.g., unbiased samples) into the training procedure to obtain more robust models. Our goal is to let a model work well in any new environment. For evaluation purposes, we introduce a new evaluation set every time, e.g., introduce ImageNet-\{A, B, C, D, \dots\}. A natural next step is to augment our network's \emph{training} with samples from ImageNet-\{A, B, C, D, \dots\} and seek new evaluation sets. We are sequentially conquering different unit tests, hoping that we eventually get a strong system that works well in any situation. But is that \emph{really} going to happen?

\textbf{Understand first, fix after.} The goal here is the same as before: make a model that works well in any new environment. For evaluation, we examine cues utilized by the model (explainability). If we understand that the model is not utilizing the right cue for recognition, then we have a way to control this. We regularize the model later to choose cues that are generalizable. We do \emph{not} evaluate whether the model works well on ImageNet-\{A, B, C, D, \dots\}, as we directly control the used cues. We regularize the model to choose generalizable cues (using feature selection). This seems to be the more scalable approach. An infinite number of unit tests will probably not solve all our problems.

\subsection{Explainability as a Base Tool for Many Applications}

There are numerous applications that require the \emph{selection of the `right' features}. In fairness, we wish to eliminate, e.g., demographic biases which requires us to select features that do not take demographic aspects into account. In the field of robustness, we also have to select powerful features to combat distribution shifts.

There are also many applications that require \emph{better understanding} and \emph{controllability}. One example is ML for science where the aim is to discover scientific facts from (usually) high-dimensional data. Here, understanding and control is the \emph{end goal}. We can also consider the task of quickly adapting ML models to downstream tasks (e.g., GPT-3 and other LLMs). If we understood what GPT-3 or other LLMs do/know, we could probably quickly adapt them to downstream tasks by only choosing the parts or subsystems responsible for useful utilities for the downstream task. In that case, we might not even need any fine-tuning.

\begin{definition}{Attribution}
Attribution can be understood as the assignment of a reason for a certain event. It is often used in the field of Explainable Artificial Intelligence (XAI) to describe attributing factors to a model's behavior in an explanation. Such factors are often selected from (1) the input's features we are explaining (be it the raw input features or intermediate feature representations of NNs), (2) the elements of the training set, or (3) the model's parameters.
\end{definition}

\begin{definition}{Explanation by Attribution}
An explanation by attribution method is a function that takes an input \(x\) and a model \(f\) and outputs the explanation of which features/training samples/parameters contribute most to the prediction \(f(x; \theta)\) of the model for input \(x\).
\end{definition}

Several applications require a \emph{better understanding of the training data}. Consider the detection of private information in a training dataset. Instead of attributing to the test data, we can also reason back to the training data. For example, if our model seems to have learned something private from user data and users can even be identified based on this information, it would be very informative to be able to trace back specific predictions to the training data (and remove private data from the training set or make sure that such information cannot leak). If an LLM outputs something that looks like someone's home address, tracking down where this information came from in the training set is very informative for those who audit the training data. Attributing to the original authors in the training data is an increasingly popular and useful task. Example questions include ``What prior art made DALL-E generate a certain image? What authors can be attributed?'' XAI can give answers to such questions.

Finally, let us discuss applications requiring \emph{greater trust}. One example is ML-human expert symbiosis where a human expert is working with ML to generate better outcomes. Trust is also needed in high-stakes decision areas: for example, finance, law, and medical applications.

\subsection{Explainability as a Data Subject's Right}

Nowadays, explanations are stipulated in law~\cite{Goodman_2017}. XAI has close ties to national security -- the research field originated from the \href{https://www.darpa.mil/program/explainable-artificial-intelligence}{DARPA XAI program} of the US, in 2016. The EU also considered AI legislation crucial -- GDPR has an article about automated decision-making, and the \href{https://artificialintelligenceact.eu}{AI act} is an even newer \emph{proposal} of harmonized rules on general artificial intelligence systems. A common theme of AI legislation is that suitable measures are needed to safeguard a data subject's rights, freedom, and legitimate interests. Data subjects have a right to request explanations in automated decision-making and to obtain human intervention. Critical decisions are made about humans based on automatized systems (ML) using their personal data, e.g., in CV preselection or loan applications. The data owner has the right to know which feature has caused, e.g., a loan rejection.

\subsubsection{Three Key Barriers to Transparency}

There are mainly three barriers to transparency. Let us briefly discuss these.

\textbf{Intentional concealment.} For example, a bank might intentionally conceal their decision procedure for loan rejection. Decision-making procedures are often kept from public scrutiny.

\textbf{Gaps in technical literacy.} For example, even if the bank is enabling insight into its decision-making procedure, people may not be able to understand the raw code. For most people, reading code is insufficient.

\textbf{Mismatch between actual inner workings of models and the demands of human-scale reasoning and styles of interpretation.} This is perhaps the most technical aspect this book seeks answers to. Human-comprehensibility was highlighted as a crucial aspect of XAI methods by several researchers~\cite{DBLP:journals/corr/Miller17a,molnar2020interpretable,belle2021principles}. If we are showing the weights of a model to a human, it is unlikely that they see some meaning. We need summarization, dimensionality reduction, and attachment of human-interpretable concepts.

We have answered ``Why is an explanation needed?'' Let us turn to ``When is an explanation needed?''

\subsection{When is an explanation needed?}

The following points are inspired by~\cite{https://doi.org/10.48550/arxiv.1702.08608,keil2006explanation}. Explanations may highlight an incompleteness/problem. In particular, explanations are typically required when something does not work as expected. When everything is working well, we usually do not question why something is working. When something does not work, we start raising questions.

\subsection{When is an explanation \emph{not} needed?}

First, we discuss a list of examples from~\cite{https://doi.org/10.48550/arxiv.1702.08608}. We also argue why this list of examples might not be descriptive enough.
\begin{itemize}
    \item \textbf{Ad servers.} Our remark is that it is a request of society in general that they should be prompted for consent if they want to see targeted ads, and also to gain insight into how profiling works.
    \item \textbf{Postal code sorting.} Even though in general, society might not care much about the inner workings of post offices, explanations might still be needed for debugging such sorting systems or for unveiling potential security risks.
    \item \textbf{Aircraft collision avoidance systems.} Again, explanations for such systems are generally not requests of society. Still, the aviation company must be in control of all situations that might arise, and for that, explanations are great tools.
\end{itemize}
The above list lacks \emph{recipients}. Whether an explanation is needed in a certain situation depends on the \emph{explainee}. A person not using the internet might, indeed, not care about ad servers. Similarly, a person who does not work as a developer for a post office might not need explanations about the sorting algorithm. Still, we can almost always find target groups for explanations about any topic.

The general reasoning of~\cite{https://doi.org/10.48550/arxiv.1702.08608} is sound: generally, we might not need explanations when
\begin{enumerate}
    \item there are no significant consequences for unacceptable results, or when
    \item the problem is sufficiently well-studied and validated in real applications that we \emph{trust} the system's decision (even if the system is not perfect).
\end{enumerate}
However, explanations are always great tools for exploratory analysis.

\section{Human Explanations}

\subsection{How do humans explain to each other?}

We discuss Tim Miller's work, titled ``\href{https://arxiv.org/abs/1706.07269}{Explanation in Artificial Intelligence: Insights from the Social Sciences}''~\cite{DBLP:journals/corr/Miller17a}.\footnote{This is a highly recommended work for those working or wishing to work in XAI.} According to~\cite{malle2006mind}, people ask for explanations for two main reasons:
\begin{enumerate}
    \item \textbf{To find meaning.} To reconcile the contradictions or inconsistencies between elements of our knowledge structures. We are trying to figure out at which principle we have contradictions. There are often contradictions between our understanding and the status quo in the outside world.
    \item \textbf{To manage social interaction.} To create a shared meaning of something, change others' beliefs and impressions, or influence their actions. Example questions include ``Why am I doing this? Why are you doing this?'' But also: ``If I believe what you are doing has a greater cause, I can also align my action to what you are doing.''
\end{enumerate}
Both are important for XAI systems.
\begin{itemize}
    \item \textbf{Finding meaning in XAI.} ``Why is this model not doing as I expect? Where is this inconsistency coming from?''
    \item \textbf{The social aspect of XAI.} We want to be able to share our way of thinking with the machine, and we expect it also to be able to do the same.\footnote{The field of Human-AI Interaction works on such methods. One possible way of knowledge exchange is through textual discussions, as seen in LLMs.}
\end{itemize}

\subsubsection{Human-to-human explanations are \dots}

\textbf{\dots contrastive.} Explanations are sought in response to particular counterfactual cases. People usually do not ask, ``Why did event \(P\) happen?'', they ask, ``Why did event \(P\) happen instead of some event \(Q\)?'' Even if the apparent format is the former, it usually \emph{implies} a hidden foil (i.e., the alternative case). \textbf{Example}: For the question ``Why did Elizabeth open the door?'', there are many possible foils. (1) ``Why did Elizabeth open the door, rather than leave it closed?'' -- a foil against the action. (2) ``Why did Elizabeth open the door rather than the window?'' -- a foil against the subject. (3) ``Why did Elizabeth open the door, rather than Michael opening it? -- a foil against the actor. A brief criticism of XAI is that the questions asked often do not have any foil in mind in general. We ask questions like, ``Why was this image categorized as \(A\)?'' It would be perhaps less ambiguous to ask, ``Why is this image categorized as \(A\), not \(B\)?'' This formulation makes the foil clear. Another way to extend the question: ``Will this image still be categorized as \(A\) even if the image is modified?'' We will see that this kind of question is implied in many XAI systems. In a sense, input gradients are asking such contrastive questions.

\textbf{\dots selective.} When someone asks for an explanation for some event, they are usually not asking for a complete list of possible causes but rather a few important reasons and causes relevant to the discussion at hand. Humans are adept at selecting one or two relevant key causes from a sometimes infinite number of causes as the explanation. If we generate all kinds of causes for explaining a single event, the causal chain can be too large and hard to handle for the explainee. The principle of simplicity dictates that the explainer should not overwhelm the explainee.

\textbf{\dots social and context-dependent.} Philosophy, psychology, and cognitive studies suggest that we are not explaining the same thing to everyone -- we change the way we explain based on whom we are talking to. The way we explain depends greatly on our model of the other person. People employ cognitive biases and social expectations. Explanations are a transfer of knowledge, presented as part of a conversation or interaction. If a person we are talking to does not know something, we are filling in the gap in their understanding. If they seem to understand the subject well, we can share less obvious causes for an event too. Explanations are thus presented \emph{relative} to the explainer's beliefs about the explainee's beliefs.

\textbf{\dots interactive.} Through the exchange of explanations and confirmation of understanding, we can continuously stay on the same page. The explainee can let the explainer know what subset causes they care about that are relevant for them. The explainer can then select a subset of that subset based on other criteria. The explainer and the explainee can interact further and argue about explanations. In XAI, there have been relatively few works on interactive explanations so far. Typically, we generate human-agnostic explanations that should work for everyone. Based on human interactions, we should be able to generate personalized explanations.

Because of these properties, there is no single correct answer to ``Why?''.

\section{Properties of Good Explanations}

\subsection{What are good explanations?}
\label{ssec:good}

From now on, we will be using more refined terminologies that are also used in XAI. The properties of a good explanation we deem most important are listed and explained below.

\textbf{Soundness/faithfulness/correctness.} The explanation should \emph{identify the true cause for an event}. This is the primary focus of current XAI evaluation: The attributions should identify the true causes (a model used) for predicting a certain label. It is also high on the list of desiderata from domain experts~\cite{lakkaraju2022rethinking}. (``What do you need from explainability methods?'') However, it is important to highlight that this is not the only criterion for a good explanation. \textbf{Example}: ``Why did you recognize a bird in this image?'' If the model points to a feature that does not contribute to its prediction of `bird', then its explanation is not sound/faithful. Another example is the following. A doctor wants to know the actual thought process of the system rather than just a likely reasoning from a human perspective. By understanding what is going on inside recognition mechanisms, we might be able to learn more than what humans are currently capable of extracting from an image.

\textbf{Simplicity/compactness.} The explanation should \emph{cite fewer causes}. A good balance is needed between soundness and simplicity (such that humans can handle the explanation).

\textbf{Generality/sensitivity/continuity.} The explanation should \emph{explain many events}. They should not only explain very specific events -- one usually seeks a general explanation. In XAI, generality means that the explanation should apply to many (similar) samples in the dataset.

\textbf{Relevance.} The explanation should be \emph{aligned with the final goal}. This criterion asks ``What do we need the explanation for?'' If we need it for fixing a system, the explanation should help us fix it. If we need explanations for understanding, a good explanation should then let us understand the event in question.

\textbf{Socialness/interactivity.} Explaining is a social process that involves the explainer and the explainee. In XAI, the explainer is the XAI method, and the explainee is the human. As mentioned before, the explanation process is dependent of the explainee. A good explanation could consider the social context and adapt and/or interact with the explainee. It should not always cite the most likely cause but also retrieve causes that are interesting for the user. These do not necessarily coincide.

\begin{figure}
    \centering
    \includegraphics[width=0.6\linewidth]{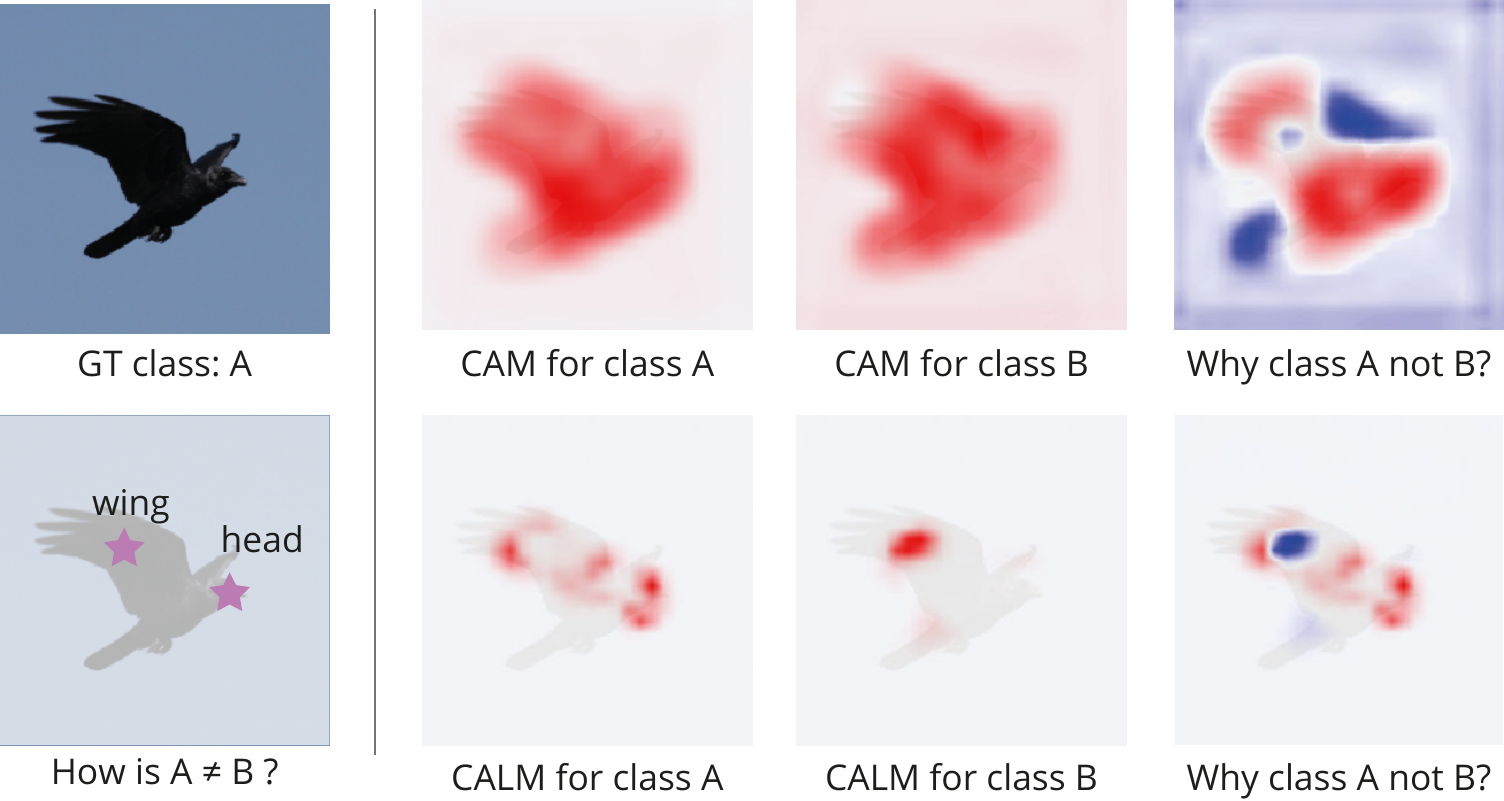}
    \caption{Example of contrastive explanations in XAI in the last column, taken from~\cite{kim2021keep}.}
    \label{fig:contrastive}
\end{figure}

\textbf{Contrastivity.} The foil needs to be clearly specified. Ideally, a method should be able to tell how the model's response changes when we change something from \(A\) to \(B\). An example of contrastivity is shown in the paper ``\href{https://openaccess.thecvf.com/content/ICCV2021/html/Kim_Keep_CALM_and_Improve_Visual_Feature_Attribution_ICCV_2021_paper.html}{Keep CALM and Improve Visual Feature Attribution}''~\cite{kim2021keep}, which we are going to discuss later in more detail. An illustration of contrastive explanations is given in Figure~\ref{fig:contrastive}.

\textbf{Human-comprehensibility/coherence/alignment with prior knowledge of the human.} The given explanation should fit the understanding and expectations of the human. It should also be presented in a format that is natural to humans.

The first property, soundness, is much more often used in XAI for evaluation. Nevertheless, the others are just as important.

\subsection{Intrinsically Interpretable Models}

Intrinsically interpretable algorithms are generally deemed interpretable and need no post-hoc explainability~\cite{https://doi.org/10.48550/arxiv.1702.08608}. A DNN is not like this. We can generate an explanation for a prediction post-hoc, making it explainable, but the DNN itself is not interpretable.

\subsubsection{Sparse Linear Models with Human-Understandable Features}

Sparse linear models make a prediction \(x\) using the following formula:
\[x = \sum_{i \in S} c_i \phi_i \qquad |S| \ll m.\]
The prediction is a simple linear combination of features \(\phi_i\) with coefficients \(c_i\). We also understand the individual \(\phi_i\)s very well, as they are human-understandable. Sparse linear models further contain a small number of coefficients: There are few enough coefficients for humans to understand the way the model works. Every feature \(\phi_i\) is contributing to the prediction by giving a factor \(c_i \phi_i\) to the sum. We know the exact contributions. We are citing every \(\phi_i\) as the cause of the outcome \(x\). This explanation is very general -- it works for any \(\phi_i\) and any value thereof.

\begin{information}{Feature Attribution vs. Feature Importance}
By construction, when \(\phi_i = 0\), the feature contributes with a factor of zero to the final prediction. By treating \(c_i\phi_i\) as our attribution score for feature \(i\), we cannot give a non-zero attribution score to features whose value is zero. Depending on what \(\phi_i\) is encoding, this might have surprising consequences. For example, when the individual features are pixel values (\(\phi_i \in [0, 1]\)), we cannot attribute to black pixels. To resolve this, we note that attribution scores are not synonymous with \emph{feature importance} -- the score we obtain is a reason for the prediction \(x\) \wrt feature \(\phi_i\), which is exactly \(\phi_i c_i\). This does not mean that this feature is unimportant, it just means that this term contributed the value 0 to the final prediction. The importance of the feature is better measured by the coefficient \(c_i\), but one can only directly interpret it as the (signed) importance of the feature in the prediction of the model if (1) the features are uncorrelated and (2) they are on the same scale (e.g., by standardization). The coefficient \(c_i\) can also be used as an attribution score: while \(c_i\phi_i\) measures the \emph{net} contribution of the feature for the prediction of \(x\), the coefficient \(c_i\) measures the \emph{relative} contribution of the feature if we slightly change that feature.
\end{information}

\subsubsection{Do these sparse linear models explain well according to our criteria?}

Let us consider the aforementioned criteria and evaluate sparse linear models according to them.

\textbf{Soundness/faithfulness/correctness.} By definition, every feature is a sound cause for the outcome, with contribution given by the terms \(c_i \phi_i\).

\textbf{Simplicity/compactness.} One can control this aspect with the number of features. Sparsity enforces the model to use few causes. We could not understand the model's decision-making process if we had millions of features.

\textbf{Generality/sensitivity/continuity.} By definition, whenever the cited causes happen, similar outcomes follow.

\textbf{Relevance.} This criterion always depends on the final goal. Is it to debug? Or to understand? For debugging purposes, we measure the quality of the explanation in terms of whether it is actually helping a human find features that are not working and whether the human can fix the system based on the explanation.

\textbf{Socialness/interactivity, Contrastivity.} One can simulate contrastive reasoning from the ground up. Instead of having \(\phi_1, \dots, \phi_{|S|}\), we can leave one out and see what happens afterward. This is the counterfactual answer for the effect of leaving a feature out. However, the models are not social and interactive by default. An additional module is required for that. The explanation is also not personalized; it does not consider the level of knowledge of the explainee. We can attach a chatbot or interactive system to make interactivity possible.

\subsubsection{Decision Trees with Human-Understandable Criteria}

In a decision tree that follows human-understandable criteria, all deciding features (criteria) follow a human-understandable concept. Unless the depth of the tree is too large, or we have too many branches, the entire tree is human-interpretable. \textbf{Example}: The task is to predict the animal breed. Features are not too many well-known properties of animals. Humans can then directly understand how the decision tree makes a particular prediction.

\section{Taxonomies of Model Explainability}

There are different ways to divide the set of explainability methods. The correlated ``axes'' of variation are as follows.

\subsubsection{Intrinsic vs. Post-hoc}

\textbf{Intrinsic interpretability} means the model is interpretable by design (i.e., sound, simple, and general at once).
The way the input is transformed into the output is interpretable. \textbf{Examples}: sparse linear models and decision trees.

\textbf{Post-hoc explainability} means the model lacks interpretability, and one is trying to explain its behavior post-hoc. \textbf{Example}: turning a DNN into a more interpretable system. The intrinsically interpretable models discussed before are very useful throughout our studies of explainability: We often turn a part of our network into a linear model and analyze that (e.g., Grad-CAM~\cite{selvaraju2017grad} in \ref{sssec:gradcam}) or approximate the whole model using a sparse linear model and explain it in a local input region (e.g., LIME~\cite{ribeiro2016should} in \ref{sssec:lime}).

\subsubsection{Global vs. Local}

\begin{figure}
    \centering
    \includegraphics[width=0.35\linewidth]{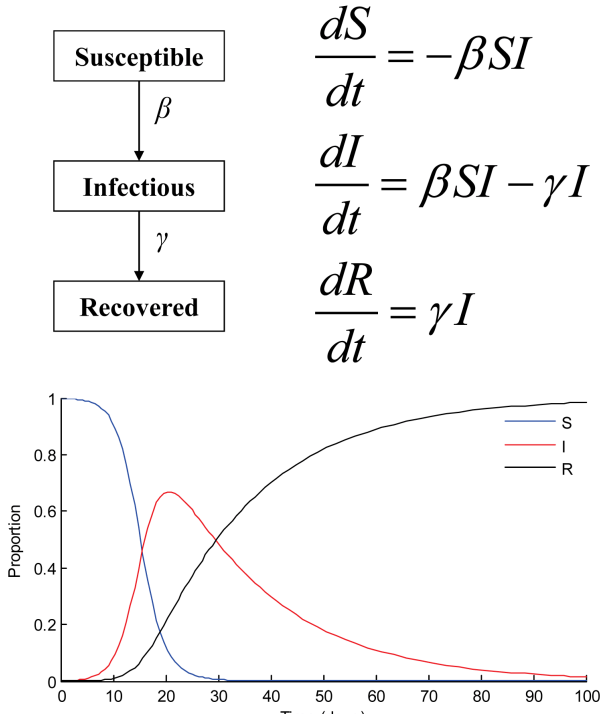}
    \caption{The SIR epidemic model is an example of a global explainability tool. The differential equations above determine the behavior of the system. Having chosen the parameters \(\beta, \gamma\), we simulate the future. Figure taken from~\cite{luz2010modeling}.}
    \label{fig:sir}
\end{figure}

\textbf{Global explainability} means the given explanation is not on a per-case basis. We do not want to understand one particular event but rather the entire system, allowing us to understand per-case decisions as well. An example of global explainability is the SIR epidemic model~\cite{kermack1927contribution}, shown in Figure~\ref{fig:sir}. Here, \(\beta, \gamma\) are rates of transitions. The system is based on the differential equations in Figure~\ref{fig:sir} that explain the entire system. We simulate the future based on our choice of \(\beta, \gamma\). This gives an overall understanding of the mechanism but is often impossible to give for complex, deep black-box models. It is particularly useful for scientific understanding and simulation of counterfactuals. (``What would happen if we changed some parameters?'')

\textbf{Local explainability} means we want to understand the decision mechanism behind a particular case/for a particular input. \textbf{Example}: ``Why did my loan get rejected?'' -- explanations for this do not lead to a global understanding of the system. Local explainability is the main focus/interest of the book. This type of analysis is feasible in somewhat sound ways even for complex models.

\subsubsection{Attributing to Training Samples vs. Test Samples}

A model is a function approximator. It is also an output of a training algorithm. The input to the training algorithm is the training data and other ingredients of the setting. We write the model prediction as a function of two variables:
\[Y = \text{Model}(X; \theta) = \text{Model}(X; \theta(X^{\text{tr}}))\]
where \(Y\) is our prediction, \(X\) is the test input, \(\theta\) are the model parameters, and \(X^\text{tr}\) is the training dataset. The prediction of our model is implicitly also a function of the training data.

We can trace back (i.e., attribute) the output \(Y\) for \(X\) to either (1) particular features \(X_i\) of the test sample \(X = [X_1, \dots, X_D]\), or (2) particular training samples \(X^{\text{tr}}_i\) in the training set \(X^\text{tr} = \{X^\text{tr}_1, X^\text{tr}_2, \dots, X^\text{tr}_N\}\).

One may also attribute the prediction to a particular parameter \(\theta_j\) or a layer, but individual parameters are often not very interpretable to humans. Usually, we ``project parameters'' onto the input space by gradient-based optimization.

\begin{information}{Correlations in the axes of variation}
Models that are intrinsically interpretable are interpretable on a global scope -- they give an understanding of the whole model. But they can also be used to explain particular decisions based on an input. While explaining local decisions is also possible, the focus is rather on the global scale.

\medskip

Methods with intrinsic interpretability also do not \emph{have to} directly attribute their predictions to anything. However, this is still often possible, e.g., in the case of sparse linear models.
\end{information}

\subsection{Soundness-Explainability Trade-off}

Explanations try to linearize a model in some way. What humans can naturally understand is a summation of a few features (i.e., a sparse linear model). There is an inherent \emph{soundness-explainability trade-off}. One extreme is the original DNN model by itself: It is by definition sound but not interpretable. Another extreme is creating a sparse linear model as the global linearization of a DNN around a particular point of interest. It cannot be sound as a global explanation but is very interpretable.

Between the two extremes, explanation methods try to linearize different bits of the model for either the entire input space (generic input) or for a small part of it. It is relatively easy to linearize the full model for a small part of the input space. (For example, LIME, discussed in Section~\ref{sssec:lime}.) It is also relatively easy to linearize a few layers of the model for generic input. (For example, Grad-CAM, discussed in Section~\ref{sssec:gradcam}.) However, it is impossible to faithfully linearize the full model for generic input. Then, we are back to global linearization.

\subsection{Current Status of XAI Techniques}

XAI research is often harshly criticized for being useless.

\begin{center}
``Despite the recent growth spurt in the field of XAI, studies examining how people actually interact with AI explanations have found popular XAI techniques to be ineffective, potentially risky, and underused in real-world contexts.''~\cite{ehsan2021expanding}
\end{center}

People working in human-computer interaction are very critical of XAI techniques in ML conferences, as they often do not take humans into account appropriately yet. It is essential to condition our mindset to help those who wish to actually use XAI techniques rather than working on techniques that look fancy or theoretically beautiful (e.g., completeness axioms, \ref{sssec:ig}).

\begin{center}
``The field has been critiqued for its techno-centric view, where ``inmates [are running] the asylum'', based on the impression that XAI researchers often develop explanations based on their own intuition rather than the situated needs of their intended audience (final goal is not taken into account). Solutionism (always seeking technical solutions) and Formalism (seeking abstract, mathematical solutions) are likely to further widen these gaps.''~\cite{ehsan2021expanding}
\end{center}

We want to move away from developing such XAI techniques and focus on the demands of those needing XAI systems.

\textbf{Note}: Formalism \emph{is} helpful (both in method descriptions and evaluation), but the most important aspect should be whether these methods actually help people. Formalism is not the end goal.

\section{Methods for Attribution to Test Features}

\begin{figure}
    \centering
    \includegraphics[width=\linewidth]{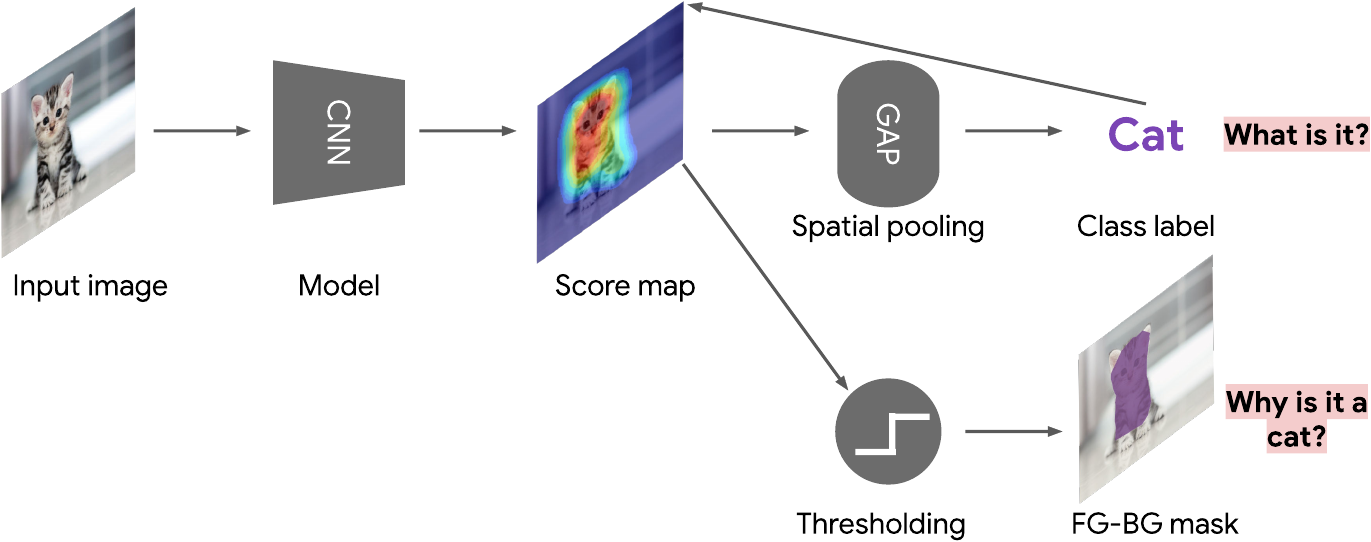}
    \caption{Simplified high-level overview of the CAM method. The model makes a prediction (`cat'), which is then used to select the appropriate channel of the Score Map that describes the attribution scores for class `cat'. Finally, an optional thresholding can be employed to make a binary attribution mask.}
    \label{fig:camsimple}
\end{figure}

So far, we have laid down what we desire from explainable ML. In this section, we discuss actual methods for extracting explanations from DNNs. In particular, we will look at methods that attribute their predictions to test features. Instead of the ``What is in the image?'' question, explanation methods seek to answer the ``Why does the model think it is the predicted object?'' question.\footnote{Curiously, some explanation methods are also good at Weakly Supervised Object Localization (WSOL), that aims to answer the ``Where is the object in the image?'' question.} For example, CAM~\cite{https://doi.org/10.48550/arxiv.1512.04150} produces a score map \wrt the predicted label, as illustrated in Figure~\ref{fig:camsimple}. We can threshold it to get a foreground-background mask as an explanation (which is not necessarily a mask \wrt the GT object location, as the network being explained might have, e.g., background or texture biases). We can also leave out thresholding and keep continuous values in the map. 

\subsection{What features to consider in attribution methods to test features?}

\begin{figure}
    \centering
    \includegraphics[width=0.4\linewidth]{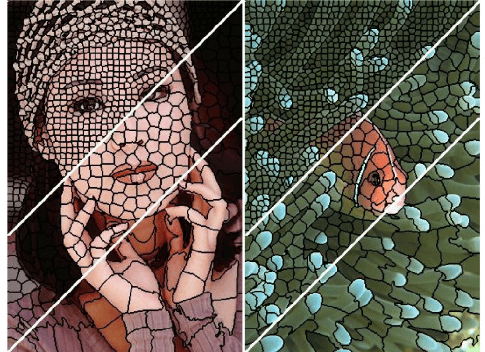}
    \caption{Illustration of superpixels of various granularities, which is a popular choice of features for attribution maps. Figure taken from~\cite{superpixels}.}
    \label{fig:superpixels}
\end{figure}

\begin{definition}{Superpixel}
Superpixels are groupings of pixels respecting color and edge similarity (that very confidently belong to the same object instance). It gives us a finer grouping than semantic segmentation (in the sense that the pixels are not grouped into only a couple of categories, but rather into many patches of pixels that closely belong together) but a coarser one than the raw pixels. An illustration is given in Figure~\ref{fig:superpixels}. There have been many improvements in superpixel technology until a few years ago. Nowadays, not many people are looking into superpixel methods. These are often used as features in explainability for images. They reduce the number of features we have to deal with without sacrificing soundness.
\end{definition}

\begin{figure}
    \centering
    \includegraphics[width=\linewidth]{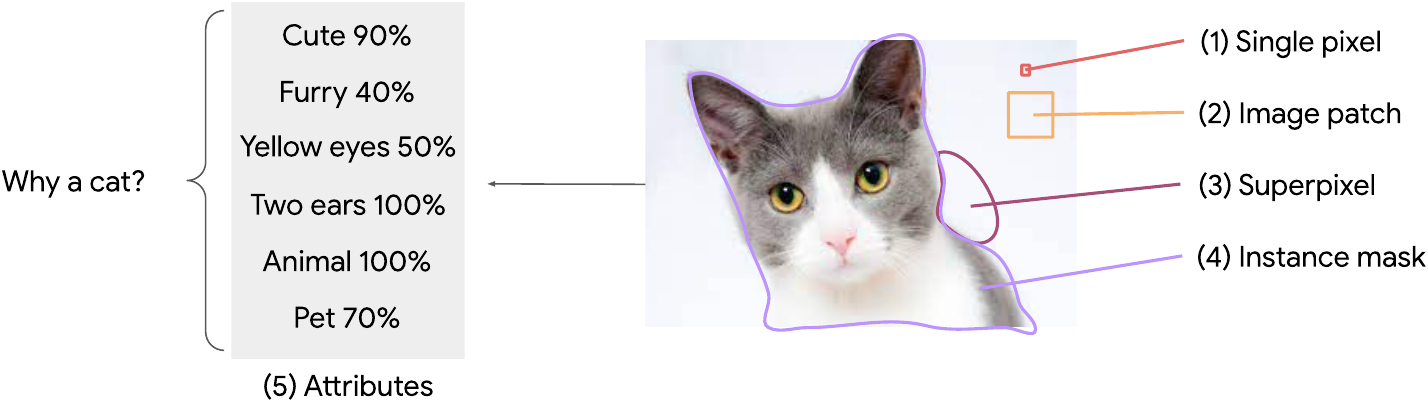}
    \caption{Illustration of several feature representations for the same image. There is a wide range of features we can attribute to.}
    \label{fig:catfeatures}
\end{figure}

We generalize the notion of a feature to any aggregation or description of the input to the model. Possible features are listed below for visual models taking an image as input. These are also illustrated in Figure~\ref{fig:catfeatures}.
\begin{enumerate}
    \item \textbf{Single pixels.}
    \item \textbf{Image patches.} We can aggregate pixels into image patches, considering each patch as a feature.
    \item \textbf{Superpixels.}
    \item \textbf{Instance mask(s).}
    \item \textbf{High-level attributes.} For example, attributes for a cat image input can be Cute, Furry, Yellow eyes, Two ears, Animal, and Pet. The values for each of these attributes can be percentages representing how fitting a certain attribute is for the input. For example, Two ears \(\rightarrow\) 100\% means the feature is maximized, i.e., the attribute perfectly fits the input. \textbf{Note}: These are \emph{not} the attribution scores corresponding to the individual attributes. Attribution scores are values describing how each of the features influences the network prediction, whereas the attributes \emph{describe} the input. The attributes can be subjective, can point to specific regions of the image, and can also describe, e.g., the general feeling of the image.
\end{enumerate}

\begin{figure}
    \centering
    \includegraphics[width=0.5\linewidth]{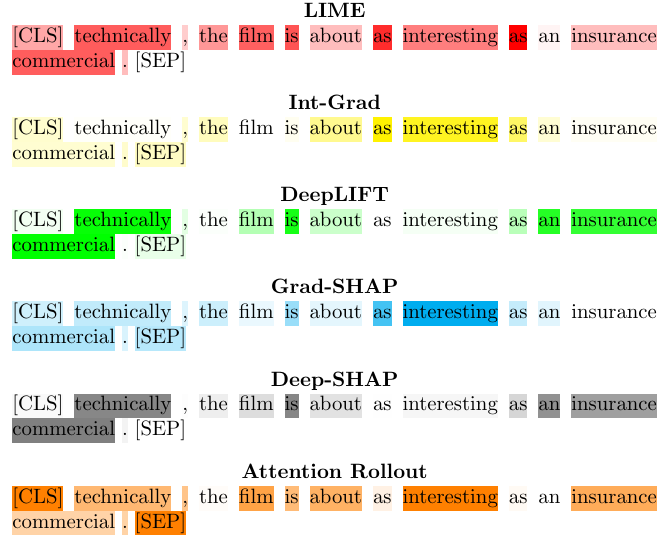}
    \caption{Sentiment analysis example. Explanation methods give significantly different attributions. ``The average Kendall-\(\tau\) correlation across all methods for this example is 0.01.''~\cite{neely2022song} Figure taken from~\cite{neely2022song}.}
    \label{fig:nlp}
\end{figure}

For natural language models taking a token sequence as inputs, we often use individual tokens/words as features. We can think about the contribution of each token (or word) towards the final prediction (e.g., sentiment analysis), as considered in the paper ``\href{https://arxiv.org/abs/2205.04559}{A Song of (Dis)agreement: Evaluating the Evaluation of Explainable Artificial Intelligence in Natural Language Processing}''~\cite{neely2022song}. Examples of attributing to individual tokens can be seen in Figure~\ref{fig:nlp}.

\textbf{Note}: Explanation methods can give significantly different results for the same input, as shown in Figure~\ref{fig:nlp}. This has also been reported in the paper titled ``\href{https://arxiv.org/abs/2202.01602}{The Disagreement Problem in Explainable Machine Learning: A Practitioner's Perspective}''~\cite{krishna2022disagreement}: local methods approximate the model at a particular test point $x$ in local neighborhoods, but there is no guarantee that they use the same local neighborhood. Indeed, since different methods use different loss functions (e.g., LIME with squared error vs. gradient maps with gradient matching), it is likely that different methods produce different explanations.

\begin{information}{Choice of Features}
If we gather all the features of an image, do we have to obtain the original image by definition? The answer is \emph{yes}; we generally wish to \emph{partition} the image with features.
\begin{itemize}
    \item For partitioning, one may choose \href{https://arxiv.org/abs/1801.00868}{panoptic segmentation}~\cite{kirillov2019panoptic}, a combination of instance segmentation and semantic segmentation. This considers both object and stuff masks (where stuff refers to, e.g., `road', `sky', or `sidewalk'). Another option is regular semantic segmentation, which can also handle various stuff categories. The \href{https://github.com/nightrome/cocostuff}{COCO-Stuff}~\cite{caesar2018cvpr} dataset gives many examples of how semantic segmentation can partition images in a detailed way.
    \item Considering only image parts corresponding to different instance masks as features is problematic, as stuff information is thrown away (we get rid of stuff categories), and we do not have a partition of the original image anymore.
\end{itemize}
\end{information}

A feature is thus a general concept. The task for feature attribution methods is to determine which feature contributes how much to the model's prediction.

\begin{figure}
    \centering
    \includegraphics[width=0.8\linewidth]{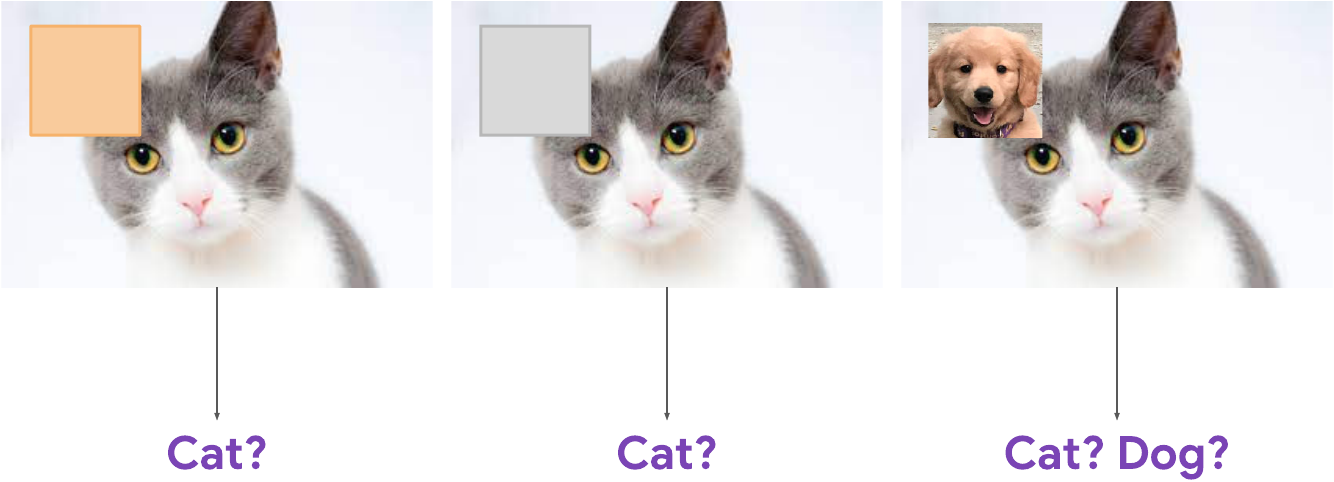}
    \caption{Three possibilities of counterfactual explanations. The left image encodes the missingness of a patch by orange pixel values. The center image encodes missingness by gray pixel values. These give answers to the question ``How would the prediction change if we removed a patch of the image?'' The right image asks a slightly different question: ``Which features contribute to predicting a cat rather than a dog?'', as the dog image does not aim to encode missingness, i.e., we cannot talk about removing the patch.}
    \label{fig:cat}
\end{figure}

In the last section, we have seen that counterfactual (i.e., contrastive) reasoning matters a lot in explaining to humans. The most basic way to explain a model's decision in a counterfactual way is by asking a question of the form ``Is the input image still predicted as a cat if this feature is missing?'' We remove a particular set of pixels from the image and see how the model's prediction changes. We have many possibilities to encode what we mean by a ``missing'' pixel. For example, we can fill them with black, gray, or even pink pixels (which are rarely seen in natural images but do not intuitively encode a baseline image). We can even choose to inpaint them based on the context. One could also ask counterfactual questions of the form ``Is the input image still predicted as a cat if this feature is replaced with something else?'' In this case, we can, e.g., insert an image of a dog in the ``missing'' patch, illustrated in Figure~\ref{fig:cat}. After carrying this out for all pixels, we get an answer to ``Which features contribute most to predicting a cat rather than a dog?'' Even in the simple setting of removing a square patch from an image, many things must be considered.

\subsection{Intrinsically interpretable models support counterfactual evaluation by design.}

In a DNN, when we change something in the input, it is highly unclear how the forward propagation is influenced to obtain the final answer. In decision trees, we can just change one attribute in any way and check how the result changes (by selecting the other branch at a corresponding attribute). We can do a full simulation quickly where we understand each part of the decision-making process. We can still do the simulation for a DNN, but we only observe the outputs (before and after the change) in an interpretable way. We have no good intuition about what changes \emph{internally}.

A sparse linear model is just a summation. Every feature contributes linearly to the final output. It is easy to interpret the relationship between the features and outputs. We know the full effect on the output of changing (or removing) one or many features. Our implicit aim is to linearize our complex models in some way for interpretability.\footnote{We might not want to linearize the entire model. Partial linearization is often used, e.g., in Grad-CAM (\ref{sssec:gradcam}) and TCAV (\ref{sssec:tcav}).} This is a common mindset of attribution methods. Because of the linear relationship between inputs and outputs, we do not have to compute differences between outputs to study counterfactual evaluation. We already know how the output changes by changing some inputs. This is highly untrue for DNNs, requiring a forward pass each time. We will see that under some quite strong assumptions, we can use the input gradient and derivative quantities for counterfactual evaluations.

\subsection{Infinitesimal Counterfactual Evaluation in Neural Networks: Saliency Maps}

We can perform the removal analysis for all input features for neural networks, e.g., using a sliding window of patches as features. This, however, takes very long for DNNs. For each image, one needs to compute \(N\) forward operations through a DNN, where
\[N = \text{number of sliding windows per image \(\times\) number of ways to alter the window content}.\]
Doing this in real-time during inference on a single sample is infeasible without sufficient computational resources for parallelization. Doing it offline for an entire dataset also takes very long if the dataset is large. One can use batching, but only a small number of samples fit on the GPU usually.

However, we can consider a special case where our \emph{features are pixels} and the \emph{perturbation is small} (infinitesimal). In this case, we can compute counterfactual analysis quickly, at the cost of the huge restriction of the perturbation size being small. 

\textbf{Example}: Consider pixel \((56, 25)\) with original pixel value: \((232, 216, 231)\). Suppose that all pixels are left unchanged except this one where the new pixel value is set to \((233, 216, 231)\).\footnote{This is the smallest possible perturbation with bit depth \(8\) -- a coarse approximation of the gradient.} Further, suppose that the original cat score was 96.5\%, but after the change, the cat score for the perturbed image decreases to 96.4\%. This seems familiar: That is exactly how we approximate the gradient numerically:
\[\frac{\partial f(x)}{\partial x_i} \approx \frac{f(x + \delta e_i) - f(x)}{\delta}, \qquad \delta \text{ small.}\]
In the ordinary sense, the gradient of the score (or probability) of the predicted class \wrt an RGB pixel is a 3D vector (as the RGB pixel itself is also a 3D vector). However, we will consider \(x \in \nR^d\) as a flattened version of an image and will also collapse the color channels. We treat the elements of the resulting vector as pixels. Therefore, the $i$th pixel direction does \emph{not} correspond to the general definition of a pixel in the following sections.

\begin{information}{Discrete Representations of Color}
The 8-bit representation is just a convention for RGB images. There exist 16- and 32-bit representations too. The RGB scale is continuous. 
\end{information}

A very inefficient way to compute the attribution of each pixel is to compute the forward pass
(number of pixels + 1) times (perturbed images plus original image, as the latter is shared in all gradient approximations) to measure pixel-wise infinitesimal contribution. A large approximate gradient signals a significant contribution of the corresponding pixel for an infinitesimal perturbation (because of a significant change in the score of the predicted class).

\textbf{Note}: Here, we consider the \emph{relative} contribution of a pixel (as we equate high contribution to a high \emph{relative change} in the network output for an infinitesimal perturbation), similarly to the sparse linear model case where the relative contribution of feature \(\phi_i\) was given by the coefficient \(c_i\). Of course, this was just a special case of the gradient for the sparse linear model case: if we differentiate \(x = \sum_{j \in S} c_j\phi_j\) \wrt \(\phi_i\), we get back \(c_i\) again. 

\begin{information}{On the Properties of Gradients}
The derivative
\[f'(x) = \lim_{h \rightarrow 0} \frac{f(x + h) - f(x)}{h}\]
is a normalized quantity. It gives the \emph{relative} change in the function output, given an infinitesimal change in the input.
\end{information}

The smart way to compute changes in the output \wrt infinitesimal perturbations: Compute one forward and one backward pass \wrt the score of the predicted class to measure attributions for this infinitesimal perturbation. This answers the question ``What will be the \emph{relative change} in the predicted score if we change a particular pixel by an infinitesimal amount?''

This leads us to the definition of \emph{Saliency/Sensitivity maps}.

\begin{definition}{Saliency/Sensitivity Map}
The saliency or sensitivity map visualizes a counterfactual attribution for an input corresponding to infinitesimal independent per-pixel perturbations. It gives us a local explanation of the model's prediction. There are two usual ways to compute it.

\medskip

Denoting the saliency map for input \(x \in \nR^{H \times W \times 3}\) and class \(c \in \{1, \dotsc, C\}\) by \(M_c(x) \in [0, 1]^{H \times W}\), the \href{https://arxiv.org/abs/1706.03825}{SmoothGrad}~\cite{https://doi.org/10.48550/arxiv.1706.03825} paper \href{https://github.com/PAIR-code/saliency/blob/master/saliency/core/visualization.py#L17}{computes} it as
\[(\tilde{M}_c(x))_{i, j} = \sum_{k} \left|\frac{\partial S_c(x)}{\partial x}\right|_{i, j, k}\]
(we take the \(L_1\) norm of each pixel), and
\[(M_c(x))_{i, j} = \min\left(\frac{(\tilde{M}_c(x))_{i, j} - \min_{k \in \{1, \dotsc, H\}, l \in \{1, \dotsc, W\}}(\tilde{M}_c(x))_{k, l}}{P_{99}(\tilde{M}_c(x)) - \min_{k \in \{1, \dotsc, H\}, l \in \{1, \dotsc, W\}}(\tilde{M}_c(x))_{k, l}}, 1\right)\]
where \(S_c(x)\) is the score for class \(c\) given input \(x\) and \(P_{99}\) is the 99th percentile. This post-processing normalizes the saliency map to the \([0, 1]\) interval and clips outlier pixels by considering the 99th percentile. Not clipping the outlier values could result in a close-to-one-hot saliency map.

\medskip

In \href{https://arxiv.org/abs/1312.6034}{Simonyan \etal (2013)}~\cite{https://doi.org/10.48550/arxiv.1312.6034} (the original saliency paper), the authors compute it as
\[(\tilde{M}_c(x))_{i, j} = \max_{k} \left|\frac{\partial S_c(x)}{\partial x}\right|_{i, j, k}\]
and the normalization method is not disclosed.
\end{definition}

\begin{definition}{First-Order Taylor Approximation}
Consider a function \(f: \nR^d \rightarrow \nR\). The first-order Taylor approximation of the function \(f\) around \(x \in \nR^d\) is
\[f(x + h) \approx f(x) + \left\langle h, \nabla f(x) \right\rangle.\]
\end{definition}

\textbf{Backpropagation linearizes the whole model around the test sample.} To see this, observe that the gradient is used to construct the first-order Taylor approximation of the model around a particular test sample, which is the tangent plane of the model around the test sample:
\[f(x + \delta e_i) - f(x) \approx \left\langle \delta e_i, \frac{\partial f(x)}{\partial x} \right\rangle = \delta \frac{\partial f(x)}{\partial x_i}\]
where \(f\) gives the score for a fixed class \(c\) that is omitted from the notation.
This tangent plane guarantees that the function output with this linearized solution will be as close as possible (in the set of linear functions) to the original function's output around the test input of interest in an infinitesimal region. We give a local (counterfactual) explanation with this linear surrogate model, as we only consider an explanation for a single test input. With this surrogate model, one can very cheaply compute input-based counterfactuals. However, these will only be faithful to the original model in a tiny region around the test input of interest.

\textbf{Note}: Our surrogate model is linear but is not guaranteed to be sparse! It can still be hard to interpret when the input dimensionality is huge. This is primarily the reason why, instead of looking at actual gradient values, we visualize the dense gradient tensors in the form of saliency maps.

\subsubsection{Summary of Infinitesimal Counterfactual Attribution}

With local gradients, we obtain
\[f(x + \delta e_i) - f(x) \approx \left\langle \delta e_i, \frac{\partial f(x)}{x} \right\rangle = \delta \frac{\partial f(x)}{x_i}\]
which measures contribution of each pixel \(i\) with an infinitesimal (\(\delta\)) counterfactual.

\subsubsection{Problem with Saliency Maps}

\begin{figure}
    \centering
    \includegraphics[width=0.5\linewidth]{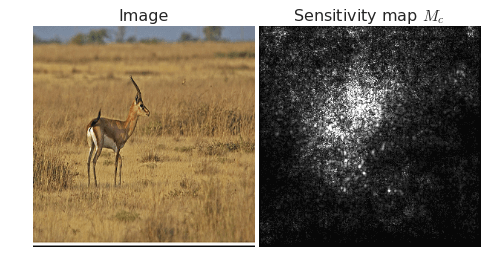}
    \caption{Example saliency map of image \(x\) \wrt the class `gazelle', taken from~\cite{https://doi.org/10.48550/arxiv.1706.03825}. Saliency maps can be challenging to interpret.}
    \label{fig:sensitivity}
\end{figure}

We visualize input gradients using saliency maps. These visualizations are not particularly helpful, as they are very noisy and hard to interpret further than a very coarse region of interest. An example is shown in Figure~\ref{fig:sensitivity}. \textbf{Note}: saliency maps are always \wrt a class \(c\). We almost always compute it \wrt the DNN's predicted class. We might ask ourselves, ``What do we actually get out of this?'' We do not even see the object in these input gradient maps. Gradient maps only represent how much \emph{relative} difference a tiny change in each pixel of \(x\) would make to the classification score for class \(c\). It is debatable whether one should measure attribution values based on such infinitesimal changes.

Negative contributions are counted as contributions here. This varies from method to method, and no ``good'' answer exists. The \(i\)th element of the gradient measures the relative response of the classification score of class \(c\) to a perturbation of the image in the \(i\)th pixel direction. If it is positive, making the pixel more intensive results in a locally positive classification score change. If it is negative, it means we reach a higher classification score if we dim the pixel. Sometimes we only want to attribute to pixels that induce a positive change in the score when made more intensive. Sometimes we also want to take negative influences into account.

\begin{information}{Gradients and Soundness}
It is a \emph{fact} that the gradient gives us the \emph{true} relative changes in the prediction considering per-pixel, independent infinitesimal counterfactuals. It is very important to not confuse this fact with the statement that gradients give perfectly sound attributions in the sense that they flawlessly enumerate the true causes for the network making a certain prediction.

\medskip

Soundness, by definition, measures whether the attribution method recites the true causes for the model to predict a certain class. As the true causes are encoded in the model weights and the forward propagation formula, which is not at all interpretable, it seems clear that \emph{no attribution method that presents significantly simpler reasoning can be perfectly sound}. Saliency maps -- that \emph{seek} to give sound counterfactual explanations \wrt an infinitesimal perturbation -- make use of such simple reasoning: linearizing the network around the input of interest and taking the rates of change as attribution scores. The linearization, the independent consideration of inputs (with which we discard the possible influence of input feature correlations on the network prediction), and the ``arbitrary'' normalization and aggregation techniques of the 3D gradient tensor are all significant simplifications that make saliency maps \emph{impossible to be completely sound}.\footnote{Strictly speaking, the linearization is not a simplification when considering infinitesimal perturbations. However, such perturbations are fictitious, and if one wants to obtain the \emph{net} changes in the network output, they have to consider small \(\delta\) values that are not exact anymore.} Even if they \emph{were} sound explanations \wrt infinitesimal perturbations, the question itself already seems oddly artificial: ``Why did the network make a certain prediction for input \(x\) compared to an infinitesimally perturbed version of it?''. There is no reason why society would demand explanations for such answers. 

\medskip

Moreover, feature attribution methods restrict the explanations to the features in the test input, but the true causes for a network to predict a certain output can also lie in the training set samples and the resulting model weights. Feature attribution methods only consider the test input features as possible causes and make crude assumptions to compute attribution scores. For the soundness of the \emph{explanation}, the attribution method has to give the exact causes of why the network made a certain prediction for an input \(x\). We argue that these exact causes cannot be encoded in general into a map measuring infinitesimal perturbations. Of course, most feature attribution explanations do not claim to provide sound explanations. Instead, they aim to highlight that, given an input, some features were more important in a certain decision than others.

\medskip

To summarize, the saliency map does not give a perfectly sound attribution map for the predictions of the model on the input of interest because it uses abstractions and simplifications to make the explanation human-understandable. 

\medskip

\textbf{Note}: The soundness of an attribution method and the counterfactual or non-counterfactual nature of explanations it gives are completely independent. For non-counterfactual explanations, a sound attribution method simply aims to give the true influence of each feature of the original input on the network prediction \emph{without} comparing to other predictions.
\end{information}

\subsection{SmoothGrad -- Smoother Input Gradients}

The natural question is: Can we get smoother maps of attributions that are more interpretable? To obtain smoother attribution maps than saliency maps, \emph{SmoothGrad}, introduced in the paper ``\href{https://arxiv.org/abs/1706.03825}{SmoothGrad: removing noise by adding noise}''~\cite{https://doi.org/10.48550/arxiv.1706.03825}, computes gradients in the vicinity of the input \(x\). It follows three simple steps:
\begin{enumerate}
    \item Perturb the input \(x\) by additive Gaussian noise.
    \item Compute the gradients of the perturbed images.
    \item Average the gradients.
\end{enumerate}

\begin{figure}
    \centering
    \includegraphics[width=0.6\linewidth]{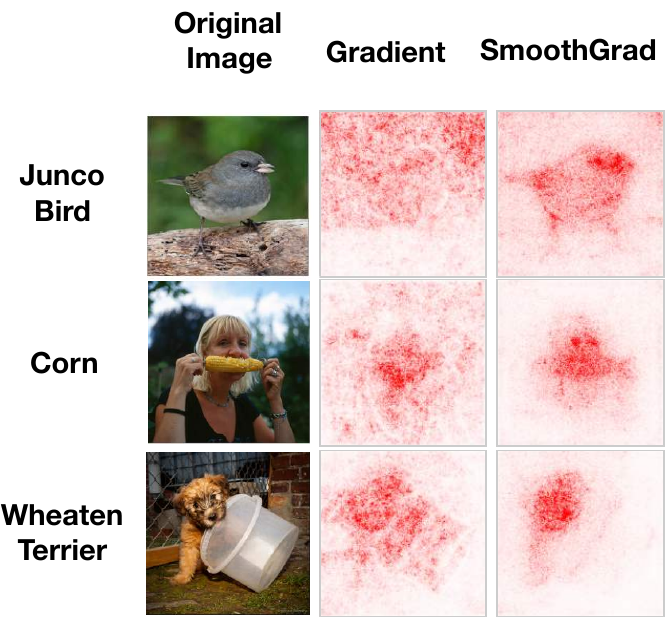}
    \caption{Qualitative comparison of SmoothGrad and saliency maps, taken from~\cite{https://doi.org/10.48550/arxiv.1810.03292}. SmoothGrad gives attribution maps that are more aligned with human expectations and more interpretable. One has to be careful with confirmation bias, though (Section~\ref{sssec:eval}).}
    \label{fig:smoothgrad}
\end{figure}

This gives us slightly less precise local attributions than the vanilla gradient (which is as local as possible). It also results in much clearer attribution maps because the added Gaussian noise and the gradient noise cancel out by averaging while the main signal remains in place. Examples are shown in Figure~\ref{fig:smoothgrad}. Combining gradients of different perturbations can reduce the noise and perhaps allow us to see more relevant attribution scores. Formally,
\[\hat{M}_c(x) = \frac{1}{n}\sum_{i = 1}^n M_c(x + \epsilon_i)\qquad \epsilon_i \sim \cN(0, \sigma^2I)\]
where care is also taken for each perturbed image \(x + \epsilon_i\) to stay in the \([0, 1]^{H \times W \times 3}\) space, as we are averaging across normalized saliency maps.

\subsubsection{Summary of SmoothGrad}

With ``less local'' gradients, we obtain
\begin{align*}
\nE_z\left[f(x + z + \delta e_i) - f(x + z)\right] &\overset{\text{Taylor}}{\approx} \nE_z\left[\delta \left\langle e_i, \nabla_x f(x + z) \right\rangle\right]\\
&= \delta \left\langle e_i, \nE_z\left[\nabla_x f(x + z)\right]\right\rangle\\
&= \delta \left(\nE_z\left[\nabla_x f(x + z)\right]\right)_i\\
&= \delta \nE_z\left[\frac{\partial}{\partial x_i}f(x + z)\right]
\end{align*}
which measures the contribution of each pixel \(i\) with an infinitesimal\footnote{In practice, we just choose a small \(\delta\) value for the tangent plane to stay faithful to the function. Another choice, as we have seen before, is to consider \(\nE_z\left[\frac{\partial}{\partial x_i}f(x + z)\right]\) as the attribution score for pixel \(i\).} counterfactual at multiple points \(x + z\) around \(x\). This expands the originally very local computation of the gradient to a slightly more global region around \(x\).

\subsection{Integrated Gradients}
\label{sssec:ig}

We will now go from local changes (simple gradients) to the inputs to more and more global changes in the hope that we obtain more sound attribution scores this way. \emph{Integrated gradients} is the middle ground between local and global perturbations. It averages over local \emph{and} global perturbations instead of perturbing only around a single point. We are linearly interpolating between two points in the input space.

In \emph{Integrated Gradients}, introduced in the paper ``\href{https://arxiv.org/abs/1703.01365}{Axiomatic Attribution for Deep Networks}''~\cite{sundararajan2017axiomatic}, we choose a base image that contains no information, \(x^0\), and consider our input image, \(x\). We linearly interpolate between \(x^0\) and \(x\) in the pixel space by slowly going from an image with no information (\(x^0\), the \emph{baseline image}) to the original image (\(x\)). We do the gradient computation at every intermediate point along the line, then average them (without weights, as the expectation is over a uniform distribution). This \emph{nearly} gives us the integrated gradients method:
\begin{align*}
\nE_{\alpha \sim \mathrm{Unif}[0, 1]}\left[f(x^0 + \alpha(x - x^0) + \delta e_i) - f(x^0 + \alpha(x - x^0))\right] &\overset{\text{Taylor}}{\approx} \nE_{\alpha}\left[\left\langle \delta e_i, \nabla_x f(x^0 + \alpha(x - x^0)) \right\rangle\right]\\
&= \delta \left\langle e_i, \nE_\alpha\left[\nabla_x f(x^0 + \alpha(x - x^0))\right] \right\rangle\\
&= \delta \left\langle e_i, \int_0^1 \nabla_x f(x^0 + \alpha(x - x^0))\ d\alpha \right\rangle.
\end{align*}
This estimates the pixel-wise contribution with an infinitesimal counterfactual (\(\delta\)), averaged over an entire line between the original input and the baseline image containing ``no information''.\footnote{Again, we can consider the attribution score with or without \(\delta\). If one includes it, one must keep it a very small number in practice for the tangent plane to stay faithful to the function. This measures the approximate absolute expected change in the output. If one does not include it (this is the usual choice), the score measures the \emph{relative} expected change in the output.}

However, in the integrated gradients method, the contribution of pixel \(i\) is computed as
\[(x_i - x^0_i) \left\langle e_i, \int_0^1 \nabla_x f(x^0 + \alpha(x - x^0))\ d\alpha \right\rangle,\]
and we derived
\[\left\langle e_i, \int_0^1 \nabla_x f(x^0 + \alpha(x - x^0))\ d\alpha \right\rangle.\]
We seemingly multiply a nicely motivated formula with pixel values. However, the integrated gradients formulation is actually the ``prettier'' formula, as it satisfies the \emph{completeness axiom}. If we sum over the contribution of all pixels \(i\), we obtain
\begin{align*}
&\sum_i (x_i - x^0_i) \left\langle e_i, \int_0^1 \nabla_x f(x^0 + \alpha(x - x^0))\ d\alpha \right\rangle\\
&= \left\langle (x_i - x^0_i)e_i, \int_0^1 \nabla_x f(x^0 + \alpha(x - x^0))\ d\alpha \right\rangle\\
&= \left\langle x - x^0, \int_0^1 \nabla_x f(x^0 + \alpha(x - x^0))\ d\alpha \right\rangle\\
&= \int_0^1 \left\langle \nabla_x f(x^0 + \alpha (x - x^0)), x - x^0 \right\rangle\ d\alpha\\
&= f(x) - f(x^0),
\end{align*}
using the fundamental theorem of line integrals (Definition~\ref{def:lineintegrals}) with \(r(\alpha) = x^0 + \alpha(x - x^0)\). In words: if we sum the pixel-wise contributions of all pixels (integrated gradients in the \(i\)th direction, multiplied by pixel differences), we get the difference between the original prediction and the baseline prediction.

The authors of~\cite{sundararajan2017axiomatic} argue that the completeness axiom is a necessary condition for a sound attribution. This axiom states that pixel-wise contributions for input \(x\) must sum up to the difference between the current model output \(f(x)\) and baseline output \(f(x^0)\). Here, the baseline image is an image without ``any information''. It represents the complete absence of signal. We measure what kind of additional information we add per pixel on top of this baseline image. The baseline image can be, e.g., an image consisting of noise or a completely black image.\footnote{A black image baseline is used in the paper. According to the authors, using a black image results in cleaner visualization of the ``edge'' features than using random noise.}

\textbf{Important downside of a black image baseline.} If we choose our baseline to be a black image, black pixels (e.g., pixels of a black camera) cannot be attributed at all, as \(x_i - x^0_i = 0\). This does not seem right. The black pixels of the camera are very likely also contributing to the model prediction of the class camera. This is different from the sparse linear model case: \(x = \sum_{i \in S}c_i\phi_i\). There, whenever an input feature \(\phi_i\) was 0 (e.g., a black pixel), it contributed to the prediction with a factor of 0, and this was the \emph{GT contribution} of this feature to the prediction. This was also a sound attribution. DNNs, however, are much more complex, and we no longer have this GT correspondence. Here, it is almost surely the case that the black pixels also contributed to the model prediction of a black camera. This problem is known as the ``missingness bias'' which we will further detail in later sections. Generally, the choice of the baseline value can be quite important. In many cases, random noise seems to be a better option. For the interested reader, the \href{https://distill.pub/2020/attribution-baselines/}{following resource} describes other options for the choice of the baseline.

\subsubsection{Results of Integrated Gradients}

\begin{figure}
  \centering
  \includegraphics[width=0.7\columnwidth]{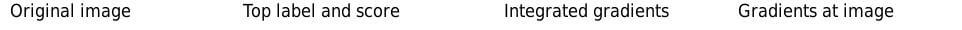}
  \includegraphics[width=0.7\columnwidth]{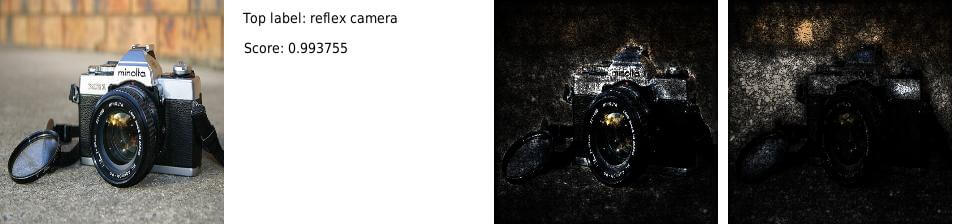}
  \includegraphics[width=0.7\columnwidth]{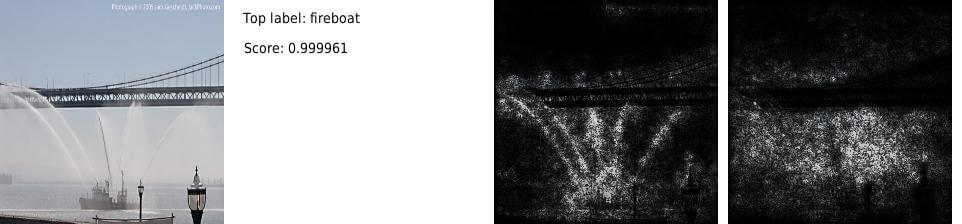}
  \includegraphics[width=0.7\columnwidth]{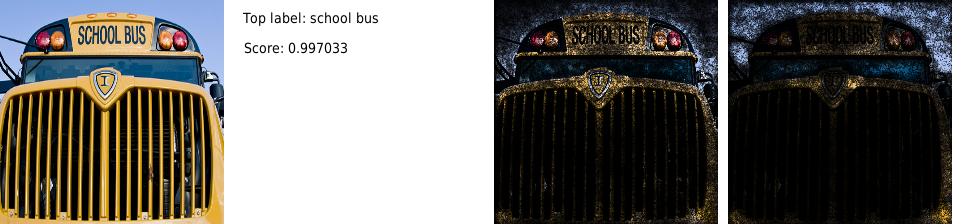}
  \caption{Qualitative comparison of Integrated Gradients and saliency maps, taken from~\cite{sundararajan2017axiomatic}. The explanations given by Integrated Gradients are more visually appealing.}
  \label{fig:integrated}
\end{figure}

The paper~\cite{sundararajan2017axiomatic} only provides an empirical evaluation of the method's soundness. Example attribution maps are shown in Figure~\ref{fig:integrated}. According to the results, the integrated gradients method nicely attributes (focuses) to the actual object regions, whereas gradients alone do not give us the ``focus'' we would expect.

We as humans deem the results sensible (which coincides with the `coherence with human expectations' property of a good explanation), as we would also focus on the regions that the method highlights. This is, however, a severe case of confirmation bias. We will discuss such biases in Section~\ref{sssec:eval}.

The attribution maps of the integrated gradients method are certainly more \emph{interpretable} than saliency maps. These show more continuous regions; thus, the explanations are more selective. However, this is just one of the evaluation criteria for a good explanation. The soundness of the explanations is only measured qualitatively, even though quantitative analysis would have been critical.

\subsection{Comparing Local and Global Perturbations -- Two Ways of Measuring Contribution}

\begin{figure}
    \centering
    \includegraphics[width=0.8\linewidth]{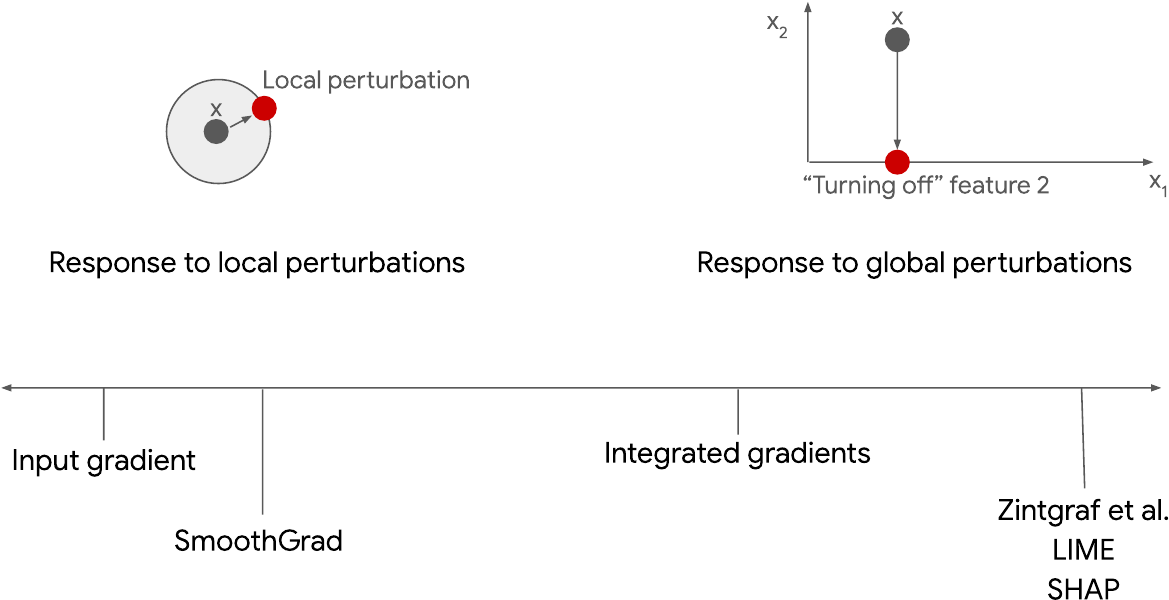}
    \caption{Spectrum of local explanation methods \wrt the nature of perturbations they employ.}
    \label{fig:spectrum}
\end{figure}

We consider two extremes in the domain of local explanation methods that aim to give counterfactual explanations: those that make \emph{local perturbations} to the input \(x\) and those that perturb the input \emph{globally}. We also consider an entire \emph{spectrum} between these two extremes. This spectrum is depicted in Figure~\ref{fig:spectrum}.

Local perturbations make very local changes to the input and measure the network's response.
\begin{itemize}
    \item \textbf{Pro}: It has well-understood properties. (The concept of a gradient.) It has no dependence on reference values.
    \item \textbf{Contra}: We only employ infinitesimal counterfactuals.
\end{itemize}
Global perturbations measure counterfactual responses by turning off features entirely in various ways.\footnote{Turning a feature off means that the features receive the baseline value, which is not necessarily zero.}
\begin{itemize}
    \item \textbf{Pro}: This can lead to meaningful counterfactual analysis. This is also a much more natural question to seek explanations for.
    \item \textbf{Contra}: Setting the reference values is hard. Such methods are computationally heavy and need further assumptions/approximations to make them efficient.
\end{itemize}
The method of integrated gradients gives a smooth interpolation between local changes and turning off features completely.
\textbf{Note}: These are all still \emph{local explainability methods}. Whether the perturbation is local or global is an independent axis of variation.

\subsection{Local \(=\) Global for (Sparse) Linear Models}
Consider a linear model
\[x = \sum_{i \in S} c_i \phi_i \qquad |S| \ll m.\]
When responding to local perturbations, the gradient of the output \wrt the feature \(i\) is \(c_i\). When responding to global perturbations, the effect of turning off feature \(i\) is \(c_i \phi_i\). (Here, we actually set the feature to zero.)

Therefore, the spectrum in Figure~\ref{fig:spectrum} collapses into a single point for linear models: we do not have any distinction between the two methods. We often try to turn some complex non-linear models into linear ones locally. Therefore, it is of crucial importance to understand linear models.

\subsection{Zintgraf \etal: Inpainting + Black-Box Computation}


The \href{https://arxiv.org/abs/1702.04595}{Zintgraf \etal (2017)}~\cite{https://doi.org/10.48550/arxiv.1702.04595} attribution method employs global perturbations -- they measure missingness by imputation. It uses the ``naive way'' of computing the forward pass several times for computing counterfactual attributions. The proposed \emph{prediction difference analysis} reflects the fundamental notion of a counterfactual explanation very well. We want to obtain
\begin{align*}
P(c \mid x_{\setminus i}) &= \sum_{x_i} P(x_i \mid x_{\setminus i})\underbrace{P(c \mid x_{\setminus i}, x_i)}_{\text{trained network}}\\
&= \nE_{P(x_i \mid x_{\setminus i})}\left[P(c \mid x_{\setminus i}, x_i)\right],
\end{align*}
which is the probability of class c according to the network after removing feature \(i\).
As we do not know the true posterior \(P(x_i \mid x_{\setminus i})\)\footnote{The works considers image data. \(P(x_i \mid x_{\setminus i})\): distribution of feature \(i\) given all the other features in the image.} over the missing feature, we approximate it using an inpainting model
\[Q_{\mathrm{inpainter}}(x_i \mid x_{\setminus i}).\]
Therefore,
\begin{align*}
P(c \mid x_{\setminus i}) &\approx \nE_{Q_{\mathrm{inpainter}}(x_i \mid x_{\setminus i})}\left[P(c \mid x_{\setminus i}, x_i)\right]\\
&\approx \frac{1}{M} \sum_{m = 1}^M P(c \mid x_{\setminus i}, x^{(m)}_i)
\end{align*}
where \(x^{(m)}_i \sim Q_{\mathrm{inpainter}}(x_i \mid x_{\setminus i})\).
Finally, we calculate the counterfactual before and after removing feature \(i\) using the \emph{weight of evidence} value:
\[\operatorname{WE}_i(c \mid x) = \log_2(\operatorname{odds}(c \mid x)) - \log_2(\operatorname{odds}(c \mid x_{\setminus i})),\]
where \[\operatorname{odds}(c \mid x) = \frac{P(c \mid x)}{1 - P(c \mid x)}.\]

\begin{figure}
    \centering
    \includegraphics[width=0.6\linewidth]{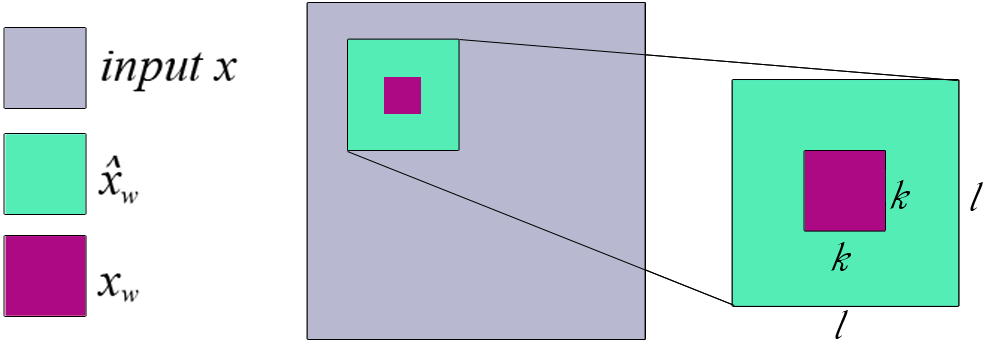}
    \caption{Illustration of the conditional independence assumptions used by Zintgraf \etal to make the conditioning tractable. A patch of size \(k \times k\) only depends on the surrounding pixels from an \(l \times l\) patch that contains the \(k \times k\) patch. Figure taken from~\cite{https://doi.org/10.48550/arxiv.1702.04595}.}
    \label{fig:zintgraf2}
\end{figure}

\begin{definition}{Mixture of Gaussians}
A Mixture of Gaussians (MoG) distribution with \(M\) components is of the form
\[P(x) = \frac{1}{M}\sum_{m = 1}^M \cN(x; \mu_m, \Sigma_m)\]
where \(\mu_m\) and \(\Sigma_m\) are the mean vector and covariance matrix of the \(m\)th component, respectively. The MoG distribution is one of the simplest \emph{multimodal} distributions.
\end{definition}

\subsubsection{Remarks for Zintgraf \etal}

In Zintgraf \etal, the features do not have to correspond to pixels. They correspond to image patches in this work. 

The weight of evidence is a signed value, as we consider evidence \emph{for} and \emph{against} the prediction. When \(\mathrm{WE}_i\) is negative for sliding window (image patch) \(i\), it is evidence \emph{against} the model's prediction. It is also often evidence \emph{for} the second-highest scoring class.

To compute the attribution scores, we could use any difference/comparison between \(P(c \mid x)\) and \(P(c \mid x_{\setminus i})\). The authors argue that using log odds is well-founded.

It is costly to do this procedure for all features \(i\). For each image, one needs to compute \(N\) forward operations for the main model + \(N\) inpainting computations, where
\[N = \text{number of sliding windows} \times \text{number of samples for inpainting}.\]

The authors propose two methods for estimating the true inpainting distributions \(P(x_i \mid x_{\setminus i})\). The first one is to assume \emph{independence} of feature \(x_i\) on other features \(x_{\setminus i}\). If we make such an assumption, we can consider the empirical distribution of feature \(x_i\) from the dataset, i.e., we replace the feature value with a different one sampled from the dataset at random. By sampling more possible feature values from the dataset (at the same image location), we Monte Carlo estimate the expectation. As the authors also state, this is a crude approximation. The second proposal of the paper is to not assume independence but to suppose that an image patch \(x_i\) of size \(k \times k\) \emph{only depends on the surrounding pixels} \(\hat{x}_i \setminus x_i\), where \(\hat{x}_i\) is an image patch of size \(l \times l\) that contains \(x_i\). An illustration is given in Figure~\ref{fig:zintgraf2}. To speed things up, the authors used a straightforward method for inpainting: a multivariate Gaussian inpainting distribution in pixel space, \href{https://github.com/lmzintgraf/DeepVis-PredDiff/blob/02649f2d8847fc23c58f9f2e5bcd97542673293d/utils_sampling.py#L146}{fit on dataset samples}. In particular, the authors calculate the empirical mean \(\mu_i\) and empirical covariance \(\Sigma_i\) of the large patch \(\hat{x}_i\) on the entire dataset, using the simplifying assumption that the distribution of the large patch \(\hat{x}_i\) (i.e., the \emph{joint} distribution of the window we want to sample from and the surrounding pixels) is a Gaussian: \(P(\hat{x}_i) = \cN(\hat{x}_i; \mu_i, \Sigma_i)\). Finally, the authors use the well-known conditioning formula for Gaussians to obtain \(P(x_i \mid \hat{x}_i \setminus x_i)\). Under their assumptions, we have
\[P(x_i \mid \hat{x}_i \setminus x_i) = P(x_i \mid x_{\setminus i}).\]
This is probably the simplest form of inpainting one could think of.

Other possibilities for the inpainting distribution: One could use a Mixture of Gaussians (MoG) or diffusion models~\cite{https://doi.org/10.48550/arxiv.2006.11239} for inpainting. However, then it would take even longer to compute the explanation for a single image. There is always a trade-off between complexity and quality.

The method of Zintgraf \etal is a \emph{local} explanation method (as it only gives an explanation for a single image) but a \emph{global} counterfactual method (because the inpainter is allowed to predict anything, not just very small perturbations compared to the original image features). Note, however, that the inpainter is only used to replace small patches -- it is still spatially local.

\subsubsection{Results of Zintgraf \etal}

\begin{figure}
    \centering
    \includegraphics[width=\linewidth]{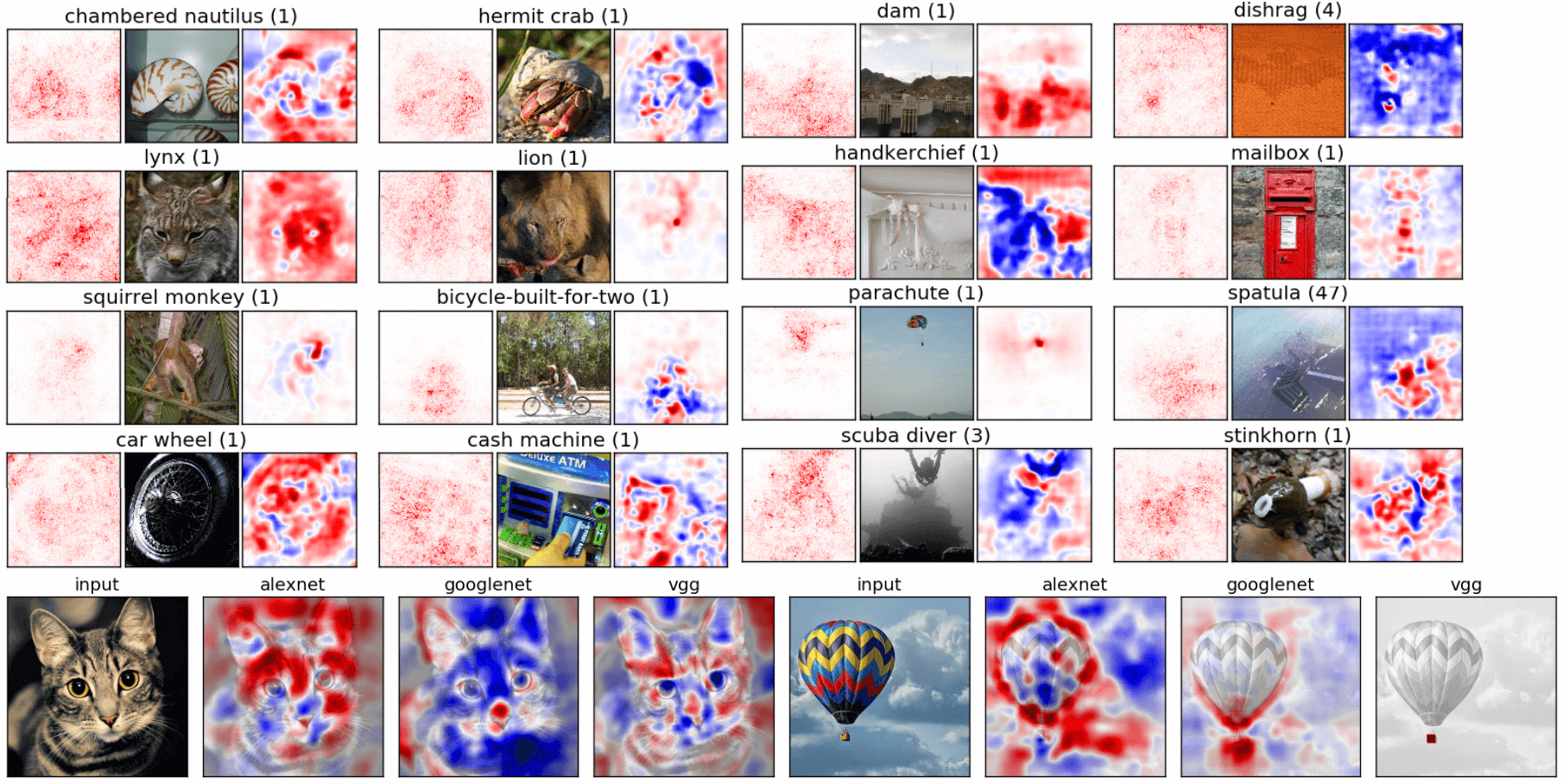}
    \caption{Results of Zintgraf \etal (2017)~\cite{https://doi.org/10.48550/arxiv.1702.04595}, taken from the paper. The attribution maps look surprisingly hard to interpret. Different architectures seem to look at notably different parts of the input image. Still, maybe it \emph{is} the genuine contribution of each feature to the network's prediction. We should not rely too much on human intuition, as that might harm our belief about the soundness of the method. It is hard to say whether this is right or wrong without a quantitative soundness evaluation.}
    \label{fig:zintgraf}
\end{figure}

We show a few attribution maps in Figure~\ref{fig:zintgraf}. We argue that this is the most promising solution for counterfactual attribution, but it is also the most computationally heavy. Let us now give some pros and cons of the method.
\begin{itemize}
    \item \textbf{Pro}: The method performs a global counterfactual analysis because of inpainting. It is also one of the few datatype-agnostic methods -- it can be applied to image, text, and tabular data inputs as well, given that an inpainter is available.
    \item \textbf{Contra}: The method is way too complex to be practical. It also depends on the inpainter, which opens a new can of worms.
\end{itemize}

\subsection{LIME: Fitting a Sparse Linear Model}
\label{sssec:lime}

LIME, introduced in the paper ``\href{https://arxiv.org/abs/1602.04938}{``Why Should I Trust You?'': Explaining the Predictions of Any Classifier}''~\cite{https://doi.org/10.48550/arxiv.1602.04938}, has been a popular method for more than five years now that is a bit more realistic than Zintgraf \etal regarding practical use. It builds a surrogate model that is explainable by definition. Given the general formulation
\[\xi(x) = \argmin_{g \in G} \cL(f, g, \pi_x) + \Omega(g)\]
where \(f\) is the original model, \(g\) is the surrogate model, \(G\) is the set of possible surrogate models, \(\pi_x\) is a measure of distance from \(x\) used to weight loss terms, and \(\Omega\) is a measure of complexity. The authors make the following choices: \(G\) should be a set of sparse linear models, and \(\Omega\) should be a sparsity regularizer for the linear model \(g\).

By optimizing the objective function, we try to make \(g\) as close to \(f\) as possible \emph{in the vicinity} of \(x\), the test input of interest, weighted by \(\pi_x\), while also keeping it sparse.

In LIME for images, we define
\begin{itemize}
    \item \(x\) as the original image,
    \item \(x'\) as the interpretable version of the original image: a binary indicator vector whether superpixel \(i\) is turned on or off (grayed out). Here, all entries are ones.
    \item \(z'\) as a sample around \(x'\) by drawing non-zero elements of \(x'\) uniformly at random. The number of draws is also uniformly sampled.
    \item \(z\) as \(z'\) transformed back to an actual image,
    \item \(f(z)\) as the probability that \(z\) belongs to the class being explained, and
    \item \(\cZ\) as the dataset of \((z, z')\) pairs.
\end{itemize}

We specify the sparse linear function \(g\) formally by
\[g(z') = w_g^\top z'\]
and the sparsity constraint by
\[\Omega(g) = \infty\bone\left(\Vert w_g \Vert_0 > K\right),\]
i.e., \(f\) should have at most \(K\) non-zero weights. The function fitting takes place around input \(x\). We let \(g\) follow \(f\) via the \(L_2\) loss on the function outputs
\[\cL(f, g, \pi_x) = \sum_{(z, z') \in \cZ} \pi_x(z) \left(f(z) - g(z')\right)^2,\]
with \(\pi_x\) making sure that we focus on fitting \(g\) to \(f\) only in the vicinity of \(x\) (we only aim for local faithfulness):
\[\pi_x(z) = \exp\left(-D(x, z)^2 / \sigma^2\right).\]
Here \(D\) is the cosine distance from \(x\) to \(z\) if the input is text, or the \(L_2\) distance for images.

\begin{figure}
    \centering
    \includegraphics[width=0.4\linewidth]{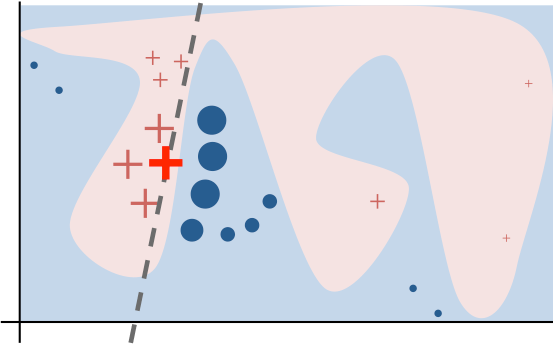}
    \caption{Toy example of LIME being fit to the bold red plus data point. The brown plus and blue circle samples are the sampled instances in the vicinity of the input being explained, \((z, z') \in \cZ\). Their size encodes their similarity with the original input, as given by \(\pi_x(z)\). The background contours encode the decision boundary of the complex model \(f\), whereas the dashed line encodes the decision boundary of \(g\). The surrogate model is locally faithful to the complex model. Figure taken from~\cite{https://doi.org/10.48550/arxiv.1602.04938}.}
    \label{fig:lime}
\end{figure}

An example of the fitting procedure is given in Figure~\ref{fig:lime}. The linear model learns to respect local changes of \(f\). This is close to taking the gradient, but we get a sparser linearization than that, which is more interpretable.

\begin{figure*}
\centering
\subfloat[Original Image]{
\includegraphics[width=0.24\textwidth]{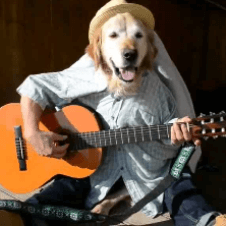}
\label{fig:original_image}}
\subfloat[Expl. \emph{Electric guitar}]{
\includegraphics[width=0.24\textwidth]{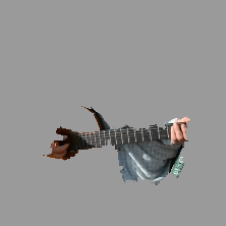}
\label{fig:electric}}
\subfloat[Expl. \emph{Acoustic guitar}]{
\includegraphics[width=0.24\textwidth]{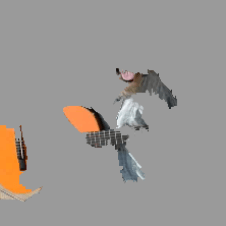}
\label{fig:acoustic}}
\subfloat[Expl. \emph{Labrador}\label{fig:labrador}]{
\includegraphics[width=0.24\textwidth]{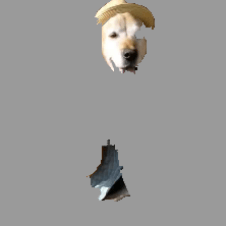}}
\caption{``Explaining an image classification prediction made by Google's Inception neural network. The top 3 classes predicted are `Electric Guitar' ($p=0.32$), `Acoustic guitar' ($p=0.24$) and `Labrador' ($p=0.21$).''~\cite{https://doi.org/10.48550/arxiv.1602.04938} The results look plausible and easy to interpret, but this should not be taken as an evaluation of soundness. Figure taken from~\cite{https://doi.org/10.48550/arxiv.1602.04938}.}
\label{fig:inception}
\end{figure*}

The workflow with LIME for images can be explained as follows.
\begin{enumerate}
    \item We pick an input \(x\) and the class to explain.
    \item We train a linear model on top of the superpixel features.
    \item We extract the surrogate model weights and check each superpixel's contribution.
    \item The superpixel corresponding to the largest weight contributes most to the class prediction in question.
\end{enumerate}
The authors do not only test the method on the actual prediction of the network. They deliberately come up with confusing images with multiple possible classes and try to explain the prediction of the network for the top \(k = 3\) predictions. This is shown in Figure~\ref{fig:inception}.

Let us discuss the pros and cons of the method.
\begin{itemize}
    \item \textbf{Pro}: The results are interpretable by design.
    \item \textbf{Contra}: (1) We only have a local sparse linear approximation that can be very different from the DNN. (2) The method is expensive, as a sparse linear model has to be fit for all images we want to be explained. (3) The reference image is assumed to be a gray image, an often-used representation of missingness. We discuss in \ref{sssec:missing} that this might be suboptimal. (4) The method is not stable. The given explanations (coefficients) are not continuous in the input and are, therefore, not general. In particular, \citeauthor{alvarezmelis2018robustness} show in the paper ``\href{https://arxiv.org/abs/1806.08049}{On the Robustness of Interpretability Methods}''~\cite{alvarezmelis2018robustness} that even explaining test instances that are very close/similar to each other leads to notably different results.
\end{itemize}

\textbf{Note 1}: The reference cannot be seen in the formulation but rather in how we construct the \(z\) samples:
\[z'_i = 0 \iff \text{superpixel \(i\) is gray}.\]
Thus, we still have an actual 0 value in ``interpretable space''; the related term does not contribute to the sum in the linear model.

\textbf{Note 2}: When we give an input to a DNN, we typically subtract the mean of the training set to center the input. So an original image that becomes a 0 input for a DNN is usually gray (the mean of the training set samples, close to constant gray for a versatile dataset). For ImageNet (and many other vision datasets), the standard practice is to \href{https://github.com/huggingface/pytorch-image-models/blob/da6644b6ba1a9a41f2815990111056bbf0b05c8e/timm/data/loader.py#L132}{subtract the mean}.

\begin{information}{Surrogate Model}
The LIME paper uses \(f(x)\) to denote the probability that \(x\) belongs to the class being explained. The surrogate \(g\) is, however, defined to be a \emph{linear} model that can, in principle, predict any real number and not just probabilities. We could have two other options for defining the surrogate model.
\begin{enumerate}
    \item Use the \emph{logit} values of the classifier as the targets for the surrogate model. This way, we are matching a real number to another (unconstrained) real number, which seems more natural. However, the coefficients of the surrogate model do not correspond to the changes in the model \emph{output} anymore, but rather to the changes in the logits that are more disconnected from the model's final decision than its predicted probabilities.
    \item Constrain the surrogate model's outputs to the \((0, 1)\) range, e.g., by using a logistic sigmoid activation function. This way, we could use any classification loss to train the surrogate model -- we are matching probabilities to probabilities. The downside is that the surrogate model outputs are not \emph{linearly} related to the outputs anymore, and the attribution scores become less interpretable.
\end{enumerate}
\end{information}

\subsection{SHAP (SHapley Additive exPlanations)}

The setup of the SHAP method, introduced in the paper ``\href{https://arxiv.org/abs/1705.07874}{A Unified Approach to Interpreting Model Predictions}''~\cite{https://doi.org/10.48550/arxiv.1705.07874}, is very similar to that of the LIME method in terms of the knowledge about the system and the input/output format. In particular, we assume a black-box system with a binary input vector \(x \in \{0, 1\}^N\) that gives us scores \(f(x) \in \nR\) for a particular class \(c\). We want to assign the contribution of each feature \(i\) to the prediction.

The input is represented by a given set of features. The binary membership indicator \(x\) is a constant one vector: in the original input, all features are present. For perturbed inputs \(z \subseteq x\), zeros and ones indicate whether the corresponding feature is present or turned off in the perturbed image. As we have binary input features, we have a clear interpretation of turning on (1) and turning off (0) features. For images, this is usually a superpixel representation, where the constant one vector is the full image, and the subsets specify which superpixels we switch off (i.e., replace with some base value) and on.

\begin{definition}{Combination}
The number of possible ways to choose \(k\) objects from \(n\) objects is
\[\binom{n}{k} = \frac{n!}{k!(n - k)!}.\]
\end{definition}

SHAP determines the individual contribution of each feature \(i\) to the prediction \(f(x)\) using the notion of Shapley values~\cite{shapley1953value}. The value is defined as
\[\phi_{f, x}(i) = \nE_{z \subseteq x: i \in z}\left[f(z) - f(z - i)\right].\]
This value gives the average contribution of feature \(i\) in all subset cases to the output of network \(f\). \(z\) is a subset of \(x\) that must include \(i\). For every subset, we analyze the effect of discarding feature \(i\). This can be thought of as a set function version of the gradient of \(f\) at \(x\) \wrt feature \(i\). The original input \(x\) is always treated as \([1, 1, \dots, 1]\) (all features are turned on), and an example of a valid sample \(z\) is \([0, 1, 0, 0, 1, 0]\) for index \(i = 2\) if \(x \in \{0, 1\}^6\). (The indexing starts from \(1\).) The possible subsets \(z\) are thus any binary vector of the same dimensionality as \(x\). It also follows from the formulation that Shapley values are \emph{signed}, unlike, e.g., saliency maps. Similarly to LIME, we give an attribution score to each feature (e.g., superpixel) \wrt turning them on/off (global counterfactual explanation). 

\textbf{Note}: The expectation in the SHAP attribution values is \emph{not} uniform across all possible \(z\)s that are subsets of \(x\). The expectation follows the procedure below:
\begin{enumerate}
    \item Sample subset size \(m\) from \(\text{Unif}\{1, \dots, |x|\}\).\footnote{Subset \(z\) must include \(i\), so the minimal size of \(z\) is 1.}
    \item Sample a subset \(z\) of size \(m\) containing feature \(i\) with equal probabilities.
\end{enumerate}
Not every subset across all subset sizes has the same probability of being picked because of sample size differences. If \(|x| = 10\), then \(\binom{9}{4} \gg \binom{9}{9}\), meaning particular small or large subsets are much more likely than particular medium-sized ones.

\begin{figure}
    \centering
    \includegraphics[width=\linewidth]{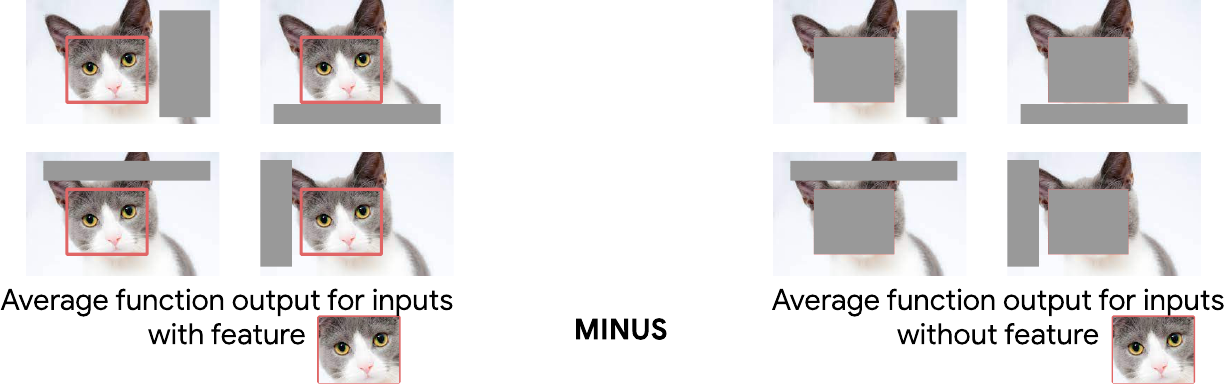}
    \caption{Illustration of the computation of Shapley values. This is equivalent to the formulation above because the expectation is linear.}
    \label{fig:shapcat}
\end{figure}

\textbf{Example}: Let us consider features as image patches. Suppose that feature \(i\) indicates the face region of the cat. To calculate the Shapley value corresponding to feature \(i\), we average the function output for all possible inputs with \(i\) switched on (other parts are free to vary), then we \emph{subtract} the average function output for all possible inputs with \(i\) switched off. The example is illustrated in Figure~\ref{fig:shapcat}.

We rewrite the expectation as
\begin{align*}
\phi_{f, x}(i) &= \nE_{z \subseteq x: i \in z}\left[f(z) - f(z - i)\right]\\
&= \frac{1}{|x|} \sum_{z \subseteq x: i \in z} \binom{|x| - 1}{|z| - 1}^{-1}\left[f(z) - f(z - i)\right]\\
&= \sum_{z \subseteq x: i \in z} \frac{(|z| - 1)!(|x| - |z|)!}{|x|!}\left[f(z) - f(z - i)\right]
\end{align*}
by leveraging that the probability of sampling \(z\) is equal to the probability of subset size \(|z|\) times the probability of choosing a particular subset of size \(|z|\).

\subsubsection{SHAP also satisfies the completeness axiom.}

\begin{claim}
If we sum over the Shapley values for all features \(i\), then we get the difference of the function value for the input of interest \(x\) and the prediction for the baseline \(0\):
\[\sum_{i} \phi_{f, x}(i) = f(x) - f(0).\]
\end{claim}
\begin{proof}
\[\sum_{i} \phi_{f, x}(i) = \sum_i \sum_{z: i \in z \subseteq x} \frac{(|z| - 1)!(|x| - |z|)!}{|x|!}\left[f(z) - f(z - i)\right].\]

Here, `\(\cdot f(z)\)' appears \(|z|\) times (\(|z| \in \{1, \dots, |x|\}\)) with a \emph{positive} sign, once for each feature \(i\) in \(z\). Its coefficient is always
\[\frac{(|z| - 1)!(|x| - |z|)!}{|x|!},\]
thus \(|z|\) times the coefficient gives \[\binom{|x|}{|z|}^{-1}.\]

Similarly, `\(\cdot f(z)\)' appears \(|x| - |z|\) times (\(|z| \in \{0, \dots, |x| - 1\}\)) with a \emph{negative} sign, once for each feature \(i\) \emph{not} in \(z\). Its coefficient is always
\[\frac{|z|!(|x| - |z| + 1)!}{|x|!}\]
as we consider \(|z| \gets |z| + 1\) in the formula, thus \(|x| - |z|\) times the coefficient gives \[\binom{|x|}{|z|}^{-1}.\]

The terms of the previous two paragraphs obviously cancel whenever \(z \notin \{0, x\}\).

For \(z = 0\), \(f(z)\) appears \(|x|\) times with a \emph{negative} sign. Its coefficient is always
\[\frac{0!(|x| - 1)!}{|x|!} = \frac{1}{|x|},\]
thus, \(|x|\) times the coefficient gives \(1\). Therefore, the term gives \(-f(0)\) in the sum.

For \(z = x\), \(f(z)\) appears \(|x|\) times with a \emph{positive} sign. Its coefficient is always
\[\frac{(|x| - 1)!0!}{|x|!} = \frac{1}{|x|},\]
thus, \(|x|\) times the coefficient gives \(1\). Therefore, the term gives \(+f(x)\) in the sum.

Finally, by summing all terms up, we indeed obtain \(\sum_i \phi_{f, x}(i) = f(x) - f(0)\).
\end{proof}
\textbf{Note}: The \(0\) vector can mean arbitrary missingness in the pixel space, just like in LIME. For integrated gradients, we had a very similar result: When we sum over all contributions from every pixel, we obtain \(f(x) - f(x^0)\). The difference is that we are not in the pixel space with SHAP.

\subsubsection{SHAP satisfies the strong monotonicity property.}

\begin{definition}{Strong Monotonicity}
Attribution values \(\phi\) satisfy the strong monotonicity property if, for every function \(f\) and \(f'\), binary input \(x\) and feature \(i\), the following holds:
\[f(z) - f(z - i) \le f'(z) - f'(z - i)\quad \forall z \subseteq x \text{ s.t. } i \in z \quad \implies \quad \phi_{f, x}(i) \le \phi_{f', x}(i).\]
In words, if the impact of deleting feature \(i\) is more significant for \(f'\) for all subsets of \(x\) containing \(i\), then the attribution value for \(f'\) on feature \(i\) must be greater than that for \(f\).
\end{definition}

The fact that SHAP satisfies the strong monotonicity property follows trivially from its formulation. This seems to be a very reasonable property\footnote{If a function values a feature a lot, then that is also reflected in the Shapley value.} but should not be deemed crucial. Below, we will see that Shapley values are special for measuring contribution.

\textbf{Uniqueness}: The attribution values \(\phi\) of SHAP are the only ones that satisfy both the strong monotonicity and the completeness axiom~\cite{young1985monotonic}. The theorem is well-known in the game theory literature. This roughly translates to: ``If we want these nice properties, we must use SHAP.'' Thus, \emph{SHAP is sufficient and necessary for these two properties to hold jointly.} The coefficients for Shapley values are, therefore, significant to be exactly these.

\subsubsection{Why do we want these properties?}

Why are strong monotonicity and completeness useful from an applicability point of view? We do not have a strong argument for why this should be the ``holy grail'' for attribution. The paper also does not give a strong reason why these properties should be strongly connected to any real-world properties. Such works that are built upon axiomatic foundations that introduce some intuitive requirements (e.g., strong monotonicity or completeness axioms) usually conclude that the only method that satisfies all the axioms is theirs. But they usually take \emph{different axioms}, which results in different formulations.  The integrated gradients method is also a unique formulation that satisfies a different set of axioms~\cite{sundararajan2017axiomatic}. Everything depends on how we choose these axioms. We do not think that any of the axioms are \emph{absolute necessities}. They are just one way to connect possible real-world needs to an actual explanation method we wish to have.

\subsubsection{Using SHAP in practice}

We approximate the Shapley values by sampling the expectation at random, according to the coefficients (choose size uniformly, choose a set of that size uniformly). This avoids traversing through the combinatorial number of subsets but introduces large variance in the Monte Carlo approximation, leading to a decreased trustworthiness of the attribution scores.

Let us consider the pros and cons of SHAP.
\begin{itemize}
    \item \textbf{Pro}: Similarly to LIME, the results are interpretable by design. The method also gives global counterfactual analysis.
    \item \textbf{Contra}: (1) We have to use efficient approximations of the Shapley values to keep tractability. Depending on the variance of our approximations, the results we obtain this way might not be faithful to the true Shapley values. (2) \citeauthor{alvarezmelis2018robustness}~\cite{alvarezmelis2018robustness} show also for SHAP that the attribution scores can change significantly in small input neighborhoods. (3) Just like in LIME, the reference image is assumed to be a gray image (the mean of the training distribution) in the paper. This might have unfavorable implications, which we will further discuss in Section~\ref{sssec:missing}.
\end{itemize}

\subsection{Defining a Missing Feature}
\label{sssec:missing}

We needed a good definition of ``no information'' for the methods discussed previously.

In \emph{integrated gradients}, we use black pixels as missing features, which is empirically justified in~\cite{sundararajan2017axiomatic}. This gradually kills information by dimming and considers the effect for each pixel integrated through the procedure.

\emph{Zintgraf \etal} use inpainting as missing features, which is, perhaps, a more sensible choice to encode missingness than any fixed color.

\emph{LIME} takes the mean pixel values to indicate missingness (which corresponds to gray pixels for most datasets of natural images).

In \emph{SHAP}, missingness is indicated the same way as in LIME. \textbf{Note}: Completeness holds when we consider a 0 \emph{vector}. It can correspond to \emph{any} image. The authors equate that to a gray image, but one could make different choices, such as black/white images or Gaussian noise. The choice of what the 0 vector encodes could also be made arbitrarily for LIME. The integrated gradients method also gives a freedom of choice in designing the baseline image. Usually, the choice is made based on results from cross-validation or qualitative analysis (the latter often being flawed). It is also important to remark that neither of the methods is restricted to images, and we have to reason about the definition of ``missingness'' for other kinds of data in the same way. For example, for tabular data, both a zero value and the mean value of the dataset make intuitive sense, but they might give different results. 

\begin{figure}
    \centering
    \includegraphics[width=0.7\linewidth]{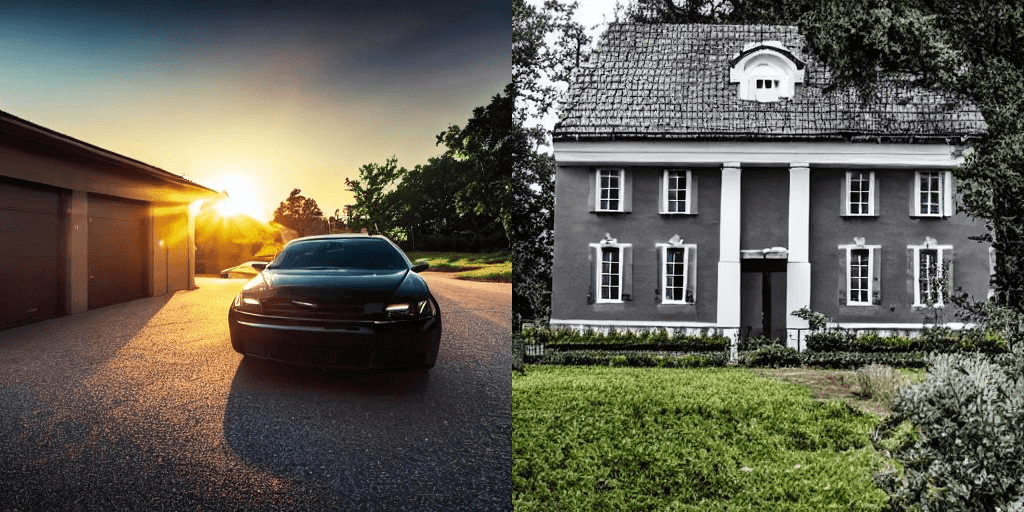}
    \caption{Example of black and gray colors -- popular choices for encoding missingness -- conveying information in images. Choosing \emph{any} fixed color to encode missingness is questionable. The images were generated by Stable Diffusion~\cite{https://doi.org/10.48550/arxiv.2112.10752}.}
    \label{fig:missingness2}
\end{figure}

As discussed previously, we consider inpainting to be the most promising approach to defining missingness. The problem with fixed missing feature values is that they can also carry information (e.g., black pixels on a car or gray pixels on a house, illustrated in Figure~\ref{fig:missingness2}), might matter a lot for the prediction but might not be attributed at all. Such pixel values can appear in natural images, yet they will automatically have a zero value in integrated gradients. This is, of course, problematic. The problem can even arise in CALM or SHAP, though perhaps not as severely as in integrated gradients: If a particular superpixel has the same constant value as the mean pixel, turning it on or off does not have any effect, so the attribution value is necessarily zero.

Using constant pixel values to encode missingness also causes problems when considering soundness evaluation methods such as remove-and-classify, introduced in Section~\ref{sssec:rac}, as it can introduce missingness bias, discussed in Section~\ref{sssec:missingness_bias}.

\begin{information}{Inpainting Models}
Language models are also often inpainting models (context prediction self-supervised learning (SSL) objective). To get performant solutions, one needs a huge model. The same goes for diffusion-based inpainting models. They are also huge pre-trained models that can synthesize more realistic images. Inpainting is not as easy as it sounds.
\end{information}

\subsection{Meaningful Perturbations}

\begin{figure}
    \centering
    \includegraphics[width=0.8\linewidth]{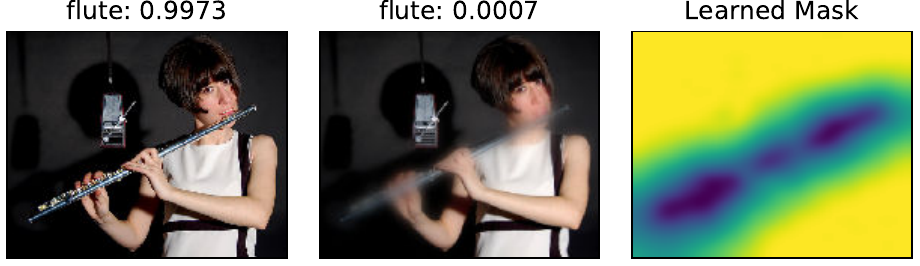}
    \caption{Example of a learned blur that results in diminished predictive performance, taken from~\cite{Fong_2017}.}
    \label{fig:learnedblur}
\end{figure}

Now, we discuss the ``\href{https://arxiv.org/abs/1704.03296}{Interpretable Explanations of Black Boxes by Meaningful Perturbation}''~\cite{Fong_2017} paper that introduces \emph{meaningful perturbations}. Instead of different colors encoding missing features, one can also use \emph{learned blurring}. Image blurring can erase information without potentially introducing some. (However, for humans, it might not be enough. Considering an image of a person playing the flute, even if we blur the flute out, a human still knows what is in their hands. However, in this paper, the authors demonstrated that DNNs do not work like this, as shown in Figure~\ref{fig:learnedblur}.)

The authors are optimizing for the \emph{blur mask}. After the optimization, the final blurred region is ideally the most important region for predicting the corresponding label. The optimization problem is
\[m^* = \argmin_{m \in [0, 1]^\Lambda}\lambda \Vert \bone - m \Vert_1 + f_c(\Phi(x_0; m))\]
where
\begin{itemize}
    \item \(m\): A continuous relaxation of a binary mask that associates each pixel \(u \in \Lambda\) with a scalar value \(m(u) \in [0, 1]\).
    \begin{itemize}
        \item \(m(u) = 1\): We do not perturb the pixel at all.
        \item \(m(u) = 0\): We perturb the pixel (region) as much as possible.
    \end{itemize}
    \item \(m^*\): Mask that erases most information from the image while also being sparse.
    \item \(\Vert \bone - m \Vert_1\): Measures the area of the erased region. As \(m\) is continuous (smooth), the magnitude matters. \(L_1\) regularization encourages the mask to be sparse. This can be considered as a relaxation of the NP-hard problem using \(\lambda \Vert \bone - m \Vert_0\) plus \(m \in \{0, 1\}^\Lambda\).
    \item \(f_c\): Classifier score for class \(c\). We want to minimize this in a regularized fashion.
    \item \(\Phi(x_0; m)\): The perturbation operator, e.g., blurring of original image \(x_0\) according to the mask \(m\):
    \[\left[\Phi(x_0; m)\right](u) = \int g_{\sigma_0 \cdot (1 - m(u))}(v - u) \cdot x_0(v)\ dv\]
    where \(\sigma_0 = 10\) is the maximum isotropic standard deviation of the Gaussian blur kernel.
\end{itemize}
The objective is fully differentiable \wrt \(m\); one can train end-to-end with Gradient Descent (GD).

\subsubsection{Use cases of meaningful perturbations}

\begin{figure}
    \centering
    \includegraphics[width=0.9\linewidth, trim=0 1em 0 1em]{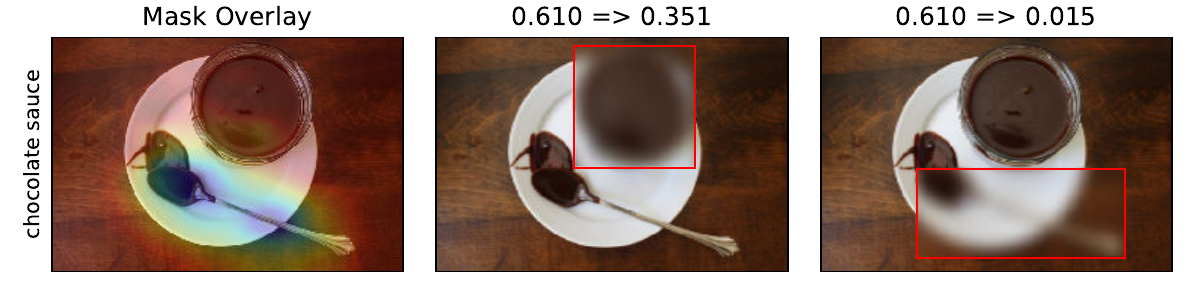}\\
    \includegraphics[width=0.9\linewidth, trim=0 1em 0 0]{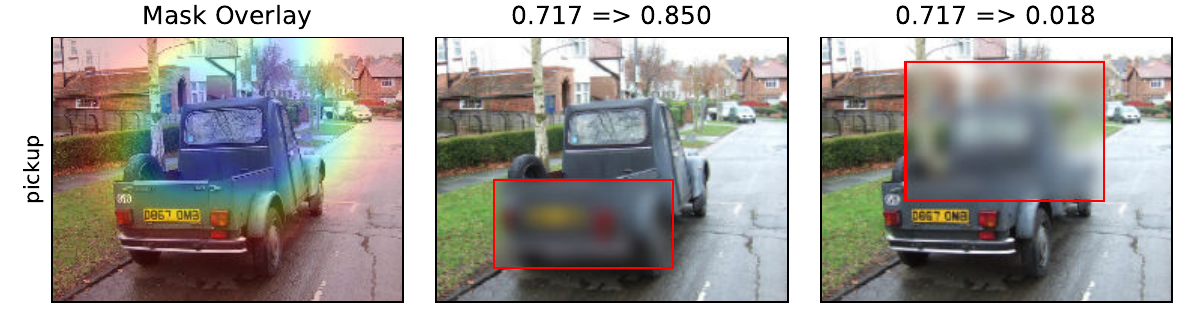}
    \caption{One can use the Meaningful Perturbations method to unveil spurious correlations. Figure taken from~\cite{Fong_2017}.}
    \label{fig:chocolate_sauce}
\end{figure}

After optimization, we can look at the learned continuous mask to see what region(s) have a large effect. This can unveil very interesting properties of our model. For example, to determine whether chocolate sauce is in the image, our model might be looking more at the spoon than the actual sauce (meaning the score decreases more for blurring this region), as depicted in Figure~\ref{fig:chocolate_sauce}. Thus, we can even detect spurious correlations with the method. (``Did my model learn the wrong association?'') After detection, we can fix them. This is much more direct than the counterfactual evaluation introduced in Section~\ref{sssec:identify}.

Considering the inherent linearity of various XAI methods (\ref{ssec:linearization}), this method does not explicitly give rise to a linear approximation of \(f(x)\), but it might be possible to obtain a linear formula in the \emph{transformed} attributions \(T(m)\) by embedding them in a non-linear fashion and still keeping them interpretable. Another possibility is that the method linearizes the model's prediction, just not in the attributions but in another property. 

\subsection{Testing with Concept Activation Vectors (TCAV)}
\label{sssec:tcav}

Let us go beyond the previous low-level features. We look into higher-level and human-understandable ones because interpretable features are more relevant for most real-life applications. Saliency maps use the gradient directly to attribute to individual pixels. If we look at saliency maps, we usually gain no information about where the important object/region is for a particular label. They are simply too noisy to read and trust and to understand a network's prediction. Even if we choose other pixel attribution methods, these are not interpretable features and do not allow us to relate to more abstract \emph{concepts}. What we really want to ask~\cite{https://doi.org/10.48550/arxiv.1711.11279}:
\begin{itemize}
    \item ``Was the model looking at the cash machine or the person to make the prediction?''
    \item ``Did the `human' concept matter?''
    \item ``Did the `glass' or `paper' concept matter?''
    \item ``Which concept mattered more?''
    \item ``Is this true for all other predictions of the same class?''
\end{itemize}
These are much more semantic questions than the previous methods can handle. This is because while most concepts can be expressed through examples/natural language, they are often impossible to explain in terms of input gradients or more sophisticated scores at the pixel/pixel aggregation level.

\begin{figure}
    \centering
    \includegraphics[width=0.8\linewidth]{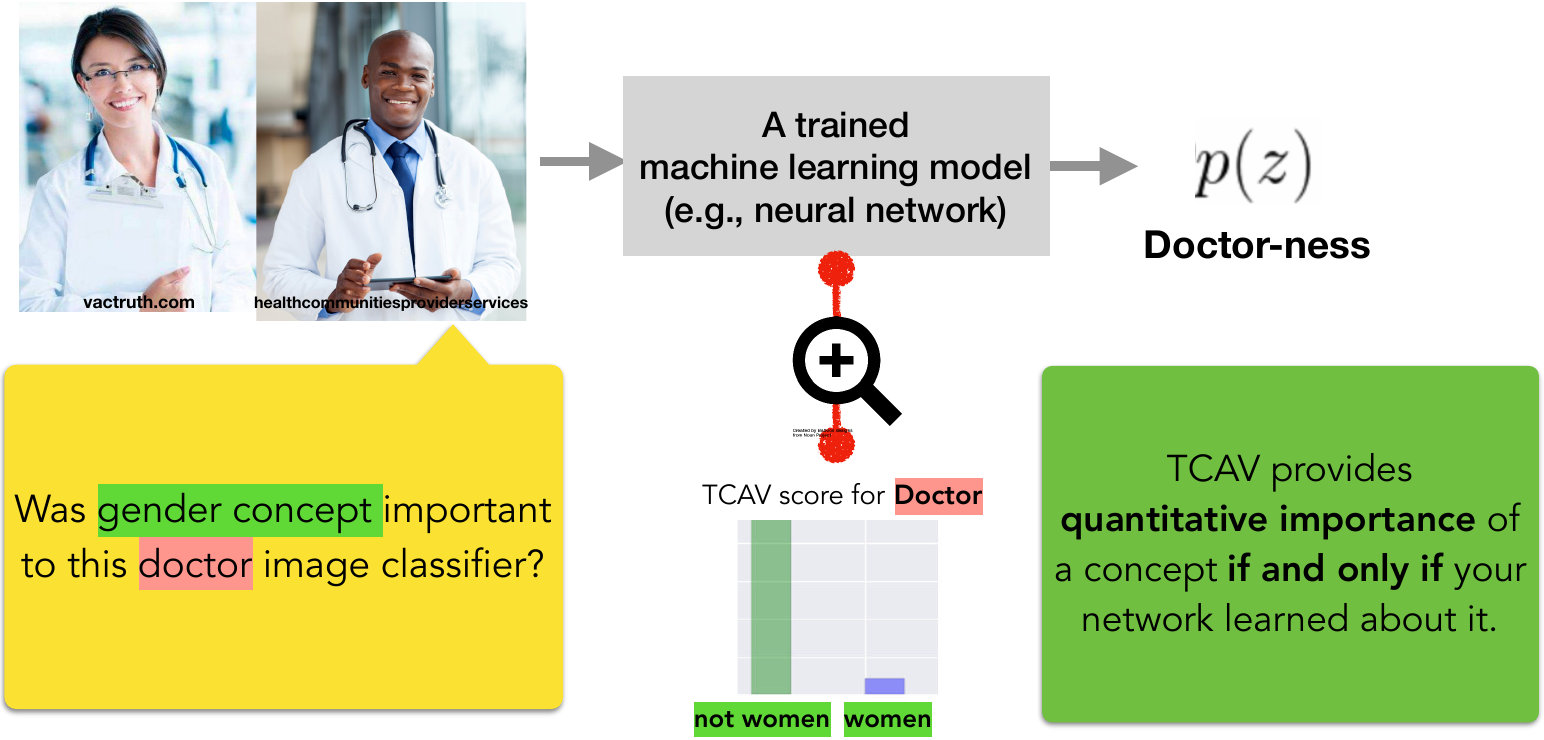}
    \caption{Overview of the TCAV method that attributes to human-interpretable concepts. Figure taken from the \href{https://beenkim.github.io/slides/TCAV_ICML_pdf.pdf}{ICML presentation slides} of~\cite{https://doi.org/10.48550/arxiv.1711.11279}. }
    \label{fig:tcav}
\end{figure}

\emph{TCAV}, introduced in the paper ``\href{https://arxiv.org/abs/1711.11279}{Interpretability Beyond Feature Attribution: Quantitative Testing with Concept Activation Vectors (TCAV)}''~\cite{https://doi.org/10.48550/arxiv.1711.11279}, is a method that allows us to ask whether an abstract concept mattered in the prediction. Figure~\ref{fig:tcav} gives an overview of the method through an intuitive example. We have a classifier with one of the classes being ``doctor''. We want to know whether some abstract concept was important in predicting \(P(z)\), the ``doctor-ness''. A concept does not have to be an explicit part of training: It can be implicitly globally encoded into the whole image. Instead of relying on gradients/pixel-wise or superpixel-wise attributions, we directly attribute to the human-understandable concept, e.g., woman/not woman.

\subsubsection{Attributing to high-level concepts}

Let us first introduce the notation used in the paper:
\begin{itemize}
    \item \(C\): concept;
    \item \(l\): layer index;
    \item \(k\): class index;
    \item \(X_k\): all inputs with label \(k\) (e.g., in the training set).
\end{itemize}

\begin{figure}
    \centering
    \includegraphics[width=0.8\linewidth]{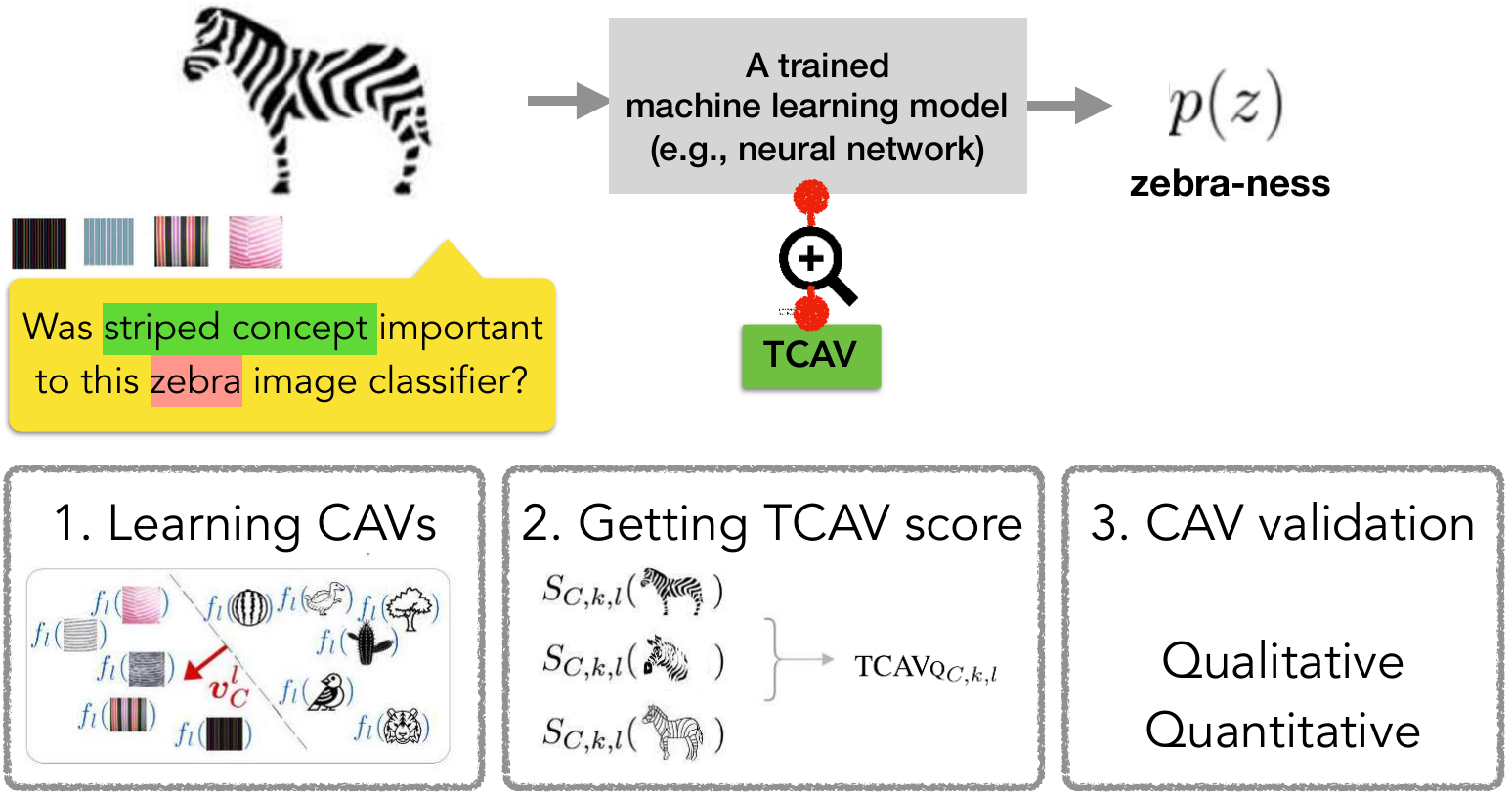}
    \caption{Individual stages of the TCAV pipeline, taken from the \href{https://beenkim.github.io/slides/TCAV_ICML_pdf.pdf}{ICML presentation slides} of~\cite{https://doi.org/10.48550/arxiv.1711.11279}. Quantitative CAV validation can be performed using statistical testing \wrt the set of random samples by validating that the distribution of the obtained TCAV scores is statistically different from that of random TCAV scores. For example, one can use a t-test.}
    \label{fig:tcavzebra}
\end{figure}

Consider the (already trained) sub-network \(f_l: \nR^n \rightarrow \nR^m\) whose output is an intermediate representation of dimension \(m\), corresponding to layer \(l\). We denote the ``remaining net'' that gives the score to class \(k\) by \(h_{l, k}: \nR^m \rightarrow \nR\). The method can be summarized as follows (Figure~\ref{fig:tcavzebra}). We prepare a set of positive and negative samples for the concept (e.g., images containing stripes and other random images). We also prepare images for the studied class (e.g., from the training set). We train a linear classifier to separate the activations of the intermediate layer \(l\) between the positive and negative samples for the concept. The Concept Activation Vector (CAV) \(v_C^l\) is the vector \emph{orthogonal to the decision boundary of the linear classifier}. This is cheap to obtain: the normal of the decision boundary is the weight vector that points into the positive class. For a particular input \(x\), we consider the \emph{directional derivative of the prediction \(h_{l, k}(f_l(x))\) \wrt the intermediate feature representation of \(x\), \(f_l(x)\), in the direction of the CAV}:
\begin{align*}
S_{C, k, l}(x) &= \lim_{\epsilon \rightarrow 0} \frac{h_{l, k}(f_l(x) + \epsilon v^l_C) - h_{l, k}(f_l(x))}{\epsilon}\\
&= \nabla_{f_l(x)} h_{l, k}(f_l(x))^\top v^l_C.
\end{align*}
We treat this as the \emph{score} of how much the concept contributed to the class prediction for this particular example. (How would it influence our predictions if we moved a tiny bit in the direction of the concept vector in the feature space?) If the directional derivative is positive, the concept positively impacts classifying the input as the class. Otherwise, the concept has a negative impact.

Finally, the TCAV score for a set of inputs with label \(k\), \(X_k\), is calculated as
\[\text{TCAV}_{Q_{C, k, l}} := \frac{\left|\left\{x \in X_k: S_{C, k, l}(x) > 0\right\}\right|}{\left| X_k \right|} \in [0, 1].\]
In words: \(\text{TCAV}_{Q_{C, k, l}}\) is the fraction of samples in the dataset with label \(k\) where the contribution of the concept was positive for the prediction of the class. This metric only depends on the sign of the scores \(S_{C, k, l}\); one could also consider the magnitude of conceptual sensitivities. The TCAV score turns the \emph{instance-specific} analysis (\(S_{C, k, l}\), local explanation method) into a more \emph{global} one, for a particular class in general (\(\text{TCAV}_{Q_{C, k, l}}\), more global explanation method). It tells us whether the \emph{presence} of the concept is important for a class in general.

\subsubsection{TCAV Results}

\begin{figure}
    \centering
    \includegraphics[width=0.5\linewidth]{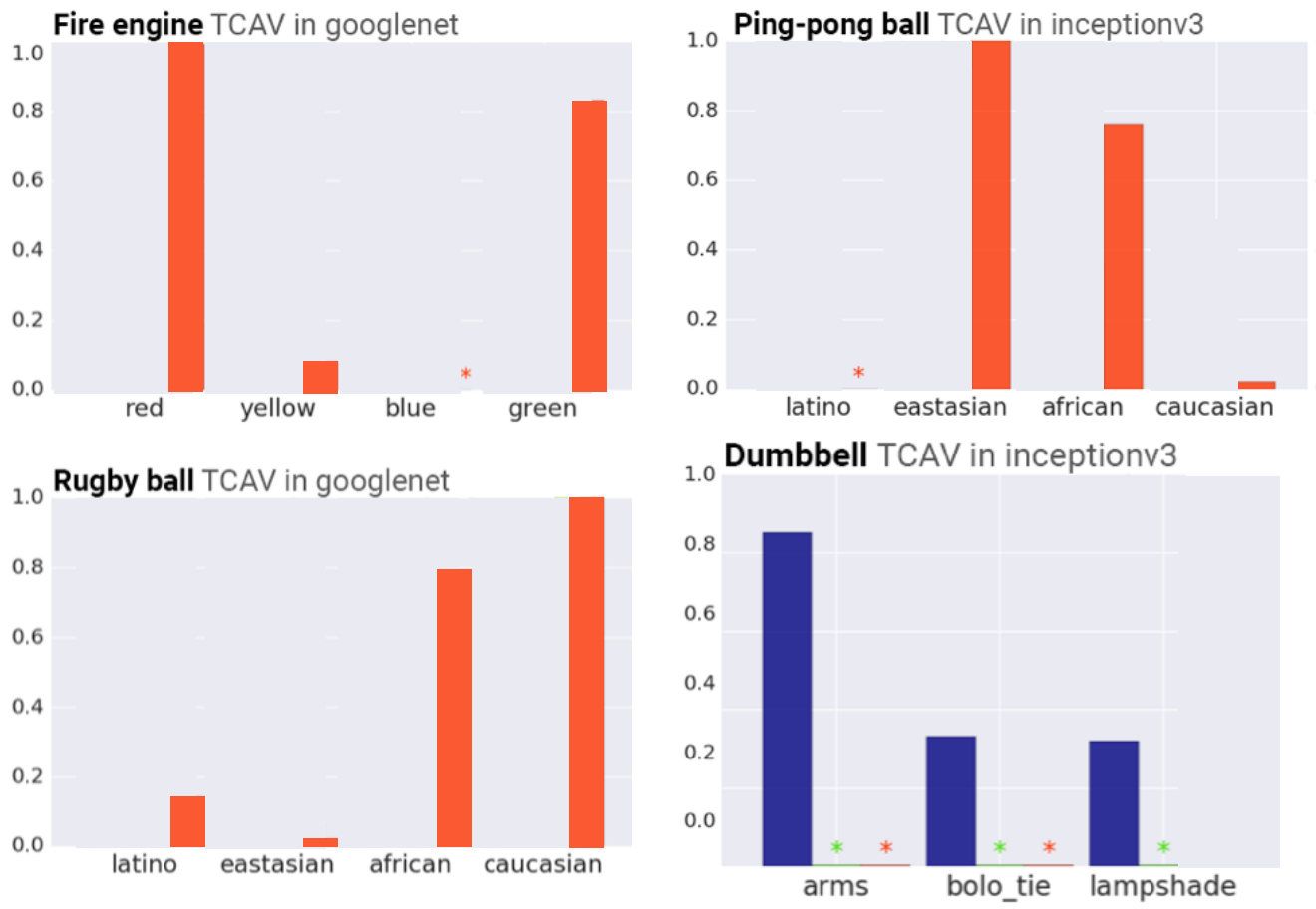}
    \caption{Qualitative results of the TCAV method on GoogLeNet and Inception-v3, taken from the \href{https://beenkim.github.io/slides/TCAV_ICML_pdf.pdf}{ICML presentation slides} of~\cite{https://doi.org/10.48550/arxiv.1711.11279}. Stars mark CAVs omitted after statistical testing \wrt different random images. One can see the concepts the model looks at to make predictions. TCAV can measure how important the presence of \{red, yellow, blue, green\} color is for the prediction of `fire engine'. The experiment results show that the red and green colors are important. This signals a strong geographical bias towards countries in the dataset with red and green fire engines. TCAV can also measure how important the presence of different ethnicities is for the prediction of `ping-pong ball'. The result of the experiments is that the East Asian and African concepts are important. This signals a strong bias towards the ethnicity of players. Agreeing with human intuition, the `arms' concept is more important for the prediction of `dumbbell' than the `bolo tie' or `lamp shape' ones.}
    \label{fig:tcavres}
\end{figure}

\begin{figure}
    \centering
    \includegraphics[width=0.9\linewidth]{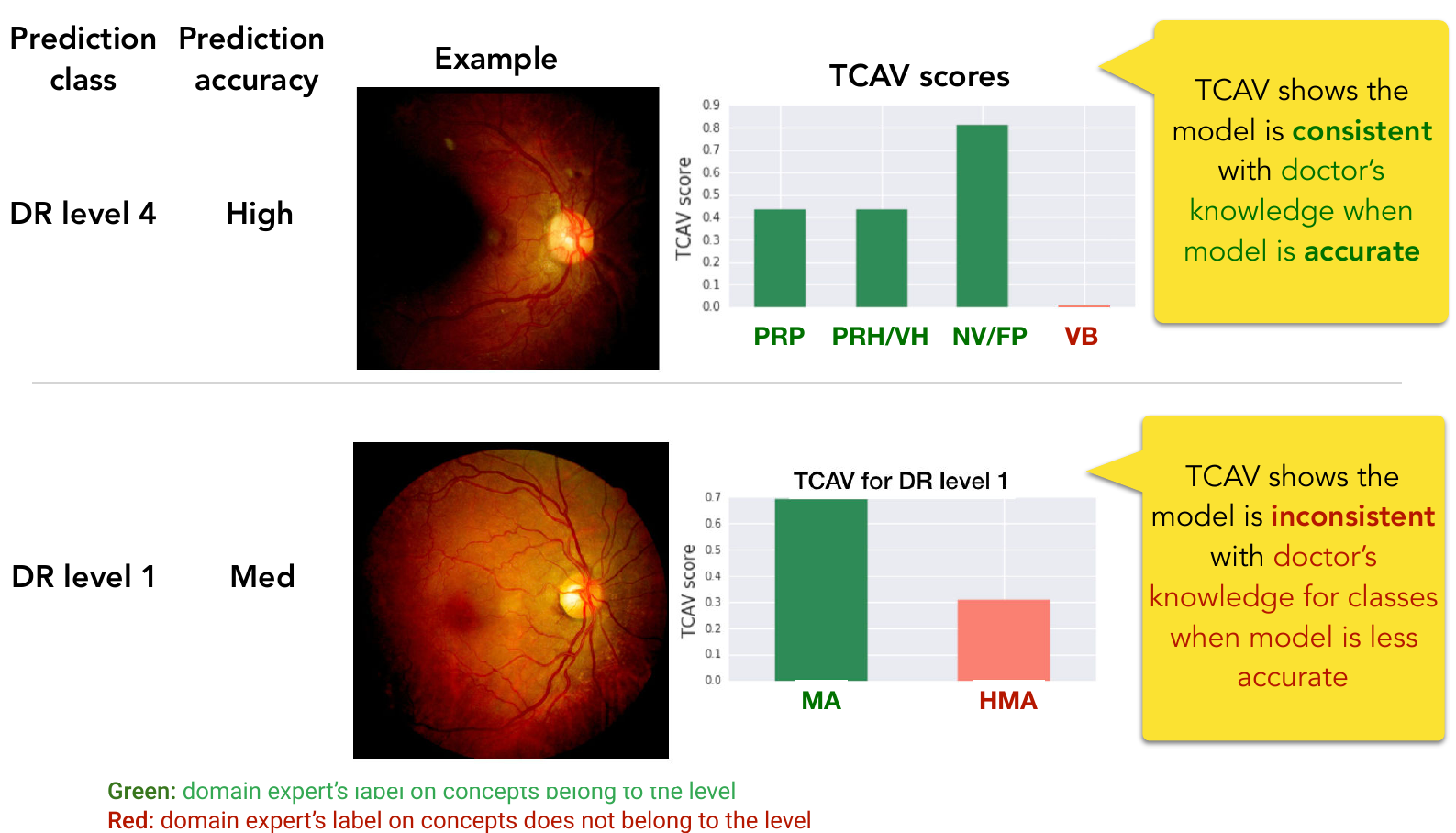}
    \caption{Results of using the TCAV method for Diabetic Rethinopathy, taken from the \href{https://beenkim.github.io/slides/TCAV_ICML_pdf.pdf}{ICML presentation slides} of~\cite{https://doi.org/10.48550/arxiv.1711.11279}. When the model is accurate, TCAV also shows that it is consistent with the doctor's knowledge: It gives high scores to features deemed by doctors as a precise cause for the prediction. When the model is less accurate, TCAV shows that the model is inconsistent with the doctor's knowledge: It gives a high score to a concept that the doctors deem not helpful to look at.}
    \label{fig:tcavdiab}
\end{figure}

Qualitative results of TCAV are shown in Figure~\ref{fig:tcavres}. TCAV can also shine in medical image analysis, as shown in Figure~\ref{fig:tcavdiab}. TCAV can streamline the interaction between humans and computers for making predictions.

Let us discuss some pros and cons of the method~\cite{molnar2020interpretable}. 
\begin{itemize}
    \item \textbf{Pro:} TCAV produces global explanations and can therefore provide insights into how the model works as a whole. It allows users to investigate any concept they define and is, therefore, flexible. 
    \item \textbf{Contra:} While the flexibility to investigate user-defined concepts is an advantage, it also has its downside: TCAV may require additional annotation/efforts to construct a concept dataset. Depending on the user's needs, TCAV may not easily scale to many concepts. Furthermore, TCAV requires a good separation of concepts in the latent space. If a model does not learn such a latent space, TCAV struggles and may not be applicable, as e.g. in shallow networks. 
\end{itemize}

\subsection{Class Activation Maps (CAM)}

\emph{CAM}, introduced in the paper ``\href{https://arxiv.org/abs/1512.04150}{Learning Deep Features for Discriminative Localization}''~\cite{https://doi.org/10.48550/arxiv.1512.04150}, is a method that attributes to interpretable intermediate features. A high-level overview of the method is shown in Figure~\ref{fig:camsimple}. CAM employs a typical CNN-based architecture with only a linear operation after calculating the intermediate score map. Up to the score map, the network is very complicated. Afterward, it is just a linear model using Global Average Pooling (GAP) and an intrinsically interpretable linear layer. The key assumption of CAM is that the attribution to pixels in the score map ``kind of'' corresponds to the attribution to original pixels. This is a huge leap of trust, but CNNs preserve localized information throughout the network (as given by the receptive field of individual neurons). Thus, the explanation \wrt the score map also roughly corresponds to the original image. Because of this, we do not have to do linearization for the earlier part of the network to attribute to pixels. We can easily find the pixel in the score map that contributes most to the final prediction. We can also do thresholding \wrt the label of choice, and then we obtain a foreground/background mask as an explanation.

\subsubsection{Original CAM Formulation}

Our training likelihood (or prediction) is
\[P(y \mid x) = \operatorname{softmax}\left(\sum_l W_{yl} \left(\frac{1}{HW} \sum_{hw} \bar{f}_{lhw}(x)\right)\right).\]
(We use NLL to train the model.) We obtain our explanation score map at test time \wrt label \(\hat{y}\) by using the formula
\[f_{\hat{y}hw} = \sum_l W_{\hat{y}l}\bar{f}_{lhw}(x).\]
That is, we weight each channel of our convolutional feature map \(\bar{f}\) \wrt the weights between channels \(l\) and class \(\hat{y}\).

The used shapes of the tensors in the above formulation are \(\bar{f}(x) \in \nR^{L \times H \times W} = \nR^{2048 \times 7 \times 7}\) for the ResNet-50 CAM uses\footnote{Note the low resolution. To overlay the score map on images, further upscaling is needed.} and \(W \in \nR^{C \times L} = \nR^{1000 \times 2048}\) where \(C = 1000\) is the number of classes (using ImageNet-1K).

\subsubsection{Simplified CAM Formulation}

We rewrite our training likelihood as
\begin{align*}
P(y \mid x) &= \operatorname{softmax}\left(\sum_l W_{yl} \left(\frac{1}{HW} \sum_{hw} \bar{f}_{lhw}(x)\right)\right)\\
&= \operatorname{softmax}\left(\frac{1}{HW} \sum_{hw} \underbrace{\sum_l W_{yl}\bar{f}_{lhw}(x)}_{f_{yhw}(x) :=}\right)\\
&= \operatorname{softmax}\left(\frac{1}{HW} \sum_{hw} f_{yhw}(x)\right)
\end{align*}
where \(f(x) \in \nR^{C \times H \times W} = \nR^{1000 \times 7 \times 7}\). After this, we trivially simplify our explanation algorithm by indexing into our last-layer feature map \(f\) that was already calculated in the forward propagation:
\[f_{\hat{y}hw} = \sum_l W_{\hat{y}l}\bar{f}_{lhw}(x).\]

We do not have to do an additional matrix multiplication to generate the score map, i.e., we do not have to do the linear computation twice. We calculate it once for the forward propagation and then reuse the intermediate result for the class of interest by taking the last-layer feature map with channel index \(=\) class of interest. This gives us the score map directly. In the original formulation, we first perform GAP, and then use the FC layer during forward propagation. In the new formulation, we have to modify our original model a bit: We exchange the GAP and FC layers and turn the FC layer into a \(1 \times 1\) convolutional layer.

The FC operation is identical to the \(1 \times 1\) convolution operation, except that FC operates on non-spatial 1-dimensional features, but \(1 \times 1\) convolution operates on spatial 3-dimensional features. We apply the same matrix multiplication for every ``pixel'' (\(\in \nR^L\)) in the spatial dimensions of the feature map. Shape of weights: \(\nR^{1000 \times 2048 \times 1 \times 1}\). This ResNet~\cite{https://doi.org/10.48550/arxiv.1512.03385} variant is fully convolutional. These have been used extensively in the era of CNNs for semantic segmentation. In this case, we are training for a pixel-wise prediction of the class; thus, we need a tensor output. Usually, the exact spatial dimensionality is not retained; we have an hourglass architecture and upscaling at the end~\cite{https://doi.org/10.48550/arxiv.1505.04597}. Mask R-CNN~\cite{https://doi.org/10.48550/arxiv.1703.06870} also predicts a binary \emph{instance} mask for each detection, and it also has to upscale to the original window size.\footnote{Nowadays, many people are using Transformer-based~\cite{https://doi.org/10.48550/arxiv.1706.03762} baselines for doing semantic segmentation. The convolutional baselines are a bit old-fashioned but are still widely used.}

\begin{figure}
    \centering
    \includegraphics[width=\linewidth]{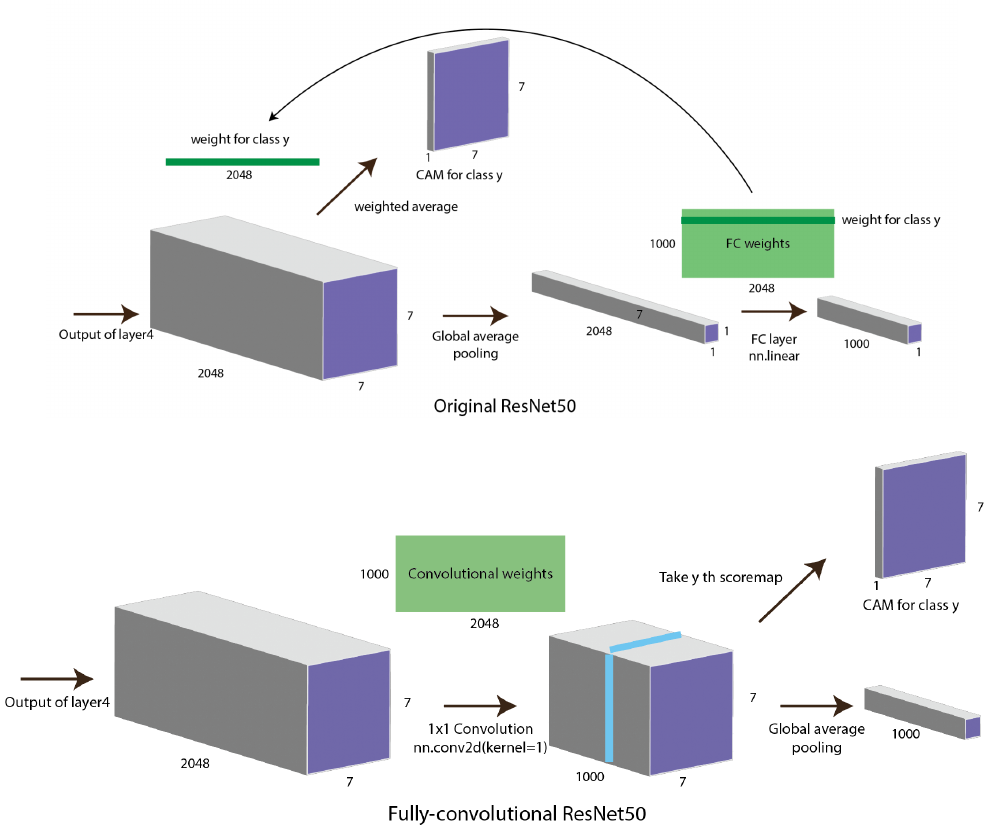}
    \caption{Comparison of the two formulations of the CAM method. Extracting the score map corresponding to any of the classes becomes significantly easier when using the bottom formulation. The overhead of having to store the tensor of shape \(1000 \times 7 \times 7\) in memory is negligible.}
    \label{fig:cam2}
\end{figure}

We compare the implementation of both approaches. With the simplified formulation, extracting the score map becomes much more straightforward. The two approaches are visualized in Figure~\ref{fig:cam2}. Python code for the original CAM formulation using PyTorch is shown in Listing~\ref{lst:original}. Similarly, Python code for the simplified CAM formulation is shown in Listing~\ref{lst:new}.

\begin{booklst}[Naive approach of computing CAM. Obtaining the score map requires complicated indexing.]{lst:original}
class ResNet(nn.Module):
    def __init__(self, block, layers, num_classes):
        super().__init__()
        self.conv1 = DNN.Conv2d(
            3, 64, kernel_size=7, stride=2, padding=3, bias=False
        )
        self.bn1 = DNN.BatchNorm2d(64)
        self.relu = DNN.ReLU(inplace=True)
        self.maxpool = DNN.MaxPool2d(kernel_size=3, stride=2, padding=1)
        self.layer1 = self._make_layer(block, 64, layers[0])
        self.layer2 = self._make_layer(block, 128, layers[1], stride=2)
        self.layer3 = self._make_layer(block, 256, layers[2], stride=2)
        self.layer4 = self._make_layer(block, 512, layers[3], stride=2)

        self.avgpool = DNN.AdaptiveAvgPool2d((1, 1))
        self.fc = DNN.Linear(512 * block.expansion, num_classes)

    def forward(self, x):
        x = self.conv1(x)
        x = self.bn1(x)
        x = self.relu(x)
        x = self.maxpool(x)

        x = self.layer1(x)
        x = self.layer2(x)
        x = self.layer3(x)
        x = self.layer4(x)

        x = self.avgpool(x)
        x = torch.flatten(x, 1)
        x = self.fc(x)
        return x

    def compute_explanation(self, x, y):
        x = self.conv1(x)
        x = self.bn1(x)
        x = self.relu(x)
        x = self.maxpool(x)

        x = self.layer1(x)
        x = self.layer2(x)
        x = self.layer3(x)
        x = self.layer4(x)  # (1, 512 * block.expansion, w, h)

        weights = self.named_modules(
        )["fc"].weight.data[y, :].unsqueeze(0).unsqueeze(2).unsqueeze(3)
        # (1, 512 * block.expansion, 1, 1)

        return torch.nansum(weights * x, dim=1)  # (1, w, h)
\end{booklst}

\begin{booklst}[Simpler approach of computing CAM. Obtaining the score map becomes trivial.]{lst:new}
class ResNet(nn.Module):
    def __init__(self, block, layers, num_classes):
        super().__init__()
        self.conv1 = DNN.Conv2d(
            3, 64, kernel_size=7, stride=2, padding=3, bias=False
        )
        self.bn1 = DNN.BatchNorm2d(64)
        self.relu = DNN.ReLU(inplace=True)
        self.maxpool = DNN.MaxPool2d(kernel_size=3, stride=2, padding=1)
        self.layer1 = self._make_layer(block, 64, layers[0])
        self.layer2 = self._make_layer(block, 128, layers[1], stride=2)
        self.layer3 = self._make_layer(block, 256, layers[2], stride=2)
        self.layer4 = self._make_layer(block, 512, layers[3], stride=2)
        self.conv_last = DNN.Conv2d(
            512 * block.expansion, num_classes, kernel_size=1
        )
        self.avgpool = DNN.AdaptiveAvgPool2d((1, 1))

    def forward(self, x):
        x = self.conv1(x)
        x = self.bn1(x)
        x = self.relu(x)
        x = self.maxpool(x)

        x = self.layer1(x)
        x = self.layer2(x)
        x = self.layer3(x)
        x = self.layer4(x)

        x = self.conv_last(x)
        x = self.avgpool(x)
        return x

    def compute_explanation(self, x, y):
        x = self.conv1(x)
        x = self.bn1(x)
        x = self.relu(x)
        x = self.maxpool(x)

        x = self.layer1(x)
        x = self.layer2(x)
        x = self.layer3(x)
        x = self.layer4(x)  # (1, 512 * block.expansion, w, h)

        x = self.conv_last(x)  # (1, num_classes, w, h)
        return x[:, y]  # (1, w, h)
\end{booklst}

\subsection{Comparison of the two CAM implementations}

Let us consider the pros and cons of the simplified CAM implementation.
\begin{itemize}
    \item \textbf{Pro}: Simpler implementation (especially for CAM computation) without changing the model performance (confirmed through numerous experiments). 
    \item \textbf{Contra}: More memory usage (but negligible). In the original formulation, after we perform GAP, we are left with a 2048D vector. In the simplified formulation we first perform the \(1 \times 1\) convolution, which results in a tensor of shape \(\nR^{1000 \times 7 \times 7}\). We need to store more floating point values for backprop (and for the CAM computation), but this is negligible compared to the total memory usage of a deep net.
\end{itemize}

\subsection{Assumptions to Make CAM Work}

As we have seen, CAM assumes an architecture in which we only perform linear operations after computing the score map. There should be a linear mapping from the feature map to the final score. (This does not hold if we also consider the softmax activation, but that is generally considered an interpretable operation, and we usually attribute to the logits.) The GAP operation is just a linear sum of \(7 \times 7\) features channel-wise, which is very interpretable. (Final prediction can be split into predictions from each of the features of the feature map. For sums, people have an excellent intuition about what is contributing by how much.)

As discussed previously, another crucial assumption of CAM is that the feature map pixels contain information specific to the corresponding input pixels. We treat each ``pixel'' of the score map as the feature corresponding to the input pixels at the same spatial location. This was empirically found to be true for CNNs because of translational equivariance. We can trace every feature pixel in the score map back to the possible range of pixels in the input that influenced that feature pixel, called the \emph{receptive field}. This tends to be huge, but the corresponding pixels ``move'' with the feature pixels via translational equivariance.

Notably, the assumptions only work to some extent. We get \emph{coarse} attribution scores upscaled to fit the input shape. This upscaling (or overlaying) is not theoretically justified, but we still get pretty sound attributions, as measured by soundness evaluation techniques. We can find worst-case examples where (because the receptive field can be much larger than the region one upscaled attribution ``pixel'' covers) the attribution map does not attribute the pixel responsible for the prediction at all. However, these are pretty artificial examples.

A spatial location of the score map usually has a very large receptive field, especially for a deep architecture like the ResNet-50. When we make use of CAM, we simply upscale it to match the input dimensions. By doing so, the input pixels that correspond to the score map ``pixels'' according to CAM can be much fewer than the number of pixels in the receptive field of a particular score map ``pixel'', i.e., the number of pixels that actually influence this score map value. CAM might be localizing more than it should. There is no guarantee that there is a nice straight mapping between the feature map pixels and the raw pixels. Researchers might have gained such insight through semantic segmentation models using a fully convolutional architecture (e.g., DeepLab~\cite{https://doi.org/10.48550/arxiv.1606.00915}) where the pixel-wise prediction is directly generated from the feature map (after upsampling). However, such models are trained with pixel-wise supervision, and so they are explicitly instructed that each feature pixel should mostly encode the content of the input area around that feature pixel (not the entire receptive field). CAM is not trained with such a signal -- only the aggregation of the pixel-wise predictions is supervised -- so there is even less guarantee for the correspondence.
In fact, the CAM activation pattern tends to reflect the shape of the receptive field. There exist architectures with non-unimodal receptive fields; for example, DeepLab has hierarchical, checkerboard-like receptive field patterns due to the dilated convolutions and increased convolution strides. As a result, CAM applied to DeepLab shows checkerboard-like activation maps even if the object is at one location of the given image.

The upsampling also introduces various possible problems with interpretability. The upsampling method also influences the attributions (e.g., nearest vs. bilinear vs. bicubic) and detaches the explanation from the extracted feature map values. This is similar to how the normalization of the CAM attributions is detached from the intuitive feature map values and how they relate to the output of the network.

We will see that more sophisticated methods (e.g., CALM~\cite{kim2021keep} in Section~\ref{sssec:calm}) still suffer from the ``handwaviness'' of the upscaling.\footnote{In CALM, the intermediate feature map elements also do not have a one-to-one correspondence to the input pixels. As we will see in Section~\ref{sssec:calm}, however, CALM resolves one of the many problems CAM has (namely, the unintuitive normalization of the attribution map).}  In summary, the foundation of CAM is questionable, and the attribution maps should be taken with a grain of salt.

The validity of the assumption that the feature map pixels correspond to the respective input pixels is unclear for Transformer-based models, as they do not have translational equivariance. It is unclear if the token-wise features after all self-attention layers actually convey the same semantic content as the corresponding tokens at the beginning. Most likely, CAM would not work well for Visual Transformers (ViTs)~\cite{https://doi.org/10.48550/arxiv.2010.11929}.

\begin{information}{Attribution Methods for Transformers}
For Transformers, various attribution methods are discussed in the paper ``\href{https://arxiv.org/abs/2202.07304}{``XAI for Transformers: Better Explanations through Conservative Propagation}''~\cite{https://doi.org/10.48550/arxiv.2202.07304}. It discusses Generic Attention Explainability (GAE), input \(\times\) gradient methods, and several other methods for Transformers.
\end{information}

\subsection{Grad-CAM -- Generalizing CAM to non-linear \(h\)}
\label{sssec:gradcam}

\begin{figure}
    \centering
    \includegraphics[width=0.9\linewidth]{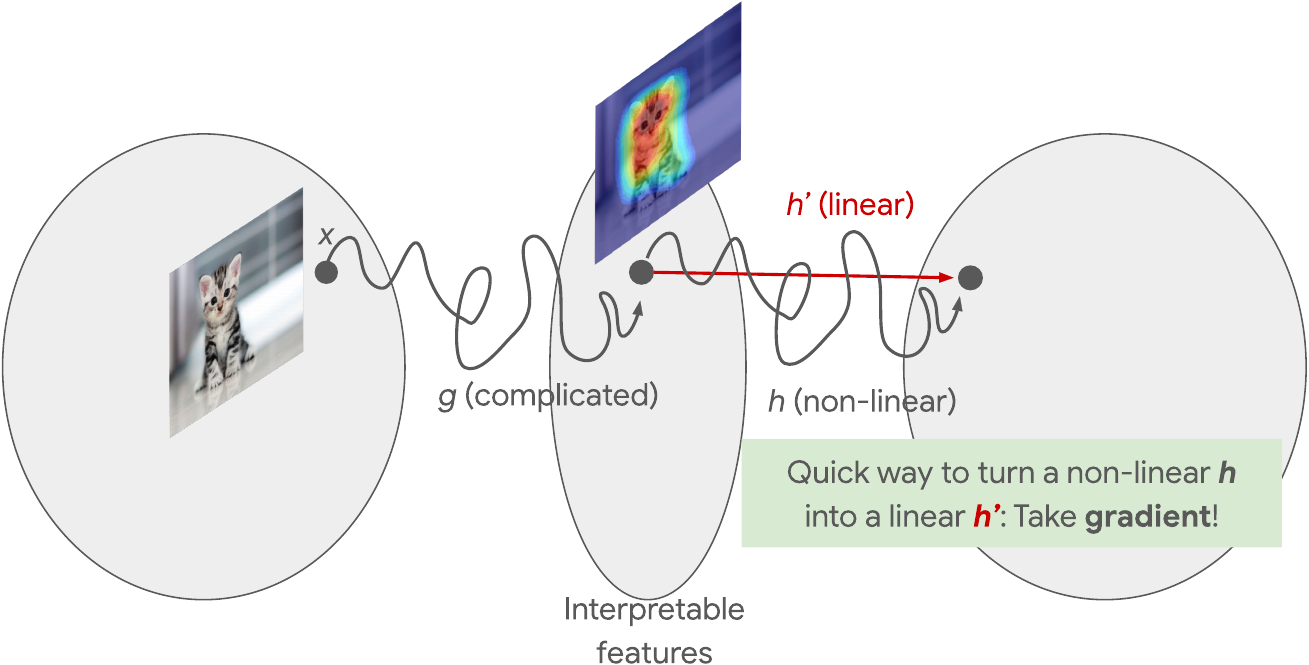}
    \caption{High-level motivation of the Grad-CAM method. In Grad-CAM, irrespective of the particular task-specific network we have on top of the convolutional feature representation, as long as it is differentiable, we can linearize it around the feature space point of interest \(g(x)\).}
    \label{fig:gradcam1}
\end{figure}

\begin{figure}
    \centering
    \includegraphics[width=\linewidth]{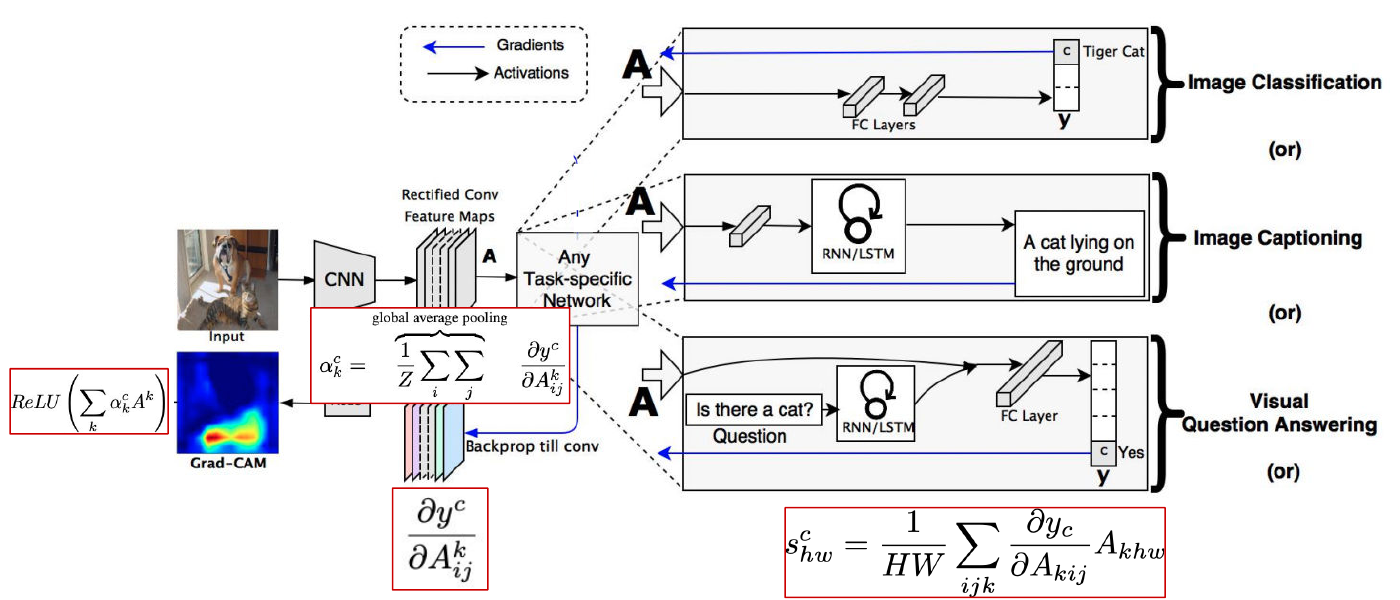}
    \caption{Detailed overview of the Grad-CAM method, a generalization of the CAM method that employs linearization. CNN denotes a fully convolutional network like the one we have seen in CAM. Until the \emph{Rectified Conv Feature Maps}, the architecture considered is the same as the one in CAM -- \(A \in \nR^{2048 \times 7 \times 7}\) is the 3D convolutional feature map we previously denoted by \(\bar{f}\). Note: \(A\) can have other shapes, too -- Grad-CAM is not restricted to fixed shapes, and neither is CAM. However, the later layers require some modifications to the CAM method. First, we consider image classification. In VGG, after the convolutional layers, the authors employ three more FC layers with activations. This results in a non-linear second part -- vanilla CAM no longer works. In 2017, LSTMs for NLP tasks (including image captioning) were the gold standards. (A lot has changed since, Transformers have taken over the field.) An LSTM also contains many non-linearities -- vanilla CAM does not work here, either. To solve this, we linearize the second part of the network architecture for each input \(x\). In particular, we calculate the 3D gradient map of the logit for class \(c\) \wrt the intermediate representation, \(\frac{\partial y^c}{\partial A} \in \nR^{2048 \times 7 \times 7}\) (meaning that \(\frac{\partial y^c}{\partial A^k_{ij}} \in \nR\)). Base Figure taken from~\cite{selvaraju2017grad}.}
    \label{fig:gradcam2}
\end{figure}

What happens if the part of our network between the feature map and the final scores is not linear? This assumption was one of the reasons why we could take an intermediate layer as a feature attribution map in CAM -- the remaining layers were linear and, therefore, intrinsically interpretable. Had they not been linear, we could have not applied CAM directly.

Conveniently, we can extend CAM using linearization techniques we have seen before. \href{https://arxiv.org/abs/1610.02391}{Grad-CAM}~\cite{selvaraju2017grad} is a follow-up method on CAM that extends it to non-linear parts between the feature map and the final scores. A high-level overview and description of the method is given in Figure~\ref{fig:gradcam1}. A detailed overview of Grad-CAM is shown in Figure~\ref{fig:gradcam2}.

\textbf{Remarks}: It is unclear why the authors apply ReLU on the weighted sum. CAM already performs max-normalization, i.e., they drop all negative values and normalize with the max value anyway. Using the notation of Figure~\ref{fig:gradcam1}, instead of having a linear \(h\) part, the authors linearize \(h\) locally around \(g(x)\) and then compute CAM \wrt this linearized network.

\textbf{Note}: People often refer to CAM as Grad-CAM in many cases because the latter is a generalization of the former. When Grad-CAM is mentioned in a paper, it could refer to CAM. It depends on the architecture it is being used on.

\begin{information}{Grad-CAM Generalizes CAM}
In CAM, we had
\[f_{\hat{y}hw} = \sum_l W_{\hat{y}l}\bar{f}_{lhw}(x).\]
Here, we have a \emph{generalization} of CAM (considering \(A = \bar{f}\)). Using the notation from Figure~\ref{fig:gradcam2},
\begin{align*}
s^c_{hw} &= \frac{1}{HW}\sum_{ijk}\frac{\partial y^c}{\partial A_{kij}}A_{khw}\\
&= \sum_k \underbrace{\frac{1}{HW}\sum_{ij}\frac{\partial y^c}{\partial A_{kij}}}_{\alpha^c_k}A_{khw}
\end{align*}
To see that this is indeed a generalization of CAM (with a twist), observe that when we have a linear second part, i.e., \[y^c = \frac{1}{HW}\sum_{hw}\sum_k W_{y^ck} A_{khw},\]
then
\begin{align*}
\frac{\partial y^c}{\partial A_{lij}} &= \frac{\partial}{\partial A_{lij}} \frac{1}{HW}\sum_{hw}\sum_k W_{y^ck} A_{khw}\\
&= \frac{1}{HW}\sum_{hw}\sum_k W_{y^ck} \frac{\partial}{\partial A_{lij}} A_{khw}\\
&= \frac{1}{HW}\sum_{hw}\sum_k W_{y^ck} \delta_{lk}\delta_{ih}\delta_{jw}\\
&= \frac{1}{HW} W_{y^cl},
\end{align*}
thus,
\[\alpha^c_k = \frac{1}{HW}\sum_{ij} \frac{1}{HW} W_{y^ck} = \frac{1}{HW} W_{y^ck},\]
which nearly gives us back the CAM formulation but has an additional scaling term \(\frac{1}{HW}\). This, however, does not matter for the final activation map because it is normalized.
This scaling factor just makes the computation a bit more stable by averaging. This shows that this method is a natural extension of CAM.
\end{information}

\subsection{Remaining Weakness of CAM}
\label{sssec:cam}

CAM is not as interpretable as we would want. While the function on top of the feature map is linear (GAP + \(1 \times 1\) convolution), that is not the end of the story. This is because when we compute CAM, we have \emph{an additional step of normalization} of the map to be in the image value range. The unnormalized score map is taken from the pre-softmax values; thus, we have no guarantee of normalization. We do not only need normalization to be in the image value range -- a fixed range for the score map is needed anyway, as otherwise, there would be no way to compare score maps consistently across many images. The model we train in CAM is
\[P(y \mid x) = \operatorname{softmax}\left(\frac{1}{HW}\sum_{hw}f_{yhw}(x)\right),\]
which is very interpretable. However, the final score map is calculated in two ways:
\[s = \begin{cases} \frac{\max(0, f^{\hat{y}})}{f^{\hat{y}}_\text{max}} & \text{ if max} \\ \frac{f^{\hat{y}} - f^{\hat{y}}_{\text{min}}}{f^{\hat{y}}_\text{max} - f^{\hat{y}}_{\text{min}}} & \text{ if min-max}\end{cases} \in [0, 1]^{H \times W}.\]
These non-linear transformations of our feature map are hard to interpret. In English, the max-version could be explained as 
\begin{center}
``The pixel-wise pre-GAP, pre-softmax feature value at \((h, w)\), measured in relative scale within the range of values \([0, A]\) where \(A\) is the maximum of the feature values in the entire image.''
\end{center}
It is clear that we could not explain this to an end user who has no knowledge of ML. They would not understand what is being shown in the score map, which is a necessary condition of the attributions to be deemed human-understandable.

A summary of problems with CAM as an attribution method is given below:
\begin{itemize}
    \item \emph{The test computational graph is not a part of the training graph.} In a sense, we are making up values later, at test time, for the score map.
    \item \emph{We only have an unintuitive description of the score map values in English.} It is difficult to explain the attribution values to clients. Another problem with the normalization method is that min-max or max normalization suffers from outliers without clipping. If one is not careful, whenever the pre-normalized scores contain outliers, the normalized score maps can become uninformative: when visualized, everything seems roughly equally important and the displayed map is not faithful anymore to the actual attribution scores.
    \item \emph{CAM also violates widely accepted ``axioms'' for attribution methods.} Details are given in the \href{https://arxiv.org/abs/2106.07861}{CALM} paper.
\end{itemize}

\subsection{Class Activation Latent Mapping (CALM)}
\label{sssec:calm}

To fix the problems introduced in Section~\ref{sssec:cam}, we discuss the paper ``\href{https://arxiv.org/abs/2106.07861}{Keep CALM and Improve Visual Feature Attribution}''~\cite{kim2021keep}. In CALM, we approach the problem with a fully probabilistic treatment of the last layers of CNNs.

\textbf{Notation}:
\begin{itemize}
    \item \(X\): Input image.
    \item \(Y\): Class label \(\in \{1, \dotsc, C\}\).
    \item \(Z\): Pixel index (location) \(\in \{1, \dotsc, M\}\) in the feature map. For example, possible values for \(Z\) are \(1, \dots, 49\) for a \(7 \times 7\) feature map. \(Z\) is a discrete random variable in the spatial feature map dimensions.
\end{itemize}
\textbf{Task}: ``Predict \(Y\) from \(X\) by looking at pixel \(Z\).'' Our prediction is based on the observations at feature location \(Z\). \(Z\) is a latent variable not observed during training. Only \(X\) and \(Y\) are observed; the training set is the same as always. In particular, we do not have GT values for \(Z\).

We use the following decomposition of the joint distribution:
\begin{align*}
P(x, y, z) &= P(y, z \mid x)P(x)\\
&= P(y \mid x, z)P(z \mid x)P(x).
\end{align*}

\begin{figure}
    \centering
    \tikz{
        \node[obs] (x) {$X$};
        \node[obs,right=of x] (y) {$Y$};
        \node[latent,above=of x,xshift=1cm] (z) {$Z$};
        \edge {x} {y} 
        \edge {x} {z} 
        \edge {z} {y} }
    \caption{Directed graphical model representation of the relationship of variables \(X, Y, Z\) in CALM.}
    \label{fig:calm1}
\end{figure}

\begin{figure}
    \centering
    \includegraphics[width=\linewidth]{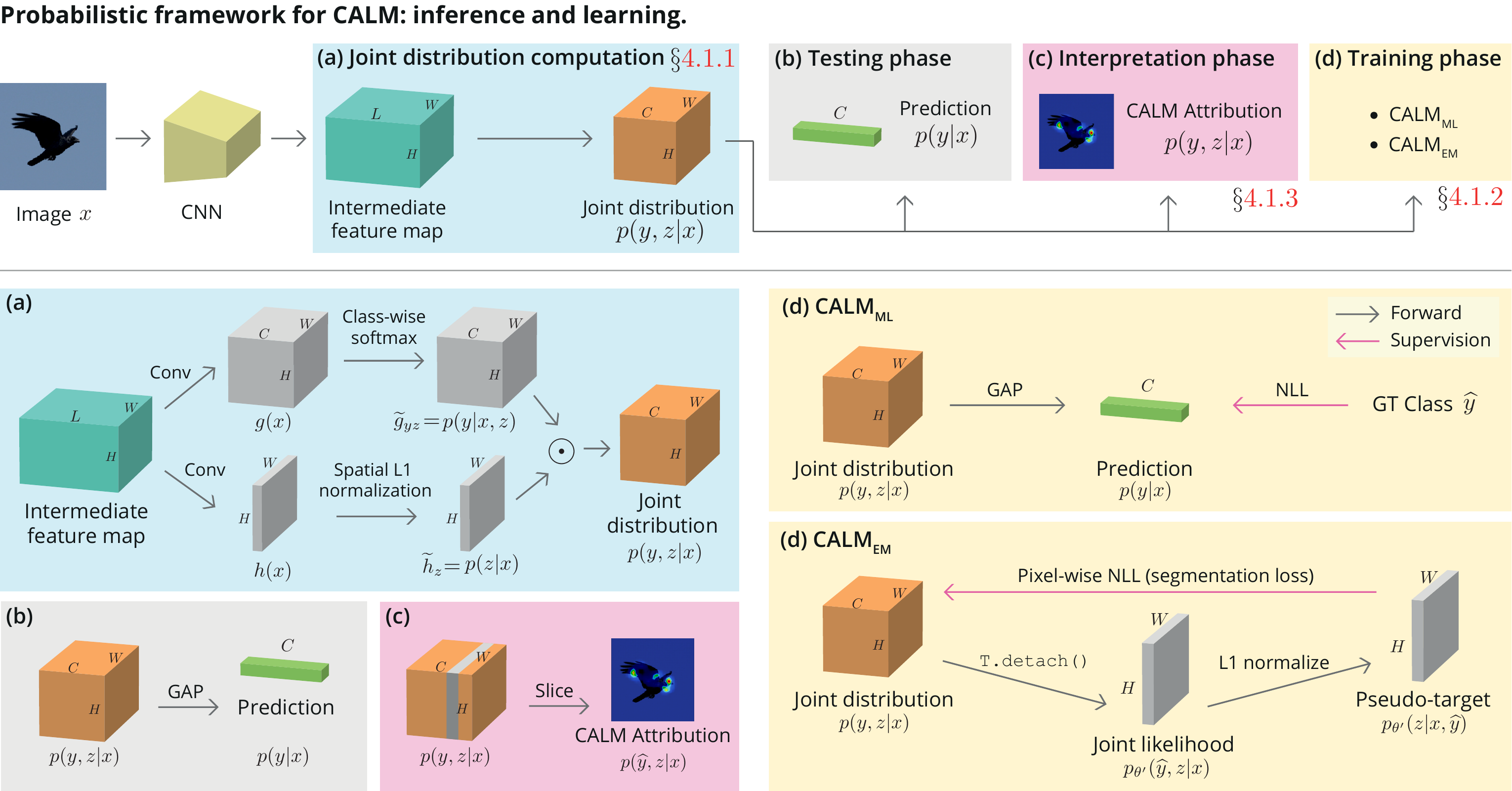}
    \caption{Detailed overview of CALM,  a fully probabilistic approach to feature attribution. The CNN used is an FCN, just like in CAM and Grad-CAM. Figure taken from~\cite{kim2021keep}.}
    \label{fig:calm2}
\end{figure}

This corresponds to the probabilistic graph (directed graphical model) illustrated in \ref{fig:calm1}. A detailed overview of CALM is provided in Figure~\ref{fig:calm2}. We discuss the individual parts of the model here.

\subsubsection{Part (a) of Figure~\ref{fig:calm2}}

We obtain the conditional joint distribution of \(Y\) and \(Z\) given \(X\). Both \(Y\) and \(Z\) are discrete random variables; thus, we can fully represent their joint distribution by a 3D tensor, where \(z\) is a 2D spatial index and \(y\) is a 1D index. Before spatial \(L_1\) normalization, we apply softplus. \(g(x)\) and \(h(x)\) are network predictions. The only requirement for the joint distribution is that the values are between \(0\) and \(1\), and they sum up to \(1\). This is enforced by the normalization before the element-wise multiplication. We could also just apply global softmax on the entire \(g(x)\) pre-activation tensor that normalizes \wrt both the class and spatial dimensions, but it did not perform well in the early experiments, according to the authors. Softmax and softplus + \(L_1\) norm are very similar: both are eventually \(L_1\) normalization, but softmax exponentiates before normalizing and softplus + \(L_1\) uses softplus before normalization. Exponentiation can sometimes be too harsh because it can blow up high values to infinity or push low values down to virtually 0. Softplus, on the other hand, is much better behaved -- the transformation is approximately linear on the positive side. For this reason, one should always consider using softplus + \(L_1\) norm when softmax blows up neural network training. It would be interesting to observe how turning softmax in Transformers into softplus + \(L_1\) norm influences the behavior of these networks.

\subsubsection{Part (b) of Figure~\ref{fig:calm2}}

We obtain the test-time prediction from the network. Similarly to CAM, we do global \emph{sum} pooling. We sum instead of averaging because the elements are probabilities, and this corresponds to marginalization.

\subsubsection{Part (c) of Figure~\ref{fig:calm2}}

We obtain the attribution map from the network for a particular input \(x\). \(\hat{y}\) is the ground truth label. For a particular location \(z\), the \emph{attribution score} \(s_z\) is
\[s_z := P(\hat{y}, z \mid x).\]
In English, the map is significantly simpler to explain than CAM:
\begin{center}
``The probability that the cue for recognition was at \(z\) and the ground truth class \(\hat{y}\) was correctly predicted for image \(x\).''
\end{center}

A nice property of this formulation is that the attribution map is well-calibrated: it lies between \(0\) and \(1\) and has a probabilistic interpretation. One can also normalize \wrt \(z\) and calculate the attribution map \wrt the predicted class to get a similar formulation as in CAM. We have a simpler way to compute a calibrated explanation score map.

\subsubsection{Part (d) of Figure~\ref{fig:calm2}}

\begin{information}{DeepLab}
\href{https://arxiv.org/abs/1606.00915}{DeepLab}~\cite{https://doi.org/10.48550/arxiv.1606.00915} is a semantic segmentation network from 2016. It was SotA on the PASCAL VOC-2012 semantic segmentation task at the time of its publication.
\end{information}

We consider two ways of training CALM: Marginal Likelihood (ML) and Expectation maximization (EM). These are typical methods to train a latent variable model. Let us discuss them in this order.

\textbf{Marginal likelihood.} This method directly minimizes the negative log-marginal likelihood. This is the usual way to train when obtaining \(P(y \mid x)\) is tractable. The NLL is simply the CE loss
\[-\log P(\hat{y} \mid x) = -\log \sum_z P(\hat{y} \mid x, z) P(z \mid x) = - \log \sum_z \tilde{g}_{\hat{y}z}\cdot \tilde{h}_z.\]

\textbf{Expectation maximization.} Segmentation methods using CNNs use it often. This is exactly how we train a DeepLab~\cite{https://doi.org/10.48550/arxiv.1606.00915} model using pixel-wise GT masks. We optimize for the joint tensor. \texttt{detach()} is needed to not have any gradient flow from the \emph{target}. We take the GT slice of the joint distribution. It is a likelihood because we apply our knowledge of what the true \(y\) is, and it is unnormalized in \(z\). \(L_1\) normalization means dividing by the sum of values in the matrix. The pseudo-target is what we want to reach with \(P(\hat{y}, z \mid x)\), as we want only the \(\hat{y}\) slice to have a positive probability. Then the joint becomes properly normalized in \(z\) when considering \(\hat{y}\). We have an entire prediction vector for every pixel in the joint. In our minds, we expand the pseudo-target into a one-hot vector (\(\hat{y}\) dimension is the pseudo-target, all other class dimensions are zeros). Then we apply a CE loss.

\subsubsection{CALM Addresses the Limitations of CAM}

CALM addresses the limitations of CAM detailed previously:
\begin{itemize}
    \item \emph{The test computational graph is a part of the training graph.} The training, test, and interpretation phases are all probabilistic.
    \item \emph{We have an intuitive description of the score map values.} 
    \item \emph{CALM respects all widely accepted ``axioms'' for attribution methods.} Exact details are discussed in the \href{https://arxiv.org/abs/2106.07861}{CALM} paper. Being probabilistic, CALM has many linear components.
\end{itemize}
While this method solves many problems with CAM, it still lacks reasoning about upscaling the score map instead of taking receptive fields into account more rigorously.

\subsubsection{Windfall features for CALM attributions}

\begin{figure}
    \centering
    \includegraphics[width=0.6\linewidth]{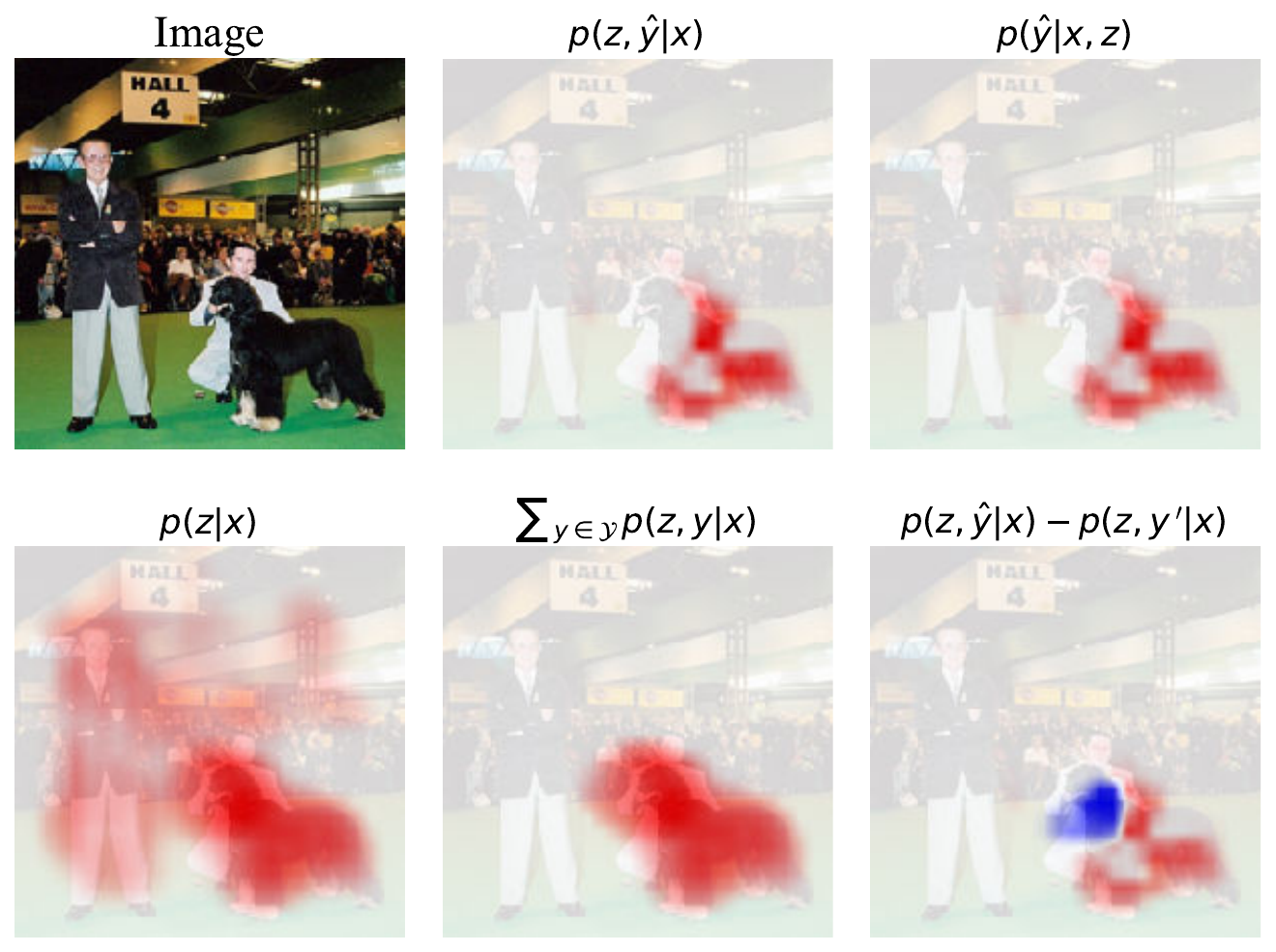}
    \caption{Examples of different windfall attributions we can obtain from the joint \(P(y, z \mid x)\) in CALM, taken from the paper~\cite{kim2021keep}.}
    \label{fig:calm3}
\end{figure}

CALM comes with numerous windfall gains.\footnote{Unexpected, large gains.} When the attribution map is well-calibrated and probabilistic, we can compute a lot of derivative\footnote{What we mean by ``derivative'' is not mathematical derivatives but computations that are derived from the probability tensors.} attributions on top of it, as illustrated in Figure~\ref{fig:calm3}. Score maps can be given, e.g., for
\begin{itemize}
    \item the GT class (first row, second image);
    \item the likelihood of the GT class (first row, third image -- the difference is the normalization factor);
    \item the predicted class (not shown);
    \item a generic class (not shown);
    \item all classes (second row, first image);
    \item multiple classes (second row, second image);
    \item and counterfactuals (second row, third image).
\end{itemize}
We discuss some of the options in detail below.

Marginalizing out all classes allows us to gain an overview. ``Where is any object that belongs to the 1k classes in ImageNet-1K?'' (Only a somewhat valid interpretation when the network is void of any spurious correlation, but even then, its prediction might only depend on small object parts that are very predictive.) ``What image regions does the network attribute to any of the classes?'' (Valid interpretation for any network.)

We can also sum scores for a subset of classes (e.g., dog, living thing, equipment, object, edible fruit, or food). Here, we sum up score maps for all dog classes (118 in ImageNet-1K). We get better-delineated boundaries for the dog meta-class.

Subtracting different score maps gives a counterfactual explanation of why we chose a class over another. The score map still makes sense; we just use different colors for the two classes.

\begin{figure}
    \centering
    \includegraphics[width=\linewidth]{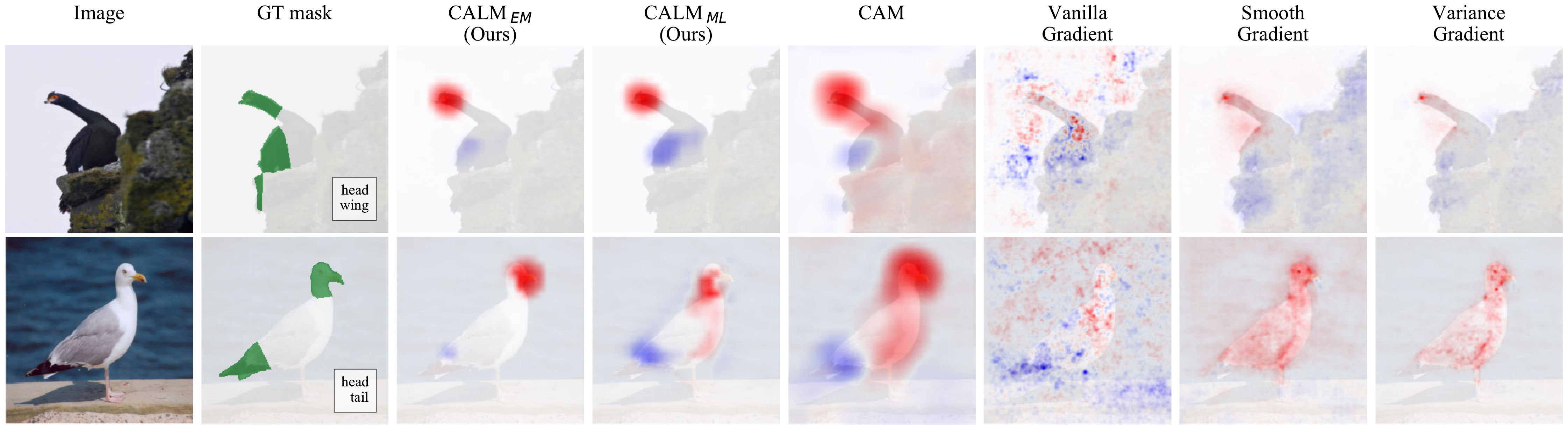}
    \caption{Qualitative comparison of CALM and other attribution methods against the GT CUB annotations, taken from the paper~\cite{kim2021keep}. In detail, the authors select the ground truth class and one that is easily confused with it (i.e., the differences appear only on a few body parts of the bird species). They want the model to give the same attributions to the body parts where the classes' (birds') attributes are the same.}
    \label{fig:calm4}
\end{figure}

Let us now turn to Figure~\ref{fig:calm4}. One can evaluate the quality of the attribution maps on the CUB dataset as follows.
\begin{center}
``We compare the counterfactual attributions from CALM and baseline methods against the GT attribution mask [on CUB]. The GT mask indicates the bird parts where the attributes for the class pair \((A, B)\) differ. The counterfactual attributions denote the difference between the maps for classes \(A\) and \(B\): \(s^A - s^B\). [...]''~\cite{kim2021keep}
\end{center}
The corresponding results are shown qualitatively in Figure~\ref{fig:calm4} and quantitatively in Table~\ref{tab:calm5}. One possible problem with the evaluation in Figure~\ref{fig:calm4} is that the attribution maps that are compared are \(P(z, \hat{y} \mid x)\) and \(P(z, \tilde{y} \mid x)\) (where \(\hat{y}\) and \(\tilde{y}\) are two similar classes), which are not normalized in \(z\). In particular, they do not even sum to the same value in \(z\), as the predicted probabilities \(P(\hat{y} \mid x\) and \(P(\tilde{y} \mid x)\) are never exactly equal for NN predictions. The paper mentions that if a pixel for both classes is equally important, the difference ideally cancels out so the counterfactual attribution map ideally focuses on pixels that affect the two classes differently. But because of the two maps not being on the same scale, even if proportionally some pixel has the same relative importance, the values are not going to cancel. Thus, the plot very likely shows that the individual attributions are already \emph{only} focusing on parts that are discriminative between the two classes. This would mean that the given reasoning for the feature maps is slightly incorrect. Without knowing the individual attribution maps, the difference is also not very descriptive. For example, for the ground truth class in the above row, the bird's head seems to be a very distinctive factor for the prediction. However, the wing might also be a factor that the network takes into consideration \emph{for} the GT class, but it is certainly taken with a higher attribution value for the alternative class because it is pale blue. Still, we do not know the exact attributions. 

CALM gives counterfactual score maps that often coincide with the GT masks on the CUB task. CALM with EM beats CAM on the CUB benchmark, as shown in Table~\ref{tab:calm5}. Both CALM variants beat other attribution methods. (This is not an evaluation of soundness. The model has all the rights to look elsewhere, e.g., because it suffers from spurious correlations.)

\begin{figure}
    \centering
    \includegraphics[width=0.8\linewidth]{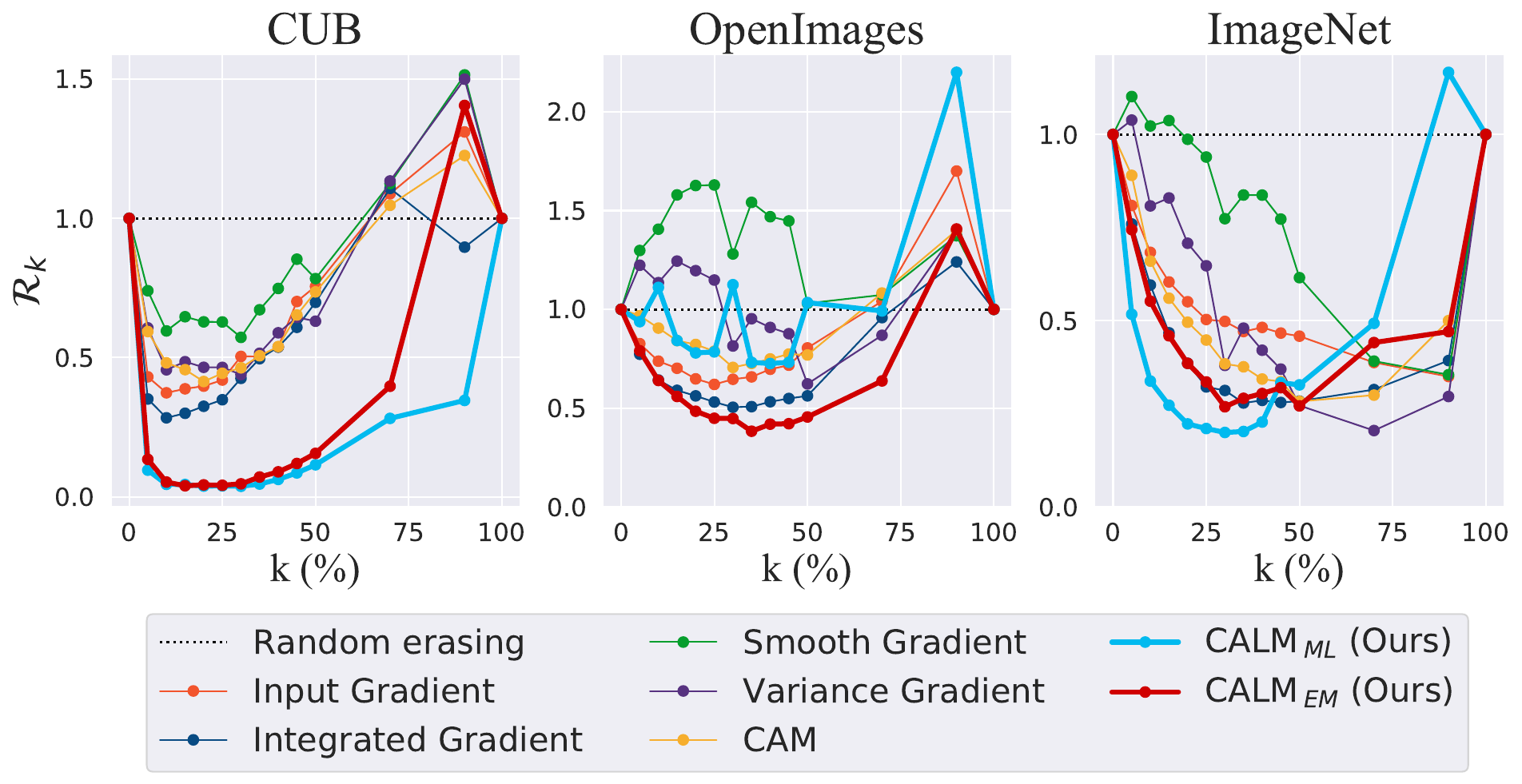}
    \caption{``\textbf{Remove-and-classify results.} Classification accuracies of CNNs when $k$\% of pixels are erased according to the attribution values $s_{hw}$. We show the relative accuracies $\mathcal{R}_k$ against the random-erasing baseline. Lower is better.''~\cite{kim2021keep} CALM performs well on the remove-and-classify benchmark. Figure taken from~\cite{kim2021keep}.}
    \label{fig:rac}
\end{figure}

The authors evaluate the soundness of their method using remove-and-classify (Figure~\ref{fig:rac}; discussed in Section~\ref{sssec:rac}). CALM performs best and seems to be the most sound.~\cite{kim2021keep}

\begin{table}
    \setlength{\tabcolsep}{.4em}
    \centering
    \caption{Quantitative comparison of CALM and other attribution methods against the GT CUB annotations, taken from the paper~\cite{kim2021keep}.}
    \label{tab:calm5}
    \small
    \begin{tabular}{ccccccc}
    \#\ignorespaces part differences && 1             & 2             & 3             &&  \\ \cmidrule{1-1} \cmidrule{3-5}
    \#\ignorespaces class pairs      && 31            & 64            & 96            &&     mean                  \\
    \cline{1-1} \cline{3-5} \cline{7-7}
    \vspace{-1em} & \\
    \cline{1-1} \cline{3-5} \cline{7-7}
    \vspace{-1em} & \\
    Vanilla Gradient           && 10.0          & 13.7          & 15.3          && 13.9                  \\
    Integrated Gradient     && 12.0          & 15.1          & 17.3	        && 15.7                  \\
    Smooth Gradient          && 11.8          & 15.5          & 18.6          && 16.5                  \\
    Variance Gradient            && 16.7          & 21.1          & 23.1	        && 21.4                  \\
    \cline{1-1} \cline{3-5} \cline{7-7}
    \vspace{-1em} & \\
    CAM                   && 24.1          & 28.3          & 32.2          && 29.6                  \\
    \(\text{CALM}_\text{ML}\)           && 23.6        &  26.7         & 28.8     && 27.3 \\ 
    \(\text{CALM}_\text{EM}\)           && \textbf{30.4} & \textbf{33.3} & \textbf{36.3} && \textbf{34.3} \\ 
    \cline{1-1} \cline{3-5} \cline{7-7}    
    \end{tabular}
\end{table}

\subsubsection{Cost to Pay in CALM}

CALM is clearly a better explainability method than CAM but is not necessarily a better classifier. CALM is changing the network structure, so it is very different from the reformulation of CAM. There, we had an equivalent formulation: The ResNet-50 architecture is fully compatible with CAM. Here, we do not have such an equivalent formulation. Now if we change the original network structure to the CALM formulation, we are changing the mathematical structure of the model. We cannot expect the same accuracy.

As shown in Table~\ref{tab:calm6}, CALM ML sometimes gains a few points of accuracy and sometimes loses a few against CAM. CALM EM sometimes becomes much worse than CAM in accuracy, sometimes stays close to CAM, and is only behind by a few percentage points. One has to be careful about the possible accuracy loss with CALM.

There is an inborn trade-off between interpretability and accuracy. The existence of this trade-off is very curious: it also means that depending on our actual needs, we might need to choose a different model. For example, losing 4\% accuracy might not be as important as gaining confidence and a better picture of how our model works. For such applications, we probably need CALM-trained models. Why must there be a trade-off between accuracy and the model's ability to explain itself? Because we limit ourselves to a smaller fraction of models if we are confined to interpretable models.

There are diverse requirements for deployment. We need to develop more diverse types of models. We should not only aim for models that perform well on a validation set but also develop slightly suboptimal models that are, e.g., interpretable or generalize very well to unseen situations. As an attribution method, there is room for improvement for CAM. CALM improves upon CAM regarding explainability. The better interpretability of CALM also contributes to better Weakly Supervised Object Localization (WSOL)~\cite{kim2021keep}, even though WSOL is not precisely aligned with explainability. Better interpretability, however, comes with a cost to pay (accuracy).

The human-interpretability of an XAI method does \emph{not} mean that we wish to make the model recognize things as humans do (human alignment). Instead, we wish to present the behavior of the model in a form that is \emph{understandable} by humans. No model will ``start thinking like humans'' by using human-interpretable XAI methods. There is no human alignment involved in the above reasoning. These two are also orthogonal axes of variation: the model might make decisions just like humans do, but the XAI method might fail to capture this. Vice versa, the XAI method might \emph{show} that the model makes decisions in a very human-aligned way, but it might just be because of the poor soundness of the method. However, there \emph{are} also occasions where we want better human alignment even at the cost of some loss in accuracy -- e.g. when the model is helping experts. This is a different trade-off, namely, the alignment-accuracy trade-off.

\setlength{\columnsep}{.3em}%
\begin{table}
\setlength{\tabcolsep}{.15em}
    \centering
    \caption{Classification accuracies of the Baseline (ResNet-50), CALM ML and CALM EM, taken from~\cite{kim2021keep}. Both formulations of CALM result in decreased accuracy in most situations. These can also be quite severe: The accuracy of \(\text{CALM}_\text{ML}\) is more than 10\% less than that of the baseline on CUB. However, there are also some situations where CALM can increase accuracy: On CUB, \(\text{CALM}_\text{EM}\) improves upon the baseline.}
    \label{tab:calm6}
    \begin{tabular}{lcccc}
    Methods && CUB & Open & ImNet \\
	\cline{1-1}\cline{3-5}\vspace{-1em}\\
	\cline{1-1}\cline{3-5}\vspace{-1em} \\
	Baseline &  & 70.6 & 72.1 & 74.5  \\
    \(\text{CALM}_\text{EM}\) &  & 71.8 & 70.1 & 70.4  \\
    \(\text{CALM}_\text{ML}\) &  & 59.6 & 70.9 & 70.6  \\
    \cline{1-1}\cline{3-5}\vspace{-1em}\\
    \end{tabular}
\end{table}

\begin{information}{Layer Norm}
\href{https://arxiv.org/abs/1607.06450}{Layer Norm}~\cite{https://doi.org/10.48550/arxiv.1607.06450} is a normalization technique that normalizes \wrt the mean and variance calculated across the feature dimension, independently for each element in the batch. On the other hand, Batch Norm calculates the mean and variance statistics across all elements in the batch. Layer Norm is widely used in Transformer-based~\cite{https://doi.org/10.48550/arxiv.1706.03762} architectures.
\end{information}

\begin{information}{Modified Backpropagation Variants}
This book does not mention some attribution methods: the class of modified backpropagation variants. There are many such methods:
\begin{itemize}
    \item LRP -- layer-wise relevance propagation~\cite{https://doi.org/10.48550/arxiv.1604.00825}
    \item DeepLIFT~\cite{https://doi.org/10.48550/arxiv.1704.02685}
    \item DeepSHAP
    \item GuidedBP~\cite{https://doi.org/10.48550/arxiv.1412.6806}
    \item ExcitationBP~\cite{https://doi.org/10.48550/arxiv.1608.00507}
    \item xxBP
    \item \dots
\end{itemize}
They are all based on some form of modification to backpropagation. They modify the gradients and do the backpropagation with some broken gradients. Eventually, we get the attribution in some intermediate feature layer or the input space. Vanilla backprop propagates gradients all the way back to the weights. However, gradients are (1) very local and (2) sometimes the function value is more important than the local variations. For example, when a function is constant, it is still contributing something to the next layer.

\medskip

We do not deal with them in this book for the following reasons:
\begin{itemize}
    \item It seems complicated to explain what the explanation shows.
    \item For new types of DNN layers, one needs to develop a new recipe for modified backpropagation. For example, for Transformers, we sometimes need to skip the layer norms with a straight-through estimator (seen in obfuscated gradients). There is no good intuition yet for how to modify backpropagation correctly across layer norms.
    \item Results depend on the implementation of the DNN. The attributions we obtain can differ for mathematically equivalent networks with different implementations. For example, we can consider two linear layers without non-linearity in between. For fixed, already trained weights, we (1) multiply them together beforehand, use the resulting matrix in a single linear layer, or (2) keep them separate for modified backprop. The results will differ because these methods modify backprop, and the separate modules are different in the two cases. This is a severe issue. We do not have the uniqueness of our attribution score.
    \item Caveat: They still show good soundness results, especially for Transformer architectures. We should have more understanding of why or how they work.
\end{itemize}
\end{information}

\subsection{Summary of Test Input Attribution Methods}

Linear models provide nice contrastive explanations. Therefore, we explored ways to linearize complex models (DNNs).

Local linearization around input \(x\) for the entire function \(f\) is employed by, e.g., input gradients, SmoothGrad, Integrated Gradients, LIME, and SHAP. The caveat is that it is hard to choose the right way to encode ``no information''. If we perform global linearization for an entire function, we just obtain a linear function that is interpretable by design but not globally sound.

We also discussed the diversity of features for contrastive explanations. One may use pixels, superpixels, instance segments, concepts (that are high-level, i.e., they cannot be represented as an aggregation of pixels), or feature-map pixels (that are aggregations of receptive field pixels with highly non-linear transformations).

We have no guarantee for Transformers that they see the same influence from the corresponding location of tokens/image patches. We cannot expect CAM to work. (Strictly speaking, we do not even have guarantees for ResNets because the receptive field does not coincide with upscaling the feature map.)

Attribution methods come with various pros and cons (depending on the method), and none of them is perfect.

\section{Explanations Linearize Models in Some Way}
\label{ssec:linearization}

As we have seen, the attributions are often based on some form of linearization of the original complex function (the DNN). This is because sparse linear models are already intuitive for humans. Let us give an overview of previously introduced methods and discuss what linearizations they employ.

\subsection{Input Gradient}

Taylor's theorem tells us that for a differentiable function \(f: \nR^d \rightarrow \nR\),
\[f(y) = f(x) + \langle y - x, \nabla_x f(x) \rangle + o(y - x);\]
thus, for very small perturbations around the input \(x\), our function is approximately linear. Taking the first-order Taylor approximation means finding the tangent plane of \(f\) at input \(x\).

\textbf{Note}: Only the input gradient linearizes the entire model \wrt the input \(x\) out of the methods in this overview. In the subsequent cases, we will observe linearization in either the \emph{attributions}, the \emph{discretized versions} of the input, or linearization of parts of the network. The only commonality is that the models are analyzed with some form of a linear model, but not necessarily a linear model \emph{\wrt \(x\)}. The point we make here is that, despite clever ways to formulate the attributions, none of the discussed methods could eventually avoid borrowing the immediate intuitiveness and interpretability of linear models. It would be an interesting research objective to try to formalize this intuition and show that any reasonable XAI method is inherently linear. (For an example of a weaker result, any method that satisfies the completeness axiom is inherently linearizing the predictions in the attributions.)

\subsection{Integrated Gradients}

This method also turns our model into a linear model for an input \(x\) around the baseline \(x^0\). Linearity is not \wrt the input \(x\) rather \wrt the attributions:
\[f(x) = f(x^0) + \sum_i a_i(f, x, x^0)\]
where
\[a_i(f, x, x^0) = (x_i - x^0_i)\left\langle e_i,  \int_{0}^1 \nabla_x f(x^0 + \alpha(x - x^0))d\alpha\right\rangle.\]
The prediction for the input \(x\) equals the prediction for the baseline image \(x^0\) plus the sum of the contribution of each pixel.

\subsection{LIME}

LIME makes an obvious, explicit linearization by approximating our possibly highly non-linear model \(f: \nR^d \rightarrow \nR\) by a sparse linear model locally:
\[g(z') = w_g^\top z'.\]

\subsection{SHAP}

SHAP also turns our model into a linear model for a binary input \(x\) around the baseline \(0\) \wrt the attribution values:
\[f(x) = f(0) + \sum_i \phi_{f, x}(i).\]
The prediction for the input \(x\) equals the prediction for the baseline plus the sum of the contribution of each feature (e.g., superpixel). By turning on/off our features, we regulate whether we include a contribution in the final prediction, so in this sense, this is a linear approximation of our model.

\subsection{TCAV}

\begin{figure}
    \centering
    \includegraphics[width=0.7\linewidth]{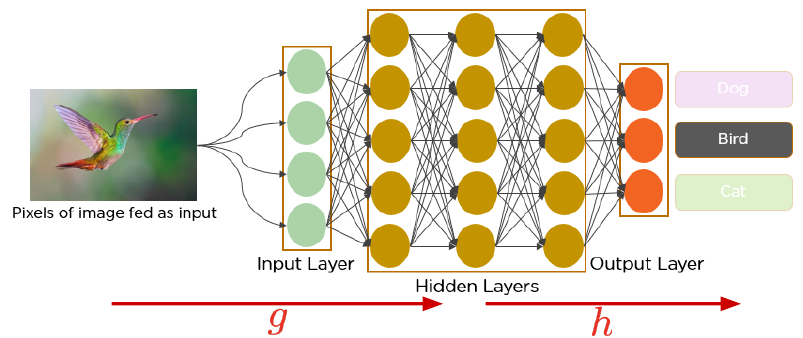}
    \caption{High-level overview of TCAV. We linearize the second part, \(h(z)\), \wrt the intermediate representation, \(z = g(x)\), making it a \emph{local} (linear) approximation of the second part of the network.}
    \label{fig:tcavlinear}
\end{figure}

In TCAV, we have
\[f(x) = h(z) = h(g(x)),\]
which is illustrated in Figure~\ref{fig:tcavlinear}.

We take the gradient of the output \wrt the intermediate layer
\[\nabla_z h(z) \big|_{z = g(x)},\]
thereby linearizing the second part of our network locally, around \(g(x)\).

\subsection{Different Types of Linearization}

We enumerated methods using linearization for explanations. Let us now see what categories of linearization we can establish.

\subsubsection{(1) Locally linear around \(x\), completely linear for \(f\).}

\begin{figure}
    \centering
    \includegraphics[width=0.7\linewidth]{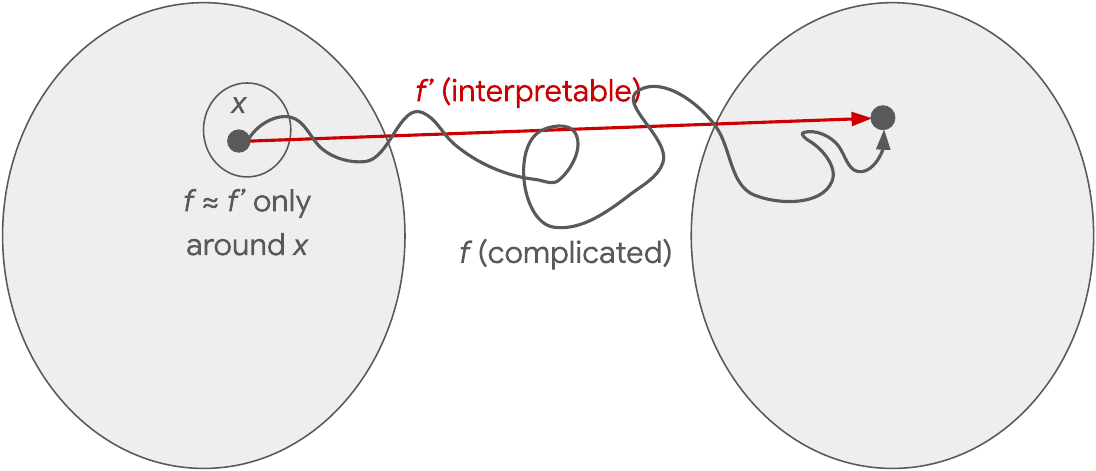}
    \caption{Local linearization around \(x\) and global linearization in \(f\). This approach is used in the Input Gradient, LIME, Integrated Gradients, and SHAP methods.}
    \label{fig:linearization1}
\end{figure}

This scenario is illustrated in Figure~\ref{fig:linearization1}. Examples include Input Gradient, LIME, Integrated Gradients, and SHAP (\emph{even though some of them employ global perturbations}). Here, the entire model is linearized, but only locally. It is quite simple to achieve: Around \(x\), we can explain everything nicely and interpretably.

\subsubsection{(2) Globally linear over \(g\)-space, partially linear for \(f\).}

\begin{figure}
    \centering
    \includegraphics[width=0.9\linewidth]{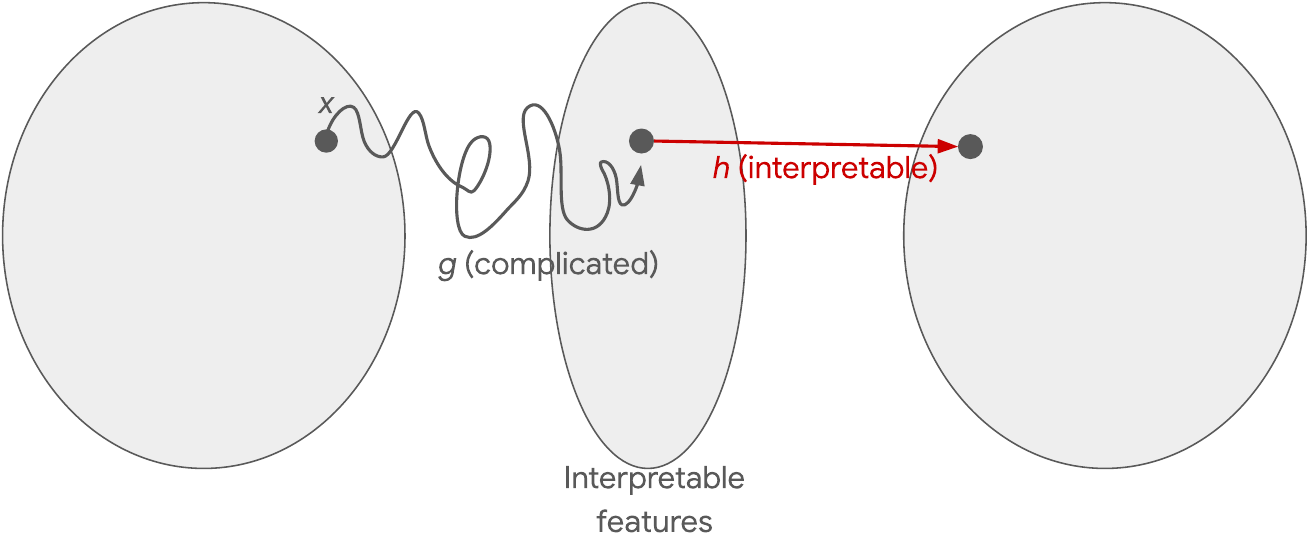}
    \caption{Global linearization over \(g\)-space, partial linearization for \(f\). Examples of methods employing this strategy include CAM and CALM.}
    \label{fig:linearization2}
\end{figure}

This scenario is shown in Figure~\ref{fig:linearization2}. Examples include CAM and CALM. CAM's/CALM's second part is already linear, therefore, there is no need to linearize it. Instead of explaining everything in terms of the input features, we are getting help from the interpretable intermediate features. The second part of the network is naturally interpretable. Therefore, we explain in terms of interpretable features without approximations, under some assumptions.

\subsubsection{(3) Locally linear around \(g(x)\), partially linear for \(f\).}

\begin{figure}
    \centering
    \includegraphics[width=0.9\linewidth]{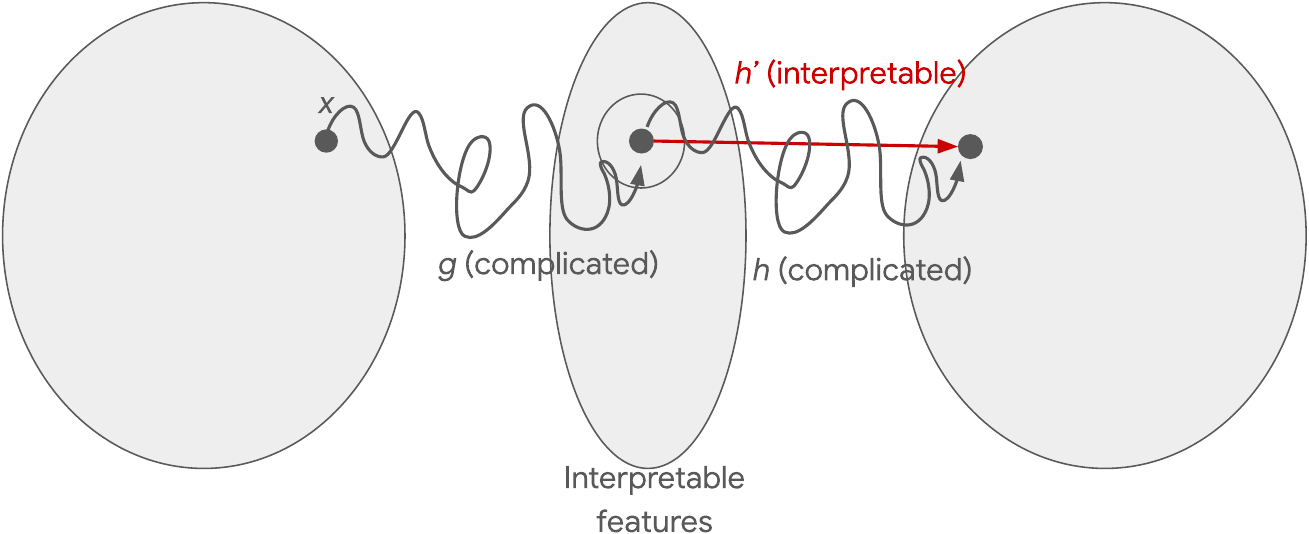}
    \caption{Local linearization around \(g(x)\) and partial linearization for \(f\). TCAV and Grad-CAM follow this approach.}
    \label{fig:linearization3}
\end{figure}

This category is illustrated in Figure~\ref{fig:linearization3}. Examples include TCAV (\(S_{C, k, l}\)) and Grad-CAM. TCAV takes all \(x\) into account that correspond to some label in the TCAV score (quite global explanations), but it linearizes the second part of the network (partial linearization) for each \(x\) separately. Whether we perform partial or total \emph{linearization} does \emph{not} depend on whether the method gives local or global \emph{explanations}. In Grad-CAM, we also only linearize the second part of the network (for a single input \(x\)).

Instead of explaining everything in terms of the input features, we are getting help from the interpretable features. We are only approximating the second part of our network through gradients. Therefore, we explain in terms of interpretable features but \emph{with approximations}.

\section{Evaluation of Explainability Methods}

First, let us discuss why we even need empirical evaluation.

\subsection{Why do we need empirical evaluation?}

The fundamental limitation of explainability is the soundness-explainability trade-off: Our explanation cannot be fully sound and fully explainable. We consider two extremes. The original model is too complex for humans to understand. This is the reason why we needed a separate explanation in the first place. We need simplifications to make humans understand. A global linear approximation makes our model interpretable again, but the model is not the same as before. The soundness of the explanations suffers a lot.

If we look at different XAI methods, they are in the trade-off frontier between soundness and explainability. One cannot say that some explanation method is conceptually perfect by design. Eventually, what matters is \emph{whether the method is serving our need} (the end goal). For that, we need empirical evaluation. We need ways to quantify different aspects of explanations in numbers.

\subsection{Types of Empirical Evaluation}

\begin{figure}
    \centering
    \includegraphics[width=0.5\linewidth]{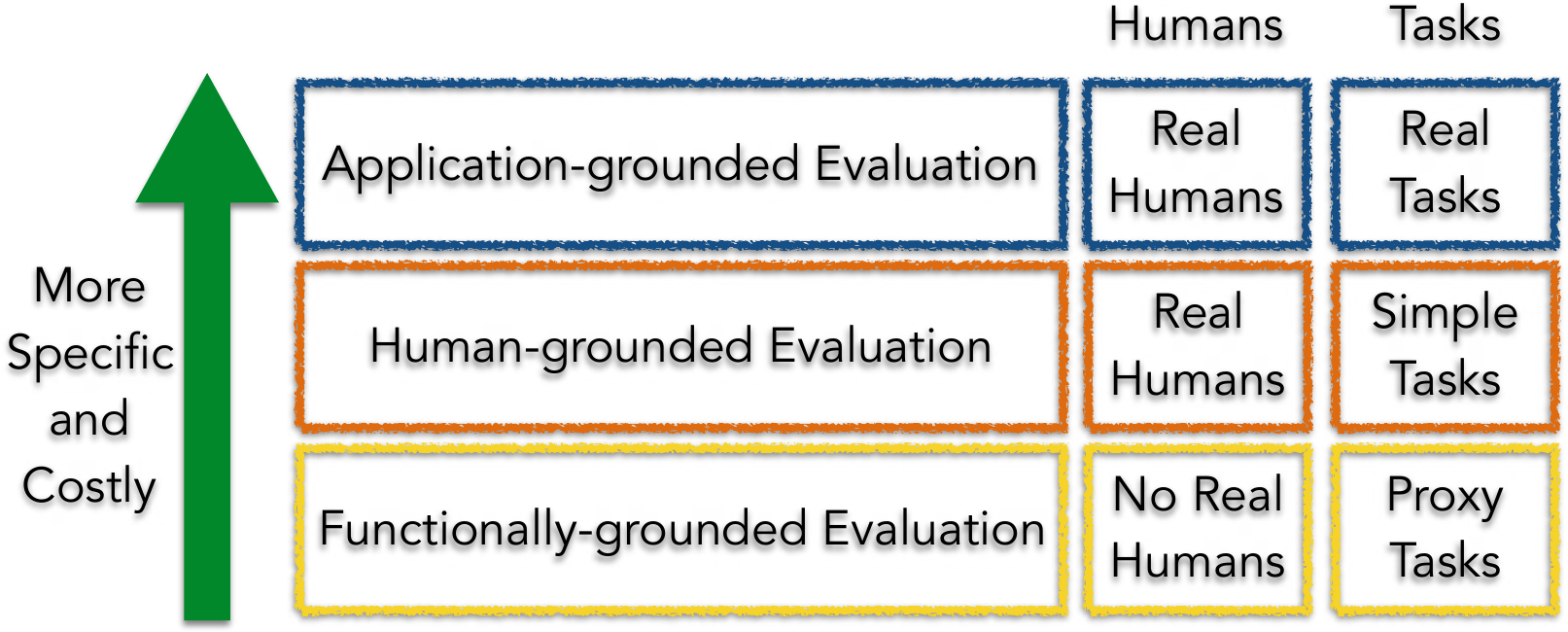}
    \caption{Comparison of three types of evaluation methods. As we go from bottom to top, the methods become more aligned with human needs but also become more expensive to carry out. Figure taken from~\cite{https://doi.org/10.48550/arxiv.1702.08608}.}
    \label{fig:doshivelez}
\end{figure}

Doshi-Velez and Kim~\cite{https://doi.org/10.48550/arxiv.1702.08608} distinguish three types of empirical evaluation: \emph{functionally-grounded}, \emph{human-grounded}, and \emph{application-grounded} evaluation (Figure~\ref{fig:doshivelez}). Let us briefly discuss each of these standard evaluation practices below.

\subsubsection{Functionally-Grounded Evaluation}

Functionally-grounded evaluation uses \emph{proxy tasks} to evaluate explanations. Here, no human subjects are required for the evaluation, making this type of evaluation appealing from a time and cost point of view. However, as explainability is necessarily human-grounded, such evaluations should only be considered in addition to human-grounded studies. \textbf{Example}: One linear model might be more sparse than another, signaling better human-interpretability~\cite{https://doi.org/10.48550/arxiv.1702.08608}. Sparsity can be evaluated without the involvement of humans.

\subsubsection{Human-Grounded Evaluation}

Human-grounded evaluation considers human subjects but conducts \emph{simple} experiments. This is desired when one wants to evaluate general aspects of the explanation that do not require domain expertise. No specific end goal is considered in such evaluation tasks, but they can still be used to judge general characteristics of explanations. \textbf{Example}: Human subjects are presented with explanation pairs and are asked to choose the `better' one~\cite{https://doi.org/10.48550/arxiv.1702.08608}.

\subsubsection{Application-Grounded Evaluation}

In application-grounded evaluation, it is measured how well an explanation method helps \emph{humans} when considering \emph{real} applications/problems. The helpfulness of an explanation method can be quantified by how much it increases human performance on a certain real task. This is the evaluation type that is most aligned with the \emph{human aspect} of explanations, but it is also the most expensive to carry out. \textbf{Example}: A computer programmer is evaluated based on how well they can fix their code after being given an explanation.

\subsection{Soundness Evaluation Techniques}
\label{sssec:eval}

As discussed in Section~\ref{ssec:good}, soundness (also referred to as faithfulness or correctness of our explanation) is arguably one of the most important and possibly the most widely used criterion. A sound explanation must identify the true cause(s) for an event. Currently, this seems to be the primary focus of XAI evaluation, but it is also \emph{not} the only criterion for a good explanation. This is crucial to keep in mind.

\begin{definition}{Confirmation Bias}
Confirmation bias is confirming the performance of our explanation method against what humans think would be the proper attribution instead of investigating further whether the model was actually basing its prediction on these causes.
\end{definition}

For measuring soundness, much previous research relied on qualitative evaluation of (potentially cherry-picked) examples. Consider the integrated gradients paper referring to the attribution maps shown in Figure~\ref{fig:integrated}:
\begin{center}
``Notice that integrated gradients are better [than input gradients] at reflecting distinctive features of the input image [for the prediction].''~\cite{sundararajan2017axiomatic}
\end{center}

\begin{figure}
    \centering
    \includegraphics[width=0.9\linewidth]{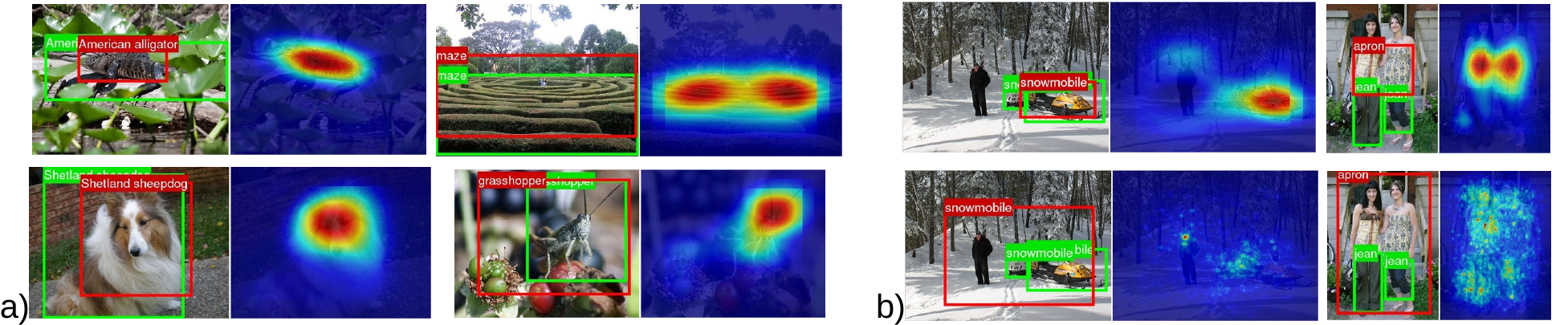}
    \caption{``a) Examples of localization from GoogleNet-GAP. b) Comparison of the localization from GooleNet-GAP (upper two) and the backpropagation using AlexNet (lower two). The ground-truth boxes are in green, and the predicted bounding boxes from the class activation map are in red.''~\cite{https://doi.org/10.48550/arxiv.1512.04150} The authors conclude that ``our CAM approach significantly outperforms the backpropagation approach.'' What they exactly mean by outperforming is not disclosed. In particular, it is questionable if any form of `outperforming' can be concluded by observing these results. Figure taken from~\cite{https://doi.org/10.48550/arxiv.1512.04150}.}
    \label{fig:camvis}
\end{figure}

Can we really conclude that for the images provided? Maybe the integrated gradients method \emph{delineates} the objects better than input gradients, but does that mean they reflect distinctive features for the model's predictions (i.e., what the model is looking at) better? That is an entirely different question.\footnote{The paper back in 2017 was not rejected for just making qualitative evaluations. The field has grown and matured a lot since then -- today, it is always a requirement to provide proper quantitative evaluations.} Another claim from the paper:
\begin{center}
``We observed that the results make intuitive sense. E.g., `und' is mostly attributed to `and', and `morgen' is mostly attributed to `morning'.''~\cite{sundararajan2017axiomatic}
\end{center}
To humans, this makes perfect sense. However, what if the model looked at a different feature for predicting these words? We argue that this is a case of confirmation bias. If we keep relying on human intuition to measure/evaluate explainability, how could we detect models that rely on new knowledge humans have not learned before? This point of view prohibits us from \emph{learning} from models. Another example from CAM (referring to a bunch of visualizations of different methods, shown in Figure~\ref{fig:camvis}):
\begin{center}
``We observe that our CAM approach significantly outperforms the backpropagation approach [...]''~\cite{https://doi.org/10.48550/arxiv.1512.04150}
\end{center}
What do they exactly mean by outperforming? When is an explanation method doing better? Can we really conclude this?\footnote{The authors likely refer to WSOL performance. However, that is just a coarse proxy for explainability methods and does not directly measure the quality of explanations in any way.} This is likely another case of confirmation bias. We can see that qualitative evaluations of soundness are susceptible to confirmation bias. This is made even more severe by the fact that \emph{no GT explanation exists in general}.

\subsubsection{Does localization evaluation make sense for soundness?}

For the quantitative evaluation of CAM, the authors measure the number of times their attribution score map corresponds to the object bounding box. They segment regions whose CAM value is above 20\% of the maximum CAM value. Then they take the tightest bounding box that covers the largest connected component in the segmentation map. Finally, they measure the IoU between this box and the GT object box of the class of choice. When \(\text{IoU} \ge 50\%\), they consider it a success. They measure the success rate on ``the'' ImageNet validation set. They find that the CAM variants perform better than the backprop variants.

\textbf{Setting}: Explanation method \(A\) finds GT object boxes better than explanation method \(B\). Does this mean that explanation method \(A\) is working better than \(B\)? \emph{We do not think so.} The model may have been looking at a non-object region to make the prediction. If that were the case, the explanation method with a lower localization score might explain the model better.

\textbf{Takeaway}: We should not evaluate according to our expectations when evaluating explanation methods.

\subsubsection{How to interpret unintuitive explanations?}

Suppose we have a case when the provided explanation differs greatly from our expectations. Does that mean that the explanation method failed while the model was working fine (it was looking at the right thing), or did the explanation method correctly expose a bug in our model (or in the data), like spurious correlation? There is no way to tell these two scenarios apart from a single visual inspection.

\subsubsection{Typical pitfalls of soundness evaluation}

Soundness aims to evaluate the following: Does the score map \(s(f, x)\) represent the true causes for \(f\) to predict \(f(x)\)? The true explanation depends on both the input \(x\) and the model \(f\). We can also calculate the attributions for the GT class \(y\) or any other class. This is usually done less in practice. Explaining something that has already happened (e.g., \(f\) predicted \(f(x)\) for \(x\)) makes more sense than explaining hypothetical situations. In this case, the true explanation depends on \(x, f,\) and \(y\). The problem with qualitative evaluation is that humans also cannot tell what the cues were that \(f\) looked at to predict a certain class. We are only looking at \(x\) and \(y\) to make the evaluation, not \(f\). This seems wrong \emph{by design}.

The problem with localization evaluation is that if we compare to GT localization, we also do not take the model \(f\) into account, only \(x\) and \(y\). This also seems wrong by design. We know already that models do not always look at foreground cues to predict classes.

The fundamental issue with evaluating soundness is that there is no GT explanation in general.\footnote{The model is, of course, sound to its own behavior. However, we cannot treat a system as its own explanation. That kills the purpose.} Humans cannot provide GT explanations. This is precisely the reason we are developing an explanation technique in the first place. If there were a GT explanation for a model, then that itself would be a good explanation, and there would be no need to study what a good explanation is and evaluate explanations. We should start from somewhere, but it is hard. We are facing a chicken-egg problem.

\subsection{Evaluation of Soundness of Explanations based on Necessary Conditions}

There is a trick that people consider to test the soundness of explanation methods. We define a few criteria that a successful explanation method must satisfy.\footnote{Without question. Full stop.}

\textbf{Example}: The explanation \(s(f, x)\) must not contain \emph{any} information if \(f\) is not a trained model (i.e., it is randomly initialized). The intuition is, ``How could any explanation contain any interesting information for an untrained model?'' Otherwise, our explanation is rather trying to please human qualitative evaluations by producing plausible explanations. Interestingly, a randomly initialized CNN achieves a better score than random guessing with a trained linear layer on top because of inductive biases. On ImageNet-1K, one can achieve \(4\%\) accuracy~\cite{https://doi.org/10.48550/arxiv.2106.05963}.\footnote{A simple linear classifier cannot perform better than random guessing.} It seems that this is probably a way too strong necessary condition. There can be some information in the score map (Why not? We do not fully know the behavior of a randomly initialized model.), but the main point is that the score map should strongly depend on function \(f\). The explanation does not have to be \emph{informationless} when a model is randomly initialized. If it turns out that the score map is independent of the model altogether, then something is wrong.

A relaxed version of the above is that when the model changes (becomes gradually randomly initialized from a trained model), we should also see notable changes in the attribution map.

\subsection{Sanity Checks for Saliency Maps}

\begin{figure}
    \centering
    \includegraphics[width=\linewidth]{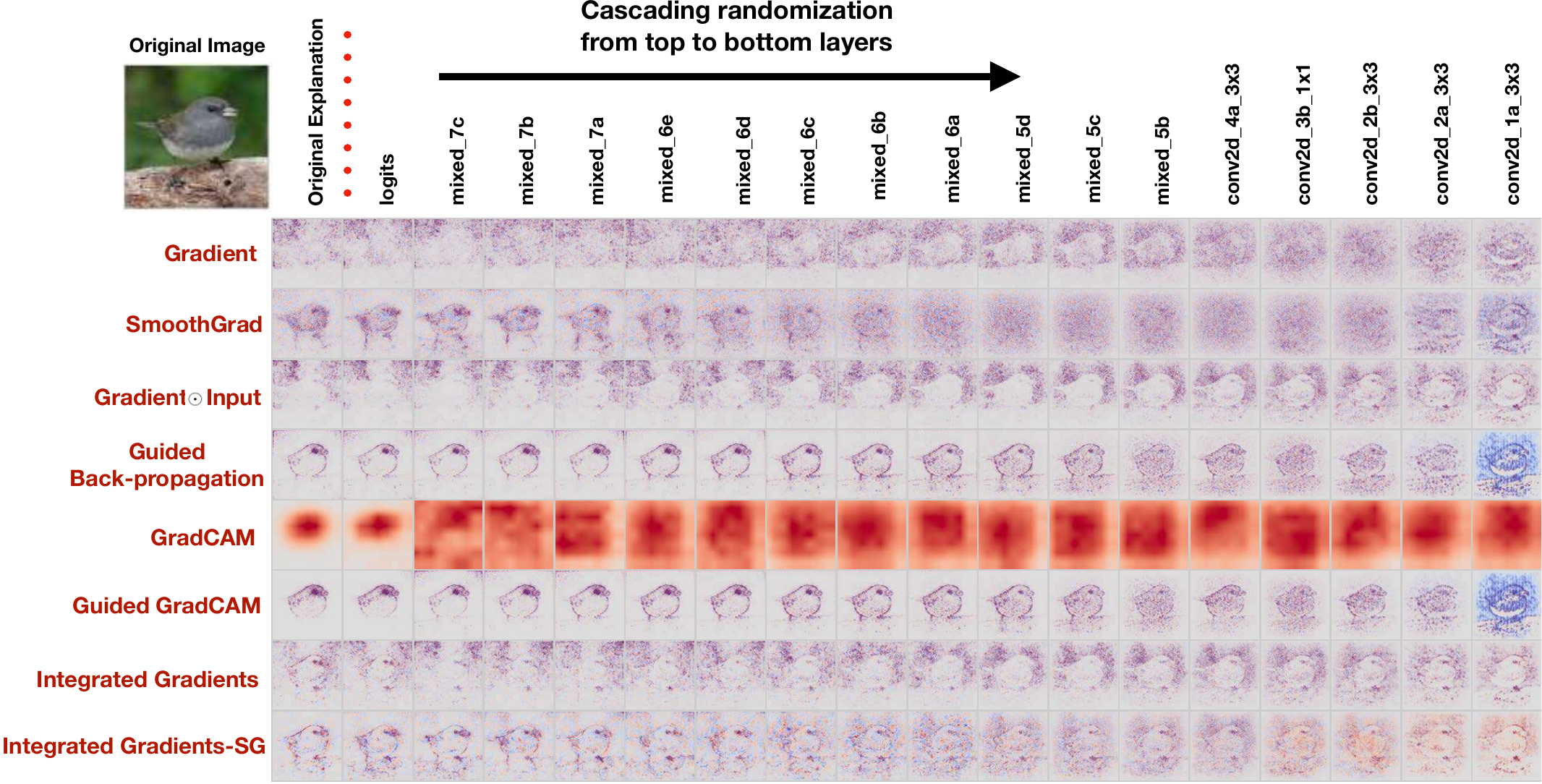}
    \caption{Results of various explainability methods on the cascading model parameter randomization sanity check. This sanity check is passed by saliency maps, SmoothGrad, and Grad-CAM. Details are discussed in the text. Figure taken from~\cite{https://doi.org/10.48550/arxiv.1810.03292}.}
    \label{fig:results}
\end{figure}

Let us discuss the paper ``\href{https://arxiv.org/abs/1810.03292}{Sanity Checks for Saliency Maps}''~\cite{https://doi.org/10.48550/arxiv.1810.03292} where the authors benchmarked various explainability methods on sanity check tasks. We highlight two of these:
\begin{enumerate}
    \item \textbf{Cascading randomization.} As we randomize the network's weights (starting from the latest layers and going toward the input layer), we should see notable changes in the explanations XAI methods give.
    \item \textbf{Label randomization.} For models trained with randomized labels (that should not learn anything meaningful), XAI methods should not highlight parts of the input that are discriminative for the \emph{original task} (without label randomization). They should return irrelevant attribution maps.
\end{enumerate}

\subsubsection{Cascading randomization}

Let us discuss Figure~\ref{fig:results} showcasing \emph{cascading normalization}. Saliency maps (called `gradient' in the Figure) exhibit large changes in the attribution map. SmoothGrad is also heavily influenced by randomization: it ``passes the check.'' Curiously, the image does not become complete noise from the initially clear attribution map, rather, it becomes a noisy edge detector.) For Gradient \(\odot\) Input, the outline of the bird is always visible: the changes are not so large. 

Guided Backpropagation~\cite{https://doi.org/10.48550/arxiv.1412.6806} shows a similar attribution map all the way, even after a global change of the model. It only becomes noisier, the edges are clear all the way. It seems like it does not take model \(f\) into account that much. For Guided Backpropagation, the authors were selling the fact that they get very nice visualizations of objects~\cite{https://doi.org/10.48550/arxiv.1412.6806}. This is true, but it does not reflect well what the model is doing. It is close to being an edge detector, but at the time when they published it, it looked like a ground-breaking technology. No one has done it before, and the results looked like the classifier had all the knowledge about where objects are. But even though it looked promising at the time, people have since then realized it does not work. It does not contain enough information about the model. Let us give a rough outline of the Guided Backpropagation method. If we use ReLU, then during backpropagation, when the pre-activation is negative, we do not backpropagate gradients. When it is positive, we just let the gradient through. In Guided Backpropagation, we do not let the gradient through when it is negative. (Like a ReLU on gradients.) This results in an AND condition: if the gradient was positive and the pre-activation was positive, then we let the gradient through. There is no justification for why this should work. And it does not, apparently.

Continuing with previously discussed methods, Grad-CAM showcases large changes in the attribution map as well. Regarding Guided Grad-CAM, we could give the same remarks as for Guided Backpropagation. This is a multiplication of the guided backpropagation score map and the CAM score map. It is natural that the method inherits lots of issues from guidance. Integrated Gradients gives very similar results to Gradient \(\odot\) Input. Attributions change only slightly -- definitely not as radically as for, e.g., SmoothGrad. Integrated Gradients-SG is very similar to Integrated Gradients, maybe even a bit worse.

\textbf{Note}: Earlier versions of~\cite{https://doi.org/10.48550/arxiv.1810.03292} give significantly different results (even more extreme).

This sanity check was set up as a necessary condition: Any explanation method (even the simplest, most naive ones) should satisfy it. If they do not, the method is unusable. It is the bare minimum requirement an explanation method has to satisfy. Nevertheless, some methods already fail to pass this simple test. Namely, Guided BP and Guided Grad-CAM are essentially edge detectors. Gradient \(\odot\) Input and Integrated Gradients are also not so convincing.

\begin{figure}
    \centering
    \includegraphics[width=0.8\linewidth]{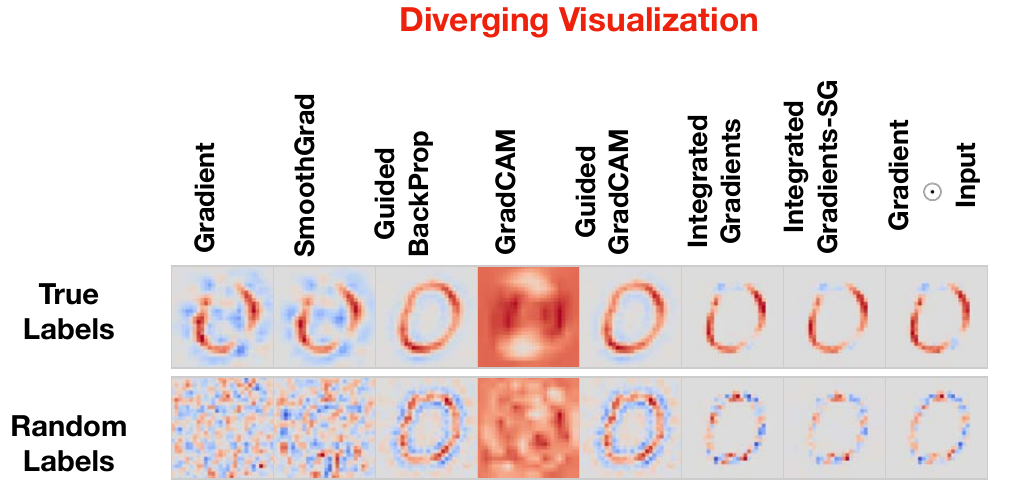}
    \caption{Results of various explainability methods on the data randomization (randomized labels) sanity check. This check is also passed by saliency maps, SmoothGrad, and Grad-CAM. Details are discussed in the text. Figure taken from~\cite{https://doi.org/10.48550/arxiv.1810.03292}.}
    \label{fig:sanity2}
\end{figure}

\subsubsection{Label randomization}

We discuss the other aforementioned sanity check the paper considered, which uses \emph{random labels} named `data randomization test'. The results can be seen in Figure~\ref{fig:sanity2}. One can also compare explanations for two models trained on MNIST with true (original) labels or random labels (control group). Random-label-trained models should return explanations without information. With Guided Backpropagation, we can clearly see the shape of 0 for random labels; it seems to give edge detection regardless of the label used for training. Guided Grad-CAM also fails the test again. Methods depending on pixel values tend to show a ``0'' shape even for random label models: Integrated Gradients\footnote{This is understandable -- it is multiplying gradient-based attribution with the pixel value differences between the image and baseline (a black image -- MNIST). If we multiply the gradient-based attribution with the image of this 0 number, we will see a 0 in the attribution map.}, Integrated Gradients-SG, and Gradient \(\odot\) Input all showcase the same problem. Gradient, SmoothGrad, and Grad-CAM look more random: We say they pass the test.

This shows that any method trying to multiply the input onto the score map is strange. Even in such cases where we should not attribute to any meaningful pixels, we see patterns in the map dependent on just the raw input image. Notice how this seemingly simple sanity check already conflicts with the theoretically justified completeness axioms.

\begin{figure}
    \centering
    \includegraphics[width=0.7\linewidth]{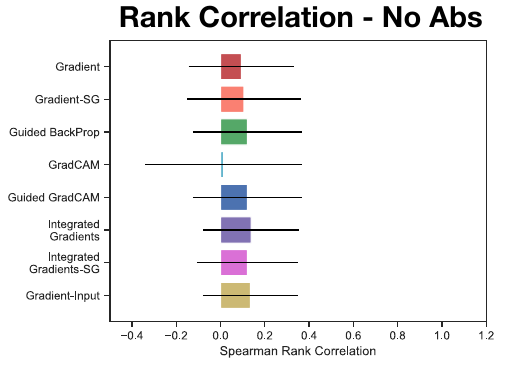}
    \caption{Spearman rank correlation barplot (without absolute values) of various explainability methods for an MLP. Grad-CAM gives convincing results. Details are discussed in the text. Figure taken from~\cite{https://doi.org/10.48550/arxiv.1810.03292}.}
    \label{fig:sanity3}
\end{figure}

\subsubsection{Quantitative results: rank correlation}

We also briefly discuss a correlation plot in the paper, shown in Figure~\ref{fig:sanity3}. How much correlation can we see between the upper and bottom rows for each method in Figure~\ref{fig:sanity2}? Grad-CAM has a rank correlation of almost 0 for pixels on average. It satisfies the overall necessary condition the best -- no correlation in attribution ranking for true/random labels. It does not seem to show any correlation between the explanation for the model trained with true labels vs. the model trained with random labels.

Grad-CAM and SmoothGrad are still generally perceived as one of the best explanation methods.

We have seen that there is no GT explanation in general, and we are facing a chicken-egg problem. However, if we are a bit creative, we can simulate samples where the GT explanation actually exists (to some high extent). We know with very high confidence where the model should be looking for these images. According to the attribution map, we then check whether it is actually looking at the ``GT part'' of the image.

\subsection{Simulation of Inputs with GT Explanations}

We discuss a possible way to simulate inputs with GT explanations from the paper ``\href{https://arxiv.org/abs/1711.11279}{Interpretability Beyond Feature Attribution: Quantitative Testing with Concept Activation Vectors (TCAV)
}''~\cite{https://doi.org/10.48550/arxiv.1711.11279}. We define three classes: zebra, cab, and cucumber.

\begin{figure}
    \centering
    \includegraphics[width=0.8\linewidth]{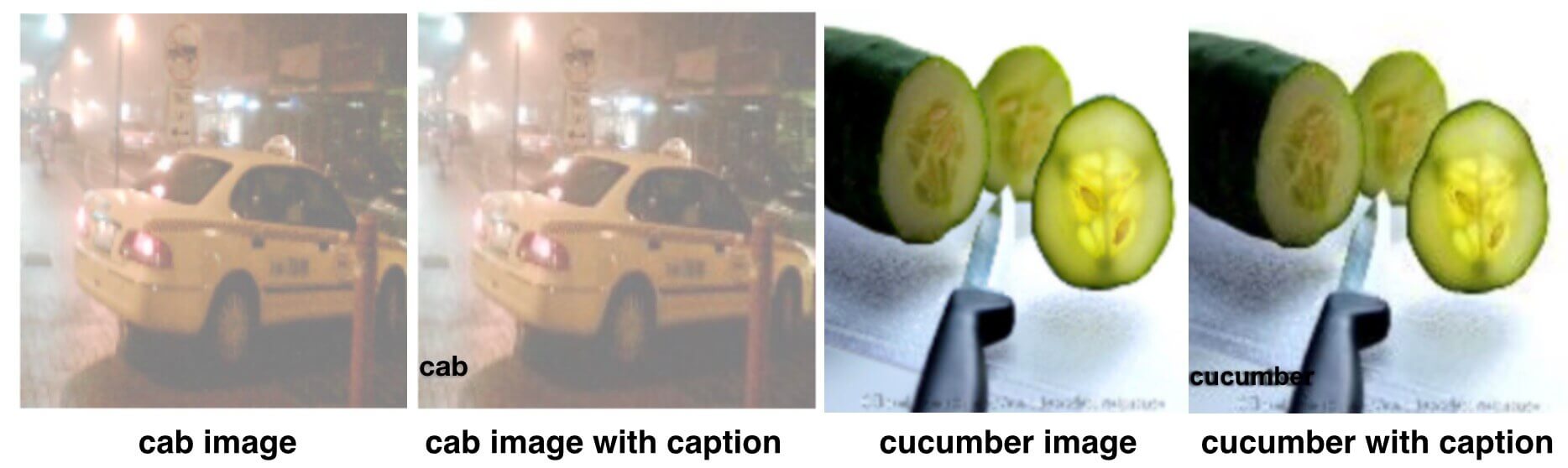}
    \caption{Samples from the dataset with ``GT attributions'' introduced in~\cite{https://doi.org/10.48550/arxiv.1711.11279}. The image label is included in the image with a noise parameter that controls the probability that the label is correct. Figure taken from~\cite{https://doi.org/10.48550/arxiv.1711.11279}.}
    \label{fig:tcavex}
\end{figure}

We provide potentially noisy captions written in the bottom left corner of the image. We have a controllable noise parameter \(p \in [0, 1]\) to control the impact of the captions. In detail, \(p\) is the probability that the caption disagrees with the image content. \(p = 0\) means there is no disagreement: an image of a cucumber would always have the caption `cucumber'. For \(p = 0.5\), each image has a 50\% chance of the caption and the image content disagreeing. Examples are given in Figure~\ref{fig:tcavex}. We have a feature selection problem (look at caption vs. image), but for low noise levels, the caption is a very prominent feature the model cannot resist looking at (refer to \emph{simplicity bias} in \ref{ssec:simplicity}). We will measure if the attribution methods are correctly picking that up.
When the noise level is high, the model cannot rely on the captions at all. Thus, we will measure if the attribution methods are correctly \textbf{not} attributing the predictions to the label.

\subsubsection{GT Attribution Results}

\begin{figure}
    \centering
    \includegraphics[width=0.8\linewidth]{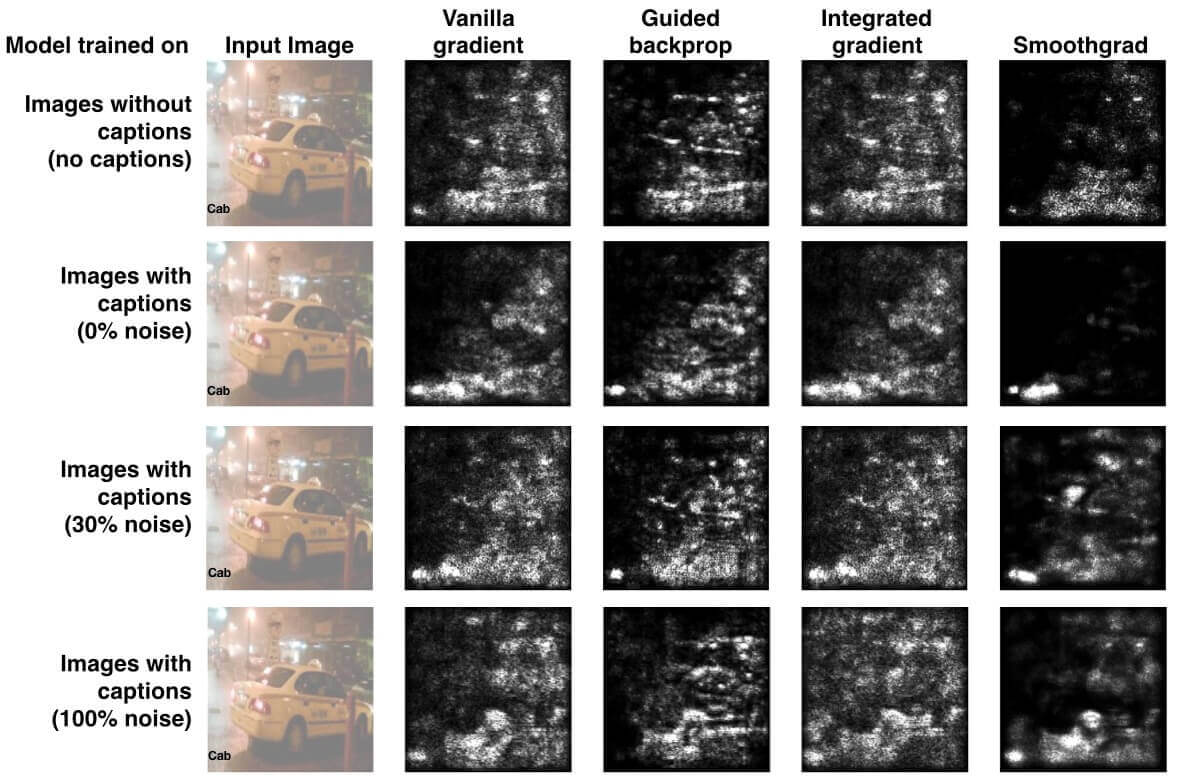}
    \caption{Results of various XAI methods on the dataset with ``GT attributions'' introduced in~\cite{https://doi.org/10.48550/arxiv.1711.11279}. The `cab' caption has to be a strong cue for recognition if \(p = 0\). Conversely, it has to be a weak cue for recognition if \(p = 1\). The test input image contains the \emph{correct} caption. For the model trained on images with captions and 0\% noise, we expect the attribution to be wholly focused on the caption. It seems like (without quantification) SmoothGrad is doing that most prominently, at least from how they show it. However, the gradient-based explanations are not well-calibrated (not ideal). Depending on how we renormalize the map, we may also get such a strong attribution to the caption for the other gradient maps. We cannot fully trust these kinds of score maps. Figure taken from~\cite{https://doi.org/10.48550/arxiv.1711.11279}. }
    \label{fig:tcavres2}
\end{figure}

Results are shown in Figure~\ref{fig:tcavres2}. Based on these results, SmoothGrad seems to be a great explainability method.

\subsection{Remove-and-Classify/Remove-and-Predict}
\label{sssec:rac}

\begin{definition}{Remove-and-Classify/Remove-and-Predict}
The Remove-and-Classify algorithm is a prevalent soundness evaluation method for feature attribution scores. Attribution scores define a ranking over features: the feature attribution explanation \(s(f, x) \in [0, 1]^{H \times W}\) ranks each feature in the input \(x\). Remove-and-Classify removes features from the test input(s) iteratively, according to the attribution ranking of the explainability method. In the most popular variant, where the feature with the highest attribution score (most important) is removed first, the explainability method with the steepest drop in classification accuracy performs best. One usually calculates the Area under the Curve (AUC) to compare explanation methods. The features might be removed one by one or in a batched manner.

\medskip

There are several variants of the Remove-and-Classify method. Compared to the variant introduced above, one might\dots
\begin{enumerate}
    \item \dots remove the features with the lowest attribution score (least important) first. In this case, the explainability method with the shallowest drop in accuracy performs best.
    \item \dots start from the base image and introduce features one by one (or in a batched way) according to the ranking of the explainability method -- either \wrt increasing or decreasing attribution score.
\end{enumerate}
Sometimes, people take the average performance on these four possible benchmark combinations.

\medskip

There is no ``correct'' choice of encoding missingness. One must be particularly careful not to introduce \emph{missingness bias} (\ref{sssec:missingness_bias}). The most prevalent removal technique for natural images and pixels as features is replacing the pixel with the mean pixel value(s) in the dataset, which is usually gray for natural images.

\medskip

\textbf{Note}: Just like in counterfactual explanation methods, grayness can still convey information -- it can be problematic to consider this the base value. The quality of this choice also depends on our task -- e.g., what if our task is to detect all gray boxes? Even though the signed distance of all data points to the mean (usually gray) image is zero on average, and the images are scattered around the mean image, it does not mean that individual gray pixels cannot contribute to a model's decision. They can be grouped into arbitrary shapes that have semantic meaning, even though a completely gray image might not convey much semantic information. Sticking to \emph{any} color has a potential pitfall.
\end{definition}

A feature attribution explanation gives a ``heat map'' of the given input. Suppose \(s(f, x)\) is sound and correctly cites the causes for the prediction (in the correct order of importance). In that case, removing the most critical feature \(i^* = \argmax_i s_i(f, x)\) will significantly decrease the score \(f(x)\) for the class in question. We measure the speed of decrease in classification accuracy as we remove pixels in the order dictated by \(s(f, x)\).

We now discuss the result of remove-and-classify shown in Figure~\ref{fig:rac} that was reported in the CALM paper. The baseline is random erasing with equal probabilities. If we remove pixels, we kill information,\footnote{As we will see in \ref{sssec:missingness_bias}, while this is generally true, there are cases where we \emph{introduce} information by encoding missingness.} so we should still see a drop in accuracy, just not as fast. The used metric is the relative drop in accuracy when erasing according to an attribution method, compared to random erasing. A method is \emph{better} if it results in a faster drop in accuracy. Methods corresponding to curves enveloping others from below are supposed to be more sound explanation methods. Sometimes, we also measure the AUC for this plot, where lower is better. One can also consider the unnormalized plot, where we do not compare against a random baseline.

Our observation is that CALM gives a sound explanation. It gives a huge drop in accuracy for pixels with high attribution scores. For SmoothGrad, the pixels attributed to being most important were not the most important ones, as we see a smaller drop.

One might ponder why most of the methods get worse than random erasing for larger values of \(k\). Filling in gray/black pixels is not the best way to kill information. It can also \emph{introduce} information. We address this in Section~\ref{sssec:missingness_bias}.

\subsection{Missingness Bias}
\label{sssec:missingness_bias}

\begin{figure}
    \centering
    \includegraphics[width=\linewidth]{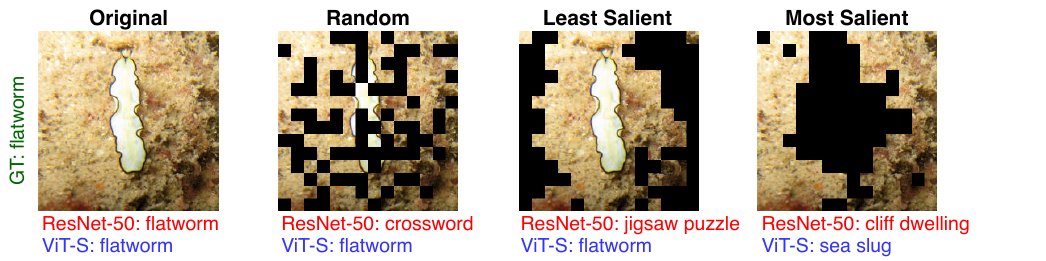}
    \caption{Illustration of the missingness bias.
        ``Given an image of a flatworm, we remove various regions of the original image. Irrespective of what subregions of the
        image are removed (least salient, most salient, or random), a ResNet-50 outputs the wrong class
        (crossword, jigsaw puzzle, cliff dwelling). A closer look at the randomly
        masked image shows that the predicted class (crossword puzzle) is not totally
        unreasonable, given the masking pattern. The model seems to rely on the masking pattern to
        make the prediction rather than the image's remaining (unmasked) portions.
        Conversely, the ViT-S either maintains its original prediction or predicts a reasonable label
        given remaining image subregions.''~\cite{https://doi.org/10.48550/arxiv.2204.08945} Figure taken from~\cite{https://doi.org/10.48550/arxiv.2204.08945}. Replacing pixels with mean values (or any fixed value) does not necessarily remove information. It may add further information (crossword) or kill unnecessary information. We also see that Transformers suffer a lot less from this phenomenon. Thus, depending on different models, the attribution methods might see different success rates. Remove-and-classify is not the perfect soundness evaluation metric. However, it is the most popular and one of the best ways to evaluate soundness.}
    \label{fig:missingness}
\end{figure}

A recent phenomenon named \emph{missingness bias} was reported in a recent paper titled ``\href{https://arxiv.org/abs/2204.08945}{Missingness Bias in Model Debugging}''~\cite{https://doi.org/10.48550/arxiv.2204.08945}. Figure~\ref{fig:missingness} aims to provide some insights. There is no common understanding of what the SotA for erasing information is. We argue that inpainting and blurring are good candidates. However, the choice of the inpainter and the exact blurring method are both hyperparameters that have important implications and might raise new problems. It is hard to explain what exactly is happening and what might be confusing textures for different architectures. It is, however, important to be aware of missingness bias and encode missing information in a suitable way.

We now address the curious behavior in Figure~\ref{fig:rac}. When we only remove information by erasing pixels, we should see the random baseline as the worst-case removing strategy (in expectation). We see the jump above the baseline in Figure~\ref{fig:rac} because the model can predict based on the ``removed'' patterns for random removal, which can introduce greater changes in classification than removing pixels in an orderly fashion. Thus, random removal might add information that confuses the model more. If we erase according to CAM, we will see something like the right of Figure~\ref{fig:missingness}.

\section{Soundness is Not The End of the Story}

There are many other criteria, like soundness, simplicity, generality, contrastivity, socialness, interactivity, but also \red{relevance}. The latter depends on the final goal. Is it to debug? Is it to understand? Is it to gain trust? This is an essential criterion, as we are not looking at explanation methods for the sake of themselves, but we rather treat them as an intermediate step towards a final goal.

\subsection{Various End Goals for Explainability}

\textbf{Model debugging as the end goal.} Here, we wish to identify spurious correlations (why a model has made a mistake) and then fix them (we have seen methods for both in \ref{sssec:identify} and \ref{sssec:overview}). This can improve generalization to OOD data. There are no successful/commercialized explanation tools yet that are specialized in debugging. It is not yet clear how to help an engineer fix a general problem with the model, and we have yet to see a \emph{successful} use case of XAI for model debugging. There is still so much more to be researched for ML explanations. Attribution does not always guarantee successful debugging.

\textbf{Understanding as the end goal.} Do humans understand the idiosyncrasies (``odd habits'') of a model? If a model is doing something odd, understanding why it is doing so could be an interesting objective. Can humans predict the behavior of a model based on the provided explanation? Do humans learn new knowledge based on the explanation? If the model is doing something new that humans cannot do yet, transferring that knowledge to humans would be essential.

\textbf{Enhancing human confidence, gaining trust as the end goal.} Does the explanation technique help persuade doctors to use ML models? Many doctors are still very averse to ML-based advice; they have no trust. Explanations could help them incorporate ML techniques. Does the explanation technique convince people to use self-driving cars (even though safety stays the same -- or, as seen, worse because of the trade-offs)?

The ``soundness'' criterion does not fully align with the previous end goals and desiderata. The current evaluation is too focused on soundness (and qualitative evaluations). Given an explanation, we still have some end goals:
\begin{itemize}
    \item ML Engineer: ``Now I know how to fix model \(f\).''
    \item Scientist: ``Now I understand the mechanism behind the recognition of cats.''
    \item Doctor: ``Now I can finally trust this model for diagnosing cancer.''
\end{itemize}
Soundness focuses only on the explanation itself, which is an intermediate step. We need evaluation with the end goal in mind. There is no way we do not have to use human-in-the-loop (HITL) evaluation at some point.

\subsection{Human-in-the-Loop (HITL) Evaluation}

\begin{definition}{Human-in-the-Loop Evaluation}
Human-in-the-loop evaluation refers to any evaluation technique for explainability that incorporates humans and measures how well the explanations help them achieve their end goals.
\end{definition}

Let us now turn to discussing human-in-the-loop (HITL) evaluation. In particular, we will consider the paper ``\href{https://arxiv.org/abs/2112.04417}{What I Cannot Predict, I Do Not Understand: A Human-Centered Evaluation Framework for Explainability Methods
}''~\cite{https://doi.org/10.48550/arxiv.2112.04417}.

\begin{figure}
    \centering
    \includegraphics[width=\linewidth]{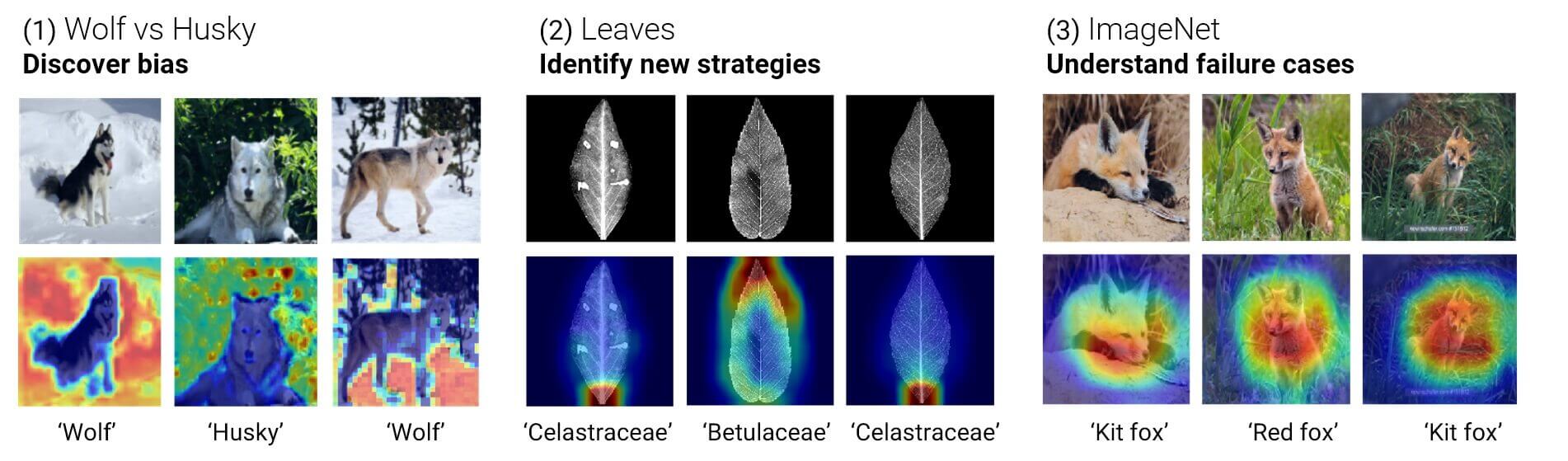}
    \caption{Overview of different tasks considered in~\cite{https://doi.org/10.48550/arxiv.2112.04417}. The authors are evaluating recent explainability methods directly through the end goals of XAI -- practical usefulness. They are trying to see whether explanations are actually helping humans in achieving their end goals. Figure taken from~\cite{https://doi.org/10.48550/arxiv.2112.04417}.}
    \label{fig:hitl1}
\end{figure}

An overview of the settings the authors consider is given in Figure~\ref{fig:hitl1}. They address three real-world scenarios, each corresponding to different use cases for XAI.
\begin{enumerate}
    \item \textbf{Husky vs. Wolf}. Here, debugging is the end goal. Can the explanations help the user identify sources of bias in the model? Examples include background bias (snow, grass) instead of the animal.
    \item \textbf{Real-World Leaf Classification problem}. Here, understanding is the end goal. Can the explanations help the user (non-expert) learn what parts of the leaf to look for to distinguish different leaf types? The humans want to adopt the strategy of the model.
    \item \textbf{Failure Prediction Problem}. Here, understanding is the end goal again. This dataset is a subset of ImageNet. It consists of images, of which half have been misclassified by the model. Can the explanations help the user understand the failure sources of the (otherwise high-performing) model?
\end{enumerate}

\begin{figure}
    \centering
    \includegraphics[width=0.8\linewidth]{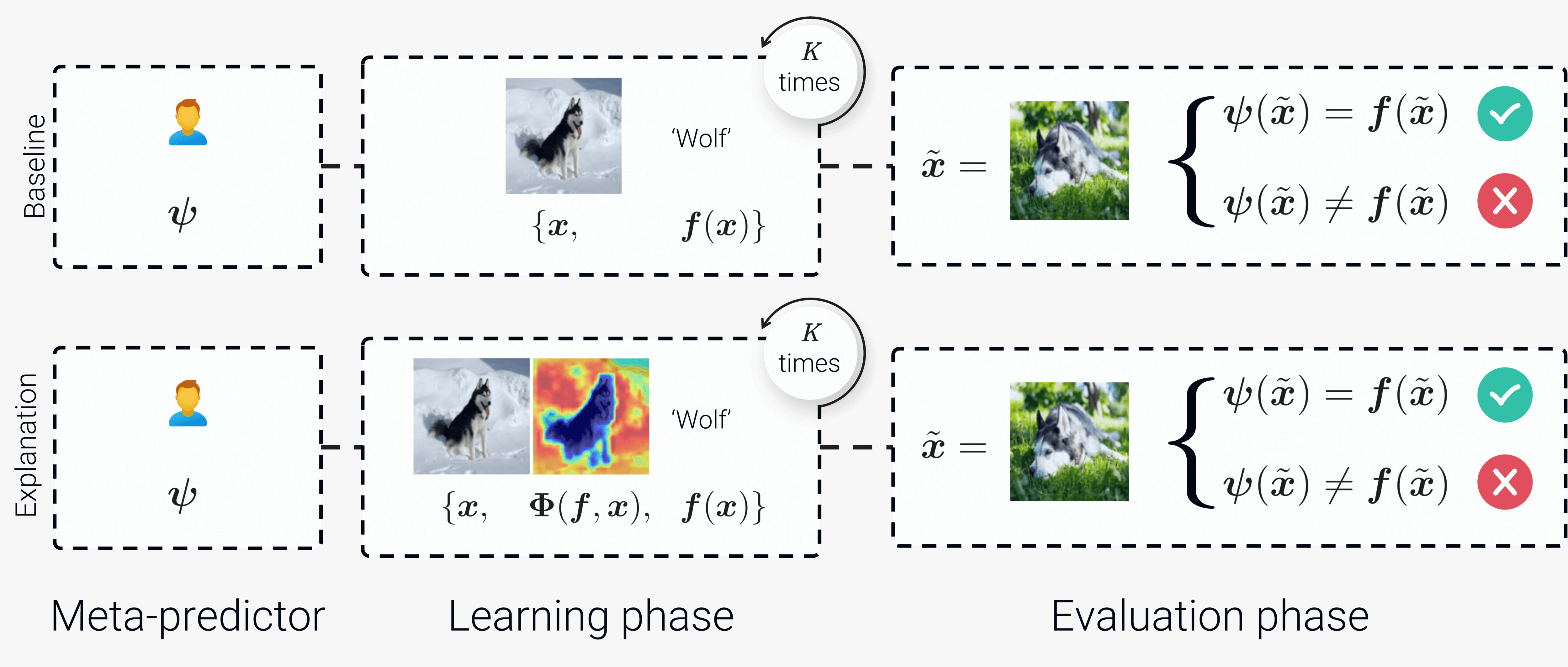}
    \caption{Overview of the stages of the method considered in~\cite{https://doi.org/10.48550/arxiv.2112.04417}. The authors use a human-centered framework for the evaluation of explainability methods. The evaluation pipeline consists of (1) the predictor \(f\), which is a black-box model, (2) an explanation method \(\Phi\), and (3) the meta-predictor, a human subject \(\psi\) whose task is to understand the behavior of \(f\) based on samples (i.e., the rules that the model uses for its predictions). First, the meta-predictor is trained using \(K\) triplets \((x, \Phi(f, x), f(x))\), where \(x\) is an input image, \(f(x)\) is the model's prediction and \(\Phi(f, x)\) is the explanation of the model's prediction. Second, for the Husky vs. Wolf and the Failure Prediction problems, the meta-predictor is evaluated on how well they can predict the model's outputs on new samples \(\tilde{x}\). This is done by comparing the meta-prediction \(\psi(\tilde{x})\) to the true prediction \(f(\tilde{x})\). For the leaf classification problem, the meta-predictor is evaluated on how well they can classify the leaves after observing the explanations. The meta-prediction \(\psi(\tilde{x})\) is compared to the GT label \(y\). Figure taken from~\cite{https://doi.org/10.48550/arxiv.2112.04417}. }
    \label{fig:hitl2}
\end{figure}

Figure~\ref{fig:hitl2} gives a detailed description of how the explainability methods are evaluated in all three scenarios. If the model makes a mistake on the evaluation image, the human should be able to pinpoint the mistake the model will make. Similarly, humans should be able to learn to classify leaves based on the knowledge encoded by the networks, and they should also be able to identify biases under the assumption that the explanation method works well. The paper uses the term \emph{simulatability}. A model is explainable when its output can be predicted following the explanations.

First, we train humans on dataset \(\cD = \{(x_i, f(x_i), \Phi(f, x_i))\}_{i = 1}^K\). For new samples, we let the humans predict the model predictions. The value \(\psi^{(K)}(x)\) is the human prediction of the model prediction after training with \(K\) samples). The Utility-K score is calculated as follows:
\[\operatorname{Utility-K} = \frac{P(\psi^{(K)}(x) = f(x))}{P(\psi^{(0)}(x) = f(x))}.\]
In words, the utility score is the relative accuracy improvement of the meta-predictor trained with or without explanations. The baseline factors out the contribution of explanations for educating humans. Humans for the baseline predictions are trained on dataset \(\cD = \{(x_i, f(x_i))\}_{i = 1}^K\). To make the evaluation meaningful for Husky vs. Wolf and Failure Prediction, the authors mixed correct and incorrect model predictions 50-50\% during evaluation.

\begin{figure}
     \centering
     \begin{subfigure}[b]{0.49\textwidth}
         \centering
         \includegraphics[width=\textwidth]{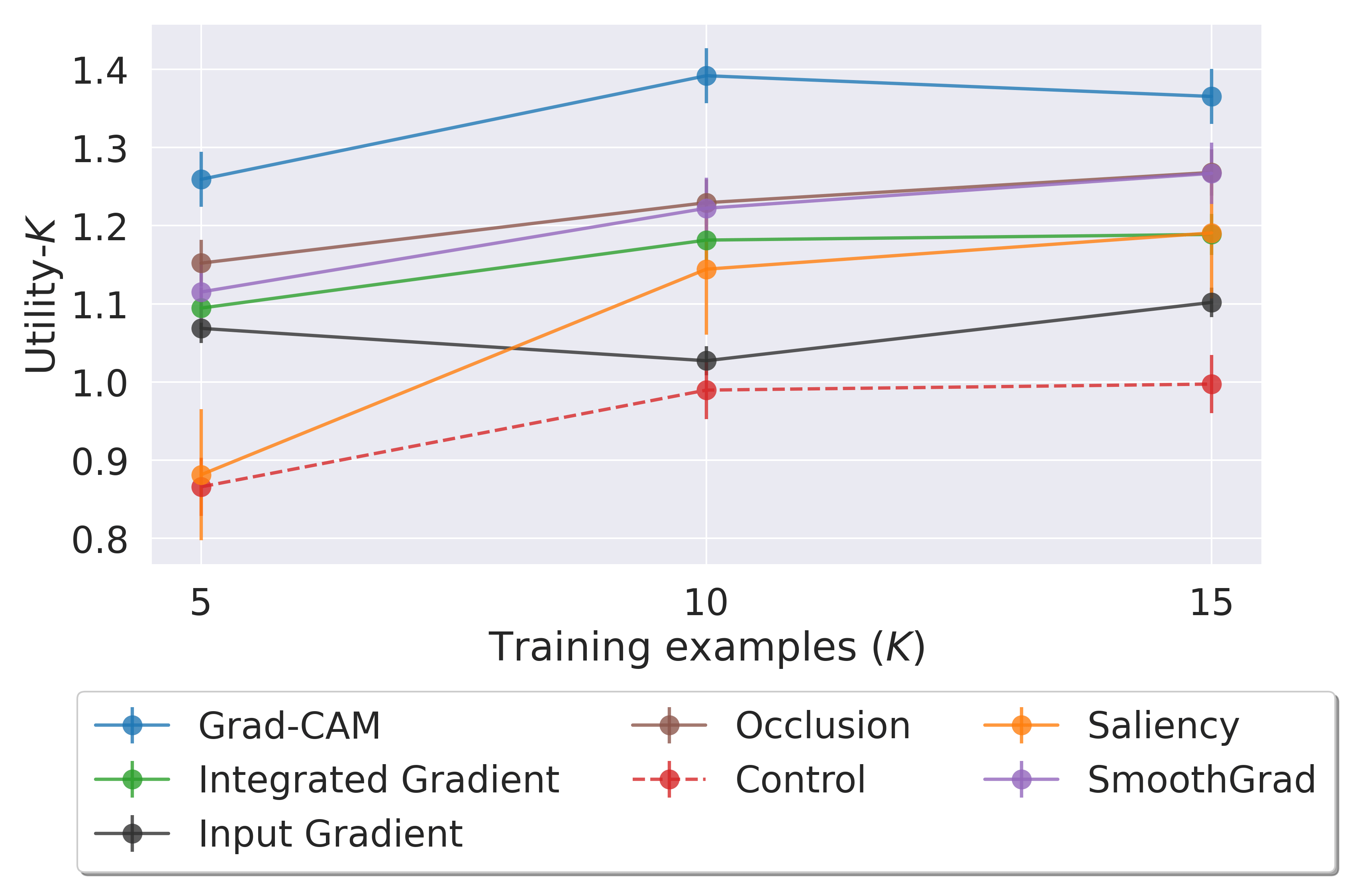}
     \end{subfigure}
     \hfill
     \begin{subfigure}[b]{0.49\textwidth}
         \centering
         \includegraphics[width=\textwidth]{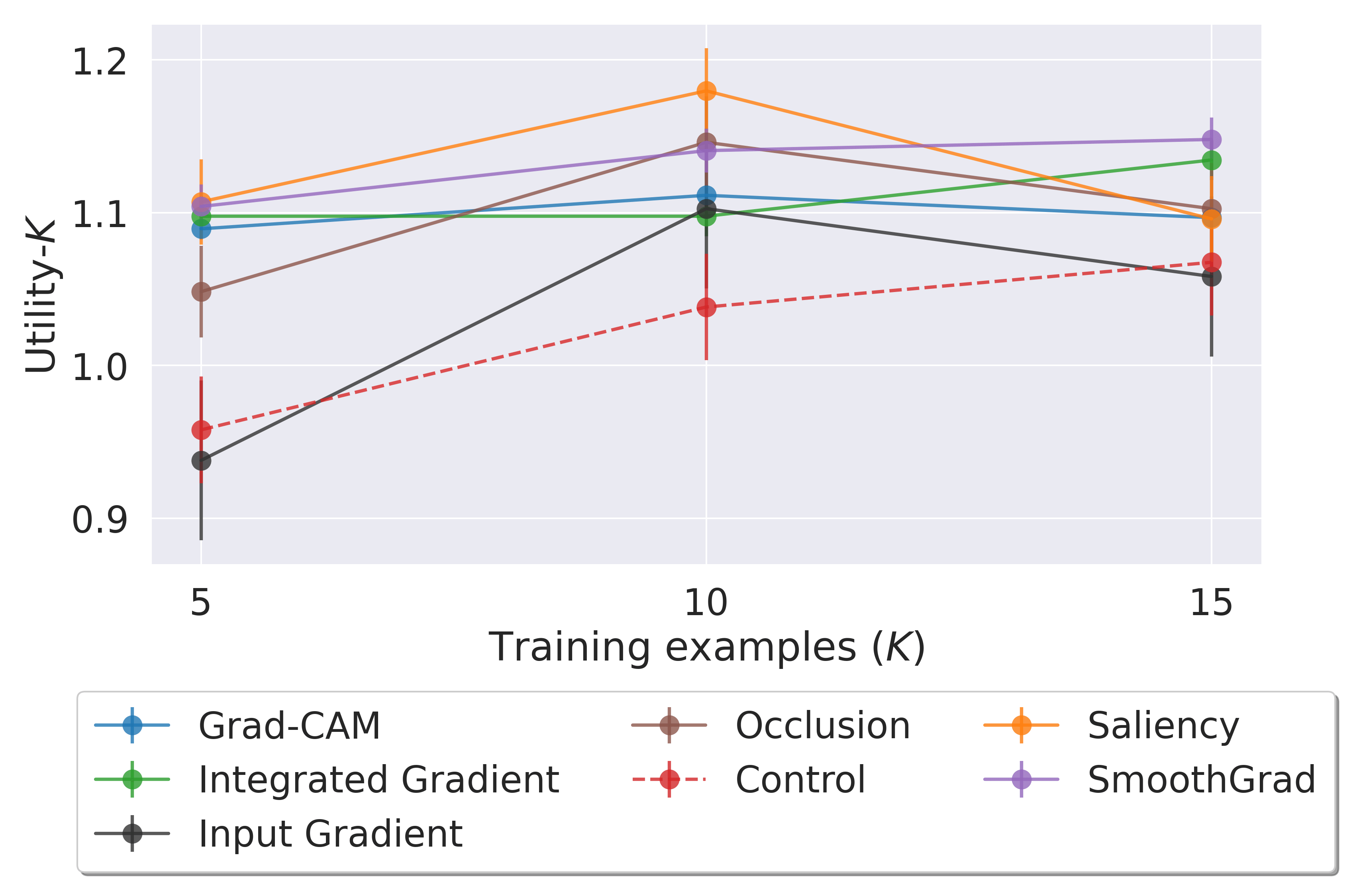}
     \end{subfigure}
    \caption{Results on the Wolf vs. Husky task (left) and the Leaf Classification task (right). For the Leaf Classification task, the Utility-K value is the normalized accuracy of the human predictor on test leaf images after observing the explanations. For the Husky vs. Wolf task, Grad-CAM, Occlusion, and SmoothGrad are seemingly useful. For the Leaf Classification task, Saliency, Smoothgrad, and Integrated Gradients seem to perform best. Figure taken from~\cite{https://doi.org/10.48550/arxiv.2112.04417}.}
    \label{fig:tasks}
\end{figure}

\subsubsection{Task (i): Husky vs. Wolf}

For Husky vs. Wolf, results are shown in the left panel of Figure~\ref{fig:tasks}. The control group is shown some score map, called bottom-up saliency, that is not an explanation (it is independent of model \(f\)). This is used to rule out the possibility that people try harder to solve the task if any explanation is provided to them. CAM achieves good results (same story as before); SmoothGrad and Occlusion are also good. All attribution methods used are better than the control `method'. More training samples mean further knowledge of what the model might be doing.

\subsubsection{Task (ii): Leaf Classification}

The results for Leaf Classification are shown in the right panel of Figure~\ref{fig:tasks}. This is an example of using ML to educate humans. In this case, Utility-K is the normalized accuracy of the meta-predictor after ``training''. Humans do not know how to distinguish these leaf types in the beginning. By showing where the model is looking (that can solve the task well), humans also learn how to classify the leaves (as they learn useful cues). SmoothGrad is consistently good for educating humans for the task. CAM helps a bit less ideally compared to SmoothGrad. Saliency also performs well but does not scale well to more training samples (\(K = 15\)). Integrated Gradients is on par with CAM and also scales better.

\subsubsection{Task (iii): Failure Prediction}

\begin{figure}
    \centering
    \includegraphics[width=0.6\linewidth]{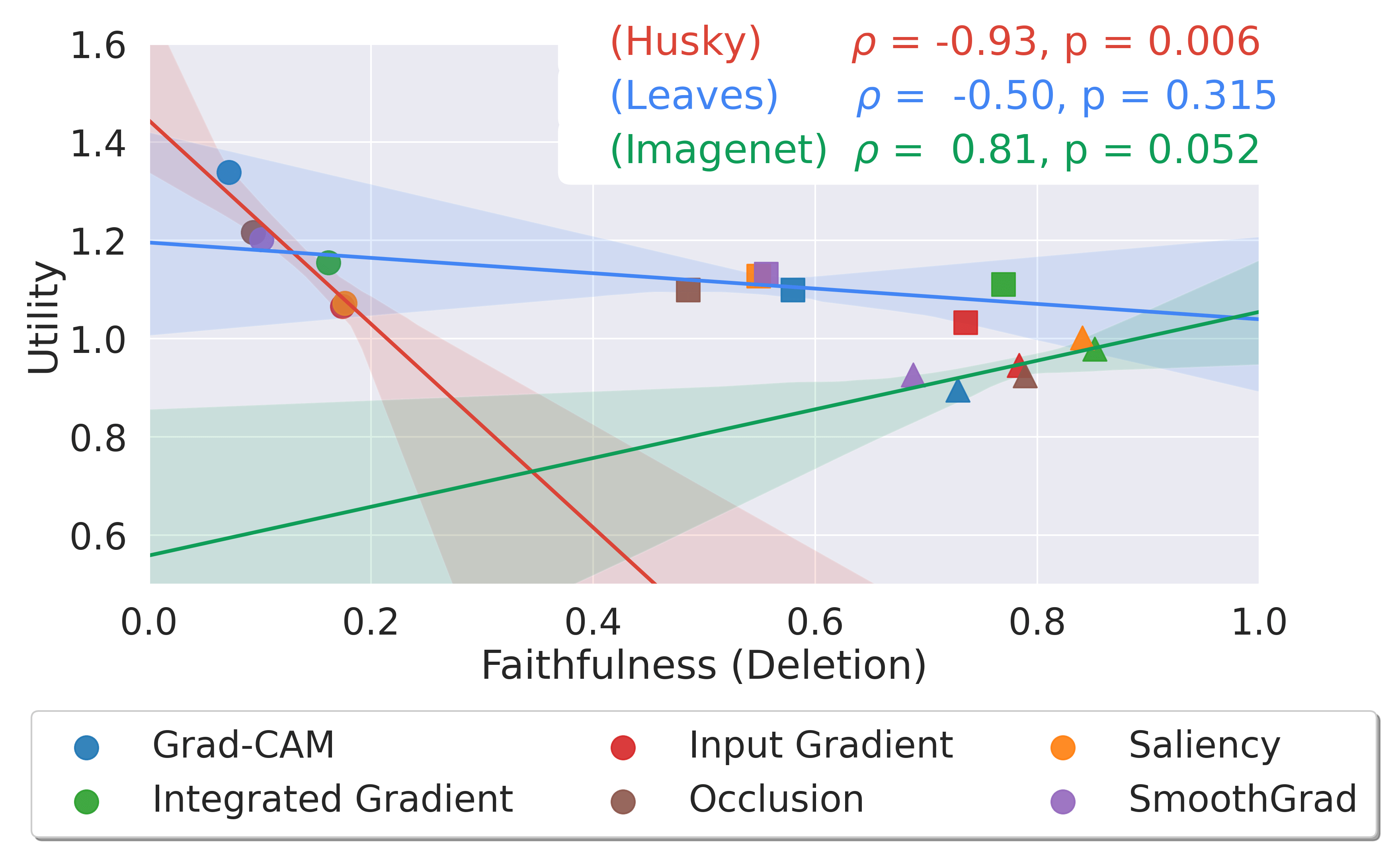}
    \caption{Correlation of the faithfulness and the utility score in~\cite{https://doi.org/10.48550/arxiv.2112.04417}. HITL Utility does not correlate in general (across datasets) with the evaluation's soundness (faithfulness). (Correlations change across tasks. It seems very random.) Faithfulness metrics are poor predictors of end-goal utility. For the Husky vs. Wolf and leaves datasets, a negative correlation can be observed, meaning \emph{high soundness might even come with the price of less end-goal utility}. Figure taken from~\cite{https://doi.org/10.48550/arxiv.2112.04417}.}
    \label{fig:hitl3}
\end{figure}

On the ImageNet dataset, \emph{none of the methods tested exceeded baseline accuracy}. These results made the authors suspicious that the explanation methods might not be sound: If the user observes explanations from a method that is not sound, it will not gain enough insight into the model's internals. The authors compared Utility scores (AUC scores under the (K, Utility-K) curve for various K values) to corresponding faithfulness scores. The results are shown in Figure~\ref{fig:hitl3}.

\subsubsection{Conclusion for HITL Evaluation}

SmoothGrad is doing a great job in helping humans with the end goals considered in the benchmark. If we care about how humans can understand attributions and learn from attributions (explanations), then soundness evaluation is not a good proxy for choosing between methods. Thus, HITL evaluation cannot be replaced with soundness evaluation.

\begin{information}{HITL}
Out of the three downstream use cases mentioned in the book, HITL evaluation seems to be more tailored toward understanding. How could it measure how much trust is given to the model? How could we measure how much a method helps fix a model (if the answer is not spurious correlations)? While HITL is a step forward compared to soundness evaluation in the sense that it measures how humans understand better, it still does not measure the end-to-end metric of how much more trust is given or how much better a model is debugged after an explanation in general. End-goal-tailored explainability is a young field with many questions to be answered.
\end{information}

\section{Towards Interactive Explanations}

Previously, we have seen that for HITL evaluations, we need to include human participants to evaluate how useful a method is for human subjects and their end goals. \emph{Interactive explanations} are also deemed necessary by decision-makers.

\subsection{A Survey on Explanations}

\begin{information}{Quant}
A quant, short for quantitative analyst, is a person who analyzes a situation or event (e.g., what assets to buy/sell in a hedge fund), specifically a financial market, through complex mathematical and statistical modeling.
\end{information}

We consider a survey for decision-makers using ML, titled ``\href{https://arxiv.org/abs/2202.01875}{Rethinking Explainability as a Dialogue: A Practitioner's Perspective}''~\cite{lakkaraju2022rethinking}. The survey aims to find answers to the question, ``What kind of features do you need from explanations?''

\subsubsection{Desiderata for Interactive XAI}

Let us now discuss the survey for domain experts using ML in detail. In particular, we consider exact statistics from the survey. \textbf{Note}: There is only a small number of respondents, but as they are experts, conducting such surveys is expensive. The quotes are imaginary and only illustrate the discussed desiderata. The list also does not mean that there are technologies already satisfying these desiderata. We are far away from many aspects still.

24/26 respondents wish to eliminate the need to learn and write the commands for generating explanations. ``We do not want to care about writing code. We need a more natural-language-based interaction with the system.''

24/26 respondents prefer methods that describe the accuracy of the explanation in the dialogues. A notion of \emph{uncertainty} is needed.

23/26 respondents wish to use explanation tools that preserve the context and enable follow-up questions. ``If we do not understand something in the previous round, we should be able to ask for follow-up explanations.'' A key characteristic of a dialog-based system is that the machine should remember previous topics/conversations.

21/26 respondents would like real-time explanations. ``Do not take several hours to answer our questions. We want an experience as if we were talking to a human.'' This is a rather basic requirement for efficiency.

17/26 respondents would let the algorithm decide which explanations to run. Users should not have to ask for a specific explainability algorithm. ``We do not wish to decide ourselves, as so many of them exist. We do not want to build a benchmark, compare all attribution methods, and decide on an appropriate one for the use case. The system should determine the best algorithm for our domain.''

\subsubsection{Key Takeaways from the Desiderata}

Decision-makers prefer \emph{interactive} explanations. Explanations are preferred in the form of \emph{natural languages}. Experts want to treat machine learning models as ``another colleague'' they can talk to. For example, a hedge fund might find good use of ML: They might wish to have a virtual human (a quant) sitting next to them who can answer questions like ``Why do you think this trend is happening?'' or ``Why did you buy/sell the stocks?'' They want to ask the models' \emph{opinion} or what they had in mind when making a decision. In particular, they want models that can be held accountable by asking why they made a particular decision through expressive and accessible natural language interactions.

\subsection{Generating Counterfactual Explanations with Natural Language}

\begin{figure}
    \centering
    \includegraphics[width=0.6\linewidth]{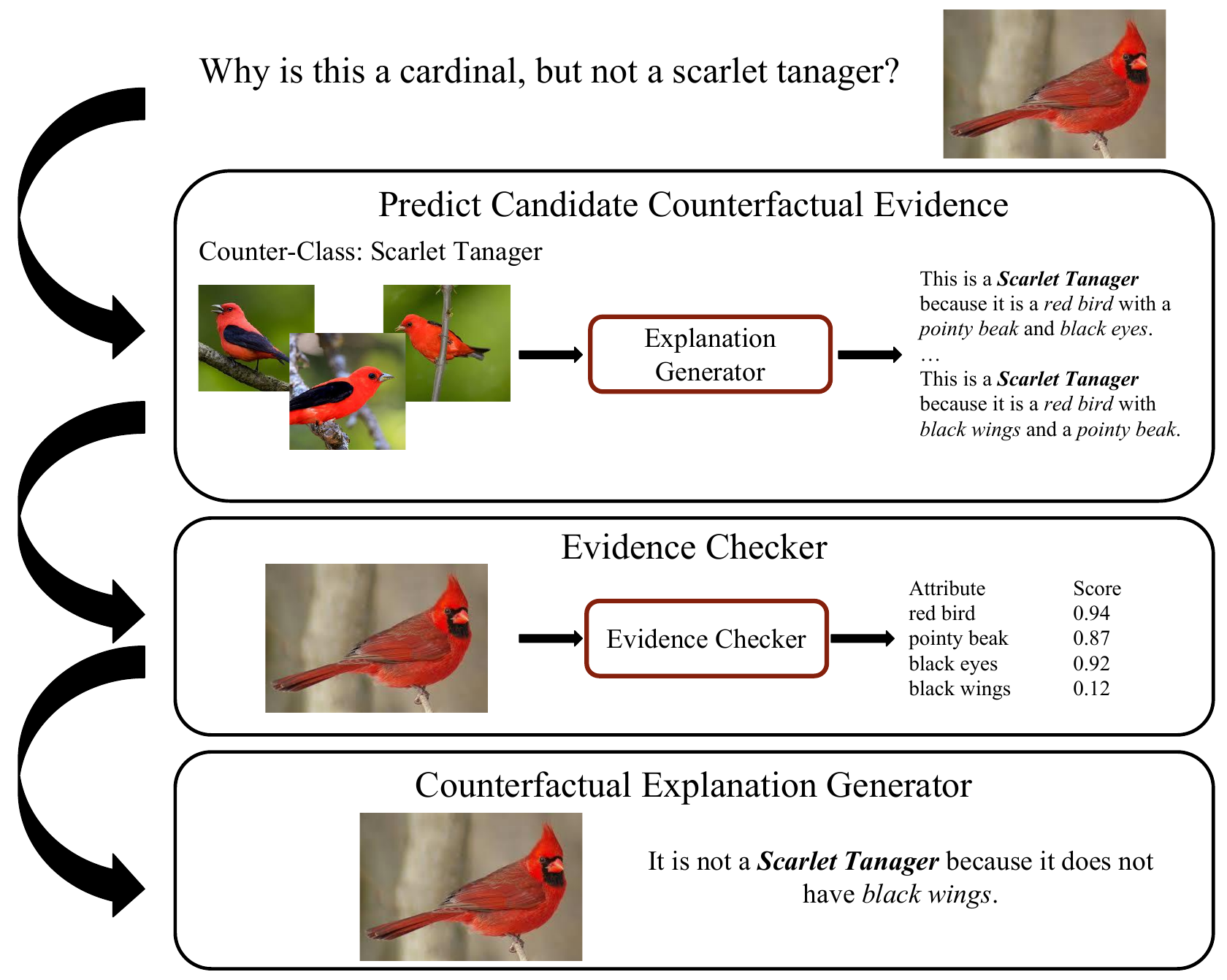}
    \caption{Overview of the counterfactual explanation pipeline of~\cite{https://doi.org/10.48550/arxiv.1806.09809}. The method allows users to generate explanations based on high-level concepts. We have a bird classifier available. There is also an explanation generator. This is different from image captioning, as image captioning only talks about what is in the image, but the explanation generator first makes a prediction (Scarlet Tanager) for the set of counter-class images and describes in natural language why it thinks it is that class. The evidence checker checks how many characteristics extracted from the explanation generator's explanations are present in the current input (i.e., it makes a list of scores). It is checking for evidence of these characteristics in the current image. The counterfactual explanation generator can use the evidence to answer the counterfactual question. Figure taken from~\cite{https://doi.org/10.48550/arxiv.1806.09809}.}
    \label{fig:natural}
\end{figure}

The ``\href{https://arxiv.org/abs/1806.09809}{Generating Counterfactual Explanations with Natural Language}''~\cite{https://doi.org/10.48550/arxiv.1806.09809} paper is a work of~\citeauthor{https://doi.org/10.48550/arxiv.1806.09809}. An overview of the method is given in Figure~\ref{fig:natural}. This is a step towards interactive explanations for humans.

\begin{figure}
    \centering
    \includegraphics[width=0.6\linewidth]{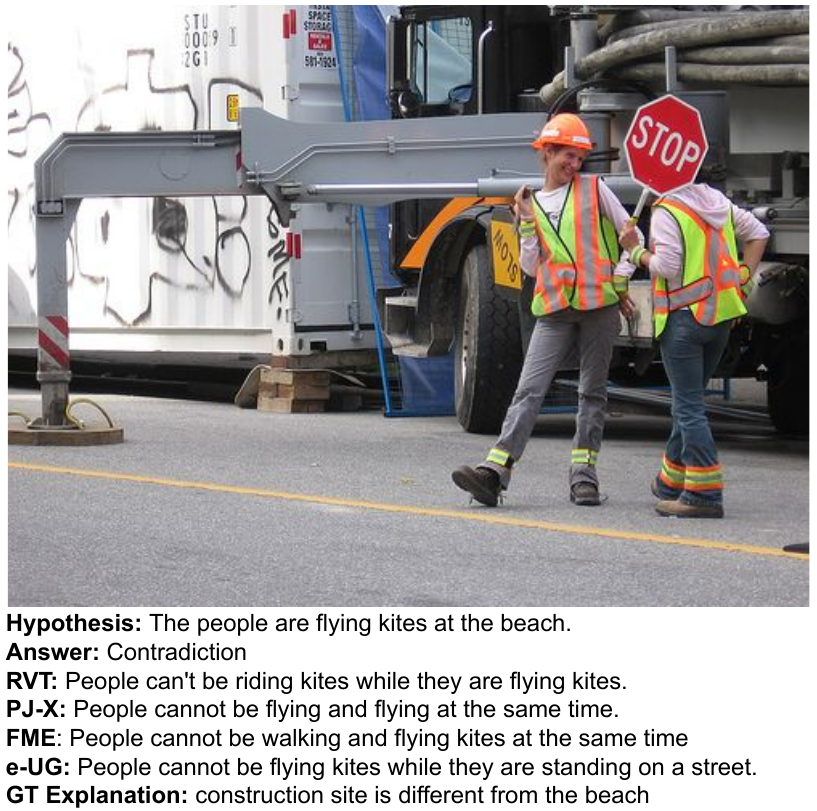}
    \caption{A test example from the e-SNLI-VE dataset~\cite{https://doi.org/10.48550/arxiv.2105.03761}. \emph{Contradiction} means the hypothesis contradicts the image content. \emph{RVT, PJ-X, FME, and e-UG} are explanation methods. They provide natural language explanations (NLEs). Explanations are not trained on any GT. (If they were, that would be another predictive model, and there is no guarantee it would explain the model's way of prediction.) They instead extract information from a vision-language (VL) model into a human language format. The \emph{GT Explanation} is a human-generated explanation for the answer collected by the authors. \textbf{Task}: Given an image and a hypothesis, decide if the hypothesis is aligned with the image. The machine also has to explain why they might be contradictory. VL-NLE models predict \emph{and} explain. Figure taken from~\cite{https://doi.org/10.48550/arxiv.2105.03761}.}
    \label{fig:evil}
\end{figure}

\begin{figure}
    \centering
     \begin{subfigure}[b]{0.29\textwidth}
         \centering
         \includegraphics[width=\textwidth]{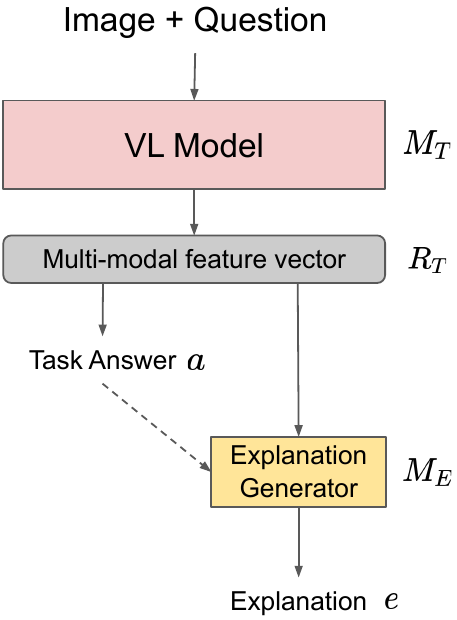}
     \end{subfigure}
     \hfill
     \begin{subfigure}[b]{0.69\textwidth}
         \centering
         \includegraphics[width=\textwidth]{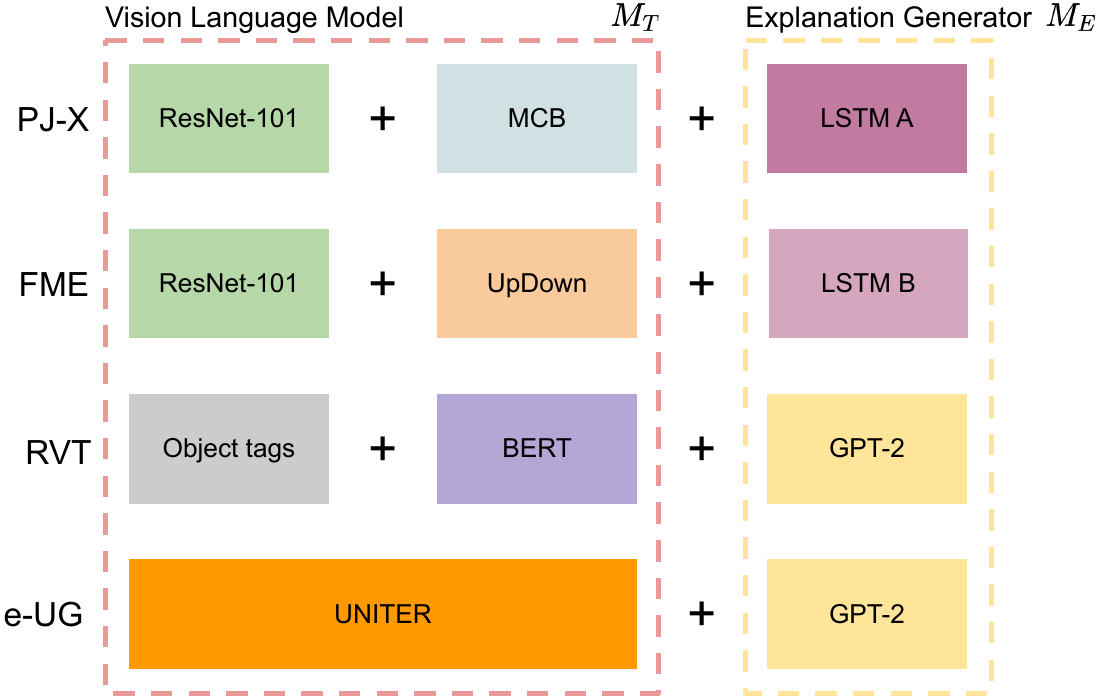}
     \end{subfigure}
    \caption{Overview of the structure of general VL models (left) and detailed subparts of individual models benchmarked in~\cite{https://doi.org/10.48550/arxiv.2105.03761}. Figure taken from~\cite{https://doi.org/10.48550/arxiv.2105.03761}.}
    \label{fig:architectures}
\end{figure}

\subsection{e-ViL}

e-ViL is an explainability benchmark introduced in the paper ``\href{https://arxiv.org/abs/2105.03761}{e-ViL: A Dataset and Benchmark for Natural Language Explanations in Vision-Language Tasks}''~\cite{https://doi.org/10.48550/arxiv.2105.03761}. A test example and the outputs of various VL models are given in Figure~\ref{fig:evil}. An overview of the architectures benchmarked in this work is shown in Figure~\ref{fig:architectures}.

\subsection{Summary of Interactive Explanations}

We only touched on interactive explanation techniques. These are very new, and there are only a few works. However, it has much potential. We recommend working in this domain. To work forward, we need to take humans into account -- whether XAI is helpful for the end user (HITL evaluation) and view XAI systems as socio-technical systems. We also want users to be able to interact with the explanation algorithm and to make the interface more natural for humans (e.g., having a natural-language-based ``chat'' about the explanation).

\section{Attribution to Model Parameters}

As we have seen, a model is a function that is an output of a training algorithm (which, in turn, is another function of the training data and other ingredients). The model takes training data as input implicitly through the training procedure. This is a hidden part of the model that is not used for explanations when we only focus on the attribution to the test sample features. It can very well happen that the model is making a bizarre decision not because of a specific feature in the test sample but because of strange (defective) training samples. It is difficult to rule this possibility out, and it is, therefore, meaningful to look at training samples.

We write the model prediction as a function of two variables:
\[Y = \operatorname{Model}(X; \theta) = \operatorname{Model}(X; \theta(\{z_1, \dots, z_n\}))\]
where \(Y\) is our prediction, \(X\) is the test input, \(\theta\) are the model parameters, and \(\{z_1, \dots, z_n\}\) is the training dataset. We use \(z\) because these can both correspond to inputs and input-output pairs. The prediction of our model is implicitly also a function of the training data.

As we discussed before, explaining our prediction against features of \(x\) is not always sufficient (but is very popular). We might also be interested in the contribution of
\begin{enumerate}
    \item individual parameters \(\theta_j\) of the model, and
    \item individual training samples \(z_i\) in the training set
\end{enumerate}
to the final prediction of the model. First, we look at the contribution of model parameters. Then, we discuss attribution methods to training samples.

\subsection{Explanation of Model Parameters \(\theta\)}

For DNNs, model parameters are simply millions of raw numbers. They are complicated to understand. Explaining a prediction \wrt these raw numbers is seemingly a tricky problem. This is in contrast with the input-level features \(x\) and labels \(y\). Inputs to a DNN are usually sensory data (image, sound, text), so humans can naturally understand them.

Thus, inputs and outputs to a DNN are often human-interpretable. However, the parameters are not, at least not directly. To understand the parameters \(\theta\), we ``project'' them onto the input space; i.e., we give visualizations of them (or explain them in text for NLP methods).

\begin{figure}
    \centering
    \includegraphics[width=0.5\linewidth]{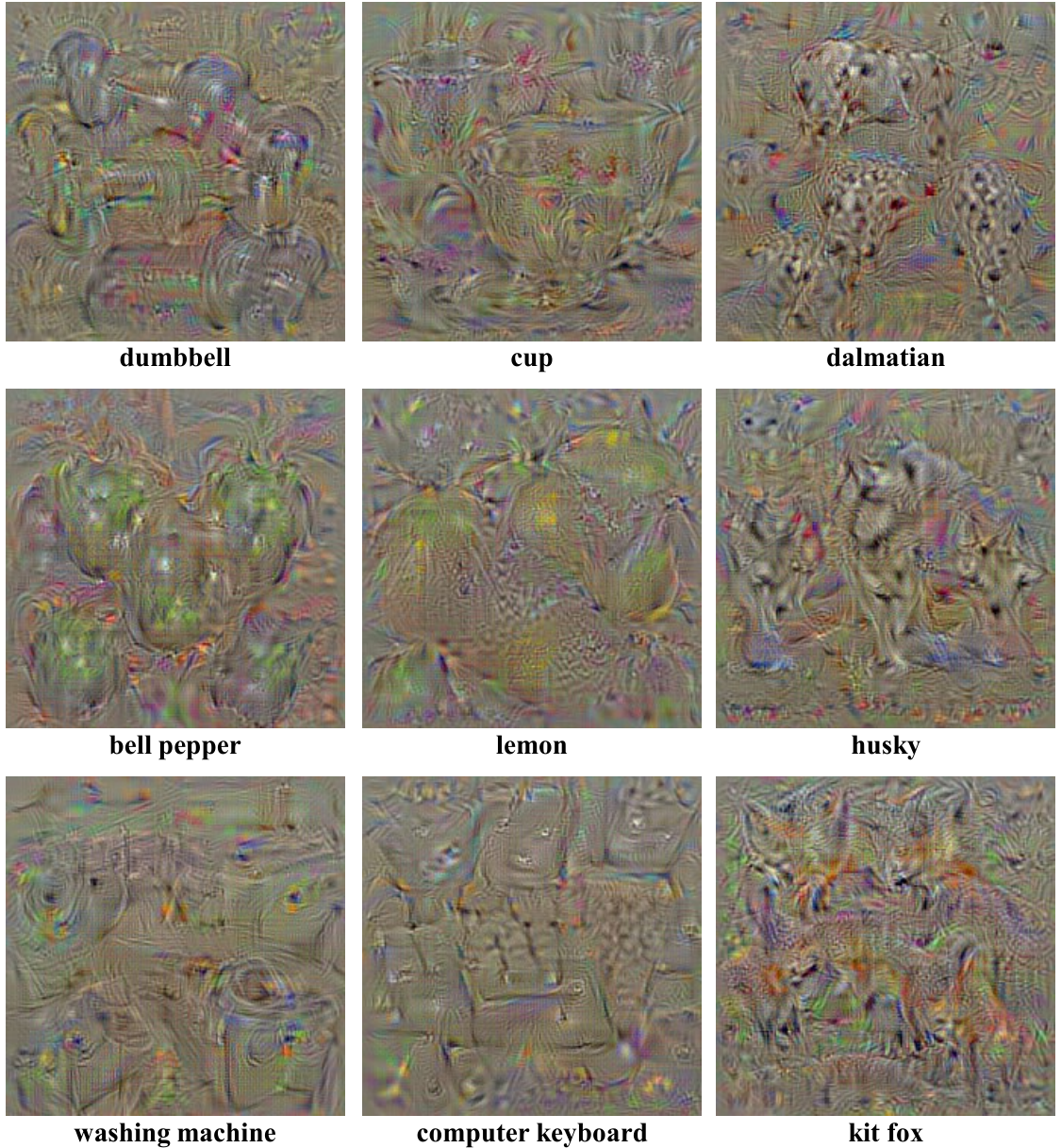}
    \caption{Various weight visualizations \wrt different target classes from~\cite{https://doi.org/10.48550/arxiv.1312.6034}. We ask ``What is the most likely image for the class dumbbell?'' from the model, or ``What excites a certain neuron most?'' One can employ several regularization techniques (e.g., TV) to make the visualizations more interpretable. For these samples, the model predicts a very high score for the respective classes. These are preliminary results from a seminal paper about turning model parameters into an image in the input space. Figure taken from~\cite{https://doi.org/10.48550/arxiv.1312.6034}.}
    \label{fig:param1}
\end{figure}

Examples for turning parameters into samples from the seminal paper ``\href{https://arxiv.org/abs/1312.6034}{Deep Inside Convolutional Networks: Visualising Image Classification Models and Saliency Maps}''~\cite{https://doi.org/10.48550/arxiv.1312.6034} are given in Figure~\ref{fig:param1}. We generate these samples by solving an optimization problem in the pixel space. We maximize the score for class \(c\) in the input space in a regularized fashion:
\[\argmax_I S_c(I) - \lambda\Vert I \Vert_2^2,\]
where \(S_c(I)\) is the prediction score (logit value, pre-activation of the output layer) for class \(c\) and image \(I\) from the network. \(L_2\) regularization prevents a small number of extreme pixel values from dominating the entire image. It results in smoother and more natural (more interpretable) images. We can also regularize the discrete image gradient (e.g., with the TV regularizer), which is also a popular choice. This mitigates the noise issue even more.\footnote{The \(L_2\)-regularized image can still be very noisy, just a bit less than the original because of the reduction in magnitude.} The objective of adversarial attack algorithms is very similar to this optimization problem. However, attacks try to minimize the score for a specific class. Here we are trying to maximize it, e.g., using gradient descent for the loss (less often used) or using gradient ascent for the logit value (popular).

\subsection{More examples of turning parameters into samples}

Let us discuss two more examples of turning parameters into samples.

\begin{figure}
    \centering
    \includegraphics[width=0.8\linewidth]{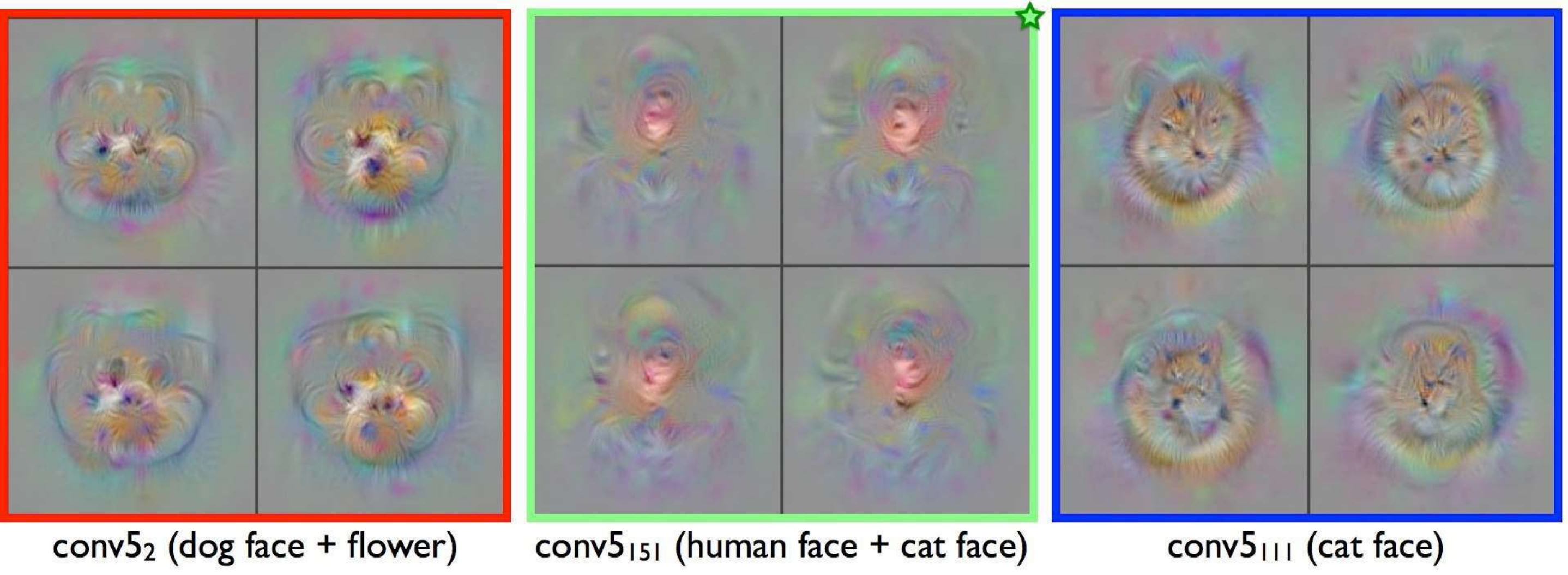}
    \caption{Visualizations of a deeper layer of an AlexNet-like architecture from~\cite{https://doi.org/10.48550/arxiv.1506.06579}. The synthesized images resemble a mixture of animals, flowers, and more abstract objects. The indices correspond to different feature channels of the conv5 pre-activation tensor. Each grid corresponds to four runs of the same optimization problem. Figure taken from~\cite{https://doi.org/10.48550/arxiv.1506.06579}.}
    \label{fig:more}
\end{figure}

First, we consider ``\href{https://arxiv.org/abs/1506.06579}{Understanding Neural Networks Through Deep Visualization}''~\cite{https://doi.org/10.48550/arxiv.1506.06579}. We can perform the previous optimization procedure on different intermediate layers as well. Instead of maximizing the score of a certain class, we maximize an intermediate feature activation for one of the units of a layer or maximize the entire layer's activation. We then recognize patterns in the generated images. These are interpreted as the patterns that the corresponding neurons have learned and respond to. The optimization problem here is
\[x^* = \argmax_x \left(a_i(x) - R_{\theta}(x)\right),\]
where \(a_i(x)\) can be an activation for a particular unit in a particular layer, or we can also maximize the mean, min, and max activation in a layer. That leads to similar results. (Not done in this work.) \(R_\theta(x)\) is the regularization term. In this work, the authors use
\[x \gets r_\theta \left(x + \eta \frac{\partial a_i(x)}{\partial x}\right),\]
which is more expressive. For example, for \(L_2\) decay one can choose \(r_\theta(x) := (1 - \theta)\cdot x\). An example collage is shown in Figure~\ref{fig:more}. Please refer to the full paper for various visualizations across many layers, which we discuss below.

When we maximize the output of an early neuron, its receptive field is usually smaller than the entire input image. Thus, when we visualize the optimized input, we will see only the small corresponding region changing in the input. The other input regions are left as we initialized them.

Typically, we will not see any interpretable pattern for many of the neurons. In many cases, people cherry-pick to generate these images. The results are also heavily dependent on the initial image of the optimization. One should be careful with how they interpret them.

At higher layers, we visualize more semantic content (e.g., cup, garbage bin, goose). To visualize entire layers, one can take one image for each channel in the corresponding layer's feature map output. As we go down the layers, we see more and more generic patterns. These are smaller, more common patterns that are found in many objects.

When we visualize the optimized inputs for the \emph{first} convolutional layer's neurons, we roughly see the filters (see \href{https://en.wikipedia.org/wiki/Gabor_filter}{Gabor filter}) of the corresponding channel for each neuron in that channel. These contain single colors or combinations of a few repetitive textures. If we use a single convolution, we just have a sparse linear network (\(S_c\) becomes linear \wrt \(I\)). The regularized activation-maximizing inputs are nearly the same as the filters themselves. If we take a channel of the filter (of shape \((3, H, W)\)) corresponding to the output channel of choice, we can directly visualize it. When doing so, we will see very similar visualizations to the visualizations of the regularized activation-maximizing inputs.

Consider a \(3 \times 3\) convolutional layer with a single channel. Then the filter \(K\) is of shape \((1, 3, 3, 3)\). The operation for a single neuron \(a\) in the output is simply
\[a = \sum \left(I^a \odot K\right),\]
where \(I^a\) is the receptive field of the neuron \(a\), of shape \((3, 3, 3)\). If we perform unregularized optimization, we obtain
\[I^{a*} = \bone(K > 0).\]
By using, e.g., \(L_2\) regularization, we roughly get \(I^{a*} \approx K\), with the outline of the generated image being the same as the corresponding filter channel.

\begin{figure}
    \centering
    \includegraphics[width=0.8\linewidth]{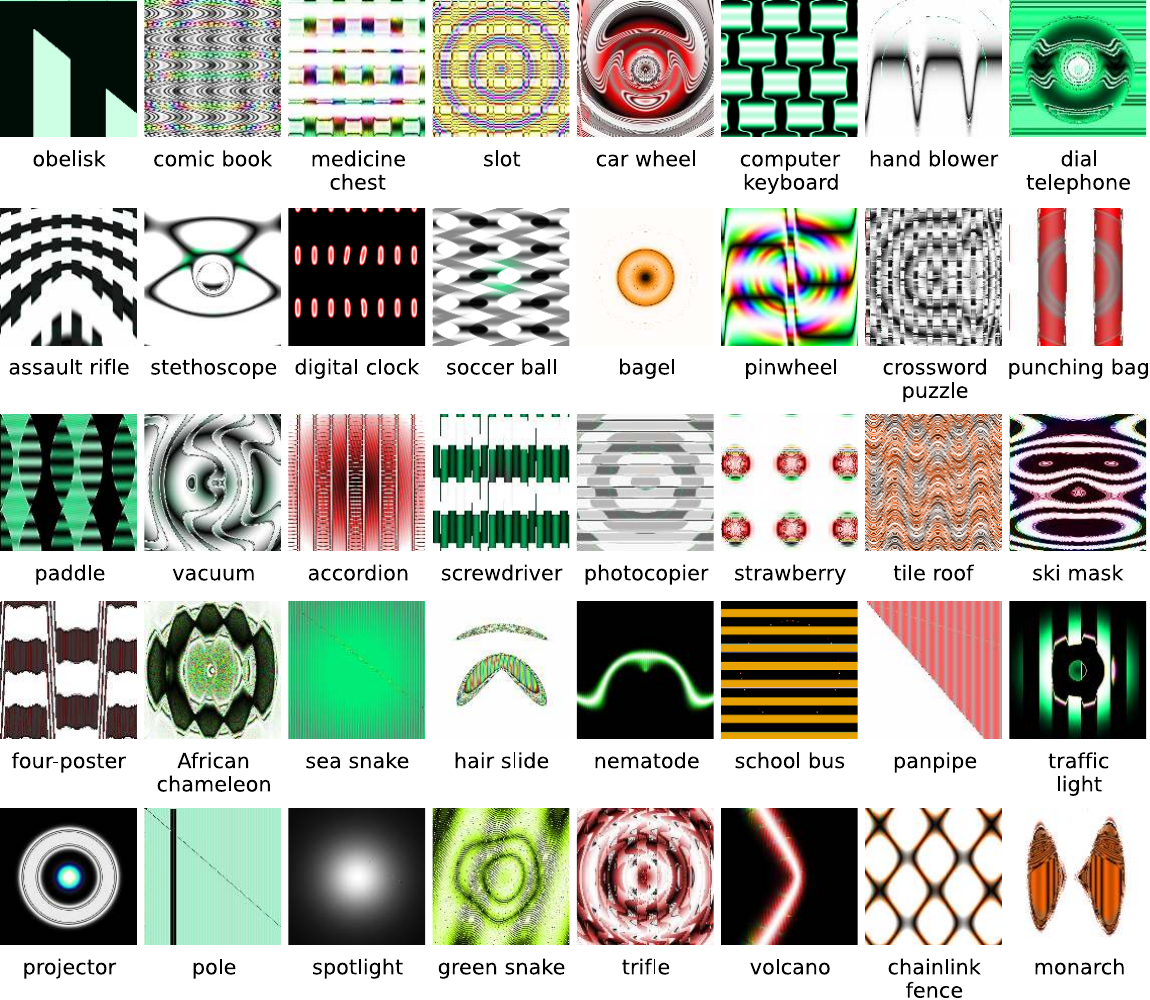}
    \caption{Visualizations of high-confidence images for different labels using the novel generation technique of~\cite{https://doi.org/10.48550/arxiv.1412.1897}. Figure taken from~\cite{https://doi.org/10.48550/arxiv.1412.1897}.}
    \label{fig:evenmore}
\end{figure}

As our second example, we look at ``\href{https://arxiv.org/abs/1412.1897}{Deep Neural Networks are Easily Fooled: High Confidence Predictions for Unrecognizable Images}''~\cite{https://doi.org/10.48550/arxiv.1412.1897}. One can use the previously introduced optimization problem in the pixel space to generate images corresponding to high activations (e.g., maximize prediction score for the class of choice). We get significantly different images depending on the regularization of the image generation (effectively, the search space). This work introduces a different generation technique that has astounding results. A teaser is shown in Figure~\ref{fig:evenmore}. Please refer to the paper for an extensive collection of visualizations.

The provided visualizations allow us to, e.g., look into the texture bias of the network. For example, the activation-maximizing input (in the modified search space) for ``baseball'' contains a very similar pattern as a baseball. However, we would not say that this is a baseball as humans. Nevertheless, the model predicts ``baseball'' with very high confidence. The technique allows us to see into the model's decision-making process, which might often be quite surprising.

In the paper's oral talk, the authors also showed that classifying images through their mobile phones gives the same result (they made an app for live demonstration). This shows that the visualizations are stable representations of the classes for the DNN in question.

The authors provide multiple visualization techniques. All visualizations correspond to highly confidently predicted generated images for the classes \(0-9\). We get astonishing results on even a simple dataset like MNIST.

The \href{https://distill.pub/2017/feature-visualization/}{linked resource} is a recommended blog post for feature visualization.

\subsection{Criticism of feature visualization}

While feature visualization can give impressive results (as can be seen in the Distill blog post), there is criticism about the utility of such methods. Even though we can find visualizations for units of a neural network that correspond to a human concept (cf. baseball example above), many visualizations are not interpretable~\cite{molnar2020interpretable}. We are not guaranteed to find something. 
Moreover, when we look at feature visualization as explanations to humans about what \emph{causes} a CNN to activate, Zimmermann \etal found that feature visualizations do not provide better insight into model behavior than e.g. looking at data samples directly~\cite{zimmermann2021well}. The visualizations are interesting but simpler approaches can provide the human with the same intuitions. 
A recent work~\cite{geirhos2023don} shows that feature visualizations are not reliable and can easily be fooled by an adversary while keeping the predictive performance of the model. They also prove that feature visualizations cannot guarantee to deliver an understanding of the model. 
Feature attribution gives an interesting tool for exploratory analysis, but it is not necessarily suitable as is for explaining model behavior to humans.

\section{Attribution to Training Samples}

This part is much more critical than attributing to individual weights, for reasons clarified just below.

\subsection{Why attribute to training samples?}

We saw that we had to eventually map our parameters onto the input space to visualize what was happening inside the model. It is a natural question to ask ``Why don't we just look at the raw ingredients for the parameters then?'' These are exactly the training samples. Model parameters are \emph{not interpretable}, so it makes sense to attribute \emph{directly} to the training samples. Model parameters \(\theta\) are also built on the training samples \(\{z_1, \dots, z_n\}\). Therefore, attributing to training samples is sufficient. Training data explanations are more likely to give more \emph{actionable} directions to improve our model. If we can trace the model's error or strange behavior back to the training samples, we can fix/add/remove training samples to resolve erratic behavior. If we find out that the model made a mistake because some of the labels were wrong, we can (1) relabel these samples that were attributed to the strange behavior, (2) remove them if the GT labeling \(P(Y \mid X = x)\) is too stochastic (faulty sample), or (3) we can even add new samples to the training set if we think there is no strong sample supporting the right behavior of the model we wish to see.

\subsection{Basic Counterfactual Question for Attribution to Training Data -- Influence Functions}

We look at influence functions, first used for deep learning in the paper ``\href{https://arxiv.org/abs/1703.04730}{Understanding Black-box Predictions via Influence Functions}''~\cite{https://doi.org/10.48550/arxiv.1703.04730}. These find influential training samples for the model prediction on a particular test sample. The influence here is an answer to the counterfactual question ``What happens to the current model prediction for a test input \(x\) if one training sample \(z_j\) was left out of the training set \(\{z_1, \dots, z_n\}\)?'' This is a minor change in the training set, as typically, the training set size is in the range of millions to billions. Leaving out one sample will generally not greatly affect the overall behavior of the model. However, for a particular test sample \(z\), we can still be interested in the training samples that made the largest impact on the test sample through the optimization procedure. Such training samples are likely to be visually similar to the test sample. We want an algorithm to measure the impact of each training sample on this particular test sample.

\subsubsection{Notation}

The notation is introduced in Table~\ref{tab:notation}. To find out \(L(z, \hat{\theta}_{\setminus j}) - L(z, \hat{\theta})\), we could retrain the model on the dataset without \(z_j\). However, this is infeasible for real-life scenarios. To study the impact of every training sample on every test sample, we would need to train (``number of training samples'' + 1) DNNs and evaluate the differences in the losses for all test samples we want to consider.

\textbf{Note}: We assume that \(\hat{\theta}\) and \(\hat{\theta}_{\setminus j}\) are \emph{global} minimizers of the respective empirical risks. This is a strong assumption, but we will see that relaxations of the resulting method still work well in practice.

\begin{table}
\centering
\caption{Notation used for discussing attribution to training samples. A large positive \(L(z, \hat{\theta}_{\setminus j}) - L(z, \hat{\theta})\) implies that \(z_j\) is very useful for test sample \(z\). Conversely, when it is a large negative number, \(z_j\) is highly harmful to \(z\).}
\label{tab:notation}
\renewcommand{\arraystretch}{1.5}
\resizebox{\textwidth}{!}{
\begin{tabular}{l l}
\toprule
\multicolumn{1}{l}{\textbf{Notation}} &  \multicolumn{1}{l}{\textbf{Description}} \\ 
\midrule
\(z = (x, y)\) & Test input and output of interest. \\
\(z_i = (x_i, y_i)\) & Training samples. \\
\(L(z, \theta)\) & Test loss for test sample \(z\) with parameters \(\theta\). \\
\(\frac{1}{n}\sum_i L(z_i, \theta)\) & Training loss (empirical risk) for training sample \(z_i\). \\
\(\hat{\theta} := \argmin_{\theta} \frac{1}{n} \sum_i L(z_i, \theta)\) & Solution parameters for the original training set. \\
\(\hat{\theta}_{\setminus j} := \argmin_{\theta}\frac{1}{n} \sum_{i \ne j} L(z_i, \theta)\) & Solution for the training set after removing \(z_j\). \\
\(L(z, \hat{\theta}_{\setminus j}) - L(z, \hat{\theta})\) & Change in the test loss for \(z\) after removing training sample \(z_j\). \\
\bottomrule
\end{tabular}}
\end{table}

\subsubsection{Lesson from attribution methods: take the gradient!}

We only had to do a single backpropagation to determine the contribution of all pixels modified by an infinitesimal amount (separately) to the infinitesimal change in the output. Here, we also take the gradient of the test loss \(L(z, \hat{\theta})\) \wrt the training sample \(z_j\), where the two values are connected through the entire optimization procedure. In particular,
\[\hat{\theta} = \argmin_\theta \frac{1}{n}\sum_i L(z_i, \theta).\]
\(\hat{\theta}\) is, therefore, a function of \(z_j\) through the optimization we employ, and the dependency between the test loss and \(z_j\) is exactly through \(\hat{\theta}\). We are interested in the change in the test loss when we make a small change in the training sample \(z_j\).

We need a few tricks to compute the gradient
\[\frac{\partial L(z, \hat{\theta})}{\partial z_j}.\]
Taking the gradient through an optimization procedure has been done in subparts of ML quite a few times. There are algorithms like ``gradient descent by gradient descent''~\cite{https://doi.org/10.48550/arxiv.1606.04474}.\footnote{This algorithm aims to find the optimal LR without cross-validation through another GD algorithm. Optimal here means good for generalization to the held-out validation set. For this, we also need backpropagation through the optimization procedure.} This is overall a great technique to know.

First, we generalize the notion of ``removal'' into a continuous procedure. Removing \(z_j\) is a discrete procedure and is thus non-differentiable. Instead, we take the loss of \(z_j\) into a separate term:
\[\hat{\theta}_{\epsilon, j} := \argmin_{\theta} \frac{1}{n} \sum_i L(z_i, \theta) + \epsilon L(z_j, \theta).\]
Note that the first term still contains a \(z_j\) term.
\begin{itemize}
    \item \(\epsilon = 0 \in \nR\): We recover the original minimizer of the training loss, \(\hat{\theta}_{0, j} = \hat{\theta} \in \nR^d\).
    \item \(\epsilon = -1/n\): We obtain our previous notion of ``removal''.
\end{itemize}
We further assume that the loss \(L\) is twice differentiable and strictly convex \wrt \(\theta\) (so that the Hessian matrix of the model parameters is PD). For DNNs, there is usually no unique \(\hat{\theta}_{\epsilon, j}\) and \(\hat{\theta}\) because of weight space symmetries and other contributing factors that make the loss landscape highly non-convex, with many equally good minima. So we further enforce strict convexity to have a unique optimum. This is a rather typical trick in research: We assume that everything is simple during theoretical derivations. In practice, we ignore the assumptions and hope our method still works.

One can obtain the following derivative (with annotated shapes for clarity):
\[\underbrace{\restr{\frac{\partial \hat{\theta}_{\epsilon, j}}{\partial \epsilon}}{\epsilon = 0}}_{\in \nR^d} \approx -\underbrace{H_{\hat{\theta}}^{-1}}_{\in \nR^{d \times d}} \underbrace{\nabla_\theta L(z_j, \hat{\theta})}_{\in \nR^d}.\]
where
\[H_{\hat{\theta}} = \frac{1}{n}\sum_i \nabla^2_\theta L(z_i, \hat{\theta}).\]
In words, \(\restr{\frac{\partial \hat{\theta}_{\epsilon, j}}{\partial \epsilon}}{\epsilon = 0}\) is the derivative of the weights \wrt \(\epsilon\), evaluated at \(\epsilon = 0\) when \(\hat{\theta}_{\epsilon, j} = \hat{\theta}\). This gives the relative change in the globally optimal weights using the original objective if we change the additional influence of \(z_j\) by an infinitesimal amount from 0. The last term is the gradient of \(z_j\) loss \wrt \(\theta\), evaluated for weights \(\hat{\theta}\). \(\nabla^2_\theta L(z_i, \hat{\theta})\) is the Hessian matrix of \(L\):
\[\left(\nabla^2_\theta L(z_i, \hat{\theta})\right)_{ij} = \frac{\partial^2 L(z_i, \hat{\theta})}{\partial \theta_i \partial \theta_j}.\]
Why is the derivative formula well-defined, i.e., why is this average Hessian matrix invertible? It is a well-known fact that the average of symmetric, positive definite (PD) matrices is symmetric PD. The Hessians are symmetric because of Schwarz's theorem (the loss has continuous second partial derivatives \wrt \(\theta\) everywhere). The Hessians are also PD (i.e., they only have positive eigenvalues, and there is strictly positive curvature in all directions) because the function is strictly convex by assumption. Therefore, \(H_{\hat{\theta}}\) is symmetric PD.

\begin{information}{Interpreting the Hessian}
If we have \(10^6\) parameters, then \(\nabla_\theta L(z_i, \hat{\theta}) \in \nR^{10^6}\) gives us how the function value changes in each principal axis direction \emph{relative} to an infinitesimal change. To obtain the relative change in the loss value in a particular input direction \(v\), one can consider
\[\nabla_\theta L(z_i, \hat{\theta})^\top v \in \nR,\quad \Vert v \Vert = 1.\]
Similarly, \(\nabla^2_\theta L(z_i, \hat{\theta}) \in \nR^{10^6 \times 10^6}\) gives us how the gradient of the loss at \(z_i\) changes in the neighborhood of \(\theta\) along all canonical axes. This is why it is a matrix. In each axis direction, we measure the relative change in the gradient vector (in each of its entries) \wrt each principal axis. To get the rate of change of the gradient (curvature) in a particular input direction \(v\), one can consider
\[v^\top \nabla^2_\theta L(z_i, \hat{\theta}) v \in \nR,\quad \Vert v \Vert = 1.\]
When Hessians are symmetric (which is almost always the case in ML settings), they are orthogonally diagonalizable. In this case, the diagonal entries of the diagonalized Hessian give the rate of change of the gradient (curvature) in the eigenvector directions. Let
\[\nabla^2_\theta L(z_i, \hat{\theta}) = Q \Lambda Q^\top\]
where \(\Lambda\) is diagonal and \(Q\) is orthogonal. Then, if \(v_i\) is the \(i\)th eigenvector direction, we have
\[v_i^\top \nabla^2_\theta L(z_i, \hat{\theta})v_i = v_i^\top Q \Lambda Q^\top v_i = v_i^\top Q \Lambda e_i = \lambda_i v_i^\top Q e_i = \lambda_i v_i^\top v_i = \lambda_i.\]
\end{information}

Our story does not end here, as we wish to see the influence of \(z_j\) on the test loss for test sample \(z\). Given the previous result, we compute the influence of sample \(z_j\) on the loss for test sample \(z\) as \(\text{IF}(z_j, z) \in \nR\),
\begin{align*}
\text{IF}(z_j, z) &:= \restr{\frac{\partial L(z, \hat{\theta}_{\epsilon, j})}{\partial \epsilon}}{\epsilon = 0}\\
&= \nabla_\theta L(z, \hat{\theta})^\top \restr{\frac{\partial \hat{\theta}_{\epsilon, j}}{\partial \epsilon}}{\epsilon = 0}\\
&= -\nabla_\theta L(z, \hat{\theta})^\top H_{\hat{\theta}}^{-1} \nabla_\theta L(z_j, \hat{\theta}).
\end{align*}
This is the formulation for IF that the referenced paper uses. The IF value gives the relative change in the test loss value if we increase \(\epsilon\) by an infinitesimal amount from \(\epsilon = 0\). This ``upweighing'' represents the removal of $z_i$ from the loss computation. It is large and positive when upweighting \(z_j\) a bit increases the loss by a lot (harmful) \(\iff\) when downweighting \(z_j\) a bit decreases the loss by a lot. It is large and negative when upweighting \(z_j\) a bit decreases the loss significantly (helpful). This formulation refers to \emph{negative influence}. \emph{As we would intuitively expect a high influence value for a sample that decreases the loss a lot, both this book and the Arnoldi paper~\cite{https://doi.org/10.48550/arxiv.2112.03052} consider the definition}
\[\text{IF}(z_j, z) = \nabla_\theta L(z, \hat{\theta})^\top H_{\hat{\theta}}^{-1} \nabla_\theta L(z_j, \hat{\theta}).\]

Let us consider some remarks. Using a first-order Taylor approximation, it is also clear that
\begin{align*}
L(z, \hat{\theta}_{\epsilon, j}) &= L(z, \hat{\theta}) + \epsilon \restr{\frac{\partial L(z, \hat{\theta}_{\epsilon, j})}{\partial \epsilon}}{\epsilon = 0} + o(\epsilon).
\end{align*}
One can study the behavior of the test loss when perturbing sample \(z_j\) by an infinitesimal amount in a clear way using the above formula. The IF formula is also very symmetrical: we are taking a modified dot product between the gradient of loss of the test sample and the training sample.

From now on, we will use the latter definition for the influence function (without the negative sign). We have discussed that \(H_{\hat{\theta}}\) is symmetric and PD. Therefore, it can be orthogonally diagonalized, i.e., we can find a rotation/mirroring such that in this new basis, the average Hessian on the training points is a diagonal matrix:
\[H_{\hat{\theta}} = Q \Lambda Q^\top\]
with an orthogonal matrix \(Q\) (with ortho\emph{normal} columns), and its inverse is given by
\[H_{\hat{\theta}}^{-1} = Q \Lambda^{-1} Q^\top.\]
Therefore,
\begin{align*}
\text{IF}(z_j, z) &= \nabla_\theta L(z, \hat{\theta})^\top H_{\hat{\theta}}^{-1} \nabla_\theta L(z_j, \hat{\theta})\\
&= \nabla_\theta L(z, \hat{\theta})^\top Q \Lambda^{-1} Q^\top \nabla_\theta L(z_j, \hat{\theta})\\
&= \left(Q^\top \nabla_\theta L(z, \hat{\theta})\right)^\top \Lambda^{-1} \left(Q^\top \nabla_\theta L(z_j, \hat{\theta})\right)\\
&= \left\langle Q^\top \nabla_\theta L(z, \hat{\theta}), Q^\top \nabla_\theta L(z_j, \hat{\theta}) \right\rangle_{\Lambda^{-1}}.
\end{align*}
To calculate \(IF(z_j, z)\), we rotate/mirror the gradient vectors to transform them into the eigenbasis of the average Hessian, then compute a generalized dot product between the gradients expressed in the eigenbasis, weighted by the corresponding diagonal entries of \(\Lambda^{-1}\) (the inverse curvatures in each direction of the eigenbasis). The dot product is, therefore, calculated in a \emph{distorted space}, where directions with the flattest curvature in the loss landscape are given more weights. To get a high influence value, having large positive values in these directions in the gradient vectors expressed in the eigenbasis is more important.

The caveat is that \emph{we might have millions or billions of parameters}. Let \(p := \text{number of parameters} = \cO(\text{millions-billions})\) and \(n := \text{number of training samples} = \cO(\text{millions-billions})\). Then, the naive \(H_{\hat{\theta}}^{-1}\) computation is \(\cO(np^2 + p^3)\) where the \(np^2\) part corresponds to computing \(H_{\hat{\theta}}\) and \(p^3\) corresponds to computing its inverse. Computing \(H_{\hat{\theta}}^{-1}\) dominates the IF computation when \(n\) is not significantly larger than \(p\). In practice, naive computation is prohibitive and infeasible.

\begin{information}{Proof of the Derivative Formula}
We start with the definitions
\[\hat{\theta} = \argmin_\theta \frac{1}{n}\sum_i L(z_i, \theta)\]
and
\[\hat{\theta}_{\epsilon, j} = \argmin_\theta \frac{1}{n} \sum_i L(z_i, \theta) + \epsilon L(z_j, \theta).\]
Following the strict convexity assumption, both of these values are unique (we consider \(\epsilon > -1/n\) s.t. all terms in the sum are strictly convex).
Fermat's theorem tells us that every extremum of a differentiable function is a stationary point. Thus, a necessary condition for the optimality of a differentiable function is that the gradient at the optimum must be 0. (This is not sufficient, however: stationary points can also be maxima and saddle points.) Therefore, the previous optimality assumptions imply
\[\nabla_\theta \left(\frac{1}{n}\sum_i L(z_i, \hat{\theta})\right) = \frac{1}{n} \sum_i \nabla_\theta L(z_i, \hat{\theta}) = 0\]
(because the gradient is a linear operator) and
\[\nabla_\theta \left(\frac{1}{n}\sum_i L(z_i, \hat{\theta}_{\epsilon, j}) + \epsilon L(z_j, \hat{\theta}_{\epsilon, j})\right) = \frac{1}{n}\sum_i \nabla_\theta L(z_i, \hat{\theta}_{\epsilon, j}) + \epsilon \nabla_\theta L(z_j, \hat{\theta}_{\epsilon, j}) = 0.\]
These are ingredients (1) and (2).

\medskip

We also make use of the Implicit Function Theorem. \(\hat{\theta}_{\epsilon, j}\) is differentiable \wrt \(\epsilon\) at \(\epsilon = 0\). (The optimal point of the modified loss is also differentiable \wrt another variable of that function.) Therefore, one can consider the first-order Taylor expansion (by linearizing \(\hat{\theta}_{\epsilon, j}\) in \(\epsilon\) around \(\epsilon = 0\)):
\[\hat{\theta}_{\epsilon, j} = \underbrace{\hat{\theta}}_{\restr{\hat{\theta}_{\epsilon, j}}{\epsilon = 0}} + \epsilon \underbrace{\restr{\frac{\partial \hat{\theta}_{\epsilon, j}}{\partial \epsilon}}{\epsilon = 0}}_{\in \nR^d} + o(\epsilon).\]
This is ingredient (3).
\begin{itemize}
    \item We use linearization often, just like when attributing to test input features.
    \item \(f(\epsilon) = f(0) + (\epsilon - 0) \cdot \frac{\partial f(0)}{\partial \epsilon} + o(\epsilon)\) is the Taylor expansion of \(f(\epsilon) := \hat{\theta}_{\epsilon, j}\) around \(\epsilon = 0\).
    \item \(o(\epsilon)\) specifies \(\lim_{\epsilon \rightarrow 0} \frac{R_1(x)}{\epsilon} = 0\). The remainder term converges to \(0\) faster than \(\epsilon\) itself.
\end{itemize}

\medskip

We compute \(\nabla_\theta L(z_i, \hat{\theta}_{\epsilon, j})\) in terms of \(\nabla_\theta L(z_i, \hat{\theta})\) as follows (by plugging in ingredient (3)). We calculate the Taylor expansion of \(\nabla_\theta L(z_i, \hat{\theta}_{\epsilon, j})\) in \(\theta\), around \(\hat{\theta}\).
\begin{align*}
\nabla_\theta L(z_i, \hat{\theta}_{\epsilon, j}) &= \nabla_\theta L \left(z_i, \hat{\theta} + \epsilon \restr{\frac{\partial \hat{\theta}_{\epsilon, j}}{\partial \epsilon}}{\epsilon = 0} + o(\epsilon)\right)\\
&= \nabla_\theta L(z_i, \hat{\theta}) + \nabla^2_\theta L(z_i, \hat{\theta}) \left(\epsilon \restr{\frac{\partial \hat{\theta}_{\epsilon, j}}{\partial \epsilon}}{\epsilon = 0} + o(\epsilon)\right) + o(\epsilon)\\
&= \nabla_\theta L(z_i, \hat{\theta}) + \epsilon \underbrace{\nabla^2_\theta L(z_i, \hat{\theta})}_{\in \nR^{d \times d}} \underbrace{\restr{\frac{\partial \hat{\theta}_{\epsilon, j}}{\partial \epsilon}}{\epsilon = 0}}_{\in \nR^d} + o(\epsilon)
\end{align*}
\begin{itemize}
    \item This formulation is, of course, given to later get rid of the \(o(\epsilon)\) terms and provide an approximation. The approximation is justified because \(\hat{\theta}_{\epsilon, j}\) is very similar to \(\hat{\theta}\) anyways for small \(\epsilon\). The difference is
    \[\epsilon \restr{\frac{\partial \hat{\theta}_{\epsilon, j}}{\partial \epsilon}}{\epsilon = 0} + o(\epsilon).\]
\end{itemize}
We plug \[\nabla_\theta L(z_i, \hat{\theta}_{\epsilon, j}) = \nabla_\theta L(z_i, \hat{\theta}) + \epsilon \nabla^2_\theta L(z_i, \hat{\theta}) \restr{\frac{\partial \hat{\theta}_{\epsilon, j}}{\partial \epsilon}}{\epsilon = 0} + o(\epsilon)\]
into the second ingredient
\[\frac{1}{n}\sum_i \nabla_\theta L(z_i, \hat{\theta}_{\epsilon, j}) + \epsilon \nabla_\theta L(z_j, \hat{\theta}_{\epsilon, j}) = 0.\]
This results in
\begin{align*}
&\frac{1}{n}\sum_i \left( \nabla_\theta L(z_i, \hat{\theta}) + \epsilon \nabla^2_\theta L(z_i, \hat{\theta}) \restr{\frac{\partial \hat{\theta}_{\epsilon, j}}{\partial \epsilon}}{\epsilon = 0} + o(\epsilon) \right)\\
&\hspace{2.8em}+ \epsilon \left(\nabla_\theta L(z_j, \hat{\theta}) + \epsilon \nabla^2_\theta L(z_j, \hat{\theta}) \restr{\frac{\partial \hat{\theta}_{\epsilon, j}}{\partial \epsilon}}{\epsilon = 0} + o(\epsilon) \right) = 0\\
&\iff \frac{1}{n} \sum_i \nabla_\theta L(z_i, \hat{\theta}) + \epsilon \frac{1}{n}\sum_i\nabla^2_\theta L(z_i, \hat{\theta})\restr{\frac{\partial \hat{\theta}_{\epsilon, j}}{\partial \epsilon}}{\epsilon = 0}\\
&\hspace{2.8em}+ \epsilon \nabla_\theta L(z_j, \hat{\theta}) + \epsilon^2 \nabla^2_\theta L(z_j, \hat{\theta})\restr{\frac{\partial \hat{\theta}_{\epsilon, j}}{\partial \epsilon}}{\epsilon = 0} + (\epsilon + 1) o(\epsilon) = 0\\
&\iff \frac{1}{n} \sum_i \nabla_\theta L(z_i, \hat{\theta}) + \epsilon \frac{1}{n}\sum_i\nabla^2_\theta L(z_i, \hat{\theta})\restr{\frac{\partial \hat{\theta}_{\epsilon, j}}{\partial \epsilon}}{\epsilon = 0} + \epsilon \nabla_\theta L(z_j, \hat{\theta}) + o(\epsilon) = 0\\
&\overset{(1)}{\iff} \epsilon \frac{1}{n}\sum_i\nabla^2_\theta L(z_i, \hat{\theta})\restr{\frac{\partial \hat{\theta}_{\epsilon, j}}{\partial \epsilon}}{\epsilon = 0} + \epsilon \nabla_\theta L(z_j, \hat{\theta}) + o(\epsilon) = 0\\
&\iff \epsilon H_{\hat{\theta}}\restr{\frac{\partial \hat{\theta}_{\epsilon, j}}{\partial \epsilon}}{\epsilon = 0} + \epsilon \nabla_\theta L(z_j, \hat{\theta}) + \underbrace{o(\epsilon)}_{\lim_{\epsilon \rightarrow 0} \frac{R_1(\epsilon)}{\epsilon} = 0} = 0\\
&\iff H_{\hat{\theta}}\restr{\frac{\partial \hat{\theta}_{\epsilon, j}}{\partial \epsilon}}{\epsilon = 0} + \nabla_\theta L(z_j, \hat{\theta}) + \underbrace{\frac{o(\epsilon)}{\epsilon}}_{\lim_{\epsilon \rightarrow 0} \frac{R_1(\epsilon)}{\epsilon^2} = 0} = 0\\
&\iff H_{\hat{\theta}}\restr{\frac{\partial \hat{\theta}_{\epsilon, j}}{\partial \epsilon}}{\epsilon = 0} + \nabla_\theta L(z_j, \hat{\theta}) + o(\epsilon^2) = 0\\
&\overset{\epsilon \text{ small}}{\implies} H_{\hat{\theta}}\restr{\frac{\partial \hat{\theta}_{\epsilon, j}}{\partial \epsilon}}{\epsilon = 0} + \nabla_\theta L(z_j, \hat{\theta}) \approx 0\\
&\iff \restr{\frac{\partial \hat{\theta}_{\epsilon, j}}{\partial \epsilon}}{\epsilon = 0} \approx  -H_{\hat{\theta}}^{-1}\nabla_\theta L(z_j, \hat{\theta})
\end{align*}
\end{information}

Generally, the focus of research in influence function computation is how to speed things up while keeping approximations accurate. Interestingly, one can speed things up a lot.

\subsection{LISSA}

LISSA~\cite{agarwal2017secondorder} is a method the authors of ``\href{https://arxiv.org/abs/1703.04730}{Understanding Black-box Predictions via Influence Functions}''~\cite{https://doi.org/10.48550/arxiv.1703.04730} use to keep the inverse average Hessian calculation tractable. The LISSA algorithm uses an iterative approximation to approximate the inverse Hessian vector product (iHVP):
\[H_{\hat{\theta}}^{-1}\nabla_\theta L(z, \hat{\theta}).\]
(Note the symmetricity of the average Hessian matrix.)
For each test point of interest, they can precompute the above vector, and then they can efficiently compute the dot product between it and
\[\nabla_\theta L(z_i, \hat{\theta})\]
for each training sample \(z_i\). This also helps with the quadratic scaling of the size of the average Hessian, as instead of computing the inverse average Hessian directly, they approximate the Matrix-Vector (MV) product through the iterative procedure.

The iterative approximation uses the fact that
\[A^{-1} = \sum_{k = 0}^\infty (I - A)^k\]
for an invertible matrix \(A\) with all eigenvalues bounded below \(1\). At the small cost of inaccuracy, we gain a lot of speedup by this. The authors have another speedup by subsampling the training data in the summation (like how we do SGD-based optimization):
\[H_{\hat{\theta}} \approx \frac{1}{|I|} \sum_{i \in I} \nabla^2_\theta L(z_i, \hat{\theta}).\]
Averaging Hessians through all training samples is infeasible. If we have a good representation of our training samples, we do not need to do a complete pass through the training samples. Random samples very likely give a good representation.

The final procedure estimates the inverse Hessian-Vector Product (HVP) as
\[H_i^{-1}v = v + (I - H_{\hat{\theta}})H_{i - 1}^{-1}v,\]
where \(H_{\hat{\theta}}\) is approximated on random batches (of size one or a small enough size), \(v = \nabla_\theta L(z, \hat{\theta})\), and \(i \in [t]\) is a particular iteration of the method (\(H_0^{-1}v = v\)). Using this technique, the authors reduce the time complexity of computing IF(\(z_j, z\)) for all training points and a single test point to \(\cO(np + rtp)\) where \(r\) is the number of independent repeats of the iterative HVP calculation (where they average the results from the \(r\) runs) and \(t\) is the number of iterations.

\textbf{Note}: LISSA already existed before the seminal IF paper -- the authors adapted it to their method.

\subsection{Arnoldi}

Arnoldi, introduced in the paper ``\href{https://arxiv.org/abs/2112.03052}{Scaling Up Influence Functions}''~\cite{https://doi.org/10.48550/arxiv.2112.03052}, is a method for speeding up influence function calculations and reducing its memory requirements. Calculating and keeping a billion-dimensional vector (number of parameters) \(H_{\hat{\theta}}^{-1}\nabla_\theta L(z, \hat{\theta})\) in memory is still very restrictive, and very coarse approximations (e.g., considering only a subset of parameters) are needed. If we consider the diagonalized formula for IF, \(H_{\hat{\theta}}\) is written as
\[H_{\hat{\theta}} = Q \Lambda Q^\top\]
and the formula is
\[\operatorname{IF}(z_j, z) = \left\langle Q^\top \nabla_\theta L(z, \hat{\theta}), Q^\top \nabla_\theta L(z_j, \hat{\theta}) \right\rangle_{\Lambda^{-1}}.\]
Here, \(Q \in \nR^{p \times p}\) which is infeasibly large for efficient use. Setting \(G = Q^\top\) to contain the \(k\) eigenvectors of \(H_{\hat{\theta}}\) that correspond to its largest eigenvalues as rows (i.e., it is the projection matrix onto the span of the ``top \(k\) eigenvectors''), we obtain
\[\operatorname{IF}(z_j, z) = \left\langle G \nabla_\theta L(z, \hat{\theta}), G \nabla_\theta L(z_j, \hat{\theta}) \right\rangle_{\Lambda_k^{-1}}\]
where \[H_{\hat{\theta}} \approx G^\top \Lambda_k G.\]
By using this formulation, we map the gradients to a much lower-dimensional (\(k\)) space, and computations (dot product) become notably faster. \(G\) and \(\Lambda_k\) are calculated once and then cached.

The top \(k\) eigenvalues of \(H_{\hat{\theta}}\) are the smallest \(k\) eigenvalues of its inverse, thus we take the eigenvalues that have the least influence in calculating IF. Very curiously, the authors report that selecting the top \(k\) eigenvalues of the inverse (corresponding to the dominant terms of the dot product) performs worse. DNN loss landscapes are highly non-convex, and the Hessian can have negative eigenvalues. The authors select the top \(k\) eigenvalues \emph{in absolute value}.

The actual Arnoldi method is much more detailed and sophisticated in obtaining the referenced matrices, but the main idea is the same as was introduced here.

\begin{information}{Using a Subset of the Parameters}
Instead of the entire model, one can also use only the final or initial layers for \(\theta\). This dramatically reduces \(p\) by orders of magnitude. It has two drawbacks: the choice of layers becomes a hyperparameter, and the viable values of the number of parameters kept
will depend on the model architecture. Using just one layer can result in
different influence estimates than those based on the whole model. This is deemed suboptimal and is not used in Arnoldi but was used in earlier work, e.g.,~\cite{https://doi.org/10.48550/arxiv.1703.04730}.
\end{information}

\begin{information}{Reducing the Search Space}
The following speedup is also compatible with Arnoldi, although the authors do not use it. (They use Arnoldi for retrieval of wrong labels. It is not aligned with the goal.) It is used in \href{https://aclanthology.org/2021.emnlp-main.808/}{FastIF}.

\medskip

\(\text{IF}(z_j, z)\) is already quite expensive. We should not calculate it for all \(j\). The end goal is usually to retrieve influential training samples \(z_j\) for the test sample \(z\). Instead of computing \(\text{IF}(z_j, z)\) for all training samples \(z_j\), we first reduce the search space (for candidate training samples that are likely to influence our test samples) via cheap, approximate search. This greatly reduces computational load. For example, we can perform \(L_2\)-distance \(k\)-NN using the last layer features to retrieve candidates. This is a typical trick for deep metric learning: We take the last layer activations from a network as a good representation of our sample in a lower-dimensional space. We compute the Euclidean distance in this space and retrieve the top \(k\) semantically similar samples from the training set to the test sample. Instead of taking the top \(k\) samples, we could also threshold by the \(L_2\) radius. Both ways reduce the search space by a lot; thus, they reduce computational time.
\end{information}

\subsection{LISSA vs. Arnoldi}

\begin{table}
    \centering
    \caption{``Retrieval of mislabeled MNIST examples using self-influence for larger CNN. For TracIn the
    $C$ value is in brackets (last or all). All methods use full models (except the LISSA run on 10\% of parameters $\Theta$).''~\cite{https://doi.org/10.48550/arxiv.2112.03052} TracIn[10] gives the best results while staying feasible to compute. RandProj is also a surprisingly strong method. Table is adapted from~\cite{https://doi.org/10.48550/arxiv.2112.03052}.}
    \label{tab:results2}
    \begin{tabular}{lrrrr}
    \toprule
      Method &    $\tilde{p}$ &    $T$, secs & AUC & AP \\
    \midrule
    LISSA, $r=10$ & - & 4900 & 98.9 & 95.0 \\
    LISSA, $r=100$ (10\% $\Theta$) & - & 32300 & 98.8 & 94.8 \\
    \midrule
    TracIn[1] & - & 5 & 98.7 & 94.0 \\
    TracIn[10] & - & 42 & \textbf{99.7} & \textbf{98.7} \\
    \midrule
    RandProj & 10 & 0.2 & 97.2 & 87.7 \\
    RandProj & 100 & 1.9 & 98.6 & 93.9 \\
    \midrule
    RandSelect & 10 & 0.1 & 54.9 & 31.2 \\
    RandSelect & 100 & 1.8 & 91.8 & 72.6 \\
    \midrule
    Arnoldi & 10 & 0.2 & 95.0 & 84.0 \\
    Arnoldi & 100 & 1.9 & 98.2 & 92.9 \\
    \bottomrule
    \end{tabular}
\end{table}

Results~\cite{https://doi.org/10.48550/arxiv.2112.03052} of Arnoldi and various other methods are given in Table~\ref{tab:results2}. The Arnoldi authors compute AUC and AP for the retrieval of wrong labels. They try to retrieve the wrongly put labels in the training set using self-influence. The task is not exactly aligned with removing training samples and retraining (precise estimation of IF) -- this is why RandProj can also perform quite well. In fact, it performs better than Arnoldi. (It does not need to give precise IF estimates!) In RandProj, \(G\) is a random Gaussian matrix: It does not correspond to the eigenvectors of the top \(k\) eigenvalues. Eigenvalues are all considered to be one. In RandSelect, the eigenvalues are also all considered to be one, and we select the (same) elements of the two gradient vectors randomly. It needs a much larger \(k\) than RandProj. Arnoldi is \(10^3-10^5\) faster than LISSA while being only a couple of percent worse on AUC and AP. TracIn~\cite{https://doi.org/10.48550/arxiv.2002.08484} performs best on AUC and AP, but Arnoldi and RandProj are an order of magnitude faster.

\subsection{TracIn}

\begin{figure}
    \centering
    \includegraphics[width=0.8\linewidth]{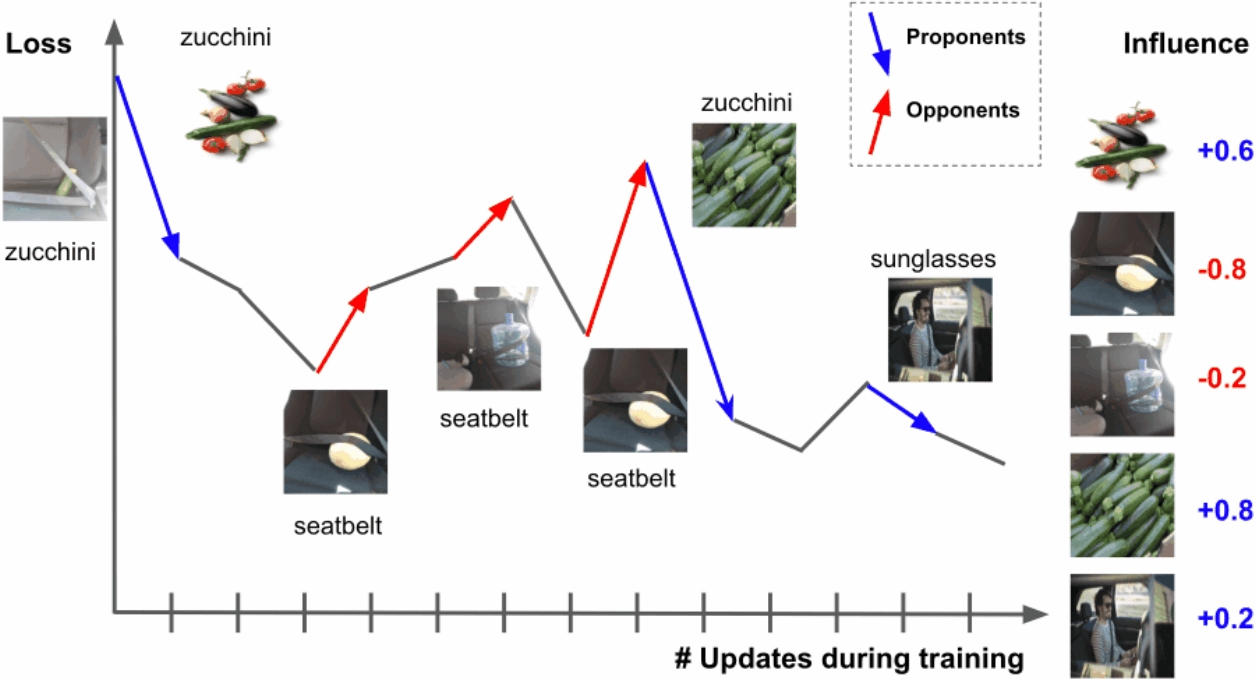}
    \caption{Loss value of a `zucchini' sample over the course of training. The initial loss value for the test sample at the beginning of training is shown on the left. Test losses are evaluated \emph{after} being presented with the shown images. Different training samples have a different impact on the test loss of interest.  When we have a similar image but the label is different, the loss goes up for the test sample. Samples that increase the loss are called opponents to the test sample of interest. Samples that decrease the loss are the proponents of a test sample. We can find proponents and opponents for each training sample during training by considering the changes in the loss value. The `Zucchini' training image results in a lower test loss, as the model learns to detect zucchinis better. The `Sunglasses' training image is similar to the seatbelt images (car interior shown) but has a non-car-related label. The model has to focus on a small part of the image to predict correctly. This implicitly helps the prediction of `zucchini'. \textbf{Note}: During training, the general trend of the loss should be downwards, but because of the noisy behavior of SGD (e.g., update after every training sample) and the possibility of overfitting, the test loss of sample \(z\) does not have to decrease at every gradient step.}
    \label{fig:tracin}
\end{figure}

Are we asking the right question in the previous set of methods to attribute to training samples? We only measure the change in the loss \wrt the optimal model, given an infinitesimal change in the weight of one of the training samples. This sounds super naive and irrelevant in practice. Who would want to introduce an infinitesimal change in the weight of one of the samples?

TracIn, presented in the paper ``\href{https://arxiv.org/abs/2002.08484}{Estimating Training Data Influence by Tracing Gradient Descent}''~\cite{https://doi.org/10.48550/arxiv.2002.08484}  is another approach from 2020: We decompose the final test loss \(L(z, \theta_T)\) of the trained model \(\theta_T\) minus the baseline loss \(L(z, \theta_0)\) of the randomly initialized model \(\theta_0\) into contributions from individual training samples. This is \emph{global linearization} of the final test loss \wrt the update steps.\footnote{It is global because one can use this linearization for \emph{any} test sample.}

Figure~\ref{fig:tracin} gives an intuitive introduction to TracIn -- it considers \emph{useful} and \emph{harmful} examples. \emph{TracIn is the Integrated Gradients for training sample attribution.}\footnote{This is because the linearization the method admits is very similar to the one of Integrated Gradients.}

The loss for the test sample at final iteration \(T\) can be written as the telescopic sum (i.e., everything cancels):
\[L(z, \theta_T) = L(z, \theta_0) + (L(z, \theta_1) - L(z, \theta_0)) + \dots + (L(z, \theta_T) - L(z, \theta_{T - 1})).\]
It is the sum of the original loss value and the loss differences between consecutive parameter updates.

Let us first consider the case when \(\theta\) is updated for every single training sample \(z_j\) (batch size \(= 1\)). There is a clear, unique assignment of which training sample affects which parameter update step. (We never do this in practice, but we assume this for simplicity.) Then there is a natural notion of contribution of \(z_j\) to the test loss \(L(z, \theta_T)\):
\[\operatorname{TracInIdeal}(z_j, z) = \sum_{t: z_j \text{ used for } \theta_t \text{ update}} L(z, \theta_t) - L(z, \theta_{t - 1}).\]
The summation is over the changes of loss for test sample \(z\), where the parameter update was done by training on a sample of interest \(z_j\). There will be millions/billions of iterations. We want to determine which of these iterations corresponds to the training sample of interest; then, we sum up these differences. This results in a completeness property (refer back to the Integrated Gradients method for test feature attribution):
\begin{align*}
L(z, \theta_T) - L(z, \theta_0) &= \sum_{t = 1}^T L(z, \theta_t) - L(z, \theta_{t - 1})\\
&= \sum_{j = 1}^n \sum_{t: z_j \text{ used for } \theta_j \text{ update}} L(z, \theta_t) - L(z, \theta_{t - 1})\\
&= \sum_{j = 1}^n \operatorname{TracInIdeal}(z_j, z).
\end{align*}
This follows from each time step corresponding to a unique training sample. Thus, we have a decomposition of (final loss - baseline loss) into individual contributions.

Of course, the critical issue with this formulation is that, in practice, we update models on a \emph{batch} of training samples. When there is a parameter update, it is hard to attribute the change in loss (due to the update) to individual training samples in the batch. 
Many training samples are involved in the difference \(L(z, \theta_t) - L(z, \theta_{t - 1})\).

Each parameter update with SGD looks as follows:
\[\theta_{t + 1} = \theta_t - \frac{\eta_t}{|B_t|} \sum_{i: z_i \in B_t} \nabla_\theta L(z_i, \theta_t)\]
where \(\eta_t\) is the learning rate at step \(t\) and \(|B_t|\) is the size of the batch at step \(t\). This is usually kept fixed, but we often do not drop the last truncated batch that has a smaller size. We average the gradients over the batch. We have a nice decomposition of the parameter update steps as a sum of individual training-sample-wise gradients for the loss (in the batch).

We rewrite the loss \(L(z, \theta_{t + 1})\) with parameters from time step \(t + 1\) as
\begin{align*}
L(z, \theta_{t + 1}) &= L\left(z, \theta_t - \frac{\eta_t}{|B_t|} \sum_{i: z_i \in B_t} \nabla_\theta L(z_i, \theta_t)\right)\\
&= L\left(z, \theta_t\right) + \left(- \frac{\eta_t}{|B_t|} \sum_{i : z_i \in B_t}\nabla_\theta L(z_i, \theta_t)\right)^\top \nabla_\theta L(z, \theta_t) + o(\eta_t)
\end{align*}
where we performed a Taylor expansion of \(L(z, \theta_{t + 1})\) around \(\eta_t = 0\) (\(f(\eta_t) = f(0) + \eta_t f'(0) + o(\eta_t)\)) or around \(\theta_t\) (\(f(\theta_{t + 1}) = f(\theta_t) + (\theta_{t + 1} - \theta_{t}) \nabla_\theta f(\theta_t) + o(\theta_{t + 1} - \theta_{t})\)). We can choose to do both because they have a linear relationship. This is an \emph{accurate approximation} because we are using a small learning rate (\(1\mathrm{e}{-3}\)), so \(\theta_{t + 1}\) is close to \(\theta_t\).

Therefore,
\[L(z, \theta_t) - L(z, \theta_{t + 1}) \approx \frac{\eta_t}{|B_t|} \sum_{i: z_i \in B_t} \nabla_\theta L(z_i, \theta_t)^\top \nabla_\theta L(z, \theta_t).\]
In words, the difference in loss values before and after the update is approximately equal to some constant times a summation over dot products of training sample gradients with the test sample gradient. There is a natural decomposition of the contribution of individual samples in the batch towards the difference in the loss. When this difference is a large positive number, it means the batch samples were useful for the test sample \(z\). This is the particular reason why we ``flip the sign'' and choose to model \(L(z, \theta_t) - L(z, \theta_{t + 1})\).

A natural notion of the contribution of sample \(z_j\) towards the difference in losses for this particular update is given by
\[\frac{\eta_t}{|B_t|}\nabla_\theta L(z_j, \theta_t)^\top \nabla_\theta L(z, \theta_t)\]
when \(z_j\) is included in batch \(B_t\) for updating \(\theta\) and 0 otherwise. Using this approach, we make attributing to individual training samples feasible in practice. This is a constant times the dot product between the test sample of interest gradient and the training sample of interest gradient. This is similar to what we have seen in the previous methods.

Summing over the entire trajectory of model updates, we define the contribution of \(z_j\) towards the loss for \(z\) as
\[\operatorname{TracIn}(z_j, z) = \sum_{t: z_j \in B_t} \frac{\eta_t}{|B_t|}\left\langle \nabla_\theta L(z_j, \theta_t), \nabla_\theta L(z, \theta_t) \right\rangle.\]
This is the final definition of TracIn, the trajectory-based influence of sample \(z_j\) towards test sample \(z\). It is simply a summation of all parameter update steps \(t\) that contained \(z_j\) in the batch. These are the only relevant terms, the others are \(0\). The smaller the loss becomes on test sample \(z\) between steps \(t\) and \(t + 1\), the more we attribute those training samples that were in the batch of step \(t\).

\subsection{TracIn vs. IF}

These two methods have very similar formulations but also some key differences.
\begin{align*}
\operatorname{IF}(z_j, z) &= \left\langle Q^\top \nabla_\theta L(z, \hat{\theta}), Q^\top \nabla_\theta L(z_j, \hat{\theta}) \right\rangle_{\Lambda^{-1}}\\
\operatorname{TracIn}(z_j, z) &= \sum_{t: z_j \in B_t} \frac{\eta_t}{|B_t|}\left\langle \nabla_\theta L(z_j, \theta_t), \nabla_\theta L(z, \theta_t) \right\rangle
\end{align*}
Both use a form of a dot product between parameter gradients for the training and test samples. It is quite impressive that the final formulations end up being so simple, but it is a natural byproduct of linearization. TracIn sums over training iterations (checkpoints) and does not use a Hessian-based distortion of the dot product (to squeeze/expand some of the eigenbasis directions). It is, therefore, cheaper because we do not need to compute the Hessian. However, it is very memory intensive. In contrast, IF considers only the final\footnote{Strictly speaking, IF considers the globally optimal parameter configuration in the formulation.} parameter and distorts the space using the average Hessian.

Using IF, we are missing out on all contributions on the way during training. Intuitively, TracIn makes more sense, but it is hard (if not impossible) to say which one is better conceptually. We can only use empirical evaluation to tell which serves our purpose better. IF has many more assumptions that are also violated in practice. The method considers globally optimal parameter configurations, strict convexity, and twice-differentiability. In practice, the eigenvalues could also become negative (saddle point) or 0 \(\implies\) invertibility does not hold when we have a 0 eigenvalue (i.e., the loss is constant in some directions). In theory, this can happen during optimization. A small epsilon has to be added.

\section{Evaluation of Attribution to Test Samples}

There are two perspectives of evaluation of such methods: (1) comparing approximate values against their GT counterparts and (2) evaluating such attribution methods based on some end goals/downstream tasks.

\subsection{Comparison of Approximation Against GT Value}

IF approximates the remove-and-retrain algorithm (remove a certain training sample, retrain, and see how much that influences the loss value for the test sample of interest). One can measure \emph{soundness} by comparing influence values against the actual remove-and-retrain baseline. This is an evaluation of soundness. To see the correspondence, consider the first-order Taylor approximation of \(L(z, \hat{\theta}_{\epsilon, j})\) again around \(\epsilon = 0\). To avoid confusion, we stick to the definition of IF where a larger positive value signals positive influence.\footnote{Interestingly, the FastIF paper considers the original IF definition without flipping the sign.} We have
\[L(z, \hat{\theta}_{\epsilon, j}) - L(z, \hat{\theta}) = \epsilon \underbrace{\restr{\frac{\partial L(z, \hat{\theta}_{\epsilon, j})}{\partial \epsilon}}{\epsilon = 0}}_{-\operatorname{IF}(z_j, z)} + o(\epsilon).\]
The notion of removal is equivalent to setting \(\epsilon = -1/n\), which is generally a very small number, so the linear approximation stays reasonably faithful to the actual loss function. We finally obtain
\[L(z, \hat{\theta}_{\setminus j}) - L(z, \hat{\theta}) \approx \frac{1}{n} \operatorname{IF}(z_j, z).\]
To benchmark IF on how faithful it is to the remove-and-retrain algorithm, we can compare the left quantity to the right one. We might be interested in removing not just one sample but a group of them. This is not modeled by the most naive version of remove-and-retrain that IF approximates.

\begin{figure}
    \centering
    \includegraphics[width=0.8\linewidth]{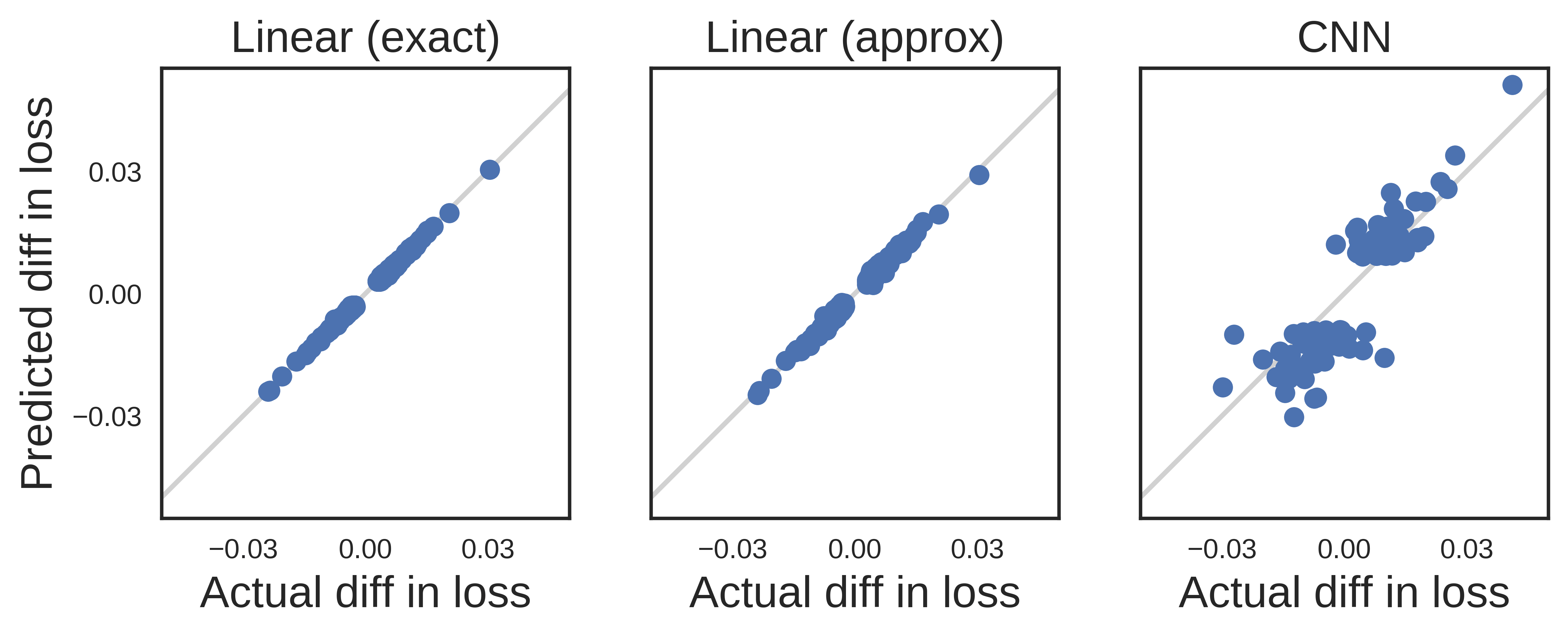}
    \caption{Comparison of the predicted difference in loss after the removal of a sample (i.e., the IF value) against the actual difference in loss. Figure taken from~\cite{https://doi.org/10.48550/arxiv.1703.04730}. The benchmark measures faithfulness to leave-one-out retraining on MNIST. For every training and test sample, we measure the change in test loss by actually removing a training sample. This is quite fast for a linear model. We also have the predicted difference in loss through the IF computation. We compare the two values. \emph{Left.} The gradient-based approximation of the influence (times \(1/n\) to model removal) gives nearly the same result as the actual difference in the loss for the linear model (logistic regression). It is a good sanity check that the exact Hessian computation performs well approximation-wise. \emph{Middle.} ``Linear (approx)'' still considers logistic regression but uses the LISSA approximations to speed up the Hessian computation. Even if we use LISSA to approximate the average Hessian, we do not lose much accuracy. \emph{Right.} To evaluate on CNNs, we must take a leap of faith. The logistic regression optimization is strictly convex, but the CNN one is, of course, not. They apply the method to a small CNN on MNIST. We can see some correlation, but many things are seemingly not working anymore. Two groups follow the overall trend, but we do not see much correlation between the actual and the predicted value \emph{within} each group. We have mixed results.}
    \label{fig:soundness}
\end{figure}

\begin{figure}
    \centering
    \includegraphics[width=0.8\linewidth]{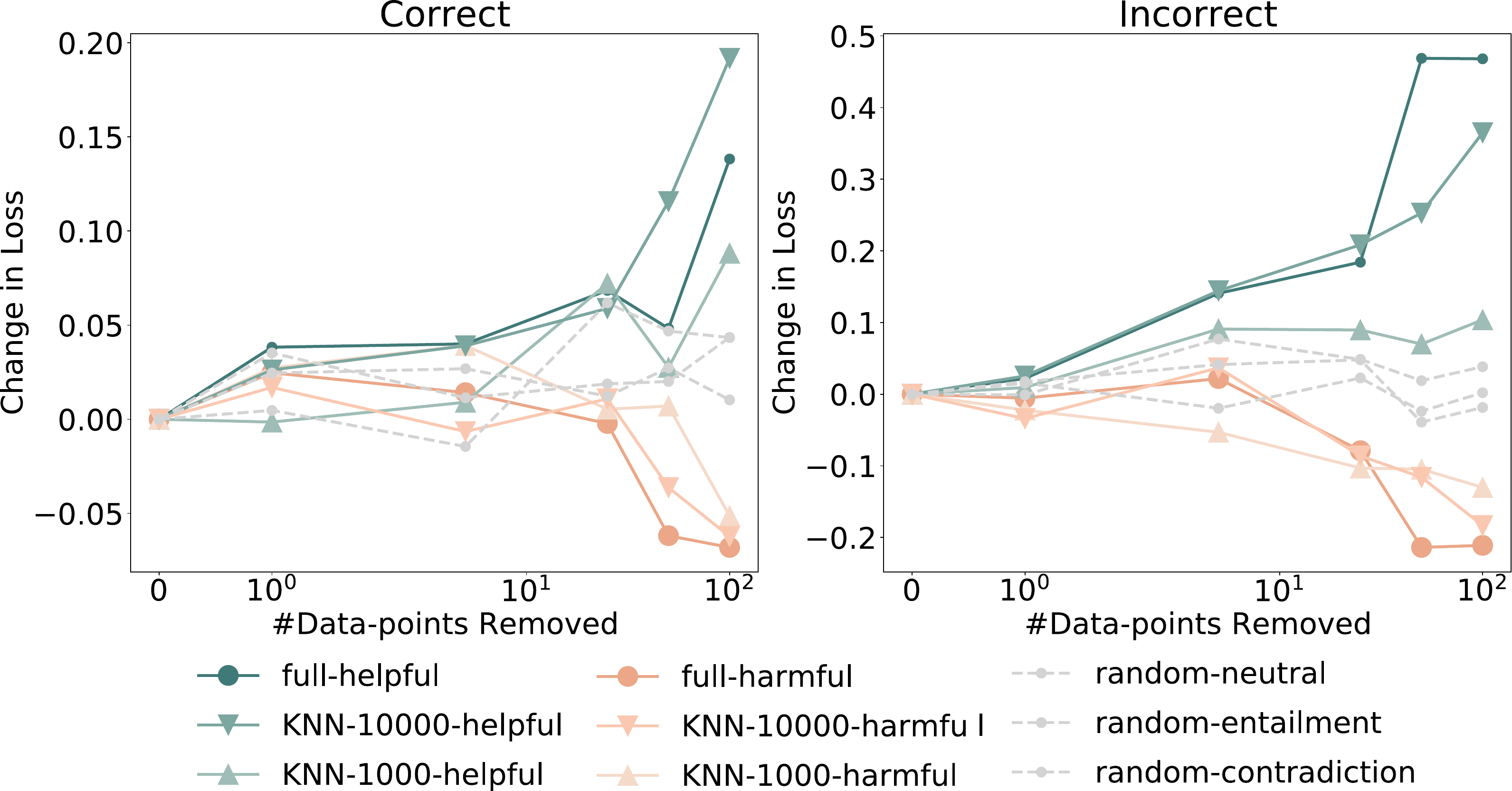}
    \caption{Leave-M-out results on \href{https://cims.nyu.edu/~sbowman/multinli/}{MNLI}~\cite{N18-1101}. 
    ``Change in loss on the data point after retraining, where we remove \(m_\text{remove} \in \{1, 5, 25, 50, 100\}\)
    data-points [either positives or negatives]. We can see that the fast influence algorithms [(the KNN versions)] produce reasonable quality estimations at just a
    fraction of computation cost.''~\cite{https://doi.org/10.48550/arxiv.2012.15781} Correct and incorrect mean that the original predictions were correct/incorrect. Figure taken from~\cite{https://doi.org/10.48550/arxiv.2012.15781}.}
    \label{fig:groupout}
\end{figure}

Soundness results of IF are shown in Figure~\ref{fig:soundness}. We can also try leaving a group of samples out from the training set and seeing how the model reacts regarding the change in the loss for a test sample. This is shown in Figure~\ref{fig:groupout} from the FastIF paper~\cite{https://doi.org/10.48550/arxiv.2012.15781}, which is yet another paper on how we can speed up IF computations. The task is MNLI, a 3-class natural language inference task with classes entailment, neutral, and contradiction. The group of samples we remove is determined by the influence values (we sort all training samples according to their influence values). The influence value has parity: it can be positive or negative. Using this book's IF definition, positive means that including the sample helps, and negative means that by including this training sample (\(\epsilon > 0\)), we are increasing the loss. The general trend is that removing helpful samples increases the loss. (It is harmful to remove the samples with a high IF value.) Similarly: removing harmful samples decreases the loss. (It is useful to remove the samples with a low IF value.) By just removing random samples, we do not see much change in the test loss. ``Full'' means we use the entire dataset. The KNN versions correspond to selecting representative samples from the training set. We can see that this can even be beneficial.

\subsection{Focus on the End Goal: Mislabeled Training Data Detection}

IF and TracIn are eventually serving certain end goals. Remove-and-retrain may not be very useful as the end goal. For example, when the actual end goal is to debug/improve the model/dataset, faithfulness to the remove-and-retrain algorithm is not of particular interest. It is just an intermediate step (a proxy) for using the method for improving models.\footnote{Suppose that a self-driving car killed a pedestrian. We need to find out which data sample was responsible for the incorrect (sequence of) predictions. Remove-and-retrain is not the end goal in this case, we do not care about how well we approximate it or whether we even approximate it at all.} We need to evaluate based on more reasonable end goals, e.g., mislabeled training data detection. We will see how people can use influence functions to detect mislabeled training samples. Checking, e.g., how well our method approximates remove-and-retrain might be good to check whether our proposed idea works. Then, we evaluate the method using the actual end goal. This can change the conclusion of which method is better for us.

\begin{definition}{Self-Influence}
Self-influence is a metric used in training sample attribution methods that measures how much contribution a particular training sample \(z_j\) has to its own loss.

\medskip

\textbf{Example}: Using influence functions, the self-influence score for sample \(z_j\) is \(\operatorname{IF}(z_j, z_j)\).  Using TracIn, we can use \(\operatorname{TracIn}(z_j, z_j)\) as a self-influence score.
\end{definition}

We make use of self-influence scores for mislabeled training data detection. If a sample is one of its kind, then it only has itself to decrease its loss, therefore, we expect a high self-influence score. Looking at that exact sample is the only way to decrease the loss of that sample. On the other hand, if the sample is just like other data points in the training set, then it is among many that decrease its loss, therefore, we expect a low self-influence score: Including it or not has little influence. Mislabeled data are typical examples of ``one of its kind'' data. As such, we expect high self-influence scores for them.\footnote{Assuming that the training set has many correctly labeled data and a few mislabeled data points (i.e., there is no systematic mislabeling).} By measuring self-influence, we should be able to tell which samples are mislabeled. 

\begin{figure}
    \centering
    \includegraphics[width=0.8\linewidth]{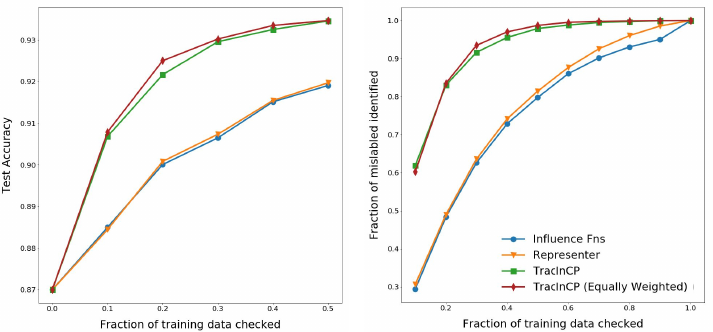}
    \caption{Results of using self-influence to detect mislabeled training samples on CIFAR. \emph{Left}. Fixing the mislabeled data found within a certain fraction of the training data results in a larger improvement in test accuracy for TracIn compared to the other methods. \emph{Right}. TracIn retrieves mislabeled samples much better than IFs. Figure taken from~\cite{https://doi.org/10.48550/arxiv.2002.08484}.}
    \label{fig:selfinf}
\end{figure}

Figure~\ref{fig:selfinf} shows benchmark results on mislabeled training data detection from the TracIn paper. Mislabeled training data detection is a typical binary detection task: we want to classify mislabeled/not mislabeled. We know the ground truth in the benchmark; we try to retrieve the mislabeled ones in the training set. We can use detection metrics like AUROC and AP (AUPR), which are typical evaluation scores for retrieval tasks. Our only feature is the attribution score. The question is, ``Is there a threshold that is extremely good for separating mislabeled samples from not mislabeled?'' In this benchmark, however, they are \emph{not} doing that. Instead, they sort training samples according to self-influence values and then decrease the threshold from top to bottom and see how many mislabeled samples are retrieved.

\begin{figure}
    \centering
    \includegraphics[width=0.6\linewidth]{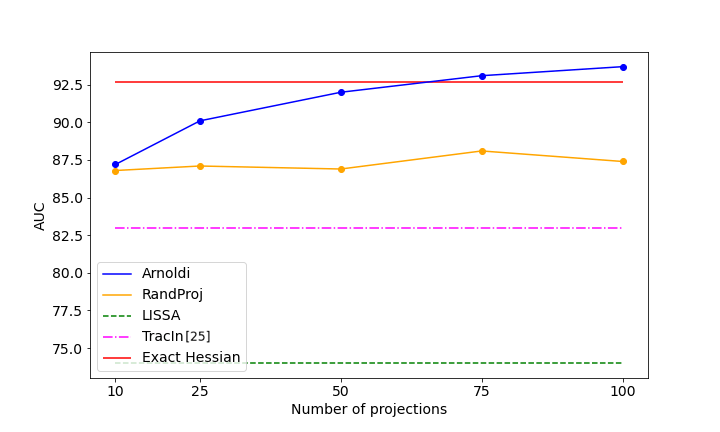}
    \caption{Results of using self-influence to detect mislabeled training samples on MNIST using a small CNN. AUC for retrieval of mislabeled MNIST examples as a function of the number of eigenvalues (projections), \(\tilde{p}\). Figure taken from~\cite{https://doi.org/10.48550/arxiv.2112.03052}.}
    \label{fig:mislabel2}
\end{figure}

We also discuss using self-influence to detect mislabeled training samples on MNIST. The results are shown in Figure~\ref{fig:mislabel2}. The task is not perfectly aligned with IF computation: the exact method can be surpassed.

Finally, we discuss the retrieval of mislabeled MNIST examples using self-influence for a larger CNN. As discussed before, AUC and AP are usual detection metrics for mislabeled samples. The results are shown in Table~\ref{tab:results2}.

\section{Applications of Attribution to Test Samples}

\subsubsection{Fact Tracing}

\begin{figure}
    \centering
    \includegraphics[width=0.5\linewidth]{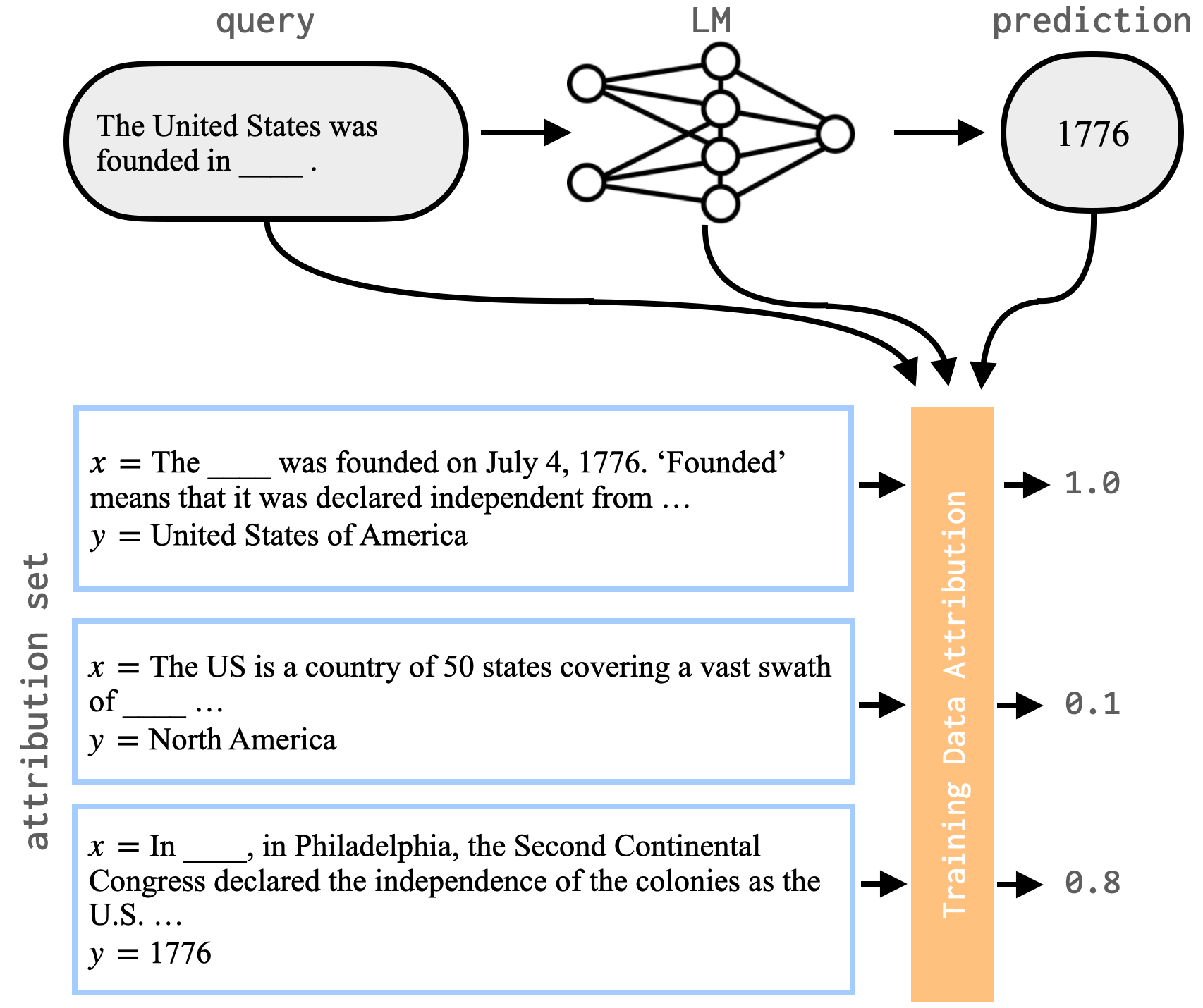}
    \caption{Illustration of using training data attribution scores for fact tracing. Figure taken from~\cite{https://doi.org/10.48550/arxiv.2205.11482}.}
    \label{fig:facttracing}
\end{figure}

We discuss fact tracing, an important application of test sample attribution, as shown in the paper ``\href{https://arxiv.org/abs/2205.11482}{Towards Tracing Factual Knowledge in Language Models Back to the Training Data}''~\cite{https://doi.org/10.48550/arxiv.2205.11482}. Suppose we have a language model that is trained to predict missing words using actual facts, and we have built a dataset with the GT fact attributions in the training set. Then we can measure fact retrieval performance: We evaluate any Training Data Attribution method on its ability to identify the so-called true proponents, i.e., the true training sample information sources. We want to retrieve the true proponents out of a large set of training examples, which is, again, a classical retrieval task. This is illustrated in Figure~\ref{fig:facttracing}.

It is a natural question to ask the model, ``Did you just make this up? Which training datum did you look at to make this decision?'' Nowadays, fact tracing matters a lot, and training data influence can be readily used for it. LLMs are critical candidates for this method. We cannot be sure how it would scale, but it is something to keep an eye out for.

\subsubsection{Membership Inference}

Given a model and arbitrary data we give to the model, we wish to see whether that data was included in the training of that model.
\begin{itemize}
    \item ``Was this image used for training the DALL-E model?''
    \item ``Was this image used for generating the current image that I got?''
\end{itemize}
Being able to answer such questions could be a nice tool for dealing with copyright issues for large-scale generative models. It would also be possible to use influence functions and training attribution in general. Suppose we had access to the training set. Then, we could use the scores to sort decreasingly and manually check whether the sample was used (soft filtering). Alternatively, if we are really searching for exact matches, we could search for matches according to the ordering given by influence function scores. Hashing already works for checking for exact matches very efficiently, but it would not work for matches that are not exact (e.g. when JPEG encoding/decoding is applied).
There are only very few papers in this area so far, but it is gaining traction. Large companies are probably also already working on this problem.

\clearpage

\chapter{Uncertainty}
\section{Introduction to Uncertainty Estimation}

Uncertainty is everywhere. Having complete information and a perfect understanding of a system can only happen in simple and closely controlled environments. The world around us is not such an environment. Humans learn to build complex internal models of uncertainty to cope with incomplete information and react robustly to events that either have not happened yet or are only partially observed.

Understanding, quantifying, and evaluating uncertainty is of crucial importance in our everyday lives, but also in fields specialized to cope with and leverage uncertainty. Examples include financial analysis, economic decisions, general statistics, probabilistic modeling, and also machine learning. Classical ML theory usually did not aim to \emph{make systems know when they do not know} -- the main goal was to find methods and solutions that work well, considering them as standalone components. These days, accuracy in most applications is not the biggest concern -- most ML solutions provide reasonably good accuracy in several tasks. Instead, there is an ever-increasing demand to quantify sources of uncertainty in ML models and make them understand their own limitations. As we will soon see, uncertainty quantification is a crucial requirement whenever we want to incorporate an ML solution into a certain pipeline.

In the Uncertainty chapter of the book, we are going to further motivate the need for uncertainty estimation, quantify sources of uncertainty, consider methods that can give us different kinds of uncertainty estimations, and learn about methods to evaluate uncertainty predictions for DNNs.

\subsection{Motivation}

We first consider a meeting with another business, based on a real story of one of the authors when they were working at a company. Teams without ML knowledge tend to downplay the difficulty of doing technical things. There are always typical subjects in such meetings:
\begin{itemize}
    \item ``Why does your AI system not return how sure it is about the output?''
    \item ``Is it not kind of trivial to make the system predict confidences?''
    \item ``We cannot plug your system into our pipeline if there is no such estimate.''
    \item ``We really need it, cannot you just do it?''
\end{itemize}
Unfortunately, solving such tasks is not at all trivial. However, they are prevalent (as there are many such requests) and valid desires; we will see methods to achieve these goals.

\subsection{Uncertainty estimation is a critical building block for many systems.}

When an ML model is part of a bigger modular pipeline, uncertainty estimation is very beneficial and often required. For an ML-based data-driven module, it is not easy to trust everything the model outputs. Such models are never perfect and extra care is needed to use the model's prediction in downstream modules. This is also true when the downstream module in question is a human -- people do not (and should not) trust every prediction of the model.

Let us suppose for a moment that the model already knows about itself how certain it is. We consider some example downstream use cases of reporting uncertainty (in later modules of the pipeline).

\textbf{Human in the loop.} We only want humans to intervene when the ML confidence is low, as human knowledge is expensive. When the model's confidence is low, the model can say, ``I am not sure about the result.'' When humans need to intervene, they can take control and handle certain requests themselves (i.e., they can fix the model's prediction).

\textbf{Risk avoidance.} When there are great risks involved in the model's task, the ML system should only act when it is confident. If the model is unsure, the processing pipeline should stop (or fall back to some other safe state), as the situation is deemed too risky. An example of this is a learning-based manufacturing robot for cars. When the robot is uncertain about its next action, there is a high risk it is going to make a mistake which could also result in it damaging or destroying the car. We want the model to be able to say, ``We should probably not take care of this input and just stop.''

\begin{figure}
    \centering
    \includegraphics[width=0.5\linewidth]{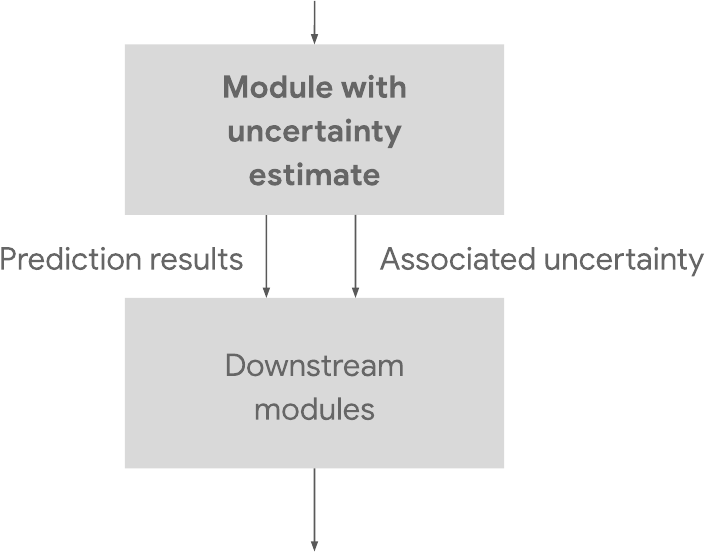}
    \caption{Simplified flowchart of the ideal integration of ML models into modular pipelines. In addition to the prediction results, we also wish to obtain associated uncertainty estimates to efficiently use the predictions in downstream tasks.}
    \label{fig:flowchart}
\end{figure}

Thus, it is very beneficial for our model to output \emph{two} predictions when it is part of a pipeline: the prediction results and also the associated uncertainty estimate(s), as illustrated in Figure~\ref{fig:flowchart}. This gives us many more choices of what to do later in the pipeline in downstream modules.

\subsubsection{When do we need confidence estimation?}

In general, confidence estimation is needed when the outputs of a model cannot be treated equally -- outputs for certain samples are more confident and some of them are less trustable. 

It is not needed when the system always returns perfect answers. Why would we need it? If such a time would come when AI systems were always giving the right answers, this study would become useless. We will learn about whether that can happen\dots\ (Spoiler: It cannot, as in almost any sensible scenario, there is some level of stochasticity we cannot get rid of.)

\subsection{Example Use Cases of Uncertainty Estimation}
The following examples of uncertainty estimation are inspired by~\cite{balajitalk}.

\subsubsection{Image search for products}

In this example, we do not consider the old Google image search. We consider products like Google Lens. Such products do not only search for similar images -- they also take the user and context into account:
\begin{center}
``Google Lens is a set of vision-based computing capabilities that can understand what you’re looking at and use that information to copy or translate text, identify plants and animals, explore locales or menus, discover products, find visually similar images, and take other useful actions. [...] Lens always tries to return the most relevant and useful results. Lens' algorithms aren't affected by advertisements or other commercial arrangements. When Lens returns results from other Google products, including Google Search or Shopping, the results rely on the ranking algorithms of those products.''~\cite{googlelens}
\end{center}
Companies are usually also very motivated to link image search results to actual products to make money. Customers can also get quick answers from such image search results.

Given an image, the task is to find the product that is shown. What should happen if the photo taken by the user is of poor quality? Regular algorithms would search for the most likely product anyway, which is usually a very poor suggestion. If the model is equipped with uncertainty estimates, when the confidence is low, it can
\begin{enumerate}
    \item ask the user to take another photo, and/or
    \item show different results from all products that could match with high probability.
\end{enumerate}
What if the photo \emph{does not contain} any product of interest? Again, regular algorithms would simply return poor results. Uncertainty estimation can allow the system to determine whether the provided photo is relevant. When there is no object of interest, the system can output suggestions such as ``User should be posing the camera differently.'' or ``Try to focus on an object of interest.'' This feedback loop can ensure that the model can perform correctly and does not mislead the user with unconfident predictions.

\subsubsection{High-stake decision making}

The prime example of a high-stake decision-making application is healthcare, where the model has to determine whether there is anything wrong with our body. We can use model uncertainty to decide when to trust the model or defer to a human. This is a crucial ability of a model in general cost-sensitive decision-making, where mistakes can potentially have huge costs. Costs include potential lawsuits, the death of a patient, or fatal road accidents. The task is to provide a binary prediction of healthy/diseased from the input image. Ideally, the model should make a prediction and output confidence estimates as well. One should only trust the model's predictions when they are confident.\footnote{For example, we might accept the model's prediction when the provided confidence estimate is above a certain tuned threshold.} When the model is not confident enough, we defer to a human. For example, we can ask a human doctor to come in and take a look.

\begin{table}
    \centering
    \caption{Example cost table for decision making in healthcare. Predicting `healthy' for a diseased person has the highest cost, as such cases can even lead to the death of a patient. Table recreated from~\cite{balajitalk}.}
    \label{tab:costtable}
    \begin{tabular}{c l c c}
      \toprule
         ~ & ~ & \multicolumn{2}{c}{True Label} \\
         ~ & ~ & Healthy & Diseased \\
         \midrule
         \multirow{3}{4em}{Action} & Predict Healthy & 0 & {\color{red}10} \\ 
         & Predict Diseased & 1 & 0 \\
         & Abstain ``I don't know'' & {\color{blue}0.5} & {\color{blue}0.5} \\
         \bottomrule
    \end{tabular}
\end{table}

\begin{figure}
    \centering
    \includegraphics[width=0.4\linewidth]{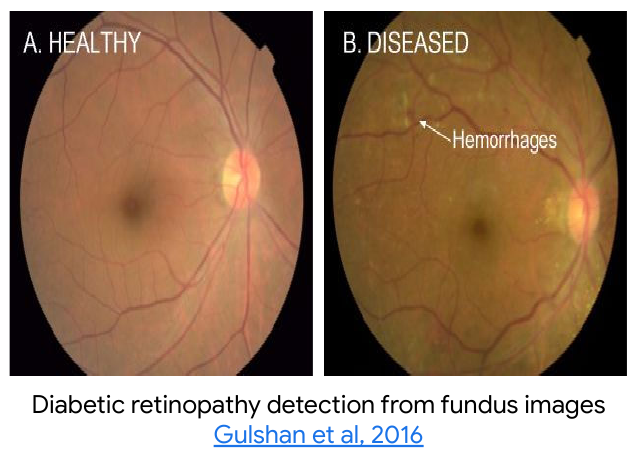}
    \caption{Diabetic retinopathy detection from fundus images. Predicting `healthy' can be catastrophic if the patient is actually diseased. Figure taken from~\cite{balajitalk}.}
    \label{fig:diabetic}
\end{figure}

In discrete cost-sensitive decision-making problems, we usually have \emph{cost tables}, depicted in Table~\ref{tab:costtable}. We have a very high stake in false negative disease diagnoses. We incur huge costs. Thus, we want to predict `healthy' only when the system is very certain, and we even prefer the answer `I don't know' over predicting false positives. Predicting `I don't know' defers to a human doctor. An example of such a scenario is diabetic retinopathy detection from fundus images~\cite{balajitalk}, illustrated in Figure~\ref{fig:diabetic}.

The field of self-driving cars also requires uncertainty estimates. It also qualifies as high-stake decision-making, as people's lives are at stake.\footnote{Self-driving and healthcare usually come in pairs when discussing high-stake ML use cases.} We do not want our current self-driving systems to drive \emph{in all cases}. In self-driving scenarios, we often experience \emph{dataset shift}. We want to make sure that our car does not crash in such cases. Examples include changes in
\begin{itemize}
    \item time of day/lighting (driving at night vs. in the morning),
    \item geographical location (inner city vs. suburban location),
    \item weather conditions (thunderstorm vs. clear weather),
    \item or traffic conditions (traffic jams, construction sites, clear highways).
\end{itemize}
In such cases, we wish to take over control and drive responsibly. By using uncertainty estimation, the car can tell us when it is uncertain.

\subsubsection{Open-set recognition}

Open-set recognition is a different scenario that is more specific to the classification task. In the development (dev) stage (Section~\ref{ssec:formal}), we can pre-define a set of classes, e.g., the 100 most popular skin condition classes. When deploying in the real world, there can be very rare diseases as test inputs for which we do not have classes. If the model predicts `normal skin' in such cases, it is very harmful. However, the other scenario is not better either: ``Well, it does not look normal, but since I need to pick one from the known cases, I will just guess Acne.'' A classification system should also be able to say, ``This is something I have not learned before, a new class. This is none of the above.'' This can either be an explicit class, or it can be signaled by low predictive confidence.

Open-set recognition considers different ways to deal with new classes in deployment. There are generally two variants of open-set recognition: models trained with or without OOD data. When they are trained with OOD data, they also usually contain a separate dimension in the output probability vector for indicating the probability of OOD (explicit introduction of the `I don't know.' class). When they are trained only with known classes, there is no data to train this extra dimension and, therefore, it is not added. Even in this case, the model can be trained to predict calibrated uncertainty estimates that can then be used to determine the `I don't know.' class in an implicit fashion. Of course, without explicit supervision, the latter case will likely produce worse results.

\begin{figure}
    \centering
    \includegraphics[width=0.5\linewidth]{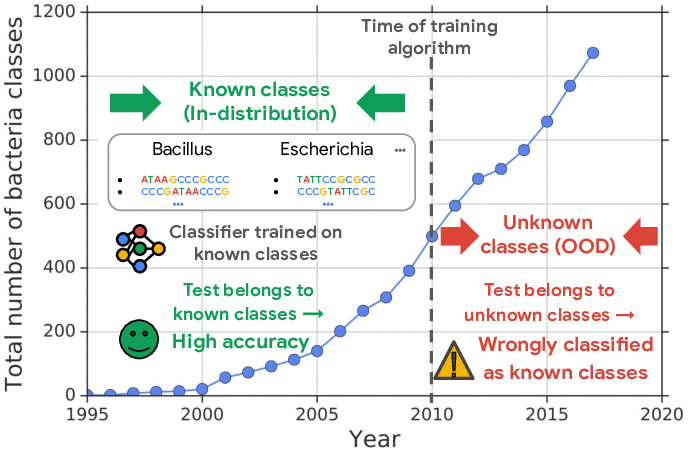}
    \caption{Example for the need for uncertainty estimation in the classification of genomic sequences. ``A classifier trained on known classes [without proper uncertainty calibration] achieves high accuracy for test inputs belonging to known classes, but can wrongly classify inputs from unknown classes (i.e., out-of-distribution) into known classes with high confidence.''~\cite{googleood} Figure taken from~\cite{googleood}.}
    \label{fig:ood}
\end{figure}

The same story goes for ``growing field'' cases. An example is the classification of genomic sequences. We discover more and more bacteria classes in biology research -- new entries are coming to our database of bacteria. We usually have high ID accuracy on known classes, but this is not sufficient. We wish to be prepared for new bacteria classes in the future (unknown classes, OOD scheme), but we can only train on classes that are currently in the database. We need to detect inputs that do not belong to any of the known classes. We wish to assign an `I don't know.' label for future cases. This scenario is depicted in Figure~\ref{fig:ood}. Samples predicted as `I don't know.' can be used later on for further training the model: we can put labels on them once we discover them. For example, we can initialize a new row in the classifier layer's weight matrix, add a new bias scalar, and then we can predict one more output class after learning to predict such samples. The keyword here is \emph{class-incremental learning}, which deals with efficiently increasing the number of classes over time without sacrificing the original classification score.

\subsubsection{Active learning}

\begin{figure}
    \centering
    \includegraphics[width=0.6\linewidth]{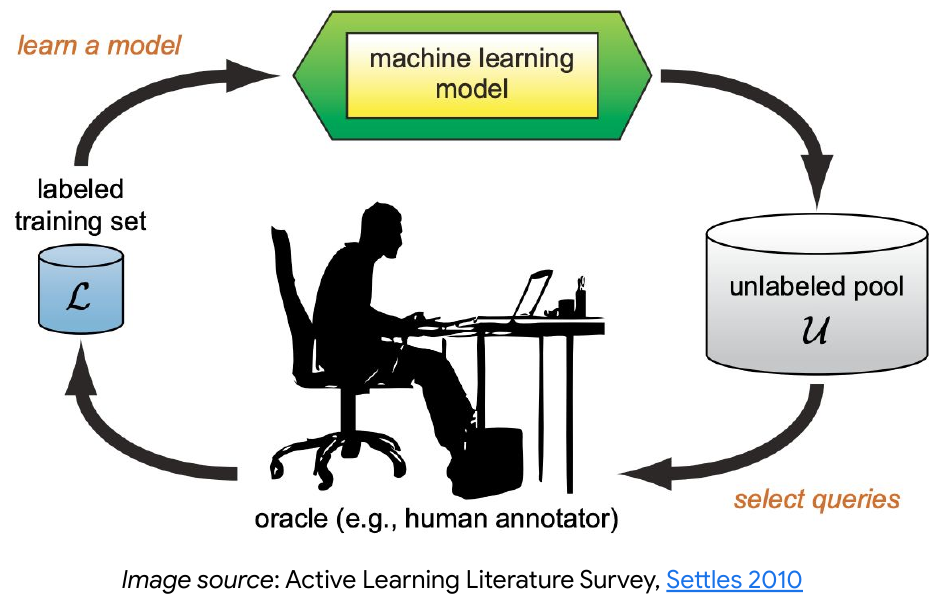}
    \caption{General overview of active learning. We can get away with labeling significantly fewer samples for our model if we label the ``right'' ones. Figure taken from~\cite{balajitalk}.}
    \label{fig:active}
\end{figure}

Active learning, illustrated in Figure~\ref{fig:active}, is concerned with finding samples to label smartly. Instead of going through a huge set of unlabeled samples to label everything, we pick the samples the model is very likely to be confused about, and then ask for human feedback on those samples in an iterative fashion. This way, we maximize the utility of humans (that are expensive). We can use model uncertainty to improve data efficiency and the model's performance in ``blind spots''. To tell which of the unlabeled samples is most likely to have the highest return when annotated by a human, we should rely on a notion of uncertainty and confidence values.

\subsubsection{Hyperparameter optimization and experimental design}

Hyperparameter optimization and experimental design are widely used across large organizations and the sciences. Such methods often employ \emph{Bayesian optimization}. Examples include photovoltaics, chemistry experiments, AlphaGo, electric batteries, and material design. 

\begin{figure}
    \centering
    \includegraphics[width=0.9\linewidth]{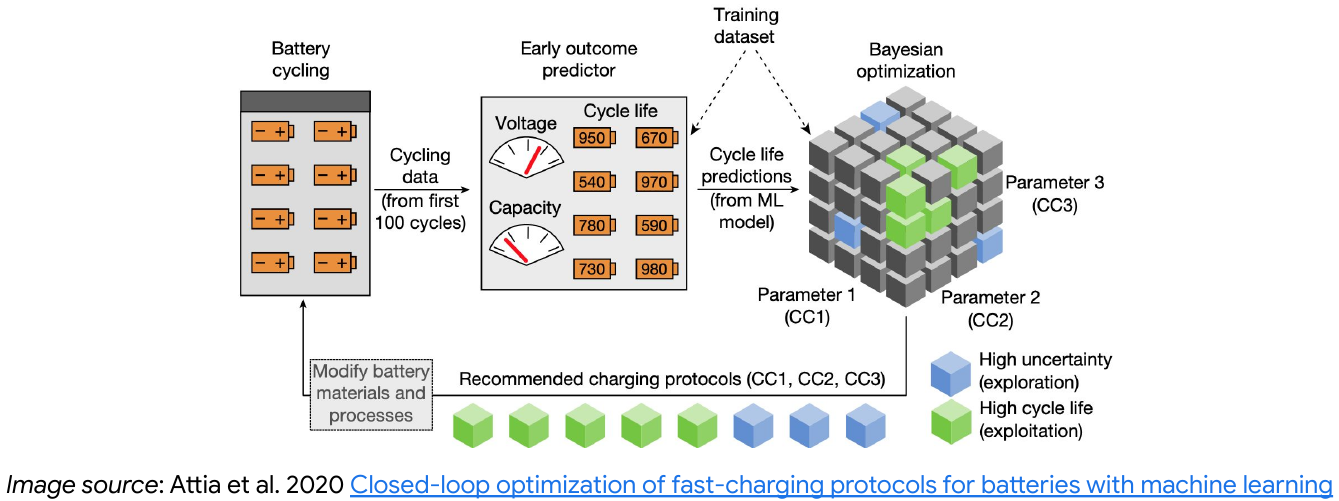}
    \caption{Role of uncertainty in optimizing battery charging protocols with ML. ``First, batteries are tested. The cycling data from the first 100 cycles (specifically, electrochemical measurements such as voltage and capacity) are used as input for an early outcome prediction of cycle life. These cycle life predictions from a machine learning (ML) model are subsequently sent to a BO algorithm, which recommends the next protocols to test by balancing the competing demands of exploration (testing protocols with high uncertainty in estimated cycle life) and exploitation (testing protocols with high estimated cycle life). This process iterates until the testing budget is exhausted. In this approach, early prediction reduces the number of cycles required per tested battery, while optimal experimental design reduces the number of experiments required. A small training dataset of batteries cycled to failure is used to train the early outcome predictor and to set BO hyperparameters.''~\cite{Attia2020ClosedloopOO} The linear model the predicts cycle life of a battery (and also gives a CI for the predictions). The GP relates protocol \(x\) to cycle life \(y\) through its internal parameters \(\theta\). Here, the GP outputs uncertainties naturally. Figure taken from~\cite{Attia2020ClosedloopOO}.}
    \label{fig:bayesopt}
\end{figure}

The setup is as follows. We are searching through a huge (combinatorial) space of possibilities for configurations/settings. For example, in a very naive hyperparameter search for an ML model, we might have
\begin{align*}
5 \text{ learning rates} &\times  4 \text{ numbers of layers} \times 5 \text{ net widths} \times 3 \text{ weight decays}\\
&\times 10 \text{ augmentations} \times 3 \text{ numbers of epochs} \times 3 \text{ optimizers} = 27000
\end{align*}
possible hyperparameter settings to iterate over. Usually, we have thousands or millions of possible combinations, even in quite simple cases. It is clearly infeasible to consider all possible configurations. Bayesian optimization reduces the uncertainty of performance in this complex landscape while also choosing performant configurations. By observing a few data points where the configurations were chosen smartly (i.e., considering the trade-off between uncertainty reduction and exploitation), it constantly updates its beliefs based on the training results of the well-studied configurations. This reduces uncertainty over time, and eventually, we find a configuration that will likely maximize our return. To explore the space most efficiently, we need a notion of uncertainty. An example use of Bayesian optimization for experimental design is shown in Figure~\ref{fig:bayesopt}.

\subsubsection{Object detection pipeline}

\begin{figure}
    \centering
    \includegraphics[width=0.6\linewidth]{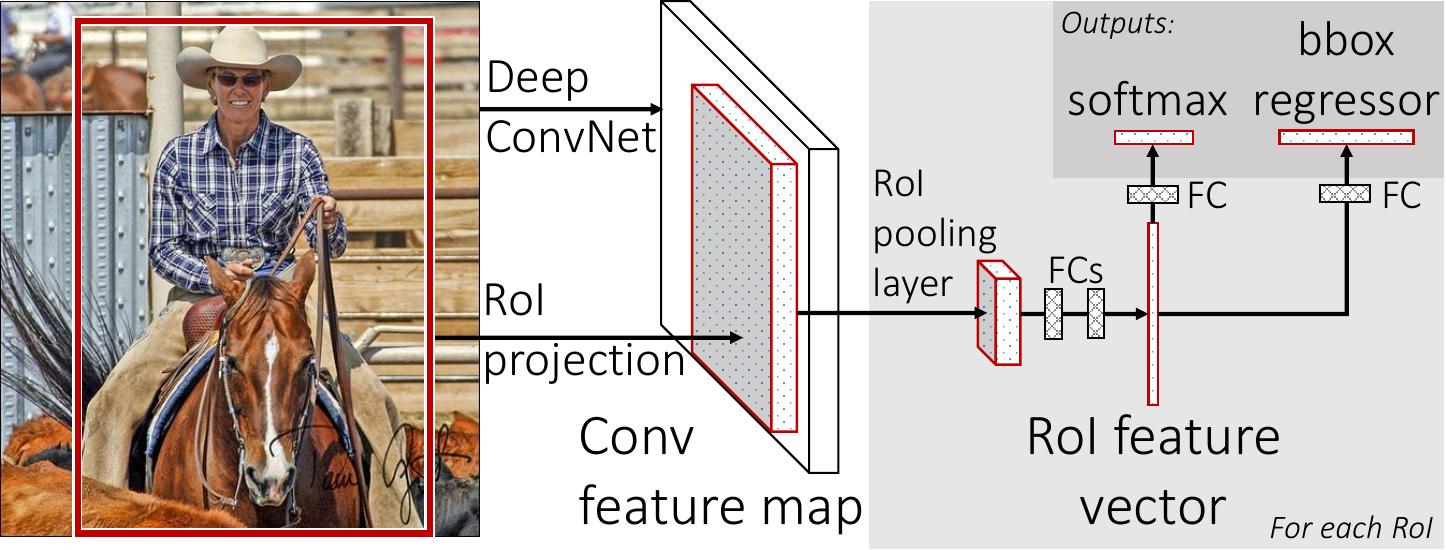}
    \caption{Fast(er) R-CNN is a renowned model in object detection. One of its distinguishing features is its modularity. When proposing bounding boxes for objects, referred to as Regions of Interest (RoIs), the method also provides a confidence or ``objectness'' score for each box. This score is crucial; it allows the system to prune less likely boxes before it refines and classifies the remaining ones, ensuring both accuracy and efficiency. Figure taken from~\cite{https://doi.org/10.48550/arxiv.1504.08083}.}
    \label{fig:faster}
\end{figure}

In object detection, we produce a bounding box and a class label for each object. Two-stage detectors (propose then refine) use multiple modules by construction. We will likely require confidence scores whenever we have multiple modules in any ML setting. Fast(er) R-CNN~\cite{https://doi.org/10.48550/arxiv.1504.08083}, the most popular object detection pipeline, is illustrated in Figure~\ref{fig:faster}. In Faster R-CNN, we have the following stages.
\begin{enumerate}
    \item Propose boxes with confidence scores. (Between \(10^3\) and \(10^6\) boxes are proposed.) This is the \emph{objectness score}.
    \item Prune boxes by thresholding \wrt the confidence/objectness scores. We return only the most likely boxes containing any objects. Then we further perform non-maximum suppression.
    \item Classify the pruned boxes and refine the boxes.
\end{enumerate}

\section{Types and Causes of Uncertainty}
\label{ssec:types}

In this section, we aim to discuss different sources of uncertainty and how they relate to each other. In particular, we will discuss the terms \emph{predictive uncertainty}, \emph{epistemic uncertainty}, and \emph{aleatoric uncertainty}. In the last paragraph of each of the subsections discussing these sources, we give an introduction to \emph{how} we can evaluate these.

\subsection{Predictive Uncertainty}
\label{ssec:predictive}

\begin{definition}{Predictive Uncertainty}
Predictive uncertainty refers to the degree of uncertainty or lack of confidence that a machine learning model has in its predictions for a given input.

\medskip

In particular, predictive uncertainty is typically referred to as the probability of the prediction's correctness. If for a fixed input sample \(x\) we define the indicator variable \(L: \Omega \rightarrow \{0, 1\}\),
\[L = \begin{cases}1 & \text{if prediction \(f(x)\) is correct} \\ 0 & \text{otherwise,}\end{cases}\]
then predictive uncertainty is usually defined as
\[c(x) = P(L = 1) = \text{probability that \(f(x)\) is correct}.\]

\medskip

\textbf{Note}: Here, \(f(x)\) denotes a single prediction from model \(f\), \emph{not} a distribution over predictions. 
\end{definition}

To summarize the above definition, predictive uncertainty tries to measure if we are likely to make an error in our prediction. Most of the ML uncertainty literature specifies two possible typical causes of predicting `I am not sure.' -- i.e., two main sources of predictive uncertainty. First, we give an informal description of these two sources, and then we discuss them in more detail.
\begin{enumerate}
    \item \textbf{Epistemic uncertainty}: ``I am not sure because I have not seen it before.''
    \item \textbf{Aleatoric uncertainty}: ``I have experienced it before, I know what I am doing, but I think there is more than one good answer to your question, so I cannot choose just one.''
\end{enumerate}

Evaluation always requires quantification -- a quantified definition of the concept. Without evaluation, we cannot progress. How should we quantify whether a specific uncertainty estimate is reasonable?  For discussing the basic evaluation of predictive uncertainty, we stick to the scalar confidence values introduced in the definition of predictive uncertainty, where we equate \(c(x)\) to the probability of the prediction's correctness. Predictive uncertainty depends on both the model and the data. In particular, it increases both when the input is ambiguous and when the model is uncertain in its parameters (arising from the undefined behavior in no-data regions of the input space). Some evaluation metrics measure the true likelihood of the model failing and compare that to the given predictive uncertainty estimation. This is a direct way to benchmark predictive uncertainty estimates. In later sections, we will consider exact methods.

\subsection{Aleatoric Uncertainty}

\begin{definition}{Aleatoric Uncertainty}
Aleatoric uncertainty is uncertainty that arises due to the inherent variability or randomness in the data or the environment. This type of uncertainty cannot be reduced by collecting more data or improving the model, as it is an intrinsic property of the system being modeled. Examples of sources of aleatoric uncertainty include measurement noise, natural variability in the data, or incomplete information.
\end{definition}

Intuitively, aleatoric uncertainty translates to ``I do not know because there are multiple plausible answers.'' For a predictive task of predicting \(Y\) from \(X\), aleatoric uncertainty takes place whenever the \emph{true distribution} \(P(Y \mid X = x)\) is non-deterministic (according to human knowledge), thus has a non-zero entropy. We have aleatoric uncertainty when \(Y \mid X = x\) has some entropy. It simply means that a sample \(x\) accommodates multiple possible \(y\)s.\footnote{In many cases, we could equivalently say that we have aleatoric uncertainty when the variance of \(Y \mid X = x\) is non-zero. However, if we want to be precise, we have to consider that variance is undefined for \emph{nominal/categorical} variables.}

\begin{figure}
    \centering
    \includegraphics[width=0.6\linewidth]{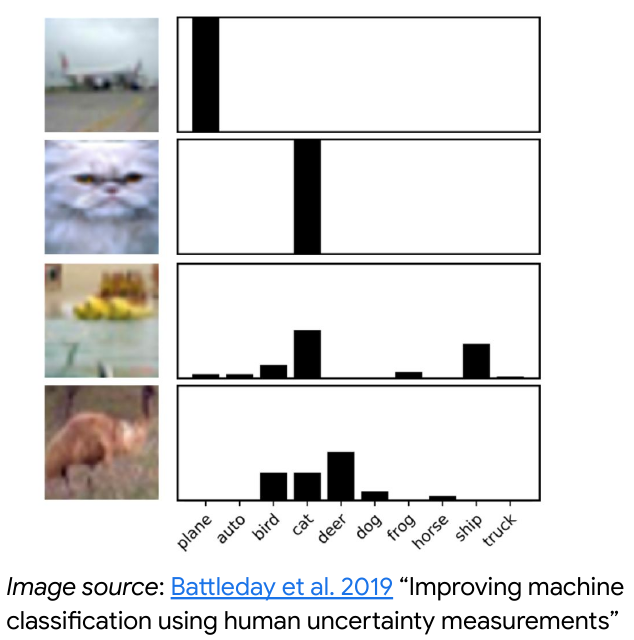}
    \caption{Example of the absence and presence of aleatoric uncertainty. Examples of images and their human choice proportions are given. For many images (upper plane and cat), the label choices are unambiguous. We have very low aleatoric uncertainty, i.e., the true \(Y \mid X = x\) has a very low entropy. The approximation of it by ten human inspectors has no entropy (deterministic); they all agree on the label. The bottom samples accommodate various labels. The single GT label does not always exist. Figure taken from~\cite{balajitalk}.}
    \label{fig:aleatoric}
\end{figure}

Examples from the CIFAR-10H~\cite{peterson2019human} dataset are shown in Figure~\ref{fig:aleatoric}. For some samples (lower ship and bird), humans are quite uncertain, even without time constraints. We have high aleatoric uncertainty; the true \(Y \mid X = x\) (according to human knowledge) has high entropy.\footnote{For general variables, multimodality is perhaps the most extreme case of aleatoric uncertainty. However, for discrete distributions (corresponding to our categorical variable \(Y \mid X = x\) here), multimodality is synonymous with having multiple possibilities, which is synonymous with having a non-zero entropy.} The approximation of it by several human inspectors (47-63 per image for the CIFAR-10H dataset~\cite{peterson2019human}) has a high entropy (non-deterministic). They have disagreements.

\begin{figure}
\centering
\subfloat{
    \includegraphics[width=0.4\textwidth]{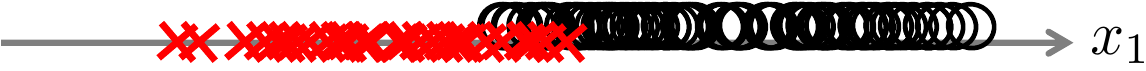}}
\quad
\subfloat{
    \includegraphics[width=0.4\textwidth]{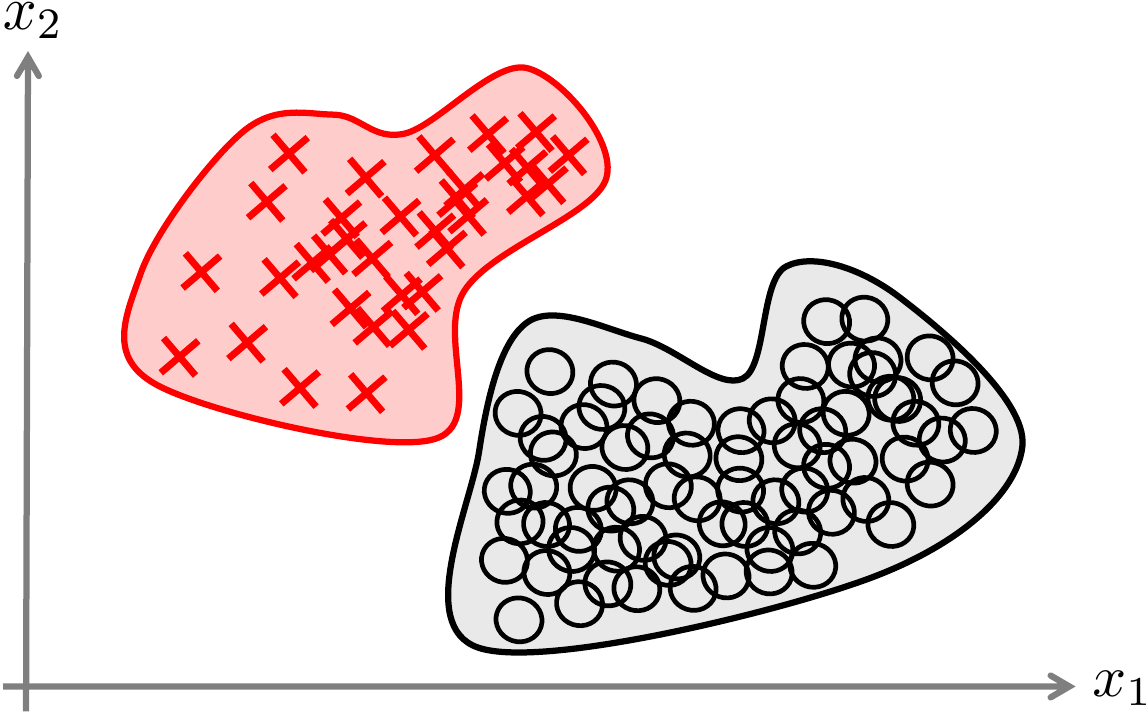}}
\caption{Missing features can introduce overlaps in two classes. When we have both features \(x_1\) and \(x_2\), points of the same label are well separated. For example, we know all the pixels in the N-digit MNIST example. We have no aleatoric uncertainty. As soon as we remove some features (e.g., a part of the image), we have overlaps between the classes. Figure taken from~\cite{H_llermeier_2021}.}
\label{fig:aleatoric2}
\end{figure}

\begin{figure}
    \centering
    \includegraphics[width=0.4\linewidth]{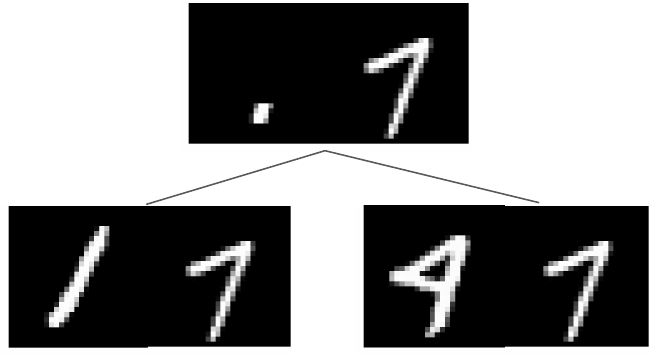}
    \caption{Sample from the N-digit MNIST dataset. There are multiple possibilities for the original image. Figure adapted from from~\cite{https://doi.org/10.48550/arxiv.1810.00319}.}
    \label{fig:hedged}
\end{figure}

\subsubsection{Many faces of aleatoric uncertainty}

First, we consider \emph{ambiguity in the observation}. This can arise, e.g., when features are missing (lack of information). An illustration of how missing features can introduce overlaps in two classes is shown in Figure~\ref{fig:aleatoric2}. This is a typical source of aleatoric uncertainty. We can also take \href{https://arxiv.org/abs/1810.00319}{N-digit MNIST samples}~\cite{https://doi.org/10.48550/arxiv.1810.00319} and consider intentionally corrupted versions of them, shown in Figure~\ref{fig:hedged}. The input goes through corruption/occlusion that removes some features. Then, multiple labels might make sense (e.g., \(41, 11\)). For larger corruptions, we might have
\[P(Y = 41 \mid X = x) = 0.5 = P(Y = 11 \mid X = x).\]
Many people also have poor handwriting, and it is generally difficult to tell a \(1\) apart from a \(7\). No artificial perturbations are required in these cases, as the observation already had an inherent ambiguity.

We can also refer back to the CIFAR images from Figure~\ref{fig:aleatoric}. When the photo of the ship was taken, it went through extra corruption (resolution reduction) to obtain thumbnails. This removes information and introduces aleatoric uncertainty. If objects are seen in the real world (all features are present), then there is probably no ambiguity.\footnote{This is not true for the handwriting case, where even if we see the handwritten digits in real life, we might be unable to tell a \(1\) apart from a \(7\).}

Out-of-focus images are also examples of ambiguity in the observation. Here, we have measurement noise. This also introduces missing features (information). We cannot tell how many people are in the image if it is severely corrupted. 

\begin{figure}
    \centering
    \includegraphics[width=0.3\linewidth]{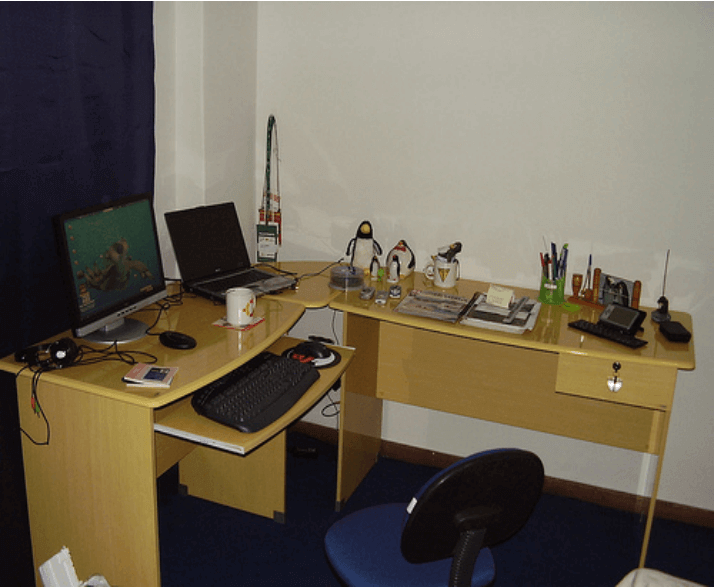}
    \caption{ImageNet-1K sample with label `desk'. Aleatoric uncertainty arises naturally because many objects corresponding to different ImageNet-1K labels are present in the image. There is no \emph{single} good answer to this task, therefore, networks should also not be overconfident in one particular prediction. Figure taken from~\cite{pmlr-v119-shankar20c}.}
    \label{fig:desk}
\end{figure}

Let us now consider \emph{ambiguity in the question}. In general, the task may be formulated so that multiple answers are naturally plausible. In the  ImageNet-1K dataset, there are several such examples. Consider an image of a desk with many objects on it, illustrated in Figure~\ref{fig:desk}. The ImageNet-1K label is `desk', but other ImageNet-1K categories also make sense: `screen', `monitor', or `coffee mug'. It is quite likely, in general, that multiple classes are present on a single image. In such cases, \(P(Y \mid X = x)\) is multimodal. This dataset is not a ``solvable'' problem, as all labels mentioned are plausible, and neither could not be deemed wrong. Annotators, in this case, will arbitrarily choose one category among them. They are only allowed to provide a single label per image. Referring back to the question of whether neural networks will ever become perfect predictors, it is now clear why the answer is negative. Inherent aleatoric uncertainty is \emph{irreducible}, and correct quantification of uncertainty is, therefore, always needed.

\begin{figure}
    \centering
    \includegraphics[width=0.5\linewidth]{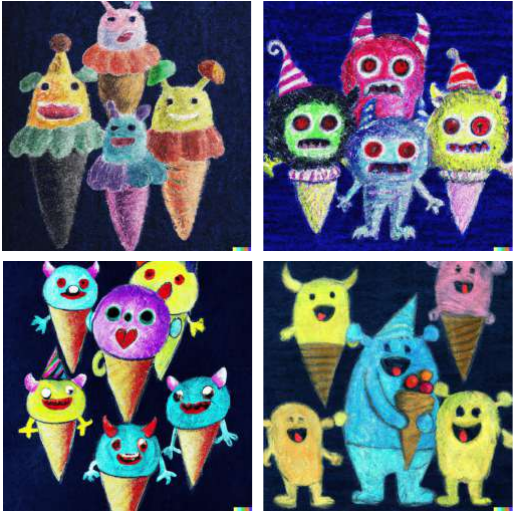}
    \caption{Four samples from DALL-E for the prompt ``crayon drawing of several cute colorful monsters with ice cream cone bodies on dark blue paper''. Each of the synthesized images is a plausible image given the prompt, leading to the presence of aleatoric uncertainty in \(Y \mid X = x\) where \(Y\) is the image and \(x\) is the exact prompt.}
    \label{fig:dalle}
\end{figure}

Another example of inherent ambiguity in the question/task is image synthesis. Consider DALL-E image synthesis for the caption ``\texttt{crayon drawing of several cute colorful monsters with ice cream cone bodies on dark blue paper}'' illustrated in Figure~\ref{fig:dalle}. Here, \(P(\text{image} \mid \text{caption})\) is highly multimodal -- we expect multiple good answers. DALL-E generates multiple plausible outputs for the caption, and all of them make sense. Thus, we have aleatoric uncertainty -- we do not have a single good answer. (Many images fit the caption, as decided by humans.) In a real dataset, we will not see the same caption twice. We do not exactly have this multitude of possible images given the same caption in the dataset, but it can be an indicator if we see a very close caption corresponding to a completely different GT image. This ``approximate multimodality'' of our outputs is also counted as aleatoric uncertainty.

In summary, when we have ambiguities and multiple plausible answers for a task, whatever the source is, we call it aleatoric uncertainty.

\subsubsection{Reducing aleatoric uncertainty}

Unfortunately, we cannot reduce aleatoric uncertainty by observing more data.\footnote{We will soon see the key difference between epistemic and aleatoric uncertainty: epistemic uncertainty \emph{can} be reduced to 0 with an infinite amount of data, sampled from the right distribution \(P(X)\) (considering underexplored regions, too).} When \(Y \mid X = x\) has a non-zero entropy, an infinite amount of data will present data samples with \emph{mixed supervision}. For the same \(x\), different supervision signals \(y\) will be given. Of course, for finite datasets like ImageNet-1K, we do not see the same image with different labels but very similar images with different labels. By seeing ambiguities multiple times, we do not reduce them. The model learns to see similarities between images and gets confused if it sees similar images but with very different labels.

To address aleatoric uncertainty, one must\dots
\begin{enumerate}
    \item \dots formulate a model architecture that accommodates multiple plausible outputs. That is normal for classifiers but not for usual regressors. They usually predict a single number/vector, not a set of plausible answers.
    \item \dots adopt a learning strategy that lets the model learn multiple plausible outputs rather than sticking to one. This is true for the CE loss for classification. However, it is not true for the \(L_2\) loss for regression! It learns the mean of the labels.
\end{enumerate}

Even though aleatoric uncertainty does not depend on the model, the only possible way to approximate it for a general test input is to use a data-driven model. Then, the focus becomes to formulate models that give reliable aleatoric uncertainty predictions. If we know the generative process (i.e., the true distribution \(P(Y \mid X = x)\)) or have multiple samples from it, then we can compare aleatoric uncertainty predictions against the true ``spread'' of \(P(Y \mid X = x)\) or the empirical spread, e.g., by comparing against its variance or entropy. Proxy tasks can also be used for benchmarking aleatoric uncertainty predictions. For example, even though aleatoric uncertainty differs from predictive uncertainty, one might want to evaluate the aleatoric uncertainty predictions on predictive uncertainty benchmarks. One reason is practicality. If we do not have access to \(P(Y \mid X = x)\), benchmarking against predictive uncertainty is better than not benchmarking at all. Another reason is correlation. Predictive uncertainty necessarily monotonically increases by increasing aleatoric uncertainty. If we assume that epistemic uncertainty (discussed in \ref{ssec:epi}) does not vary too much on the test samples, we can use the true predictive uncertainty values as ground truth for \emph{ranking} the test samples, and we can measure how well the ranking based on aleatoric uncertainty estimates agrees with it. This is a strong assumption, and such an evaluation is usually used as a heuristic.

\subsection{Epistemic Uncertainty}
\label{ssec:epi}

Epistemic uncertainty is uncertainty from lack of experience: ``I do not know because I have not experienced it.'' Let us first consider an example of epistemic uncertainty in a binary classification setting to motivate the formalism that follows.

\subsubsection{Example of Epistemic Uncertainty: Training Data for Binary Classification}

\begin{figure}
    \centering
    \includegraphics[width=0.4\linewidth]{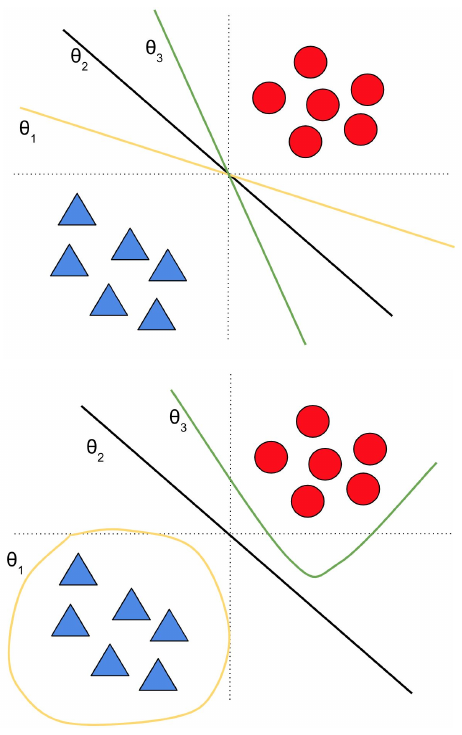}
    \caption{Example of the presence of epistemic uncertainty arising from underexplored data regions. The dataset accommodates many models. Models can be from the same hypothesis class (e.g., linear classifiers in the top subfigure or belong to different hypothesis classes (bottom subfigure). To increase our certainty in the ``correct'' model from the model (= hypothesis) space, we wish to obtain more data from the underexplored regions. Figure taken from~\cite{balajitalk}.}
    \label{fig:epistemic}
\end{figure}

We consider a toy example that showcases the presence of epistemic uncertainty, shown in Figure~\ref{fig:epistemic}. There are several possible classifiers compatible with the data we have observed. While they agree on the data we have observed, we are epistemically uncertain about how to classify points where the models disagree. We wish to sample data from underexplored regions\footnote{We still want to stay on the data manifold -- sampling from underexplored regions that are very implausible is not useful.} to increase our certainty in the choice of the model.

\subsubsection{Formal Treatment of Epistemic Uncertainty}

Let us consider a more formal definition of epistemic uncertainty than the intuitive description given at the beginning of Section~\ref{ssec:epi}.

\begin{definition}{Epistemic Uncertainty}
Epistemic uncertainty is a reducible source of uncertainty that arises due to a lack of knowledge or information. This type of uncertainty can be reduced by collecting more data or improving the model class, as it is a result of the limitations of the current knowledge or understanding of the process being modeled. Examples of sources of epistemic uncertainty include model parameter uncertainty or model \emph{structure} uncertainty.
\end{definition}

During learning, we ``reduce the possible list of models'' to ones that agree with the data (Figure~\ref{fig:bayesian}). One popular way of encoding a ``list of plausible models'' is via the uncertainty over network parameters in \emph{Bayesian machine learning}:
\[\text{\stackanchor{No experience}{Prior over parameters}} \rightarrow \text{\stackanchor{Observations}{Likelihood of data}} \rightarrow \text{\stackanchor{Prediction based on experience}{Posterior over parameters}}\]
We start from our prior knowledge. The prior that encodes our initial beliefs about plausible models is usually broad and has many possibilities for \(\theta\). We have high epistemic uncertainty in regions with no observations, so in the beginning, we have high epistemic uncertainty in general. Then, we accumulate observations. By doing so, we reduce epistemic uncertainty. The likelihood of the data \(\cD\) is a stack of likelihoods of each data point \(X_i\) (IID assumption). By merging our prior knowledge with the observations, we obtain our \emph{posterior beliefs}. Finally, we can make our prediction based on our posterior beliefs using the posterior predictive distribution.

\[P(\theta \mid \cD) \propto P(\theta)P(\cD \mid \theta) \overset{\mathrm{IID}}{=} P(\theta) \prod_{i = 1}^n P(X_i \mid \theta)\]
Typically, the entropy for \(\theta\) decreases with multiple observations.

\subsubsection{Model Misspecification and Effective Function Space}

The uncertainty arising from the restriction of the model class we are learning over (e.g., all linear models or all GPs), i.e., the uncertainty about \emph{choosing the right model family}, is a part of epistemic uncertainty.\footnote{It could also be treated as a separate source of uncertainty when considering a different definition of epistemic uncertainty.} 

\begin{definition}{Model Misspecification}
Model misspecification in ML happens when the inductive biases and prior assumptions injected into the model disagree with the (usually stochastic) process that generated the data.
\end{definition}

We leave model misspecification out in the remainder of the book, always assuming that the model class includes the true \(P(Y \mid X = x)\) so that the epistemic uncertainty can be reduced to 0.\footnote{Most existing uncertainty quantification methods also do not model misspecification as an additional source of uncertainty.} We quickly formalize this below.

\begin{definition}{Function Space}
The function space corresponding to a neural network architecture is the set of all functions we can represent using different parameterizations of the architecture:
\[\cH = \left\{ f_\theta\colon \cX \rightarrow \cY \middle| \theta \in \Theta\right\}\]
where
\begin{itemize}
    \item \(\theta\) is a particular parameterization,
    \item \(\cX\) is the input space and \(\cY\) is the output space,
    \item and \(\Theta\) is the space of all possible parameterizations. For example, for a linear regressor with input \(x \in \nR^n\), \(\Theta = \nR^n\).
\end{itemize}
The above definition does \emph{not} consider the training algorithm, the regularizers, or the optimizer.
\end{definition}

\begin{definition}{Effective Function Space}
The effective function space of a neural network is a subspace of the function space that the network can represent. It is a set of functions that the network can actually learn or achieve, given the training procedure, optimization algorithm, and other hyperparameters.

\medskip

The effective function space of a neural network is influenced by the dataset. For example, a dataset with high noise may require more regularization or early stopping to prevent overfitting, which may limit the effective function space of the network. Conversely, a well-structured and informative dataset may allow the network to explore a wider effective function space as we vary the dataset.

\medskip

The effective function space of a neural network is also influenced by the optimization algorithm and the training procedure. Different optimization algorithms, such as stochastic gradient descent or Adam, may converge to different local minima (or saddle points), which may affect the set of optimal parameters that the network can achieve. Similarly, the training procedure, such as the choice of learning rate, batch size, or data augmentation, may affect the set of optimal parameters that the network can reach.
\end{definition}

When leaving model misspecification out, we can give an alternative definition for epistemic uncertainty: Epistemic uncertainty arises when multiple models out of our \emph{effective} function space can fit the training data well. So, epistemic uncertainty is uncertainty in the \emph{set of plausible models} (but not a property of each individual model). But let's return to the gist of it:

\subsubsection{Epistemic uncertainty is reducible.}

Considering the appropriate distribution, epistemic (= model) uncertainty \emph{vanishes} (reduces to 0) in the limit of infinite data (= observations).\footnote{This depends on what we consider a ``model''. If we consider the models as the parameters, then this statement is subject to model identifiability. For DNNs, because of weight space symmetries and other factors, many models can correspond to the same function. If we equate models to the functions, then this statement always holds.} One can thus completely rule out specific models in the limit, and in fact, we can uniquely determine which model is the right one, i.e., which one ``generated the data''.\footnote{We emphasize that this only holds under the simplifying assumptions we (and many other authors) make in this book; namely that the generative model is contained in the effective function space.}

\begin{definition}{Data Manifold}
Informally speaking, the data manifold is a region of the input space where elements look more natural and realistic.

\medskip

As a more formal definition, data manifold refers to the underlying geometric structure of the (usually high-dimensional) data that is being modeled. It describes the intrinsic, underlying structure of the data in a lower-dimensional space that captures the essential features and relationships between the data points.

\medskip

The data manifold is typically assumed to be smooth and continuous, and it is usually modeled as a lower-dimensional submanifold embedded in the high-dimensional feature space. The dimensionality of the data manifold is determined by the number of intrinsic degrees of freedom in the data, which is almost always lower than the dimensionality of the original feature space, especially in the case of sensory data.
\end{definition}

If the data distribution we sample from does not cover specific areas of interest in the input space (the \emph{data manifold}), then we will still have uncertainty there in limit. It is, therefore, important to sample from \emph{underexplored regions} \(P(X)\) of the data manifold that are still realistic but underrepresented in the original training data to achieve this (OOD samples). However, we do not care about images that are purely Gaussian noise or that are away from the data manifold. As soon as we collect and label many OOD samples, we can reduce epistemic uncertainty as much as we like.\footnote{New types of realistic OOD data (e.g., counterfactual data) did not matter so much before, so they were not collected. This is precisely the reason they \emph{stayed} OOD. With the rising popularity of the field of ML robustness, these samples also matter a lot (refer back to OOD generalization), so we want to perform well on these samples, too.}

\textbf{Example}: Active learning reduces epistemic uncertainty efficiently by acquiring supervision on underexplored samples. We can use epistemic uncertainty to sample from regions where the model needs the most samples. For such a scheme, the model must provide us with well-calibrated epistemic uncertainty estimates.

\subsubsection{Example Sources of Epistemic Uncertainty in Practice}

Let us first consider two possible sources of epistemic uncertainty that often arise in practice.

\textbf{Distribution shifts.} For example, a self-driving car was mostly trained on daylight videos, but it is deployed in a night scenario. On OOD samples, we (usually) have high epistemic uncertainty.

\textbf{Novel concepts.} For example, new objects, words, or classes (open set recognition). These naturally have high epistemic uncertainty (but not always -- this highly depends on the employed inductive biases).

For epistemic uncertainty, many definitions exist (e.g., refer to \cite{shaker2021ensemblebased, valdenegrotoro2022deeper, lahlou2023deup}), and it is not exactly clear what the best way is to properly benchmark such estimates. One possibility is to employ proxy tasks that should be reasonably well correlated with epistemic uncertainty. Another possibility is to consider a binary OOD/not OOD prediction task. This is only a proxy task for epistemic uncertainty because the true ``OOD-ness'' of a sample is independent of any model. However, we still expect epistemic uncertainty to be higher on OOD samples, so the use of such benchmarks is justified to some extent. This is further discussed in \ref{sssec:connection_ood}.

\subsection{Epistemic vs. Aleatoric Uncertainty}

Aleatoric uncertainty is \emph{data uncertainty}. It means there is a \emph{multiplicity of possible answers}. When class-conditional distributions overlap, \(P(Y \mid X = x)\) has a considerable entropy.\footnote{For example, if we have two class-conditional Gaussians, we necessarily have variance/uncertainty in the largely overlapping region, but it reduces considerably outside of this region.} Aleatoric uncertainty is inherent to the data distribution.

Epistemic uncertainty is \emph{model uncertainty}. It means there is a \emph{multiplicity of possible models}. It arises from underexplored data regions. Epistemic uncertainty is inherent to the dataset that allows multiple possible hypotheses.

Treating epistemic and aleatoric sources of uncertainty separately is not only done for philosophical reasons. If we only obtained new samples based on regions with high \emph{predictive} uncertainty, it could very well happen that the epistemic uncertainty was actually \emph{low} in that region but a high \emph{irreducible} value of aleatoric uncertainty caused the high predictive uncertainty. For the sake of intuition, we might consider predictive uncertainty as simply the sum of epistemic and aleatoric uncertainty. Later we will see that the decompositions are not this straightforward and require many assumptions. However, we can still say in general that both aleatoric and epistemic uncertainty influence predictive uncertainty.

In essence, both types of uncertainty arise from the data. While epistemic uncertainty depends on the set of plausible models, aleatoric uncertainty does not, as it only depends on the entropy/variance of the true \(Y \mid X = x\) variable. This also agrees with the statement that epistemic uncertainty is reducible, while aleatoric is inherent to the data generating process and, therefore, is irreducible. However, they both \emph{influence} predictive uncertainty.

In general, it is hard to disentangle general predictive uncertainty into aleatoric and epistemic sources and is an open research topic. \cite{shaker2021ensemblebased, valdenegrotoro2022deeper, lahlou2023deup} 

\begin{figure}
    \centering
    \includegraphics[width=0.8\linewidth]{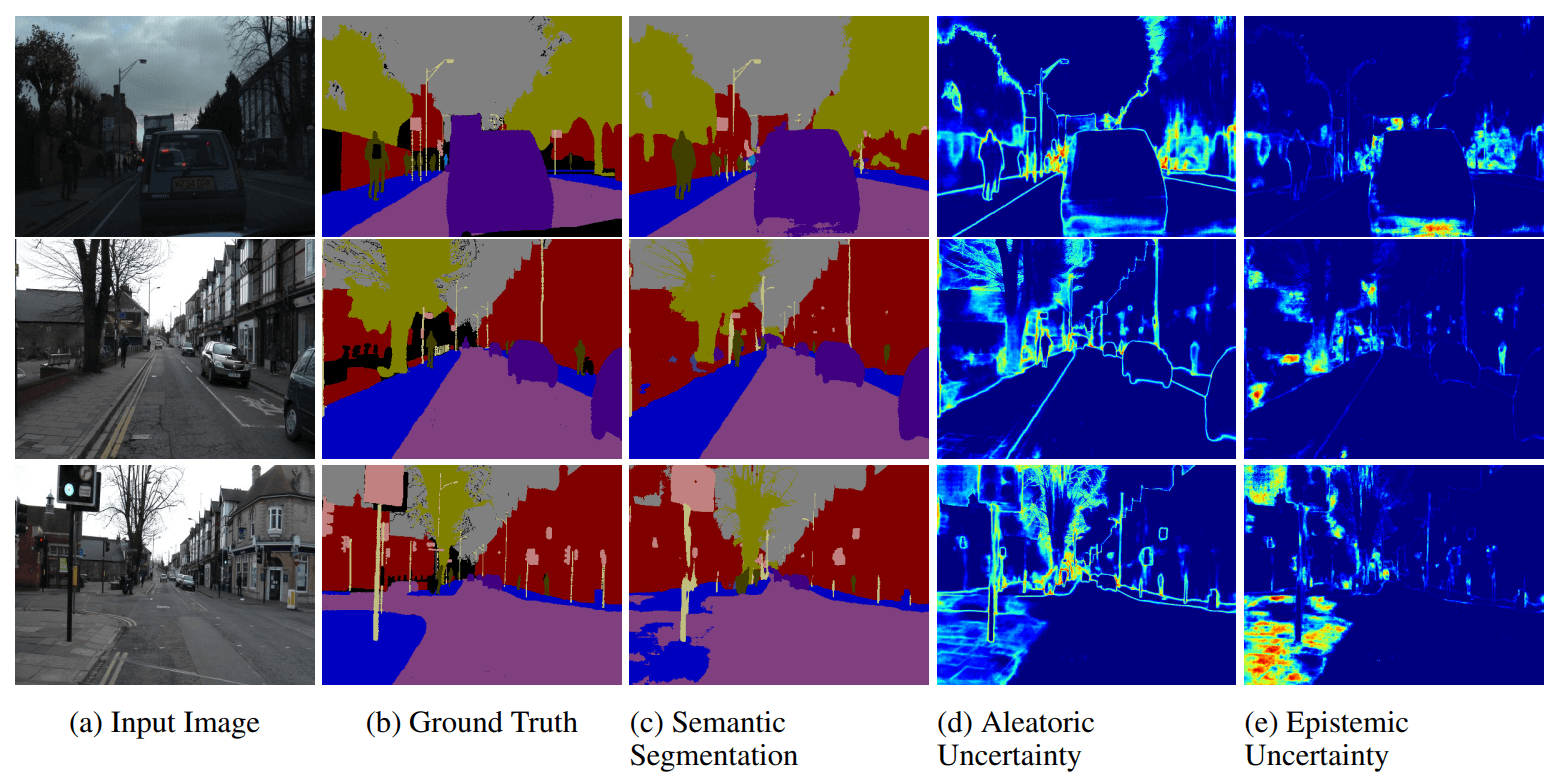}
    \caption{Example application of epistemic and aleatoric uncertainty estimation in computer vision. These two sources of uncertainty are fundamentally different, which is further highlighted by the uncertainty maps in the Figure. Figure taken from~\cite{https://doi.org/10.48550/arxiv.1703.04977}.}
    \label{fig:cv}
\end{figure}

\begin{information}{Epistemic vs. Aleatoric Uncertainty in Computer Vision}
We consider a method that models both epistemic and aleatoric uncertainty in computer vision, introduced in the paper ``\href{https://arxiv.org/abs/1703.04977}{What Uncertainties Do We Need in Bayesian Deep Learning for Computer Vision?}''~\cite{https://doi.org/10.48550/arxiv.1703.04977}. This is illustrated in Figure~\ref{fig:cv}. The task is semantic segmentation, which is pixel-wise classification. This method is capable of measuring epistemic and aleatoric uncertainty at the same time.

\medskip

Aleatoric uncertainty arises at boundaries between classes (e.g., pavement/road). People annotate pixel-wise, and mistakes usually take place around boundaries. Mixed supervision around boundaries leads to high aleatoric uncertainty in these regions.

\medskip

Epistemic uncertainty arises at parts of the image the model has not seen before. It seems that the model has not seen similar pavements before.
\end{information}

\section{Connection of Uncertainty Estimates to Earlier Chapters}

The subfields of Trustworthy Machine Learning are very interconnected. Here, we briefly discuss some of the connections to OOD generalization and explainability.

\subsection{Connection of Epistemic Uncertainty to OOD Generalization}
\label{sssec:connection_ood}

Epistemic uncertainty and OOD generalization have many connections, though they should not be treated interchangeably, as discussed previously. Still, epistemic uncertainty should be high for OOD samples. If we have access to \(M\) models in the form of an ensemble, then the epistemic uncertainty for a sample \(x\) is closely linked to the diversity of predictions \(f_1(x), \dots, f_M(x)\) by the set of trained models. Let us assume that we have a diagonal dataset (Section~\ref{ssec:spurious}) and multiple plausible models that are fit to this dataset. As the models are all trained on the training samples (ID data), they all perform well on the training samples (given sufficient expressivity). However, as the models still differ, they will generally not agree on off-diagonal samples. This is emphasized even more if the models are \emph{regularized} to be diverse. Therefore, the off-diagonal samples will have high output variance (high epistemic uncertainty), and the training samples will have very low output variance (low epistemic uncertainty). We can, therefore, measure epistemic uncertainty by training multiple models and seeing how much they agree on a particular sample. This is the essence of Bayesian ML: training multiple models simultaneously more smartly and efficiently, and checking their divergence on certain test samples.

\subsection{Connection of General Uncertainty Estimation to Explainability}

When a model returns its predictive confidence or other uncertainties and is well-calibrated, it is a great way for the user to learn about the model and the output. Such uncertainty estimates are great explanation tools. Some interesting questions that relate explainability to uncertainty estimation are listed below.
\begin{itemize}
    \item How uncertain was the model?
    \item Due to which factor was the model uncertain? (If there are multiple factors, see above.)
    \item What additional training data will make the model more confident? (What regions suffer from high epistemic uncertainty?)
\end{itemize}

\subsection{Trustworthiness and Confidence Estimates}

A critical component of the trustworthiness of an ML system is the ``truthfulness'' of the confidence estimates \(c(x)\). The most popular demand for \emph{predictive uncertainty} estimates is that \(c(x)\) must quantify the actual probability of the model to get the prediction right (known as true predictive uncertainty). A model needs to address two tasks now: (1): Predicting the GT label \(y\), and (2): Predicting the correctness of the prediction \(L\). Then, we want to obtain \(c(x) = P(L = 1)\).

\section{Formats of Uncertainty}

Let us first consider different approaches people use to represent/estimate uncertainty. What is the appropriate data format for uncertainty? In the following sections, we will refer to confidence and uncertainty ``interchangeably'', with ``\(\text{confidence} = 1 - \text{uncertainty}\)''.

\subsubsection{The simplest form: a scalar.}

The model \(f\) on input \(x\) produces an output \(f(x)\) \emph{and} a scalar confidence score \(c(x) \in [0, 1]\), where
\[c(x) = \text{probability that \(f(x)\) is correct}.\]
This is a type of \emph{predictive uncertainty} which subsumes aleatoric and epistemic uncertainty (and also the model not being expressive enough). 

\textbf{Note}: Whenever we have a scalar in \([0, 1]\), we can treat it as a probability.

\begin{figure}
    \centering
    \includegraphics[width=0.3\linewidth]{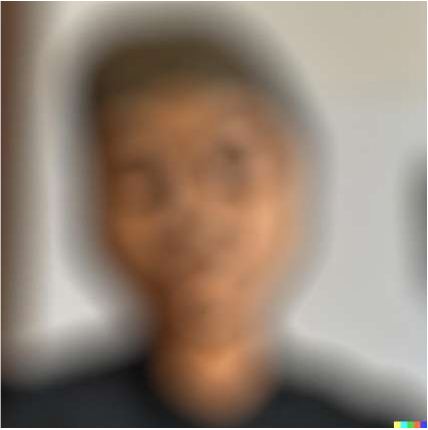}
    \caption{A blurry image of a person, generated by DALL-E. The blurriness corresponds to high aleatoric uncertainty.}
    \label{fig:blur}
\end{figure}

\subsubsection{A vector.}

The model can also report \(c(x) \in \nR^d\), an array of scalars, as a representation of uncertainty. The question we ask is ``Which attributes/features/concepts does the model lack confidence in?'' We attach a confidence value to each attribute (evidence) of the sample.

Consider a \emph{person identification task}. Let the prediction \(f(x) := \text{person name}\). A possible input \(x\) is shown in Figure~\ref{fig:blur}. We might obtain the following confidence values over various evidence:
\begin{align*}
c_{\text{hair color}} &= 0.99 & \text{(we kind of see it)}\\
c_{\text{eye color}} &= 0.39 & \text{(we cannot see it well)}\\
c_{\text{ear shape}} &= 0.1 & \text{(we are not sure at all)}.
\end{align*}
The value $c$ can model predictive uncertainty like here (how sure the model is in the correctness of its prediction, broken down into confidences along various evidence), but analogously also aleatoric uncertainty (how much variance does the true \(Y \mid X = x)\) have along various evidence), or epistemic uncertainty (how much uncertainty there is arising from the lack of observations in the sample along various evidence). For these, different evaluations exist.

\subsubsection{A matrix and a vector.}

This section is inspired by the work ``\href{https://arxiv.org/abs/2101.05068}{Probabilistic Embeddings for Cross-Modal Retrieval}''~\cite{https://doi.org/10.48550/arxiv.2101.05068}. Uncertainty cannot only arise in the outputs of discriminative models that aim to model \(Y \mid X = x\). If we want to embed our data into a lower-dimensional space using probabilistic methods, modeling uncertainty has several advantages. We discuss probabilistic embeddings in Section~\ref{sssec:representation_learning}; here, we only consider the representation of uncertainty.

One can have \(c(x) = \left[\mu_\theta(x), \Sigma_\theta(x)\right]\) interpreted as parameters of a distribution/density. The prediction of the network is a distribution, not a single point. We obtain the posterior over the embedding (probabilistic embeddings), which represents aleatoric uncertainty:
\[P(z \mid x) = \cN\left(\mu_\theta(x), \Sigma_\theta(x)\right)\]
with
\[\Sigma_\theta(x) \in \nR^{D \times D}.\]
The network outputs a Gaussian for each \(x\), just like a Gaussian Process (GP) would. \(\Sigma_\theta(x)\) is a representation of the aleatoric uncertainty in the embedding (covariance of \(\cN\)). This is a more complicated way of uncertainty representation.

\subsubsection{A ``disentangled'' representation.}

We consider the work ``\href{https://arxiv.org/abs/1703.04977}{What Uncertainties Do We Need in Bayesian Deep Learning for Computer Vision?}'~\cite{https://doi.org/10.48550/arxiv.1703.04977} to highlight the possibility to separately obtain aleatoric uncertainty estimations \(c_{\mathrm{al}}(x)\) and epistemic uncertainty estimations \(c_{\mathrm{ep}}(x)\).\footnote{These only approximate the true aleatoric and epistemic uncertainties. Their faithfulness is subject to evaluation.} Then, we can give our \emph{approximate} predictive uncertainty as \(c(x) = c_\mathrm{al}(x) + c_\mathrm{ep}(x)\).

Consider a regression task, and in particular, the problem of monocular depth estimation, where the network has to output per-pixel depth estimates from a single image. Suppose that we have a distribution \(Q(W)\) over the weights \(W\) of the model by using dropout (discussed in detail in Section~\ref{sssec:dropout}), and each model outputs a mean prediction and a variance term that measures aleatoric uncertainty. In the referenced paper, the authors calculate these as
\begin{align*}
c_\mathrm{al}(x) &= \frac{1}{T} \sum_{t = 1}^T \hat{\sigma}_t^2 \approx \nE_q\left[\hat{\sigma}_t^2\right]\\
c_\mathrm{ep}(x) &= \frac{1}{T} \sum_{t = 1}^T \hat{y}^2_t - \left(\frac{1}{T}\sum_{t = 1}^T \hat{y}_t\right)^2 \approx \operatorname{Var}_q\left[\hat{y}\right]
\end{align*}
where \[\left\{\hat{W}_t\right\}_{t = 1}^T \sim Q(W), \qquad\left[\hat{y}_t, \hat{\sigma}_t^2\right] = f^{\hat{W}_t}(x).\]
\(c_\mathrm{al}(x)\) is the average learned spread (variance) of \(Y \mid X = x)\) by the ensemble members and \(c_\mathrm{ep}\) is the variance among the ensemble predictions. Here, \(\hat{y}_t\) is a single output scalar, corresponding to the mean prediction of model \(t\) for a particular input pixel. These uncertainties are calculated for all pixel-wise depth predictions \(\hat{y}_t\) of the different networks \(\left\{f^{\hat{W}_t} \middle| t \in \{1, \dots, T\} \right\}\). Thus, when performing monocular depth estimation, we have as many aleatoric and epistemic uncertainty scalars as there are input pixels.

The Bayesian training for a single input image \(x\) is then performed by minimizing the following loss function. This is learned loss attenuation (attenuating the \(L_2\) loss with the learned weight of error \(\sigma^2\)).
\[\cL_{\mathrm{BNN}}(\theta) = \frac{1}{D} \sum_{i = 1}^D \left[\frac{1}{2\hat{\sigma}_i^2} (y_i - \hat{y}_i)^2 + \frac{1}{2}\log \hat{\sigma}_i^2\right],\]
where \[\hat{W} \sim Q(W), \qquad \left[\hat{y}, \hat{\sigma}^2\right] = f^{\hat{W}}(x).\]
The likelihood is Gaussian and heteroscedastic (\wrt pixels and samples). \(\hat{y}\) is the predicted mean, and \(\hat{\sigma}^2\) is the predicted variance (aleatoric uncertainty), both vectors with as many dimensions as there are pixels. \(q\) is the approximate posterior over the weights modeled by dropout, which corresponds to epistemic uncertainty. Not only does the formulation allow for modeling epistemic and aleatoric uncertainty, but it also improves accuracy.

\section{Proper Scoring Rules}

We discuss a useful and general framework for training and measuring uncertainty estimates: the framework of \emph{proper scoring rules}. Considering the simplest case of scalar uncertainty estimates, we generally want to learn a value \(c(x)\) for a particular sample \(x\) that corresponds to the true probability (be it predictive, epistemic, or aleatoric uncertainty). Luckily, there is a class of scores/losses that \emph{ensures this automatically}.

In subsections \ref{sssec:motivation}, \ref{sssec:logprob}, and \ref{sssec:brier}, we do not make connections to ML concepts, such as the correctness of prediction \(L = 1\). We will simply aim to match a predicted probability \(q\) of a binary event \(Y = 1\) to the true probability \(p\) of it. Later, in subsection \ref{sssec:role}, we will see that this is indeed very useful for matching probabilities corresponding to different sources of uncertainty in neural networks.

\subsection{Motivation: Binary Forecasting Task}
\label{sssec:motivation}

Consider a simple weather forecasting task. We let subjects bet on the probability of rain tomorrow, which is a binary random variable \(Y: \Omega \rightarrow \{0, 1\}\) according to the distribution \(P(Y)\). The prediction is the scalar \(q \in [0, 1]\).  We want to encourage the prediction of the correct probability among people. To this end, we give \(S(q, Y)\) USD to subjects. \(S\) is a function of the reported probability \(q\) and the true outcome \(Y\). \(Y = 1\) if it actually ends up raining and \(Y = 0\) otherwise. Let us assume that the subjects are rational, i.e., they maximize the expected money they get. We want to give the maximum amount of money to people who predict the actual probability of rain. How should we design \(S\)?

The expected reward for the subject is 
\[\nE_{P(Y)} S(q, Y) = S(q, 0)P(Y = 0) + S(q, 1)P(Y = 1),\]
as \(Y\) is a binary random variable. Depending on the actual outcome, we get a different amount of money. We wish to find a function \(S\) such that
\[\max_{q \in [0, 1]} \nE_{P(Y)} S(q, Y)\]
is attained iff \(q = P(Y = 1)\), i.e., the predicted probability truly represents the probability of rain. That is, \[\nE_{P(Y)} S(q, Y) \le \nE_{P(Y)} S(P(Y = 1), Y)\ \forall q \in [0, 1]\]
and the equality implies \(q = P(Y = 1)\). Such a function is called a \emph{strictly proper scoring rule}, formally defined below.

\begin{definition}{Proper/Strictly Proper Scoring Rule}
Let us consider a function \(S\colon \cQ \times \mathcal{Y} \rightarrow \nR\) where \(\cQ\) is a family of probability distributions over the space \(\mathcal{Y}\), called the label space. For a particular distribution \(Q(Y) \in \cQ\) and a sample \(y\) from a GT distribution \(P(Y)\), the function \(S\) outputs a real number.

\medskip\medskip

\textbf{Proper Scoring Rule}

\medskip

\(S\) is called a proper scoring rule iff
\[\max_{Q \in \cQ} \nE_{P(Y)} S(Q, Y) = \nE_{P(Y)} S(P, Y),\]
i.e., \(P\) is one of the maximizers of \(S\) in \(Q\).

\medskip\medskip

\textbf{Strictly Proper Scoring Rule}

\medskip

\(S\) is a \emph{strictly} proper scoring rule iff
\begin{itemize}
    \item \(S\) is a proper scoring rule and
    \item \(\argmax_{Q \in \cQ} \nE_{P(Y)} S(Q, Y) = P\) is the \emph{unique} maximizer of \(S\) in \(Q\).
\end{itemize}

\medskip

\textbf{Note}: The family of distributions \(\cQ\) can be a parameterized distribution with parameters \(\theta \in \nR^n\) that uniquely define the distribution. In this case, the scoring rule can also be defined over the space parameters \(\Theta\) instead of the space of distributions \(\cQ\).
\end{definition}

According to the note in the definition of proper scoring rules, instead of considering the family of Bernoulli distributions \(\cQ\) in the above example, we considered its parameter \(q\) for working with proper scoring rules.

Luckily, many often-used loss functions fulfill this criterion.\footnote{Strictly speaking, the negative loss functions fulfill this criterion, as scores are meant to be \emph{maximized}.} We will look at some examples below.

\subsection{The Log Probability is a Strictly Proper Scoring Rule}
\label{sssec:logprob}

Define
\[S(q, y) \overset{(1)}{:=} \begin{cases}\log q & \text{if } Y = 1 \\ \log(1 - q) & \text{if } y = 0\end{cases} \overset{(2)}{=} y\log q + (1 - y)\log (1 - q).\]
Using this definition, a very confidently wrong prediction gives \(-\infty\) ``reward''. The ``reward'' is non-positive in this case. We can think of it as ``I will take less money from you if you get the prediction right.''

\textbf{Note}: The two expressions (1) and (2) above have different domains. The first one has \(D_{S} = \left((0, 1] \times \{1\}\right) \cup \left([0, 1) \times \{0\}\right),\) whereas the second one has \(D_{S} = (0, 1) \times \{0, 1\}.\) If we take the expectation of both expressions \wrt \(Y\), both domains become
\(D_{\nE_{P(Y)}S} = (0, 1) \times \{0, 1\}.\)

The expected reward for the subject is
\(\nE_{P(Y)} S(q, Y) = P(Y = 0)\log(1 - q) + P(Y = 1) \log q.\)

\begin{claim}
\(S\) defined above is a strictly proper scoring rule.
\end{claim}

\begin{proof}
For the score \(S\) to be well-defined, we have to restrict its domain to \(D_S := (0, 1) \times \{0, 1\}.\)
(Otherwise, we could obtain ``\(0 \cdot -\infty\)'' parts in the expectation below. The case distinction formulation of the score makes \(S(1, 1)\) and \(S(0, 0)\) also well-defined, but the expectation below would \emph{not} be well-defined if we included \(q \in \{0, 1\}\).)

Let \(a := P(Y = 1)\). Then \(\nE_{P(Y)} S(\cdot, Y)\colon (0, 1) \rightarrow \mathbb{R}\),
\begin{align*}
\nE_{P(Y)} S(q, Y) &= P(Y = 0)S(q, 0) + P(Y = 1)S(q, 1)\\
&= (1 - a) \cdot \log(1 - q) + a\cdot \log(q).
\end{align*}

To show that \(S\) defined above is a strictly proper scoring rule, we can leverage the first-order condition for optimality \wrt \(q \in (0, 1)\) when \(a \in (0, 1)\).
\begin{gather*}
\frac{\partial}{\partial q} \nE_{P(Y)} S(q, Y) = -\frac{1-a}{1-q} + \frac{a}{q} \overset{!}{=} 0\\
\iff\\
\frac{a}{q} = \frac{1-a}{1-q}\\
\iff\\
a - aq = q - aq\\
\iff\\
a = q\\
\iff\\
P(Y = 1) = \hat{P}(Y = 1).
\end{gather*}
\(q = a\) is the only stationary point when \(a \in (0, 1)\). To verify that it corresponds to the global maximizer of \(\nE_{P(Y)} S(q, Y)\), we can use the second derivative test:
\[\frac{\partial^2}{\partial q^2} \nE_{P(Y)} S(q, Y) = -\frac{\overbrace{1-a}^{> 0}}{\underbrace{(1-q)^2}_{> 0}} - \frac{\overbrace{a}^{> 0}}{\underbrace{q^2}_{> 0}} < 0,\]
which verifies that \(\nE_{P(Y)} S(q, Y)\) is strictly concave in \(q\) for \(a \in (0, 1)\) and \(q = P(Y = 1)\) is thus the unique maximizer.

Strictly speaking, when \(a \in \{0, 1\}\), there are no stationary points of the above formulation as \(q \in (0, 1)\), according to the domain of the score. However, in these cases, we can trivially simplify \(\nE_{P(Y)} S(q, Y)\), which allows us to extend the domain to allow \(q = a\) even in these extreme cases:
\begin{align*}
a = 0\colon\qquad &\nE_{P(Y)} S(q, Y) = \log(1 - q), &\text{unique maximizer is } q = 0,\\
a = 1\colon\qquad &\nE_{P(Y)} S(q, Y) = \log(q), &\text{unique maximizer is } q = 1.
\end{align*}

This concludes the proof that \(S\) is a strictly proper scoring rule.
\end{proof}

\subsection{The Brier Score is a Strictly Proper Scoring Rule}
\label{sssec:brier}

Define \(S(q, y) := -(q - y)^2\) where \(q\) is our belief in a binary event \(Y = 1\), and \(y\) is an actual outcome of the event (0 or 1) according to random variable \(Y\). The reward is higher when our belief matches the outcome. But in proper scoring maximization, we want to maximize the \emph{expectation} in random variable \(Y\) (and also in \(X\), considering an entire data distribution \(P(X)\) and not just a single sample \(x\)). The expected reward for the subject is
\(\nE_{P(Y)} S(q, Y) = -P(Y = 0)q^2 - P(Y = 1)(1 - q)^2.\)

\begin{claim}
\(S\) defined above is a strictly proper scoring rule.
\end{claim}

\begin{proof}
Analogous to the proof of the log probability being a strictly proper scoring rule.
\end{proof}

\subsection{Role of Proper Scoring Rules}
\label{sssec:role}

A proper scoring rule encourages a subject to report the true probability \(p\) of some binary event \(Y = 1\) as \(q\). As such, it also encourages them to report their true beliefs, as this corresponds to their best approximation of the true probability. Intuitively, it does not make sense to lie. Now we turn away from considering general binary events \(Y = 1\) and consider a use case of proper scoring maximization for ML. In particular, we can use proper scoring maximization to encourage a model to choose its confidence value \(c(x)\) such that it is equal to the probability of getting the prediction for sample \(x\) right (\(L = 1 \iff Y = \hat{Y}\)).

In the case of ML models, predicting the random variable \(L\) implicitly conditioned on \(x\) is a binary classification task of whether we are going to make a correct prediction. The original problem of predicting \(Y \mid X = x\) can be multi-class classification as well.

\subsection{Binary Cross-Entropy for True Predictive Uncertainty}

\begin{definition}{Binary Cross-Entropy (BCE) Loss}
Consider a classifier \(f\colon \cX \rightarrow [0, 1]\) that, for a particular input \(x \in \cX\), predicts the probability of \(x\) belonging to class 1, i.e., \(P(Y = 1 \mid X = x)\). For a GT label \(y\) sampled from \(P(Y \mid X = x)\), the Binary Cross-Entropy (BCE) loss is defined as
\[\cL(f, x, y) = \begin{cases} -\log f(x) & \text{if } y = 1 \\ -\log(1 - f(x)) & \text{otherwise.} \end{cases}\]
This is the most prominent loss for binary classification when training DNNs.
\end{definition}

Consider a binary prediction problem of classifying into classes 0 and 1. Let \(f(x) \in [0, 1]\) be the predicted probability of model \(f\) for class 1 on sample \(x\). It follows that \(1 - f(x) \in [0, 1]\) is the prediction of the model for class 0.
We predict class 1 when \(f(x) \ge 0.5\). Otherwise, we predict class 0.

We define our confidence measure as \(c(x) := \max \left(f(x), 1 - f(x)\right)\), called the \emph{max-probability} or max-prob confidence estimate between classes 0 and 1. It is easy to see that \(c(x) \in [0.5, 1]\). Other confidence estimates also exist, such as entropy-based ones. These also consider probabilities of other classes. (Implicitly, max-prob does, too.)

We wish to make sure that \(c(x)\) estimates the probability of the prediction being correct (\(L = 1\)). As seen in \ref{sssec:logprob}, we can encourage the model to report \(c(x) = P(L = 1)\) (the true predictive uncertainty) by letting the model maximize the log probability proper scoring in expectation of \(L\).
 
\begin{claim}
The negative of the BCE loss is a proper scoring rule for \(c(x) := \max \left(f(x), 1 - f(x)\right)\) to report the true predictive certainty \(P(L = 1)\).
\end{claim}

\begin{proof}
According to the definition of the log probability proper scoring rule,
\[S(c, L) := \begin{cases} \log c(x) & \text{if } L = 1 \\ \log (1 - c(x)) & \text{if } L = 0.\end{cases}\]

One can observe that
\begin{itemize}
    \item \(f(x) < 0.5, Y = 0 \iff L = 1 \land S(c, L) = \log c(x) = \log (1 - f(x))\);
    \item \(f(x) < 0.5, Y = 1 \iff L = 0 \land S(c, L) = \log(1 - c(x)) = \log f(x)\);
    \item \(f(x) \ge 0.5, Y = 0 \iff L = 0 \land S(c, L) = \log (1 - c(x)) = \log (1 - f(x))\);
    \item \(f(x) \ge 0.5, Y = 1 \iff L = 1 \land S(c, L) = \log c(x) = \log f(x)\).
\end{itemize}
Therefore,
\[S(c, L) = \begin{cases} \log c(x) & \text{if } L = 1 \\ \log (1 - c(x)) & \text{if } L = 0\end{cases} = \begin{cases} \log f(x) & \text{if } Y = 1 \\ \log (1 - f(x)) & \text{if } Y = 0.\end{cases}\]
Maximizing the expectation of the above encourages the true predictive uncertainty when our confidence measure is \(c(x) = \max(f(x), 1 - f(x))\). This is exactly the log-likelihood criterion for binary classification. Maximizing this reward on a training set is equivalent to minimizing the BCE loss (negative log-likelihood).
\end{proof}

\textbf{Conclusion}: BCE encourages not only the correctness of classification \(f(x)\) but also the truthfulness of the max-prob confidence \(c(x) = \max (f(x), 1 - f(x))\). BCE is excellent in this regard.

\subsubsection{Remarks for binary cross-entropy}

When the prediction is correct, \(\log c(x)\) reward is given. As \(c(x) \ge 0.5\), we can, at worst, obtain \(\log 0.5\) reward when our prediction is correct. When the prediction is incorrect, but \(c\) is very large, we can obtain an arbitrarily negative reward. We can see the role of aleatoric uncertainty, as \(Y\) is random. We can also see the role of epistemic uncertainty, as \(P(Y = f(x))\) depends on whether the model has seen such a sample already or not.

\textbf{Note}: Looking at the log probability proper scoring rule, one might mistakenly think that naively setting \(c(x) = 1\) is enough to maximize the expected reward on sample \(x\) when the model is correct according to one labeling. However, \(L\) is a random variable because \(L = \bone(Y = f(x))\) and \(Y\) is a random variable. There is an inherent stochasticity in \(L\) whenever \(P(Y \mid X = x)\) has a non-zero entropy: We want \(c(x)\) to maximize the \emph{expected} reward, not just the reward for one particular observation of \(L\).

\subsubsection{Proper Scoring Maximization on Finite Datasets}

When performing ERM, we have no expectation over the loss. We have deterministic \((x, y)\) pairs in our training set and minimize BCE on the batches. (Multiples \emph{can} be present in the dataset with different labels. Very similar inputs can also correspond to different labels. But every \((x, y)\) pair we have is deterministic.) In this case, we have no guarantee of recovering the true predictive uncertainty \(P(L = 1)\) for all samples. We only have the guarantee of recovering the empirical probabilities \(\hat{P}(L = 1)\) based on our dataset. We also have no guarantees of how faithful our predictive uncertainty scores are on unseen (e.g., OOD) samples, as we can arbitrarily overfit our predictive uncertainty predictions. This is important to keep in mind.

Therefore, the truthfulness of the max-prob confidence estimates is only encouraged \wrt the empirical probability of correctness on the training set. When we consider the idealistic case of having infinitely many samples from \(P(X)\) (i.e., we optimize the expectation), then we have the guarantee that \(c(X)\) will recover \(P(L = 1)\) for all samples \(X \sim P(X)\).

By optimizing the BCE, our model also becomes better on the training samples (until a certain point, given by how expressive the model is). Therefore, the well-calibratedness -- as measured by log probability proper scoring -- and the accuracy usually improve hand-in-hand.\footnote{Accuracy is usually highly correlated with the negative loss. However, not all calibration metrics have such a high correlation with accuracy.} We saw above that BCE encourages the prediction of the true probability of correctness. We can consider two corner cases here, depending on the expressivity of our model.
\begin{enumerate}
    \item Consider a shallow model, such as a logistic regression classifier. Further, assume that the dataset's generative model is non-linear; there is model misspecification. Unfortunately, even in the limit of infinite data, training with the BCE loss (and in general with any negative proper scoring rule) \emph{does not ensure} that we get well-calibrated predictive uncertainty estimates. Proper scoring rules only guarantee that they are maximized at the GT distribution in expectation. They do not give any guarantees for calibration when this maximizer cannot be attained in our function class. However, when our estimator is consistent, we are guaranteed to have calibrated predictive uncertainty estimates in the limit of infinite data when using strictly proper scoring rules.
    \item Now, let us assume that we have a very expressive model: one that is capable of fitting to the generative model extremely well. When trained with the BCE loss, in the limit of infinite data, the model will give very accurate predictive uncertainty estimates. If we consider a case with low aleatoric uncertainty, these estimates will be very confident in the model being correct -- and the model will indeed be correct most of the time.
\end{enumerate}
It is hard to create an expressive model using only this criterion that is well-calibrated but inaccurate, as both are optimized simultaneously.

\subsection{Multi-Class Cross-Entropy (CE) for True Predictive Uncertainty}
\label{ssec:ce_pu}

\begin{definition}{Multi-Class Cross-Entropy (CE) Loss}
Consider a classifier \(f\colon \cX \rightarrow \Delta^{K}\) that, for a particular input \(x \in \cX\), predicts an element of the \((K-1)\)-dimensional probability simplex, i.e., predicts a vector of probabilities corresponding to each class. For a GT label \(y\) sampled from \(P(Y \mid X = x)\), the (multi-class) Cross-Entropy (CE) loss is defined as
\[\cL(f, x, y) = -\log f_y(x).\]
This is the most prominent loss for multi-class classification when training DNNs.
\end{definition}

In multi-class classification, we usually use CE as our loss function. We will see that it also encourages the correct predictive confidence.

Let \(f(x) \in \nR^K\) be a vector of probabilities for each class \(k \in \{1, \dotsc, K\}\). That is, \(\forall i \in \{1, \dotsc, K\}\colon\) \(f_i(x) \ge 0\) and \(\sum_{i = 1}^K f_i(x) = 1\). We can define our confidence measure as the max-probability among class probabilities: \(c(x) := \max_k f_k(x).\) Then, just like before, we could apply the log probability proper scoring rule. This rewards the model for how correct it is on its own most likely prediction. But notice the following, using the shorthand \(k_\mathrm{max} := \argmax_k f_k(x)\):
\begin{align*}
&S(c, L)\\
&= \begin{cases} \log c(x) & \text{if } L = 1 \\ \log (1 - c(x)) & \text{if } L = 0\end{cases}\\
&= \begin{cases} \log \max_k f_k(x) & \text{if } Y = k_\mathrm{max} \\ \log \sum_{k \ne k_\mathrm{max}} f_k(x) & \text{if } Y \ne k_\mathrm{max}\end{cases}\\
&= \begin{cases} \log f_Y(x) & \text{if } Y = k_\mathrm{max} \\ \log\left(f_Y(x) + \sum_{k: k \notin \{Y, k_\mathrm{max}\}} f_k(x)\right) & \text{if } Y \ne k_\mathrm{max} \end{cases}\\
&\ge \log f_Y(x).
\end{align*}

The proper scoring rule \(S\) for \(L = 1\) can be bounded from below with \(\log f_Y(x)\), i.e., the log probability the model assigns to the true class.  The negative log probability \(-\log f_Y(x)\) is the CE loss, one of the most widely used losses for training classifiers. Maximizing the lower bound \(\log f_Y(x)\) (minimizing the CE loss) encourages \(c(x) = \max_k f_k(x)\) to be the truthful predictive uncertainty (either \(\hat{P}(L = 1)\) or \(P(L = 1)\), depending on whether we consider the expectation or its Monte Carlo (MC) approximation). While in general, when maximizing a lower bound, we do not have any guarantee that we also maximize the original objective, we can prove just that here: In Section~\ref{ssec:proper_au_pu}, we will prove that this lower bound is \emph{also} a strictly proper scoring rule for the correctness of prediction (thereby saving the CE loss's reputation). In that chapter, we will also uncover important relationships between proper scoring rules for predictive uncertainty and aleatoric uncertainty.

\subsection{Strictly Proper Scoring Rules can Behave Differently}

We have now discovered two strictly proper scoring rules for the correctness of prediction: the log probability of the model's most likely class and the log probability of the true class. Which one should we use? The important bit is that being strictly proper does not necessarily mean that they are also good training objectives. When training deep neural networks, we are solving a highly non-convex optimization problem. Different objectives might induce noisier and more complex loss surfaces: It could be that one of the scoring rules provides a better regularization of the loss surface (which, perhaps, is smoother). In that sense, it is also meaningful to empirically compare the two scores.

\begin{information}{Benchmarking Strictly Proper Scoring Losses}
Let us compare training with the objective
\[\cL_1(f(x), y) = \begin{cases} -\log \max_k f_k(x) & \text{if } y = \argmax_{k} f_k(x) \\ -\log \sum_{k \ne \argmax} f_k(x) & \text{if } y \ne \argmax_k f_k(x),\end{cases}\]
to usual CE training using
\[\cL_2(f(x), y) = -\log f_y(x).\]
This experiment is conducted in the \href{https://colab.research.google.com/drive/1Y5HZSD7lMBulUrraftGP6YTSbxJR_k73?usp=sharing}{linked notebook}.
For a toy dataset like MNIST, a shallow CNN (3 convolutional layers) fits the training data very well with both losses and produces equivalent results across the ECE, log probability, and Brier Score metrics. However, training with \(\cL_1(f(x), y)\) converges slower, even after tuning hyperparameters to have a fair comparison. 

\medskip

\href{https://colab.research.google.com/drive/1OR0KDD9JC2aoBaHK0Fb25leA9X-g3iGS?usp=sharing}{Comparing} the losses on a slightly more realistic dataset, CIFAR-10, the model is not expressive enough to get close to interpolating the training dataset. The network trained with CE achieves an accuracy of around 67\%. The \(\cL_1(f(x), y)\) loss variant converges even slower than before, and plateaus much earlier. Even after hyperparameter tuning, it only reaches an accuracy of 54\% on average. Even though the solution sets are identical, the loss surface corresponding to \(\cL_1(f(x), y)\) is considerably noisier. Regarding calibration, the Brier Score and log-probability scores are higher for the NLL-trained network (which is partly expected because it also has a considerably higher accuracy) but the ECE value for the \(\cL_1(f(x), y)\) loss network is very slightly better. Checking how the uncertainty estimates perform in predicting aleatoric uncertainty would also be a curious research objective.

\medskip

In conclusion, \(\cL_1(f(x), y)\) \emph{can} train a model, but generally with worse accuracy and predictive uncertainty estimates (as measured by proper scoring rules). This might come as a surprise, given that minimizing a proper scoring loss directly tries to optimize the metric we evaluate on. However, numerical optimization can be quite unintuitive and is generally unpredictable. Not all strictly proper scoring rules are equally good training objectives.
\end{information}

\subsection{Multi-Class Brier Score}

Some researchers also report the multi-class Brier score:\footnote{Some people refer to scores even when lower is better. To discuss a unified overview in this book, we refer to scores when we wish to `maximize', and refer to losses when we wish to `minimize'. There is a trivial correspondence between scores and losses when taking reciprocals or negatives.}
\[S(f(x), y) = -(1 - f_{y}(x))^2 - \sum_{k \ne y} f_k(x)^2.\]

\begin{claim}
The above multi-class Brier score provides a lower bound on the Brier score for the max-prob confidence estimate, \(S(c, l) = -(c(x) - l)^2.\) where \(l\) is a realization of the Bernoulli random variable \(L\).
\end{claim}

\begin{proof}
\begin{align*}
S(c, l) &= -(c(x) - l)^2\\
&= \begin{cases} -(c(x) - 1)^2 & \text{if } l = 1 \\ -c(x)^2 &\text{if } l = 0 \end{cases}\\
&= \begin{cases} -\left(\max_k f_k(x) - 1\right)^2 &\text{if } y = \argmax_k f_k(x) \\ -\left(\max_k f_k(x)\right)^2 &\text{if } y \ne \argmax_k f_k(X) \end{cases}\\
&= \begin{cases} -\left(1 - f_y(x)\right)^2 &\text{if } y = \argmax_k f_k(x) \\ -\left(\max_k f_k(x)\right)^2 &\text{if } y \ne \argmax_k f_k(x) \end{cases}\\
&\ge \begin{cases} -\left(1 - f_y(x)\right)^2 &\text{if } y = \argmax_k f_k(x) \\ -\sum_{k \ne y}f_k(x)^2 &\text{if } y \ne \argmax_k f_k(x) \end{cases}\\
&\ge -\left[(1 - f_y(x))^2 + \sum_{k \ne y} f_k(x)^2\right]\\
&= S(f(x), y).
\end{align*}
\end{proof}

Perhaps unsurprisingly, this lower bound is, in fact, also a strictly proper scoring rule for the correctness of prediction. We will show this in Section~\ref{ssec:proper_au_pu}.

\begin{information}{Can learning theory be used for uncertainty guarantees?}
We have not yet seen learning theory used for uncertainty prediction. In learning theory, we have many results based on the 0-1 loss and binary classification. In predictive uncertainty, we also have a binary classification problem: Is the prediction correct or not? However, it is not a standalone classification problem. First, we make a prediction, and then based on that, we can make the meta-output of whether the prediction was correct. It would be interesting to have such results, but it is very underexplored at the moment.
\end{information}

\subsection{Empirical Evaluation of Predictive Uncertainties}

\subsubsection{Using a Test Set to Measure Generalization}

As discussed previously, a good objective does not necessarily imply that the final trained model behaves nicely if we train with that objective. For the training set samples, it trivially does. However, we can still arbitrarily overfit to training set samples (during the optimization, anything can go wrong) and be very confidently wrong on test samples. The model then fails to represent its uncertainty generally. This is already problematic for ERM without uncertainty quantification. So we need some metrics to evaluate the uncertainty estimates on test sets.

\subsubsection{Using Proper Scoring Rules to Evaluate Predictive Uncertainties}

We need empirical evaluation for predictive uncertainty. For empirical evaluation, we always need a sensible evaluation metric. And what metric could be better than one for which we know it achieves its minimum if and only if the prediction is correct? (Strictly) proper scoring rules to the rescue!

\textbf{Log probability.} As the log probability is a strictly proper scoring rule for the correctness of prediction, we can use the average CE (NLL) over the test samples as the evaluation metric (where lower is better) for multi-class classification:
\[\cL_\mathrm{NLL} = -\frac{1}{N_\mathrm{test}}\sum_{i = 1}^{N_\mathrm{test}} \log f_{y_i}(x_i).\]

Luckily, many papers report NLL tables besides, say, accuracy or RMSE. This allows judging the correctness of confidence predictions.

In NLP, people use perplexity instead of CE (especially for language models, used in benchmarks), which is very similar to CE:
\begin{align*}
\cL_\mathrm{NLL} &= -\frac{1}{N_\mathrm{test}}\sum_{i = 1}^{N_\mathrm{test}} \log f_{y_i}(x_i)\\
\cL_\mathrm{Perplexity} &= 2^{-\frac{1}{N_\mathrm{test}}\sum_{i = 1}^{N_\mathrm{test}} \log_2 f_{y_i}(x_i)}
\end{align*}

The perplexity is the exponentiated NLL value, using base 2 in both the exponential and the logarithm.\footnote{The reader can easily convince themselves that the perplexity is independent of the common base of the exponential and logarithm.} It shows the same information but is generally deemed more intuitive because of the following reasons.
\begin{enumerate}
    \item Perplexity can be interpreted as the weighted average branching factor of a language model~\cite{10.5555/555733}. In the context of language models, the branching factor refers to the number of words that can follow a given context (with non-zero probability). The word `weighted' is used because the language model usually assigns different probabilities to different words that can follow -- perplexity takes this into consideration.
    A lower perplexity means the language model is less ``perplexed'' or less uncertain, i.e., it is more confident in its predictions. This intuition can be easier to understand compared to the raw log-likelihood.
    \item Exponentiating with base 2 ``undoes'' the \(\log_2\) operation, bringing the metric back into the probability space.
\end{enumerate}

\begin{figure}
\centering
\includegraphics[width=\linewidth]{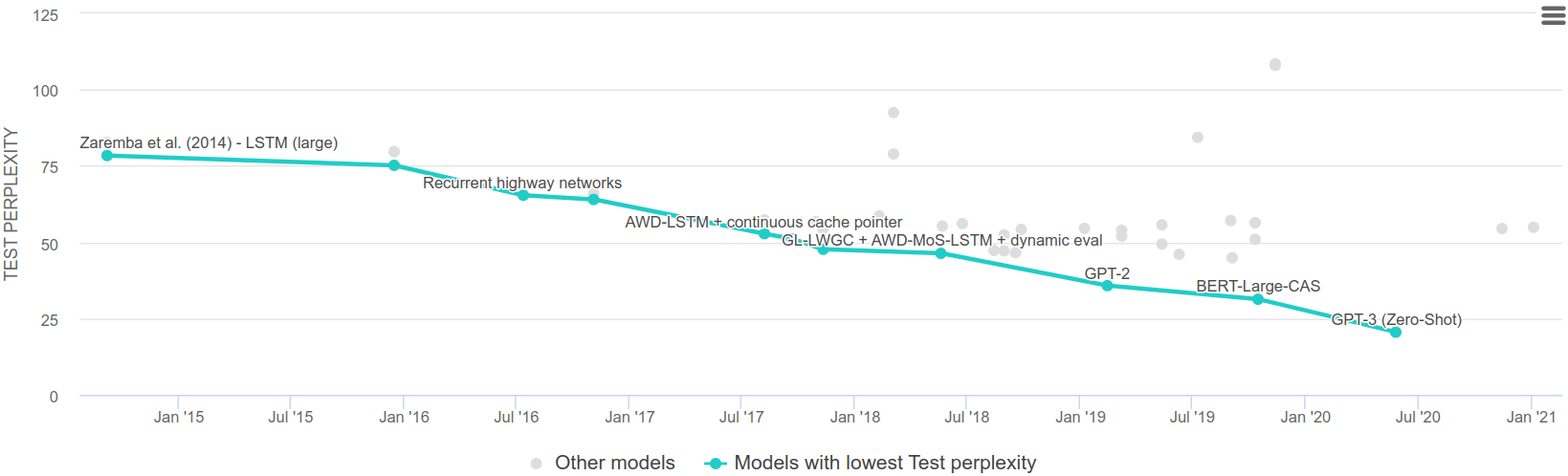}
\caption{Leaderboard of perplexity of Penn Treebank on 04.03.2023~\cite{perplexityleaderboard}. Test perplexity shows a decreasing trend with increasing model capacity.}
\label{fig:perplexity}
\end{figure}

\textbf{Note}: One can verify that larger LLMs seem to have lower test perplexities, meaning they \emph{seemingly} give better predictive uncertainty estimates (Figure~\ref{fig:perplexity}. However, the NLL and perplexity metrics mix calibration with accuracy (see above). Therefore, we should only conclude that larger LLMs fit the data distribution better, which is not a surprising outcome.

\textbf{Multi-class Brier score.} As the multi-class Brier score is also a proper scoring rule for the correctness of prediction, we can evaluate our predictions using the loss
\[\cL_\mathrm{Brier} = \frac{1}{N_\mathrm{test}} \sum_{i = 1}^{N_\mathrm{test}} \left[(1 - f_{y_i}(x_i))^2 + \sum_{k \ne y_i} f_k(x_i)^2\right].\]

\textbf{Remarks for the two previous examples.} The lower \(\cL_\mathrm{NLL}\) and \(\cL_\mathrm{Brier}\) are, the better our predictive uncertainty estimates are. However, there are a few important things to keep in mind.
\begin{enumerate}
    \item We do not know the lowest possible value of these values in expectation over the data generating process. It depends on the aleatoric uncertainty \(P(Y \mid X = x)\) on samples \(X \sim P(X)\).\footnote{If we do not take an expectation but still have mixed supervision (different labels for the same input \(x\)), the lowest possible value is, again, non-zero.}
    \item The NLL can be challenging to interpret. If we take its exponential, then we \emph{roughly} get the average probability assigned to the correct class -- not exactly because of the order of sum and exp. For the correctness of prediction, this still does not give rise to an intuitive explanation. Further, it is unbounded from above and bounded from below by the true aleatoric uncertainty, which is generally unknown.
    The Brier score can be easier to interpret in this regard. 
    \item In general, proper scoring rules for predictive uncertainty using max-prob \emph{mix good calibration with good accuracy}. Notably, this is not the case for ECE (Section~\ref{sec:calibration}) that can capture calibration \emph{independently} from accuracy.
    \item The pointwise Bayes predictor (the predictor with the minimal pointwise risk), \(P(Y \mid X = x)\), is a maximizer of these scoring rules with a max-prob confidence estimate, but it is also a maximizer of proper scoring rules for aleatoric uncertainty. Therefore, epistemic uncertainty is not taken into account -- proper scoring only gives statements in expectation over labels, and the Bayes predictor necessarily has an epistemic uncertainty of zero as it only models aleatoric uncertainty.

\end{enumerate}

\section{A New Notion of Calibration}
\label{sec:calibration}

We have seen that proper scoring rules can be used to define a notion of calibration, but their values are often hard to interpret. In this section, we discuss an easily interpretable notion of calibration. However, we will also see that, unlike proper scoring rules, it can be cheated.

\subsection{Evaluating Calibration}

Let us first discuss how we can \href{https://arxiv.org/abs/1706.04599}{evaluate calibration}~\cite{https://doi.org/10.48550/arxiv.1706.04599}, quantifying it in an alternative way compared to proper scoring rules. Let the input be \(x \in \cX\), the output be \(y \in \cY = \{1, \dots, K\}\) (multi-class classification problem) and the model output be
\[h(x) = (\hat{y}, c(x)),\]
which is a pair of the class prediction and the confidence estimate, respectively. \(c(x)\) does not have to be a max-prob confidence estimate.

\begin{definition}{Perfect Calibration}
A model is \textit{perfectly calibrated} if \(P(\hat{Y} = Y \mid C = c) = c\quad \forall c \in [0, 1].\)
\end{definition}
Intuitively, for confidence level \(c\), the probability of correct prediction should be \(c\), as the confidence level should faithfully reflect the probability of correctness. This is very similar to what we meant by the correct prediction of predictive uncertainty.

\textbf{Example for the empirical probability in practice}: Predictions for any sample in our dataset with confidence score \(c = 0.8\) must only be correct \(80\%\) of the time. A rough outline of a procedure that checks for this (refined later) can be given as follows.
\begin{enumerate}
    \item \textit{Collect all samples in the test dataset with confidence score \(c = 0.8\).}
    \item Compute accuracy across all samples.
    \item Check whether this gives us \(80\%\) accuracy.
\end{enumerate}

\begin{definition}{Model Calibration}
\textit{Model calibration} is defined as
\[\nE_{{c} \sim C}\left[\left|P(\hat{Y} = Y \mid C = c) - c\right|\right] = \int \left|P(\hat{Y} = Y \mid C = c) - c\right| dC(c).\]
\end{definition}
Informally, model calibration quantifies the deviation of our model from perfect calibration. Of course, in practice, we do not have access to the data generating process and, therefore, cannot compute model calibration. If we resort to empirical probabilities, a problem with the rough outline we discussed above is that we never have samples with exactly the same confidence scores, so we cannot calculate the model's accuracy on them this way. An easy fix is to \emph{introduce binning}. The Expected Calibration Error (ECE) metric does exactly that.

\begin{definition}{Expected Calibration Error (ECE)}
\textit{Expected Calibration Error} is a finite approximation of model calibration that uses binning:
\[\mathrm{ECE} = \sum_{m = 1}^M \frac{|B_m|}{n} \left|\mathrm{acc}(B_m) - \mathrm{conf}(B_m)\right|\]
where
\begin{align*}
\mathrm{acc}(B_m) &= \frac{1}{|B_m|} \sum_{i \in B_m} \bone\left(\hat{y}_i = y_i\right),\\
\mathrm{conf}(B_m) &= \frac{1}{|B_m|} \sum_{i \in B_m} c_i.
\end{align*}
\end{definition}
The ECE measures the deviation of the model's confidence predictions from the corresponding actual accuracies on a test set. It is a weighted average of bin-wise miscalibration. \(\mathrm{acc}(B_m)\) is the proportion of correct predictions (the accuracy) in the \(m\)th bin, and \(\mathrm{conf}(B_m)\) is the average confidence in the \(m\)th bin. We take the average of the confidences to ensure we follow the actual confidence values in this range more precisely. Further, we weight by the bin size for the correct approximation of the expectation: \(\hat{C}(c) = \frac{|B_m|}{n}\).

Computing the ECE in practice can be done as follows.
\begin{enumerate}
    \item Train the neural network on the training dataset.
    \item Create predictions and confidence estimates using the test data.
    \item Group the predictions into \(M\) bins (typically \(M = 10\)) based on the confidences estimates. Define bin \(B_m\) to be the set of all predictions \((\hat{y}_i, c_i)\) for which it holds that
    \[c_i \in \left(\frac{m - 1}{M}, \frac{m}{M}\right].\]
    \item Compute the accuracy and confidence of each bin \(B_m\) using the above formulas for \(\mathrm{acc}(B_m)\) and \(\mathrm{conf}(B_m)\).
    \item Compute the ECE by taking the mean over the bins weighted by the number of samples in them.
\end{enumerate}

\begin{information}{Relationship of the above metrics}
What we would ideally want to achieve is that the model returns \emph{truthful predictive uncertainty estimates}, i.e., $c(x) = P(L=1 \mid x) \forall x$. However, that is impossible to measure. So we measure a necessary (not sufficient!) condition: If the model always returns truthful predictive uncertainty estimates, then it also needs to be \emph{perfectly calibrated} (across all $x$ that have the same $c(x)$. 

\medskip

This condition is quantified by the \emph{model calibration}: The model calibration is zero if and only if the model is perfectly calibrated. To measure this in practice, we need to approximate it by the \emph{ECE}. This is basically a discretized version of the model calibration integral.

\medskip

Due to the approximation, we cannot theoretically guarantee that an ECE of 0 implies a model calibration of 0 or vice versa (and, in fact, we show how to game both below). But an ECE close to zero means the model calibration should also be close to zero. This, in return, at least checks one of the boxes a model with truthful predictive uncertainties has to fulfill. It is the best we can do in practice.
\end{information}

While the ECE is a useful metric, for high-risk applications we might be interested in worst-case metrics. The \emph{Maximum Calibration Error} computes such a worst-case discrepancy.

\begin{definition}{Maximum Calibration Error}
The \textit{Maximum Calibration Error} is a useful metric for high-risk applications:
\[\mathrm{MCE} = \max_{m \in \{1, \dotsc, M\}} \left|\mathrm{acc}(B_m) - \mathrm{conf}(B_m)\right|.\]
\end{definition}

MCE computes the maximal bin-wise miscalibration (difference between empirical accuracy and average confidence value). This might be a very pessimistic metric if for \[m' := \argmax_{m \in \{1, \dotsc, M\}} \left|\mathrm{acc}(B_m) - \mathrm{conf}(B_m)\right|,\] \(\frac{|B_{m'}|}{M}\) is very small, depending on our end goal. For high-risk applications, we could also define the worst-case ECE per class if our concern is per-class performance.

\subsection{Gaming the ECE Metric}

ECE is usually a good \emph{indicator} of whether something is fairly well-calibrated. Its main advantage is that ECE scores are often more interpretable and intuitive than proper scoring rules, as they denote deviations from the perfect calibration in a bounded manner: The ECE is a number between 0 and 1. It tells us how much we are deviating from the \(x = y\) line \wrt a weighted average. In comparison, NLL scores can be arbitrarily large. When we consider the log probability, the sign flips, which can be confusing. We cannot immediately tell what is good or bad. It is difficult to interpret what the numbers mean, and it heavily depends on the scoring rule of choice. 

Although it has many nice properties, \emph{the ECE is not a proper scoring rule}. One can easily achieve \(\text{ECE} = 0\) (the minimal value) even when the model is not reporting the true predictive uncertainties. This can give us a false sense of calibration and can kill the purpose of the metric. In particular, if we predict a constant \(c\) for all samples, where \(c = P(\hat{Y} = Y)\) is the global accuracy of the model on the data distribution. Then the conditional probability is only defined for \(c = P(\hat{Y} = Y)\), as this is the only value with a positive measure (i.e., we have a Dirac measure at the global accuracy), and for this value, the definition holds by construction. To game the ECE metric, one does not even need access to labeled validation data. All one needs to know is the prior probability of correctness, \(P(\hat{Y} = Y)\). The same trick can game the more theoretical notion of model calibration.

Therefore, perfect calibration does \emph{not} imply that \(c(x) = P(L = 1)\), i.e., that \(c(x)\) is the GT probability of predicting the output correctly for all individual inputs \(x\). Predictive uncertainties can be arbitrarily incorrect per sample (\(c(x) \ne P(L = 1)\)). This is because the \emph{conditional} probability $P(\hat{Y} = Y \mid C=c)$ \emph{aggregates} all samples with the same value $c(x)$. As long as this group has the correct accuracy on average, it is considered perfect. The intention of ECE and related metrics is still to ensure \(c(x) = P(L = 1)\), but they fail to fully encode this requirement. This can be exploited to, e.g., win competitions and benchmarks.

Another important drawback of the ECE metric is that it depends on the binning. Using twenty bins gives us a different score than using ten. There should be an agreed-upon number of bins across papers and methods. This is usually ten but there are several papers using different numbers as well. However, fixing it is probably not a good idea in the long run: There are pros and cons of fixing the number of bins. Eventually, models will be making more and more correct predictions. We should probably make binning more fine-grained near the \(90\% - 100\%\) confidence range, as there will probably be a lot more samples there.

\subsection{Reliability Diagrams}

\begin{figure}
    \centering
    \includegraphics[width=0.5\columnwidth]{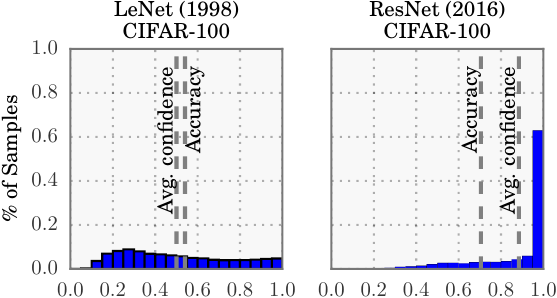}
    \includegraphics[width=0.5\columnwidth]{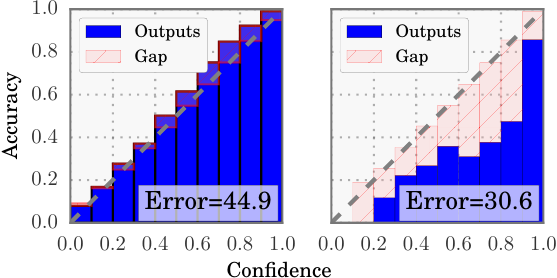}
    \caption{Example reliability diagrams (bottom Figures) with confidence histograms (top Figures). Error refers to the Top-1 accuracy of the models, not the calibration error. For each bin, \(\mathrm{acc}(B_m)\) in blue and \(\mathrm{conf}(B_m) - \mathrm{acc}(B_m)\) in red are plotted as a barplot. The well-calibrated model's (left) gaps are nearly all smaller than the less well-calibrated model's (right) gaps. In the right plot, the bin indices and accuracies for the bins do not match well. The ResNet is much more accurate and also much less well-calibrated. Figure taken from~\cite{https://doi.org/10.48550/arxiv.1706.04599}.}
    \label{fig:reliability}
\end{figure}

Instead of quantifying calibration in a single number, we can also \emph{visualize} how well-calibrated a model is by leveraging \emph{reliability diagrams} (Figure~\ref{fig:reliability}).

\begin{definition}{Reliability Diagram}
A reliability diagram is a visualization of model calibration that uses binning. It is calculated as follows.
\begin{enumerate}
    \item Bin through different confidence values and take the mean accuracy per bin on the test set: for each bin, calculate \(\mathrm{acc}(B_m)\) and \(\mathrm{conf}(B_m) - \mathrm{acc}(B_m)\) as defined previously.
    \item Visualize the discrepancies between the bin-wise accuracies and confidences using a barplot.
\end{enumerate}
\end{definition}

Reliability diagrams allow us to judge whether a model is under- or overconfident (or a mixture). While ECE only concerns the distance to the true \(c\), the diagram tells us whether the actual accuracy is higher or lower than the model predicts. If the line is above, then the model is \emph{underconfident}. If it is below, it is \emph{overconfident} (as in Figure~\ref{fig:reliability}).

Reliability diagrams also allow us to look at the MCE, while ECE can often hide that. But they do not allow inferring the ECE because we do not know the bin sizes (the weights). Seemingly large discrepancies might be weighted with a negligible weight if only a couple of samples are in those bins. If the model on the right had a tiny gap for the last bin, it could have a lower ECE value than the one on the left. Even if the weights are reported as histograms along with the reliability diagrams (the original paper did this, but most follow-ups drop this), the reliability diagram might still give the wrong impression \emph{at first glance}. 

\begin{figure}
    \centering
    \includegraphics[width=0.5\linewidth]{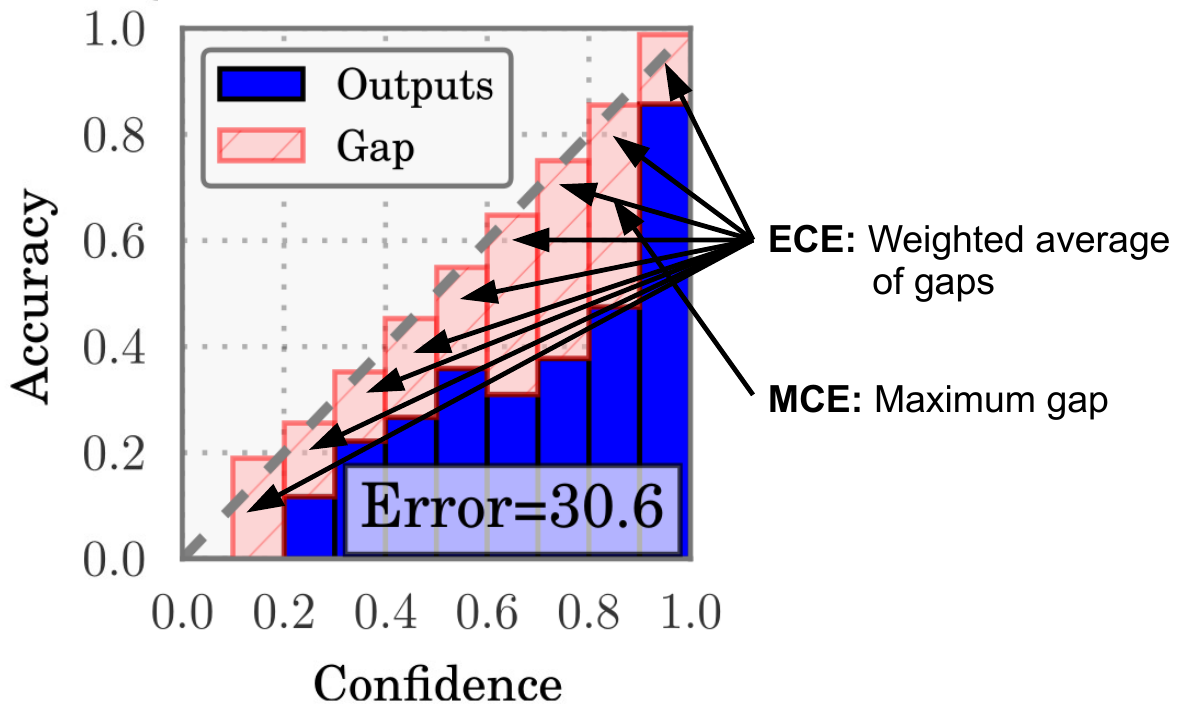}
    \caption{Connection between the reliability diagram and the ECE, MCE scores. Accuracy: \(P(\hat{Y} = Y \mid C = c)\), confidence: \(C = c\). Figure taken from~\cite{fluri}.}
    \label{fig:connection}
\end{figure}

The connection between reliability diagrams and the ECE and MCE scores can be seen in Figure~\ref{fig:connection}.
Note that the plot starts at 0.1 and not at 0. This is not a coincidence: If we use the max-prob class as a prediction, its lowest possible \(c\) can only be \(1/K\). This becomes even more visible when we only have 10 or 2 classes.

For binary classification, there is also a second definition of reliability where the y-axis shows the probability of the positive class. Thus, it always starts at 0 and does not include the mind-flip that the confidence may also be the probability of the 0 class. However, it requires a different mind-flip: An underconfident model, in this case, would have an S-shaped diagram. In the definition of~\cite{https://doi.org/10.48550/arxiv.1706.04599} above, an underconfident model has a curve that is always above the line. This version of a reliability diagram is common in traditional statistics, where classes are not equal, but the 1 class is more important. So, if one sees a binary reliability diagram, it is better to double-check its axis labels.

\section{Summary of Evaluation Tools for the Truthfulness of Confidence}

Let us provide a collection of evaluation tools for the truthfulness of confidence (predictive uncertainty).

\subsubsection{Proper Scoring}

As we have seen before, one can use the negative log-likelihood (NLL) loss or the log probability scoring rule on the test dataset to evaluate the truthfulness of predictive uncertainty estimates. Similarly, one can use the Brier score or its multi-class variant on a test dataset. These are all proper scoring rules/losses for the correctness of prediction.\footnote{The NLL loss and the multi-class Brier score are also strictly proper for aleatoric uncertainty (i.e., the recovery of \(P(Y \mid X = x)\)), as we will see in Section~\ref{ssec:au_classification}.}

\subsubsection{Metrics Based On Model Calibration}

One can use the ECE score for an expected deviation from perfect calibration (in a binned fashion). For high-risk applications where we are concerned with the ``worst-case bin,'' one can also employ the MCE score.

It is also possible to visualize calibration by using reliability diagrams. However, it is also important to plot confidence histograms, as reliability diagrams alone can be misleading.

These metrics/visualization tools are all used for predictive uncertainty (correctness of prediction \(L = 1\)).

\section{Excourse: How well-calibrated are DNNs?}

Let us consider some findings from the literature on DNN calibration.

\subsection{On Calibration of Modern Neural Networks}

We discuss the seminal paper titled ``\href{https://arxiv.org/abs/1706.04599}{On Calibration of Modern Neural Networks}''~\cite{https://doi.org/10.48550/arxiv.1706.04599}. In particular, we refer to Figure~\ref{fig:reliability}. Both LeNet and ResNet are trained with the NLL loss, which is the negative of a lower bound of a proper scoring rule for multi-class predictive uncertainty under max-prob. According to the Figure, LeNet is relatively well-calibrated, and ResNet performs worse than LeNet regarding calibration.

It is important to note that this finding is not a general observation. ResNet-50s usually perform well on calibration benchmarks~\cite{galil2023learn}. Training procedures and best practices since this work have also improved considerably, which might have compounding effects on the results shown in Figure~\ref{fig:reliability}.

\begin{information}{The Use of ResNets in Modern DL}

In medium-sized models, ResNets are still among the top performers (see the ``\href{https://arxiv.org/abs/2302.11874}{What Can We Learn From The Selective Prediction And Uncertainty Estimation Performance Of 523 Imagenet Classifiers}'' paper~\cite{galil2023learn}. They are often used in practice as ``the smallest possible model that still allows experimenting with DL.''
\end{information}

\begin{figure}
    \centering
    \includegraphics[width=\linewidth]{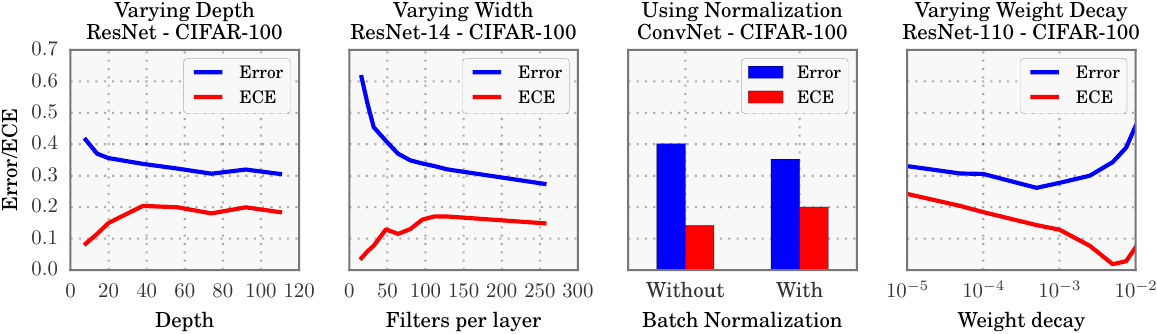}
    \caption{Influence of depth, filters per layer, batch normalization, and weight decay on the error and calibration of different ConvNet architectures. Figure taken from~\cite{https://doi.org/10.48550/arxiv.1706.04599}.}
    \label{fig:miscal}
\end{figure}

\textbf{Why is this the case?} Let us consider Figure~\ref{fig:miscal}. Greater model capacity is known to improve model generalizability~\cite{goodfellow2016deep}. We can see a decrease in error as the capacity increases.\footnote{Curious readers might find the phenomenon of benign overfitting in the highly overparameterized regime interesting.} However, it also leads to greater miscalibration. We can see an increase in ECE. In particular, increasing the depth or the number of filters per layer (``width'') both result in decreased calibration.

\begin{figure}
    \centering
    \includegraphics[width=0.6\linewidth]{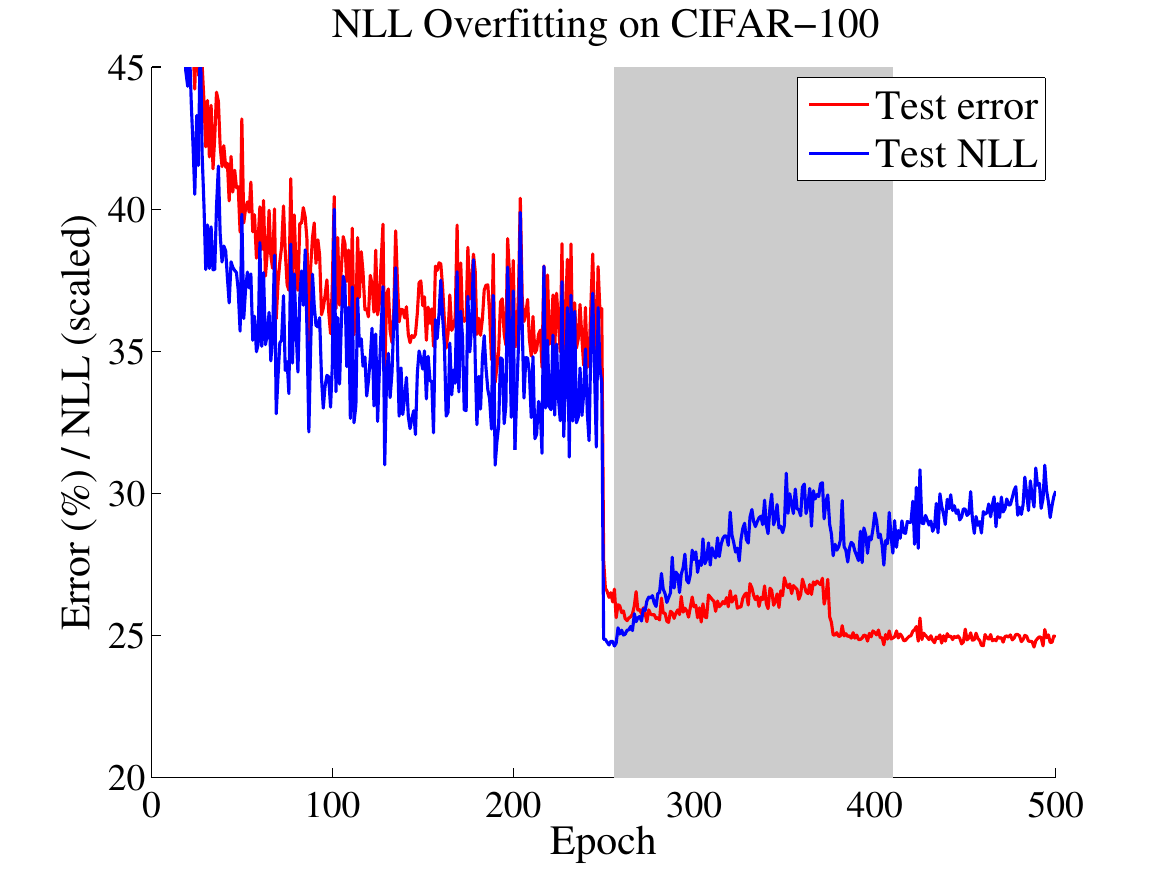}
    \caption{Test error and NLL of ResNet-110 over a training run. While the test NLL starts to overfit (i.e., uncertainty estimates become less calibrated), the error keeps decreasing. NLL is scaled in order to fit the Figure. Note the scheduled LR drop at epoch 250. Figure taken from~\cite{https://doi.org/10.48550/arxiv.1706.04599}.}
    \label{fig:miscal2}
\end{figure}

Let us now turn to Figure~\ref{fig:miscal2}. We can measure predictive uncertainty faithfulness with the test NLL. At epoch 250, we have a scheduled LR drop. Both the test error and test NLL decrease a lot. The grey area is between epochs in which the best validation loss and validation error are produced. The test NLL tends to increase after epoch 250. It shows the overfitting of \(c(x)\) to the training samples. It does not go back to epoch 250 levels, not even after the scheduled LR drop at epoch 375. The test error also shows a little overfitting, as it increases by \(1-2\%\) after epoch 250. However, it drops again after the scheduled LR drop at epoch 375, surpassing epoch 250 levels. The authors draw the following conclusions. ``In practice, we observe a disconnect between NLL and accuracy, which may explain the miscalibration in [Figure~\ref{fig:miscal}]. This disconnect occurs because neural networks can overfit to NLL without overfitting to the 0-1 loss. We observe this trend in the training curves of some miscalibrated models. [Figure~\ref{fig:miscal2}] shows test error and NLL (rescaled to match error) on CIFAR-100 as training progresses. Both error and NLL immediately drop at epoch 250, when the learning rate is dropped; however, NLL overfits during the remainder of the training. Surprisingly, overfitting to NLL is beneficial to classification accuracy. On CIFAR-100, test error drops from 29\% to 27\% in the region where NLL overfits. This phenomenon renders a concrete explanation of miscalibration: the network learns better classification accuracy at the expense of well-modeled probabilities. We can connect this finding to recent work examining the generalization of large neural networks. Zhang et al. (2017) observe that deep neural networks seemingly violate the common understanding of learning theory that large models with little regularization will not generalize well. The observed disconnect between NLL and 0-1 loss suggests that these high capacity models are not necessarily immune from overfitting, but rather, overfitting manifests in probabilistic error rather than classification error.''~\cite{https://doi.org/10.48550/arxiv.1706.04599}

\subsection{Modern Results on Model Calibration}

\begin{figure}
    \centering
    \includegraphics[width=\linewidth]{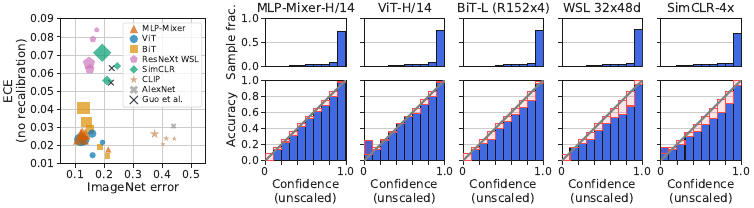}
    \caption{The ViT, BiT, and MLP-Mixer architectures are well-calibrated and accurate. \emph{Left.} ECE is plotted against classification error on ImageNet for various classification models. \emph{Right.} Confidence distributions and reliability diagrams of various architectures on ImageNet. ``Marker size indicates the relative model size within its family. Points labeled
``Guo et al.'' are the values reported for DenseNet-161 and ResNet-152 in Guo et al. (2017).''~\cite{https://doi.org/10.48550/arxiv.2106.07998} Figure taken from~\cite{https://doi.org/10.48550/arxiv.2106.07998}.}
    \label{fig:vit}
\end{figure}

For more recent models,~\cite{galil2023learn} provides an extensive calibration analysis. Several MLP-Mixers~\cite{tolstikhin2021mlpmixer} (fully connected vision models), ViTs~\cite{https://doi.org/10.48550/arxiv.2010.11929} (vision transformers), and BiTs~\cite{kolesnikov2020big} (ResNet-based models) are among the most calibrated \emph{and} accurate models, considering both the NLL loss and the ECE. In particular, knowledge-distilled variants of these usually perform better. This disagreement with the previous study shows that there is no unanimous agreement on the matter of model calibration in the literature.

ViT and Mixer are \href{https://arxiv.org/abs/2106.07998}{reported to be well-calibrated}~\cite{https://doi.org/10.48550/arxiv.2106.07998} in other works as well, however, as shown in Figure~\ref{fig:vit}. Notably, no recalibration is performed for the Figure. ``Several recent model families (MLP-Mixer, ViT, and BiT) are both highly accurate and well-calibrated compared to prior models, such as AlexNet or the models studied by
Guo et al. (2017). This suggests that there may be no continuing trend for highly accurate modern
neural networks to be poorly calibrated, as suggested previously. In addition, we find that a recent zero-shot model, CLIP, is well-calibrated given its accuracy.''~\cite{https://doi.org/10.48550/arxiv.2106.07998}

Calibration depends a lot on the architecture family. There are huge differences even between ConvNet-variants.

\textbf{Remark}: The decrease in ECE values for recent NN-variants \emph{could} also be attributed to them being trained on more data. However, the authors of~\cite{https://doi.org/10.48550/arxiv.2106.07998} find that ``Model size, pretraining duration, and pretraining dataset size cannot fully explain differences
in calibration properties between model families.'' (Well-calibratedness has a lot to do with overfitting. Increasing the number of training samples could result in better ECE on its own. However, this is apparently not the deciding factor.)

``The poor calibration of
past models can often be remedied by post-hoc recalibration such as temperature scaling (Guo et al.,
2017), which raises the question of whether a difference between models remains after recalibration. We
find that the most recent architectures are better calibrated than past models even after temperature scaling.''~\cite{https://doi.org/10.48550/arxiv.2106.07998}

\subsection{Easy Fix for Better ECE: Temperature Scaling}

Let us discuss \href{https://arxiv.org/abs/1706.04599}{Temperature Scaling}~\cite{https://doi.org/10.48550/arxiv.1706.04599}. For DNN classifiers, one could fix their calibration via post-processing on the softmax outputs. Suppose that the model output \(f(x)\) is the result of a softmax operation over logits \(g(x)\):
\[f(x) = \operatorname{softmax}(g(x)) \in \nR^K.\]
Softmax converts the logits to parameters of a categorical distribution. We define temperature scaling with the temperature \(T > 0\) as follows:
\[f(x; T) = \operatorname{softmax}(g(x) / T).\]
In words, we divide each logit value by \(T\).

When \(T \downarrow 0\), the elements of the argument of the softmax explode to infinity, the differences between the \(\argmax\) and the other elements increase more and more. Thus, the output of softmax, \(f(x; T)\), becomes a one-hot vector. (As the difference grows, we are stressing the argmax value more and more.)

When \(T \rightarrow \infty\), the elements of the argument of the softmax go to 0. The differences between the elements decrease more and more. Thus, the output of softmax, \(f(x; T)\), becomes uniform.

One can find the \(T > 0\) that returns the best ECE score over a validation set. We let the model's predictive confidence be
\[\left\{\max_k f_k(x_i; T)\right\}_{i = 1, \dots, N_\mathrm{val}}\]
over the validation set and search for the \(T > 0\) that minimizes the ECE. We can perform a grid search over different \(T\) values and find the one that works best.

\textbf{Temperature scaling improves calibration quite dramatically.} Results are shown in Table~\ref{tab:temp}. \(T = 1\) usually results in suboptimal ECE results; the models are not well-calibrated. \(T = T^*_\mathrm{val}\) (after performing the search over the val set) results in sub-\(2\%\) ECE values in general, whereas before the average was around \(8\)-\(10\%\). This is a nice and easy fix.\footnote{ECE here is calculated with 15 bins. We can already see that \(M = 10\) is not consistently applied through papers, though it is a popular choice.}

\begin{table}
    \centering
    \caption{Comparison of temperature scaling with an untuned baseline. Temperature scaling can lead to a drastic improvement in calibration. Table adapted from~\cite{https://doi.org/10.48550/arxiv.1706.04599}.}
    \label{tab:temp}
    \begin{tabular}{c c c c}
    \toprule
              Dataset &            Model & Uncalibrated (\(T = 1\)) & Temp. Scaling (\(T = T_\text{val}^*\)) \\
    \midrule
                Birds &        ResNet 50 &        9.19\% &         {\bf1.85\%}\\
                 Cars &        ResNet 50 &         4.3\% &         2.35\%\\
             CIFAR-10 &       ResNet 110 &         4.6\% &         0.83\%\\
             CIFAR-10 &  ResNet 110 (SD) &        4.12\% &         {\bf0.6\%}\\
             CIFAR-10 &   Wide ResNet 32 &        4.52\% &         {\bf0.54\%}\\
             CIFAR-10 &      DenseNet 40 &        3.28\% &         {\bf0.33\%}\\
             CIFAR-10 &          LeNet 5 &        3.02\% &         {\bf0.93\%}\\
            CIFAR-100 &       ResNet 110 &       16.53\% &         {\bf1.26\%}\\
            CIFAR-100 &  ResNet 110 (SD) &       12.67\% &         0.96\%\\
            CIFAR-100 &   Wide ResNet 32 &        15.0\% &         {\bf2.32\%}\\
            CIFAR-100 &      DenseNet 40 &       10.37\% &         1.18\%\\
            CIFAR-100 &          LeNet 5 &        4.85\% &         {\bf2.02\%}\\
             ImageNet &     DenseNet 161 &        6.28\% &         {\bf1.99\%}\\
             ImageNet &       ResNet 152 &        5.48\% &         {\bf1.86\%}\\
                 SVHN &  ResNet 152 (SD) &        0.44\% &         0.17\%\\
                  \midrule
              20 News &            DAN 3 &        8.02\% &         4.11\%\\
              Reuters &            DAN 3 &        0.85\% &         0.91\%\\
           SST Binary &         TreeLSTM &        6.63\% &         1.84\%\\
     SST Fine Grained &         TreeLSTM &        6.71\% &         2.56\%\\
    \bottomrule
    \end{tabular}
\end{table}

\section{Do we really need proper scoring?}

\subsection{Ranking Condition}

The previous proper scoring rules for the correctness of prediction demanded that $c(x) = P(L = 1 \mid X = x)$ be their optimal value, i.e., that confidences directly give the probabilities of correctness. Calibration followed a similar principle. Let us now consider slightly weaker \href{https://arxiv.org/abs/1610.02136}{ranking conditions}.
\begin{center}
If \(P(L = 1 \mid x_1) > P(L = 1 \mid x_2)\) then \(c(x_1) > c(x_2)\).
\end{center}
That is, we want to have the confidence values in the right order. Instead of requiring \(c(x)\) to be equal to the actual probability, we only require that the ranking is preserved. If this condition holds, there exists a monotonic calibration function \(g\colon \nR \rightarrow \nR\) such that \(g(c(X)) = P(L = 1 \mid X)\) for input variable \(X\). That is, the ranking condition is almost the same as the calibration condition, up to a monotonic transformation. (We, of course, would have to find this \(g\) as a post-processing step if we wanted truthful predictive uncertainties.) This is more approachable than requiring DNNs to be outputting the true confidence values. And it is, in fact, sufficient for many applications, such as when we filter out too-uncertain examples via a threshold. 

Based on this intuition, people have produced different metrics for quantifying the ranking condition. Essentially, we have two ingredients:
\begin{enumerate}
    \item \textbf{Confidence estimates.} \(c_i := c(x_i) \in \nR\) is the \emph{unnormalized} confidence value for test sample \(x_i\).
    \item \textbf{Correctness of prediction.} \(L_i := \bone(\argmax_k f_k(x_i) = y_i) \in \{0, 1\}\) for test sample \(x_i\).
\end{enumerate}
Instead of trying to estimate the true predictive uncertainty \(p_i\) from \(L_i\) and comparing ranking (we can do this with binning), one may use the raw binary \(L_i\) to benchmark the \(c_i\) estimates. In ECE, we binned the confidence values (restricted to \([0, 1]\)) and took the average of the \(L_i\)s in the bin, which was our estimate of \(p_i\) (very coarse). Now we simply use the raw binary values and benchmark how predictive the confidence estimates are for the \(L_i\) values per sample.

We turn the task into a binary detection task for \(L_i\), where the only feature is \(c_i\). The question is: Can \(c_i\) tell us anything about the prediction correctness?

\subsection{Binary Detection Metrics}

Given features \(c_i\) and target binary labels \(L_i\) as well as a threshold \(t \in \nR\), we predict 1 (``correct'') when \(c_i \ge t\) and 0 when \(c_i < t\). This lets us define the following index sets:
\begin{align*}
\text{True positives: }\mathrm{TP}(t) &= \left\{i: L_i = 1 \land c_i \ge t\right\}\\
\text{False positives: }\mathrm{FP}(t) &= \left\{i: L_i = 0 \land c_i \ge t\right\}\\
\text{False negatives: }\mathrm{FN}(t) &= \left\{i: L_i = 1 \land c_i < t\right\}\\
\text{True negatives: }\mathrm{TN}(t) &= \left\{i: L_i = 0 \land c_i < t\right\}\\
\mathrm{Precision}(t) &= \frac{|\mathrm{TP}(t)|}{|\mathrm{TP}(t)| + |\mathrm{FP}(t)|}\\
\mathrm{Recall}(t) &= \frac{|\mathrm{TP}(t)|}{|\mathrm{TP}(t)| + |\mathrm{FN}(t)|}.
\end{align*}
Informally, precision tells us how pure our positive predictions are at threshold \(t\). Out of the positively predicted samples, how many were correct? Similarly, recall tells us how many of the actual positive samples in the dataset are recalled (predicted positive) at threshold \(t\).

\begin{figure}
    \centering
    \includegraphics[width=0.5\linewidth]{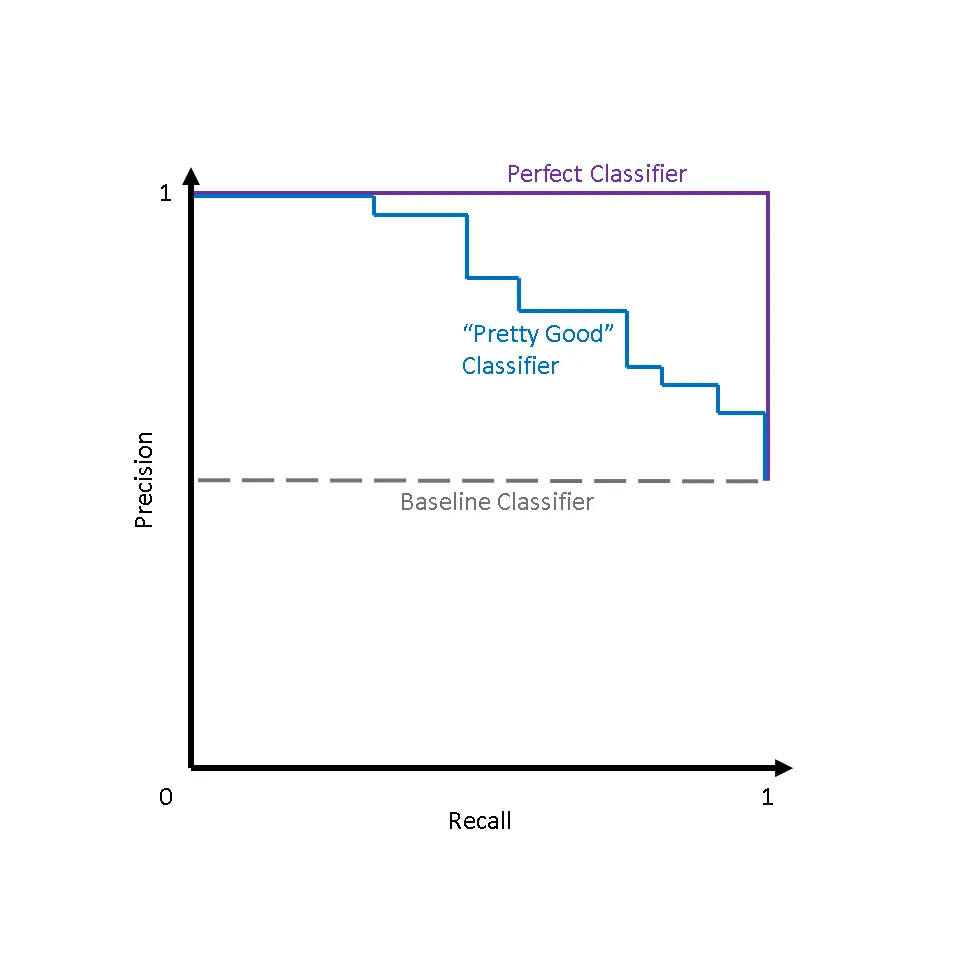}
    \caption{Example precision-recall curve that showcases a random classifier, a perfect one, and one in between. Figure taken from~\cite{steen}.}
    \label{fig:pr}
\end{figure}

One can draw a curve for \(\mathrm{Precision}(t)\) and \(\mathrm{Recall}(t)\) for all possible thresholds $t$ from \(-\infty\) to \(+\infty\) or, for a probability \(c_i\), from \(0\) to \(1\). This is the \emph{precision-recall curve}, shown in Figure~\ref{fig:pr}.

As we go on the recall axis from left to right, we observe the following values for precision and recall. First, we predict all samples as negative. In this case, precision is undefined. Then we recall the sample with the highest \(c_i\) that is actually positive. Recall is almost 0, and precision is 1. We continue\dots, and at the last point, we recall all actual positive samples (i.e., the recall is one). As we predict everything to be positive, the precision is the fraction of true positive samples. This point is always on the line of the random detector.

To summarize this curve, we can compute the area under the precision-recall curve (AUPR). This is a metric for how well we are predicting (how correct our predictions are based on \(c_i\) values). For the perfect detector, \(\mathrm{AUPR} = 1\). While we recall all the actual positive samples, we also never recall actual negative samples. For a random detector, \(\mathrm{AUPR} = P(L = 1)\) where \(P(L = 1)\) is the ratio of positive samples in the dataset. AUPR can be calculated in two ways: AUPR-Success is the method we discussed above. In AUPR-Error, we use errors (\(L = 0\)) as the positive class. Both are often reported together for predictive uncertainty evaluation.

A drawback of the AUPR is that the random classifier's performance depends on \(P(L = 1)\). For example, if \(P(L = 1) = 0.99\) (i.e., the test set is severely imbalanced), then AUPR is already \(99\%\) for a random detector. It lacks the resolution to see the improvement above the random detector baseline.

The Receiver Operating Characteristic (ROC) curve fixes this. It compares the following quantities:
\begin{align*}
\mathrm{TPR}(t) = \mathrm{Recall}(t) &= \frac{|\mathrm{TP}(t)|}{|\mathrm{TP}(t)| + |\mathrm{FN}(t)|} = \frac{|\mathrm{TP}(t)|}{|\mathrm{P}|}\\
\mathrm{FPR}(t) &= \frac{|\mathrm{FP}(t)|}{|\mathrm{FP}(t)| + |\mathrm{TN}(t)|} = \frac{|\mathrm{FP}(t)|}{|\mathrm{N}|}.
\end{align*}
Here, FPR tells us how many of the actual negative samples in the dataset are recalled (predicted positive) at threshold \(t\). This is ``1 - the recall for the negative samples.''

\begin{figure}
    \centering
    \includegraphics[width=0.5\linewidth]{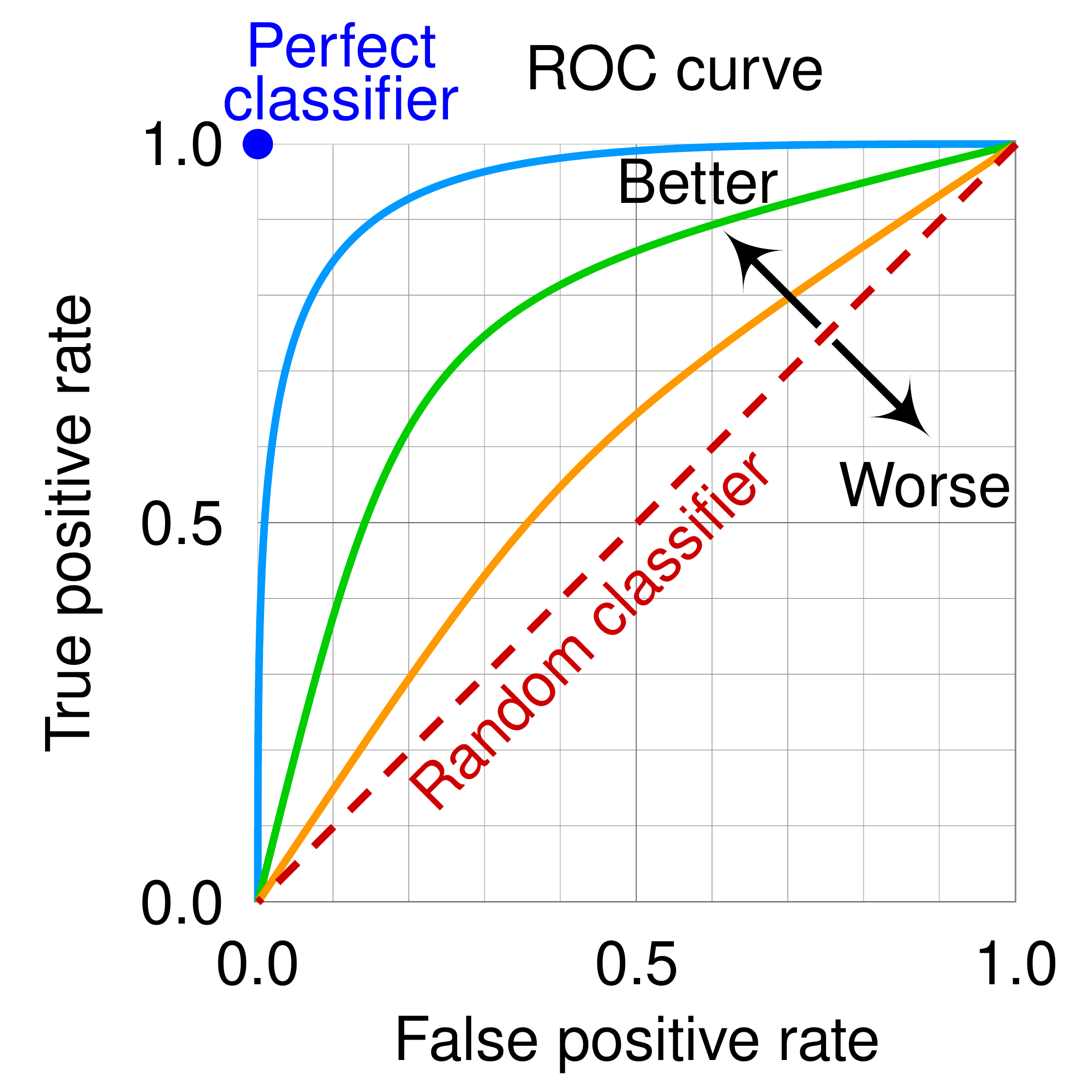}
    \caption{Example ROC curve showing results for a perfect classifier, a random one, and ones in between. Figure taken from~\cite{roc}.}
    \label{fig:roc}
\end{figure}

Similarly to the Precision-Recall curve, one can draw a curve of \(\mathrm{TPR}(t)\) and \(\mathrm{FPR}(t)\) for all \(t\) from \(-\infty\) to \(+\infty\) or, for a probability \(c_i\), from \(0\) to \(1\). This is the \emph{ROC curve}, shown in Figure~\ref{fig:roc}.

As we go on the FPR axis from left to right, the FPR and TPR values change as follows. First, we predict all samples as negative. There, TPR is 0, and FPR is 0. We continue until the last point, where we predict all samples as positive. There, TPR is 1, and FPR is 1. 

The area under the ROC curve (AUROC) can be computed as a summary metric. The AUROC has a nice interpretation: It gives the probability that a correct sample ($L=1$) has a higher certainty $c(x)$ than an incorrect one. This very much captures our ranking goal. For the perfect ordering \(\text{AUROC} = 1\). And, interestingly, for a random order, \(\text{AUROC} = 0.5\), regardless of \(P(L = 1)\). \emph{This makes AUROC the recommended metric over AUPR, especially on unbalanced datasets.}

\section{\(c(x)\) as Non-Predictive Uncertainty}

So far, we have expected \(c(x)\) to be an estimate of the predictive (un)certainty -- whether the model is going to get the answer right or wrong. \(c(x)\), the \emph{confidence estimate}, was required to be a good representation of the likelihood of getting the answer right (\(L = 1\)).

However, we have discussed two more equally important uncertainties: the epistemic and the aleatoric components. We can design benchmarks for each of these sources separately, i.e., measure the quality of a particular \(c(x)\) as the predictor for other factors (i.e., not predictive uncertainty anymore). Here, we impose no restrictions on the estimator \(c(x)\) we might use, only that it returns a probability \(\in [0, 1]\) for a binary prediction task. Possibilities for non-predictive uncertainty benchmarks are listed below.

\textbf{Is the sample \(x\) an OOD sample?} In this case, we can treat \(c(x)\) is an OOD detector. This is not perfectly aligned with predictive uncertainty. Even if a sample is OOD, the model might get the answer confidently right, and even if it is ID, the model can be unconfident. It is rather a measure of epistemic uncertainty.

\textbf{Is the sample \(x\) severely corrupted?} Corruption is related to predictive uncertainty, but they are not perfectly aligned: The level of corruption in an input sample is only one source of uncertainty. This aspect has close ties to aleatoric uncertainty.

\textbf{Does the sample \(x\) admit multiple answers?} This is -- by definition --  aleatoric uncertainty.

As can be seen, these questions are more closely related to identifying particular aspects of uncertainties tied to either epistemic or aleatoric sources.

\subsection{\(c(x)\) as an OOD Detector}

We write \(Y\) for the binary variable indicating
\begin{itemize}
    \item \(Y = 1\) if \(x\) is from outside the training distribution.
    \item \(Y = 0\) if \(x\) is from inside the training distribution.
\end{itemize}
Then, we would expect high \(P(Y = 1)\) for higher uncertainty values \(1 - c(x)\). That is, \(c(x)\) shall be a good estimator for epistemic uncertainty. \(c(x)\) can be, again, treated as a feature for the binary prediction of OOD-ness. We may then evaluate \(c(x)\) for its OOD detection performance with AUPR or AUROC. These are evaluation metrics we already know from predictive uncertainty that are generally used for \emph{ranking} uncertainties. In the literature for OOD detection or general uncertainty estimation, we often see OOD detection performances reported in terms of area under curve metrics.

\subsection{\(c(x)\) as a Multiplicity Detector}

This is a much less popular choice. We write \(Y\) for the binary variable indicating
\begin{itemize}
    \item \(Y = 1\) if the true label for \(x\) has multiple possibilities, maybe because of inherent ambiguity in the task or due to corruption.
    \item \(Y = 0\) if there exists a unique label for \(x\).
\end{itemize}
Then, we would expect high \(P(Y = 1)\) for higher uncertainty values \(1 - c(x)\). That is, \(c(x)\) shall be a good estimator for aleatoric uncertainty (whether the sample accommodates more than one answer). \(c(x)\) can be, again, treated as a feature for the binary prediction of aleatoric uncertainty. We may then evaluate \(c(x)\) for its multiplicity detection performance with AUPR or AUROC.

\subsection{Summary of Evaluation Methods so far for Uncertainty}

For \emph{predictive uncertainty} (whether the model is going to get the prediction right), we have seen (1) proper scoring rules such as log probability and Brier score, (2) metrics based on model calibration such as ECE, MCE, and reliability diagrams (that are more intuitive metrics), and (3) ranking (or ``weak calibration'') using AUROC or AUPR. The third approach uses different thresholds for the retrieval of correctly predicted samples.

If we only care about \emph{epistemic uncertainty}, it makes sense to consider the downstream proxy task of OOD detection to measure the quality of our uncertainty estimates. We can measure OOD detection performance using AUROC or AUPR. Plotting the ROC or precision-recall curves can also be insightful.

For \emph{aleatoric uncertainty}, one might want to look at multiplicity/corruption detection. Detection performance can be, again, measured by AUROC or AUPR.

\section{Estimating Epistemic Uncertainty}

As we have seen, epistemic uncertainty means we are unsure about our prediction because several models could fit the training data (because we have not experienced enough training data to distinguish the correct from the incorrect model.) There are two possibilities:
\begin{enumerate}
    \item The size of our training set is too small, and so the variance of our estimator is too high.
    \item The training data distribution does not cover some meaningful regions in the input space; there are some underexplored areas.
\end{enumerate}

Epistemic uncertainty has a close connection with Bayesian machine learning. A great tool for dealing with multiple possibilities in maths is probability theory:
\[P(\theta \mid \cD) \propto P(\theta)P(\cD \mid \theta) = P(\theta) \prod_{i = 1}^N P(x_i \mid \theta).\]
A posterior distribution over the parameter space is the Bayesian way of saying, ``This space accommodates multiple possible solutions after observing the training set and taking our prior beliefs into consideration.'' A ``wider'' distribution means higher uncertainty regarding the true model. (The one that ``generated'' the dataset.)

It can be instructive to consider the ``input space point of view'': We are adding more and more observations to underexplored regions of the input space. These give more and more supervision: We are narrowing down the possible range of \(\theta\)s based on the observations. This should ideally be happening with Bayesian ML as we observe more data.

\subsection{Space of Model Parameters \(\theta\)}

\begin{figure}
\centering
\includegraphics[width=0.9\linewidth]{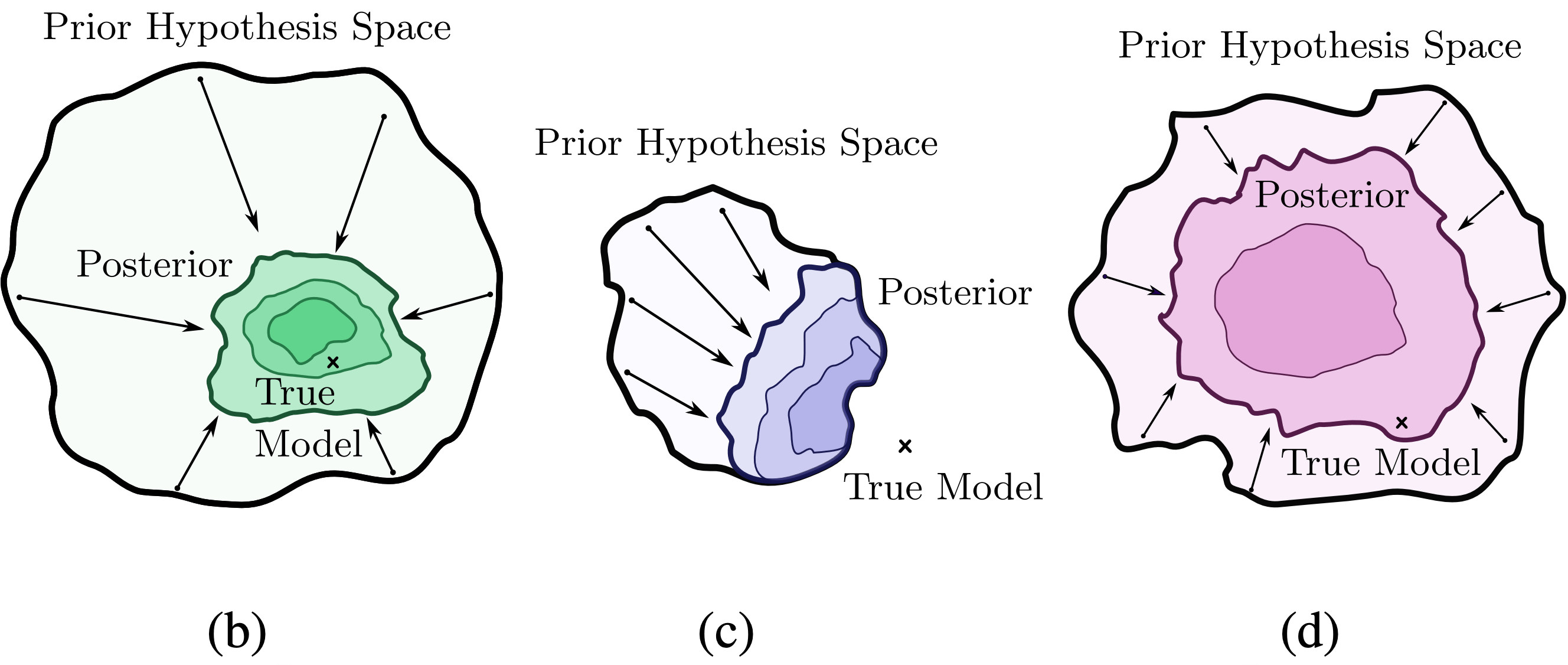}
\caption{Different scenarios for optimization in the hypothesis space. ``(b) By representing a large hypothesis space, a model can contract around a true solution, which in the real world is often very sophisticated. (c) With truncated support, a model will converge to an erroneous solution. (d) Even if the hypothesis space contains the truth, a model will not efficiently contract unless it also has reasonable inductive biases.''~\cite{https://doi.org/10.48550/arxiv.2002.08791} Figure taken from~\cite{https://doi.org/10.48550/arxiv.2002.08791}.}
\label{fig:bayesian}
\end{figure}

This space is at the center of our attention in Bayesian ML. The notion of parameters \(\theta\) is often interchangeably used with weights \(w\) and, sadly, also with functions, models, or hypotheses \(h\). Using Bayesian inference
\[P(\theta \mid \cD) \propto P(\theta)P(\cD \mid \theta) = P(\theta) \prod_{i = 1}^N P(x_i \mid \theta),\]
we are narrowing down our hypothesis space from the wide prior space by observing more and more data until we arrive at the final posterior. We hope this distribution contains the true model (the one that actually ``generated'' the dataset) with high probability. Figure~\ref{fig:bayesian}(b) is the ideal visualization of what should happen with Bayesian ML.

\subsection{Approximate Posterior Distribution Families}

In the previous section, we discussed why a posterior over models is a great way to represent multiple possibilities. In most cases, however, the true posterior (i.e., the one given by Bayes' rule) over our weights/models is intractable. (The prior specification is also often left implicit.) Therefore, we have to make some approximations to our true posterior. We need to define the distributional format of our posterior approximations; in other words, the approximate posterior distribution family. The posterior \(P(\theta \mid \cD)\) can be thought of as an infinite set of models (using sensible priors). We denote our approximation by \(Q_\phi(\theta)\), where \(\phi\) are the parameters of this parametric distribution. The posterior is often approximated without explicitly specifying the prior.

\begin{definition}{Dirac Delta Measure}
The Dirac delta is a generalized function over the real numbers whose value is zero everywhere except at zero. For our purposes, it represents the fact that we only have one possible parameter configuration \(\theta\) in our posterior. Formally, it is a measure. Without going too much into Lebesgue integration theory, the gist is that it acts like a Kronecker delta.

\medskip

\textbf{Note}: This definition only acts as an intuitive description of the Dirac measure. Interested readers should refer to measure theoretical treatments of the notion.
\end{definition}


\(Q_\phi(\theta)\) can be, e.g., \dots
\begin{itemize}
    \item \textbf{\dots a generic multimodal distribution.} For example, it can be a Mixture of Gaussians (MoG), but any other distribution can be chosen. A MoG with an appropriate number of modes is enough to cover any continuous distributions if we allow an arbitrary number of modes.
    \item \textbf{\dots a uni-modal Gaussian distribution.} Many people like to use this for computational simplicity and tractability.
    \item \textbf{\dots a sum of Dirac delta distributions.} Some people use such semi-deterministic \(Q_\phi(\theta)\)s.
    \item \textbf{\dots a single Dirac delta distribution.} This takes us back to deterministic ML. A deterministic posterior approximation means a single point estimate for \(\theta\) (MLE, MAP).
\end{itemize}
Of course, under any sensible prior belief and problem setup, the true posterior \(P(\theta \mid \cD)\) will never be a sum of Dirac deltas. Nevertheless, we might use it as an \emph{approximation} to the true posterior. In this section, we will always \emph{approximate} the true posterior \(P(\theta \mid \cD)\) \wrt either an implicit or explicit prior distribution.

\subsubsection{Deterministic vs. Bayesian ML}

There is a whole spectrum between probabilistic Bayesian ML and deterministic ML. We may also recover the original deterministic ML formulation by choosing our approximate posterior family to be the family of Dirac deltas. Thus, the Bayesian framework is a generalization of deterministic ML. One could express various forms of posterior uncertainty by considering different approximate posterior distribution families.

\emph{Deterministic ML} first optimizes a single model (parameter set) over the training set, \(\theta^*(\cD)\). Then, for a test sample, it predicts the label as
\[P(y \mid x, \cD) = P(y \mid x, \theta^*(\cD)).\]
We use only this single model to produce the output for the input of interest. From the Bayesian perspective, this is equivalent to having a Dirac posterior. As epistemic uncertainty arises from the existence of multiple plausible models, but we only consider a single one in deterministic ML, we cannot represent epistemic uncertainty using deterministic ML (i.e., we treat is as 0).

\emph{Bayesian ML} finds a distribution of models, \(Q_\phi(\theta \mid \cD)\), the approximate posterior over the models after observing the training data. Think of Bayesian ML as training an infinite number of models simultaneously (whenever our approximate posterior does not only accommodate a finite set of models).

\subsubsection{Quantifying Epistemic Uncertainty}

Now, we have the most important ingredient to represent epistemic uncertainty: a set of models. However, measuring the diversity of this set directly is hard. Therefore, people usually look at the averaged prediction of the models, formalized as follows. For a test sample, Bayesian ML predicts the label using Bayesian Model Averaging (BMA)/marginalization:
\[P(y \mid x, \cD) = \int P(y \mid x, \theta) \underbrace{Q_\phi(\theta)}_{\approx P(\theta \mid \cD)}\ d\theta = \nE_{Q_\phi(\theta)}\left[P(y \mid x, \theta)\right].\]
Thus, we take the average prediction from the approximate posterior distribution (the voting from an ``infinite number of models'') at test time.
This can be further approximated as
\[P(y \mid x, \cD) \overset{\mathrm{MC}}{\approx} \frac{1}{M} \sum_{i = 1}^M P(y \mid x, \theta^{(i)}), \qquad \theta^{(i)} \sim Q_\phi(\theta).\]
The entropy \(\nH(P(y \mid x, \cD)\) or the max-prob for classification \(\max_k P(Y = k \mid x, \cD)\) are popular choices to quantify epistemic uncertainty.

\textbf{Intuition of BMA.} We expect the outputs of all models in the posterior to be similar on the training data, as we explicitly train the models on the training set. When we have a test sample in the training data region, we expect \(P(y \mid x, \theta)\) (i.e., the vector of probabilities in classification) to be similar across the models, as the sample will probably lie on the same side of the decision boundaries of the models (which gets tricky to think about in multi-class classification). The models will also be confident in the predictions (up to aleatoric uncertainty), having been trained on similar samples. Therefore, the BMA output \(P(y \mid x, \cD)\) will show high confidence (e.g., it will have max-prob = \(99\%\)). When we have a test sample in an underexplored region, we expect the individual \(P(y \mid x, \theta)\)s to be divergent, as nothing forces the models' decision boundaries to agree in these regions (as we have not trained on samples from these regions).\footnote{Here, we also need the model posterior we obtain to represent a diverse set of plausible models.} Therefore, the models give divergent answers (i.e., the max-prob indices are different). The BMA output will show low confidence: e.g. max-prob = \(59\%\). Averaging/integrating gives us a mixture, and the maximal value of the mixture will be more smoothed out. Even if the individual models are overconfident, the average output will not be. By averaging, the arg max can even become different from all individual arg maxes. For example, in the case of a discrete set of models,
\[\operatorname{avg}\left((0.51, 0.01, 0.48), (0.01, 0.51, 0.48)\right) = (0.26, 0.26, 0.48).\]
To provide further intuition for why the BMA can represent epistemic uncertainty: Models are sure about different things; when we average their outputs, it makes the ensemble more unsure. Thus, we get better epistemic uncertainty prediction. \textbf{Note}: The BMA output still contains the aleatoric uncertainty represented by the individual models -- the BMA represents predictive uncertainty (both epistemic and aleatoric uncertainty) in the most precise sense.


\subsection{Ensembling}

Since ensembling is hugely successful in practice, we will focus a bit more on it in the next sections. Ensemble learning is usually done as follows (popularized by Balaji \etal~\cite{https://doi.org/10.48550/arxiv.1612.01474}).
\begin{enumerate}
    \item Select \(M\) different random seeds. These are different starting points for the optimization in the parameter space.
    \item Train the \(M\) models regularly, using either a bagged dataset for each model (where we sample with replacement from the original training set) or the original training set.
\end{enumerate}
As the loss landscape is highly non-convex, we usually end up with different local minima depending on where we start. Therefore, we usually get a diverse set of models. Random seeds also control the noise on the objective function (loss landscape) itself, not only the starting points on ``the'' landscape. The seeds influence \dots
\begin{itemize}
    \item \textbf{\dots the formation of batches of training samples for SGD.} If we change the seed, we change the batching, as the reshuffling of the dataset is seeded differently.
    \item \textbf{\dots the random components of the data augmentation process.} Therefore, the actual loss landscape is also changed. We almost always perform data augmentation.
    \item \textbf{\dots the random network components, such as Dropout, DropConnect, or Stochastic Depth.} In DropConnect (2013)~\cite{pmlr-v28-wan13}, instead of dropping out activations (neurons), we drop connections between neurons in subsequent layers. Stochastic Depth (2016)~\cite{https://doi.org/10.48550/arxiv.1603.09382} shrinks the network's depth during training, keeping it unchanged during testing. It randomly drops entire ResBlocks during training:
    \[H_l = \mathrm{ReLU}(b_l f_l(H_{l - 1}) + \mathrm{Id}(H_{l - 1}))\]
    where \(b_l\) is a binary random variable.
\end{itemize}
Changing the random seed, therefore, changes many things, which usually encourages enough diversity in our ensemble. Using bagging to obtain separate training sets for each model further encourages diversity.

\subsubsection{BMA with Ensembles}
In \emph{model ensembling}, we train several deterministic models on the same (or subsampled) data simultaneously. In this case, our posterior approximation becomes a mixture of \(M\) Dirac deltas, where \(M\) is the number of models in our ensemble. We claim that this is Bayesian.
\[Q_\phi(\theta) = Q_{\theta^{(1)}, \dots, \theta^{(M)}}(\theta) = \frac{1}{M} \sum_{m = 1}^M \delta(\theta - \theta^{(m)}).\] 

After training the \(M\) models, BMA boils down to taking the average over the ensemble members' predictions
\begin{align*}
P(y \mid x, \cD) &= \int P(y \mid x, \theta) P(\theta \mid \cD)\ d\theta\\
&\approx \int P(y \mid x, \theta) Q_\phi(\theta)\ d\theta\\
&= \int P(y \mid x, \theta) \frac{1}{M} \sum_{m = 1}^M \delta(\theta - \theta^{(m)})\ d\theta\\
&= \frac{1}{M}\sum_{m = 1}^M \int P(y \mid x, \theta) \delta(\theta - \theta^{(m)})\ d\theta\\
&= \frac{1}{M}\sum_{m = 1}^M P(y \mid x, \theta^{(m)}).
\end{align*}
If the reader is not well versed in measure theory, the last equality can be considered a part of the Dirac measure's definition.\footnote{When considering the Dirac measure, one should write \(\int P(y \mid x, \theta) d\delta(\theta - \theta^{(m)})\), which is a rigorous form of Lebesgue integration.} This corresponds to averaging the predictions of individual models. \(y\) can be a scalar value in regression, one particular class in a classification setting, or even the whole class distribution.

Previously, we have discussed an intuitive explanation for why the BMA can represent epistemic uncertainty. On the side, ensembles also often provide better accuracy. The intuition here is that single predictors make different mistakes and overfit differently. This noise cancels out by averaging, and we get a better test accuracy. This phenomenon is formalized and widely used in statistics: Readers might find the various techniques for bootstrap aggregation (or bagging) interesting.

Let us collect the pros and cons of ensembles.
\begin{itemize}
    \item \textbf{Pro}:
    \begin{itemize}
        \item Conceptually simple -- run the training algorithm \(M\) times and average outputs.
        \item Applicable to a wide range of models -- from linear regression to ChatGPT.
        \item Parallelizable -- if we have a lot of computational resources, we can train multiple models simultaneously on different cluster nodes (GPUs).
        \item Performant -- ensembles are not only able to represent epistemic uncertainty but are also often more accurate.
    \end{itemize}
    \item \textbf{Contra}:
    \begin{itemize}
        \item Ensembles do not realize the full potential of Bayesian ML (no infinite number of models, no connectivity between the models).
        \item Space and time complexities scale linearly with \(M\). If we have a limited number of GPUs, we must wait until the previous model finishes training (the same holds even for evaluation). Compute scales linearly even if we parallelize, time might not. To summarize, this does not scale nicely. However, we often share some weights to increase the number of models we include in the ensemble (e.g., to infinity). We will discuss several approaches to training an ``infinite number of models'' below.
    \end{itemize}
\end{itemize}
Finally, we note that ensembling roughly approximates the true posterior that is given \wrt the weight initialization scheme, which is our implicit prior. In other methods, we have no such connections, and the (implicit) prior remains undisclosed.

\subsection{Dropout}
\label{sssec:dropout}

Having a combinatorial number of models during training sounds familiar. We have used dropout for model training for quite some time. When using dropout~\cite{JMLR:v15:srivastava14a}, we sample the dropout masks in every iteration, so a different model is being trained at every iteration. The models are, of course, very correlated. Every time we are training our net with different neurons missing. This is an ensemble of many models. We train each of them for just a couple of steps, but they are so similar that optimizing one model translates over to improving the other models too. On the spectrum of Bayesian methods, dropout is between the sum of Diracs (training a few models) and the variational approach (that trains an infinite number of models). Some people say dropout is Bayesian.

The dropout objective is
\[\frac{1}{N} \sum_{n = 1}^N \log P(y_n \mid x_n, s \odot \theta).\]
This is a simple CE loss over the training dataset, but we turn on/off each weight dimension randomly in each iteration: \(s^{(i)} \sim \mathrm{Bern}(s \mid p)\). This is very similar to the BBB data term. However, we draw \(\theta(s) = s \odot \theta\) from a huge discrete, categorical distribution, not a Gaussian.

What makes this interesting to create ensemble predictions is that we can also use dropout at inference time, as introduced in the paper ``\href{https://arxiv.org/abs/1506.02142}{Dropout as a Bayesian Approximation: Representing Model Uncertainty in Deep Learning}''~\cite{https://doi.org/10.48550/arxiv.1506.02142}. Eventually, any configuration of turning on/off parameters may fit the given training data well (resulting in low NLL loss). This does not have to be the case for non-training data: there will be disagreements between models for OOD samples. The method is good for detecting such OOD samples. Although we have always trained many models simultaneously using dropout, we have not taken advantage of that during inference. To apply dropout at inference time (test time), we do BMA across different Bernoulli mask choices (different weight samples) and average the predictions:
\begin{align*}
P(y \mid x, \cD) &= \int P(y \mid x, \theta)P(\theta \mid \cD)\ d\theta\\
&\approx \int P(y \mid x, \theta)Q_\phi(\theta)\ d\theta\\
&\overset{\mathrm{MC}}{\approx} \frac{1}{K} \sum_{k = 1}^K P(y \mid x, \theta^{(k)})\qquad s^{(k)} \sim \mathrm{Bern}(s \mid p), \theta^{(k)} = s^{(k)} \odot \theta.\\
\end{align*}

\subsection{Evaluation of Ensembling and Dropout in Practice}

Let us discuss a paper on \href{https://arxiv.org/abs/1612.01474}{evaluating ensembling and dropout in practice}~\cite{https://doi.org/10.48550/arxiv.1612.01474}.

\subsubsection{Results of Ensembling}

\begin{figure}
    \centering
    \includegraphics[width=0.5\linewidth]{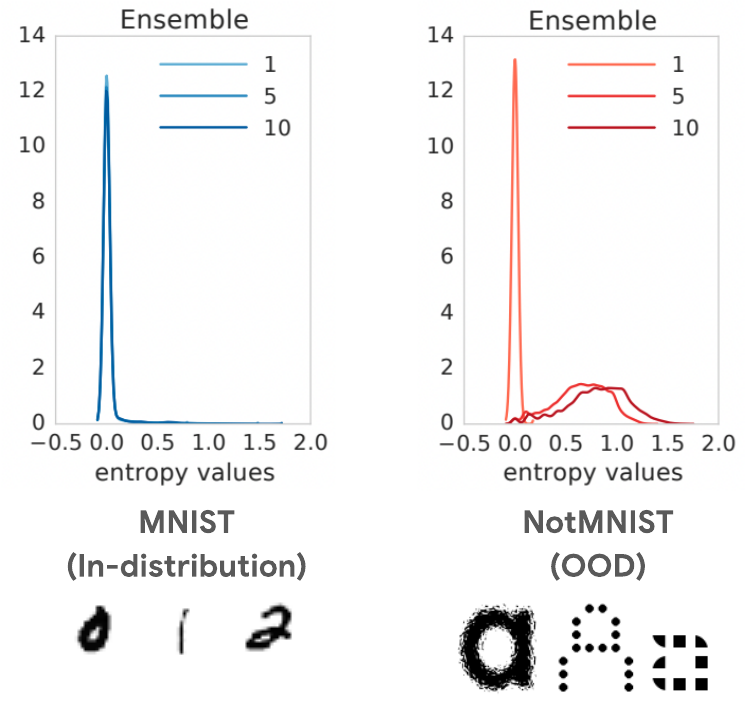}
    \caption{Entropy values on ID and OOD datasets with a varying number of models in the ensemble. Ensembling results in higher entropy values on OOD samples. Base figure taken from~\cite{https://doi.org/10.48550/arxiv.1612.01474}.}
    \label{fig:ensembling3}
\end{figure}

We start with Figure~\ref{fig:ensembling3}. In distribution, the spread of the output categorical distribution is minimal, no matter how many models we have in the ensemble. The prediction is nearly always close to a one-hot vector. The spread here is measured by the entropy of the distribution. Entropy 0 means the categorically distributed random variable is constant, \(p\) is a true one-hot vector. High entropy means \(p\) is close to being uniform. Out of distribution, a single model still produces close to 0 entropy values. However, the ensemble over more and more models has increasing entropy on the OOD samples. 

\begin{table}
    \centering
    \caption{Quantitative results of ensembling on ImageNet. All considered metrics improve with more models. Both the NLL and Brier scores correlate calibration with accuracy. Table taken from~\cite{https://doi.org/10.48550/arxiv.1612.01474}.}
    \label{tab:ensembling4}
    \begin{tabular}{ccccc}
    \toprule
    M & Top-1 error & Top-5 error & NLL & Brier Score \\
    & \% &  \% & & $\times10^{-3}$ \\
    \midrule
    1 & 22.166 &  6.129 & 0.959 & 0.317 \\
    2 & 20.462 &  5.274 & 0.867 & 0.294 \\
    3 & 19.709 &  4.955 & 0.836 & 0.286 \\
    4 & 19.334 &  4.723 & 0.818 & 0.282 \\
    5 & 19.104 &  4.637 & 0.809 & 0.280 \\
    6 & 18.986 &  4.532 & 0.803 & 0.278 \\
    7 & 18.860 &  4.485 & 0.797 & 0.277 \\
    8 & 18.771 &  4.430 & 0.794 & 0.276 \\
    9 & 18.728 &  4.373 & 0.791 & 0.276 \\
    10 & 18.675 &  4.364 & 0.789 & 0.275 \\
    \bottomrule
    \end{tabular}
\end{table}

Finally, let us consider the quantitative results of Table~\ref{tab:ensembling4} from an ImageNet experiment. Ensembling also works at the ImageNet scale. Adding more members to the ensemble decreases error and increases accuracy. (Training on NLL also improves accuracy, not just uncertainty estimates.) Test NLL and Brier scores also improve by increasing the number of models. One could conclude that we obtain better predictive and aleatoric uncertainties. However, it could also be the case that the improvements in these scores are just due to the higher accuracy. Drawing conclusions from proper scoring rule values is, therefore, tricky.

\subsubsection{Comparison of Ensembling and Dropout}

\begin{figure}
    \centering
    
    \includegraphics[width=\linewidth]{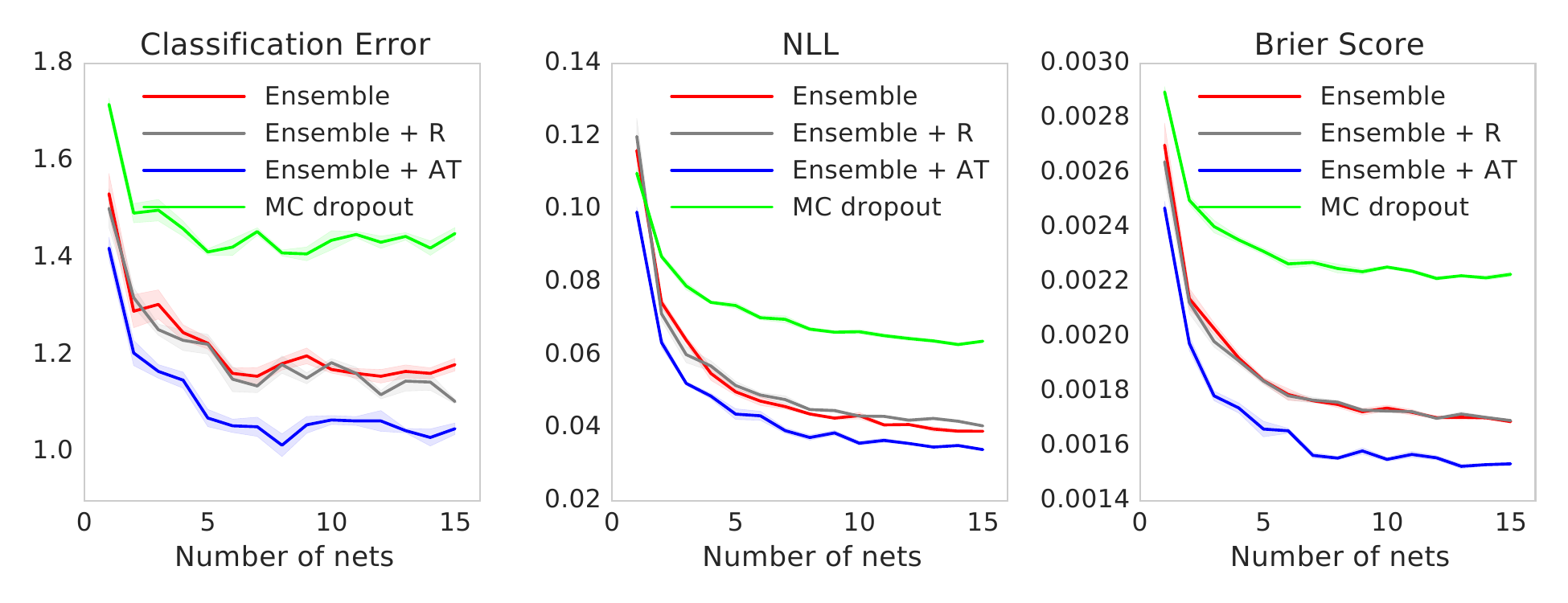}
    \vspace{1em}
    
    \includegraphics[width=\linewidth]{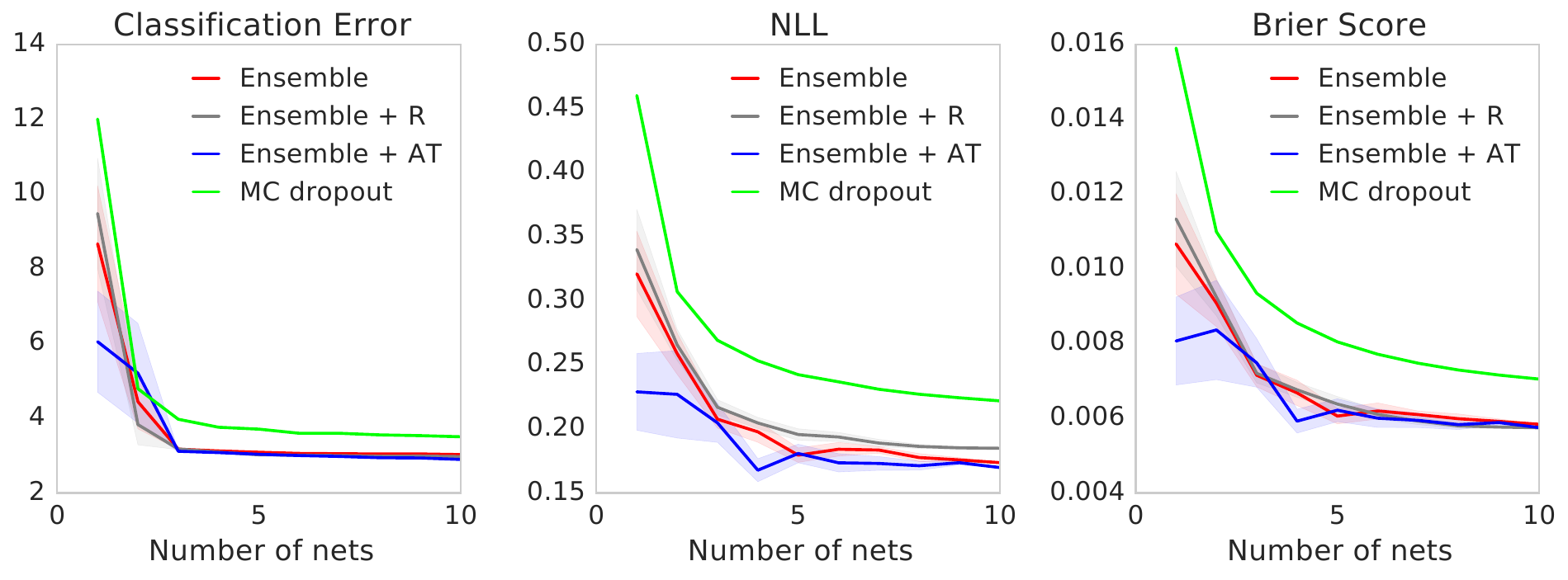}
    
    \caption{\textit{Top.} Evaluation of epistemic uncertainty estimation methods on the MNIST dataset using a 3-layer MLP. \textit{Bottom.} Evaluation on the SVHN dataset using a VGG-style convnet. In both cases, ensembling improves both accuracy and proper scoring metrics. Dropout plateaus earlier and gives suboptimal results. AT: Adversarial training added. R: Random signed vector added (baseline, no difference). Figure taken from~\cite{https://doi.org/10.48550/arxiv.1612.01474}.}
    \label{fig:ensembling}
\end{figure}

Ensembling and dropout seem to be plausible ways to represent epistemic uncertainty. Let us now focus on the top part of Figure~\ref{fig:ensembling}. We take the NLL and the Brier Score of the true label. 

\textbf{Ensembling.} As we add more and more nets to the ensemble, we see a decrease in the classification error (or, equivalently, an increase in accuracy). This is not surprising, as everyone is doing ensembling to get better accuracies. Ensembling with more models also seems to produce better aleatoric and predictive uncertainty estimation. We can conclude this because, for multi-class classification, the log probability scoring rule and the multi-class Brier score are strictly proper scoring rules for both \emph{predictive} and \emph{aleatoric} uncertainty estimation using max-prob. Therefore, by measuring the log-likelihood and the multi-class Brier score, we are also measuring how far away we are from perfect aleatoric uncertainty prediction. \textbf{Note}: By training on NLL, we encourage each model to give correct predictive uncertainties on the training set, and we also ensemble to get correct epistemic uncertainties. The models usually generalize better by ensembling, and we also get better predictive uncertainties on the test samples. Ensembling seems to work for a small dataset and a simple neural network.

\textbf{Dropout.} Sampling more and more nets from dropout seems to plateau quite early and at notably worse values than what we can achieve by ensembling. These days, MC dropout is treated as a method that does not really work. Many people are critical of it.

\textbf{Note}: Aleatoric uncertainty cannot be reduced by ensembling or using dropout: It is completely independent of the model. However, the model posterior might become better at \emph{modeling} the aleatoric uncertainty.


Let us now turn to the bottom part of Figure~\ref{fig:ensembling} that evaluates a VGG-style ConvNet on SVHN (street view house numbers). Ensembling is also scalable to large models and ``large'' datasets. We can use ensembling for any model: We simply have to average the outputs.

\subsection{Training an Infinite Number of Models -- Bayes By Backprop}

Now, we consider a method for training an infinite number of models, called \href{https://arxiv.org/abs/1505.05424}{Bayes By Backprop}.
Training an infinite number of models is possible when the approximate posterior is a continuous distribution (e.g., Gaussian). Expressing infinite possibilities with a finite number of parameters can be easily achieved using parameterized probability distributions. This work explicitly models \(P(\theta)\) and approximates the true posterior \wrt this prior and the training likelihood by
\[P(\theta \mid \cD) \approx \cN\left(\theta \mid \mu^*(\cD), \Sigma^*(\cD)\right) =: Q_\phi(\theta),\]
where * denotes the \(\mu\) and \(\Sigma\) values attained by training on dataset \(\cD\). We simply model the mean and variance, assuming the posterior is approximately Gaussian. Instead of training \(\theta\) directly, we are training \(\mu\) and \(\Sigma\) on \(\cD\). \(\theta\)s are just samples from the Gaussian. One can choose \(P(\theta)\) arbitrarily. However, to keep things closed-form, one usually also chooses a Gaussian.

\begin{figure}
    \centering
    \includegraphics[width=0.4\linewidth]{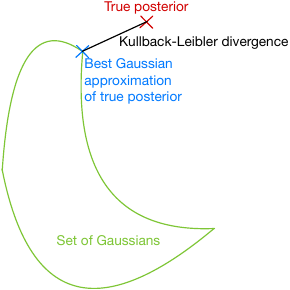}
    \caption{Informal illustration of the Bayes By Backprop optimization problem. The procedure aims to find the best Gaussian approximation of the true posterior. The use divergence between the two is the KL divergence.}
    \label{fig:gaussian}
\end{figure}

Of course, we know that the true posterior \wrt the chosen prior is likely not a Gaussian. It is usually much more complex. Nevertheless, we may still search for the best Gaussian describing the posterior. This is called \emph{variational approximation/inference}. We minimize the ``distance'' between our true posterior and the Gaussian approximation:
\[\min_{\mu, \Sigma} d\left(\cN\left(\mu(\cD), \Sigma(\cD)\right), P(\theta \mid \cD)\right).\]
A popular choice for measuring the divergence (not distance!) between two distributions is the Kullback-Leibler (KL) divergence. With that choice, our problem becomes
\[\min_{\mu, \Sigma} \mathrm{KL}\left(\cN\left(\mu(\cD), \Sigma(\cD)\right)\ \Vert\ P(\theta \mid \cD)\right).\]
Training this directly is impossible, as we do not know the true posterior. However, we can still derive an equivalent optimization problem that does not require us to calculate the exact posterior. Figure~\ref{fig:gaussian} illustrates this optimization problem.

Using the fact that
\begin{align*}
\log \frac{1}{P(\theta \mid \cD)} &= -\log P(\theta \mid \cD)\\
&= -\log \frac{P(\cD \mid \theta)P(\theta)}{P(\cD)}\\
&= -\log P(\cD \mid \theta)P(\theta) + \log P(\cD)\\
&= \log \frac{1}{P(\cD \mid \theta)P(\theta)} + C
\end{align*}
and
\begin{align*}
\log P(\cD \mid \theta) &\overset{\mathrm{IID}}{=} \log \prod_{n = 1}^N P(x_n, y_n \mid \theta)\\
&= \log \prod_{n = 1}^N \left(P(y_n \mid x_n, \theta)P(x_n \mid \theta)\right)\\
&= \sum_{n = 1}^N \left(\log P(y_n \mid x_n, \theta) + \log P(x_n)\right) & (x_n \indep \theta)\\
&= \sum_{n = 1}^N \log P(y_n \mid x_n, \theta) + \underbrace{\sum_{n = 1}^N\log P(x_n)}_{C'},
\end{align*}
we rewrite our training objective as
\begin{align*}
&\mathrm{KL}\left(\cN\left(\mu(\cD), \Sigma(\cD)\right) \ \Vert\  P(\theta \mid \cD)\right)\\
&= \int \cN\left(\theta \mid \mu, \Sigma\right) \log \frac{\cN\left(\theta \mid \mu, \Sigma\right)}{P(\theta \mid \cD)}\ d\theta\\
&= \int \cN\left(\theta \mid \mu, \Sigma\right) \log \frac{\cN\left(\theta \mid \mu, \Sigma\right)}{P(\cD \mid \theta)P(\theta)}\ d\theta + C\\
&= \int \cN\left(\theta \mid \mu, \Sigma\right) \log \frac{\cN\left(\theta \mid \mu, \Sigma\right)}{P(\theta)}\ d\theta - \int \cN\left(\theta \mid \mu, \Sigma\right) \log P(\cD \mid \theta)\ d\theta + C\\
&= \mathrm{KL}\left(\cN(\mu, \Sigma) \ \Vert\  P(\theta)\right) - \underbrace{\nE_{\theta \sim \cN(\mu, \Sigma)} \log P(\cD \mid \theta)}_{\mathrm{CE}}\ +\ C\\
&= \mathrm{KL}\left(\cN(\mu, \Sigma) \ \Vert\  P(\theta)\right) - \sum_n \nE_{\theta \sim \cN(\mu, \Sigma)} \log P(y_n \mid x_n, \theta) + C\\
&\overset{\mathrm{MC}}{\approx} \mathrm{KL}\left(\cN(\mu, \Sigma) \ \Vert\ P(\theta)\right) - \frac{1}{K}\sum_{n = 1}^N \sum_{k = 1}^K  \log P(y_n \mid x_n, \theta^{(k)}) + C
\end{align*}
where \(\theta^{(k)} \sim \cN(\mu, \Sigma)\) and we collapse all terms into \(C\) that do not contain \(\mu\) and \(\Sigma\), the parameters we optimize. In the MC sampling, usually, we take \(K = 1\) for training. This is usually fine because we are MC estimating the expected gradient anyway, with a small batch size (SGD). This expectation approximation can also be made coarse, as noise in SGD was shown to be a regularizer and promote better generalization~\cite{https://doi.org/10.48550/arxiv.2101.12176}.

Our final optimization problem is thus
\[\min_{\mu, \Sigma} \mathrm{KL}\left(\cN(\mu, \Sigma) \ \Vert\ P(\theta)\right) - \frac{1}{K}\sum_n \sum_k  \log P(y_n \mid x_n, \theta^{(k)}) \qquad \theta^{(k)} \sim \cN(\mu, \Sigma).\]
The first term is the prior term, the regularizer. The second term is the data term, the likelihood. We took conceptual, rigorous steps to justify what we are deriving, but this equation makes sense on its own as well.

This is already a convenient loss function, but we want to make it \emph{more DNN-friendly}. We have complete freedom to choose the prior for the KL term. We only need to encode our beliefs through our prior, which can be anything. (This, of course, influences the true posterior but not the true model that generated the data. We want the true model to have high density in the true posterior.) In the parameter space, there are many symmetries; equivalent solutions are spread across the entire space. Regardless of which part of the space we choose, it is very likely that we will find a suitable solution locally. This might serve as a weak justification of the choice of a standard normal distribution as the prior:\footnote{The fact that Gaussianity makes integrals more tractable and the \(L-2\) regularization a Gaussian prior imposes is widely known to work well are more convincing arguments}
\[P(\theta) := \cN\left(\theta \mid 0, I\right).\]
We also restrict our posterior to Gaussians with diagonal covariance matrices:
\[\Sigma = \operatorname{diag}(\sigma^2).\]
(The full covariance matrix with full degrees of freedom would introduce many computational problems.) Thus, we approximate \(P(\theta \mid \cD)\) with a heteroscedastic diagonal Gaussian. Then the KL divergence can be given in closed form, as it is between two multivariate Gaussians:
\[\operatorname{KL}\left(\cN(\mu, \operatorname{diag}(\sigma^2))\ \Vert\ \cN(\theta \mid 0, I)\right) = \frac{1}{2} \sum_i \left[\mu_i^2 + \sigma_i^2 - \log \sigma_i^2 - 1\right].\]
The only remaining problem why the loss is not DNN-friendly is that the loss does not depend on \(\mu\) and \(\Sigma\) straightforwardly. We have to sample from a distribution parameterized by \(\mu, \Sigma\), which is not differentiable in the naive way \wrt \(\mu, \Sigma\). The reparameterization trick is used here to detach \(\mu\) and \(\Sigma\) from the randomness in the approximate posterior. We compute the model parameter via
\[\theta = \mu + \sigma \odot \epsilon,\]
where \(\odot\) means pointwise multiplication and \(\epsilon \sim \cN(0, I)\). We only have to sample \(\epsilon\)s (the random part which does not depend on \(\mu\), \(\Sigma\)) and push it through the above transformation to obtain the \(\theta\) values. This separates the randomness and backpropagation.

Lastly, we need to ensure that the \(\sigma\) vector is always positive. It cannot be just an unbounded parameter, like usual. We counteract this by parameterizing \(\rho\) instead (which is a normal \texttt{nn.Parameter}), which may take on negative values too, and setting
\[\sigma := \operatorname{softplus}(\rho) = \log (1 + \exp(\rho)) > 0\]
where all operations are element-wise.
The actual softplus function also has a hyperparameter \(\beta\) -- we keep everything minimal here. Therefore, we obtain a closed-form, differentiable loss for \(\mu, \sigma\) without any constraints. An example PyTorch code for BBB in a network with a single linear layer is given in Listing~\ref{lst:bbblinear}. We consider multi-class logistic regression in the BBB formulation. It can be trained with backpropagation and SGD.

\begin{booklst}[PyTorch code for a linear BBB classifier. The reparameterization trick and the special loss can be implemented in just a few lines of code.]{lst:bbblinear}
class BBBLinear(nn.Module):
    def __init__(self, input_dim, output_dim):
        super().__init__()
        self.mu = nn.Parameter(
            torch.tensor(input_dim, output_dim).uniform_(-0.1, 0.1)
        )
        self.rho = nn.Parameter(
            torch.tensor(input_dim, output_dim).uniform_(-3, 2)
        )
        # Sizes: number of weights in the model.

    def forward(self, x):
        eps = torch.randn_like(self.mu)  # requires_grad is not propagated, K = 1
        sigma = F.softplus(self.rho)
        theta = self.mu + sigma * eps
        return x @ theta  # logits

    def compute_loss(self, logits, targets):
        # K = 1, negative sum of log-probs
        neg_log_likelihood = F.cross_entropy(
            logits, targets, reduction="sum"
        )
        sigma = F.softplus(self.rho)
        kl_prior = 0.5 * (
            self.mu ** 2 + sigma ** 2 - torch.log(sigma ** 2) - 1
        ).sum()
        return kl_prior + neg_log_likelihood
\end{booklst}

Variational approximation (which justifies what we are doing theoretically) consists of a prior KL term and a likelihood term. For the likelihood, we sample a parameter \(\theta\) from the infinite possibilities of models at every iteration. This is the ``secret sauce'' for training an infinite number of models simultaneously while sharing weights (\(\mu, \Sigma\)) and saving computation. To separate the sampling operation from BP (i.e., to have gradient flow to the parameters of the approximate posterior), we use the reparameterization trick. Frequently, we need to clip parameter values to a certain range -- we use softplus to ensure \(\sigma > 0\).

After training the model with the given formulation, we obtain the optimal parameters \(\mu^*, \Sigma^*\) for our Gaussian approximation. We then compute the BMA based on the learned approximate posterior as
\begin{align*}
P(y \mid x, \cD) &= \int P(y \mid x, \theta)P(\theta \mid \cD)\ d\theta\\
&\approx \int P(y \mid x, \theta)Q_{\mu^*, \Sigma^*}(\theta)\ d\theta\\
&= \int P(y \mid x, \theta)\cN(\theta \mid \mu^*, \Sigma^*)\ d\theta\\
&\overset{\mathrm{MC}}{\approx} \frac{1}{K} \sum_{k = 1}^K P(y \mid x, \theta^{(k)})
\end{align*}
where \(\theta^{(k)} \sim \cN(\mu^*, \Sigma^*)\).

\textbf{Note}: All models with high mass in our posterior make sense. This is a huge statement, as we can sample infinitely many models, meaning we have an entire nice \emph{region} of models in the parameter space. At test time, BBB works the same as ensembles. However, we always draw a new set of \(\theta\)s, and we truly integrate over all \(\theta\)s of our approximate posterior. In contrast, the \(\theta\)s in ensembles are fixed.

\subsubsection{Gaussian posterior approximations are restrictive\dots Why is this better than ensembles?}

Gaussian posterior approximation is a different way to model the posterior than the sum of Diracs. The training procedure and the form of the approximation (meaning of the posterior space) are different.
\begin{itemize}
    \item \textbf{Pro}: We can think about confidence intervals for the approximate posterior (in such a high-D space still; one number for each weight). It is also meaningful to ensure that a certain area around \(\mu\) is always a solution. If we care about getting a region in the parameter space where everything is a solution, then this has more edge. This is a huge volume (a subset of a million-D space) compared to ensembles (that have no volume, as they are just points).
    \item \textbf{Contra}: We have to specify an explicit prior with which the problem remains tractable.
\end{itemize}
In general, this is not a \emph{better} solution than the ensemble method. The ensemble method does not give a variational approximation and basically samples from the true posterior \wrt the weight initialization prior.

\subsubsection{An overview of training an infinite number of models}

One possible recipe for training an infinite number of models is as follows.
\begin{enumerate}
    \item Sample \(\theta^{(k)} \sim \cN(\mu, \Sigma)\) and train that model (\(\mu, \Sigma\)) with the likelihood and the KL term. This is training an infinite number of models represented by \(\cN(\mu, \Sigma)\) at once.
    \item After training, we trust \(\cN(\mu^*, \Sigma^*)\) (the approximate posterior) to represent a good set of plausible models (of infinite cardinality) that generally work well for the training data. If those models disagree on some sample \(x\) (i.e., \(K^{-1}\sum_k P(y \mid x, \theta^{(k)})\) has high uncertainty), then \(x\) is likely to be alien to these models. (The epistemic uncertainty is high, the sample is likely to be OOD and from an unseen region.)
\end{enumerate}

\begin{information}{Regularization Term}
Why is there no \(\lambda\) term in the BBB objective formulation to balance the effect of the two terms? We can do it, it is a more general formulation. Here we did not augment the derived formula with any hyperparameters. However, this effect is already controlled by the prior variance to some extent (not exactly, check the effect in the KL term). This is a nice gain from the probabilistic formulation.
\end{information}

\begin{figure}
    \centering
    \includegraphics{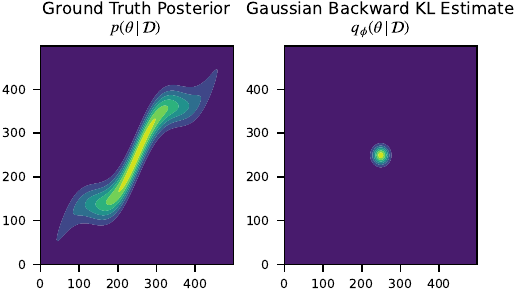}
    \caption{Gaussian posterior approximation using the reverse Kullback-Leibler divergence. The resulting approximate posterior only fits a small high-density region of the true posterior.}
    \label{fig:elongated}
\end{figure}

\begin{information}{Choosing a Diagonal \(\Sigma\) in BBB}
Choosing a diagonal covariance matrix for our Gaussian approximation can be questionable if our true posterior is elongated in some directions. This is illustrated in Figure~\ref{fig:elongated}.

We generally cannot know whether this will happen in advance without extensive investigation in random directions. We force our variational posterior under the true posterior because we use the \emph{reverse KL divergence} as our objective (true posterior is the second argument of KL), which ``squishes'' our approximate posterior into regions of the true posterior with high density. With the \emph{forward KL divergence} (approximate posterior is the second argument of KL), the exact opposite happens: we want to have high density with our approximate posterior wherever the true posterior has high density.
\end{information}

\begin{information}{BBB looks just like VAEs. What is different?}
These are called variational because they use variational inference. They consider a complex posterior (over an unobserved (not a training sample) variable \(\theta\) or \(z\), intermediate latent values) and use a tractable family \(Q\) to approximate it. In VAEs we want to maximize \(P(x \mid \theta)\), and we approximate the intractable \(P(z \mid x, \theta)\) with \(Q_\phi(z \mid x) = \cN(z \mid \eta_\phi(x), \Lambda_\phi(x))\). The general idea is the same: estimate the posterior over the latent variable given the training data. Build a KL distance between the true posterior and the approximate posterior. (Eventually, we get a KL term against the prior plus a data term. We train the parameters of the approximate posterior. In VAEs~\cite{https://doi.org/10.48550/arxiv.1312.6114}, we also train the decoder \(\theta\), not just the encoder \(\phi\).) However, in VAEs, we optimize the ELBO (evidence lower bound); here, we optimize an MC approximation of the true objective. Another difference: In BBB, we are performing variational inference over the parameters we have to train, but in VAEs, we are doing inference over the intermediate outputs \(z\), not the parameters. In VAEs, we use MLE to learn the parameters \(\theta\) and \(\phi\). In BBB, our entire problem is about the variational approximation of the true posterior.

\medskip

Variational approximations happen in many contexts. As an analogy, we can train a DNN for some problem, and it is always the same story\dots Well, it is, but several things are different. 

VAEs are from ICLR 2014. BBB is from ICML 2015. BBB actually cites the VAE paper.
\end{information}

\subsection{Weight Space}

We recommend looking at loss landscape visualizations. Imagining these when discussing Bayesian ML and loss landscapes, in general, makes the topics a lot easier to interpret. It is nice to share these visualizations in our heads. In particular, we refer to two videos, \href{https://www.youtube.com/watch?v=aq3oA6jSGro}{The Loss landscape} and \href{https://www.youtube.com/watch?v=As9rW6wtrYk}{Loss Landscape Explorer 1.1}. We can see a visualization of traveling on the loss landscape to a local minimum. The latter video shows the loss landscape explorer 1.1, which can explore the loss landscape live on real data.

The way these visualizations are created is discussed in the FAQ session of the \href{https://losslandscape.com/faq/}{webpage} of the authors.
\begin{center}
    ``How to deal with so many dimensions? It is very challenging to visualize a very large number of dimensions. If we want to understand the shape of the loss landscapes, somehow we need to reduce the number of dimensions. One of the ways in which we can do that is by using a couple of random directions in space, random vectors that have the same size of our weight vectors. Those two random directions compose a plane. And that plane slices through the multidimensional space to reveal its structure in 2 dimensions. If we then add a 3rd vertical dimension, the loss value at each point in that plane, we can then visualize the structure of the landscape in our familiar 3 dimensions. (Visualizing the Loss Landscape of Neural Nets, Li \etal)''~\cite{losslandscape}
\end{center}
It is easy to make such landscape visualizations look nice by ``cheating'': One can pick random directions until they get something that is visually appealing, then report only these as cherry-picked results. Of course, this is academic malpractice, but its possibility should always be considered, especially when reviewing novel works.

\subsection{Training a Curve of an Infinite Number of Models}

\begin{figure}
    \centering
    \begin{subfigure}{0.32\textwidth}
        \includegraphics[width=\textwidth]{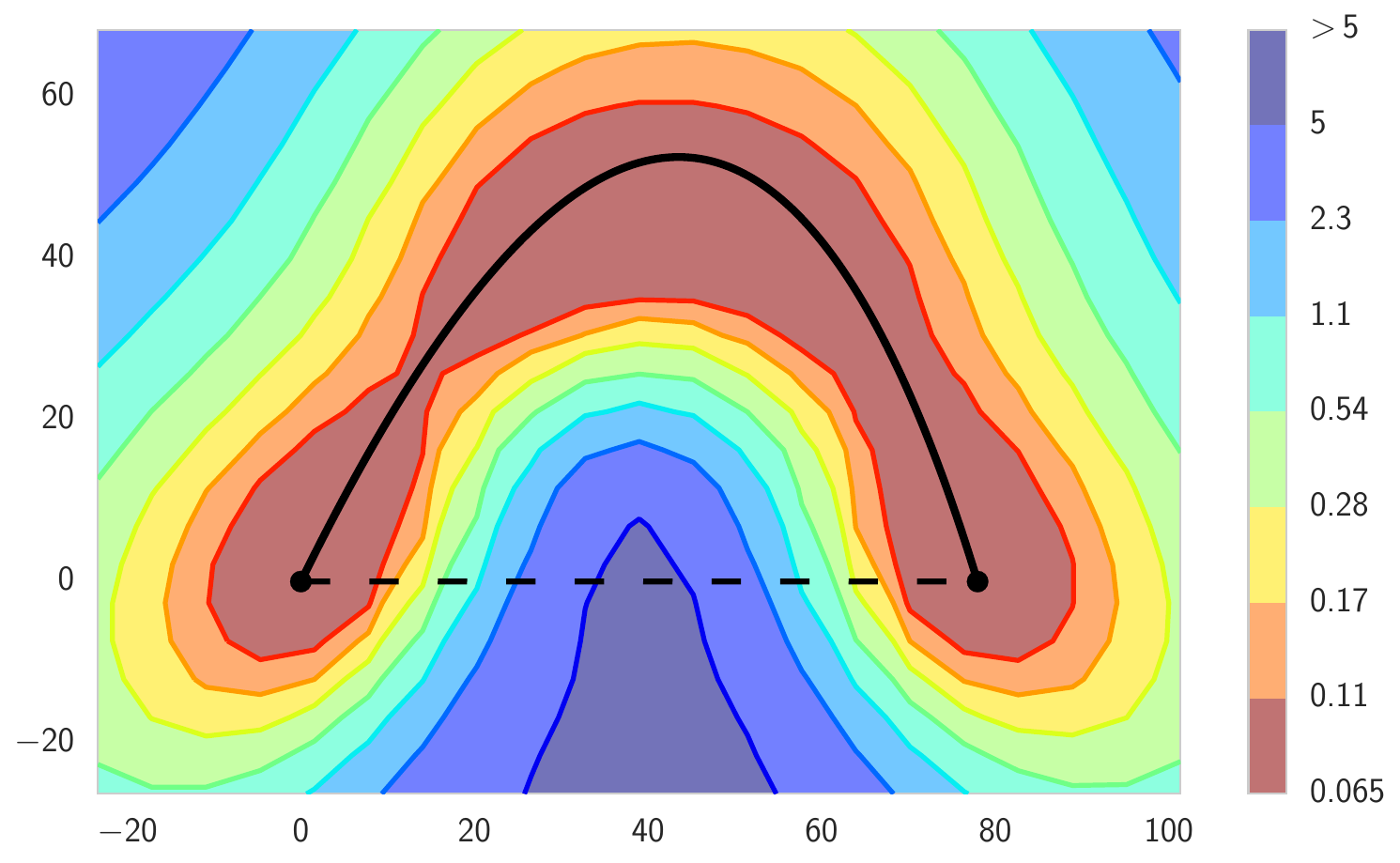}
    \end{subfigure}
    ~
    \begin{subfigure}{0.32\textwidth}
        \includegraphics[width=\textwidth]{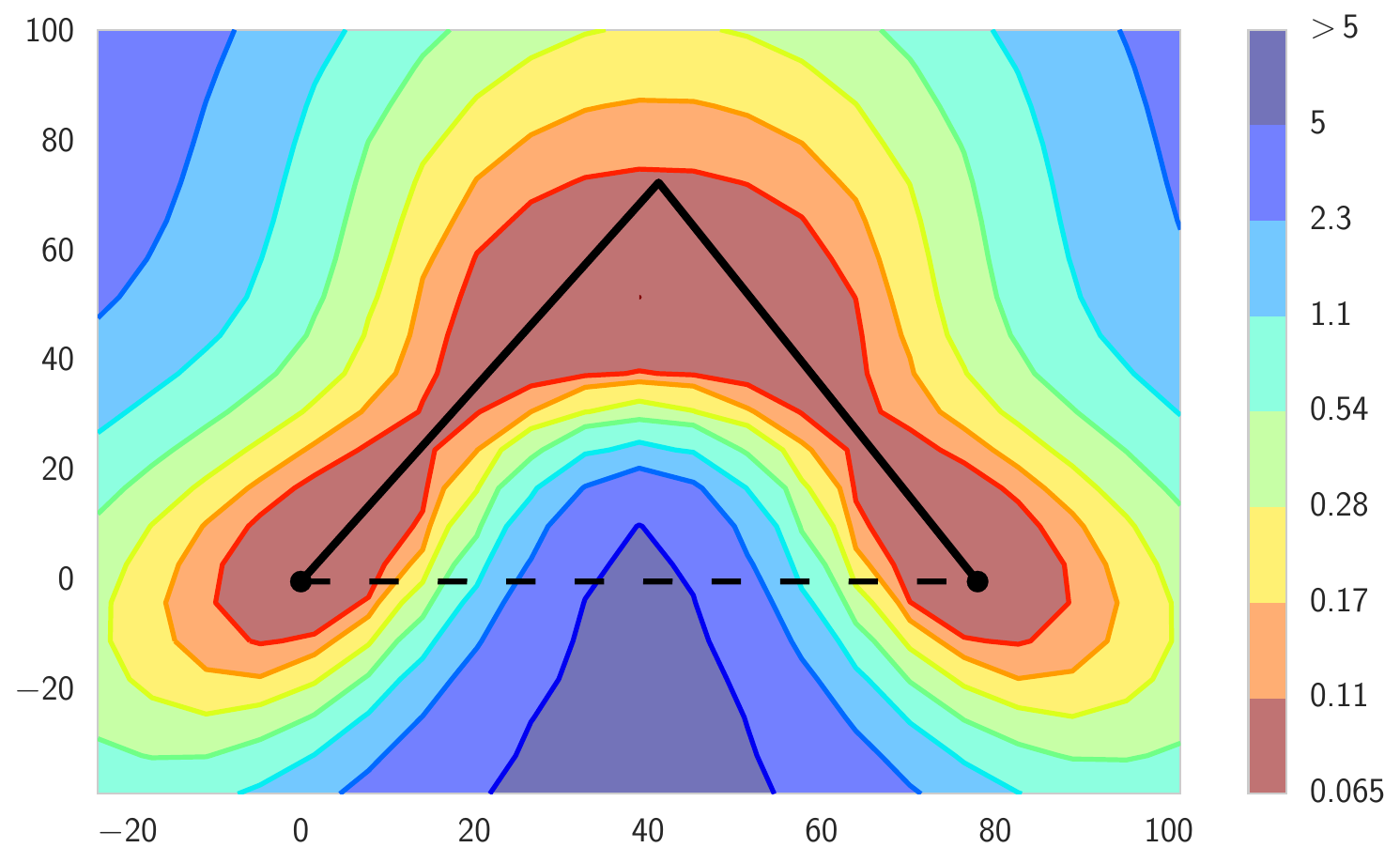}
    \end{subfigure}
    \caption{Training loss surface of a Resnet-164 model on CIFAR-100. \emph{All} models on the obtained curves have low training loss. The training loss is an \(L_2\)-regularized CE loss. Two different parametric curves are shown after optimization. Figure taken from~\cite{https://doi.org/10.48550/arxiv.1802.10026}.}
    \label{fig:curve}
\end{figure}

\begin{figure}
    \centering
    \includegraphics[width=0.7\linewidth]{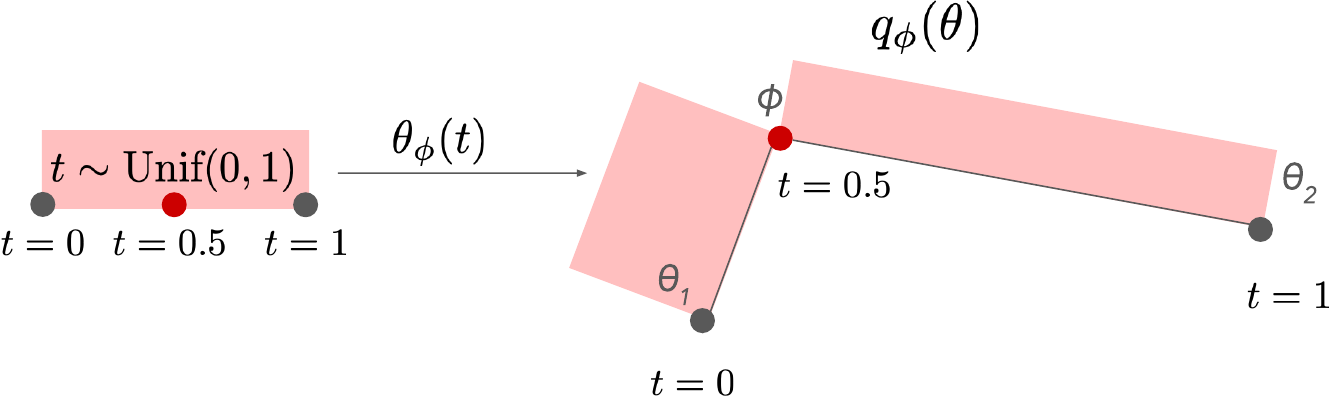}
    \caption{Piecewise uniform distribution over a piecewise linear curve, treated as the approximate posterior \(Q_\phi(\theta)\).}
    \label{fig:lineseg}
\end{figure}

Based on loss landscape visualizations, we can find creative new ways to train an infinite number of models. For example, we can \href{https://arxiv.org/abs/1802.10026}{parameterize a curve}~\cite{https://doi.org/10.48550/arxiv.1802.10026} between two trained models in the parameter space. This is illustrated in Figure~\ref{fig:curve}. Previously, we fit a Gaussian around a point in the parameter space (BBB). We might also think about training more global connections between two, possibly faraway points in the parameter space. We can get an x-y cut of the parameter space, where the training loss values are indicated by colors. Here, x and y indicate two selected axes from the parameter space. They are determined by the third trained point of the curve.

Here, one trains not just a single \(\theta\), but a continuous set of \(\theta\)s along the curve. \footnote{This is also of measure 0, just like the ensemble posterior approximation.} On the plots, we see an infinite number of models that perform well on the training set. Along the curves, the training loss is always very low. Perhaps all points along the curve are good solutions for the training set. This is also Bayesian, as we are training an infinite number of models according to a learned parametric approximate posterior.

We train the curve above as follows. (This is the same story as before.)
\begin{enumerate}
    \item Train two independent models \(\theta_1\) and \(\theta_2\) on different seeds. They are fixed throughout and treated as constants. They are the endpoints of our parametric curve.
    \item We parameterize a curve via a third model \(\phi\). We define the curve via the line segments \(\theta_1 - \phi\) and \(\phi - \theta_2\).
    \[\theta_\phi(t) = \begin{cases} 2(t\phi + (0.5 - t)\theta_1) & \text{if } t \in [0, 0.5) \\ 2((t - 0.5)\theta_2 + (1 - t)\phi) & \text{if } t \in [0.5, 1]\end{cases},\]
    which is a bijection between \(\theta_\phi(t)\) and \(t\) (if \(\theta_1 \ne \theta_2\)).
    \item We model a piecewise uniform distribution over our parametric curve embedded in a high-D space. This is shown in Figure~\ref{fig:lineseg}.\footnote{The two line segments are equally probable. Therefore, the piecewise density values differ when \(\phi\) is not equidistant to the two parameters.}
    \item At each iteration, sample one parameter from the piecewise uniform distribution over the curve at a time. We sample \(t \sim \operatorname{Unif}[0, 1]\); then, \(\theta_\phi(t)\) is a sample of the approximate posterior \(Q_\phi(\theta)\).
    \item \(\phi\) is optimized such that any model on the curve has low training loss. The trained curve is supposed to be a subset of the solution set of the training loss function. We use the reparameterization trick to separate randomness and backprop to the parameters of the distribution that describe the curve.
\end{enumerate}

The optimization problem is as follows (which is typical BNN training).
\[\min_\theta \nE_{\theta \sim Q_\phi(\theta)}\left[-\frac{1}{N}\sum_n \log P(y_n \mid x_n, \theta)\right].\]
The objective function can be rewritten as follows using the reparameterization trick:
\[\nE_{t \sim \operatorname{Unif}[0, 1]} \left[-\frac{1}{N}\sum_n \log P(y_n \mid x_n, \theta_\phi(t))\right]\]
with the curve defined above. The curve is piecewise linear \wrt \(t\) (not differentiable at \(t = 0.5\)) but is entirely linear \wrt \(\phi\) (\(t\) is just a fixed parameter then), so it is differentiable everywhere.

It is also easy to see that the procedure is differentiable after selecting \(t\), as that is the only source of randomness. Sampling \(t\) and obtaining the actual parameter \(\theta_\phi(t)\) are well separated by design. We do not have to use the reparameterization trick, it is ``already used''.

\begin{figure}
    \centering
    \includegraphics[width=0.6\linewidth]{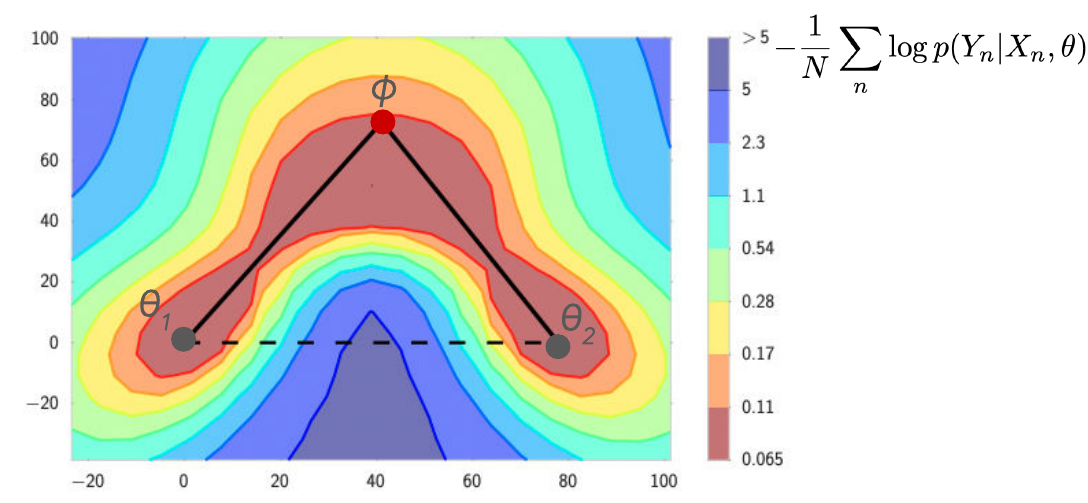}
    \caption{Sampling uniformly from the curve during training ensures that all models on the curve have a low training loss. The 2D training loss surface slice is plotted in which the parameterized curve resides. The authors argue that the parameters in the middle of the curve tend to \emph{generalize} better than the endpoints (i.e., their test loss is lower than those of the endpoints, for being embedded in the middle of a wider basin of the loss surface). Base figure taken from~\cite{https://doi.org/10.48550/arxiv.1802.10026}.}
    \label{fig:lowloss}
\end{figure}

The training ensures that every point \(\theta\) on the surface has low training loss. This is empirically verified in Figure~\ref{fig:lowloss}, where the training loss values are plotted. All losses are below \(\approx 0.11\) on the curve. A general observation is that almost all pairs of independently trained models \((\theta_1, \theta_2)\) for DNNs are connected through a third point \(\phi\) in a low-loss ``highway'' that we can easily find. This gives an interesting intuition for the loss landscape: Most solutions in the DL landscape are connected by some piecewise linear curve. This is not so surprising: We have millions/billions of dimensions to choose from. We can likely find a 2D cut of the loss in which there exists a parametric curve parameterized by \(\phi\) that connects the two endpoints with a low training loss.

The NN has a vast capacity (many dimensions) to accommodate an infinite number of solutions globally rather than around a certain point. Previously, we have shown that it is possible to train an infinite set of models around a specific point locally (Gaussian posterior). Here, we are expanding that idea to global traversal of the parameter space. This was the first work that showed that it is possible.

After training the model with the given formulation, we compute the BMA based on the learned approximate posterior as
\begin{align*}
P(y \mid x, \cD) &= \int P(y \mid x, \theta)P(\theta \mid \cD)\ d\theta\\
&\approx \int P(y \mid x, \theta)Q_{\phi}(\theta)\ d\theta\\
&\overset{\mathrm{MC}}{\approx} \frac{1}{K} \sum_{k = 1}^K P(y \mid x, \theta_\phi(t^{(k)}))
\end{align*}
where \(t \sim \operatorname{Unif}[0, 1]\). 

\begin{figure}
    \centering
    \includegraphics[width=\linewidth]{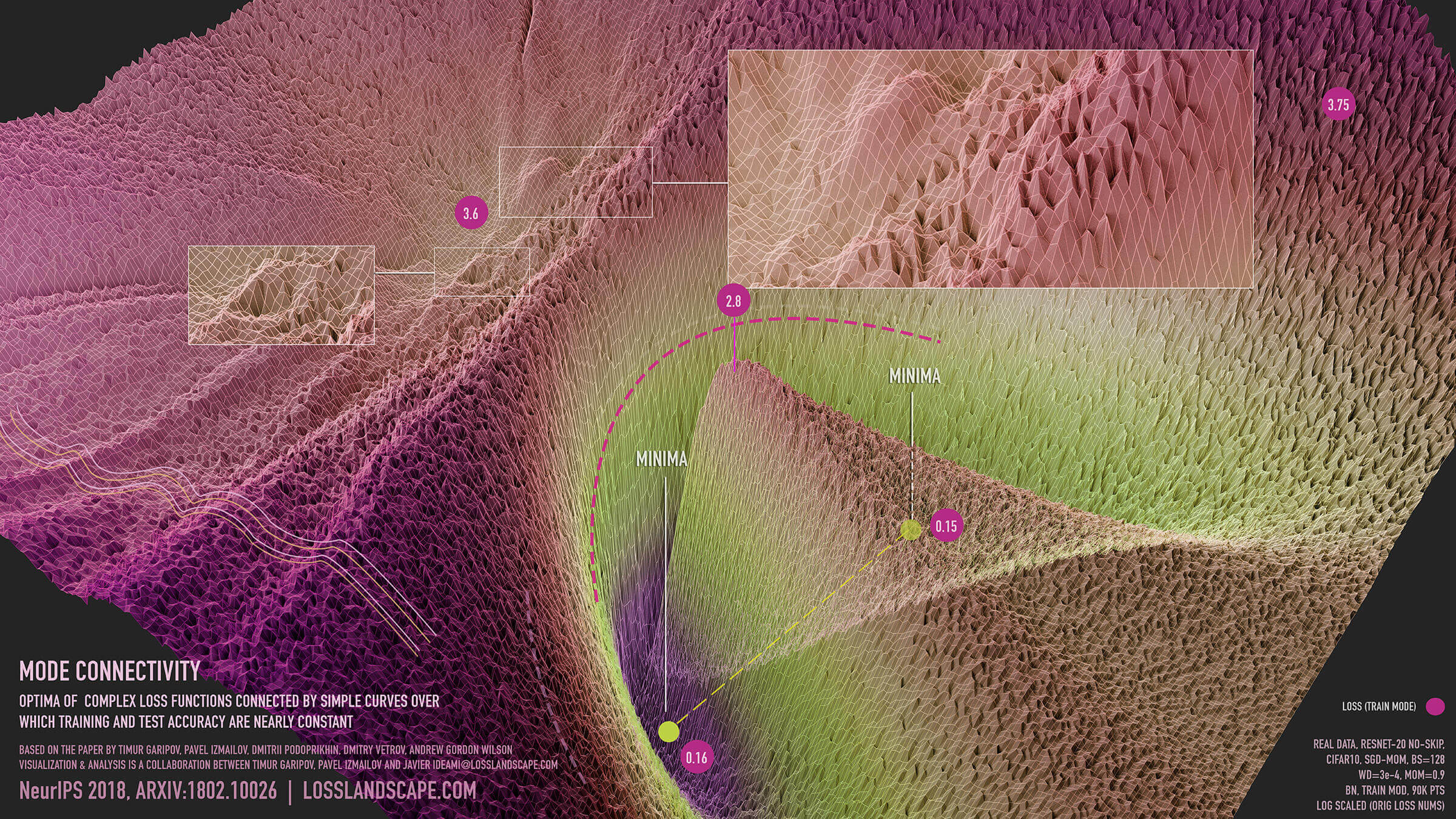}
    \caption{Mode connectivity visualization. The direct path between the two local minima contains high-loss parameter configurations as well. However, we can find a line connecting the two where \emph{all} configurations result in low loss. Figure taken from~\cite{losslandscape}.}
    \label{fig:mode}
\end{figure}
A visualization of mode connectivity is given in Figure~\ref{fig:mode}. We have two solutions that can be connected by some curve in the parameter space. On the curve, \emph{test accuracy is also nearly constant}. This has strong implications for generalization.

\begin{figure}
    \centering
    \includegraphics[width=\linewidth]{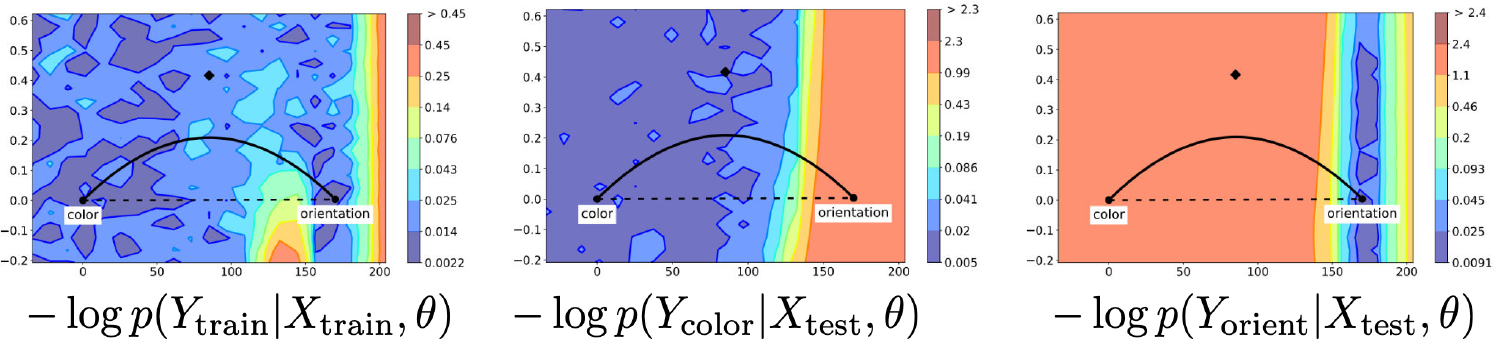}
    \caption{Negative log-likelihood of the diagonal training set and unbiased test set with different labels. \emph{Left.} On the training set, the found curve of models has a low loss. One of the endpoints is a color-biased model, the other is orientation-biased. Therefore, one can obtain a curve of models that interpolates between two models with different biases. \emph{Middle.} When considering a test set with color labels, the color-biased endpoint performs much better, as expected. However, there are many models on the curve that also perform well. There is a relatively quick shift between color-biased and orientation-biased models on the curve. There are also more color-biased models (as the blue area is larger). \emph{Right.} On the same test set with orientation labels, the orientation-biased endpoint performs well, and also a small region of models on the curve (corresponding to the blue region). Base figure taken from~\cite{https://doi.org/10.48550/arxiv.2110.03095}.}
    \label{fig:surprise}
\end{figure}

\begin{information}{Is this method Bayesian?}
This is not a purely Bayesian approach: The true posterior \(P(\theta \mid \cD)\) is approximated by the density \(Q_\phi(\theta)\), which is obtained by maximum likelihood (we maximize the likelihood of the dataset \wrt \(\phi\)). So, we do not consider any prior beliefs over the parameters, and indeed it is probably unlikely that the posterior would be anything close to being a curve if we chose our prior as something like a Gaussian. This work does not care about the prior. It samples some initial models, trains them, fits the curve, and treats it as a posterior approximation. It still has many nice properties and allows interesting insights into the parameter space and the loss surface.

\medskip

However, Bayesian in this book refers to training infinitely many models, not performing Bayesian inference using the prior + likelihood formulation. For a true Bayesian, the prior matters a lot. For the purpose of this book, it does not. We also see ensembling as a Bayesian method. All we are doing is approximating the otherwise intractable true posterior in various ways, sometimes taking an explicit prior into account, sometimes not. This is a common interpretation in the field, and is hard to connect it to any rigorous Bayesian theory.
\end{information}

We are \textbf{not} using a variational approximation of the true posterior: We do not have an explicit prior, and the training objective also does not take any prior into account, as we are performing maximum likelihood estimation over the third parameter of the curve.

\begin{information}{Further Surprise}
We can find \href{https://arxiv.org/abs/2110.03095}{further surprises}~\cite{https://doi.org/10.48550/arxiv.2110.03095} when considering models biased to different cues. This is shown in Figure~\ref{fig:surprise}. Even heterogeneous pairs of models \((\theta_1, \theta_2)\) can be connected with some curve, where heterogeneous means that the two solutions are biased to different attributes. 

\medskip

In the training data, color and orientation labels coincide; we have a diagonal dataset (Section~\ref{ssec:spurious}). We can use either of the cues to get low training loss. Here, \(\theta_\mathrm{color}\) refers to a model biased to color (i.e., the usual solution we get), and \(\theta_\mathrm{orientation}\) corresponds to a model biased to orientation (which is an unusual solution). \textbf{Note}: We have two sets of inputs: \(X_\mathrm{train}\) and \(X_\mathrm{test}\). However, for \(X_\mathrm{test}\), we consider two labeling schemes: one \wrt color and one using orientation as the task cue. Therefore, the loss landscapes are different for these three datasets in total. The parameter x-y cut is shared across these datasets.

\medskip

The axes are chosen as follows. The starting point is two models with different biases. We train a third model using the formulation above. We obtain three points in a million-D space that determine a unique plane (2D subspace) that contains all three models. The other dimensions are hidden in the plots. The negative log-likelihood is plotted for all models (parameter configurations) in this plane.

\medskip

On the left, it is possible to connect these very different solutions with a curve on the training set landscape: The loss for the training data is very low for the entire curve of models, as for the training set, it does not matter which cue our model chooses.

\medskip

In the middle, when we consider the color test set and the \emph{same} curve, we have many models with a low loss, i.e., many models on the curve are biased towards color (as the blue region is rather large). However, an entire region of models that had low training loss suddenly has high test loss (right, orange part of the middle plot): This shows that these models learned spurious correlations \wrt the color labeling scheme.

\medskip

On the right, when we consider the orientation test set (i.e., we only change the labels compared to the middle plot) and the \emph{same} curve, we have a lot fewer models on the curve with a low loss, i.e., models that are biased towards orientation: The blue (low-loss) region is rather small. The yellow region shows a transition from color-biased models to orientation-biased models, and all color-based models have a high loss (red region) on the orientation task.

\medskip

It is nice to see that the space of color-biased solutions is much larger than that of the orientation-biased solutions. This probably explains why if we simply train a model, it is more likely to get a color-biased solution than solutions biased to other cues. This is a volumetric POV for explaining why color is a more favored bias for the models than other cues.

\medskip

\textbf{Another example}: Frogs being the foreground cue and swamps being the background cue. The training samples consist of frogs in swamps. The middle plot would then correspond to pictures of swamps that do not necessarily contain a frog (unbiased dataset). Here, looking at the foreground does not solve the problem. Many more models are biased toward the background than the foreground.
\end{information}

\subsection{Stochastic Weight Averaging}

\begin{figure}
    \centering
    \includegraphics[width=0.4\linewidth]{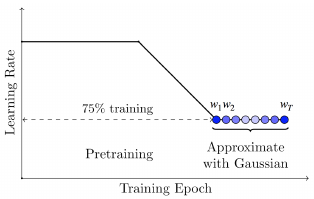}
    \caption{Informal overview of Stochastic Weight Averaging. We give an approximate posterior by considering parameter configurations from the later 25\% of the epochs. Not all of these models have necessarily converged. Figure taken from~\cite{andrewgw}.}
    \label{fig:sgdrand}
\end{figure}

\begin{figure}
	\centering
	\begin{subfigure}[u]{0.4\textwidth}
		\includegraphics[width=\textwidth]{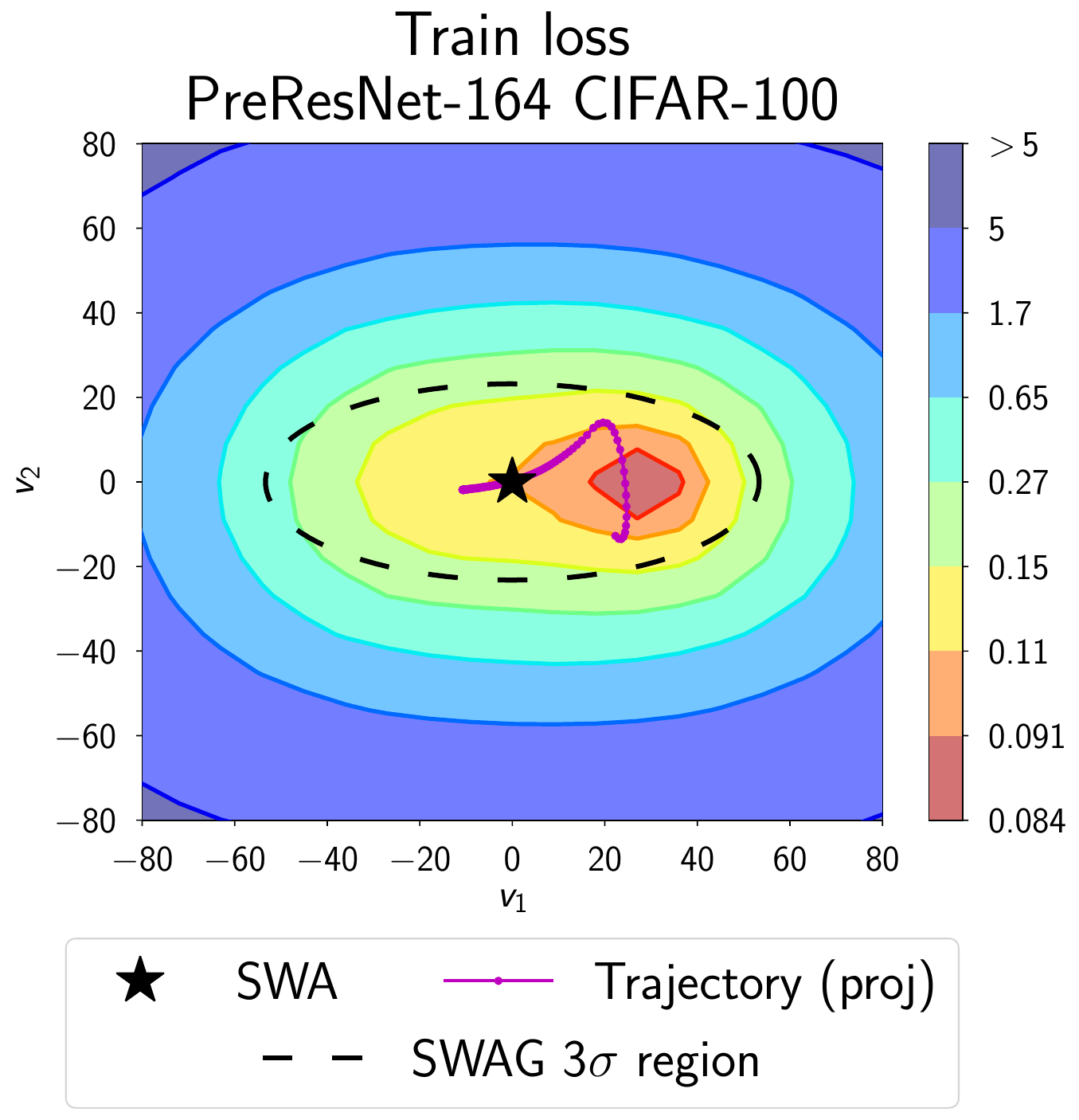}
	\end{subfigure}
	\begin{subfigure}[u]{0.4\textwidth}
		\includegraphics[width=\textwidth]{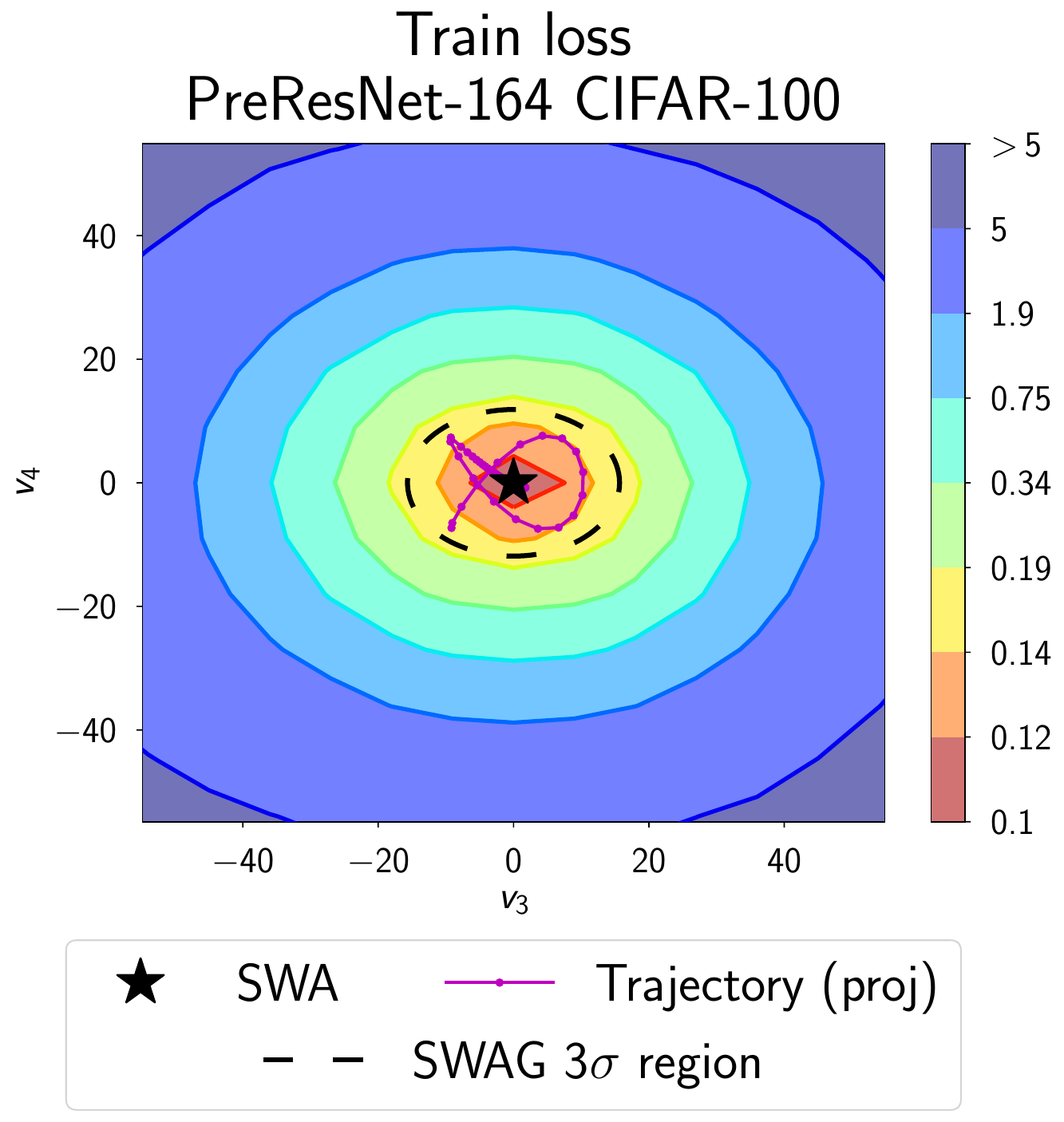}
	\end{subfigure}
	\caption{
		``\textbf{[Left]:} Posterior joint density surface in the plane spanned by eigenvectors of SWAG 
		covariance matrix corresponding to the first and second largest eigenvalues and
		\textbf{Right:} the third and fourth largest eigenvalues. 
		All plots are produced using PreResNet-164 on CIFAR-100. The SWAG distribution projected onto these directions fits the geometry of the posterior density remarkably well.''~\cite{https://doi.org/10.48550/arxiv.1902.02476} Figure taken from~\cite{https://doi.org/10.48550/arxiv.1902.02476}.
	}
	\label{fig:sgdrand2}
\end{figure}

We can exploit the randomness in SGD. This is another cheap source of Bayesian ML. The method is called \href{https://arxiv.org/abs/1902.02476}{Stochastic Weight Averaging}~\cite{https://doi.org/10.48550/arxiv.1902.02476} (SWA). An informal overview is given in Figure~\ref{fig:sgdrand}. In SGD, we usually use an LR schedule. When the LR is sufficiently reduced, solutions are not moving too much around a certain point in space. We treat the set of points (models) towards the end of training (i.e., when the model roughly converged and the models are indeed plausible under the data) as samples from some Gaussian (see Figure~\ref{fig:sgdrand2}). This is the whole idea behind SWA with a Gaussian (SWAG). 

SGD thus inherently trains a large number of models. The SGD trajectory is noisy because of the small batches. The training's final few iterations (epochs) can be treated as samples from the approximate posterior distribution. This is the MCMC point of view of training and sampling, first introduced in ``\href{https://www.stats.ox.ac.uk/~teh/research/compstats/WelTeh2011a.pdf}{Bayesian Learning via Stochastic Gradient Langevin Dynamics}''~\cite{welling2011bayesian}, published way before this paper, in 2011. SGLD is an MCMC method to train and sample a posterior. The intuition is the same as what we discuss here. We treat the current procedure as if we were MCMC sampling the posterior (because of the noise from SGD) that is determined by the loss landscape. (If we do not regularize, we only have an uninformative prior, and the loss landscape is the negative log-likelihood.)

A visualization of SWAG is shown in Figure~\ref{fig:sgdrand2}. Based on the mean and variance of the models of the last couple of epochs, we give a Gaussian approximation. Now, we do not use a variational approximation to get this Gaussian: We do not minimize KL divergences.

The method's assumption is that the posterior is approximately Gaussian:\footnote{The corresponding priors (of multiple experiments) are specified in~\cite{https://doi.org/10.48550/arxiv.1902.02476}, e.g., \(L_2\) regularization.}
\begin{align*}
Q(\theta) &\approx P(\theta \mid \cD)\\
Q(\theta) &= \cN(\theta \mid \mu(\cD), \Sigma(\cD)).
\end{align*}

The Gaussian parameters are computed from the parameters of the last \(L\) epochs (= iterations): \(\theta_1 \dots, \theta_L\).
\begin{align*}
\mu(\cD) &= \frac{1}{L} \sum_l \theta_l\\
\Sigma(\cD) &= \frac{1}{L} \sum_l \theta_l \theta_l^\top - \left(\frac{1}{L} \sum_l \theta_l\right)\left(\frac{1}{L} \sum_l \theta_l\right)^\top.
\end{align*}

\subsubsection{SWAG is not so scalable.}

The problem with the above formulation is the full empirical covariance matrix: For a mid-sized network with a few million parameters, computing and storing this matrix becomes infeasible.

\emph{SWAG-Diag} uses a diagonal approximation of SWAG. The only difference is how the covariance matrix is approximated. As expected from its name, SWAG-Diag uses a diagonal approximation:
\[\Sigma(\cD) = \operatorname{diag}\left(\frac{1}{L}\sum_l \theta_l^2 - \left(\frac{1}{L} \sum_l \theta_l\right)^2\right),\]
where the squaring operations are element-wise.

\begin{figure}
    \centering
    \includegraphics[width=\textwidth]{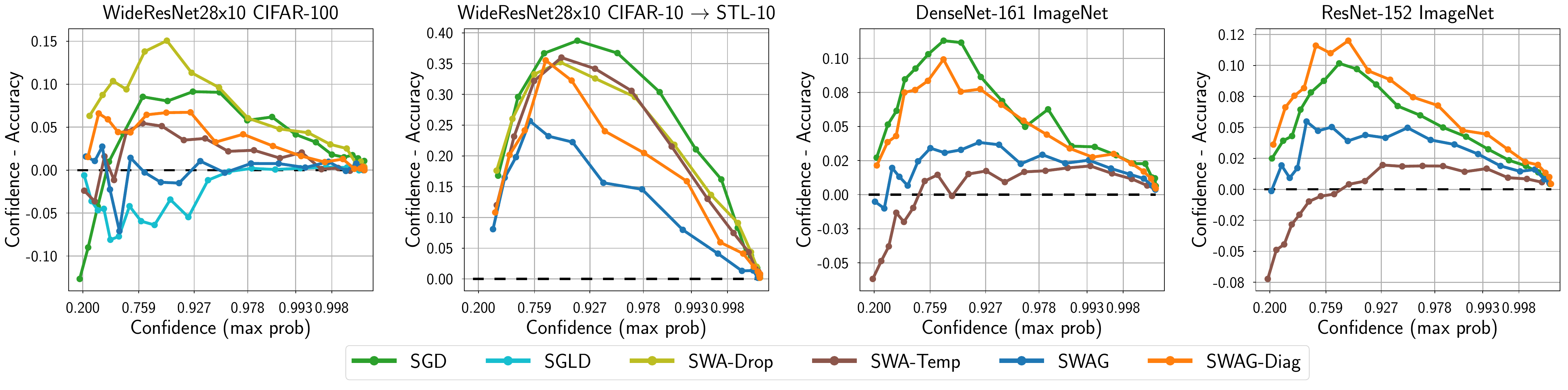}	
    \caption{
    ``Reliability diagrams for WideResNet28x10 on CIFAR-100 and transfer task; 
    ResNet-152 and DenseNet-161 on ImageNet. Confidence is the value of the max softmax output. [...]
    SWAG is able to substantially
    improve calibration over standard training (SGD), as well as SWA. Additionally, SWAG significantly outperforms temperature
    scaling for transfer learning (CIFAR-10 to STL), where the target data are not from the same distribution as the training data.''~\cite{https://doi.org/10.48550/arxiv.1902.02476}. Figure taken from~\cite{https://doi.org/10.48550/arxiv.1902.02476}.}
    \label{fig:cal_curv}
\end{figure}

When using SWAG-Diag, we do not need to train \(M\) different models (like in an ensemble setup), nor do we need to calculate a full covariance approximation (like vanilla SWAG). We only need to train normally and give a Gaussian approximation based on the last few epochs. This is very easy and comes at almost no cost. We can, e.g., do it on ImageNet, or could do it for ChatGPT too. A comparison of SAWG and SWA with other methods is shown in Figure~\ref{fig:cal_curv}. SGD corresponds to the standard training of a single model. Unlike a reliability diagram, the plots directly show the deviation from the line of perfect calibration. Therefore, closer to 0 is better. SWAG-Diag is sadly very similar to SGD regarding the reliability diagram -- SWAG is way better than SWAG-diag, even on ImageNet. It seems that SWAG-diag only scales better computationally, but the results do not follow.

\subsection{On the Principledness of Bayesian Approaches}

Bayesian approaches look principled. They \emph{are} principled, given that lots of assumptions are actually true:
\begin{itemize}
    \item We have a sensible prior that does not make learning infeasible (e.g., the true model (parameter configuration) is outside the support of the prior) or inefficient (e.g., the true model is in the tail, so we need a huge dataset to have high mass at the true model in the approximate posterior).
    \item The posterior follows the assumed distribution (e.g., a Gaussian). Of course, the posterior will seldom be truly Gaussian. This is a huge assumption.\footnote{Bayesians usually claim that at least they are open with their assumptions. Frequentists \emph{also use priors}, but implicitly, which makes them less principled in the Bayesian sense.}
\end{itemize}
In high-dimensional parameter spaces (millions/billions), it is challenging to guarantee those criteria. To ensure that our posterior is concentrated around the true model, we need many samples (which is a foundational problem, not a shortcoming of approximations). To recover the true posterior, we need it to be in the approximate family. Even to \emph{verify} correctness, we would need many samples from the true posterior (an exponentially scaling number in the number of dimensions), especially for complex distributions. This is infeasible for deep learning.

\section{Non-Bayesian Approaches to Epistemic Uncertainty: Measuring Distances in the Feature Space}

We have seen that we can give epistemic uncertainty estimates (by measuring, e.g., the variance of the predictions) when training an infinite (or large) number of models. In principle, however, we do not require training an infinite number of models. Let us remember our basic requirement for epistemic uncertainty: \(c(x)\) is expected to be low when \(x\) is away from seen examples (OOD). Hence, we can also try to estimate epistemic uncertainty by measuring the distance between test sample \(x\) and training samples in the feature space.

\subsection{Mahalanobis Distance}

\begin{figure}
    \centering
    \includegraphics[width=\linewidth]{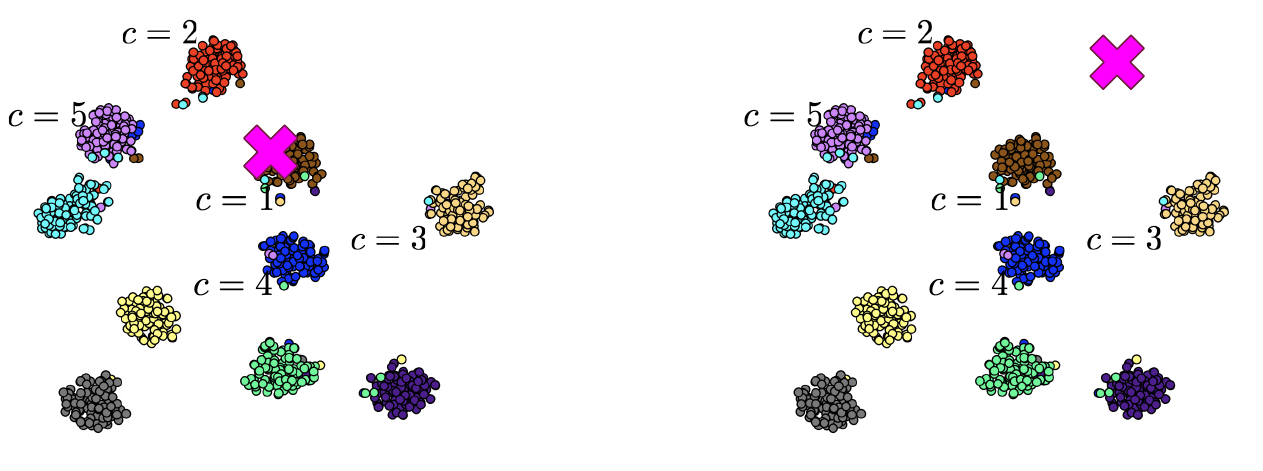}
    \caption{Feature space representation of training samples, where classes are encoded by color \textit{Left cross.} Test features are close to training sample features of class 1 (one of the clusters). This is an in-distribution (ID) test sample. \textit{Right cross.} Test features are not close to training sample features of any class (neither of the clusters). This is an OOD test sample. Base figure taken from~\cite{https://doi.org/10.48550/arxiv.1807.03888}.}
    \label{fig:distance}
\end{figure}

Let us discuss ``\href{https://arxiv.org/abs/1807.03888}{A Simple Unified Framework for Detecting Out-of-Distribution Samples and Adversarial Attacks}''~\cite{https://doi.org/10.48550/arxiv.1807.03888}. We want to measure the closeness of a test sample \(x\) to one of the classes in the feature space for OOD detection. To give a feature representation to each class, we consider the training samples in the feature space (e.g., the penultimate layer of DNNs, with dimensionality \(\approx\) 1000). This is illustrated in Figure~\ref{fig:distance}.

In Figure~\ref{fig:distance}, we computed the distance of our test sample to all training samples. As we do not want to keep all training sample features for future reference, we compute the mean and covariance for each class in the feature space based on all samples in the training set, then approximate each class by a heteroscedastic, non-diagonal Gaussian:
\begin{align*}
\mu_k &= \frac{1}{N_k} \sum_{i: y_i = k} f(x_i)\\
\Sigma_k &= \frac{1}{N_k} \sum_{i: y_i = k} (f(x_i) - \mu_k)(f(x_i) - \mu_k)^\top.
\end{align*}

The authors of~\cite{https://doi.org/10.48550/arxiv.1807.03888} also consider ``tied cov'' to simplify computations by unifying the covariance across classes:
\begin{align*}
\Sigma &= \frac{1}{N} \sum_k N_k \Sigma_k\\
&= \frac{1}{N} \sum_k \sum_{i: y_i = k} (f(x_i) - \mu_k)(f(x_i) - \mu_k)^\top.
\end{align*}
This is \emph{not} the same as calculating the covariance matrix for the entire dataset, as the individual class means are preserved. This is the weighted average of all covariance matrices where the weights are \(N_k / N\). Every class has a different number of samples. We then measure the \textbf{Mahalanobis distance} between a test sample \(x\) and the Gaussian for class \(k\) as
\[M(x, k) = (f(x) - \mu_k)^\top \Sigma^{-1}(f(x) - \mu_k).\]

\begin{information}{Interpretations of the Mahalanobis Distance}
There are two ways to think about the Mahalanobis distance. (1) It is roughly the  NLL of the test sample given class \(k\) (up to a constant). (2) It is the \(L_2\) distance between sample \(x\) and the class mean, weighting every dimension by the precision (inverse covariance) matrix. This is a distorted \(L_2\) distance that weights directions with large precision (small variance) more. Both interpretations are useful to keep in mind.
\end{information}

Then we define the confidence measure \(c(x)\) based on the smallest Mahalanobis distance to a Gaussian:
\[c(x) := - \min_c M(x, c).\]
According to this definition, \(c(x)\) is low when \(x\) is OOD (i.e., \(\min_k M(x, k)\) is high) and analogously \(c(x)\) is high when \(x\) is ID (i.e., \(\min_k M(x, k)\) is low).

\textbf{Note}: This method was designed for detecting OOD samples. It did not consider aleatoric uncertainty, which is also present in the estimates. The field is now aware that it can also influence predictive uncertainty. One could, e.g., measure aleatoric uncertainty as the ratio of distances of the two closest distributions. When it is approximately one, we have high aleatoric uncertainty, because we are split between the classes.

\subsubsection{Results of Mahalanobis Distance}

To discuss the Mahalanobis distance's ability to detect OOD samples, we consider Table~\ref{tab:mah}. This showcases the detection accuracy of OOD samples under various metrics. AUPR in and out correspond to whether the ID or OOD class is considered positive. The max-prob confidence measure is not suitable for OOD detection as much. The performance in detecting OOD samples is considerably better for the Mahalanobis distance in the feature space. This is, of course, just an example. One should not conclude that the Mahalanobis distance is always a better confidence measure than max-prob, just for this particular setup.

\begin{table}
\centering
\caption{Comparison of max-probability and Mahalanobis distance for distinguishing OOD samples. The Mahalanobis distance gives better results on OOD detection. Detection performance is measured using a ResNet trained on CIFAR-10. The OOD dataset is SVHN. For the Mahalanobis distance method, no feature ensembling and input pre-processing is used. Table adapted from~\cite{https://doi.org/10.48550/arxiv.1807.03888}.}
\label{tab:mah}
\begin{tabular}{ccccccccc}
\toprule
\(c(x)\)
& \begin{tabular}[c]{@{}c@{}}TNR \\at TPR 95\% \end{tabular}
& \begin{tabular}[c]{@{}c@{}}AUROC \end{tabular} 
& \begin{tabular}[c]{@{}c@{}}Detection \\accuracy \end{tabular} 
& \begin{tabular}[c]{@{}c@{}}AUPR \\in \end{tabular} 
& \begin{tabular}[c]{@{}c@{}}AUPR \\out \end{tabular} \\\midrule
Max-probability & 32.47 & 89.88 & 85.06 & 85.40 & 93.96 \\ \midrule
Mahalanobis & 54.51 & 93.92 & 89.13 & 91.56 & 95.95 \\ \bottomrule
\end{tabular}
\end{table}

We also consider a more fine-grained evaluation of the Mahalanobis distance's ability to detect OOD samples in Table~\ref{tab:mah2}. Here, various distribution changes are considered where the confidence measure correctly has to tell ID and OOD apart. The baseline is, again, max-prob. The superiority of the Mahalanobis distance in this OOD detection setup is clear. \textbf{Note}: Generally, one needs a good set of in-dist datasets and a diverse set of OOD datasets to measure OOD detection performance.

\begin{table}
\centering
\caption{Accuracies expressed in percentages on distinguishing OOD data under various distribution changes. The Mahalanobis distance consistently outperforms the max-prob confidence measure. The detection accuracies are reported for the OOD samples. Table adapted from~\cite{https://doi.org/10.48550/arxiv.1807.03888}.}
\label{tab:mah2}
\resizebox{0.7\columnwidth}{!}{\begin{tabular}{ccccc}
\toprule
\multirow{3}{*}{\begin{tabular}{c} In-dist \\ (model) \end{tabular}}  & \multirow{3}{*}{OOD} & \multicolumn{3}{c}{Validation on OOD samples}\\
 &  & TNR at TPR 95\%        & AUROC        & Detection acc. \\
 \cline{3-5} & & \multicolumn{3}{c}{Max-prob / Mahalanobis} \\ \midrule
\multirow{3}{*}{\begin{tabular}[c]{@{}c@{}} CIFAR-10 \\(DenseNet) \end{tabular}}     
&  SVHN         
& \multicolumn{1}{c}{40.2 / {\bf 90.8}}
& \multicolumn{1}{c}{89.9 / {\bf 98.1}}
& \multicolumn{1}{c}{83.2 / {\bf 93.9}}\\
&  TinyImageNet 
& \multicolumn{1}{c}{58.9 / {\bf 95.0}}
& \multicolumn{1}{c}{94.1 / {\bf 98.8}}
& \multicolumn{1}{c}{88.5 / {\bf 95.0}}\\
&  LSUN        
& \multicolumn{1}{c}{66.6 / {\bf 97.2}}
& \multicolumn{1}{c}{95.4 / {\bf 99.3}}
& \multicolumn{1}{c}{90.3 / {\bf 96.3}}\\\midrule
\multirow{3}{*}{\begin{tabular}[c]{@{}c@{}} CIFAR-100 \\(DenseNet) \end{tabular}}   
&  SVHN         
& \multicolumn{1}{c}{26.7 / {\bf 82.5}}
& \multicolumn{1}{c}{82.7 / {\bf 97.2}}
& \multicolumn{1}{c}{75.6 / {\bf 91.5}}                \\
&  TinyImageNet 
& \multicolumn{1}{c}{17.6 / {\bf 86.6}}
& \multicolumn{1}{c}{71.7 / {\bf 97.4}}
& \multicolumn{1}{c}{65.7 / {\bf 92.2}}\\
&  LSUN         
& \multicolumn{1}{c}{16.7 / {\bf 91.4}}
& \multicolumn{1}{c}{70.8 / {\bf 98.0}}
& \multicolumn{1}{c}{64.9 / {\bf 93.9}}\\  \midrule
\multirow{3}{*}{\begin{tabular}[c]{@{}c@{}} SVHN \\(DenseNet) \end{tabular}} 
&  CIFAR-10     
& \multicolumn{1}{c}{69.3 / {\bf 96.8}}
& \multicolumn{1}{c}{91.9 / {\bf 98.9}}
& \multicolumn{1}{c}{86.6 / {\bf 95.9}}\\
&  TinyImageNet 
& \multicolumn{1}{c}{79.8 / {\bf 99.9}}
& \multicolumn{1}{c}{94.8 / {\bf 99.9}}
& \multicolumn{1}{c}{90.2 / {\bf 98.9}}             \\
&  LSUN         
& \multicolumn{1}{c}{77.1 / {\bf 100}}
& \multicolumn{1}{c}{94.1 / {\bf 99.9}}
& \multicolumn{1}{c}{89.1 / {\bf 99.3}} \\  \midrule
\multirow{3}{*}{\begin{tabular}[c]{@{}c@{}} CIFAR-10 \\(ResNet) \end{tabular}}     
&  SVHN         
& \multicolumn{1}{c}{32.5 / {\bf 96.4}}
& \multicolumn{1}{c}{89.9 / {\bf 99.1}}
& \multicolumn{1}{c}{85.1 / {\bf 95.8}} \\
&  TinyImageNet 
& \multicolumn{1}{c}{44.7 / {\bf 97.1}}
& \multicolumn{1}{c}{91.0 / {\bf 99.5}}
& \multicolumn{1}{c}{85.1 / {\bf 96.3}} \\
&  LSUN         
& \multicolumn{1}{c}{45.4 / {\bf 98.9}}
& \multicolumn{1}{c}{91.0 / {\bf 99.7}}
& \multicolumn{1}{c}{85.3 / {\bf 97.7}}
\\ \midrule
\multirow{3}{*}{\begin{tabular}[c]{@{}c@{}} CIFAR-100 \\(ResNet) \end{tabular}}     
&  SVHN         
& \multicolumn{1}{c}{20.3 / {\bf 91.9}}
& \multicolumn{1}{c}{79.5 / {\bf 98.4}}
& \multicolumn{1}{c}{73.2 / {\bf 93.7}}\\
&  TinyImageNet 
& \multicolumn{1}{c}{20.4 / {\bf 90.9}}
& \multicolumn{1}{c}{77.2 / {\bf 98.2}}
& \multicolumn{1}{c}{70.8 / {\bf 93.3}} \\
&  LSUN         
& \multicolumn{1}{c}{18.8 / {\bf 90.9}}
& \multicolumn{1}{c}{75.8 / {\bf 98.2}}
& \multicolumn{1}{c}{69.9 / {\bf 93.5}}\\ \midrule
\multirow{3}{*}{\begin{tabular}[c]{@{}c@{}} SVHN \\(ResNet) \end{tabular}}  
&  CIFAR-10     
& \multicolumn{1}{c}{78.3 / {\bf 98.4}}
& \multicolumn{1}{c}{92.9 / {\bf 99.3}}
& \multicolumn{1}{c}{90.0 / {\bf 96.9}} \\
&  TinyImageNet 
& \multicolumn{1}{c}{79.0 / {\bf 99.9}}
& \multicolumn{1}{c}{93.5 / {\bf 99.9}}
& \multicolumn{1}{c}{90.4 / {\bf 99.1}}\\
&  LSUN         
& \multicolumn{1}{c}{74.3 / {\bf 99.9}}
& \multicolumn{1}{c}{91.6 / {\bf 99.9}}
& \multicolumn{1}{c}{89.0 / {\bf 99.5}}\\ \bottomrule
\end{tabular}}
\end{table}

\subsection{Other types of distances than Mahalanobis: RBF kernel}

We discuss the work ``\href{https://arxiv.org/abs/2003.02037}{Uncertainty Estimation Using a Single Deep Deterministic Neural Network}''~\cite{https://doi.org/10.48550/arxiv.2003.02037} to highlight another distance-based uncertainty estimator.

Here, instead of computing the Mahalanobis distance \wrt the class Gaussians, we first compute the \(L_2\) distance between test sample and centroid of class \(k\), \(\mu_k = \frac{1}{N_k} \sum_{i: y_i = k} f(x_i)\):
\begin{align*}
d(x, k) = \Vert f(x) - \mu_k \Vert_2^2.
\end{align*}
Then, we compute the RBF kernel value for class \(k\) as
\[K_k(f(x), \mu_k) = \exp\left(-\frac{d(x, k)}{2\sigma^2}\right)\]
where \(\sigma\) is a hyperparameter. This is a special case of the Mahalanobis distance where the covariance is isotropic (hence the name ``radial'' basis function), and we take the squared \(L_2\) norm.

The kernel value has a nice property:
\[K_k(f(x), \mu_k) \in (0, 1]\]
where higher values indicate greater similarity. This is more interpretable than the Mahalanobis distance, and it also has a nice interpretation as a probability. All \(\sigma\) does is to control the temperature of this distribution.
Finally, we define our confidence level as
\[c(x) := \max_k K_k(f(x), \mu_k).\]
Low confidence indicates an OOD sample: \(c(x)\) can be interpreted as the probability of \(x\) \emph{not} being OOD. Conveniently, \(c(x) \in (0, 1]\), thus, we can apply a proper scoring rule and train using the resulting criterion. In the derivation below, we exclude the case of \(K_k(f(x), \mu_k) = 1\) and also simplify notation to just \(K_k\). The negative log probability scoring rule for the max-RBF similarity is given by
\begin{align}\label{eq:loss}
\cL = \begin{cases} - \log \max_k K_k & \text{if } Y_{\argmax_k K_k} = 1 \\ -\log(1 - \max_k K_k) & \text{if } Y_{\argmax_k K_k} = 0\end{cases}
\end{align}
where \(Y\) is a one-hot (random) vector for the GT class. As \(Y\) is a one-hot vector, \(Y_{\argmax_k K_k} = 1\) means that the prediction is \emph{correct}, whereas \(Y_{\argmax_k K_k} = 0\) shows an \emph{incorrect} prediction.

When the prediction is correct, we gain \(\log c(x)\) reward (or lose \(-\log c(x)\) reward). If we were very confident, we would gain the most. This encourages the network to have a high \(c(x)\), i.e., make the feature representation of \(X\) even closer to the current centroid. We are optimizing correct predictive uncertainty estimation by \(c(x)\). When the prediction is incorrect, we gain \(\log (1 - c(x))\) reward. We repel the current centroid.

We upper bound Equation~\ref{eq:loss} with a familiar loss function, BCE. When \(Y_{\argmax_c K_c} = 1\) (upper branch), we write
\[-\log \max_k K_k = -\sum_k Y_k\log K_k\]
because \(Y\) is a one-hot vector. We also have
\[-\log (1 - \max_k K_k) = -\log \min_k (1 - K_k) = \max_k - \log(1 - K_k)\]
where we used for the last equality that \(\log\) is monotonically increasing. When \(Y_{\argmax_k K_k} = 0\) (lower branch), this can be bounded from above as
\begin{align*}
\max_k \underbrace{-\log(\underbrace{1 - K_k}_{\in (0, 1)})}_{\in (0, +\infty)} &\le \sum_{k: Y_k = 0} -\log(1 - K_k)\\
&= \sum_k -(1 - Y_k)\log(1 - K_k).
\end{align*}
Thus, we finally have that
\begin{align*}
\cL &\le \begin{cases} \overbrace{-\sum_k Y_k \log K_k}^{> 0} & \text{if } Y_{\argmax_k K_k} = 1 \\ \underbrace{-\sum_k (1 - Y_k)\log(1 - K_k)}_{> 0} & \text{if } Y_{\argmax_k K_k} = 0\end{cases}\\
&\le -\sum_k \left(Y_k \log K_k + (1 - Y_k) \log(1 - K_k)\right).
\end{align*}
The authors of~\cite{https://doi.org/10.48550/arxiv.2003.02037} optimize this upper-bound proxy loss on a finite (deterministic) dataset \(\{(x_i, y_i)\}_{i = 1}^N\). The loss is the sum of BCEs of one-vs-rest classifications where \(K_k\) is our predicted probability of membership of class \(k\) for sample \(x\). This advocates the use of this form of BCE for optimizing our classifiers. This encourages correct predictive uncertainty (\(L = 1\)) reports for \(c(x) := \max_k K_k(f(x), \mu_k)\).

A remaining problem is that the class centroids \(\mu_k = \frac{1}{N_k}\sum_{i: y_i = k}f(x_i)\) are continuously updated during training. These are needed for all \(K_k\) and for all \(x\). Suppose we recompute centroids every time we update our parameters. In that case, we will have a very noisy training procedure, as the targets (centroid means) are constantly moving, and we are trying to chase after them for the right class for each sample. Also, recalculating these for the entire dataset after every network update is infeasible. To solve both, we use a moving average for more stable centroid estimation at each iteration and more stable training:
\begin{align*}
N_k &\gets \gamma N_k + (1 - \gamma)n_k\\
m_k &\gets \gamma m_k + (1 - \gamma) \sum_{i \in \mathrm{minibatch}: y_i = k}f(x_i)\\
\mu_k &\gets \frac{m_k}{N_k}.
\end{align*}
where
\begin{itemize}
    \item \(N_k\) is the ``soft'' number of samples per class in mini-batch: It is the moving average of the number of samples per class in mini-batch. This changes over iterations; we also need to smooth this out.
    \item \(m_k\) is the moving average of the sum of class \(k\) sample features.
    \item \(\mu_k\) is the average feature location (centroid) for class \(k\).
    \item \(n_k\) is the number of samples per class in the current mini-batch.
    \item \(\gamma \in [0.99, 0.999]\) corresponds to the momentum term in the moving average. To make learning stable, it is chosen to be quite high.
\end{itemize}

\textbf{Note}: When a training sample has high aleatoric uncertainty, it will be positioned between likely centroids at the end of training. When a training sample has low aleatoric uncertainty, it will be very closely clustered to the correct class. When an OOD sample comes, it will have low confidence. However, we can also get low confidence for samples with high aleatoric uncertainty. We cannot distinguish these two cases based on the confidence value. This work just attributes low confidence to epistemic uncertainty.

\subsubsection{Results of RBF Kernel}

\begin{figure}
    \centering
    \includegraphics[width=0.5\linewidth]{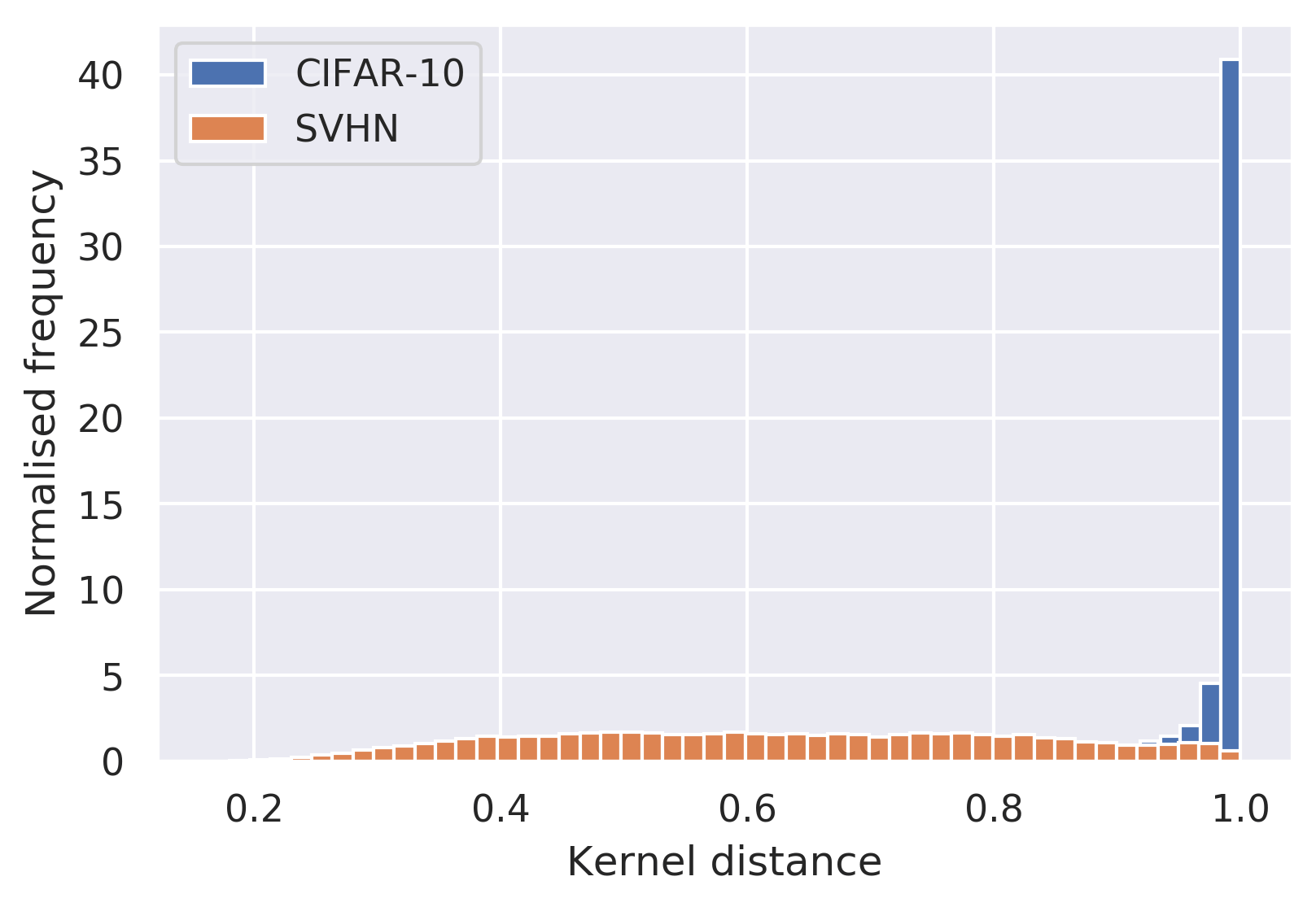}
    \caption{Results of RBF kernel confidence estimation. In distribution, the kernel similarities are all high, i.e., the mapped samples are concentrated in high-density regions of the feature space. Out of distribution, the similarities are mixed, meaning there are a variety of samples ``the model is not familiar with.'' ID dataset is CIFAR-10, OOD dataset is SVHN. Figure taken from~\cite{https://doi.org/10.48550/arxiv.2003.02037}.}
    \label{fig:rbf}
\end{figure}

We first discuss Figure~\ref{fig:rbf} that showcases qualitative results. The confidence estimate successfully distinguishes the two sources of data (ID, OOD). In distribution, the maximal kernel similarity\footnote{The x-axis label reads ``kernel distance'' but is actually the kernel similarity. Distance is low when similarity is high.} is very high, and samples are well clustered in the feature space. Out of distribution, samples tend to have different maximal kernel similarities than one. We qualitatively conclude that \(c(x) = \max_c K_c(f(x), \mu_c)\) is a good indicator of how to separate OOD samples from ID samples after training the network. One can find the best separating threshold (or just report AUROC or AUPR).

We also discuss quantitative results shown in Table~\ref{tab:quant}. Results are quantified using the AUROC score. (``Can we separate ID from OOD based on \(c(x)\) predictions?'') DUQ corresponds to deterministic uncertainty quantification using the confidence score \(c(x) = \max_k K_k(f(x), \mu_k)\). The name highlights that they do not have to stochastically train multiple (or even an infinite number of) models to obtain epistemic uncertainty estimates. LL ratio is a method we do not discuss. `Single model' denotes DUQ trained with softmax-cross-entropy. It uses the same \(c(x)\) formulation but is trained with the usual softmax-cross-entropy loss. As shown in the Table, the method gives good results after training with the proposed objective (DUQ).

\begin{table}
    \centering
    \caption{AUROC results on FashionMNIST, with MNIST being the OOD set. DUQ (using \(c(x) = \max_k K_k(f(x), \mu_k)\)) outperforms most methods.``Deep Ensembles is by
    Lakshminarayanan et al. (2017), Mahalanobis Distance by
    Lee et al. (2018), LL ratio by Ren et al. (2019). Results marked by
    (ll) are obtained from Ren et al. (2019), (ours) is implemented using
    our architecture. Single model is our architecture, but trained with
    softmax/cross entropy.''~\cite{https://doi.org/10.48550/arxiv.2003.02037} Table taken from~\cite{https://doi.org/10.48550/arxiv.2003.02037}.}
    \label{tab:quant}
    \begin{tabular}{lc}
    \toprule
    Method                      & AUROC \\ \hline
    DUQ                         & 0.955 \\
    LL ratio (generative model) & 0.994 \\
    Single model                & 0.843 \\
    5 - Deep Ensembles (ours)   & 0.861 \\
    5 - Deep Ensembles (ll)     & 0.839 \\
    Mahalanobis Distance (ll)   & 0.942 \\
    \bottomrule
    \end{tabular}
\end{table}

\subsection{Summary of Modeling Epistemic Uncertainty}

We have seen two general ways of modeling epistemic uncertainty. In Bayesian ML, we train a set of models simultaneously. We measure their disagreement during inference through Bayesian model averaging (BMA). We can also choose to measure distances in the feature space. In particular, we can compute the distance to the closest class centroid in the feature space to get a sense of how surprising an input sample is. Both have been successfully applied to the problem of OOD detection, which is a proxy task for epistemic uncertainty.

\section{Modeling Aleatoric Uncertainty}

As we have seen, aleatoric uncertainty refers to ``I do not know because there are multiple plausible answers.'' This happens when true label \(y\) is not a deterministic function of input \(x\), as multiple possibilities could be an answer for input \(x\).

Below, we give a high-level overview of the ingredients we will use to represent aleatoric uncertainty.

\subsection{Roadmap to Representing Aleatoric Uncertainty}
\label{ssec:roadmap}

There are \emph{two ingredients} that are used together for the recipe of representing aleatoric uncertainty.

\textbf{Architecture.} Formulate a model architecture that accommodates multiple possible outputs. We should prepare, e.g., a probabilistic output where our model outputs the parameters of this output distribution rather than a single prediction.

\textbf{Loss function.} Taking a proper scoring rule for matching the predicted output distribution to the one dictated by the dataset (examples are discussed in Sections~\ref{ssec:au_classification} and~\ref{ssec:au_regression}) is often sufficient.

Let us follow our recipe and extend proper scoring rules to more generic distributions, as this will allow us to recover truthful aleatoric uncertainty estimates. We start with matching output distributions in classification.

\subsection{Aleatoric Uncertainty In Classification}
\label{ssec:au_classification}

As discussed previously, aleatoric uncertainty refers to the inherent variability of the labels, i.e., the non-deterministic nature of the data generating process. To \emph{represent} aleatoric uncertainty, we would like our model to output a \emph{distribution} which is faithful to \(P(Y \mid X = x)\). 

\subsubsection{Proper Scoring Rules to the Rescue, Again}

So far, our discussion centered around binary distributions, where we tried to match a confidence value \(c(x)\) to the true probability of an event, such as \(P(L = 1)\), where \(L\) represents the correctness of prediction. Here, both \(c(x)\) and \(P(L = 1)\) corresponded to the parameters of respective Bernoulli distributions. By matching the Bernoulli parameters, we were also matching the Bernoulli distributions. To achieve this, we leveraged (strictly) proper scoring rules.

We now extend the notion of proper scoring rules to general discrete distributions. 
In particular, we want to match a distribution \(Q\) (a categorical distribution encoded by a vector of probabilities) to the true discrete distribution \(P\). Let \(y\) be a sample of distribution \(P(Y)\) -- for example, the GT class index. We then define the scoring rule as a function \(S(Q, y)\). Arguments are \(Q\) (the predicted distribution) and \(y\) (a sample from true distribution \(P(Y)\)). This scoring rule is \emph{strictly proper} when the expected score \(\nE_{P(Y)} S(Q, Y)\) is maximized iff \(Q \equiv P\) (i.e., when the distributions match). \(S\) may also be described as a function of \(Q\)'s parameters (e.g., the parameter vector of a categorical distribution or the parameter of a Bernoulli distribution) rather than \(Q\) itself.

If we want, we can then further compress \(Q\) into a scalar. The aleatoric confidence can be, e.g., given by the max-prob \(\max_k Q(Y = k)\), or by the entropy of the predicted distribution, \(\nH(Q)\).

Let us discuss some popular proper scoring rules for matching predicted categorical distributions, encoded by softmax outputs \(f(x)\), to the GT distributions with GT probabilities \(P(Y = y \mid X = x)\ \forall y \in \cY\) from which we can only sample.

\subsubsection{Log Probability Scoring Rule}

The log probability scoring rule (negative CE) for categorical distributions is defined as
\[S(f, y) = \sum_k y_k \log f_k(x) = \log f_y(x),\]
where \(y\) is the true class.\footnote{\(y\) is often used to denote both a one-hot vector of a class and the class label. This is just an abuse of notation.} It can be shown that \(S\) defined this way is a strictly proper scoring rule, i.e.
\[\nE_{P(Y)}S(f, Y)\]
is maximal iff
\[f_k(x) = P(Y = k \mid x)\ \forall k \in \{1, \dotsc, C\}.\]
This is great news! Many DNNs already minimize a NLL loss of the form 
\[\cL = -\sum_k y_k \log f_k(x) = -\log f_y(x),\]
which means they are already matching their predictions to the aleatoric uncertainty of a data source. Since it is a proper scoring rule, in the expectation of \(Y\), we encourage our DNN to predict \(f(x)\) that correctly represents the spread of \(P(Y \mid X = x)\) in the training set.

\subsubsection{Multi-Class Brier Scoring Rule}

To match a probability vector encoding a categorical distribution to the true distribution \(P(Y \mid X = x)\), we can also use the Brier scoring rule. Consider a predicted probability vector \(f(x) \in [0, 1]^K\) with \(\sum_{k=1}^K f_k(x) = 1\) and a categorical random variable \(Y \in \{1, \dots, K\}\). The multi-class Brier scoring rule is defined as
\[S(f(x), y) = -(1 - f_y(x))^2 + f_y(x)^2 - \sum_{k=1}^K f_k(x)^2.\]

\begin{claim}
The multi-class Brier score is a strictly proper scoring rule for aleatoric uncertainty.
\end{claim}

\begin{proof}
First, we rewrite \(\nE_{P(Y \mid X = x)} S(f, Y)\) as
\begin{align*}
\nE_{P(Y \mid X = x)} S(f, Y) &= \sum_{k=1}^K P(Y = k)\left[-(1 - f_k(x))^2 + f_k(x)^2 - \sum_{l=1}^K f_l(x)^2\right]\\
&= \sum_{k=1}^K P(Y = k) \left[-f_k(x)^2 + 2f_k - 1 + f_k(x)^2 - \sum_{l=1}^K f_l(x)^2\right]\\
&= -\sum_{k=1}^K \left[P(Y = k)(1 - 2f_k(x)) + \sum_{l=1}^K P(Y = k)f_l(x)^2\right]\\
&= -\sum_{k=1}^K P(Y = k)(1 - 2 f_k(x)) - \sum_{l=1}^K f_l(x)^2\\
&= -\sum_{k=1}^K \left[P(Y = k)(1 - 2f_k(x)) + f_k(x)^2\right]
\end{align*}
for all \(f \in \Delta^K\) which is the (\(K - 1\))-dimensional probability simplex.

A necessary condition for the maximizer of the Brier scoring rule in expectation is as follows.

\(\forall r \in \{1, \dots, K\}, f \in \Delta^K\):
\begin{align*}
\frac{\partial}{\partial f_r}\left(-\sum_{k=1}^K \left[P(Y = k)(1 - f_k(x)) + f_k(x)^2\right]\right) &= -\sum_{k=1}^K \frac{\partial}{\partial f_r}\left[P(Y = k)(1 - 2f_k(x)) + f_k(x)^2\right]\\
&= -\frac{\partial}{\partial f_r}\left[P(Y = r)(1 - 2f_r(x)) + f_r(x)^2\right]\\
&= -(-2P(Y = r) + 2f_r(x)) \overset{!}{=} 0\\
&\iff f_r(x) = P(Y = r).
\end{align*}
As \(\frac{\partial}{\partial f_r} \left(-[-2P(Y = r) + 2f_r(x)]\right) = -\left(0 + 2 \right) = -2 < 0\ \forall r \in \{1, \dots, K\}, f \in \Delta^K\), the above, \(f(x) \equiv P(Y \mid X = x)\) is the unique maximizer of the multi-class Brier scoring rule's expectation. Therefore, it is strictly proper.
\end{proof}

\subsubsection{Using softmax with the NLL Loss}
Let us discuss the most popular setup for classification that uses the steps introduced in Section~\ref{ssec:roadmap}.

Using a softmax output with the NLL loss is by far the most common activation and loss function for (multi-class) classification problems. Luckily, it is also designed to handle the aleatoric uncertainty in the true \(P(y \mid x)\) distribution, which is potentially multimodal (according to humans). \textbf{Note}: NLL loss = CE loss = softmax CE loss = log-likelihood loss = negative log probability for classification.

\textbf{Ingredient 1.} The softmax output \(f(x)\) for input image \(x\) has the right dimensionality (number of classes) to represent any \(P(y \mid x)\). \(f(x)\) outputs the parameter vector \(p\) of the output categorical distribution. Therefore, the architectural condition is satisfied. The model is ready to represent aleatoric uncertainty.

\textbf{Ingredient 2.} Is the method also \emph{encouraged} to represent the \emph{true} aleatoric uncertainty? We consider the loss function \(-\log f_Y(x)\). We have seen that, in expectation of \(Y\), it guides the model to produce the GT distribution \(P(y \mid x)\).

\subsubsection{Toy experiment with the NLL loss}

Importantly, our DNN is \emph{not} encouraged to be overconfident (nearly one-hot) for \(\argmax_k P(Y = k \mid X = x)\) when using the NLL loss, as many people suggest. The loss encourages the model to produce distributions \(f(x)\) with variance when the true distribution also has a non-zero variance. This is, of course, considering infinite data. For finite datasets where the model usually does not see two labelings for a single data point, it can arbitrarily overfit to the given labeling for each datum (given sufficient expressivity). This makes the model \emph{not} predict the true aleatoric uncertainty for a sample (only the empirical probability), which can result in the model being extremely overconfident in one of the possible answers. This can be possibly mitigated by adversarial training (increase region of class \(y\) prediction) or by regularizing the model based on how many data points we have.

\begin{figure}
    \centering
    \includegraphics[width=0.6\linewidth]{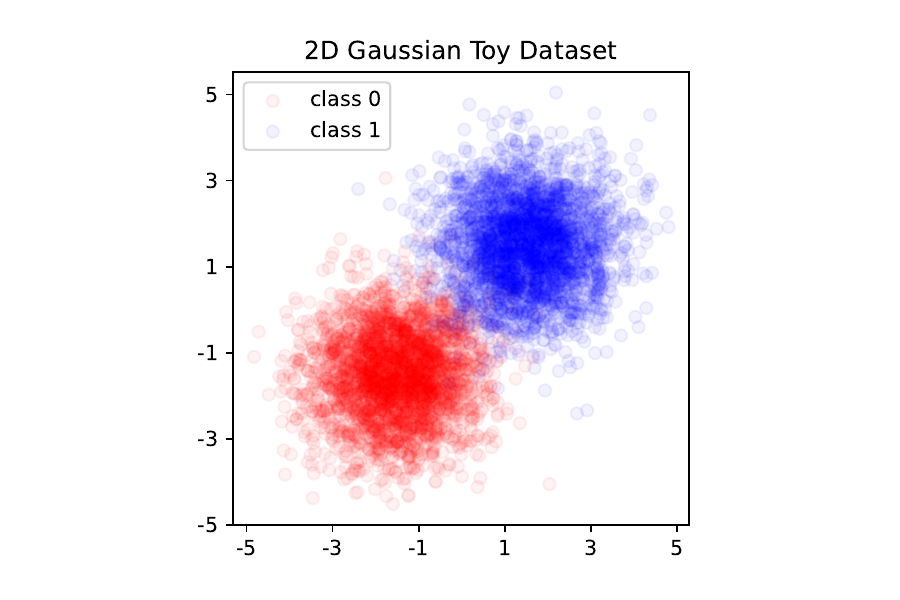}
    \caption{Homoscedastic 2D class-wise Gaussian dataset in a binary classification setting, used for the experiment in the \href{https://colab.research.google.com/drive/1ao7oyRoye2uPnfk7NFhz5AujH-jAMxAd?usp=sharing}{notebook}.}
    \label{fig:dset}
\end{figure}

\begin{figure}
    \centering
    \includegraphics[width=0.6\linewidth]{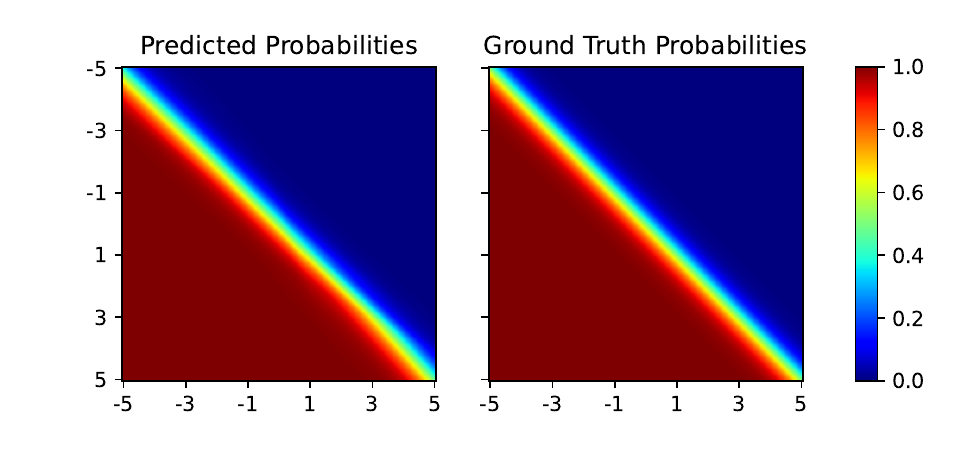}
    \caption{Predicted and ground truth probabilities from the experiment in the \href{https://colab.research.google.com/drive/1ao7oyRoye2uPnfk7NFhz5AujH-jAMxAd?usp=sharing}{notebook}. The two probability maps are almost indistinguishable.}
    \label{fig:predgt}
\end{figure}

We provide a \href{https://colab.research.google.com/drive/1ao7oyRoye2uPnfk7NFhz5AujH-jAMxAd?usp=sharing}{notebook} to clean up the possible source of the misconception of the NLL encouraging overconfidence. The dataset is generated from two homoscedastic 2D Gaussians with a small overlap near \((0, 0)\) (Figure~\ref{fig:dset}). The task is binary classification. Since we know the Gaussians that generate each class, we can calculate the true probability that a sample \(x\) is of class 0:
\begin{align*}
P(Y = 0 \mid X = x) &= \frac{P(X = x \mid Y = 0)P(Y = 0)}{P(X = x \mid Y = 0)P(Y = 0)+P(X = x \mid Y = 1)P(Y = 1)}\\
&= \frac{P(X = x \mid Y = 0)}{P(X = x \mid Y = 0) + P(X = x \mid Y = 1)}\\
&= \frac{\cN(x \mid \mu_0, \Sigma_0)}{\cN(x \mid \mu_0, \Sigma_0) + \cN(x \mid \mu_1, \Sigma_1)}
\end{align*}
where we assumed a uniform label prior. This is just the ratio of the likelihood of \(x\) being a part of class 0 and the total likelihood of it being a part of any of the two. We visualize the predicted (\(f_0(x)\)) and GT (\(P(y \mid x)\)) probabilities in Figure~\ref{fig:predgt}.

We have very high GT label certainty for the lower and upper triangles. On the diagonal, the GT label certainties are close to 0.5, signaling high aleatoric uncertainty.
The question is: If we train with this data, does a 2-layer DNN predict something close to this after applying sigmoid? The model will observe mixed supervision near the class boundary \(x_1 + x_2 = 0\). (It is also not expressive enough to overfit to the training set and produce incorrect aleatoric uncertainties. We have enough data points.) Such mixed supervision and the NLL objective result in the correct estimation of \(P(y \mid x)\). 

\begin{figure}
    \centering
    \includegraphics[width=0.6\linewidth]{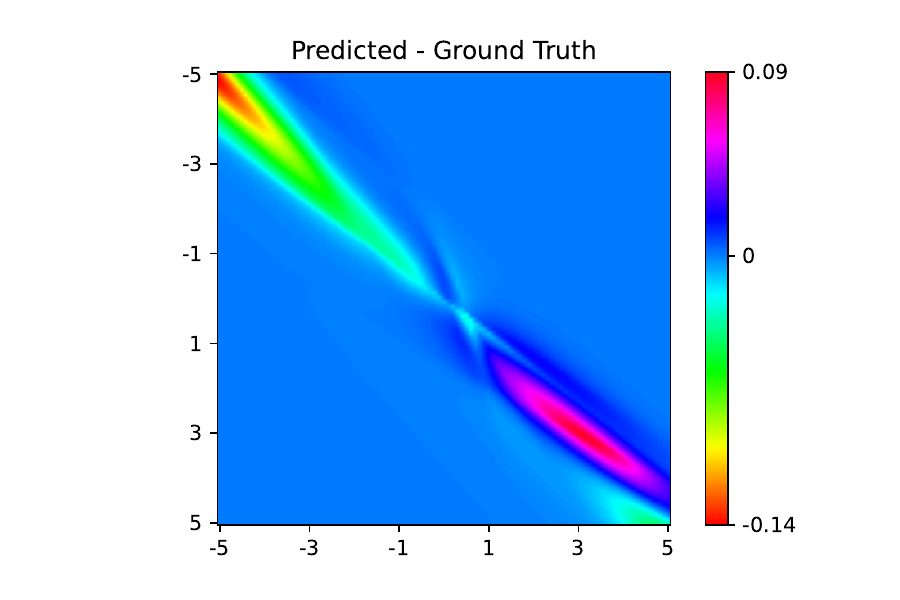}
    \caption{Difference of predicted and ground truth probabilities from the experiment in the \href{https://colab.research.google.com/drive/1ao7oyRoye2uPnfk7NFhz5AujH-jAMxAd?usp=sharing}{notebook}. The differences are mostly negligible, but there are slight deviations around the corners (where data was more sparse).}
    \label{fig:res2}
\end{figure}

The model outputs closely resemble the true \(P(Y = 0 \mid X = x)\) at nearly all \(x\) values. The model \emph{can} learn correct aleatoric uncertainty estimation (as supported by the theory of proper scoring rules). We can see the pointwise difference between \(f_0(x)\) and \(P(Y = 0 \mid X = x)\) in Figure~\ref{fig:res2}.

The pointwise differences are minor (note the scale). The popular softmax + CE design is already capable of handling aleatoric uncertainty. We have differences in regions with a few training samples. Here the model struggles a bit to find the correct values. This also has close ties with epistemic uncertainty.

\subsection{Detour: Proper Scoring Rules for Aleatoric and Predictive Uncertainty}
\label{ssec:proper_au_pu}

Let us now refer back to Section~\ref{ssec:ce_pu}, where we established the fact that the negative cross-entropy loss is a lower bound to a strictly proper scoring rule for the correctness of prediction. We also briefly mentioned that it was, in fact, not only a lower bound: The negative cross-entropy loss also guarantees the exact recovery of the true probability of correctness, \(P(L = 1)\). As the relationship between proper scoring rules for aleatoric uncertainty and predictive uncertainty is quite subtle, in this section, we wish to prove the aforementioned result and formalize this relationship.

\subsubsection{We can just use an aleatoric uncertainty proper scoring rule for the correctness of prediction.}

Informally, we will show that any strictly proper scoring rule for aleatoric uncertainty acts as a strictly proper scoring rule for the correctness of predictions as well when using max-prob confidence estimates.

\begin{claim}\label{clm:log_correctness}
If \(f(x) = \argmax_{p \in \Delta^K} \nE_{P(Y \mid X = x)} S(p, Y)\) for a strictly proper scoring rule \(S\), then \(c(x) = \max_k f_k(x) = P(L = 1)\) is the unique maximizer of any expected strictly proper scoring rule for \(P(L = 1)\) for all \(x \in \cX\). 
\end{claim}

\begin{proof}
As \(S(f(x), y)\) is strictly proper for aleatoric uncertainty, the unique maximizer of the expected score is \(f(x) = P(Y \mid X = x) \in \nR^{K}\). We need to show that \(c(x) = \max_{k \in \{1, \dots, K\}} P(Y = k \mid X = x)\) maximizes any expected strictly proper scoring rule for the correctness of prediction, i.e., \(c(x) = P(L = 1)\). To see this, one can observe that
\[P(L = 1) = P\left(Y = \argmax_{k \in \{1, \dots, K\}} P(Y = k \mid X = x)\right) = \max_{k \in \{1, \dots, K\}} P(Y = k \mid X = x).\]
This means that the \(c(x)\) we obtained is also a maximizer for any expected strictly proper scoring rule for the correctness of prediction, concluding our proof.
\end{proof}

In the above proof, we did not assume any particular strictly proper scoring rule. One could, e.g., choose the log probability scoring rule. In that case, training a model with the CE loss (NLL) minimization objective encourages both the correct prediction of the classes (original role) and the truthful report of predictive uncertainty using the max-probability confidence estimates \(c(x) = \max_k f_k(x)\), and we do not only maximize a lower bound: we have strict properness. However, we could just as well choose any other strictly proper scoring rule, and the relationship would be preserved.

\subsubsection{Sometimes, we can also use a predictive uncertainty proper scoring rule for aleatoric uncertainty.}

Below, we show that the relationship also works backward in binary classification but not in multi-class classification.

\begin{claim}
In binary classification, if \(c(x) = \max(f(x), 1 - f(x))\) is the unique maximizer of any expected strictly proper scoring rule for \(P(L = 1)\) (i.e., \(c(x) = P(L = 1)\)), then
\[f(x) = \argmax_{p \in [0, 1]} \nE_{P(Y \mid X = x)} S(p, Y)\] for any (strictly) proper scoring rule \(S\) for aleatoric uncertainty for all \(x \in \cX\).
\end{claim}

\begin{proof}
Let $x$ be arbitrary. \(c(x)\) being the unique maximizer of a strictly proper scoring rule for $P(L = 1 \mid X = x)$ implies \(c(x) = P(L = 1 \mid X = x)\). Let us denote
\[\hat{y} = \begin{cases}1 & \text{if } f(x) \ge 0.5 \\ 0 & \text{ otherwise}\end{cases}.\]
One can observe that
\[P(L = 1 \mid X = x) = P(Y = \hat{y} \mid X = x),\]
therefore, we write
\begin{equation}\label{eq:1}
c(x) = \max(f(x), 1 - f(x)) =  P(Y = \hat{y} \mid X = x).
\end{equation}

Suppose that \(\hat{y} = 1\). Then, \(\max(f(x), 1 - f(x)) = f(x)\) and \(P(Y = \hat{y} \mid X = x) = P(Y = 1 \mid X = x)\). Therefore, Equation~\ref{eq:1} becomes
\[f(x) = P(Y = 1 \mid X = x),\]
i.e., we recover the GT probability of class 1. The GT probability of class 0 is also recovered as \(1 - f(x) = 1 - P(Y = 1 \mid X = x) = P(Y = 0 \mid X = x)\).

Now, suppose that \(\hat{y} = 0\). Then, \(\max(f(x), 1 - f(x)) = 1 - f(x)\) and \(P(Y = \hat{y} \mid X = x) = P(Y = 0 \mid X = x)\). Here, Equation~\ref{eq:1} simplifies to
\[1 - f(x) = P(Y = 0 \mid X = x),\]
i.e., we recover the GT probability of class 0. The GT probability of class 1 is recovered analogously to the previous case.

Therefore, for any (strictly) proper scoring rule \(S\) for aleatoric uncertainty, we indeed obtain
\[f(x) = \argmax_{p \in [0, 1]} \nE_{P(Y \mid X = x)} S(p, Y)\]
by definition.
\end{proof}

This does not hold for the multi-class case. Here, we can denote
\[\hat{y} = \argmax_{k \in \{1, \dots, K\}} f_k(x).\]
Then, as \(c(x) = \max_{k \in \{1, \dots, K\}} f_k(x)\) and \(P(L = 1) = P(Y = \hat{y}),\) the necessary and sufficient criterion \(c(x) = P(L = 1)\) for the maximizer of the expected score can be rewritten as
\begin{equation}\label{eq:2}
\max_{k \in \{1, \dots, K\}} f_k(x) = P\left(Y = \hat{y}\right).
\end{equation}
Here, the probability vector \(f(x)\) has \(K\) elements, but only the maximal element is matched to the GT probability of the predicted class. This can have surprising consequences that were not observed in the binary case because by matching one probability there, we matched the other one as well.

Predicting a suboptimal class (one with non-maximal GT probability) \emph{does not} lead to a contradiction of Equation~\ref{eq:2}. One can convince themselves that \(f(x) := (0.32, 0.33, 0.35) \in \Delta^3\) with \(P(Y \mid X = x) = (0.5, 0.15, 0.35) \in \Delta^3\) satisfies the above constraint, meaning it is a maximizer of the expected score, even though the resulting prediction is incorrect. To understand this seemingly strange behavior, we note that \emph{the maximizer \(\argmax_{k \in \{1, \dots, K\}} f_k(x)\) can change over the course of optimization}. In fact, under any sensible loss (e.g., cross-entropy) and a network of enough capacity, the maximizer usually becomes the GT class (at least on ID samples). Optimizing not only over the maximum probability (\(\max_{k \in \{1, \dots, K\}} f_k(x)\)) but also the maximizer of it (\(\argmax_{k \in \{1, \dots, K\}} f_k(x)\)) results in the desired behavior of our network predicting the correct class and recovering the true \(\max_{k \in \{1, \dots, K\}}P(Y = k \mid X = x)\) as the corresponding predicted probability.

Still, in the multi-class case, even though we can match the probabilities corresponding to the maximizer class of \(P(Y \mid X = x)\), we get no guarantees that the remaining probabilities will also be matched. For example, \(f(x) := (0.5, 0.15, 0.35) \in \Delta^3\) with \(P(Y \mid X = x) = (0.5, 0.3, 0.2) \in \Delta^3\) (1) satisfy the constraint, (2) predict the correct class, but (3) the probabilities are not matched. Therefore, the equivalence of max-prob proper scoring rules for predictive uncertainty and aleatoric uncertainty only holds in the binary case. In the multi-class case, we cannot establish such a correspondence.

\subsubsection{Wait, what about epistemic uncertainty?}

One intuitive explanation for the above results is that proper scoring rules only give statements about the recovery of the true probabilities \emph{in expectation}, or in other words, in the limit of infinitely many labels for a given input \(x\). In this limit, epistemic uncertainty naturally vanishes for this particular input \(x\), as we have full information about the (conditional) data generating process \(P(Y \mid X = x)\). Therefore, it makes sense that proper scoring for predictive uncertainty reduces to that for aleatoric uncertainty.

\subsection{Aleatoric Uncertainty in Regression}
\label{ssec:au_regression}

In regression, our true distribution \(P(y \mid x)\) is continuous and has variance in all sensible settings (think of measurement noise, to begin with). Thus, \(Y\) is a continuous random variable. We now assume that the true distribution follows a normal distribution:
\[P(y \mid x) = \cN(y \mid \mu(x), \sigma(x)^2I).\]
The usual setting is that \(\sigma^2\) does not depend on \(x\), i.e., it is \emph{homoscedastic}. However, that would make aleatoric uncertainty prediction uninteresting, since it would return constant values for all $x$. Thus, we allow $\sigma(x)$ to vary with \(x\), i.e., it is \emph{heteroscedastic}. We will see that heteroscedastic regression can model this aleatoric uncertainty (when trained right).

\begin{figure}
    \centering
    \includegraphics[width=0.6\linewidth]{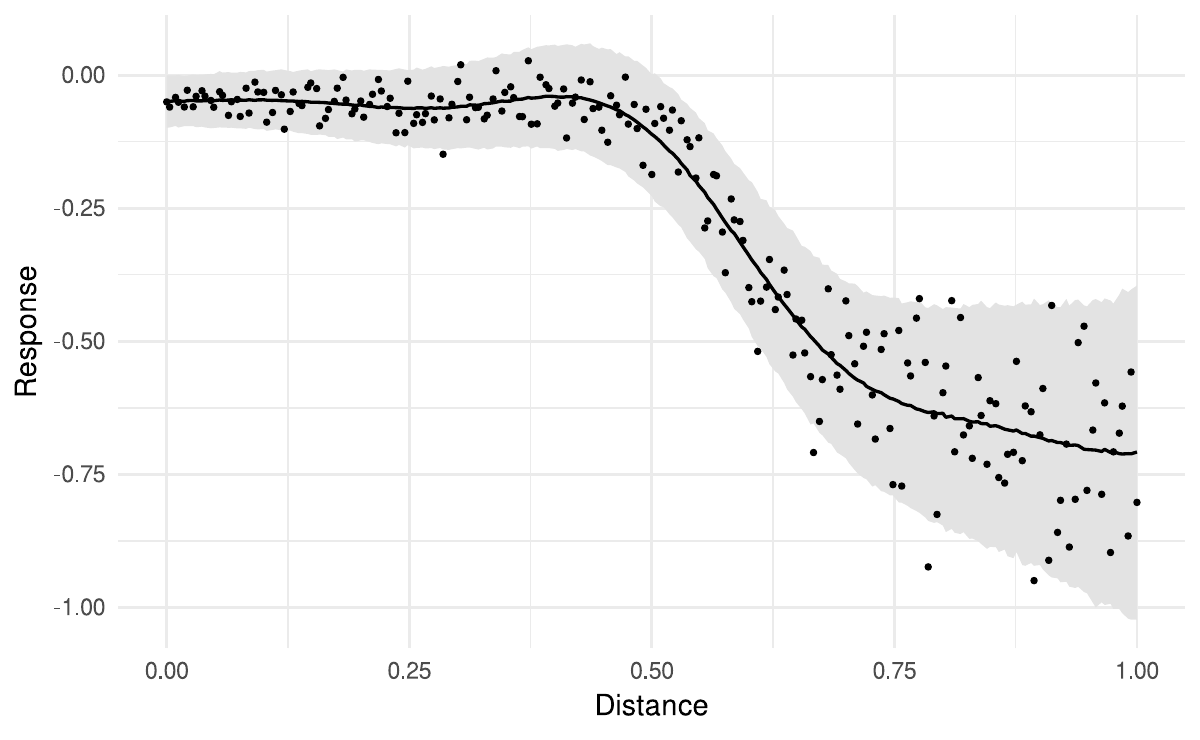}
    \caption{Example heteroscedastic regression dataset. Figure taken from~\cite{ivanukhov}.}
    \label{fig:heterosc}
\end{figure}

Heteroscedastic regression can be nicely visualized in 1D as shown in Figure~\ref{fig:heterosc}. Each \(x\)-slice corresponds to a Gaussian distribution with a different \(\sigma\). \(\sigma^2(x)\) is the variance (level of noise) of \(P(y \mid x)\) that depends on \(x\). Our goal is to correctly represent not only \(\mu(x)\) (what we usually do) but also \(\sigma^2(x)\) at every \(x\).

\subsubsection{Example of connecting epistemic and aleatoric uncertainty in heteroscedastic regression}

\begin{figure}
    \centering
    \includegraphics[width=0.8\linewidth]{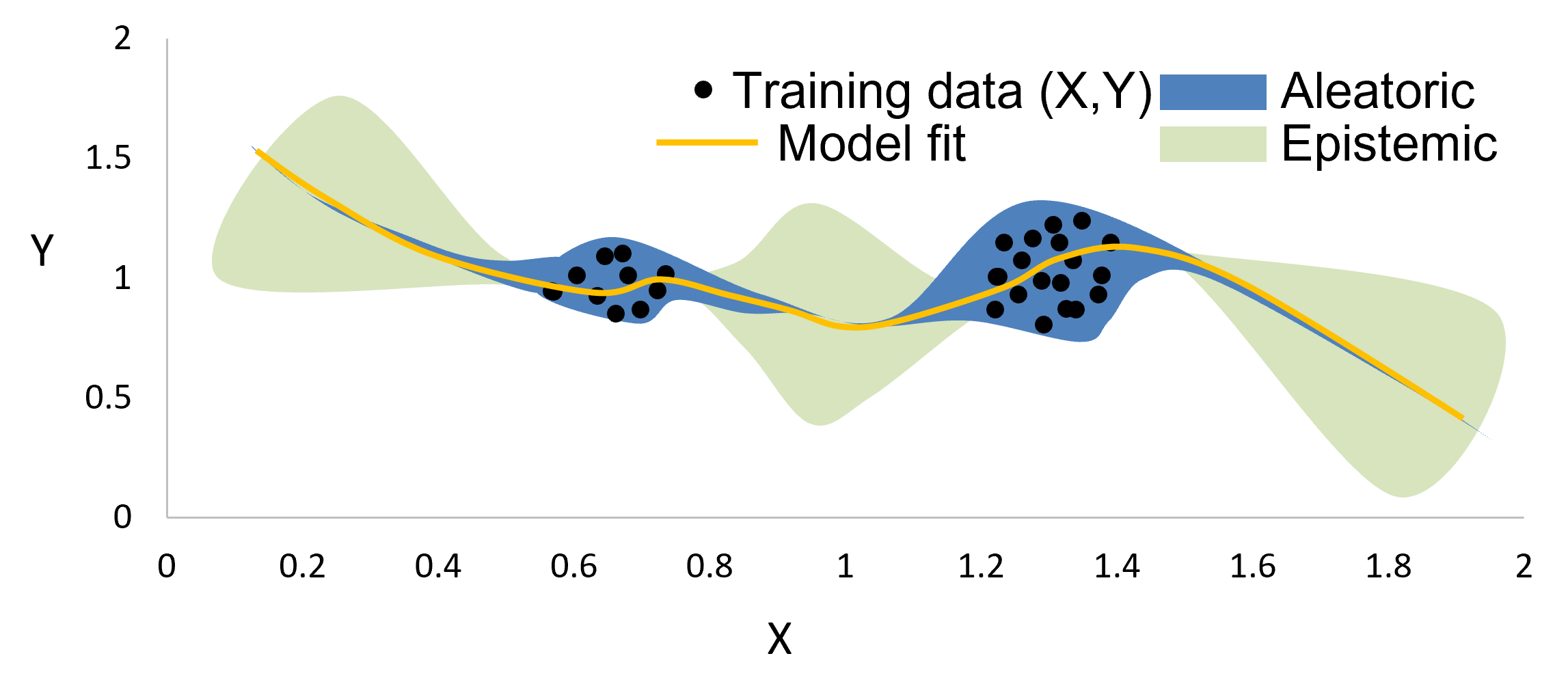}
    \caption{Connection of epistemic and aleatoric uncertainty. In distribution, we can still have high aleatoric uncertainty as the generating process can be arbitrarily noisy. Out of distribution, epistemic uncertainty increases. Figure taken from~\cite{https://doi.org/10.48550/arxiv.2107.01557}.}
    \label{fig:heterosc2}
\end{figure}

We briefly discuss an illustration from the work ``\href{https://arxiv.org/abs/2107.01557}{Leveraging Graph and Deep Learning Uncertainties to Detect Anomalous Trajectories}''~\cite{https://doi.org/10.48550/arxiv.2107.01557}, shown in Figure~\ref{fig:heterosc2}. On the left, we see no training data, resulting in high epistemic uncertainty. To the right of this region, we see aleatoric uncertainty with heteroscedasticity. Here, we have more training data which results in less epistemic uncertainty. However, label noise leads to large aleatoric uncertainty. (In underexplored regions, we, of course, might also have high aleatoric uncertainty, but we have no observations to estimate that.)

\subsubsection{Another example application of heteroscedastic regression: monocular depth estimation}

\begin{figure}
    \centering
    \includegraphics[width=0.8\linewidth]{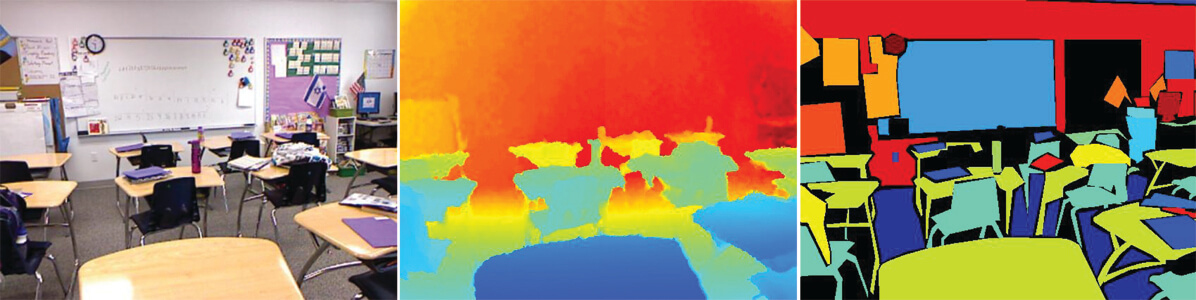}
    \caption{Depth estimation is a field that can benefit from uncertainty estimates. On the left, we see an image from an RGB camera. In the middle, we see a preprocessed (metric) depth map where objects farther away are represented by red color. On the right, per-pixel labels are given for the image using \emph{semantic segmentation}. Figure taken from~\cite{nyu}.}
    \label{fig:nyu}
\end{figure}

A sample from the \href{https://cs.nyu.edu/~silberman/datasets/nyu_depth_v2.html}{NYU Depth Dataset V2}~\cite{nyu} is shown in Figure~\ref{fig:nyu}. A single image does not contain all information about which point is at what distance (perspective projection loses information about how far away the projected point was). A black box that does not reflect any light means that there is \emph{no way} we can figure out its depth in the image. Thus, we have an inherent ambiguity of depth corresponding to a single image: we are greeted by aleatoric uncertainty. Also, the ``deeper'' the object, the more noise there is in the true depth value (noisy label). Furthermore, depending on the position in the image, we have different levels of aleatoric uncertainty -- we need per-pixel estimates. This leads to heteroscedastic regression with aleatoric uncertainty: We want to predict a depth map (regression problem) and represent the corresponding aleatoric uncertainties. One could, e.g., do this by using a Gaussian noise model where the aleatoric uncertainty is represented by the standard deviation of the Gaussian at a particular pixel.

\subsubsection{Log Probability Scoring Rule}

To follow our recipe from Section~\ref{ssec:roadmap} in a regression setting, let us extend the definition of scoring rules to matching a continuous distribution \(Q(Y)\) to the true continuous distribution \(P(Y)\).
Let \(y\)  be a sample of \(P(Y)\). We define the scoring rule as a function \(S(Q, y)\). It is strictly proper when its expected score \(\nE_{P(Y)} S(Q, Y)\) is maximized iff \(Q \equiv P\), i.e., the aleatoric uncertainty estimation is correct.
A reasonable expectation at this point is that if we train with some proper scoring rule for the correct output distribution, we will be able to ``solve'' the problem of aleatoric uncertainty.

The log probability scoring rule for continuous distributions is defined by \(S(Q, y) = \log Q(y)\), the log density of the sample. And indeed, one can show that it is strictly proper \wrt \(P\).

\subsection{Using Proper Scoring Rules to Recover \(\cN(\mu(x), \sigma^2(x)I)\)}

Let us assume a regression setting where our actual conditional distribution is a \emph{heteroscedastic} (depends on \(x\)), \emph{isotropic} (same variance in all directions, defined by a single number \(\sigma^2(x)\)) Gaussian:
\(P(y \mid x) = \cN(y \mid \mu(x), \sigma^2(x)I)\).

We wish to train a model with output
\(f(x) = \left(\hat{\mu}(x), \hat{\sigma}(x)\right)\). Our model has two heads, one for predicting the mean, and one for predicting the standard deviation in each direction, at each point \(x\). Our predicted distribution is then
\(Q(y \mid x) = \cN\left(y \mid \hat{\mu}(x), \hat{\sigma}^2(x)I\right)\).
As can be seen, our approximate distribution is also a heteroscedastic, isotropic Gaussian, just like the true conditional distribution. Therefore, in principle, we can recover the true distributions.

For a fixed input \(x\), the log probability scoring rule yields
\[-S(q, y) = -\log Q(y \mid x) = -\log \left(\frac{1}{\sqrt{(2\pi)^d(\hat{\sigma}^2(x))^d}}\exp\left[-\frac{1}{2\hat{\sigma}^2(x)} \Vert y - \hat{\mu}(x) \Vert_2^2\right]\right)\]
where
\(y \sim P(Y \mid x) = \cN(Y \mid \mu(x), \sigma^2(x)I)\)
is a sample from the true conditional distribution. This leads to the equation
\[-S(q, y) = \frac{1}{2\hat{\sigma}^2(x)}\Vert y - \hat{\mu}(x)\Vert_2^2 + \frac{d}{2}\log \hat{\sigma}^2(x) + C.\]

\textbf{Minimizing the above quantity has a very intuitive interpretation.} We use a scaled \(L_2\) loss for fitting \(\hat{\mu}\) to the GT label (i.e., we perform MLE on the mean). The \(L_2\) loss is weighted by the sample-wise weight \(\frac{1}{2\hat{\sigma}^2(x)}\). When the model says \(x\) is ambiguous (it is having a hard time with it, uncertain about the prediction), i.e., \(\hat{\sigma}^2\) is high, the loss penalizes less for the wrong \(\hat{\mu}\), as the weight is small. The \(\log \hat{\sigma}^2(x)\) term prevents the model from saying that every sample is hard (ambiguous). This is the balancing term (regularizer) that gives a large penalty when the model is always uncertain about the correct value for \(\hat{\mu}(x)\). Otherwise, the model could be extremely uncertain for all samples, and that could result in an arbitrarily small objective value. This forces the model only to be uncertain (high variance in predicted conditional) when it has to be (because of target ambiguities). In such cases, the first term does not penalize much, allowing deviations from the (disagreeing) labels. The model learns to say, ``I don't know.'' when the true conditional has high variance, too (conflicting labels). Here, we are still performing MLE, but now both over the means and standard deviations per sample.

\subsubsection{Results of Proper Scoring on a Toy Dataset}

\begin{figure}
    \centering
    \includegraphics[width=0.5\linewidth]{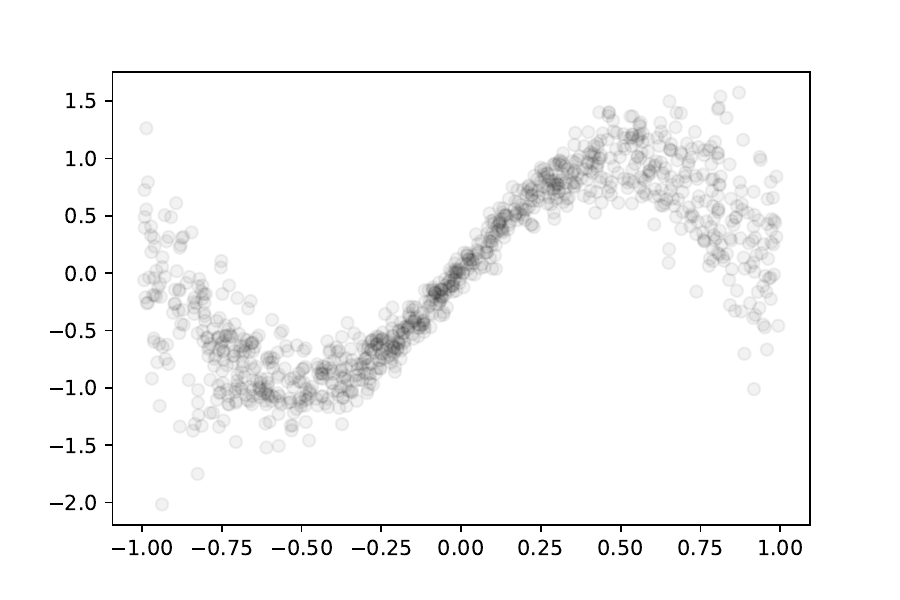}
    \caption{Toy regression dataset with heteroscedastic aleatoric uncertainty from the \href{https://colab.research.google.com/drive/12Eds3sX9Ago38iS9CXcMgEEt36Ab5xuq?usp=sharing}{notebook}.}
    \label{fig:notebook}
\end{figure}

In this section, we aim to verify that this training gives us good aleatoric uncertainty predictions in practice. To this end, a three-layer FCN is fit to the dataset in Figure~\ref{fig:notebook}. Aleatoric uncertainty is present in the dataset. The uncertainty is also heteroscedastic, i.e., we have varying degrees of mixed supervision during training. For the ``same'' \(x\), we will see many different values of \(y\). This can confuse the model, as it often gets marginally different GT labels for very similar inputs. However, the model has a way to express its per-sample confusion: via the predicted standard deviation \(\hat{\sigma}(x)\). The model does not have enough capacity to overfit to every single sample. There are also enough data points to counteract this.

\begin{figure}[ht]
    \centering
    \begin{minipage}[b]{0.48\linewidth}
        \centering
        \includegraphics[width=\linewidth]{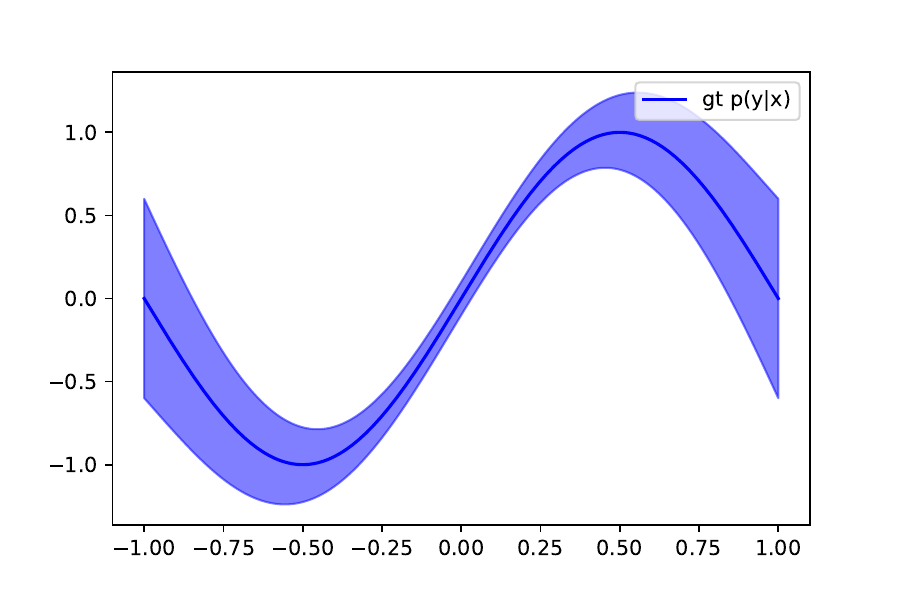}
        \caption{Ground truth aleatoric uncertainties from the \href{https://colab.research.google.com/drive/12Eds3sX9Ago38iS9CXcMgEEt36Ab5xuq?usp=sharing}{notebook}.}
        \label{fig:gt}
    \end{minipage}%
    \hfill
    \begin{minipage}[b]{0.48\linewidth}
        \centering
        \includegraphics[width=\linewidth]{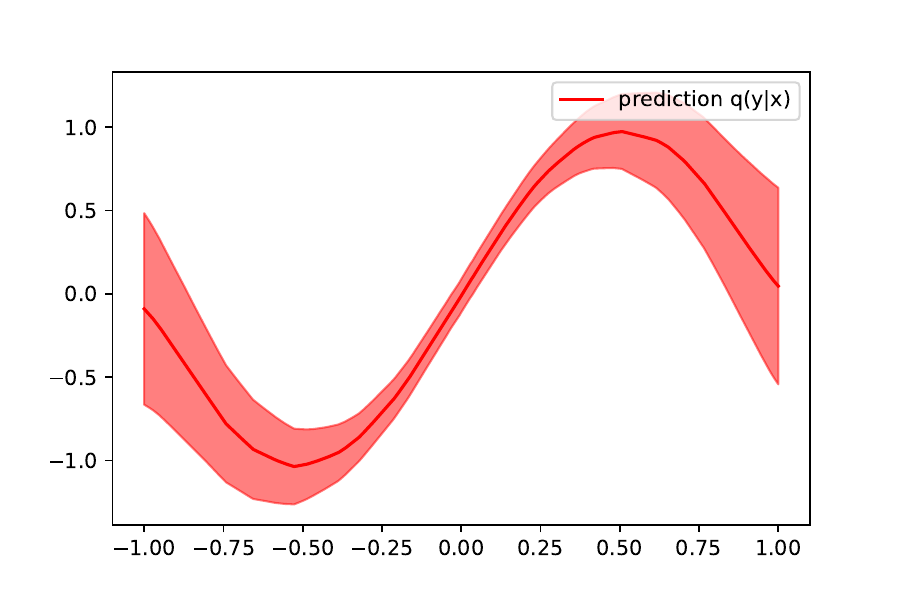}
        \caption{Predicted aleatoric uncertainties from the notebook.}
        \label{fig:pred}
    \end{minipage}
\end{figure}

The ground truth and predicted conditionals (\(x\)-slices) are shown in Figures~\ref{fig:gt} and \ref{fig:pred}, respectively. The predictions were attained by minimizing the loss introduced above, i.e., performing MLE over means and homoscedastic standard deviations. Training with the introduced proper scoring rule really enables the recovery of the correct conditional distribution \emph{given enough data}. The predicted means and standard deviations match well with their GT counterparts: The model represents faithful aleatoric uncertainty estimates. Whenever we have to represent aleatoric uncertainty in our regression problem, we recommend using this loss function or similar probabilistic formulations.

\subsection{Aleatoric Uncertainty for Complex Output: Multimodal Future Prediction}

\begin{figure}
    \centering
    \includegraphics[width=0.5\linewidth]{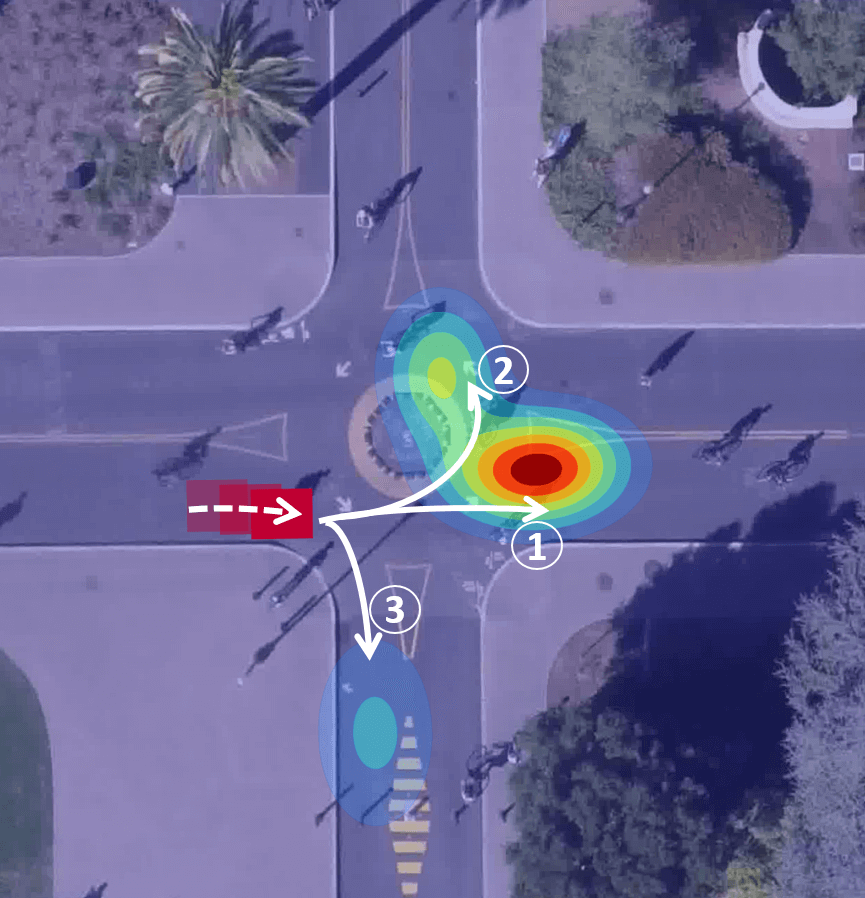}
    \caption{Presence of aleatoric uncertainty in future prediction.
            ``Given the past images, the past positions of an object
            (red boxes), and the experience from the training data, the approach predicts a multimodal distribution over future states of that
            object (visualized by the overlaid heatmap). The vehicle is most
            likely to move straight (1), but could also continue on the roundabout (2) or turn right (3).''~\cite{Makansi_2019}
            Figure taken from~\cite{Makansi_2019}.}
    \label{fig:intersection}
\end{figure}

So far, we have only seen a 1D regression problem and a 2D classification problem. What about more complex outputs and more complex conditional distributions \(P(Y \mid X = x)\)? We briefly discuss modeling aleatoric uncertainty for \href{https://arxiv.org/abs/1906.03631}{multimodal future prediction}~\cite{Makansi_2019}, illustrated in Figure~\ref{fig:intersection}. The task is future prediction, which also matters a lot for self-driving cars: Where will another car go in the next 5 seconds? Where will a pedestrian move in the next 10 seconds? The task inherently has aleatoric uncertainty, as there is no single correct answer for future position prediction in most cases. In regression, we observed the same thing. However, here, the task also inherently has a multimodal true conditional distribution \(P(Y \mid X = x)\) over future positions for many image inputs \(x\), as vehicles can have multiple equally plausible trajectories at intersections, roundabouts, and even on highways. In the training set, we will see many cases of mixed supervision. There will be many cases where a vehicle will take turns (1), (2), or (3) as well in Figure~\ref{fig:intersection}, coming from the same direction. Therefore, we need something better than a single Gaussian to approximate the true multimodal conditional distribution. We have to accommodate all the complexity that can happen in the future.

\subsection{Aleatoric Uncertainty for Complex Output: Image Inpainting}

\begin{figure}
    \centering
    \includegraphics[width=0.7\linewidth]{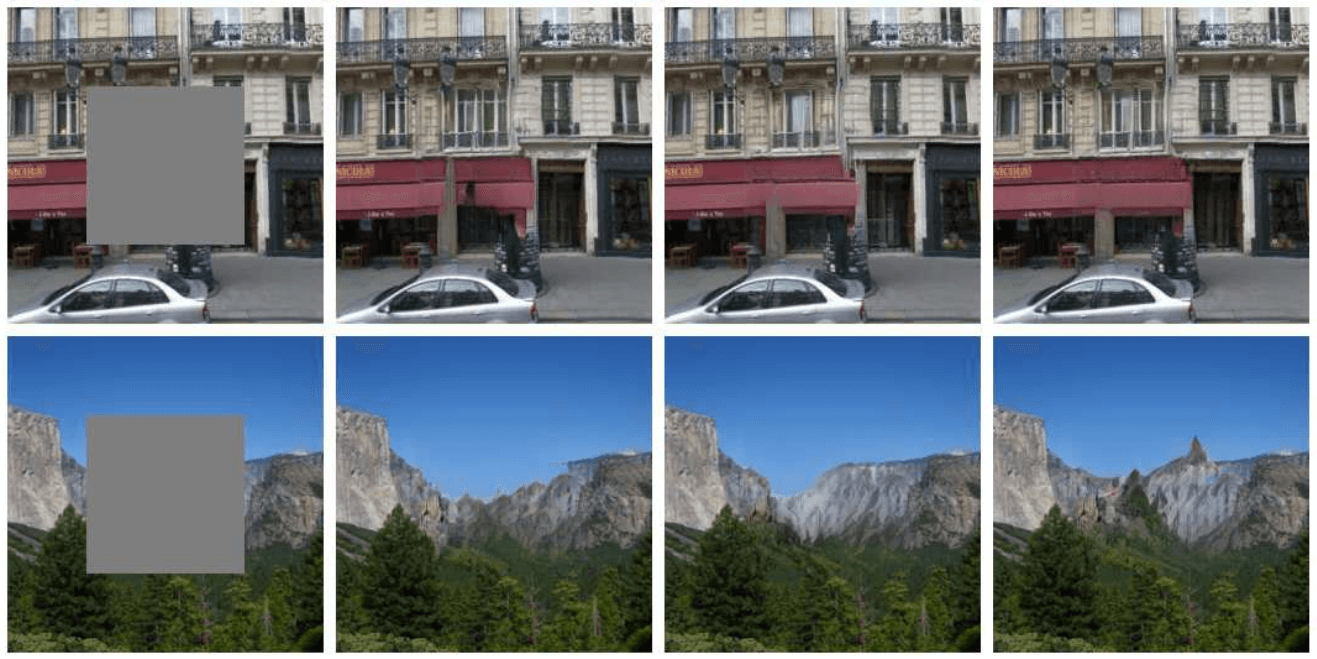}
    \caption{Presence of aleatoric uncertainty in image inpainting. There are multiple plausible reconstructions of the gray areas. Figure taken from~\cite{zhao2020uctgan}.}
    \label{fig:inpainting}
\end{figure}

Now, we turn to model aleatoric uncertainty for \href{https://openaccess.thecvf.com/content_CVPR_2020/papers/Zhao_UCTGAN_Diverse_Image_Inpainting_Based_on_Unsupervised_Cross-Space_Translation_CVPR_2020_paper.pdf}{image inpainting}~\cite{zhao2020uctgan}, as shown in Figure~\ref{fig:inpainting}. The task is inpainting. It inherently has aleatoric uncertainty, as there is no single correct answer for missing parts of images.

However, we have thousands of dimensions now, as we do not model a 2D position stochastically; rather, we model an entire missing patch in the image stochastically. (I.e., we have a \emph{joint} distribution of pixels in the missing patch, conditioned on the current input.) We have a large degree of freedom for the missing image patch based on the data manifold. It is complicated to fit a faithful parametric distribution in thousand-dimensional spaces. Usually, we assume a simple parametric distribution. However, such simplifications harm the faithfulness of the distribution we fit. For example, a Gaussian centered at a plausible fill does not explain all possible variations but also has high density for implausible window contents. The true conditional distribution is highly multimodal and complex.

In Figure~\ref{fig:inpainting}, we see samples according to the density \(P(y \mid x)\) for two images, which shows us that this distribution is indeed highly multimodal, as all inpainted regions are plausible. We can also think of this as a regression problem with per-pixel aleatoric uncertainty predictions, but we still have to consider thousands of dimensions. By doing so, we assume independence between the output variables (missing pixels in the patch), which is a notable simplification.

\subsection{Aleatoric Uncertainty for Complex Output: 2D to 3D}

\begin{figure}
    \centering
    \includegraphics[width=0.5\linewidth]{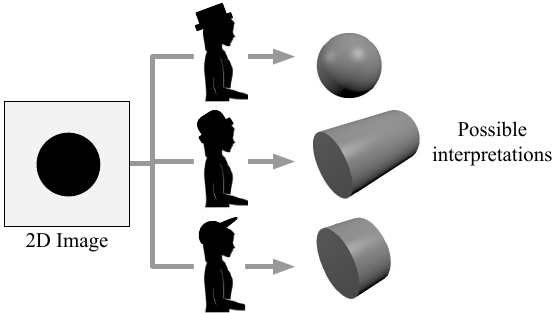}
    \caption{Presence of aleatoric uncertainty in 2D to 3D ``projection''. Infinitely many 3D objects can give rise to the same 2D projection. Figure taken from~\cite{https://doi.org/10.48550/arxiv.2008.10634}.}
    \label{fig:2d3d}
\end{figure}

Let us now consider the task of \href{https://arxiv.org/abs/2008.10634}{2D to 3D}~\cite{https://doi.org/10.48550/arxiv.2008.10634} ``projection''. An intuitive illustration is shown in Figure~\ref{fig:2d3d}. Here, the task is providing 3D reconstruction from a 2D projection. This is highly aleatoric: many different shapes give the same 2D projection. Predicting the 3D shape from a 2D projection is ill-posed. Different experts predict different latent 3D shapes that could have resulted in the given 2D projection. We, again, have ``thousands of degrees of freedom'' (actually, we have infinitely many) in the shapes that give the same 2D projection. If we consider thousands of dimensions of variation, it is complicated to fit a parametric distribution. Therefore, the space of all possible solutions of 3D shapes is an ultra-high-dimensional space with multimodality. We need to handle such complexity as well.

\subsection{Common Challenges in Aleatoric Uncertainty}

The presence of aleatoric uncertainty signals that there is no single good answer. We have mixed supervision during training: \(Y \mid X = x\) has a non-zero variance. Moreover, \emph{the distribution of possible answers is not simple}. Considering continuous outputs, it is challenging to estimate the true density \(P(y \mid x)\) with a Gaussian \(Q(y \mid x)\) outside the realm of simple problems. Usually, we have a multimodal true density \(P(y \mid x)\), and we often also have a high-dimensional output variable \(Y \mid X = x\). Predicting multimodal distributions over high-dimensional output spaces (without simplifications arising from independence assumptions) is extremely hard.

\subsection{MoGs with Fixed Variance for Aleatoric Uncertainty Estimation in Regression Problems}

When the true conditional distribution \(P(Y \mid X = x)\) is multimodal, a natural choice for approximating the density \(P(y \mid x)\) is to use a Mixture of Gaussians (MoG). They have much capacity to model a lot of different distributions. To simplify derivations and the distributions we model, we assume a fixed variance (homoscedasticity) over all inputs and Gaussians in the mixture. We also only model an isotropic mixture, meaning that we encode the variance of each Gaussian by the same, single fixed number. Isotropic MoGs with fixed variance can be written as
\[Q(y \mid x) = \frac{1}{M} \sum_m Q_m(y \mid x)\]
where
\[Q_m(y \mid x) = \cN(y \mid f_m(x), \sigma^2I).\]
\(Q(y \mid x)\) is the \emph{estimated} multimodal conditional distribution. MoGs are ``universal approximators'' as \(\sigma^2 \downarrow 0\) and \(M \rightarrow \infty\). We can approximate almost any density in the space with this approximator.

\begin{information}{Alternative Interpretation of MoGs}
MoGs have close connections to kernel regression in non-parametric statistics (in particular, the RBF kernel). It also has connections to GPs: the individual mixture components can be GPs. However, in kernel regression, \(M\) means the number of data points in the dataset: As \(M \rightarrow \infty\), we can recover the true function (the Bayes regressor). Here we are not talking about this interpretation.
\end{information}

\begin{figure}
    \centering
    \includegraphics[width=0.5\linewidth]{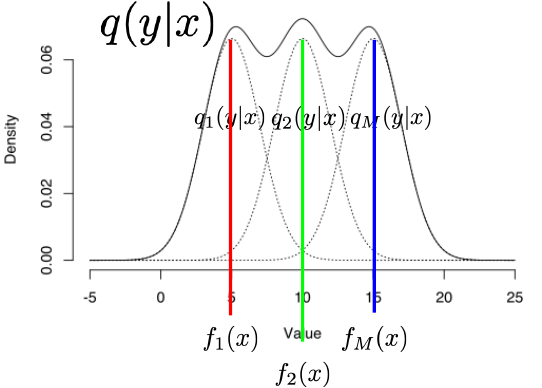}
    \caption{Gaussian mixture with three modes corresponding to three model outputs.}
    \label{fig:mog}
\end{figure}

An illustration of MoGs used for our purposes is shown in Figure~\ref{fig:mog}.

\begin{figure}
    \centering
    \includegraphics[width=0.4\linewidth]{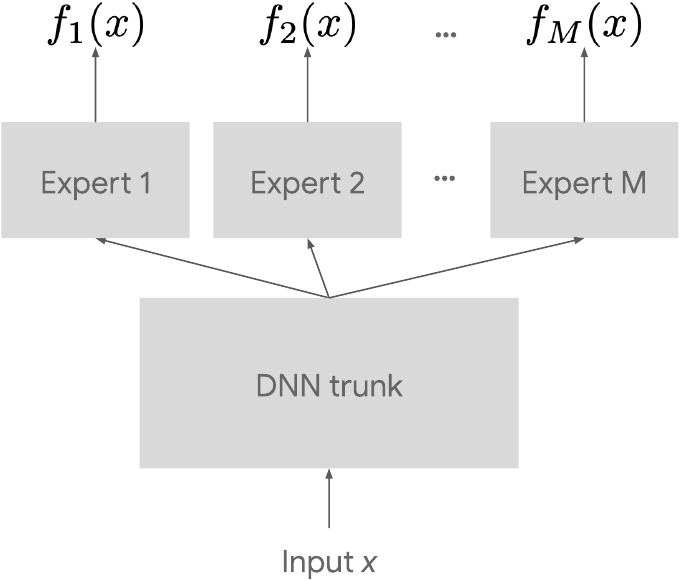}
    \caption{Illustration of separately parameterized expert heads on a common DNN trunk.}
    \label{fig:experts}
\end{figure}

For practical purposes to model MoGs, it makes sense to consider the ``experts'' interpretation and a multi-head DNN, as shown in Figure~\ref{fig:experts}.  We have a common DNN trunk and \(M\) experts (heads) on top of the common DNN trunk. (``Experts'' is the word used in the field.) The MoG is described by the \(M\) experts, with each component centered around the output of the corresponding head. We also assign parameters separately to different experts -- each expert reports their own hypothesis on the correct output.

\subsubsection{Training MoGs for Aleatoric Uncertainty Estimation}

How should we train the model to capture the diversity of \(P(y \mid x)\) that we will see in the training data? The architecture clearly allows us to do so, but we need to ensure that we encourage this behavior during training.

We have seen that the log probability is a proper scoring rule for capturing a continuous \(P(y \mid x)\). We plug our approximate distribution into the negative log probability (NLL) to obtain
\[\cL := -\log Q(y \mid x) = -\log \left(\frac{1}{M} \sum_m Q_m(y \mid x)\right).\]
This is the negative log of the MoG PDF evaluated at GT sample \(y\). We aim to write down the gradient of this loss function \wrt the weights of individual experts in terms of the losses of the individual heads. The motivation for this will become clearer later. The individual gradients tell us what the signal is we are sending to the individual experts for training their parameters. We will soon see that the total loss \(\cL\) of a single sample decomposes very nicely into a weighted sum of individual losses of the experts. Thus, the total gradient also has a nice interpretation. We will also see that the K-diverse loss has many connections to this MoG NLL formulation.

\subsubsection{Decomposing the Total Gradient}

For the shared backbone, the gradient calculation is simple and coincides with how we regularly train DNNs: We backpropagate to the backbone parameters and calculate
\[\frac{\partial \cL}{\partial \theta_{\mathrm{backbone}}},\]
then perform a gradient update as usual.

However, the gradients of the experts give rise to insightful intuitions we wish to discuss in detail. To this end, let us analyze the MoG loss in terms of the losses of \emph{individual experts}:
\[\cL_m := -\log Q_m(y \mid x).\]
Consider expert \(l \in \{1, \dots, M\}\) that has weights \(\theta_l\) that are separated from the DNN trunk. By noting that
\[Q_m(y \mid x) = \frac{1}{\sqrt{(2\pi)^d(\sigma^2)^d}}\exp\left(-\frac{1}{2\sigma^2}\Vert y - f_m(x) \Vert_2^2\right),\]
we write
\begin{align*}
\frac{\partial}{\partial \theta_l}\cL &= -\frac{\partial}{\partial \theta_l} \log \left(\frac{1}{M} \sum_m Q_m(y \mid x)\right)\\
&= -\frac{\frac{\partial}{\partial \theta_l} \frac{1}{M}\sum_m Q_m(y \mid x)}{\frac{1}{M}\sum_m Q_m(y \mid x)}\\
&= -\frac{\sum_m \frac{\partial}{\partial \theta_l} Q_m(y \mid x)}{\sum_m Q_m(y \mid x)}\\
&= -\frac{\sum_m \frac{\partial}{\partial \theta_l} \exp\left(-\frac{1}{2\sigma^2}\Vert y - f_m(x) \Vert_2^2\right)}{\sum_m \exp\left(-\frac{1}{2\sigma^2} \Vert y - f_m(x) \Vert_2^2\right)}\\
&= \frac{\exp\left(-\frac{1}{2\sigma^2}\Vert y - f_l(x) \Vert_2^2\right)}{\sum_m \exp\left(-\frac{1}{2\sigma^2} \Vert y - f_m(x) \Vert_2^2\right)} \cdot \frac{\partial}{\partial \theta_l}\frac{1}{2\sigma^2}\Vert y - f_l(x) \Vert_2^2\\
&= w_l \cdot \frac{\partial}{\partial \theta_l}\left(-\log\exp\left(-\frac{1}{2\sigma^2} \Vert y - f_l(x) \Vert_2^2\right)\right)\\
&= w_l \cdot \frac{\partial}{\partial \theta_l} \left(-\log \left(Q_l(y \mid x) \cdot \sqrt{(2\pi)^d(\sigma^2)^d}\right)\right)\\
&= w_l \cdot \frac{\partial}{\partial \theta_l} \left(-\log Q_l(y \mid x)\right)\\
&= w_l \cdot \frac{\partial}{\partial \theta_l} \cL_l,
\end{align*}
where we defined
\[w_l := \frac{\exp\left(-\frac{1}{2\sigma^2}\Vert y - f_l(x) \Vert_2^2\right)}{\sum_m \exp\left(-\frac{1}{2\sigma^2} \Vert y - f_m(x) \Vert_2^2\right)}\]
the softmax over individual expert outputs. It is important to note here that in any implementation, we need a stopgrad operation for calculating \(w_l\), as we are doing inference (using the model parameters) to calculate the weight itself. To break \(w_l\) further apart, we can see that it is of the form
\begin{align*}
w_l &= \frac{\exp\left(-\frac{1}{2\sigma^2}\Vert y - f_l(x) \Vert_2^2\right)}{\sum_m \exp\left(-\frac{1}{2\sigma^2} \Vert y - f_m(x) \Vert_2^2\right)}\\
&= \frac{\exp\left(\log Q_l(y \mid x) + \log\left(\sqrt{(2\pi)^d(\sigma^2)^d}\right)\right)}{\sum_m \exp\left(\log Q_m(y \mid x) + \log\left(\sqrt{(2\pi)^d(\sigma^2)^d}\right)\right)}\\
&= \frac{\exp\left(-\cL_l\right)}{\sum_m \exp\left(-\cL_m\right)}\\
&= \mathrm{softmax}\left(\left[-\cL_m\right]_{m = 1}^M\right)_l.
\end{align*}
Therefore, the weights are given by a softmax over the individual negative losses. What is this encoding precisely? Softmax gives an approximation (a soft version) of the \(\argmax\) operation.\footnote{Many people consider the name a misnomer because of this and propose to use the term ``softargmax'' instead.} When \(-\cL\) is maximal, that entry in the loss vector gets a high probability. The other entries get exponentially lower values. One can also write
\[\cL = \sum_l \operatorname{stopgrad}(w_l) \cdot \cL_l + C\]
which is derived below. When an expert has a small loss on a sample \(x\), it gets a large weight and, thus, a large gradient update. If an expert performs poorly on some sample, they will not learn it later either (with high probability). This allows the best-performing model to perfect its prediction the most on the sample in question. The models, therefore, become more and more diverse as the training progresses. As the models are randomly initialized, they will probably perform well on different samples to begin with, and their diversity is reinforced through the loss formulation throughout the training procedure.

\subsubsection{Effect of the Fixed Variance \(\sigma^2\)}

Let us discuss the effect of the fixed variance \(\sigma^2\) on \emph{diversity}. We can also write the weight \(w_l\) as
\[w_l = \frac{\exp\left(\mathrm{logit}_l / \sigma^2\right)}{\sum_m \exp\left(\mathrm{logit}_l / \sigma^2\right)}\]
where \(\mathrm{logit}_l = -\frac{1}{2}\Vert y - f_l(x) \Vert_2^2\) (\wrt the softmax). The \(\sigma^2\) parameter acts as a \emph{temperature} parameter for the softmax operation to calculate the weights. We divide the logit values by this before applying \(\exp\). As \(\sigma^2 \rightarrow \infty\), the softmax operation gives a uniform distribution. We are training every expert equally. This is almost like training an ensemble but with a shared backbone. We only have minor differences between experts. When \(\sigma^2 = 1\), we have no temperature scaling in the softmax. As \(\sigma^2 \downarrow 0\), the softmax result ends up being a one-hot vector indicating the argmax among negative losses. Here we are only training the most correct expert in this case, and we completely stop gradients to other experts. This results in a very diverse set of experts.

To summarize, for high-dimensional and structural, possibly multimodal \(P(y \mid x)\), \(Q = \text{MoG}\) (with different heads of the network) and proper scoring lead to K-diverse and smooth K-diverse loss functions. The loss favors more correct solutions among the set of experts. It encourages the models to produce more diverse solutions. The choice of \(\sigma^2\) controls how diverse we want the experts to be. It is a hyperparameter.

\begin{information}{Derivation of the Alternative Formulation of the Log-Likelihood MoG Loss \(\cL\)}
The loss we start out from is
\begin{align*}
\mathcal{L} &= -\log\left(\frac{1}{M}\sum_{l}Q_l(y \mid x)\right)\\
&= -\log\left(\frac{1}{M}\sum_{l} \frac{1}{\sqrt{(2\pi \sigma^2)^d}}\exp\left(-\frac{1}{2\sigma^2}\Vert y - f_l(x) \Vert_2^2\right)\right).
\end{align*}
We want to show that
\[\mathcal{L} = \sum_{l} \operatorname{stopgrad}(w_l)\cdot \mathcal{L}_{l} + c,\]
where
\[w_l = \frac{\exp\left(-\frac{1}{2\sigma^2}\Vert y - f_l(x) \Vert_2^2\right)}{\sum_m \exp\left(-\frac{1}{2\sigma^2}\Vert y - f_m(x) \Vert_2^2\right)}\]
and
\[\mathcal{L}_l = -\log Q_l(y \mid x).\]
Let us calculate the gradient of both formulations (LHS and RHS) \wrt the entire network's weights:
\begin{align*}
\frac{\partial}{\partial \theta}\mathcal{L} &= -\frac{\frac{1}{M}\sum_l \frac{1}{\sqrt{(2\pi \sigma^2)^d}}\frac{\partial}{\partial \theta} \exp\left(-\frac{1}{2\sigma^2}\Vert y - f_l(x) \Vert_2^2\right)}{\frac{1}{M}\sum_{l} \frac{1}{\sqrt{(2\pi \sigma^2)^d}}\exp\left(-\frac{1}{2\sigma^2}\Vert y - f_l(x) \Vert_2^2\right)}\\
&= -\frac{\sum_l \frac{\partial}{\partial \theta} \exp\left(-\frac{1}{2\sigma^2}\Vert y - f_l(x) \Vert_2^2\right)}{\sum_{l} \exp\left(-\frac{1}{2\sigma^2}\Vert y - f_l(x) \Vert_2^2\right)}\\
&= -\sum_l \frac{\frac{\partial}{\partial \theta} \exp\left(-\frac{1}{2\sigma^2}\Vert y - f_l(x) \Vert_2^2\right)}{\sum_{m} \exp\left(-\frac{1}{2\sigma^2}\Vert y - f_m(x) \Vert_2^2\right)}\\
&= -\sum_l \frac{\exp\left(-\frac{1}{2\sigma^2}\Vert y - f_l(x) \Vert_2^2\right)}{\sum_{m} \exp\left(-\frac{1}{2\sigma^2}\Vert y - f_m(x) \Vert_2^2\right)} \frac{\partial}{\partial \theta}\left(-\frac{1}{2\sigma^2}\Vert y - f_l(x) \Vert_2^2\right)\\
&= -\sum_l w_l \frac{\partial}{\partial \theta}\left(-\frac{1}{2\sigma^2}\Vert y - f_l(x) \Vert_2^2\right)\\
&= \sum_l w_l \frac{\partial}{\partial \theta}\left(\frac{1}{2\sigma^2}\Vert y - f_l(x) \Vert_2^2\right),
\end{align*}
and similarly,
\begin{align*}
&\frac{\partial}{\partial \theta}\left(\sum_{l} \operatorname{stopgrad}(w_l)\cdot \mathcal{L}_{l} + c\right)\\
&= \sum_l w_l \frac{\partial}{\partial \theta} \mathcal{L}_l\\
&= \sum_l \frac{\partial}{\partial \theta}\left(-\log \frac{1}{\sqrt{(2\pi \sigma^2)^d}}\exp\left(-\frac{1}{2\sigma^2}\Vert y - f_l(x) \Vert_2^2\right)\right)\\
&= \sum_l \frac{\partial}{\partial \theta}\left(\log \sqrt{(2\pi \sigma^2)^d} + \frac{1}{2\sigma^2}\Vert y - f_l(x) \Vert_2^2\right)\\
&= \sum_l w_l \frac{\partial}{\partial \theta}\left(\frac{1}{2\sigma^2}\Vert y - f_l(x) \Vert_2^2\right).
\end{align*}
This necessarily means that the two expressions only differ in constants (\wrt the weights). However, they \emph{do} differ in constants, as we saw that we canceled some zero-gradient terms in the second expression but did not cancel any in the first.
\end{information}

\subsection{K-Diverse Loss}

The special case of MoG NLL training when \(\sigma^2 \downarrow 0\) is called the \emph{k-diverse loss} or \emph{min-loss} in literature.
\[\cL = \sum_l \operatorname{stopgrad}(w_l) \cdot \cL_l \xrightarrow[]{\sigma^2 \downarrow 0} \sum_l \bone(l = \argmax_p w_p) \cdot \cL_l = \min_l \cL_l,\]
as \(\argmax_p w_p = \argmin_p \cL_p\). The total loss is the minimal loss over the set of experts in this case. The gradient is only backpropagated through the most correct expert for each sample. That is, only the expert corresponding to the best answer gets updated. For a fixed input sample, we nearly always train the same expert. This explains why we call the heads ``experts'': Every head becomes an expert for a subset of samples. The model is incentivized to give diverse answers to every question. This is because only one of the predictions from the heads has to be correct to achieve a very low loss value. This is, of course, a very cheap shortcut for diversity. However, the loss achieves its aim: It diversifies the output predictions.

We see different forms of the min-loss in the literature, hidden under different notations. We will discuss some of these in the following sections.

\subsubsection{DiverseNet}

The \href{https://arxiv.org/abs/2008.10634}{DiverseNet}~\cite{https://doi.org/10.48550/arxiv.2008.10634} method uses the K-diverse loss:
\[\cL_\mathrm{div} = \sum_{(x, \cY) \in \cD} \sum_{y \in \cY} \min_{c \in \cC} \ell\left(f(c, x), y\right)\]
where \(\ell\) is the loss function, \(c\) is the expert (hypothesis/head) index, and \(y\) corresponds to multiple GT labels for each \(x\), sampled according to \(P(y \mid x)\).

This is the same K-diverse loss we saw before; the difference is that we always have multiple labels too, that are sampled from \(P(y \mid x)\).

The authors of~\cite{https://doi.org/10.48550/arxiv.2008.10634} want DiverseNet to output diverse solutions for each input \(x\). However, they also want each output to be sensible to avoid shortcuts like predicting a constant digit with each expert on MNIST with ten experts. To this end, they also employ a catch-up loss
\[\cL_{\mathrm{catchup}} = \frac{1}{|\cC|} \sum_{(x, \cY) \in \cD} \max_{c \in \cC} \min_{y \in \cY} \ell(f(c, x), y)\]
that encourages the worst-performing model on the sample to improve as well.

\begin{figure}
    \centering
    \includegraphics[width=\linewidth]{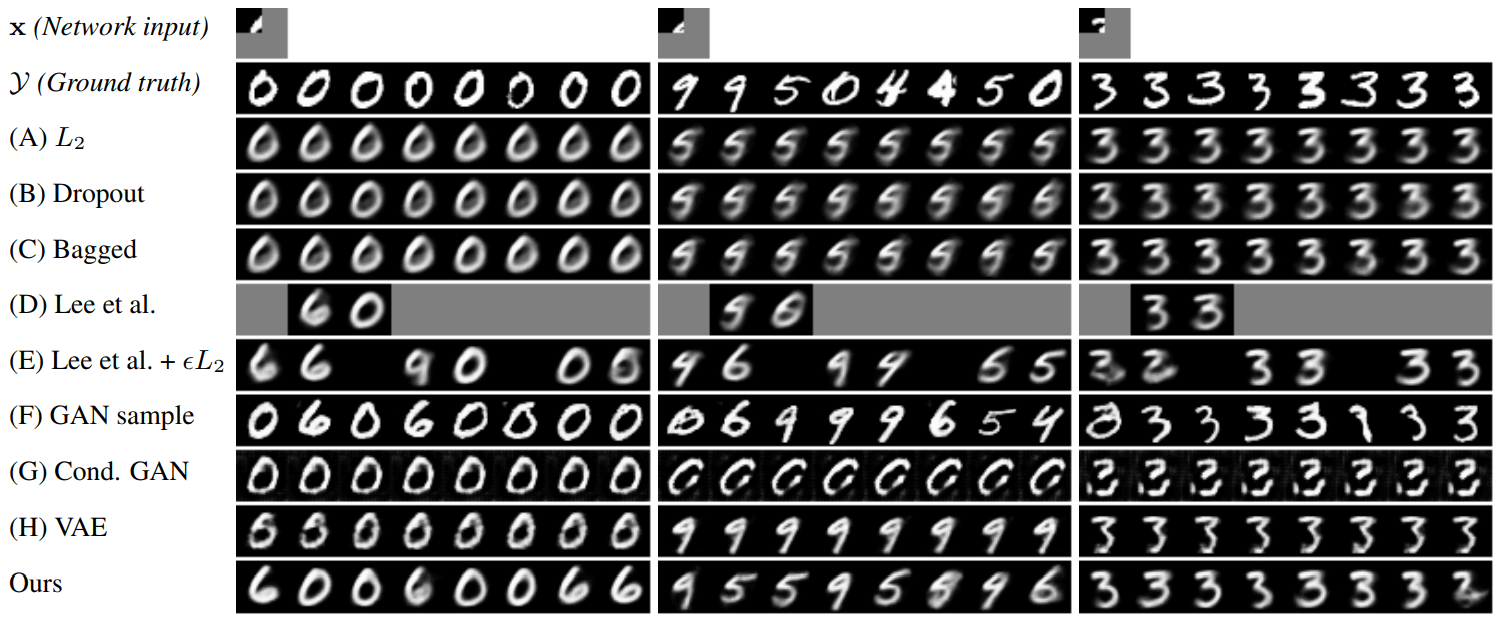}
    \caption{Predictions of various models on the MNIST outpainting task. ``[...] For each model, [...], we make eight predictions on each image. The ground truth row shows the \(\cY\) values in \(\text{MNIST}_{\text{OCC}}\), found from the test set using nearest neighbor lookup -- the left-most value in \(\cY\) is the image from which \(x\) was generated. [...]''~\cite{https://doi.org/10.48550/arxiv.2008.10634} Figure taken from~\cite{https://doi.org/10.48550/arxiv.2008.10634}.}
    \label{fig:mnist_results}
\end{figure}

If we train with the final loss
\[\cL = \cL_\mathrm{div} + \beta \cL_\mathrm{catchup}\]
with a tuned \(\beta\) parameter, we perform well on their proposed MNIST outpainting task, as illustrated in Figure~\ref{fig:mnist_results}. The input here is the upper right fraction of an MNIST image. The output is the completed MNIST image. The authors consider many different loss functions and argue that only theirs is giving a notably diverse set of solutions. Their proposed loss is a good choice when such a high aleatoric is present in most samples.

\begin{figure}
    \centering
    \includegraphics[width=\linewidth]{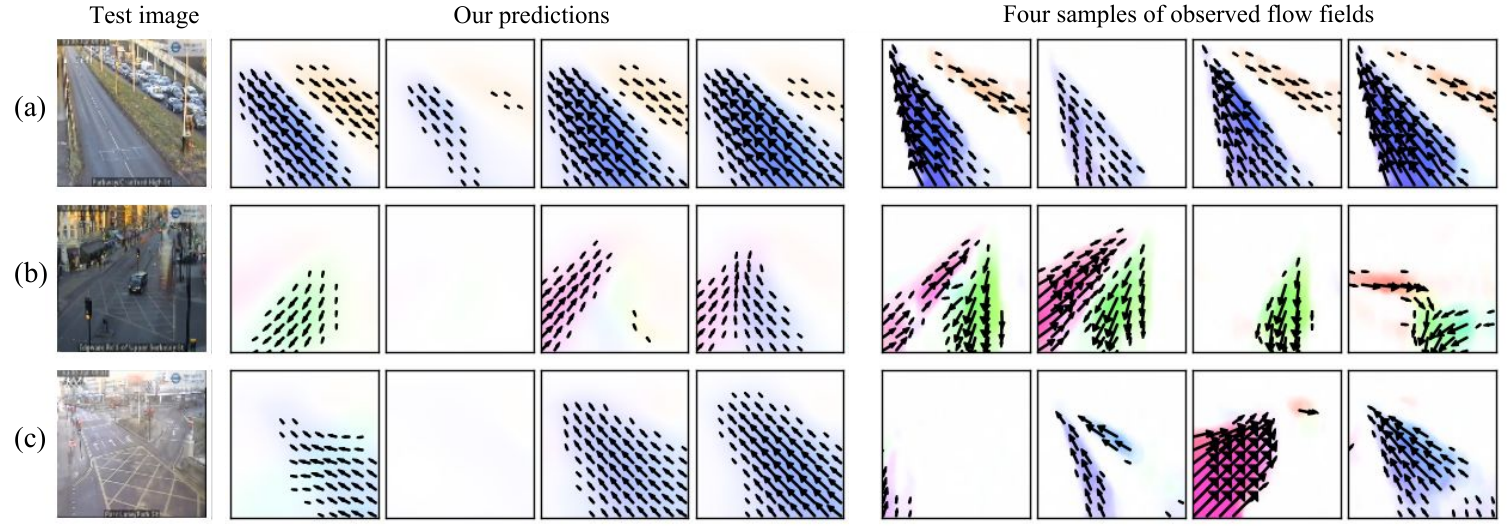}
    \caption{Results on the traffic flow prediction task, which is inherently aleatoric. The GT flow maps are notably diverse. The model makes very accurate aleatoric uncertainty predictions in some cases, but there are also several incorrect predictions. Figure taken from~\cite{https://doi.org/10.48550/arxiv.2008.10634}.}
    \label{fig:flow_results}
\end{figure}

We also discuss Figure~\ref{fig:flow_results}, where the authors apply their model to a traffic direction prediction task. Given an image, the task is to predict the possible passage of traffic in the future from the same view. The GT \(P(y \mid x)\) flow fields from the same view show that the task has large inherent aleatoric uncertainty. The authors argue that their method gives a diverse set of predictions for the possible flow of traffic that are also truthful to the observations.

\begin{figure}
    \centering
    \includegraphics[width=\linewidth]{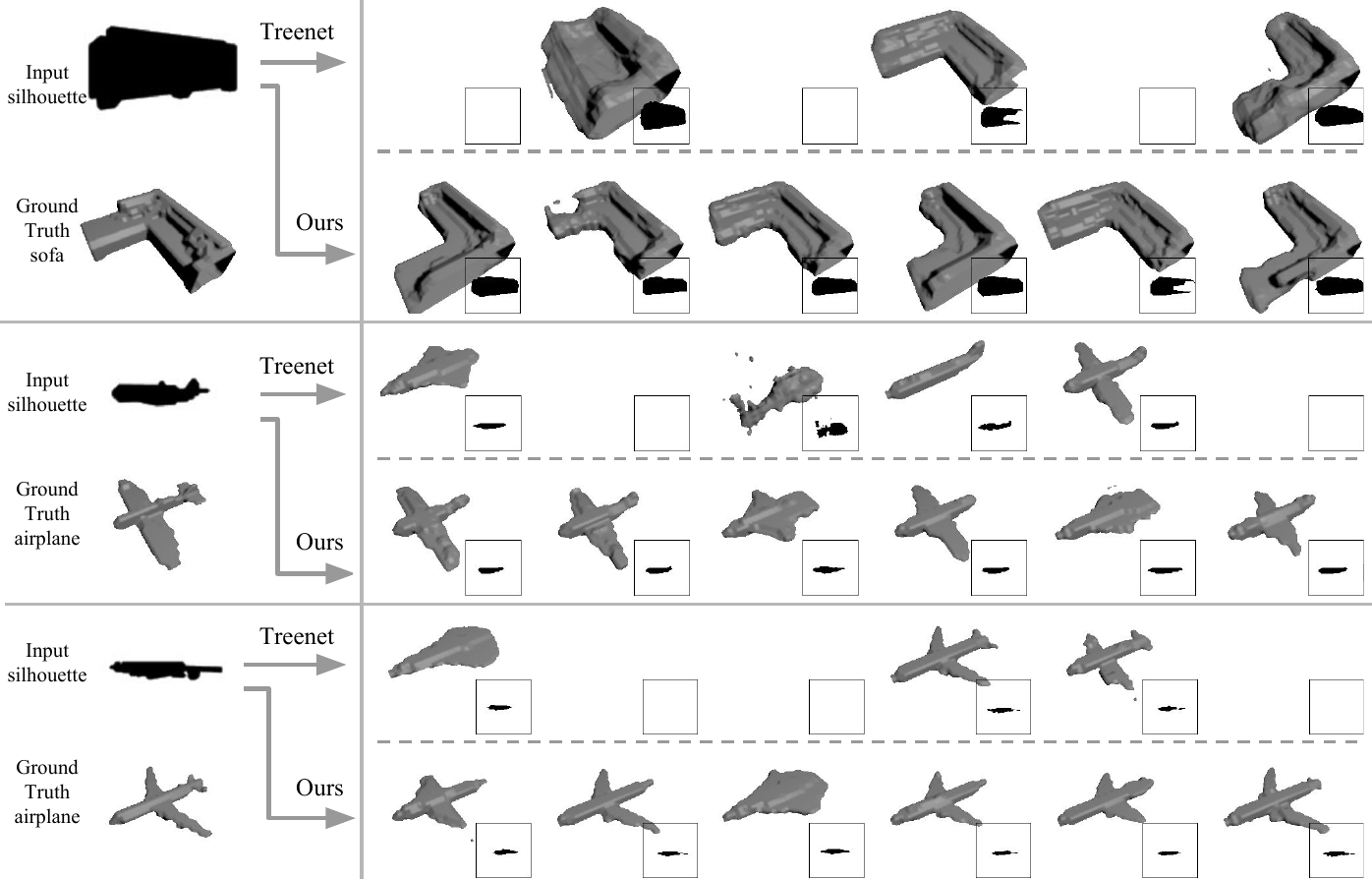}
    \caption{Results on the 2D to 3D reconstruction task. The reconstructed objects are quite diverse in appearance and reasonably plausible. Figure taken from~\cite{https://doi.org/10.48550/arxiv.2008.10634}.}
    \label{fig:reconstr_results}
\end{figure}

Finally, we consider the 2D to 3D reconstruction task, illustrated in Figure~\ref{fig:reconstr_results}. The input is a silhouette: a 2D projection with removed texture. The task is to reconstruct the 3D shape from the 2D silhouette, which we have seen to induce many ambiguities previously. The method produces a diverse set of possibilities that are all quite plausible.

\subsubsection{Mixture Density Networks}

The min-loss is also hidden in the work on \href{https://arxiv.org/abs/1906.03631}{improved mixture density networks}~\cite{Makansi_2019}. They apply the \emph{Winner-Takes-All (WTA)~\cite{guzman-rivera2012multiple} loss} (which is the same as our previous K-diverse loss) to their novel problem of future prediction:
\begin{align*}
\cL_\mathrm{WTA} &= \sum_{k = 1}^K w_k \ell(h_k, \hat{y})\\
w_i &= \bone\left(i = \argmin_k \Vert \mu_k - \hat{y} \Vert\right).
\end{align*}
where \(k\) is the expert (hypothesis) index, \(\mu_k\) is the prediction of expert \(k\), \(\hat{y}\) is the GT (!) location, and \(\ell\) is the loss function. The weight term is a Kronecker delta indicating the most correct solution out of the set of experts (i.e. when the predicted mean is closest to the GT location). The winner is the expert with minimal prediction distance. That expert is the only one trained on that sample; it takes the full penalty.

\begin{figure}
  \centering
  \resizebox{0.6\linewidth}{!}{%
  \setlength{\tabcolsep}{0.7pt}%
  \begin{tabular}{ccc}
      \includegraphics[width=0.2\textwidth, height=1.1in]{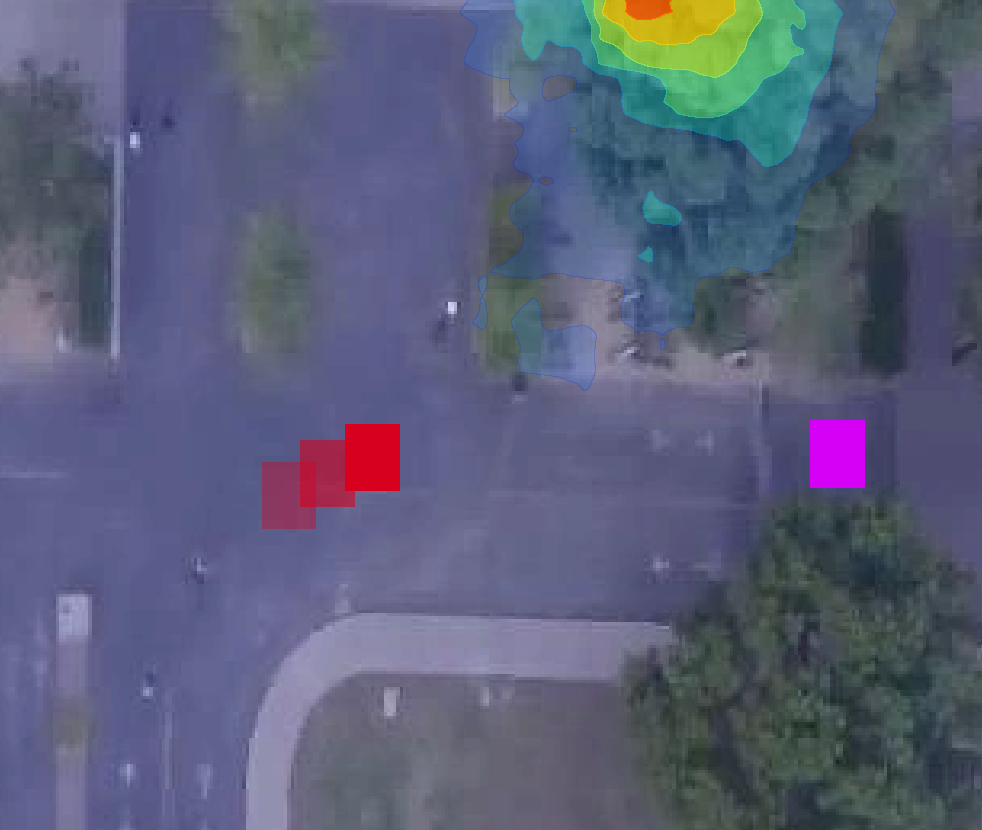}  & 
      \includegraphics[width=0.2\textwidth, height=1.1in]{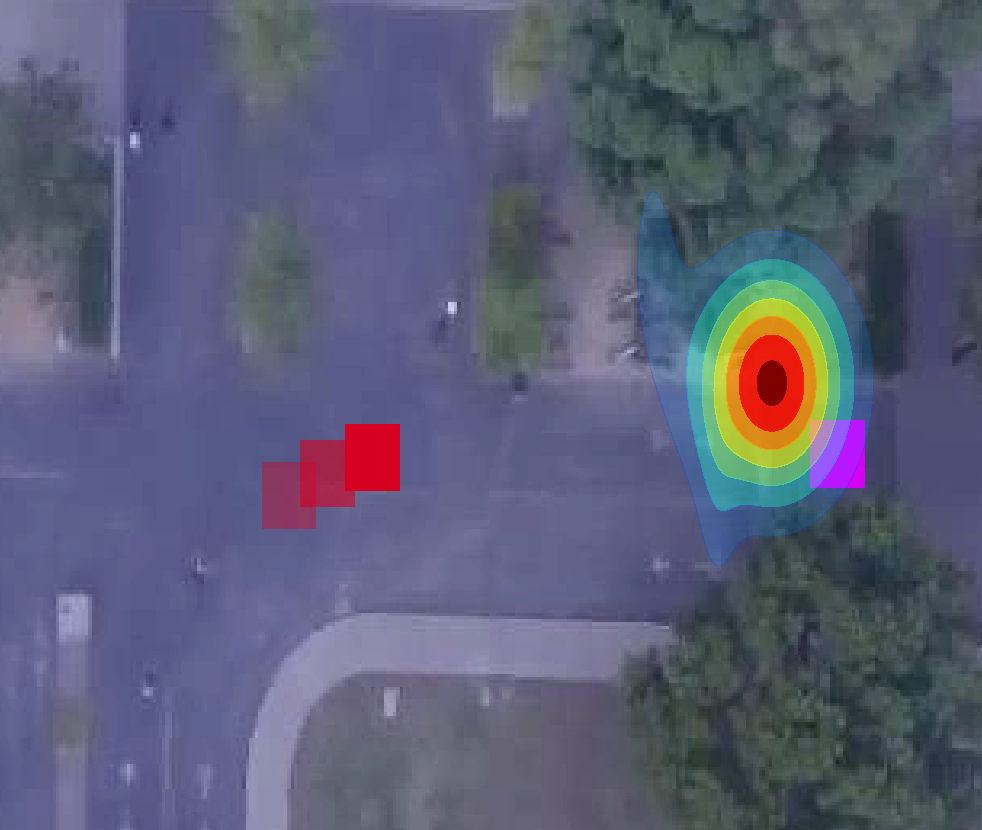}  &
      \includegraphics[width=0.2\textwidth, height=1.1in]{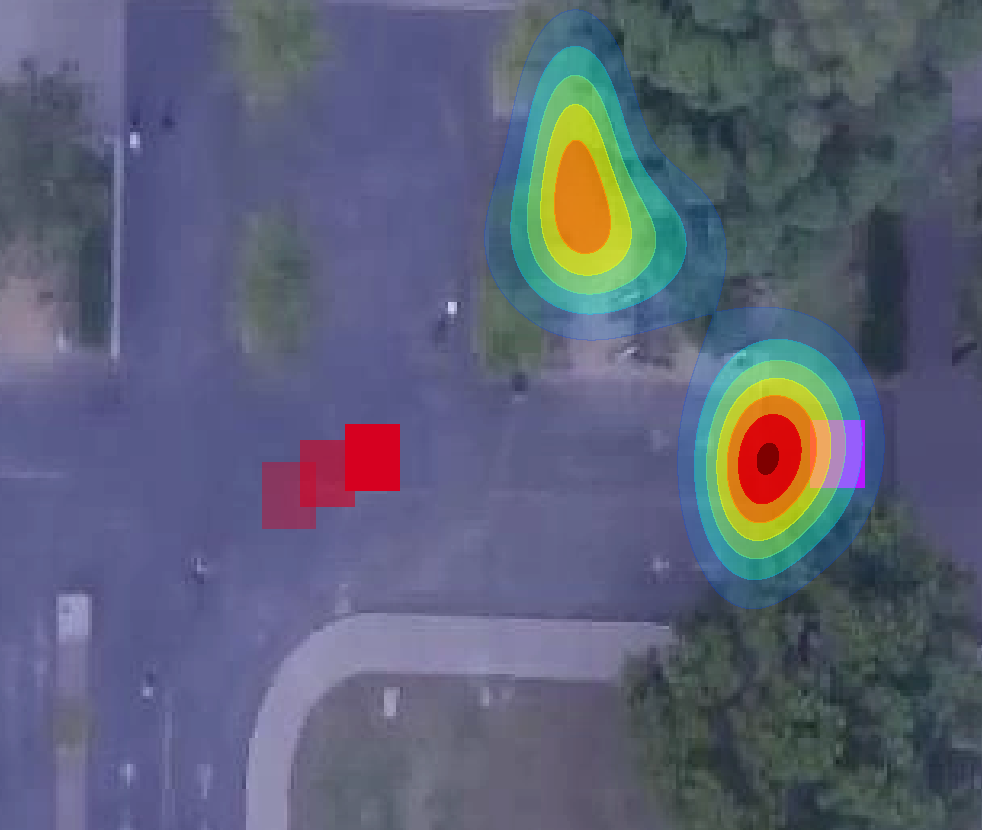}  
      \\    
      \includegraphics[width=0.2\textwidth, height=1.1in]{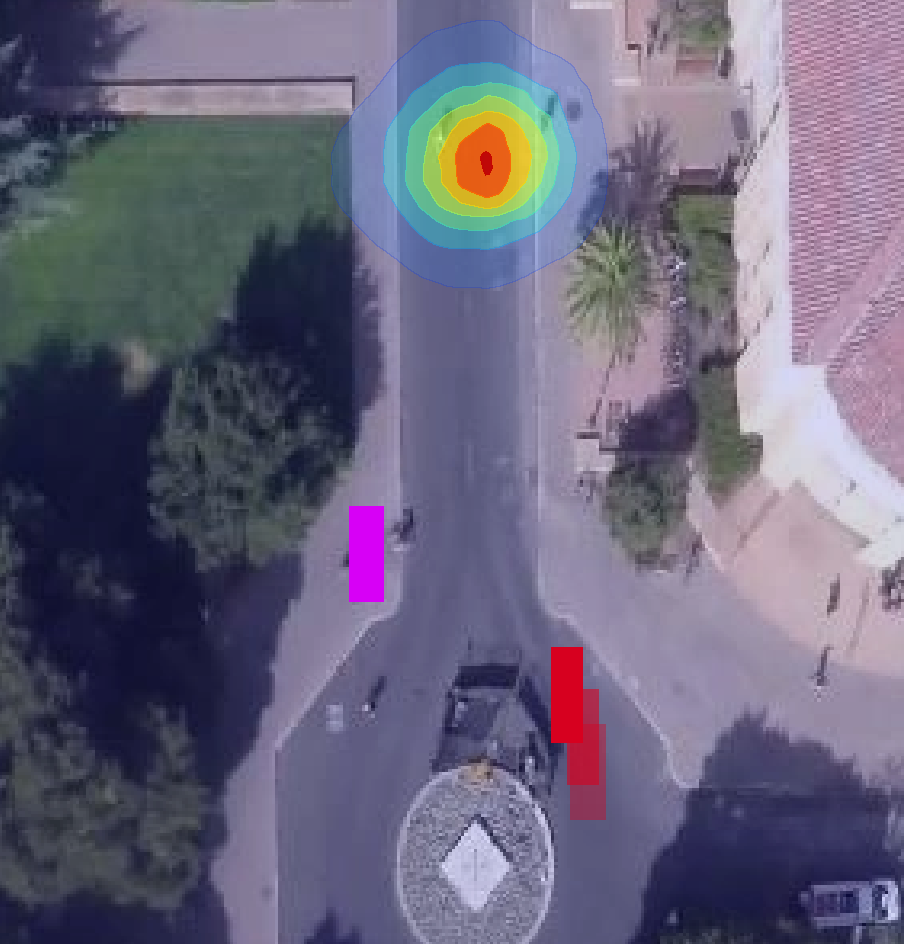}  & 
      \includegraphics[width=0.2\textwidth, height=1.1in]{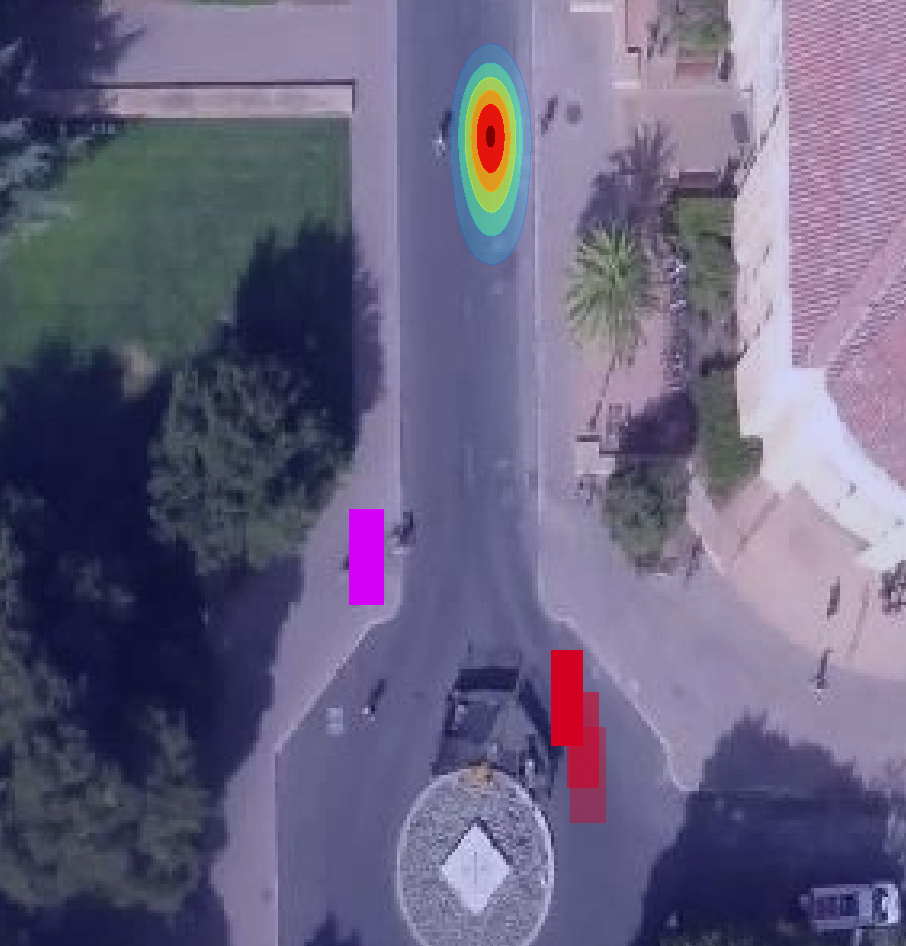}  &
      \includegraphics[width=0.2\textwidth, height=1.1in]{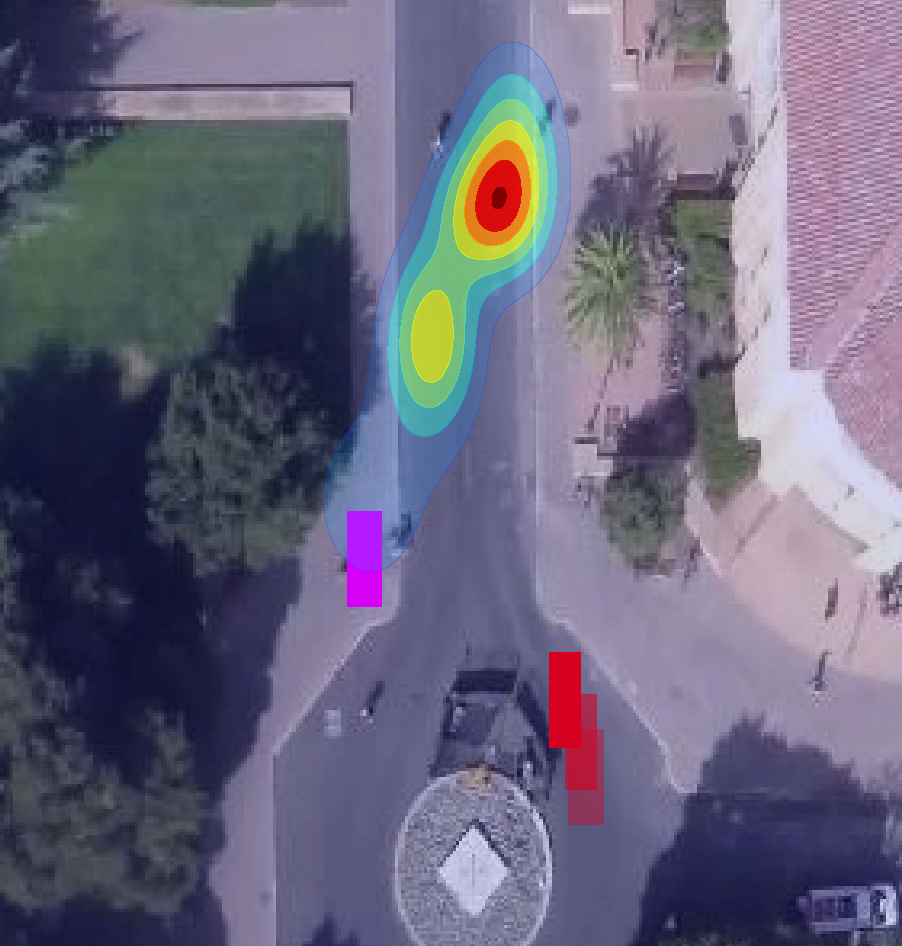}  
      \\
      \includegraphics[width=0.2\textwidth, height=1.1in]{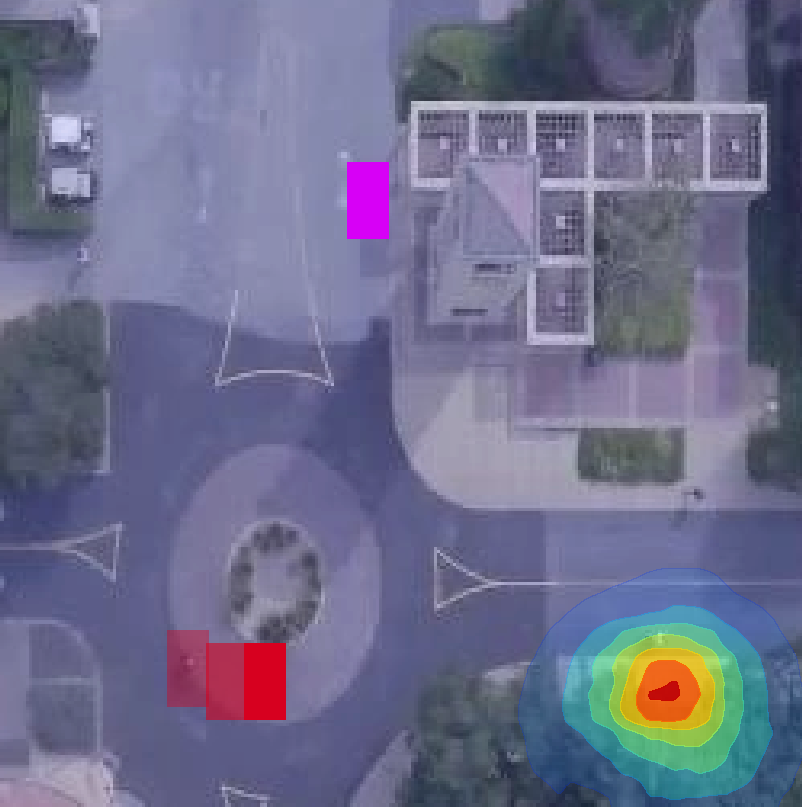}  & 
      \includegraphics[width=0.2\textwidth, height=1.1in]{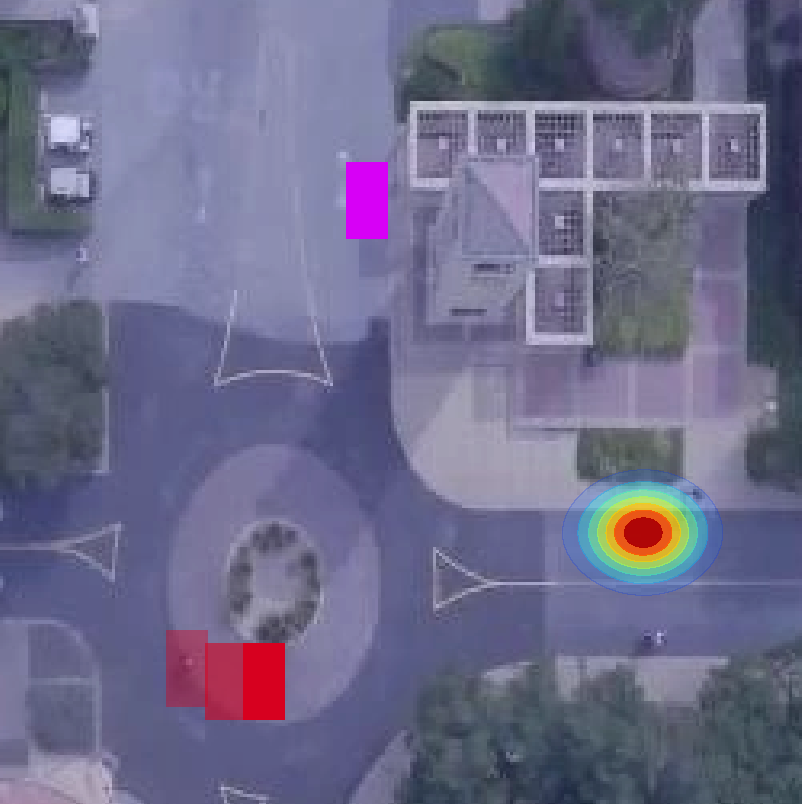}  &
      \includegraphics[width=0.2\textwidth, height=1.1in]{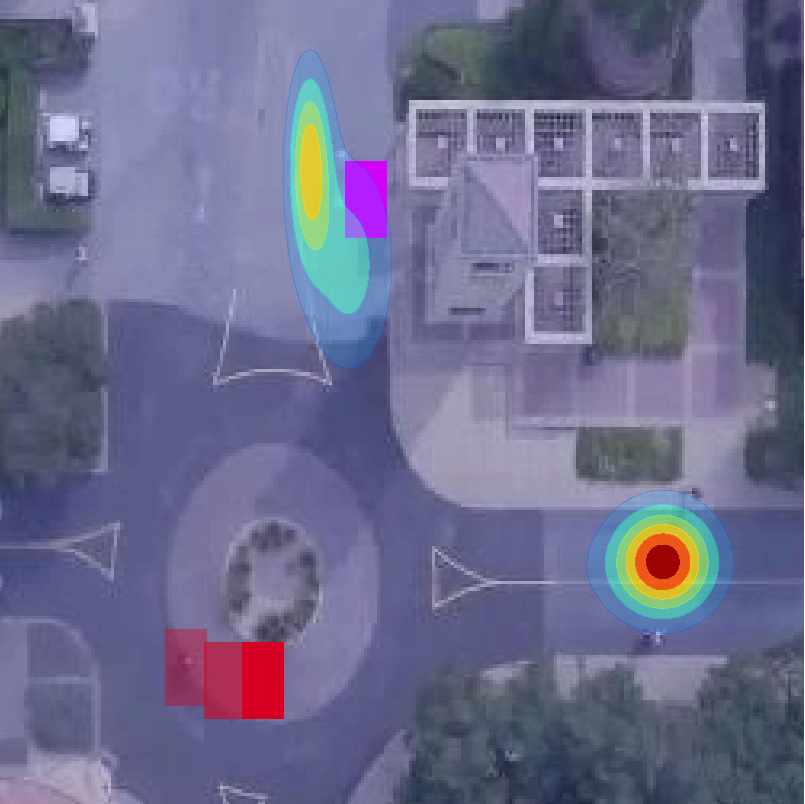}  
      \\
      \small{Non-Parametric} & \small{MDN} & \small{EWTAD-MDF} \\
\end{tabular}%
}%
   \caption{
   Qualitative evaluation of different probabilistic methods on future location prediction. Non-parametric and MDN are the baselines, and EWTAD-MDF is the proposed method. ``Given three past locations of the target object (red boxes), the task is to predict possible future locations. A heatmap overlay is used to show the predicted distribution over future locations, while the ground truth location is indicated with a magenta box. Both variants of the proposed method capture the multimodality better, while MDN and non-parametric methods reveal overfitting and mode-collapse.''~\cite{Makansi_2019} Figure taken from~\cite{Makansi_2019}.} 
    \label{fig:mdn}
\end{figure}

Let us discuss Figure~\ref{fig:mdn}. As shown by the authors, training with the WTA loss results in more diversified and more multimodal solutions. The red boxes correspond to the current and previous frame locations. The magenta boxes denote the future frame GT location. The isoprobability contours correspond to the predicted densities \(Q(y\mid x)\) over the future frame location by the methods. Ideally, the prediction should enclose the actual location and be diverse (multimodal). The method of~\cite{Makansi_2019} seems to be superior to the baselines. The min-loss is good for enforcing diversity in ill-posed tasks.

So far, we have considered the prediction of higher-dimensional outputs in regression tasks using DNNs while modeling aleatoric uncertainty. In the last section, we will see use cases of aleatoric uncertainty in representation learning.

\section{Aleatoric Uncertainty in Representation Learning}

Aleatoric uncertainty also has close ties to representation learning. Let us first consider what we mean by a representation.

\begin{definition}{Representation}
A representation is a mapping of input \(x\) to some space of latent variables \(z\). The semantic distance between two inputs (e.g., images), \(d(x_1, x_2)\), is encoded through more straightforward metrics in the latent space \(\cZ\). An example is the Euclidean distance:
\[d(x_1, x_2) = \Vert z_1 - z_2 \Vert_2.\]
This is possible in principle because the mapping from the input space to the space of latent variables can be highly non-linear, making it potentially expressive enough to extract semantic meaning through the embedding procedure.
\end{definition}

\subsection{Prelude: Representation Learning}
\label{sssec:representation_learning}

\begin{definition}{Retrieval Task}
The retrieval task in representation learning refers to returning inputs \(x\) from a dataset that are semantically similar to a test input \(x_\mathrm{test}\), as defined by the task. One can perform this efficiently if they possess a performant feature embedding \(f\), as they only have to compute the feature embeddings of the dataset samples once. Then they only have to search for k-NNs in the latent space. Finally, e.g., using a simple hash table, one can return the original dataset samples corresponding to the k-NN feature embeddings. An example of a system solving a retrieval task is \href{https://images.google.com/}{Google Images}.
\end{definition}

For learned representations, we can consider (1) the retrieval task detailed above, but also (2) the downstream fine-tuning of our feature embedding on a classification problem. These both become much more manageable and computationally more efficient if we already have a robust embedding.

For retrieval, instead of having to iterate through all images \(x_i\) in the dataset, one can index them by passing them through the feature embedding and saving the feature vectors. For comparing semantic distances, one just has to use cosine similarity or \(L_2\) distance in the latent space, which is very efficient to compute. Obtaining k-NNs can be done very efficiently nowadays, e.g., using \href{https://github.com/facebookresearch/faiss}{\texttt{faiss}}~\cite{johnson2019billion}.

\begin{figure}
    \centering
    \includegraphics[width=0.8\linewidth]{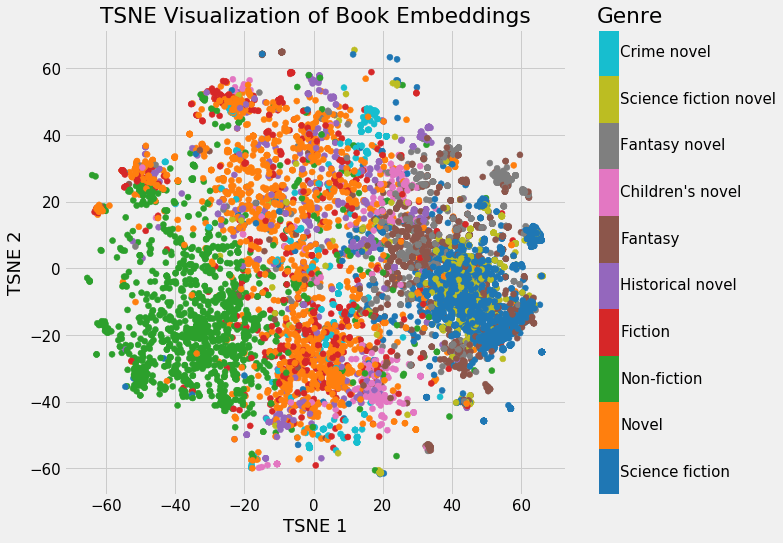}
    \caption{Illustration of t-SNE word embeddings for books. Books of the same genre are often clustered together, although no label supervision is used in the method. Figure taken from~\cite{devopedia}.}
    \label{fig:wordembedding}
\end{figure}

Suppose we are only interested in a low-dimensional, unsupervised, interpretable representation of our book dataset samples (Figure~\ref{fig:wordembedding}). In that case, we can obtain a 2D t-SNE representation of different books where similarities in the original space are aimed to be preserved in the embedding space. Instead of saving all characters in every book, we embed them in a 2D space where the (distance-based) similarities are as preserved as possible. The original samples do not have to be saved. As can be seen, this representation gives a coarse clustering \wrt the genre, even though the method did not have access to genre labels. Thus, the 2D location encodes the genre of the book to some extent.

\subsubsection{Training a representation}

Our goal is to estimate a deterministic mapping \(z = f(x)\) (construct the latent space) such that the simple distance metric (e.g., \(L_2\) distance or cosine similarity) in the latent space reflects semantic distances (class membership) of the corresponding inputs. The ingredients (i.e., the training data) usually come in two forms.
\begin{itemize}
    \item \textbf{Pairs.} \((x_1, x_2, y = 1)\) when \(x_1\), \(x_2\) are from the same category or \((x_1, x_2, y = 0)\) when they are not. 
    \item \textbf{Triplets.} \((x_a, x_p, x_n)\), where \(x_a\) is the anchor image, \(x_p\) is a positive match to it (should be mapped closer), and \(x_n\) is a negative match to it (should be mapped farther). Here we do not have to consider categories at all (discrete); we just have to specify relative similarities (relative distances in the latent space). As we are only supervising with relative distances to train the embedding, all we know is that \(x_a\) and \(x_p\) should be closer in the latent space than \(x_a\), \(x_n\). This can, however, encode more subtle relationships than the previous supervision technique: we cannot only specify which samples belong to the same group, but we can also specify which groups are more similar and which are less, and which samples are more similar in the same group.\footnote{This is a somewhat idealistic description of triplet supervision where the negative mining is proportional to the semantic similarity. In practice, people consider two extremes: either using a classification dataset to provide positive/negative samples (very coarse) or using self-supervised learning where only the image itself is considered positive to itself (too fine-grained). This still works in practice, though.}
\end{itemize}
In either case, we only get a very coarse description of where each sample should be mapped in the latent space. We have to construct the latent space where, e.g., the Euclidean distance should encode semantic similarity using this very coarse training signal. The latent embedding \(z\) is never observed (not part of supervision); it has to be inferred.

\subsection{Aleatoric Uncertainty in Representation (Learning)}

\subsubsection{Aleatoric uncertainty can be present in representation learning.}

In a typical embedding space of 2-digit MNIST images, very clear samples of classes 17 and 47 with no ambiguity are mapped to two quite different points in the embedding space. When we observe a new sample of class 17 with no ambiguity, it should be ideally mapped close to the previous 17 sample (maybe not very close if the handwriting is quite different). It is quite hard to confuse 17 with 47, so even if the handwriting differs between the 17 samples, they should still be reasonably far away from 47. The same should happen with a new sample of class 47 with no ambiguity, just with the classes flipped in the previous scenario.

Now, consider a sample where a crucial part of the first digit is deleted. Both 17 and 47 are plausible true classes, i.e., we have aleatoric uncertainty. It is now unclear where this image should be put in the latent embedding space. Putting it close to 47 makes sense; this could have been a 47. However, putting it close to 17 makes sense as well for the same reason. Putting it in the middle is not really a good solution: it does not signal closeness to either of the two plausible classes, even though it is plausibly very close to \emph{both} of them. This is impossible to communicate with deterministic embeddings. Whenever the inverse generative process \(P(Z \mid X)\) is noisy, deterministic embeddings cannot faithfully represent the true relationship of an embedding and an image.\footnote{\(P(X \mid Z)\) might even be deterministic, but \(P(Z \mid X)\) can still be stochastic, and vice versa.}

We argue that a good solution is not to consider point embeddings (deterministic functions of \(x\)) but rather an entire distribution over latent variables given an input \(x\) whose distribution is multimodal (e.g., a MoG). Such a probabilistic embedding can put two different components in two different parts of the space if our latent is uncertain (torn between two categories). For, e.g., a Gaussian noise image, we would ideally get a probabilistic embedding spread across the space, not just torn between two distributions. This represents what the underlying latent generator of \(X\) -- i.e., the latent variable \(Z\) -- \emph{could} have plausibly been, whereas a point embedding in the middle of plausible embedding regions represents what latent code the noisy image was \emph{surely} generated from.\footnote{This does not model the noise in the inverse generative process. Rather, it tries to assign noisy images to single points in the embedding space as well, which is not a reasonable probabilistic interpretation.}

\begin{figure}
    \centering
    \includegraphics[width=0.8\linewidth]{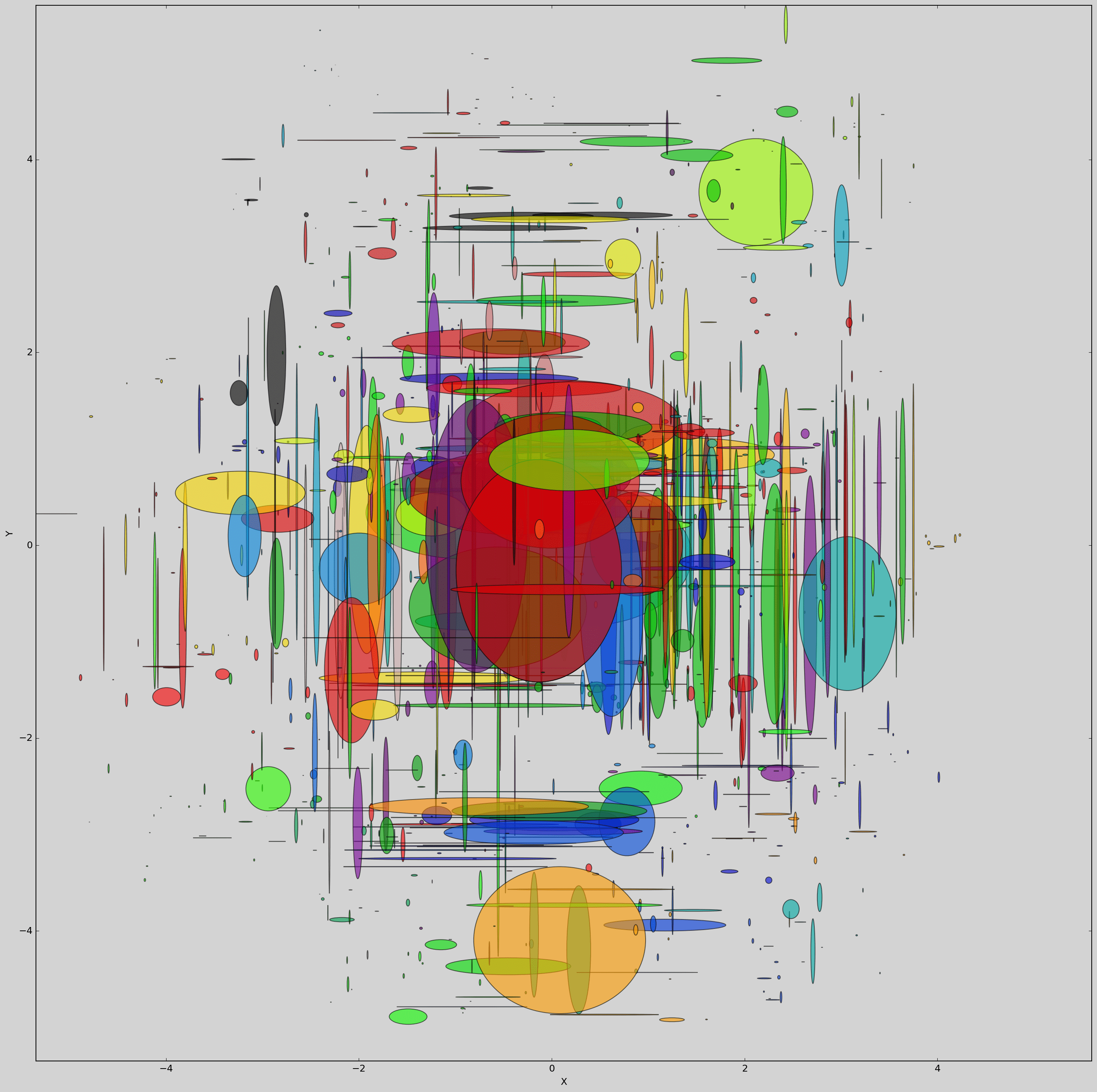}
    \caption{Estimated latent heteroscedastic Gaussian distributions \(P(Z \mid X)\). Figure taken from~\cite{https://doi.org/10.48550/arxiv.1810.00319}.}
    \label{fig:gaussianemb}
\end{figure}

\subsubsection{Problem statement for representation of aleatoric uncertainty in representation learning}

We discuss ``\href{https://arxiv.org/abs/1810.00319}{Modeling Uncertainty with Hedged Instance Embedding}''~\cite{https://doi.org/10.48550/arxiv.1810.00319}. The goal is to estimate latent distributions \(P(Z \mid X = x)\). \(Z\) is a latent variable here; it could be, e.g., estimated by a heteroscedastic diagonal Gaussian. Then, probabilistic embeddings look like in Figure~\ref{fig:gaussianemb}. These embeddings are 2D Gaussians for corrupted test images from 2-digit MNIST. It is desired that corrupted test images have embeddings that spread density across larger regions of the space than clean images.

The ingredients (training data) of the method are pairs: \((x_1, x_2, y = 1)\) when \(x_1\), \(x_2\) are from the same category or \((x_1, x_2, y = 0)\) when they are not. Note again that \(Z\) is not observed -- it is a latent variable.

Estimating \(P(Z \mid X = x)\) is, thus, even more challenging than estimating \(P(Y \mid X = x)\) with complete observations (the typical aleatoric uncertainty case, where \((x, y)\) pairs are available in a large dataset). Here we do not have any observations of \(Z\). We have to learn good representations from noisy observations and a lack of observations of the true latents.

How to handle representation learning with aleatoric uncertainty? Just as in deterministic representation learning, we will compute a quantity that expresses how likely \(x_1\) and \(x_2\) are to be similar. The catch is to introduce a latent distribution \(Q(Z \mid X = x)\) in the middle for each input \(x\). This estimates the noisy inverse generative process $P(Z \mid X = x)$. We then marginalize over this distribution, i.e., compute the similarity for all possible latents the image may represent:
\begin{align*}
Q(Y = 1 \mid x_1, x_2) &= \int Q(Y = 1, z_1, z_2 \mid x_1, x_2)\ dz_1dz_2\\
&= \int Q(z_1, z_2 \mid x_1, x_2)Q(Y = 1 \mid z_1, z_2, x_1, x_2)\ dz_1dz_2\\
&= \int Q(z_1 \mid x_1, x_2)Q(z_2 \mid x_1, x_2)Q(Y = 1 \mid z_1, z_2)\ dz_1dz_2\\
&= \int Q(z_1 \mid x_1)Q(z_2 \mid x_2)Q(Y = 1 \mid z_1, z_2)\ dz_1dz_2
\end{align*}
where we used the following assumptions in order:
\begin{align*}
Z_1 &\indep Z_2 \mid X_1, X_2\\
Y &\indep X_1 \mid Z_1, Z_2\\
Y &\indep X_2 \mid Z_1, Z_2\\
Z_1 &\indep X_2 \mid X_1\\
Z_2 &\indep X_1 \mid X_2.
\end{align*}
We marginalize over the predicted posteriors. \(Q(Y = 1 \mid z_1, z_2)\) is a distance-based probability. The distance of \(z_1\) and \(z_2\) should tell us whether they correspond to the same category.

To make this DNN-friendly (computational-graph-wise), we model the probabilistic embeddings by parameterizing heteroscedastic, isotropic Gaussians:
\[Q(z \mid x) = \cN(z \mid \mu(x), \sigma^2(x)I)\]
where \(\mu(\cdot)\) and \(\sigma^2(\cdot)\) are DNNs. (We predict the parameters of the latent Gaussian from the input \(X\).) We parameterize the match probability
\[Q(Y = 1 \mid z_1, z_2) = \mathrm{sigmoid}\left(-a \Vert z_1 - z_2 \Vert_2 + b\right) \in (0, 1)\]
with some smooth calibration function (sigmoid of affine transformation) with learnable parameters \(a > 0\) and \(b\). We want to keep a notion of Euclidean distance in the latent space. Therefore, we let the match probability depend on the Euclidean distance in some way. For example, \(b = 0\) is not a good choice: In this case, distance 0 means 0.5 probability in the match. Thus, the simple sigmoid is not well-calibrated to the 0-1 range. \(a > 0\) is required to ensure that \(a\) is not flipping the sign of distance. We want a closer distance to encode more similarity; therefore, we need \(-a\) to be negative. The exact linear scaling and translation have to be calibrated.

To make the formulation even more DNN-friendly, we perform a Monte Carlo sampling to approximate the integral
\begin{align*}
Q(Y = 1 \mid x_1, x_2) &= \int Q(z_1 \mid x_1)Q(z_2 \mid x_2)Q(Y = 1 \mid z_1, z_2)\ dz_1dz_2\\
&= \nE_{Z_1 \sim Q(Z_1 \mid X_1 = x_1), Z_2 \sim Q(Z_2 \mid X_2 = x_2)}\left[Q(Y = 1 \mid z_1, z_2)\right]\\
&\overset{\mathrm{MC}}{\approx} \frac{1}{K^2} \sum_{k_1, k_2} Q(Y = 1 \mid z_1^{(k_1)}, z_2^{(k_2)})
\end{align*}
where \(z_i^{(k_i)} \sim Q(z_i \mid x_i) = \cN\left(z_i \mid \mu(x_i), \sigma^2(x_i)I\right), i \in \{1, 2\}\). We sum over the match probability values for every possible pair of samples from each distribution. With that, we obtain the final match probability estimate.

We differentiate through the sampling back to the network parameters via the reparameterization trick:
\begin{enumerate}
    \item We sample \(\epsilon_i^{(k_i)} \sim \cN(0, I)\).
    \item We set \(z_i^{(k_i)} = \sigma(x_i) \cdot \epsilon_i^{(k_i)} + \mu(x_i)\).
\end{enumerate}
This way, sampling is separated from backpropagation.

Finally, the NLL loss is minimized \wrt the parameters of our neural network (MLE):
\begin{align*}
\cL(x, y) &:=\begin{multlined}[t]-y\log\left(\frac{1}{K^2} \sum_{k_1, k_2} Q(Y = 1 \mid z_1^{(k_1)}, z_2^{(k_2)})\right)\\ - (1 - y)\log\left(1 - \frac{1}{K^2} \sum_{k_1, k_2} Q(Y = 1 \mid z_1^{(k_1)}, z_2^{(k_2)})\right)\end{multlined}\\
&= -\log\left(\frac{1}{K^2} \sum_{k_1, k_2} Q(Y = y \mid z_1^{(k_1)}, z_2^{(k_2)})\right)
\end{align*}
where \(y \in \{0, 1\}\) is the GT label for the \((x_1, x_2)\) pair and \(z_i^{(k_i)} \sim Q(z_i \mid x_i) = \cN\left(z_i \mid \mu(x_i), \sigma^2(x_i)I\right)\).

Although the NLL was a proper scoring rule for many of the previous problems, we now have no guarantee of propriety. It is not guaranteed that we recover the true latent distribution, not even if the true distribution is in our parametric family \(\cQ\). This is because, unlike previously, where we said that if we use log-prob that guarantees the recovery of the true distribution in expectation, now we are talking about latent variables, for which we do not have many guarantees (so far). But we will come back to this in Section~\ref{ssec:guarantee}.

\subsubsection{Overview of the method}

\begin{figure}
    \centering
    \includegraphics[width=\linewidth]{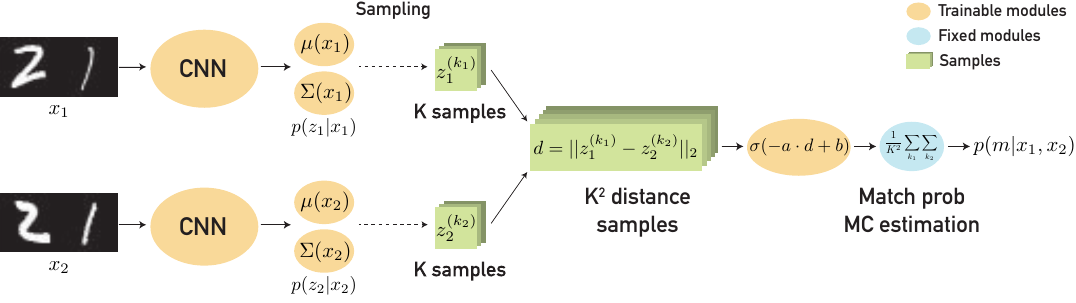}
    \caption{Overview of~\cite{https://doi.org/10.48550/arxiv.1810.00319}, a method for representing aleatoric uncertainty in learned embeddings. CNNs predict the mean and covariance corresponding to the latent posterior conditioned on the input. These are used to calculate semantic distances between the two inputs that are further transformed into a match probability. Figure taken from~\cite{https://doi.org/10.48550/arxiv.1810.00319}.}
    \label{fig:overview}
\end{figure}

An overview of the method is shown in Figure~\ref{fig:overview}. Here, a CNN outputs the posterior parameters. Sampling is differentiable via the reparameterization trick. The learned sigmoid gives us the match probabilities: we average over \(K^2\) of them. Finally, we take the negative log of the final match probability \wrt the true label \(y \in \{0, 1\}\).

\subsubsection{Results on [N \(\in\) \{2, 3\}]-digit MNIST}

\begin{figure}
    \centering
    \includegraphics[width=0.5\linewidth]{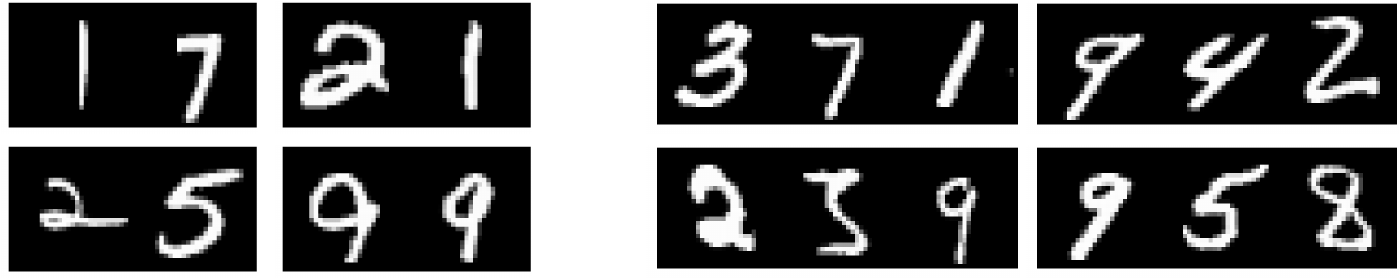}
    \caption{Samples from [N \(\in\) \{2, 3\}]-digit MNIST. Digits from the original MNIST dataset are concatenated together.}
    \label{fig:ndigit}
\end{figure}

Samples from [N \(\in\) \{2, 3\}]-digit MNIST are shown in Figure~\ref{fig:ndigit}. Statistics of the used datasets are shown in Table~\ref{tab:ndigit}. We perform data corruption which introduces more aleatoric uncertainty. The clean inputs are original MNIST samples; corrupted inputs contain randomly occluded parts. This leads to ambiguities and variance in \(P(y \mid x)\).

\begin{table}
	\footnotesize
	\centering
	\caption{Statistics of [N \(\in\) \{2, 3\}]-digit MNIST. Table taken from~\cite{https://doi.org/10.48550/arxiv.1810.00319}.
	}
	\label{tab:ndigit}
	\begin{tabular}{*{8}{c}}\toprule
		\vspace{-1em} \\
		Number 						&& Total & Training & Unseen Test 	& Seen Test & Training&Test \\
		Digits						&& Classes & Classes& Classes	& Classes & Images & Images\\
		\vspace{-1em} \\
		\cline{1-1} \cline{3-8}
		\vspace{-1em} \\
		2 && 100 & 70 & 30& 70 & 100\,000& 10\,000 \\  
		3 && 1000 & 700 & 100& 100 & 100\,000& 10\,000 \\  
\bottomrule
	\end{tabular}
\end{table}	

\begin{figure}
    \centering
    \includegraphics[width=\linewidth]{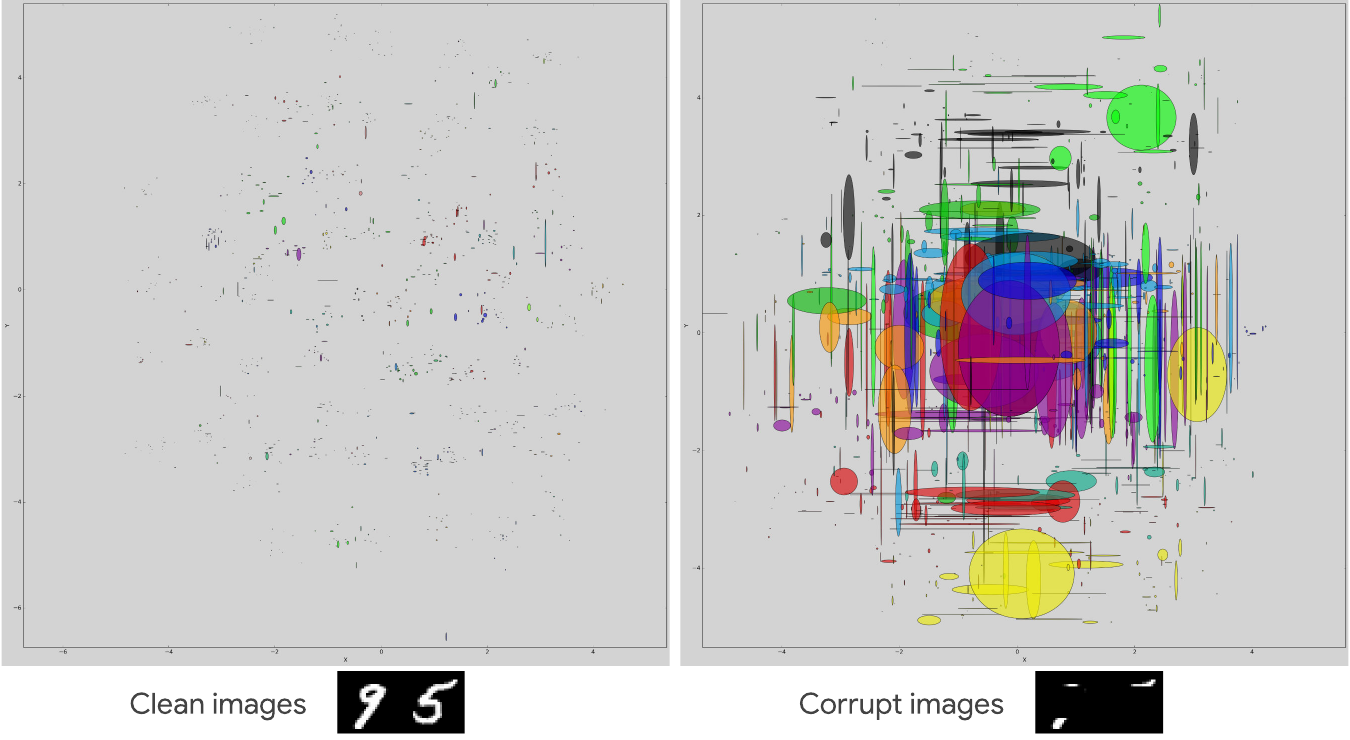}
    \caption{Comparison of 2D embedding distributions of clean (left) and corrupted (right) images. As expected, corrupted images also have a latent posterior with significantly more spread. Base figure taken from~\cite{https://doi.org/10.48550/arxiv.1810.00319}.}
    \label{fig:2dhedged}
\end{figure}

In Figure~\ref{fig:2dhedged}, we compare the embedding distributions of clean and corrupt images from 2-digit MNIST. 2D Hedged Instance Embeddings (MoGs with a single component) are used. Although there was no guarantee regarding the correct recovery of the true latent posterior, we do see a high correlation between the image's cleanliness and the embeddings' variance. For clean images, we have barely any variance in the embeddings. For corrupted images, however, we see larger variances in the latent posteriors. Here, we do not have an isotropic Gaussian posterior family, which makes the method more expressive.

\begin{figure}
    \centering
    \includegraphics[width=\linewidth]{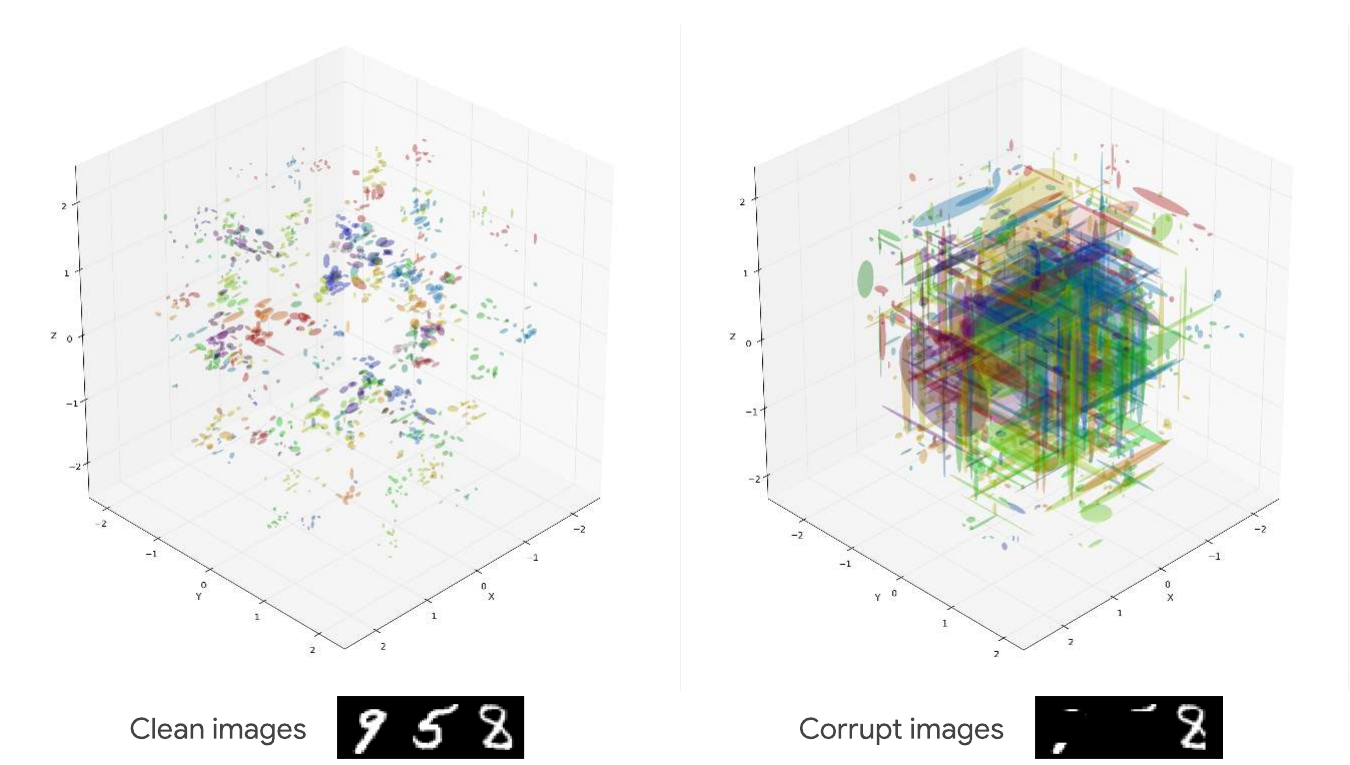}
    \caption{Comparison of 3D embedding distributions of clean and corrupted images. Corrupted images have latent posteriors with more spread. Base figures taken from~\cite{https://doi.org/10.48550/arxiv.1810.00319}.}
    \label{fig:3dhedged}
\end{figure}

In Figure~\ref{fig:3dhedged}, we turn to the 3-digit MNIST dataset and use 3D Hedged Instance Embeddings. We observe the same trend as in Figure~\ref{fig:3dhedged}, supporting that these observations are quite general when training with this method.

\begin{figure}
    \centering
    \includegraphics[width=0.5\linewidth]{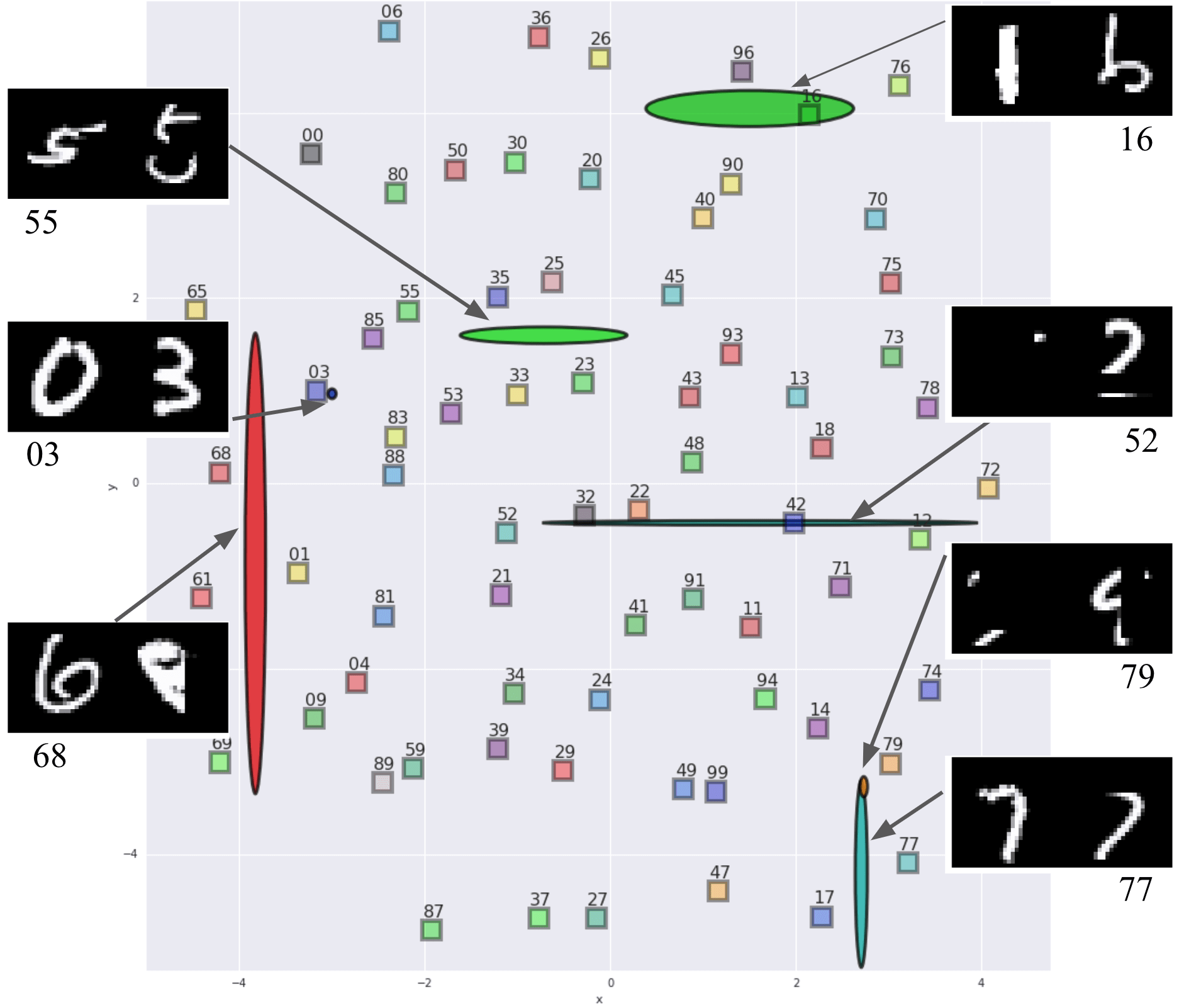}
    \caption{Visualization of 2D embedding distributions of particular clean and corrupt images. The distributions are aligned across plausible numbers (reconstructions). Figure taken from~\cite{https://doi.org/10.48550/arxiv.1810.00319}.}
    \label{fig:2dhedged2}
\end{figure}

Finally, let us discuss Figure~\ref{fig:2dhedged2}, a visualization of the embedding distributions of particular clean and corrupt images. The squares correspond to the centroids of each class in the 2D embedding space. (Average mean of the posterior for the same class using clean samples.) For the clean 03 image, the latent posterior is a very narrow Gaussian close to the centroid of class 03. For the corrupted `68' image, the Gaussian has a lot more variance. For the second digit, the network is uncertain whether it is 8, 1, or 9. It hedges its bets across the three class centroids corresponding to 61, 68, and 69. For the corrupted `52' image, the Gaussian is, again, broad. This is because nearly the entire 5 is blocked out. The model finds 52, 32, 22, 42, 12, and 72 all plausible. Humans would also find these plausible. We also see that the neighboring centroids are well structured, allowing the model to hedge its bets across the plausible labels well (as the model was also \emph{trained} with corrupted samples). In general, similar images are embedded in neighboring 2D regions.

The variance of the Gaussians not only encodes aleatoric uncertainty but also shows us a good set of plausible classes the sample could belong to (according to human inspection). It kind of recovers the true \(P(z \mid x)\) in that sense, as it empirically recovers all latent variations. Still, we have no guarantees.

\textbf{Note}: The latent dimensionality does not have to correspond to the number of digits. However, this way, we get interpretable embeddings (but nothing forces the embeddings to be axis-aligned \wrt the digits). 

\subsection{Multimodal Joint Representations and Aleatoric Uncertainty}

\begin{figure}
    \centering
    \includegraphics[width=0.8\linewidth]{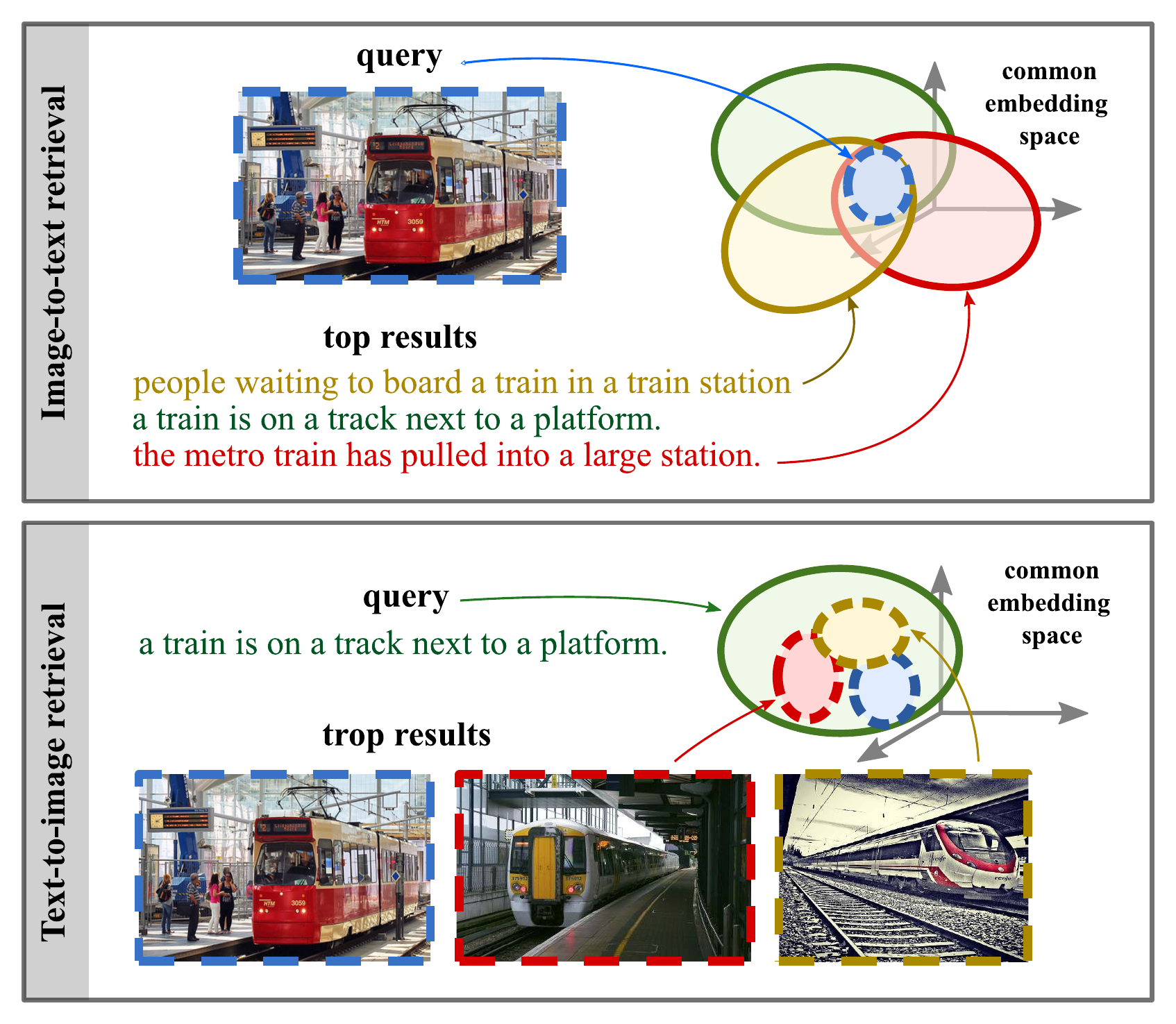}
    \caption{Visualization of the joint embedding space of images and texts, and their inherent uncertainty. Many images correspond to the same query, but also many queries correspond to the same image. Figure taken from~\cite{https://doi.org/10.48550/arxiv.2101.05068}.}
    \label{fig:multimodal}
\end{figure}

The work on \href{https://arxiv.org/abs/2101.05068}{aleatoric uncertainty in multimodal joint representations}~\cite{https://doi.org/10.48550/arxiv.2101.05068} extends the previous work to more scenarios. The idea was that aleatoric uncertainty could also naturally arise in multimodal joint embeddings. An overview of text-to-image and image-to-text retrieval is given in Figure~\ref{fig:multimodal}. The common embedding space accommodates both image and text embeddings. Given an image, we might be interested in what captions it (likely) corresponds to. However, given the image, we have multiple possibilities for the captions that correctly describe the image. Probabilistic embeddings, therefore, make more sense than point embeddings for multimodality. This also holds the other way around: Multiple different images make sense for the same query. To model this multiplicity, one can embed every image and caption into a Gaussian in the embedding space.

\begin{figure}
    \centering
    \includegraphics[width=\linewidth]{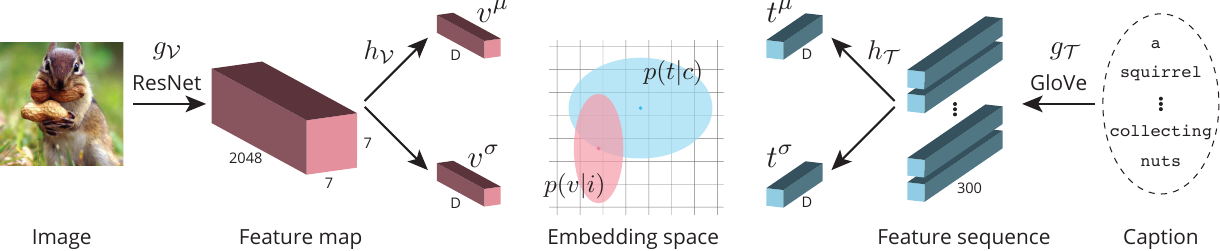}
    \caption{High-level overview of representing aleatoric uncertainty in multimodal joint embeddings. The method is quite similar to Hedged Instance Embeddings~\cite{https://doi.org/10.48550/arxiv.1810.00319}. Figure taken from~\cite{https://doi.org/10.48550/arxiv.2101.05068}.}
    \label{fig:multimodal2}
\end{figure}

We can train probabilistic representations using the same loss as before, as illustrated in Figure~\ref{fig:multimodal2}. We sample \(K\) times from each Gaussian, measure Euclidean distances, perform a calibration, use MC integration, and use NLL on top. This trains reasonably well; the authors also ended up with nice embeddings for these joint embeddings.

\subsection{Theoretical Guarantees for Recovering \(P(Z \mid X)\)}
\label{ssec:guarantee}

\begin{figure}
    \centering
    \includegraphics[width=\linewidth]{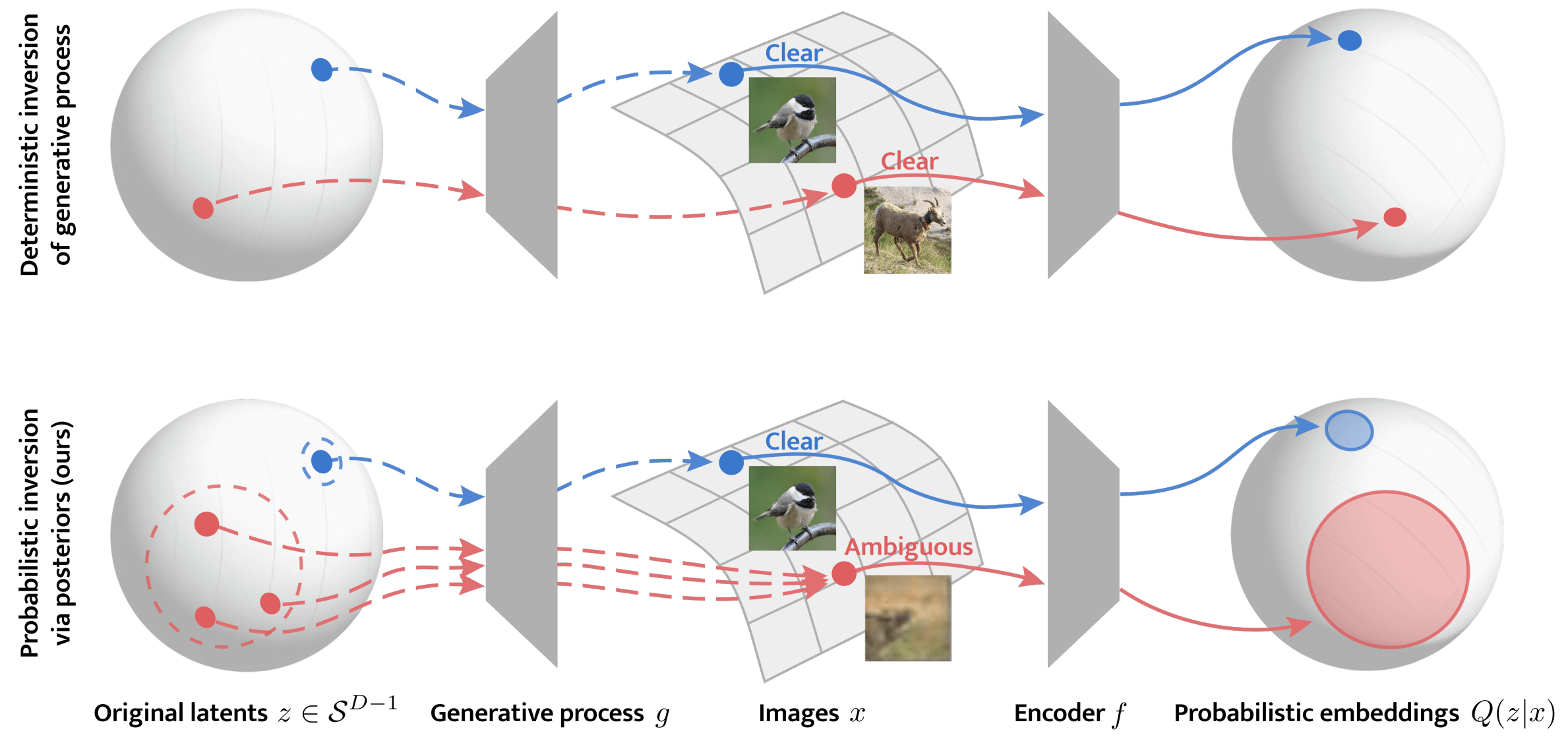}
    \caption{Overview of~\cite{https://doi.org/10.48550/arxiv.2302.02865}.
        ``Deterministic encoders embed images to points in the latent space. This recovers the latent vectors that generated them, up to a
            rotation (top). However, if an image is ambiguous, multiple possible latents could have generated it (bottom). An encoder
            trained with MCInfoNCE correctly recovers this posterior of the generative process, up to a rotation, from contrastive supervision.''~\cite{https://doi.org/10.48550/arxiv.2302.02865}
        Figure taken from~\cite{https://doi.org/10.48550/arxiv.2302.02865}.}
    \label{fig:guarantee1}
\end{figure}

Finally, let us discuss some theoretical guarantees for recovering the true latent posterior (up to a rotation and flip) that we hinted at previously. This is a work of Kirchhof \etal on \href{https://arxiv.org/abs/2302.02865}{theoretical guarantees for recovering \(P(Z \mid X = x)\)}~\cite{https://doi.org/10.48550/arxiv.2302.02865}. An overview of the method is given in Figure~\ref{fig:guarantee1}. Here, we only observe images. Our job is to recover the true latent posteriors \(P(Z \mid X = x)\) by only observing which pairs of images are positive and which are not. We assume that images are coming from the true latent space (\(P(X \mid Z = z)\), generation).

The used loss function is a probabilistic version of a representation learning objective (InfoNCE), called MCInfoNCE (where MC stands for Monte Carlo):
\begin{align*}
    \mathcal{L} :=& \hspace{-0.34cm} \mathop{\mathbb{E}}\limits_{\substack{x \sim P(x) \\ x^+ \sim P(x^+|x) \\ x^-_m \sim P(x^-), m=1, \dotsc, M}} \hspace{-1cm} \left( L_f\left(x, x^+, \{x_m^-\}_{m=1, \dotsc, M}\right)\right) \hspace{-0.5mm}\text{, with} \\
    L_f := & -\log \hspace{-1.4cm} \mathop{\mathbb{E}}\limits_{\substack{z \sim Q(z|x) \\ z^+ \sim Q(z^+|x^+) \\ z^-_m \sim Q(z^-_m|x^-_m), m=1, \dotsc, M}} \hspace{-0.8cm} \left( \frac{e^{\kappa_\mathrm{pos} z^\top z^+}}{\frac{1}{M} e^{\kappa_\mathrm{pos} z^\top z^+} + \frac{1}{M} \sum\limits_{m=1}^M e^{\kappa_\mathrm{pos} z^\top z^-_m}} \right)\hspace{-0.5mm}. \label{form:loss}
\end{align*}
This is the loss function we should use to recover the true latent posterior. Of course, there are some assumptions to guarantee the recovery, e.g., the true posterior has to be a von Mises-Fisher distribution, we are sampling from some uniformity, and the positives are determined by the distance from the point.

The theorem that gives the guarantee for the recovery of the true posterior (under some assumptions) is as follows.
\begin{theorem}[$\mathcal{L}$ identifies the correct posteriors] \label{eq:final}
Let $\mathcal{Z} = \mathcal{S}^{D-1}$ and $P(z) = \int P(z|x) dP(x)$ and $\int Q(z|x) dP(x)$ be the uniform distribution over $\mathcal{Z}$. Let $g$ be a probabilistic generative process [...]. Let $g$ have vMF posteriors $P(z|x) = \operatorname{vMF}(z;\mu(x), \kappa(x))$ with $\mu: \mathcal{X} \rightarrow \mathcal{S}^{D-1}$ and $\kappa: \mathcal{X} \rightarrow \mathbb{R}_{> 0}$. Let an encoder $f(x)$ parametrize vMF distributions $\operatorname{vMF}(z; \hat{\mu}(x), \hat{\kappa}(x))$. Then $f^* = \arg\min_f \lim_{M \rightarrow \infty} \mathcal{L}$ has the correct posteriors up to a rotation of $\mathcal{Z}$, i.e., $\hat{\mu}(x) = R \mu(x)$ and $\hat{\kappa}(x) = \kappa(x)$, where $R$ is an orthogonal rotation matrix, $\forall x \in \{x \in \mathcal{X} | P(x) > 0\}$.
\end{theorem}

\begin{figure}
    \centering
    \includegraphics[width=0.8\linewidth]{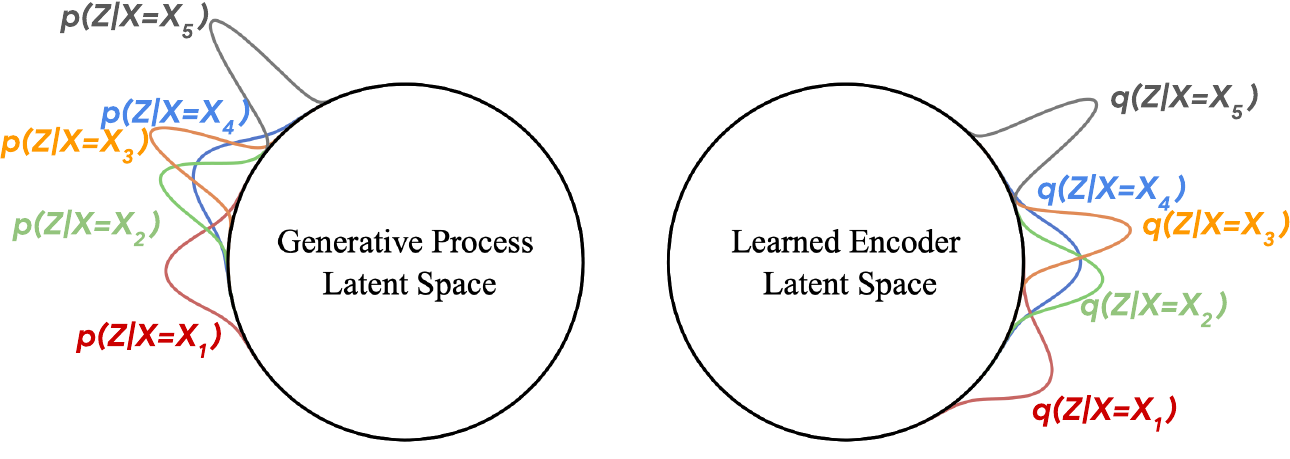}
    \caption{Visualization of the degree to which~\cite{https://doi.org/10.48550/arxiv.2302.02865} can recover the true embedding posteriors: up to a rotation and a flip. Base figure taken from~\cite{https://doi.org/10.48550/arxiv.2302.02865}.}
    \label{fig:guarantee4}
\end{figure}

Pay special attention to the $\argmin$ formulation: This statement is, once more, similar to a strictly proper scoring rule. In summary, it says that the minimizer \(Q(Z \mid X)\) of the loss (think maximizer of the score) has true latent posteriors \(P(Z \mid X)\) (up to rotation and flip), as visualized in Figure~\ref{fig:guarantee4}.

\clearpage

\chapter{Evaluation and Scalability}
\section{Benchmarks and Evaluation}

In this section, we will see common pitfalls of evaluations in trustworthy machine learning.

\subsection{Why do we do evaluation?}

Evaluation enables the ranking of methods. We have a 1D line to put different methods at different positions. We can design new methods that are better than previous ones (\wrt the metric) and advance the field. We often compare to prior state-of-the-art (SotA) methods, but comparing a method's performance against human performance often also makes sense. Sometimes there is also a derived theoretical upper bound for performance, either from previous works or our current work. When a model goes over a theoretical upper bound, one has to explain how that is possible. Either there is a bug in the evaluation, the upper bound is flawed, or the model assumes a different set of ingredients than the upper bound.

It is essential to talk about evaluation because it is hard to do it right. There have been many cases in the literature where the evaluation was wrong, and the field had to pay a huge price for that.

\subsection{What are the costs of wrong evaluation?}

\begin{figure}
\centering
\begin{subfigure}[b]{0.49\linewidth}
\includegraphics[width=\textwidth]{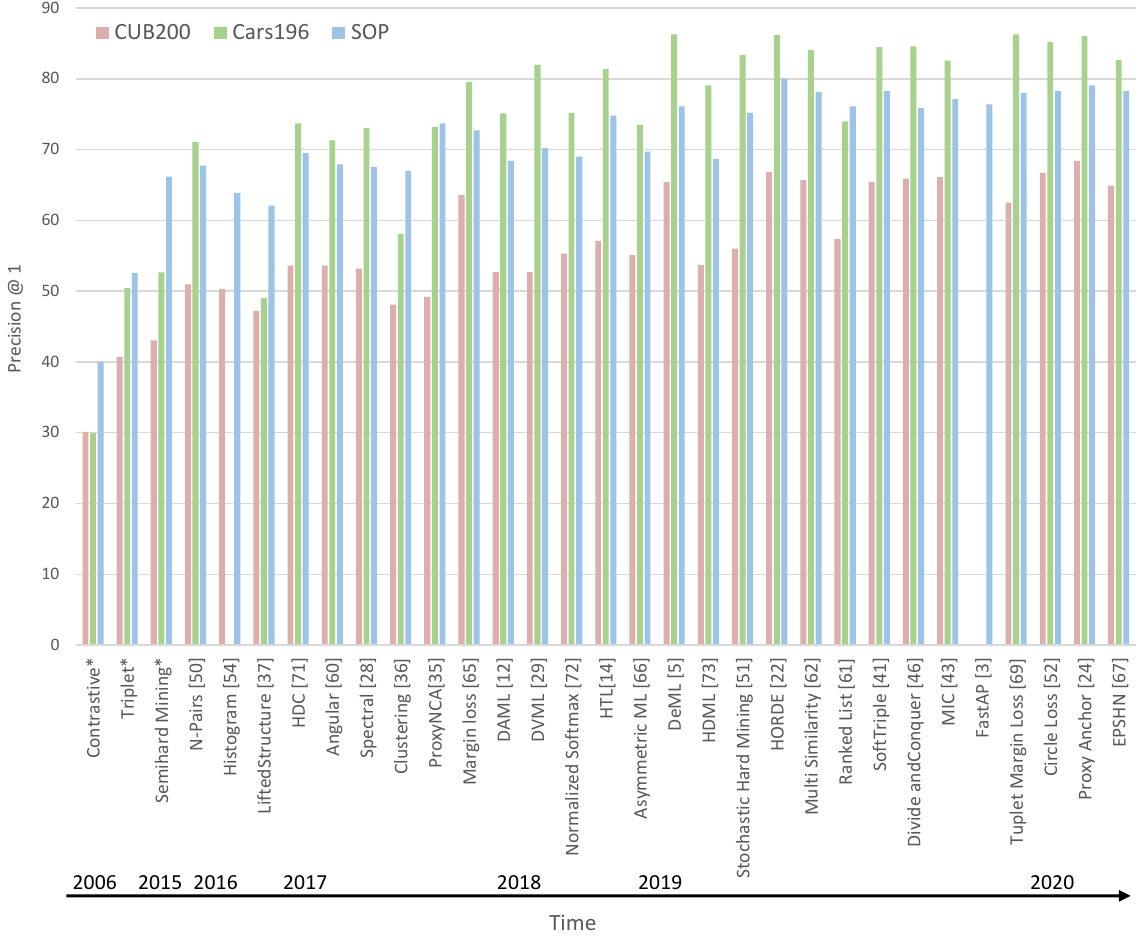}
\caption{The trend according to papers}
\end{subfigure}
\hfill
\begin{subfigure}[b]{0.49\linewidth}
\includegraphics[width=\textwidth]{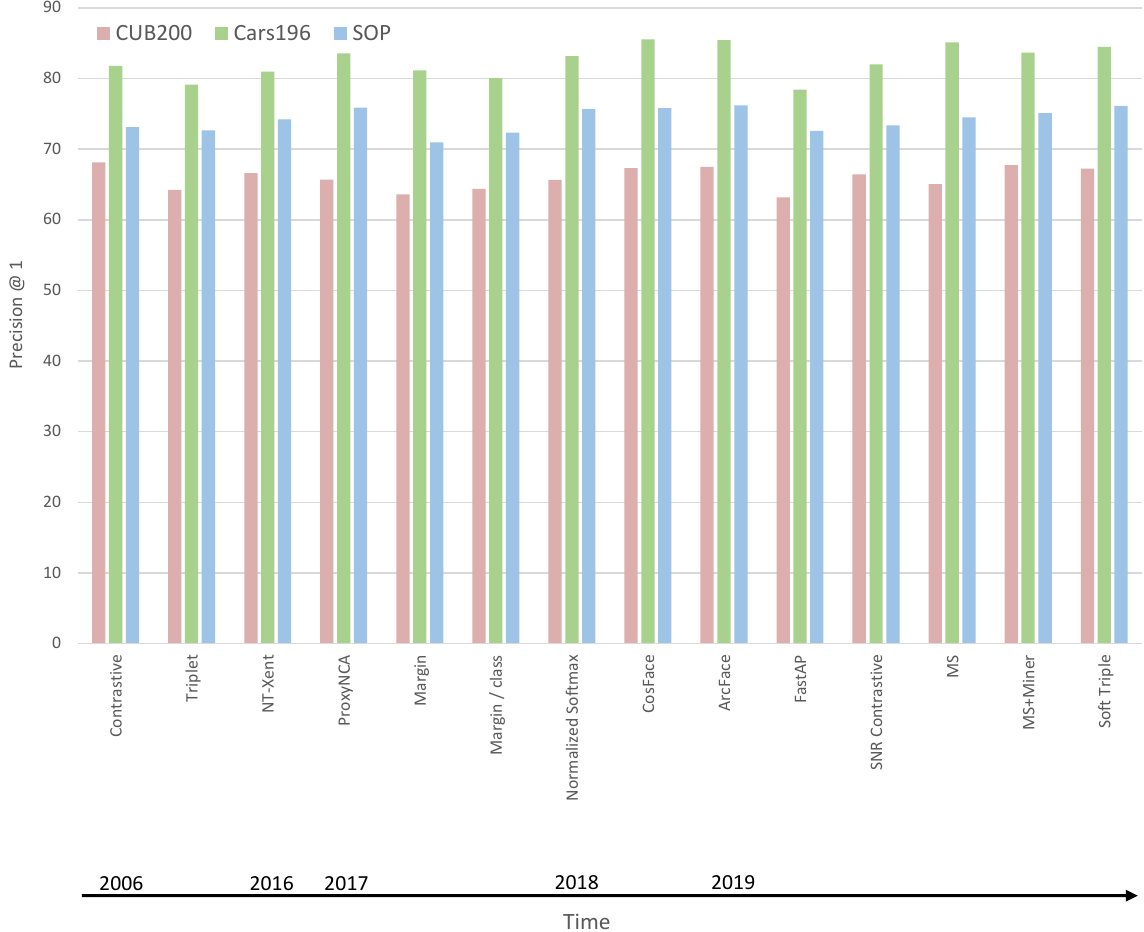}
\caption{The trend according to reality}
\end{subfigure}
\caption{``Papers versus Reality: the trend of Precision@1 of various methods over the years. In a), the baseline methods have * next to them, which indicates that their numbers are the average reported accuracy from all papers that included those baselines.''~\cite{https://doi.org/10.48550/arxiv.2003.08505} In reality (b)), the baseline contrastive loss has comparable performance to all other methods, and no real progress can be seen.}
\label{fig:claims}
\end{figure}

For example, we consider a \href{https://arxiv.org/abs/2003.08505}{metric learning benchmark}. The ``expectation vs. reality'' check is shown in Figure~\ref{fig:claims}. We first discuss what the papers claim over time. The colors correspond to different datasets for measuring metric learning performance. The contrastive loss is the starting point of metric learning methods that the later works built upon. This is the standard method we would use to learn a deep metric representation space. Over time, people have developed complicated tricks to improve upon the baselines and the previous year's SotA method. Importantly, \emph{there is a clear upward trend}.

However, the actual reality is \emph{much worse}. The paper unveils many details of the unfair comparisons that lead to the distorted results seen above. In particular, if we tune the hyperparameters super well for contrastive learning and the recent (so-called) SotA methods and consider a fair comparison of them, we get almost the same performance among the methods.

\textbf{The costs of the wrong evaluation above are severe.}

For researchers, 4+ years of effort was put into pursuing the wrong evaluation protocol where we do not unify the set of ingredients among the methods (e.g., the effort put into fine-tuning previous works). We have a false sense of improvement over time. This also translates to opportunity cost: What if they worked on other ``real'' challenges and were satisfied with contrastive loss instead of working on all these complicated methods?

Practitioners need to select the loss function for their business problems. They waste time looking into all these recent methods, although the most straightforward solution (contrastive loss) probably gives them a good result and requires much less human effort to get it working. This leads to a misinformed selection of methods based on the wrong ranking. They suffer the cost of neglecting a simple solution that works equally well.

\begin{information}{Similar ``evaluation scandals'' in many CV and ML tasks}
We consider a list of similar cases in ML where poor evaluation wasted human effort and money. Typically the papers unveiling the problems with the evaluations tend to be very entertaining to read and interesting; thus, we recommend reading them. They can also be very valuable for practitioners who want an unbiased and correct evaluation of methods they can choose from.
\begin{itemize}
    \item \textbf{Face detection}: Mathias \etal ``\href{https://link.springer.com/chapter/10.1007/978-3-319-10593-2_47}{Face Detection without Bells and Whistles}''~\cite{mathias2014face}. ECCV'14.
    \item \textbf{Zero-shot learning}: Xian \etal ``\href{https://openaccess.thecvf.com/content_cvpr_2017/html/Xian_Zero-Shot_Learning_-_CVPR_2017_paper.html}{Zero-Shot Learning -- The Good, the Bad and the Ugly}''~\cite{xian2017zero}. CVPR'17.
    \item \textbf{Semi-supervised learning}: Oliver \etal ``\href{https://proceedings.neurips.cc/paper/2018/hash/c1fea270c48e8079d8ddf7d06d26ab52-Abstract.html}{Realistic Evaluation of Deep Semi-Supervised Learning Algorithms}''~\cite{oliver2018realistic}. NeurIPS'18.
    \item \textbf{Unsupervised disentanglement}: Locatello \etal ``\href{https://proceedings.mlr.press/v97/locatello19a.html}{Challenging Common Assumptions in the Unsupervised Learning of Disentangled Representations}''~\cite{locatello2019challenging}. ICML'19.
    \item \textbf{Image classification}: Recht \etal ``\href{http://proceedings.mlr.press/v97/recht19a.html}{Do ImageNet Classifiers Generalize to ImageNet?}''~\cite{https://doi.org/10.48550/arxiv.1902.10811} ICML'19.
    \item \textbf{Scene text recognition}: Baek \etal ``\href{https://openaccess.thecvf.com/content_ICCV_2019/html/Baek_What_Is_Wrong_With_Scene_Text_Recognition_Model_Comparisons_Dataset_ICCV_2019_paper.html}{What is Wrong with Scene Text Recognition Model Comparisons? Dataset and Model Analysis}''~\cite{baek2019wrong}. ICCV'19.
    \item \textbf{Weakly-supervised object localization}: Choe \etal ``\href{https://openaccess.thecvf.com/content_CVPR_2020/html/Choe_Evaluating_Weakly_Supervised_Object_Localization_Methods_Right_CVPR_2020_paper.html}{Evaluating Weakly-Supervised Object Localization Methods Right}''~\cite{choe2020evaluating}. CVPR'20.
    \item \textbf{Deep metric learning}: Musgrave \etal ``\href{https://link.springer.com/chapter/10.1007/978-3-030-58595-2_41}{A Metric Learning Reality Check}''~\cite{https://doi.org/10.48550/arxiv.2003.08505}. ECCV'20.
    \item \textbf{Natural language QA}: Lewis \etal ``\href{https://aclanthology.org/2021.eacl-main.86.pdf}{Question and Answer Test-Train Overlap in Open-Domain Question Answering Datasets}''~\cite{https://doi.org/10.48550/arxiv.2008.02637}. ArXiv'20.
\end{itemize}
These papers cover many domains in CV and NLP in general.
\end{information}

\subsection{``Recipes'' for Wrong Benchmark Evaluation}

What are the typical patterns in wrong benchmarking/evaluation? We provide an incomplete list of possible failure modes.

\subsubsection{Everyone writes their own evaluation metric code.}

Even if things are mathematically the same, when it comes to coding, everyone has different ways of handling corner cases. There are non-trivial code-level details in some evaluation metrics. For example, for computing average precision (AP = AUPR), how should we handle precision values for high-confidence bins where the threshold is very high, and thus there are no positive predictions at all? In such cases
\[\operatorname{Precision}(p) = \frac{|\operatorname{TP}(p)|}{|\operatorname{TP}(p)| + |\operatorname{FP}(p)|} = \frac{0}{0}.\]
This is undefined. Some argue that it should be considered 0, some decide to use 1, and others say it should be excluded from the integral computation (for calculating the AP).
There must be some agreement on handling such cases in practice. What probably works best in these cases is to have an evaluation server or a library for computing the metrics. That way, we can ensure that all methods use the same implementations of metrics.

\subsubsection{Confounding multiple factors when comparing methods.}

\begin{figure}
    \centering
    \includegraphics[width=0.8\linewidth]{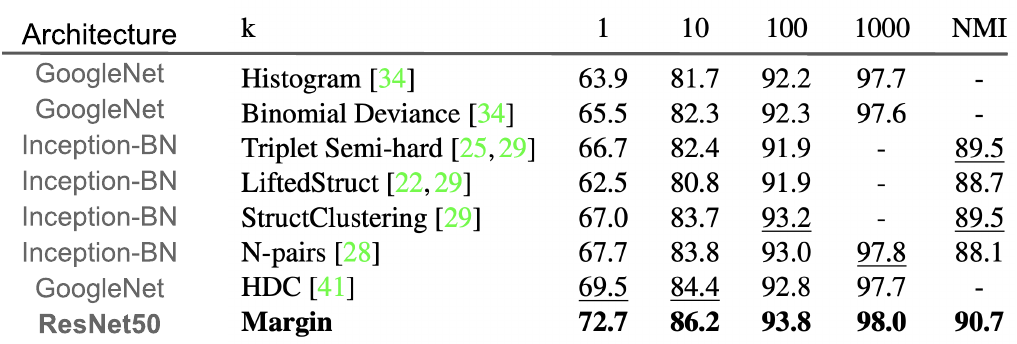}
    \caption{Example of a scenario where some part of the setup stays hidden (the used architecture) that would clarify an unfair comparison of methods. The first column is not shown in the original paper, although it is very influential of the results. Base figure taken from~\cite{https://doi.org/10.48550/arxiv.1706.07567}.}
    \label{fig:confound}
\end{figure}

An example is shown in Figure~\ref{fig:confound}. Consider the paper ``\href{https://arxiv.org/abs/1706.07567}{Sampling Matters in Deep Embedding Learning}''~\cite{https://doi.org/10.48550/arxiv.1706.07567} They argue in a benchmark that their novel loss function is bringing them gains. But do the improvements really come from the loss function? They do not disclose that the architectures used for training with the respective losses were different. In particular, for training with their loss, they used a more modern architecture (ResNet-50) than for the others (GoogleNet, Inception-BN -- archaic). Then it is naturally expected that a Resnet-50 performs better than a GoogleNet. By confounding multiple factors, it becomes hard to rank the losses alone.

\subsubsection{Hiding extra resources needed to make improvements.}

We mention the work ``\href{https://arxiv.org/abs/1904.01906}{What Is Wrong With Scene Text Recognition Model Comparisons? Dataset and Model Analysis}''~\cite{https://doi.org/10.48550/arxiv.1904.01906} If we only care about accuracy, we might be missing the other important axis: computational cost and efficiency. When we only look at an accuracy plot, we are more inclined to select the method with the highest accuracy. However, if we also considered the inference time (latency) or other computational costs, maybe we would want a different method than the one with the highest accuracy (that is a bit less accurate but much faster).

\subsubsection{Training and test samples overlap.}

This problem is illustrated in Table~\ref{tab:overlap}. We consider the paper ``\href{https://aclanthology.org/2021.eacl-main.86.pdf}{Question and Answer Test-Train Overlap in Open-Domain Question Answering Datasets}''~\cite{https://doi.org/10.48550/arxiv.2008.02637}. In this work, a general problem is highlighted where a fraction of the test sets overlap with the training set for the natural language Q\&A task.

\begin{table}
    \centering
    \caption{``Fractions of open-domain [question answering] test sets that overlap with their training sets.''~\cite{https://doi.org/10.48550/arxiv.2008.02637}.}
    \label{tab:overlap}
    \begin{tabular}[t]{lcc}
    \toprule
    Dataset & \multicolumn{1}{ m{0.1\textwidth}}{\centering \% Answer overlap} &\multicolumn{1}{m{0.1\textwidth}}{\centering \% Question overlap}  \\
    \midrule
    Natural Questions & 63.6 & 32.5 \\
    TriviaQA & 71.7& 33.6 \\
    WebQuestions & 57.9 & 27.5 \\
    \bottomrule
    \end{tabular}
\end{table}

Sometimes, our evaluation set is contaminated: We see many test samples during training. The Natural Questions, TriviaQA, and WebQuestions datasets are popular benchmarks for the Q\&A task in NLP. It turns out that for all three datasets, there is a \(> 50\%\) answer overlap (up to \(70\%\)!) with the test answers. By memorizing the training answers, it becomes much easier for the model to produce a good answer at test time. Questions are also overlapping quite a bit, as shown in Table~\ref{tab:overlap2}. The authors show that the models solve the task by memorizing rather than generalizing. Many models achieve \(0\%\) accuracy for no overlap samples.

\begin{table}
    \centering
    \caption{Model performances (exact match scores) in different partitions of the test set. The question/answer overlaps are significant and severe. Table taken from~\cite{https://doi.org/10.48550/arxiv.2008.02637}.}
    \label{tab:overlap2}
    \setlength{\tabcolsep}{2pt}
    \begin{tabular}[t]{ll|cccc}
    \toprule
    \multicolumn{2}{c|}{\multirow{2}{*}{Model}} &  \multicolumn{4}{c}{Open Natural Questions} \\
    \multicolumn{2}{c|}{} &  \multicolumn{1}{m{0.7cm}}{\centering \scriptsize{Total}} &\multicolumn{1}{m{1cm}}{\centering \scriptsize{Question Overlap}} & \multicolumn{1}{m{1cm}}{\centering \scriptsize{Answer Overlap Only}}&  \multicolumn{1}{m{1cm}}{\centering \scriptsize{No Overlap}} \\
    \midrule
    \multirow{2}{1.7cm}{Closed book}& \scriptsize{T5-11B+SSM} &36.6 & 77.2 & 22.2 & 9.4\\
    & BART                                                  &26.5 & 67.6 & 10.2 & 0.8\\
    \midrule
    \multirow{2}{1.7cm}{Nearest Neighbor}& Dense            &26.7 & 69.4 & 7.0 & 0.0\\
    & TF-IDF                                                &22.2 & 56.8 & 4.1 & 0.0\\
    \bottomrule
    \end{tabular}
\end{table}

\subsubsection{Lack of validation set.}

\begin{figure}
  \centering
  \begin{subfigure}{0.48\textwidth}
    \includegraphics[width=\linewidth]{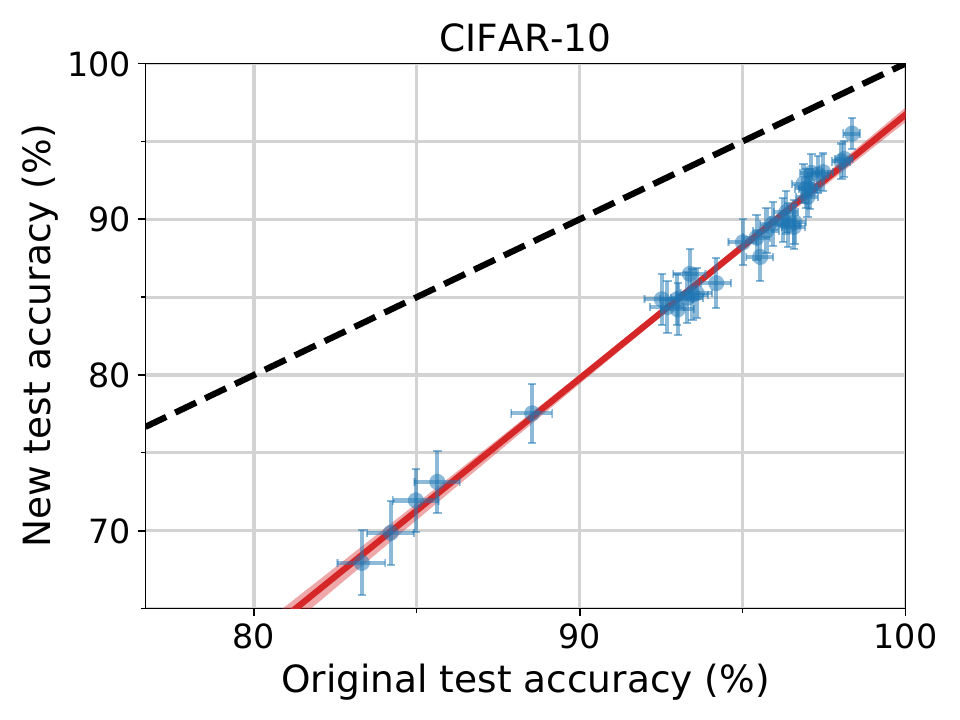}
  \end{subfigure}
  \hfill
  \begin{subfigure}{0.48\textwidth}
    \includegraphics[width=\linewidth]{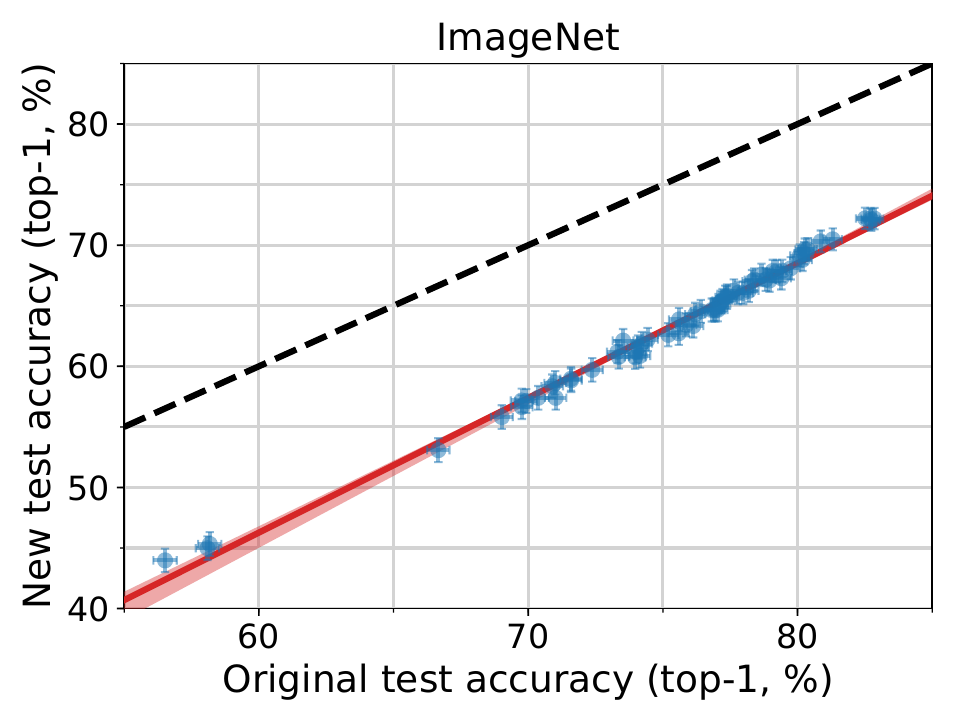}
  \end{subfigure}
  \begin{subfigure}{\textwidth}
    \vspace{-.15cm}
    \centering
    \includegraphics[width=.75\linewidth]{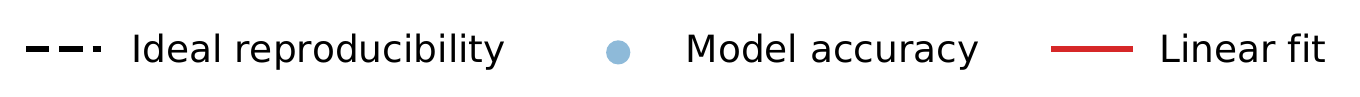}
  \end{subfigure}
  \caption{Comparison of model accuracy on the original test sets vs. the new test sets collected by the authors. Ideal reproducibility is the line of identity -- the performances should not differ at all if the models are not overfit. This is not the case: All models are overfit to both CIFAR-10 and ImageNet. Figure taken from~\cite{https://doi.org/10.48550/arxiv.1902.10811}.
  }
  \label{fig:imagenet_overfit}
\end{figure}

This problem is shown in Figure~\ref{fig:imagenet_overfit}. Sometimes, there is no published validation set. The CIFAR and ImageNet classification benchmarks lack validation sets. To be precise, ImageNet has a validation but not a test set. Therefore, people are using the validation set as a test set. This brings us many problems. When there is an improvement on the ImageNet validation set benchmark, it is usually the pointwise samples in the validation set that are addressed rather than the general image classification task. There have been questions like ``Are we solving ImageNet or image classification?''.

Another question for the same narrative is ``\href{https://arxiv.org/abs/1902.10811}{Do ImageNet Classifiers Generalize to ImageNet?}''~\cite{https://doi.org/10.48550/arxiv.1902.10811}. In this case, the design choices and hyperparameter tuning are performed over the test set, spoiling its measure of generalization. There is some evidence that ImageNet classifiers do not generalize to ImageNet, and they are overfitted to the test set. The same holds for CIFAR. The authors of the referenced paper collected another ImageNet validation set, following the same collection procedure. They found that compared to the original ImageNet validation set, the version 2 validation set accuracy is notably lower. If we plot the performances of individual models on the original validation set against the corresponding performances on the version 2 validation set as a scatter plot, it tends to follow a line below the \(x = y\) line, indicating that the models do not seem to generalize well. The models' dropping performances on new samples from the same distribution is evidence of overfitting the design choices to the test set over time.

\begin{information}{Practical Pointers for Failure Modes}
When we start a new research project in a particular field, how should we find the common failure modes, and how can we avoid them in evaluation?

\medskip

A good first thing to check is whether there is any paper about the fair comparison of all the methods we are interested in / we want to improve upon.

\medskip

If not, we have three choices.
\begin{enumerate}
    \item Write such a paper ourselves to ``unify all the numbers''. This takes the most work, but it can be very rewarding.
    \item Say that we trust the benchmark because we think there are not so many complicated ingredients involved in the setup; therefore, there is not so much room for the researchers to confound multiple factors during evaluation, e.g., by introducing architectural changes. When the task and the ingredients are both simple, we might want to trust the benchmark.
    \item Choose to stay skeptical and leave the field until someone performs a trustworthy unified evaluation.
\end{enumerate}
\end{information}

It is crucial to do the evaluation right; otherwise, we are losing much money, time, and research effort. There are currently many domains where this is going sideways, as seen from the list above.

\section{Scalability}

\begin{figure}
    \centering
    \includegraphics[width=\linewidth]{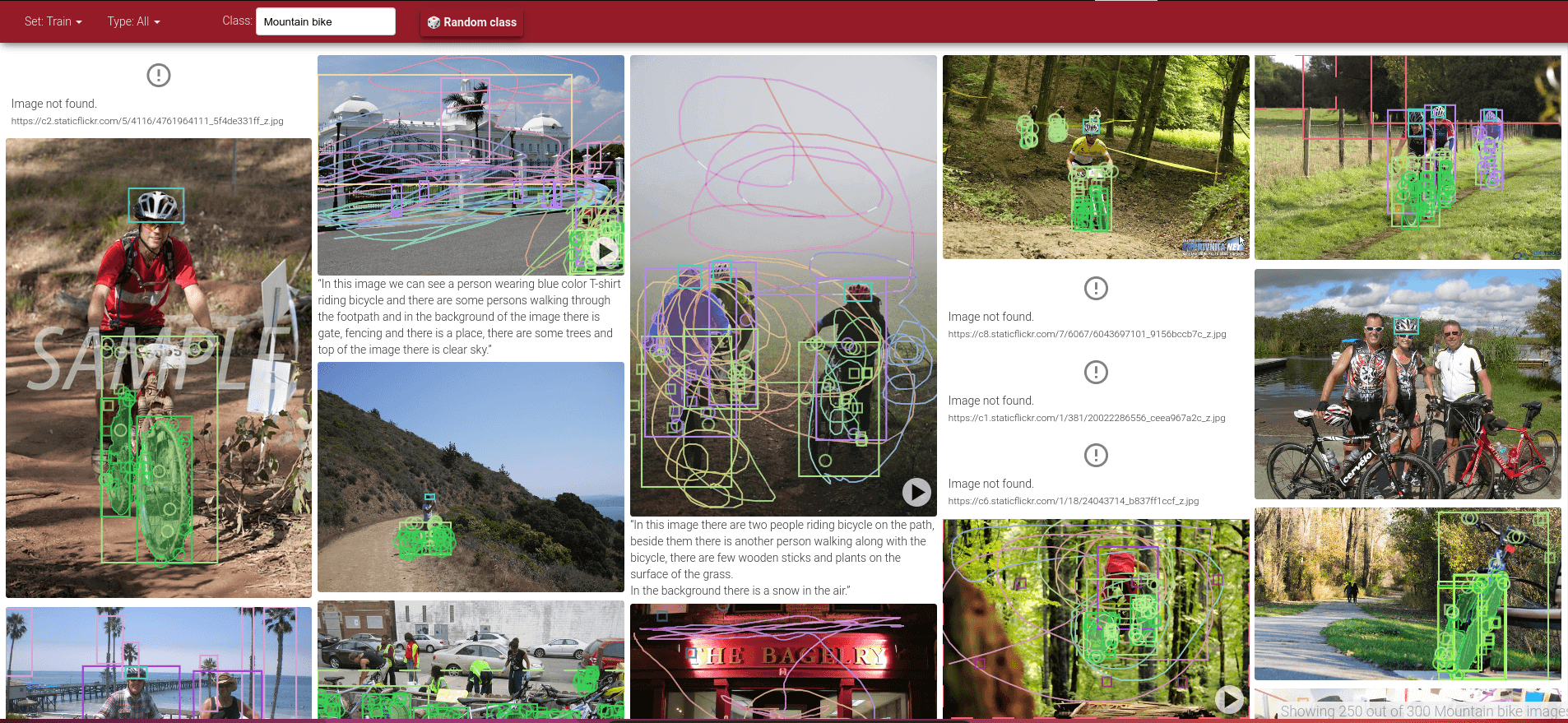}
    \caption{Sample of the Open Images Dataset (V7)~\cite{OpenImages2}.}
    \label{fig:openv7}
\end{figure}

If we look at TML papers, TML is often studied with ``toy'' datasets. These have the following properties:
\begin{itemize}
    \item Low-dimensional data (\(\le\) order of 1000 dims per sample).
    \item Small number of training samples (\(\le\) order of 100k samples).
    \item Benefit: More extensive and precise labels are available per sample. For example, we have all kinds of attributes labeled for the sample (not just the task label but also other attribute labels like the domain or bias label).
    \item They make quick evaluation possible.
    \item Controlled experiments are also possible.
    \item This kind of dataset accommodates complicated methods with many hyperparameters. Typically we can bring in many of them and tune them the right way to generate the best results here.
\end{itemize}
Example datasets used in OOD generalization can be seen in Figure~\ref{fig:domainbed}. The real impact, however, comes from results on large-scale datasets. These have the following properties:
\begin{itemize}
    \item High-dimensional data (\(\ge\) order of 10k dims per sample).
    \item Large number of training samples (\(\ge\) order of 1M samples). These days this is not that large-scale either; we can go up to 1B samples if we have the resources.
    \item High-quality labels are dearer. Often a large portion of our data is even unlabelled or very noisily labelled.
    \item The validation of an idea on large-scale datasets may take days-weeks. We cannot validate hyperparameter settings super frequently.
    \item It is hard to analyze the contributing factors. We do not have all the labels we had for the toy dataset, and we also do not have the time and resources to determine which kind of factor contributes to the performance. It is hard to gain knowledge and insights from this kind of data. Therefore, we need some simple methods without many design choices (few hyperparameters).
\end{itemize}
An example large-scale dataset is the Open Images Dataset (V7), illustrated in Figure~\ref{fig:openv7}.

\subsection{Possible Roadmap to Scaling Up TML}

There is a possibility to combine toy-ish data and real data to make an impact. This is the method of scaling up from toy data to real data.

First, we work with toy data (e.g., MNIST). Here we have to be creative and propose new ideas (potentially complicated methodology) through quick experiments and tuning hyperparameters. Our goal is to understand why things work.

Based on the insights from toy data and a set of candidate tools, we can go to real data (scaling up). Here we have to identify and remove unnecessary complexities based on knowledge from toy data and aim for something simple. Based on the understanding obtained on toy datasets, we make a good guess about what will work on real data.

The point here is that eventually, to scale up, we need something simple. Of course, going simple is not always easy. We will see some examples where simple wins. There are tons more on arXiv and Twitter.

\subsection{Simple Wins}

\begin{table}
\begin{center}
\caption{``Average out-of-distribution test accuracies for all algorithms, datasets, and model selection criteria included in the initial release of DomainBed. [...]''~\cite{DBLP:journals/corr/abs-2007-01434} Tuned fairly, ERM -- the simplest method -- is not worse at all than other complicated methods.}
\label{tab:lost}
\resizebox{\textwidth}{!}{%
\begin{tabular}{lcccccccc}
\toprule
\multicolumn{8}{c}{{Model selection method: Test-domain validation set \textit{(oracle)}}} \\
\midrule
\textbf{Algorithm}    & \textbf{CMNIST}       & \textbf{RMNIST}     & \textbf{VLCS}             & \textbf{PACS}             & \textbf{Office-Home}       & \textbf{TerraInc}   & \textbf{DomainNet} & \textbf{Avg} \\
\midrule
ERM                       & 58.5 $\pm$ 0.3            & 98.1 $\pm$ 0.1            & 77.8 $\pm$ 0.3            & 87.1 $\pm$ 0.3            & 67.1 $\pm$ 0.5            & 52.7 $\pm$ 0.2            & 41.6 $\pm$ 0.1 & 68.9\\
IRM                       & 70.2 $\pm$ 0.2            & 97.9 $\pm$ 0.0            & 77.1 $\pm$ 0.2            & 84.6 $\pm$ 0.5            & 67.2 $\pm$ 0.8            & 50.9 $\pm$ 0.4            & 36.0 $\pm$ 1.6 & 69.2\\
DRO                       & 61.2 $\pm$ 0.6            & 98.1 $\pm$ 0.0            & 77.4 $\pm$ 0.6            & 87.2 $\pm$ 0.4            & 67.7 $\pm$ 0.4            & 53.1 $\pm$ 0.5            & 34.0 $\pm$ 0.1 & 68.4\\
Mixup                     & 58.4 $\pm$ 0.2            & 98.0 $\pm$ 0.0            & 78.7 $\pm$ 0.4            & 86.4 $\pm$ 0.2            & 68.5 $\pm$ 0.5            & 52.9 $\pm$ 0.3            & 40.3 $\pm$ 0.3 & 69.0\\
MLDG                      & 58.4 $\pm$ 0.2            & 98.0 $\pm$ 0.1            & 77.8 $\pm$ 0.4            & 86.8 $\pm$ 0.2            & 67.4 $\pm$ 0.2            & 52.4 $\pm$ 0.3            & 42.5 $\pm$ 0.1 & 69.1\\
CORAL                     & 57.6 $\pm$ 0.5            & 98.2 $\pm$ 0.0            & 77.8 $\pm$ 0.1            & 86.9 $\pm$ 0.2            & 68.6 $\pm$ 0.4            & 52.6 $\pm$ 0.6            & 42.1 $\pm$ 0.1 & 69.1\\
MMD                       & 63.4 $\pm$ 0.7            & 97.9 $\pm$ 0.1            & 78.0 $\pm$ 0.4            & 87.1 $\pm$ 0.5            & 67.0 $\pm$ 0.2            & 52.7 $\pm$ 0.2            & 39.8 $\pm$ 0.7 & 69.4\\
DANN                      & 58.3 $\pm$ 0.2            & 97.9 $\pm$ 0.0            & 80.1 $\pm$ 0.6            & 85.4 $\pm$ 0.7            & 65.6 $\pm$ 0.3            & 51.6 $\pm$ 0.6            & 38.3 $\pm$ 0.1 & 68.2\\
C-DANN                    & 62.0 $\pm$ 1.1            & 97.8 $\pm$ 0.1            & 80.2 $\pm$ 0.1            & 85.7 $\pm$ 0.3            & 65.6 $\pm$ 0.3            & 51.0 $\pm$ 1.0            & 38.9 $\pm$ 0.1 & 68.7\\
\bottomrule
\end{tabular}}
\end{center}
\end{table}

\subsubsection{OOD Generalization}

Consider addressing the \href{https://arxiv.org/abs/2007.01434}{OOD generalization}~\cite{DBLP:journals/corr/abs-2007-01434} problem and the fair evaluation shown in Figure~\ref{tab:lost}. Tuned fairly, ERM -- the simplest method -- is not worse at all than other complicated methods. (ERM: Training on the whole combined dataset without domain label.) We discussed DRO and DANN in this book.

\begin{figure}
    \centering
    \resizebox{0.5\linewidth}{!}{
            \input{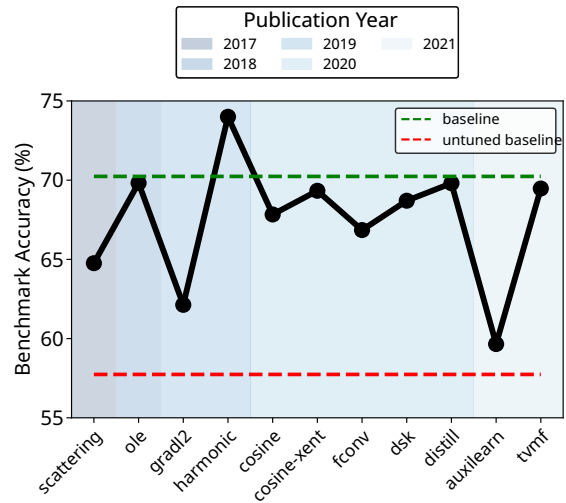}}
    \caption{``Accuracy of state-of-the-art methods and baselines on the proposed benchmark. The untuned baseline (red dashed line) is trained with default hyper-parameters, i.e., a learning rate of 0.1, and weight decay of \(10^{-4}\). Conversely, for all other methods, including the baseline (green dashed line), we performed hyper-parameter optimization.
    Methods are ordered on the x-axis according to their publication year.''~\cite{Brigato_2022} Seemingly, a lot of human effort went to waste: One could have tuned the baseline instead of coming up with new methods that give almost identical or worse results.}
    \label{fig:untuned}
\end{figure}

\subsubsection{Loss Functions}

We have a lot of \href{https://arxiv.org/abs/2212.12478}{different variants of loss functions}~\cite{Brigato_2022} we can use for training a classifier. An interesting contrast between a tuned and an untuned baseline is shown in Figure~\ref{fig:untuned}. Here, the baseline is vanilla CE. The red line is what papers report as the performance of vanilla CE. They say their method works better than vanilla CE. However, it does not. They just did not tune the baseline properly. We should always do it as well as we possibly can and be completely clear about our methodology.

\textbf{Note}: It is true that nowadays, reviewers are much more careful with checking evaluation setups than in 2017. However, one can always be not 100\% clear about how they performed the evaluation. They can say that they did the tuning, but they can hide the fact that they did so, e.g., using a tiny search window. They can also, e.g., leave out weight decay from the baseline (``Who cares about weight decay\dots''). Weight decay actually turns out to be quite important. Papers that properly tune hyperparameters for fair comparison tend to recognize the importance of weight decay. We can see a mismatch between what people generally know and what is actually true. By not being 100\% descriptive, people can still sneak in papers to conferences by saying they did the hyperparameter search, leaving out subtle and important details (What they did not do right.)

\subsubsection{Weakly-Supervised Object Localization}

\begin{figure}
    \centering
    \includegraphics[width=\linewidth]{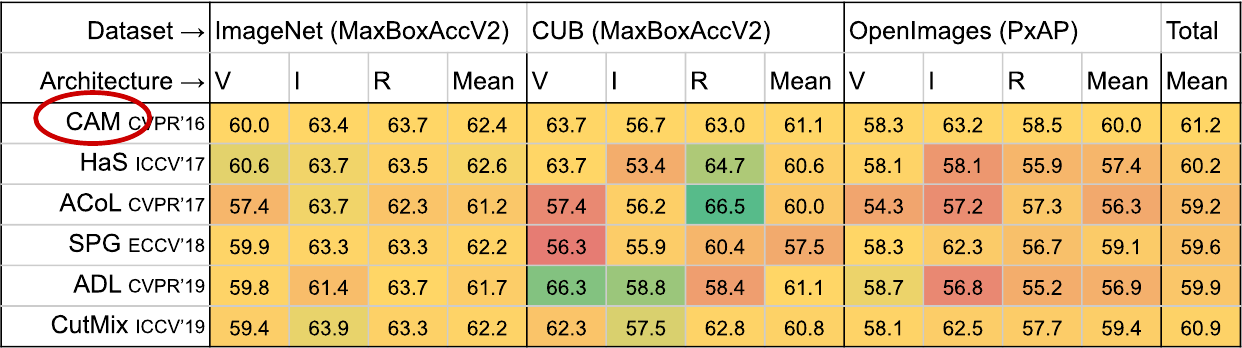}
    \caption{Re-evaluation of various methods on WSOL. CAM, a method from 2016, is still the best when tuned appropriately. Table adapted from~\cite{https://doi.org/10.48550/arxiv.2007.04178}.}
    \label{fig:camevaluation}
\end{figure}

\href{https://arxiv.org/abs/2007.04178}{Weakly-Supervised Object Localization}~\cite{https://doi.org/10.48550/arxiv.2007.04178} is another field that had such a scandal. In Figure~\ref{fig:camevaluation}, we see a table adapted from the authors' work. The coverage of the paper's re-evaluation is extensive. CAM has been the simplest method for WSOL for a while now.  All the other papers reported better results than CAM in general. However, when we do everything correctly with the same set of ingredients, CAM is the best method on average.

\subsection{One Right Way to Tune Hyperparameters}

The rule of thumb we propose is random search.
\begin{enumerate}
    \item Set a sensible range of hyperparameters by searching the exponential space. From this heuristic, we could already see that, e.g., \(10^{-20}\) for the learning rate is not sensible at all; no learning happens there. We cut off parts of the parameter space that are not sensible.
    \item Once we have this search space defined by sensible ranges of 5-10 hyperparameters, we perform a random search with a fixed number of iterations/samples (between different methods). Random search in practice is excellent. The intuition for that: In practice, not all parameters contribute equally to the performance. In particular, some might not contribute to it at all, only a selection of them. In that case, randomly searching the exponential grid is already good enough because all irrelevant dimensions will not contribute anyway, and all the search samples in this exponential grid are effectively just searching in the relevant dimensions of the hyperparameter space.
\end{enumerate}

Of course, if many hyperparameters contribute to the final result, we might want to consider more principled techniques, e.g., Bayesian Optimization. However, the authors have never used it for hyperparameter search, only random search.

Suppose the viable hyperparameter regions are super non-convex and very wiggly. In that case, this range-based approach might not work, as we will probably lose a lot of reasonable solutions by cutting the space. We hope this is not the case and the loss is close to unimodal in the space of hyperparameters we are optimizing over. We also assume the independence of the involved hyperparameters. (And that many of these do not matter, actually.)

\section{Transition from ``What'' to ``How''}

Now let us consider future research ideas from the authors. We walk about these from the perspective of going from the ``What'' to the ``How'' question we have seen before. In ML 2.0 we learn \(P(X, Z, Y)\) from \((X, Y)\) (``What'') data. If we look at the ingredients, there is a historical artifact. For ML 1.0, we trained on \((X, Y)\) data, but for ML 2.0, we are still using the same ingredients for solving the derivative problems. Is this right? Is this going to work? We argue that the answer is likely no.

\subsection{Our Vision}

\begin{figure}
    \centering
    \includegraphics[width=0.8\linewidth]{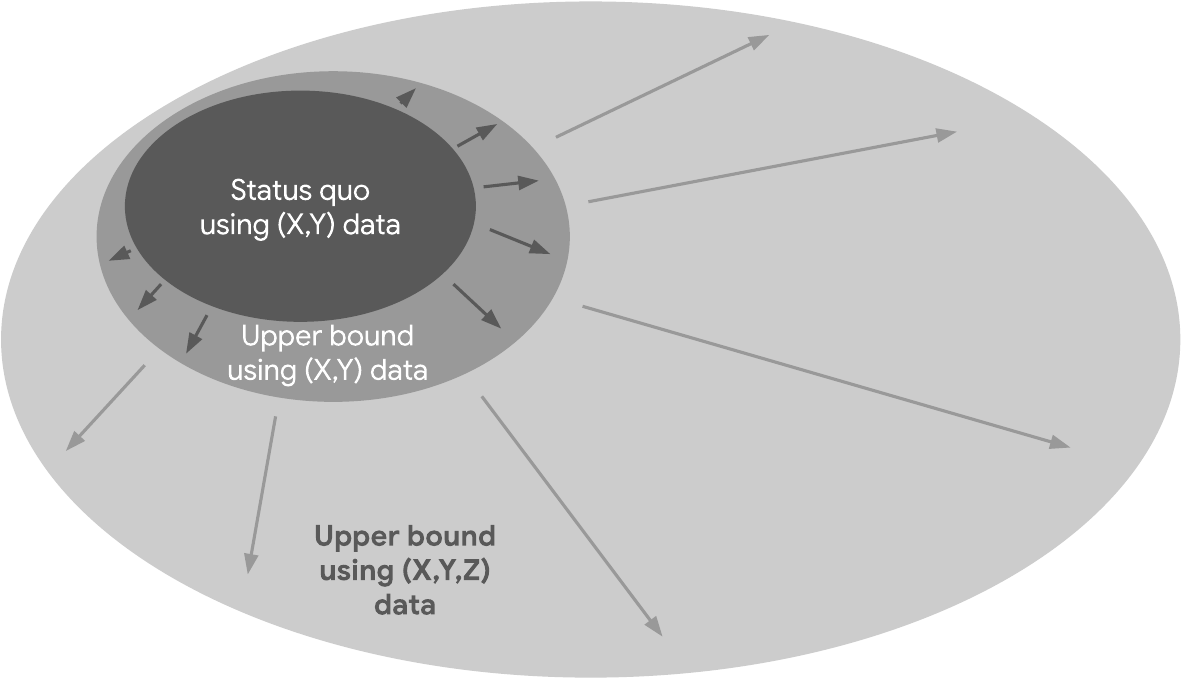}
    \caption{Different perspectives of different settings. \((X, Y)\) data correspond to limited ingredients. The upper bound of performance using such data is also very limited. Using \((X, Y, Z)\) opens much broader perspectives.}
    \label{fig:vision}
\end{figure}

We can go simple in terms of the method, but what could be really interesting in the future is to collect new types of datasets for scalability and trustworthiness. We discuss Figure~\ref{fig:vision}.

The inner two ovals correspond to the \emph{benchmarking} approach. This is the typical approach we have been discussing so far. In a fixed benchmark, everyone uses the same ingredients: an \((X, Y)\) dataset. If we allow them to use more ingredients, it is no longer a benchmark. It is unfair. (Although that is what people do sometimes still.) The \emph{goal} is to compete for the highest accuracy by using the ingredients most efficiently and smartly. We want to generate the maximal performance from a limited set of ingredients. The \emph{key contribution} is usually the learning algorithm. This approach used to work well for ``What'' problems. However, we think we should probably use new types of data (new ingredients) that also involve \(Z\). If we discuss with reviewers, we learn that this ``learning algorithm contribution + using the same ingredients'' is the default mode of thinking for many people.

The outer oval corresponds to the \emph{data hunting} approach. We are still doing some competition, but we are not confined to using the same ingredients. Searching for the ingredient itself is part of the game. When we allow people to use new ingredients, we invite creative new ways to find cheap sources of information that could give us hints about the \(Z\) data from all kinds of places. Competitors are allowed to use other ingredients: \((X, Y) + Z\). The \emph{source of value} is the discovery of new, efficient data sources. This is the future of addressing the ``How'' task. This is also the general research direction the authors of this book want to pursue in the future.

\subsection{Data as Compressed Human Knowledge}

Data are a \emph{compression of human knowledge}. When training an ML model, there are typically two sources of human knowledge.

The first comes from the data and is embedded in the ML model through training. There is a transition of the abstract concept of knowledge in the real world (from annotators) into a dataset in the computational domain. Usually, the dataset contains labels crowdsourced by some annotators.

\begin{figure}
    \centering
    \includegraphics[width=0.8\linewidth]{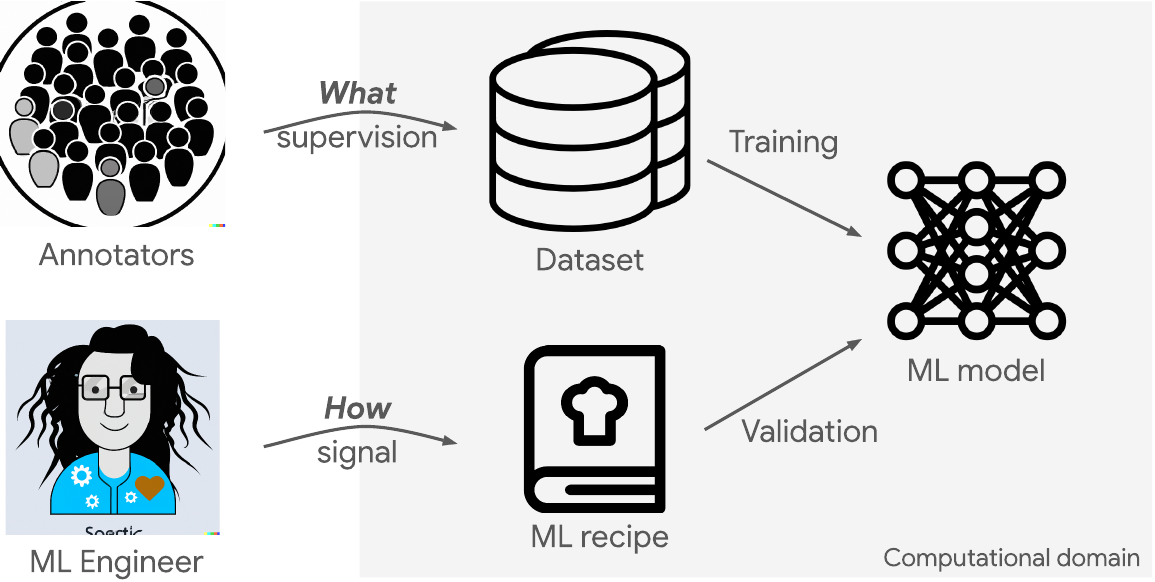}
    \caption{The currently dominating types/sources of supervision. Annotators only give ``What'' supervision through the labeling process. The ``How'' signal is only supplied by ML engineers through the recipe of creating and training the model.}
    \label{fig:compressed1}
\end{figure}

The second source of human knowledge comes from the validation loop. There is a transition of knowledge of the ML engineer into the recipes for training an ML model that they develop over time. A general overview of this setup is shown in Figure~\ref{fig:compressed1}. This is typically how people are addressing ``How'' problems now: through the ML engineer's knowledge. Through many validations, we can find the right setup and design to achieve the ``How'' tasks. They use the same kind of \((X, Y)\) dataset and rely on the ML engineers to encode business intentions, such as:
\begin{itemize}
    \item ``We need more transparency in the model.''
    \item ``We need more robustness.''
    \item ``We need better OOD generalization.''
\end{itemize}

\begin{figure}
    \centering
    \includegraphics[width=0.8\linewidth]{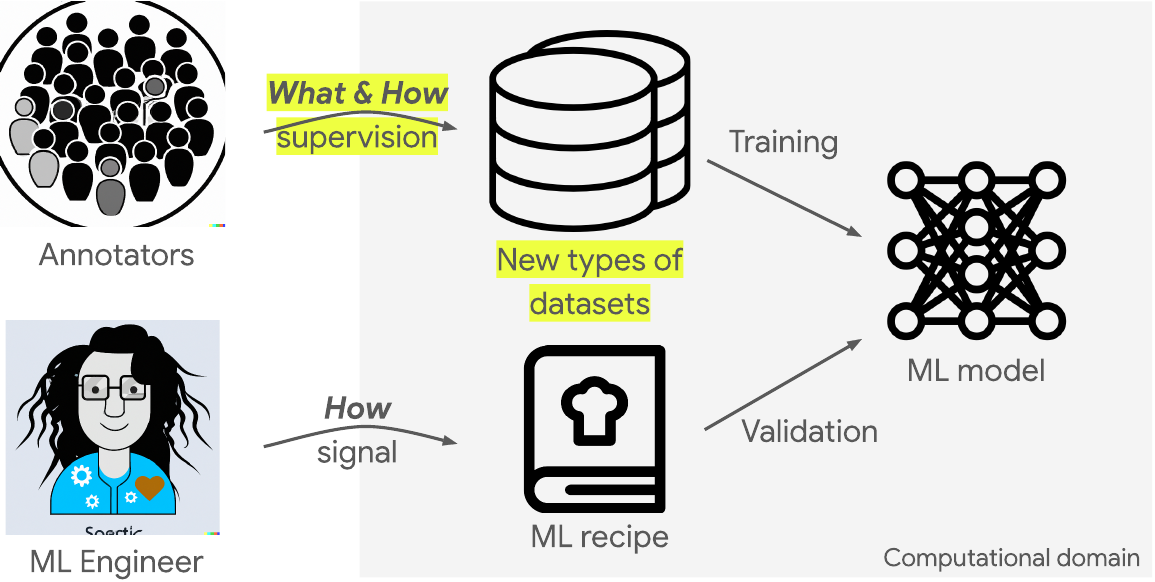}
    \caption{A possible future paradigm for types/sources of supervision. Here, the annotators also provide ``How'' supervision, which can lead to much more robust models.}
    \label{fig:compressed2}
\end{figure}

We argue that in the future, we should also look for methods or datasets for addressing the ``How'' problem \emph{from the dataset side}. In the future, ``How'' will probably also be handled through data collection. This is illustrated in Figure~\ref{fig:compressed2}. We wish to not only collect ``What'' supervision from the annotators but also information related to the ``How'' task. This way, we obtain a new type of dataset that could be very interesting to the community.

\begin{figure}
    \centering
    \includegraphics[width=0.6\linewidth]{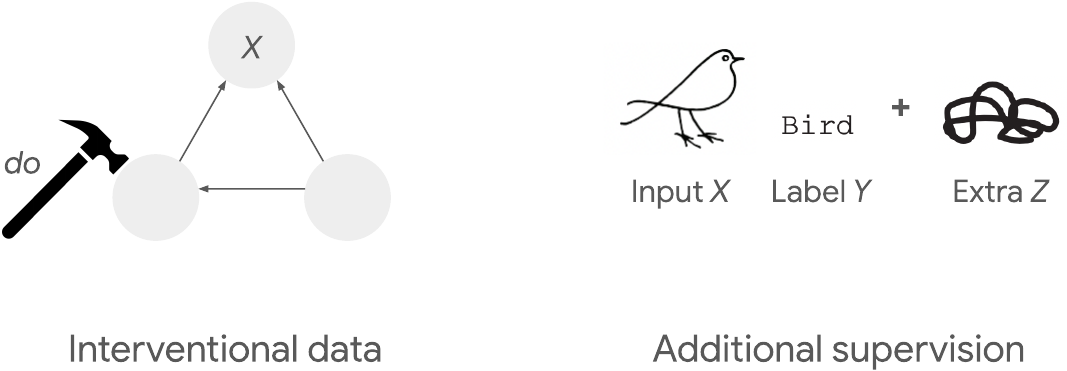}
    \caption{Different types of ``How'' data. Interventional data specify the ``How'' aspect by breaking spurious correlations that lead to the incorrect selection of cues. Additional supervision provides explicit new information to specify our needs more thoroughly.}
    \label{fig:howdata}
\end{figure}

We will now specify two examples of ``How'' data: We will consider \emph{interventional data} and \emph{additional supervision} on top of our standard annotations. An illustration is given in Figure~\ref{fig:howdata}.

\subsection{Interventional Data}

\begin{figure}
    \centering
    \includegraphics[width=0.8\linewidth]{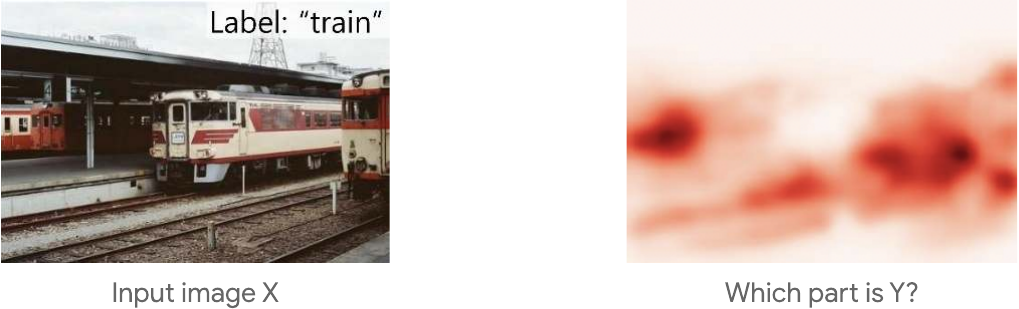}
    \caption{Example (input, attribution map) pair that highlights spurious correlations between the label `train' and the rails. Considering most natural images, the model can get away with looking at the rails because it is quite uncommon to see a train without rails. However, this choice of cue is misspecified and does not lead to robust generalization. Base figure taken from~\cite{https://doi.org/10.48550/arxiv.2203.03860}.}
    \label{fig:train}
\end{figure}

We will discuss the paper ``\href{https://arxiv.org/abs/2203.03860}{Weakly Supervised Semantic Segmentation Using Out-of-Distribution Data}''~\cite{https://doi.org/10.48550/arxiv.2203.03860}.
First, let us consider an example of the spurious correlation between trains and rails. If we visualize where our model is looking for the class `train', we are probably going to get something like in Figure~\ref{fig:train}. The models often look a lot on the rail pixels. This is a well-known problem. The reason is that if we collect data naturally arising from the way people take pictures, then we will probably see many images where the trains are on the rails. Models can recognize trains based on rails, leveraging spurious correlation. The learned ``How'' by the model is wrong. The existence of spurious correlations already indicates that interventional data does not arise naturally in natural data. (If they arose naturally, they would have been part of the training data already, and there would not be any spurious correlation at all.) However, we did not encode ``How'' requirements in our dataset. \emph{Therefore, we cannot expect our model to get it right.}

\begin{figure}
    \centering
    \includegraphics[width=0.6\linewidth]{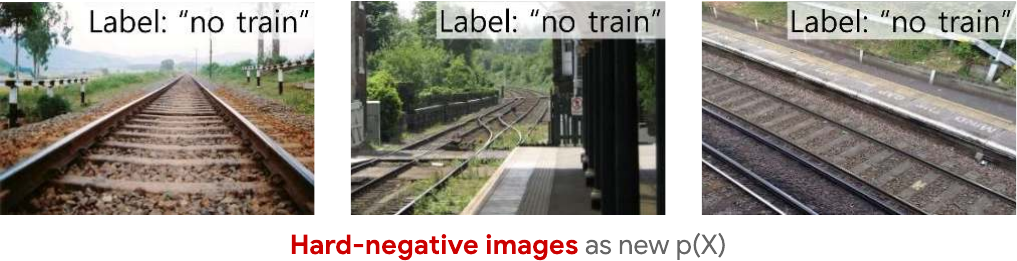}
    \caption{Example hard-negative samples \wrt train samples, containing no trains but still including rails. Base Figures are taken from~\cite{https://doi.org/10.48550/arxiv.2203.03860}.}
    \label{fig:hardneg}
\end{figure}

One way to combat the problem of spurious correlation between rail and train is to introduce interventional data. If we are more cautious when collecting data, we can also collect hard-negative images (rail with no train). This is illustrated in Figure~\ref{fig:hardneg}. Hard-negative images are (1) hard because they target spurious correlations, so the models employing such spurious correlations will get them wrong; and (2) negative because there is no train in the image. Here, we explicitly target the possible bias -- we eliminate ``rail'' from a plausible set of cues for detecting trains.

\begin{figure}
    \centering
    \includegraphics[width=0.8\linewidth]{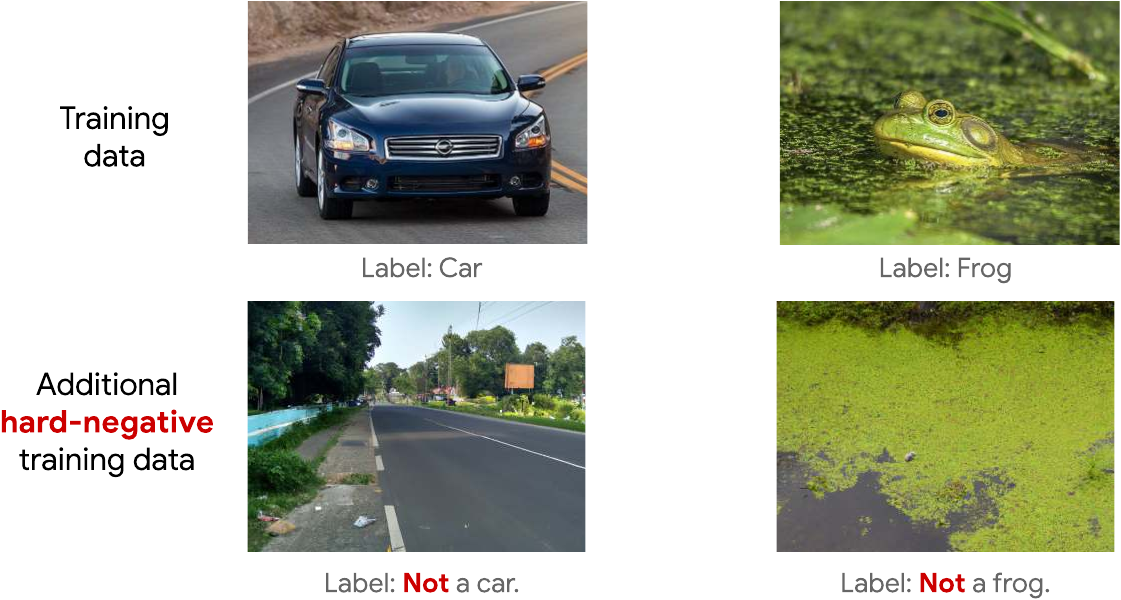}
    \caption{Addition of hard-negative samples \wrt `car' and `frog', containing their corresponding biases but not the objects themselves.}
    \label{fig:morehardneg}
\end{figure}

More examples of interventional data are shown in Figure~\ref{fig:morehardneg}.
Of course, as discussed, this kind of data does not arise very often naturally; thus, there should be a way to go and find them. We need a data crawling mechanism that supplies such examples.

\subsubsection{Efficiently Collecting an Interventional Dataset}

\begin{figure}
    \centering
    \includegraphics[width=0.7\linewidth]{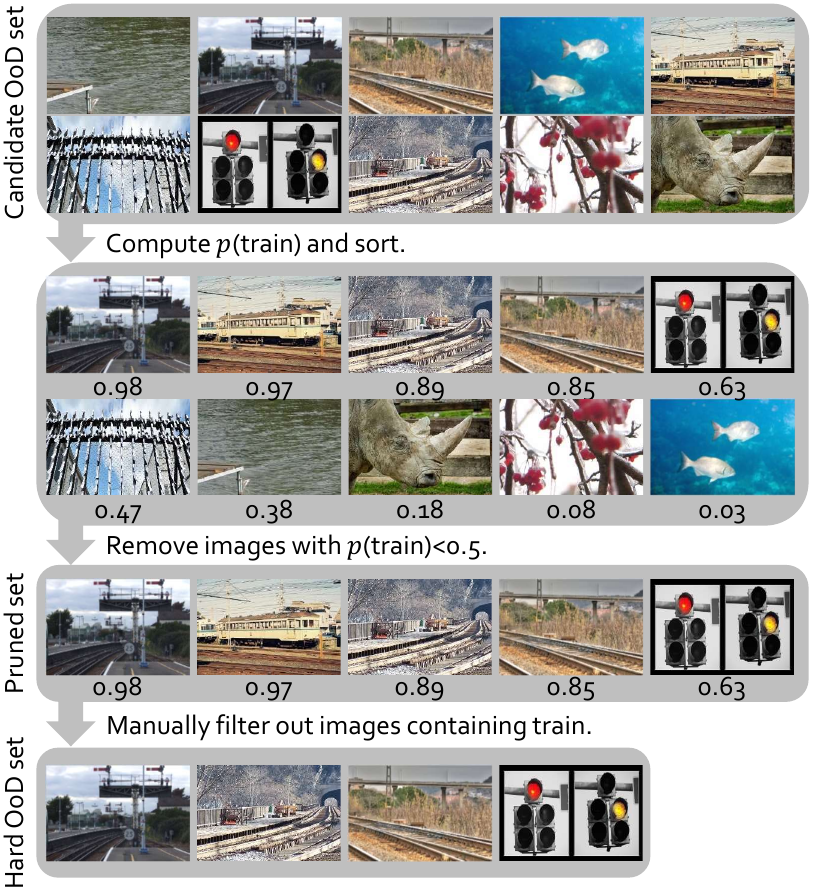}
    \caption{Possible procedure of collecting hard OOD samples. Figure taken from~\cite{https://doi.org/10.48550/arxiv.2203.03860}.}
    \label{fig:collect}
\end{figure}

Our task is to find hard-negative samples for the `train' class in our running example. We can do this as illustrated in Figure~\ref{fig:collect}. The candidate dataset does not fully have to be OOD; it just has to be a large dataset of images. Of course, the more purely OOD the original dataset is, the more efficiently we can collect relevant data from it. We compute \(p(\text{train})\) by running all images in the dataset through our classifier. We only keep images with a `train' score above a certain threshold. This set will not look as nice as in the figure in practice. There will be a lot of true positives as well. For manual filtering, we need human labor (HITL). Humans are the sources of hard-negative knowledge. There is no solution to finding a clarification of such spurious correlations without requiring human knowledge. The name ``hard OOD dataset'' is equivalent to ``hard false positive dataset'' and also to ``hard negative dataset.''

We need to minimize our costs whenever we use human labor because it is expensive. The cost depends on two dimensions:
\begin{enumerate}
    \item \textbf{How long does it take a human to remove all the true positive images?} This is very cheap, as the annotators do not have to draw a bounding box/segmentation map or classify the image into 1k classes. It is an easy binary decision (Y/N).
    \item \textbf{How many hard-negative images are needed per class?} Not many at all. Considering only one image per class, the mIoU (which is a measure for telling how much spurious correlation we have, higher is better) with foreground increases by 2\%. Using 100 images per class, the mIoU with foreground increases by 3\%. We have poor mIoU without any hard negatives. However, one hard-negative image per class already helps a lot, apparently. If we take more hard negatives per class, we get diminishing returns as the mIoU performance saturates.
\end{enumerate}

Interventional data are, therefore, a cheap source of ``How'' information. For the Pascal VOC dataset with 20 classes, we only need 20 new hard-negative samples to improve mIoU quite a bit. This is a low-hanging fruit for new types of data.

\begin{figure}
    \centering
    \includegraphics[width=\linewidth]{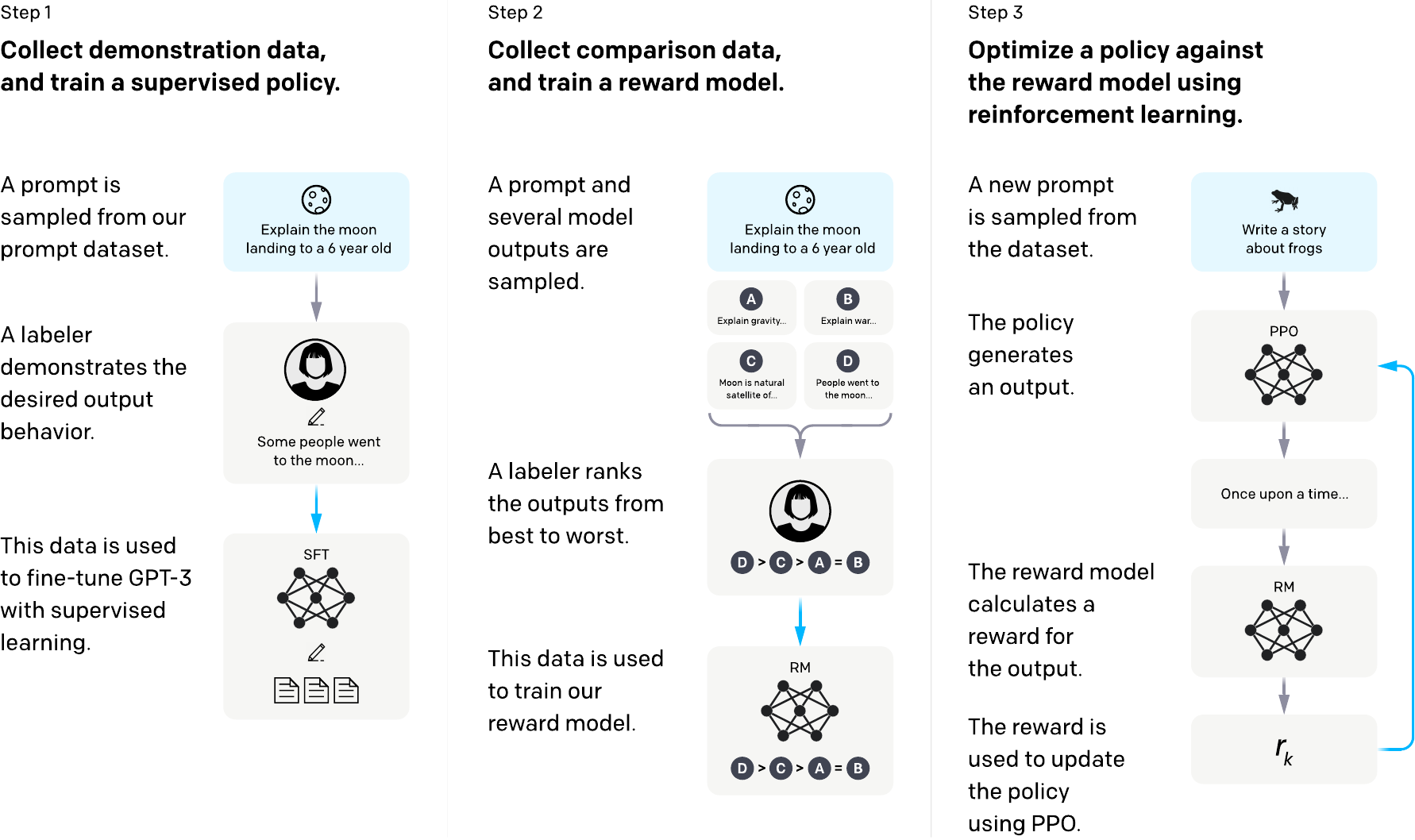}
    \caption{Three-step fine-tuning procedure of ChatGPT. HITL is crucial for aligned ML models. Figure taken from~\cite{chatgpt}.}
    \label{fig:chatgpt}
\end{figure}

\textbf{Interventional data collection is gaining momentum now.} One example is ChatGPT's fine-tuning, illustrated in Figure~\ref{fig:chatgpt}. The research field seems to return to the HITL paradigm. This is good because it is the only way to solve this problem. HITL is used for both InstructGPT and ChatGPT. These improve upon the original GPT-3 in terms of the safety features precisely because they also use HITL to fine-tune the models further. Researchers developing these systems know that humans are the ultimate source of the ``How'' information. We are also shifting the distribution (data or output) a little bit to what humans would consider more appropriate/relevant as answers during a chat. This introduces an intervention in the data generation process; we use a novel data source for further training. And this is what matters: There are all kinds of issues around LLMs, like inappropriate outputs and jailbreak (making LLMs output inappropriate things). Humans can teach LLMs ``how to behave.'' Instead of web crawling, one can use humans to generate samples. This improves trustworthiness considerably. On the left of Figure~\ref{fig:chatgpt}, humans are used to generate possible answers to questions. This is quite labor-heavy. On the right, humans are only used to rank the outputs of models based on their preferences. This ranking can be used for further fine-tuning with RLHF. This is less labor-heavy and is quite scalable.

\subsection{Introducing Additional Supervision}
\href{https://www.youtube.com/watch?v=AAoFT9xjI58}{ImageNet annotation} is performed as follows. First, annotators receive an object category or concept at the top of a webpage. Then they have to click on images containing the concept. Some images from the candidate image set are selected, and some are not. (This is already a pre-filtered set of images that might correspond to the concept.) When we do this, we obtain a set of images for every selected concept. These are then used for training the model.

\begin{figure}
    \centering
    \includegraphics[width=0.9\linewidth]{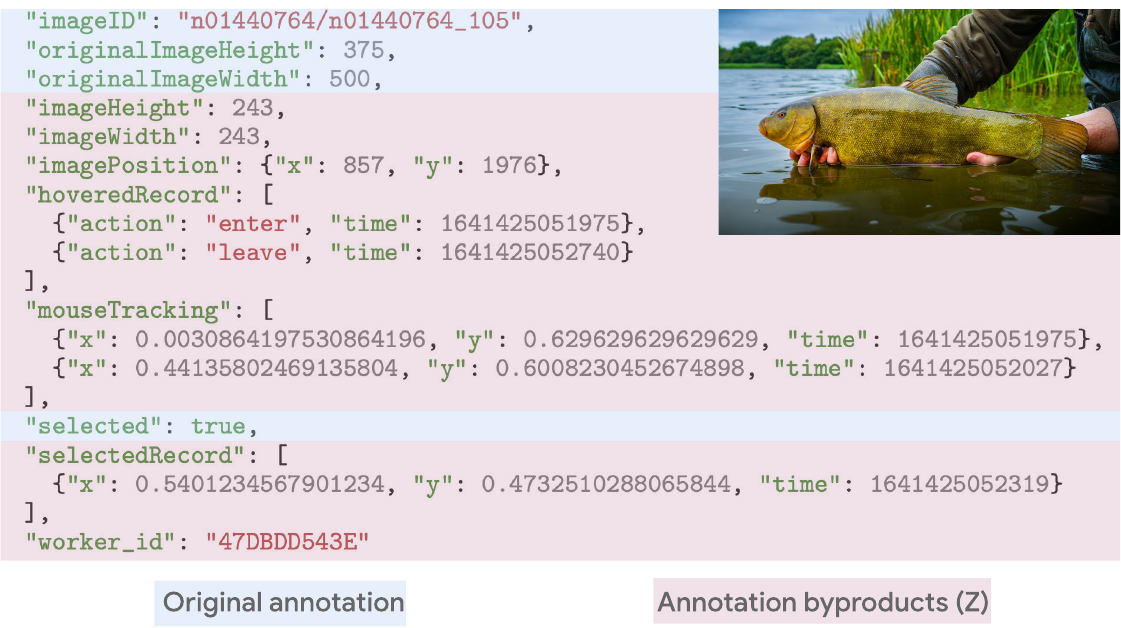}
    \caption{Additional, potentially useful meta-data from the annotation procedure. \emph{Blue:} Original annotation ImageNet data collectors have considered so far. This is wasting a ton of auxiliary supervision. \emph{Red:} The annotation byproducts may be irrelevant and noisy, but we should not throw them away, as they can also be informative. We want to use them to improve our model (e.g., by obtaining new ingredients for uncertainty estimation).}
    \label{fig:additional}
\end{figure}

However, the action of annotation also contains valuable information in terms of the mouse track, click location, time annotators took between clicks, the full time needed to go through the set of images, and many other factors. We can efficiently collect additional supervision. This is shown in Figure~\ref{fig:additional}. Annotation byproducts can be leveraged in several ways. The work of Han~\etal~\cite{han2023neglected} gives a thorough demonstration of how these can be used along with task supervision.

\begin{figure}
    \centering
    \includegraphics[width=0.7\linewidth]{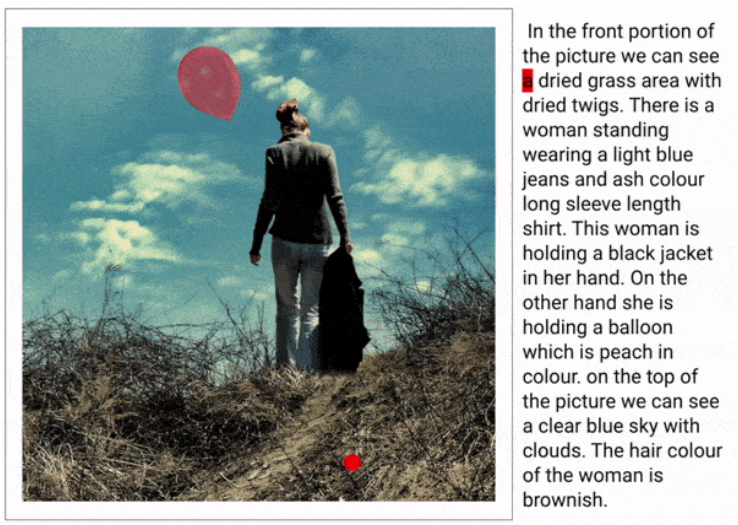}
    \caption{Labeling process of OpenImages Localized Narratives. They try to collect as much information as possible for every image. The human annotator speaks out what they see in the image. As they describe every object, they need to hover over the image part they are talking about. For every word, we have a corresponding location in the image. They record the mouse trajectory and voice (1-to-1 correspondence between what they say and what they point to). Then, the voice recording is transcribed into text. This results in huge captions compared to COCO. Figure taken from~\cite{openimages}.}
    \label{fig:localization}
\end{figure}

There are also a lot of parallel efforts from other groups to obtain additional supervision. One is OpenImages Localized Narratives, illustrated in Figure~\ref{fig:localization}. This is way more information and supervision compared to traditional image captioning or object localization datasets. Multimodal annotations are rich in the ``How'' information content in general. The annotation contains much new information. We should consider how to best exploit this information for the ``How'' problems. There is not much research on this yet; we are fortunate to work on this now and make an impact.

\subsection{Method-centric vs. Dataset-centric Solutions}

There are two general ways to solve problems, both with pros and cons, detailed below.

\textbf{Method-centric solutions.} These have cheap initial costs. One can use existing benchmarks and training sets. One just needs to devise a clever new method (e.g., loss, architecture, optimizer, regularizer). Typically we end up with highly complex methods because all simple methods have been tried out already. For these complicated methods, we need a lot of computational resources and human brain time. This potentially has enormous costs. We usually do not consider brain time cost as much compared to, e.g., annotation cost. Development is expensive and requires many runs to validate hyperparameters. Furthermore, such solutions are upper-limited by the information cap defined by the benchmark. (What supervision do we have?) As such, scaling up often fails. The complex tricks do not work anymore. We need a lot of effort and experiments to prune down the method into something simpler that scales well.

\textbf{Dataset-centric solutions.} These are relatively new and exciting approaches. It has large initial costs: 10k - 10M EUR for a large-scale dataset. A few thousand might be enough for a small-scale dataset, but it is meaningful when our budget goes up to 100k EUR. (This way, we can obtain a larger dataset and/or better supervision.) Once built, it brings huge utility to the public. (Everyone can use it to create new methods; it has a huge impact.) One could also expect good transferability of pre-trained models to other tasks. (We can pre-train a model on the dataset and open source it: this is also a huge contribution to the field. They can just download it without needing to train it from scratch.) Notably, there is no information cap (only creativity cap and budget cap). If we have more information available, the method itself can be quite simple. We can just use a vanilla loss/architecture, which will often work best (we have seen that simple methods often work best). Such methods also scale better and are easier to use, as they come with fewer hyperparameters usually.

We think that simple methods with new kinds of data will bring us the biggest gain in the future.

\href{https://chat.openai.com/chat}{ChatGPT} proposed the following closing statement for the book: ``Let us harness the power of machine learning to make a difference. Let us make an impact through machine learning.'' We could not agree more.

\appendix

\chapter{Calculus Refresher}
We consider a couple of exercises for calculating partial derivates \wrt vectors and matrices. One particularly useful object for calculating partial derivatives is the Kronecker delta, a function of two variables.

\begin{definition}{Kronecker Delta}
\[\delta_{ij} = \begin{cases} 0 & \text{if } i \ne j \\ 1 & \text{if } i = j. \end{cases}\]
\end{definition}

Matrix multiplication, which is also important to be able to solve the exercises later, is defined as follows.

\begin{definition}{Matrix Multiplication}
Let \(A \in \nR^{m \times n}, B \in \nR^{n \times p}, C \in \nR^{m \times p}\).
\[C = AB \iff C_{ij} = \sum_{k = 1}^n A_{ik} B_{kj}\ \forall i \in \{1, \dots, m\}, j \in \{1, \dots, p\}.\]
\end{definition}

Let us consider the gradient operator.

\begin{definition}{Gradient}
Let \(f: \nR^n \rightarrow \nR\). Then
\[\nabla f: \nR^n \rightarrow \nR^n, \left(\nabla f\right)_i = \frac{\partial f}{\partial x_i}.\]

Sometimes, the argument is explicit: \(\nabla_x f\). In \(\nabla_x f(x)\), the subscript indicates ``which variable'' and the argument indicates ``where to evaluate''. 

\medskip

\textbf{Example}: \(f: \nR^n \times \nR^m \times \nR^l \rightarrow \nR\). Then \(\nabla_z f: \nR^n \times \nR^m \times \nR^l \rightarrow \nR^l\). We often abuse notation and use the variable name to indicate the position of the argument \wrt which we take the gradient. Often this is clear from the context.

\medskip

The following notations are questionable. They are both abusing the abuse of notation.
\begin{itemize}
    \item \(\nabla_z f(x + z, z^2, y)\). This notation is unclear. According to general use, the \(z\) in the subscript should refer to the position. However, we also have an explicit variable \(z\) that can be confusing. One should either use different symbols as the arguments and/or one should write everything down nicely using partial derivatives. Combining the two, one might first declare that \(f\) is a function of variables (placeholders) \(x'\), \(y'\), and \(z'\), and then write
    \[\restr{\frac{\partial f}{\partial x}}{x' = x + z, y' = z^2, z' = y}.\]
    
    \item \(\nabla_{x + y + z} f(x + y + z, x)\). This notation is incorrect. One should use the subscript to refer to the position, and again, either use different symbols as the arguments or write everything down using partial derivatives.
\end{itemize}
\end{definition}

Lastly, we present a simple rule for taking partial derivatives of a tensor element \wrt another tensor element.

\begin{definition}{Derivative of a Tensor Element \wrt Another Element}
\[\frac{\partial v_{i_1,\dots,i_n}}{\partial v_{j_1,\dots,j_n}} = \prod_{k = 1}^n\delta_{i_kj_k}.\]
\end{definition}

We are now ready to solve the first exercise.

\begin{task}{Gradient of Squared $L_2$ Norm}
Show that for $x \in \nR^n$,
\[\nabla_x \Vert x \Vert^2 = 2x.\]
\end{task}

$\forall i \in \{1, \dots, n\}$:
\begin{align*}
\left(\nabla_x \Vert x \Vert^2\right)_i &= \frac{\partial}{\partial x_i} \sum_{j = 1}^n x_j^2\\
&= \sum_{j = 1}^n \frac{\partial}{\partial x_i} x_j^2\\
&= \sum_{j = 1}^n \delta_{ij} 2x_j\\
&= 2x_i.
\end{align*}

Let us define the trace operator for real matrices.

\begin{definition}{Trace}
The trace of a square matrix $A \in \nR^{n \times n}, n \in \nN$ is defined as
\[\operatorname{tr}(A) = \sum_{i = 1}^n a_{ii}.\]
\end{definition}

The second exercise is as follows.

\begin{task}{Gradient of Trace of Matrix Multiplication}
Show that for $A \in \nR^{n \times m}, B \in \nR^{m \times n}$,
\[\nabla_A \operatorname{tr}(AB) = B^\top.\]
\end{task}
$\forall i, j \in \{1, \dots, n\}$:
\begin{align*}
\left(\nabla_A \text{tr}(AB)\right)_{ij} &= \frac{\partial}{\partial A_{ij}} \sum_{p = 1}^n (AB)_{pp}\\
&= \frac{\partial}{\partial A_{ij}} \sum_{p = 1}^n \sum_{q = 1}^m A_pq B_qp\\
&= \sum_{p = 1}^n \sum_{q = 1}^m \frac{\partial}{\partial A_{ij}} A_pq B_qp\\
&= \sum_{p = 1}^n \sum_{q = 1}^m \delta_{ip} \delta_{jq} B_qp\\
&= B_{ji}.
\end{align*}

Our last exercise is to compute the gradient of a quadratic form.

\begin{task}{Gradient of Quadratic Form}
Show that for $x \in \nR^n, A \in \nR^{n \times n}$,
\[\nabla_x x^\top A x = (A + A^\top)x.\]
\end{task}
$\forall i \in \{1, \dots, n\}$:
\begin{align*}
\left(\nabla_x x^\top A x\right)_i &= \frac{\partial}{\partial x_i} \sum_{p, q = 1}^n x_p A_{pq} x_q\\
&= \sum_{p, q = 1}^n \frac{\partial}{\partial x_i} \left(x_p A_{pq} x_q\right)\\
&= \sum_{p, q = 1}^n \delta_{ip} A_{pq} x_q + \sum_{p, q = 1}^n \delta_{iq} x_p A_{pq}\\
&= \sum_{q = 1}^n A_{iq} x_q + \sum_{p = 1}^n x_i A_{iq}\\
&= (Ax)_i + (A^\top x)_i\\
&= ((A + A^\top)x)_i.
\end{align*}


\clearpage

\tcblistof[\chapter*]{definition}{List of Definitions}
\clearpage

\chapter*{Bibliography}
\printbibliography[heading=none]

\end{document}